\documentclass[twoside,11pt]{article}

\usepackage{amsmath,amsthm,mathtools,bm,mathrsfs}
\usepackage{bbm}
\PassOptionsToPackage{numbers,sort&compress}{natbib}
\let\jmlrsavedbibliographystyle\bibliographystyle
\renewcommand{\bibliographystyle}[1]{}
\usepackage[preprint]{jmlr2e}
\let\bibliographystyle\jmlrsavedbibliographystyle
\bibpunct{[}{]}{,}{n}{,}{,}
\usepackage{indentfirst}
\usepackage[toc]{appendix}
\usepackage{array}
\usepackage{multirow}
\usepackage{tabularx}
\usepackage{pdflscape}
\usepackage{afterpage}
\usepackage[figuresright]{rotating}
\usepackage{enumitem}
\usepackage{anyfontsize}
\usepackage[outline]{contour}
\usepackage[ruled]{algorithm2e}%
\usepackage{algorithmic}
\SetAlgoCaptionSeparator{.}

\usepackage{etoolbox}

\newlength{\algoinset}
\newcommand{\SetAlgoInset}[1]{%
    \setlength{\algoinset}{#1}%
    \setlength{\algomargin}{\algoinset}%
    \SetAlCapHSkip{\algoinset}%
}
\SetAlgoInset{6pt}

\makeatletter
\BeforeBeginEnvironment{algorithmic}{%
    \setlength{\linewidth}{\hsize}\addtolength{\linewidth}{1.5em}%
    \addtolength{\linewidth}{-\algoinset}%
    \@rightskip\z@skip \rightskip\z@skip
}
\makeatother
\AtBeginEnvironment{algorithmic}{%
    \setlength{\abovedisplayskip}{4pt}\setlength{\belowdisplayskip}{4pt}%
    \setlength{\abovedisplayshortskip}{4pt}\setlength{\belowdisplayshortskip}{4pt}%
}

\makeatletter
\let\NM@algocf@makecaption@ruled\algocf@makecaption@ruled
\renewcommand{\algocf@makecaption@ruled}[2]{%
    \begingroup
    \hsize\algocf@ruledwidth \advance\hsize-\algoinset%
    \NM@algocf@makecaption@ruled{#1}{#2}%
    \endgroup
}
\makeatother

\hypersetup{colorlinks=true,linkcolor=blue,citecolor=blue,urlcolor=blue}

\newcommand{\NMlandscapepage}[1]{%
    \ifnum\value{page}=\getpagerefnumber{#1}\relax
        \pdfpageattr{/Rotate 90}%
    \fi
}
\AddToHook{shipout/before}{%
    \NMlandscapepage{tab:2}%
    \NMlandscapepage{tab:1}%
}

\usepackage{thm-restate}
\makeatletter
\newcommand{\restatethmlike}[2]{%
    \begingroup
    \renewcommand{\thetheorem}{\csname the#1\endcsname}%
    \let\c@theorem\c@thmt@dummyctr
    \let\theHtheorem\theHthmt@dummyctr
    #2*%
    \endgroup
}
\makeatother
\newcommand{\restateproposition}[1]{\restatethmlike{proposition}{#1}}

\newcommand{\E}{\mathbb{E}}
\newcommand{\R}{\mathbb{R}}

\renewcommand{\S}{\mathbb{S}}

\newcommand{\calL}{\mathcal{L}}

\newcommand{\BY}{\boldsymbol{Y}}
\newcommand{\BZ}{\boldsymbol{Z}}

\newcommand{\argmin}{\operatorname*{arg\,min}}
\newcommand\norm[2][1]{\left\|#2\right\|_{#1}}

\newcommand\abs[1]{\left|#1\right|}
\newcommand\where[2]{\left.#1\right|_{#2}}
\newcommand{\diffeomorphismto}{\xrightarrow{\sim}}
\newcommand{\Setminus}[2]{{\left.#1\middle\backslash #2\right.}}
\newcommand{\Setmod}[2]{{\left.#1\middle/ #2\right.}}
\newcommand\RomanNum[1]{\uppercase\expandafter{\romannumeral #1}}

\newcommand\trace{\mathrm{tr}}
\newcommand{\target}{\mathcal{Y}}

\newcommand{\dd}{\mathrm{d}}
\newcommand{\DD}{\mathrm{D}}

\newcommand{\Normal}{\mathcal{N}}
\newcommand{\U}{\operatorname{U}}
\newcommand{\KL}[2]{\operatorname{KL}\left(#1\middle\|#2\right)}
\newcommand{\TV}[2]{\operatorname{TV}\left(#1,#2\right)}
\newcommand{\Cov}{\operatorname{Cov}}
\newcommand{\Var}{\operatorname{Var}}

\newcommand{\sg}{\operatorname{sg}}
\newcommand{\iidsim}{\overset{\mathrm{i.i.d.}}{\sim}}

\newcommand{\DensityTangentFieldFirst}{P}
\newcommand{\DensityTangentFieldSecond}{Q}
\newcommand{\pdfspace}{\mathscr{P}}
\newcommand{\velspace}{\mathscr{V}}

\newcommand{\velmancan}{\velspace_{\mathrm{can}}}

\newcommand{\energyspace}{\mathscr{E}}
\newcommand{\terminalmap}{\mathcal{T}}

\newcommand{\canonicalmap}{\mathcal{C}}
\newcommand{\projection}{\mathcal{R}}

\newcommand{\flow}{\Phi}

\newcommand{\retraction}{\mathrm{Retr}}

\newcommand{\iteration}{\mathcal{I}}
\newcommand{\Stein}{\mathcal{S}}
\newcommand{\Diss}{\mathcal{D}}
\newcommand{\grad}{\operatorname{grad}}
\newcommand{\gradFR}{\grad_{\mathrm{FR}}}

\newcommand{\const}{\mathrm{const}}
\newcommand{\base}{\mathrm{base}}

\newcommand{\fisher}{\mathrm{FR}}
\newcommand{\Id}{\mathrm{Id}}
\newcommand{\weakto}{\xrightarrow{\mathrm{w}}}

\usepackage{calc}
\newlength{\eqbylen}
\newcommand{\eqby}[1]{%
  \mathrel{\mathop{%
    \settowidth{\eqbylen}{$\scriptstyle #1$}%
    \addtolength{\eqbylen}{0.0em}%
    \vcenter{\offinterlineskip
      \hbox to \eqbylen{\leaders\hrule height .45pt\hfill}
      \kern.35ex
      \hbox to \eqbylen{\leaders\hrule height .45pt\hfill}
    }%
  }\limits^{#1}}%
}
\newcommand{\aeq}{\eqby{\mathrm{a.e.}}}

\usepackage{tikz}
\usetikzlibrary{positioning,calc,arrows.meta,decorations.markings}
\usepackage{tikz-cd}

\usepackage{subcaption}

\makeatletter
\renewcommand{\p@subfigure}{\thefigure.}
\makeatother
\DeclareCaptionLabelFormat{subletter}{(\alph{subfigure})}
\newsavebox{\bafbox}
\newlength{\bafexcess}
\newenvironment{bottomalignedfloat}
  {\begin{lrbox}{\bafbox}\begin{minipage}{\textwidth}}
  {\end{minipage}\end{lrbox}%
   \setlength{\bafexcess}{\dimexpr\ht\bafbox+\dp\bafbox-\textheight\relax}%
   \ifdim\bafexcess>0pt\relax\vspace*{-\bafexcess}\fi
   \usebox{\bafbox}}

\definecolor{velouter}{RGB}{222,233,247}
\definecolor{velinner}{RGB}{239,245,252}
\definecolor{pdfouter}{RGB}{247,228,228}
\definecolor{pdfinner}{RGB}{252,240,240}
\definecolor{parfill}{RGB}{239,233,247}
\definecolor{flowouter}{RGB}{239,233,247}%
\definecolor{flowinner}{RGB}{247,244,251}
\definecolor{canmangreen}{RGB}{222,255,222}

\theoremstyle{definition}
\newtheorem{definition}[theorem]{Definition}
\newtheorem{remarkx}[theorem]{Remark}
\newenvironment{remark}
  {\pushQED{\qed}\remarkx}
  {\popQED\endremarkx}
\newtheorem{assumption}[theorem]{Assumption}
\theoremstyle{plain}

\usepackage{lastpage}

\firstpageno{1}

\begin{document}
\pagenumbering{roman}

\title{Newton Matching for Generative Modeling:\\A Unified Framework for Fine-Tuning and Sampling}

\author{\name Zeyang Li \email zeyang@mit.edu \\
    \name Yunan Wang \email wangyn25@mit.edu\\
    \name Paolo Giaretta \email pgiarett@mit.edu\\
    \name Navid Azizan \email azizan@mit.edu\\
        \addr Massachusetts Institute of Technology%
       }

\editor{
}

\maketitle

\newcommand{\correspondencetext}{Correspondence to Zeyang Li and Yunan Wang.}
\newlength{\correspondencefnwidth}
\setlength{\correspondencefnwidth}{2.25in}
\renewcommand{\footnoterule}{\kern-3pt\hrule width \correspondencefnwidth \kern 2.6pt}
{\makeatletter\long\def\@makefntext#1{\noindent #1}\makeatother
\let\thefootnote\relax\footnotetext{\correspondencetext}}

\begin{abstract}%
Diffusion and flow models have achieved remarkable success, enabled in large part by conditional matching, which reduces distribution learning to a scalable supervised regression problem when samples from the target distribution are available. However, many important tasks, such as fine-tuning and sampling, fall into the opposite regime, in which the target probability density is specified only up to a normalizing constant and direct samples are unavailable. Specifically, we consider target densities of the form \(\pi(x) \propto \mu(x)e^{\tau r(x)}\), where $\mu$ denotes the reference factor, \(r\) denotes the reward function, and \(\tau\) controls the emphasis on the reward. \(\mu = \rho^{\mathrm{base}}\) denotes the normalized terminal density from the pretrained model in the fine-tuning task, whereas \(\mu \equiv 1\) in the sampling task.

Early approaches cast these tasks as reinforcement learning over the generative dynamics, lifting a terminal-density problem to trajectory-level optimization while leaving the conditional-matching structure largely unexploited. More recent work instead seeks to preserve the supervised-regression form. Reweighting-based methods retain the standard matching objective but can become unstable when the proposal poorly covers the target density. Another emerging class draws samples from the current model and uses reward-based feedback to construct the next regression update. These methods, however, arise from disparate principles and intertwine model representation, sampling construction, and feedback signal. Consequently, their regression objectives rarely reveal the underlying population-level update, obscuring which differences are fundamental, which reflect alternative exact realizations, and which arise from approximations that alter the update. More fundamentally, how a single stage transforms the terminal density, whether it improves the objective, and whether the resulting iteration converges remain largely uncharacterized.

In this paper, we develop \emph{Newton Matching}, a unified framework for fine-tuning and sampling. Newton Matching rests on two paradigm shifts in perspective. First, rather than treating a method as an isolated regression loss, we view learning as an iterative optimization process: the current model generates samples, each stage updates its terminal density based on these samples, and the resulting density determines the next model iterate. Second, we exploit the conditional-expectation structure of standard matching to define the optimization domain. Although a terminal density can be realized by infinitely many generative dynamics, standard conditional matching selects a distinguished representation---the population minimizer that would be recovered if samples from that density were available---which we call its canonical model. Canonical models retain the regression structure that makes diffusion and flow training scalable. Under the compatible smooth-realization assumptions, their velocity representations form a manifold in one-to-one correspondence with terminal densities. On this shared domain, fine-tuning and sampling become iterative optimization problems, and our framework disentangles three questions that existing methods often conflate: which distributional direction should govern the iteration, how a finite-stepsize update along that direction transforms the terminal density, and how that transformation can be realized through matching-style regression. To specify a principled direction, Newton Matching chooses the reverse Kullback--Leibler (KL) objective, equips the terminal-density manifold with the Fisher--Rao metric and the mixture connection, and transports both geometric structures to the canonical manifold. Remarkably, the reverse-KL Hessian under the mixture connection coincides with the Fisher--Rao metric, so the Newton direction is exactly the negative Fisher--Rao gradient. This direction is generated by the regularized reward, which combines the task reward with the density-ratio correction required by the prescribed target, and admits equivalent covariance and gradient representations, both learnable through sample-wise matching without importance sampling or backpropagation through the full generation trajectory. A Newton Matching stage first takes a damped or full tangential step and then canonicalizes this intermediate outcome, replacing it with the canonical model associated with the same terminal density. This composition defines a canonical retraction and yields an exact finite-stepsize characterization of the induced density update, from which we establish strict reverse-KL descent throughout the full admissible stepsize range, global convergence under mild conditions, and local quadratic convergence of the full-step iteration. We then translate the ideal tangential update into scalable algorithms. Covariance and gradient representations of the Newton direction can each be combined with forward or reverse regression-pair constructions, yielding multiple sample-wise regression objectives with the same population minimizer. To broaden the computational design space, we further develop approximate Newton Matching, which allows the tangential update to be approximated in various principled ways. We introduce critical-point consistency to distinguish approximations whose tangential updates vanish exactly when the current terminal density equals the prescribed target. Finally, this modular decomposition recovers representative existing methods as exact realizations, critical-point-consistent approximations, or variants that alter the underlying objective, thereby demystifying methods derived from seemingly disparate principles and furnishing a principled design space for new algorithms.
\end{abstract}


\setcounter{tocdepth}{2}
{\hypersetup{linkcolor=black}\tableofcontents}
\clearpage
\pagenumbering{arabic}

\section{Introduction}

Diffusion \citep{sohl2015deep, ho2020denoising} and flow models \citep{lipman2023flow} have achieved remarkable success in image and video generation \citep{rombach2022highresolution,ho2022video,esser2024scaling}, and their applications are rapidly expanding to broader domains \citep{chi2025diffusion,watson2023denovo,zeni2025generative}. At their core, these models learn a time-dependent generative process that gradually transports a simple prior distribution, typically Gaussian noise, into a complex target distribution. Their success, however, stems not only from the expressive power of these generative dynamics, but also from the remarkably simple and scalable way in which they can be learned. When samples from the target distribution are available, both model families share a common conditional-matching structure: intermediate states are constructed analytically along a prescribed noising process, and the model is trained by supervised regression against tractable sample-wise targets. At the population level, the optimal regressor is precisely the conditional expectation that drives the desired probability transport. This conditional-expectation structure reduces the difficult problem of learning a complex distribution to a scalable supervised regression problem.

Many important tasks, however, violate precisely the assumption that makes this training recipe possible: samples from the desired target density are unavailable. In reward-based fine-tuning, one starts from a pretrained generative model and seeks to shift probability mass toward high-reward outputs. In sampling from an unnormalized density, the target is specified by an energy function. Although usually studied separately, both problems can be expressed through the shared target
\begin{equation*}
    \pi(x)\propto\mu(x)e^{\tau r(x)},
\end{equation*}
where \(\mu\) denotes the reference factor, \(r\) is the reward function, and the inverse temperature \(\tau>0\) controls the strength of the tilt. For fine-tuning, \(\mu=\rho^\base\) is the terminal density induced by the pretrained model; for sampling, \(\mu\equiv1\).
These two formulations encompass a broad range of applications. For the fine-tuning formulation, in which the terminal density of a pretrained model supplies the reference factor, representative examples are reward alignment of text-to-image models \citep{fan2023dpok,black2024training} and offline reinforcement learning \citep{wang2020critic,wang2023diffusion}. For the sampling formulation, in which the target is specified directly as a Boltzmann distribution, representative examples are equilibrium configuration sampling \citep{noe2019boltzmann} and maximum-entropy online reinforcement learning \citep{haarnoja2018soft,li2026reverse}. In this paper, we focus on theory and algorithms at the level of these two formulations rather than on any particular application; accordingly, our results apply whenever the desired target can be expressed in either form.

In both cases, learning takes place in a \emph{dataless} regime: the target density $\pi(x)\propto\mu(x)e^{\tau r(x)}$ is specified through its defining factors, rather than through samples drawn from it. Therefore, the supervised construction that underpins the efficacy of diffusion and flow training can no longer be used directly: there are no target endpoints from which to form training pairs, and the generative dynamics must instead be learned from feedback evaluated on samples produced by the current or base model. Turning this indirect signal into a stable and scalable learning procedure that recovers the prescribed target density is highly nontrivial.

Across the literatures on fine-tuning and sampling, a common early strategy is to recast the dataless learning problem as reinforcement learning (RL). At a high level, these approaches can be divided into discrete-time and continuous-time formulations. In discrete time, the discretized generative trajectory is modeled as a Markov decision process (MDP): intermediate noisy samples serve as states, denoising transitions define the policy, and the terminal reward guides policy optimization \citep{fan2023dpok,black2024training,liu2025flowgrpo,xue2025dancegrpo,hwang2026value}. In continuous time, the analogous formulation is stochastic optimal control, where the drift or score is treated as the control and optimized against a terminal reward or energy, typically together with a pathwise control cost or a divergence from a reference process \citep{tzen2019theoretical,zhang2022path,vargas2023denoising,berner2024optimal,richter2024improved,uehara2024finetuning,zhao2025score,han2025stochastic}. These formulations offer a straightforward way to propagate terminal feedback through the generative dynamics, but they treat diffusion and flow models as generic controlled processes. The problem of learning the prescribed terminal density is thereby lifted to a trajectory-level optimization problem, through transition-wise objectives in discrete time or path-space objectives in continuous time. Since many dynamics can realize the same terminal law, this lift requires path-level choices beyond the target specification, which can bias the endpoint distribution or tie the method to particular noise schedules and samplers. Meanwhile, the specialized structure underlying scalable diffusion and flow training remains largely unused.

More recently, a growing line of work has sought to preserve the native regression structure of diffusion and flow training in the dataless regime. The most straightforward strategy retains the usual matching objective and incorporates the task signal through sample reweighting. For fine-tuning, this amounts to weighting matching losses by rewards or advantages evaluated on samples from a reference distribution \citep{zhang2025energy,fan2025online,xue2026advantage}; for sampling, it corresponds to posterior-mean estimation of the target score or velocity via self-normalized importance sampling \citep{akhoundsadegh2024iterated,li2026reverse}. In both settings, insufficient coverage of the desired target can cause the weights to concentrate on a small fraction of samples, thereby reducing the effective sample size and leading to high-variance, unstable updates.
To avoid the pitfalls of importance sampling, a more promising direction is to derive model-dependent regression targets from samples generated by the current model. This direction has emerged from various perspectives, including stochastic optimal control \citep{domingo2025adjoint,havens2025adjoint,liu2025adjoint,bergmeister2026reinforce}, contrastive learning \citep{zheng2026diffusionnft}, fixed-point iteration \citep{havens2026flow,blessing2026bridge}, and velocity evolution under exponential tilting \citep{potaptchik2026tilt}. Collectively, these works demonstrate the feasibility of retaining the simple and scalable regression form of standard generative model training, even in the absence of target samples.

Despite this progress, the emerging landscape remains conceptually fragmented. Even the basic design choices are tightly intertwined: the model may be represented by a score, velocity, or drift; constructing a regression pair involves choosing whether the endpoint is obtained from an ODE or SDE sampler and whether the corresponding intermediate state is taken from the same trajectory or sampled from a prescribed noising kernel; and the regression target may depend on the reward itself, its gradient, or an adjoint quantity. These choices arise from distinct, often ad hoc and technically intricate derivations, and the distributional meaning of the resulting regression objectives is rarely transparent from their final form. Consequently, the objectives alone do not reveal which differences are fundamental, which are merely alternative realizations of the same population-level update, and which arise from approximations that alter its distributional meaning.
More fundamentally, these methods are often presented as generic regression losses, even though in practice they define an iterative model-update procedure: the current model generates samples, regression produces an updated model, and the cycle repeats. The iterative behavior of these methods remains largely uncharacterized, even at the level of a single stage: it remains unclear how the terminal density is updated, whether the procedure moves the model density toward \(\pi(x)\propto\mu(x)e^{\tau r(x)}\), or whether it yields consistent improvement, let alone whether the full iterative procedure converges.

Motivated by these gaps, we develop a unified framework for dataless fine-tuning and sampling that reveals the shared structure underlying scalable matching methods that do not rely on importance sampling. Our starting point is a simple but powerful observation: the conditional-expectation structure is not merely a convenient way to train diffusion and flow models; it selects a distinguished model representation for every terminal density. Indeed, the same terminal density can be generated by infinitely many dynamics. Searching over all such representations introduces unnecessary redundancy and does not by itself ensure that intermediate models retain the structure needed for scalable regression. We therefore restrict attention to the specific models, whether expressed as velocities, scores, or drifts, that standard diffusion or flow training would recover at the population level if samples from their terminal densities were available. We call them \emph{canonical models} and show that they form a structured geometric family, the \emph{canonical manifold}, embedded in the generic model space and in one-to-one correspondence with terminal densities. Fine-tuning and sampling can then be viewed as iterative optimization over this manifold. At each stage, the current model is updated toward the target density using a direction that admits a conditional-expectation representation and can therefore be learned through sample-wise regression in the same form as conditional flow matching. The updated terminal density is then expressed again through its canonical model, providing the appropriate starting point for the next stage. With this fundamental shift in perspective, the central question is no longer how to design a particular regression loss, but which principled optimization direction should govern the iteration, how a finite-stepsize update along that direction changes the terminal density, and how the resulting update can be realized through scalable matching.

\begin{figure}[!p]
    \centering
    \begin{bottomalignedfloat}
    \captionsetup[subfigure]{labelsep=space,skip=-9pt}
    \begin{subfigure}{\linewidth}
        \centering
        \captionsetup{skip=-12.8pt}
        \input{figures/overview_a.tikz}
        \caption{Geometric illustration of Newton Matching. $\terminalmap:\velspace\to\pdfspace$ is the terminal-density map, $\canonicalmap:\pdfspace\diffeomorphismto\velmancan$ is the canonical-velocity map, and $\projection:\velspace\to\velmancan$ is the canonical projection. The target density $\pi\in\pdfspace$ is defined as $\pi(x)\propto \mu(x)e^{\tau r(x)}$, where the reference $\mu\equiv1$ and $\mu\in\pdfspace$ correspond to sampling and fine-tuning, respectively. $r$ is the raw reward and $\tilde{r}^\rho=r-\frac{1}{\tau}\log\frac{\rho}{\mu}$ is the regularized reward. $\Gamma^{\rho,\tilde{r}^\rho}$ and $\xi^{\rho,\tilde{r}^\rho}$ define the Newton directions of reverse KL at $v^\rho\in\velmancan$ and $\rho\in\pdfspace$, respectively.}
        \label{fig:1}
    \end{subfigure}
    \par\vspace{7pt}

    \begin{subfigure}{\linewidth}
        \centering
        \input{figures/overview_b.tikz}
        \caption{Tangential update. At stage $k$, given $v^{\rho_k}$, obtain $\bar{v}^{k+1}=v^{\rho_k}+\eta_k\Gamma^{\rho_k,\tilde{r}^{\rho_k}}$.}
        \label{fig:2}
    \end{subfigure}
    \par\vspace{7pt}

    \begin{subfigure}{\linewidth}
        \centering
        \input{figures/overview_c.tikz}
        \caption{Canonicalization. At stage $k$, given $\bar{v}^{k+1}$, obtain ${v}^{\rho_{k+1}}=\projection(\bar{v}^{k+1})$.}
        \label{fig:3}
    \end{subfigure}
    \caption{Overview of Newton Matching, which iteratively solves the optimization problem $\min_{v\in\velmancan}\,\KL{\terminalmap(v)}{\pi}$ by applying Newton's method on the canonical manifold.}
    \label{fig:4}
    \end{bottomalignedfloat}
\end{figure}

We answer these questions with \emph{Newton Matching}, whose overall architecture is illustrated in Figure~\ref{fig:4}. Newton Matching casts fine-tuning and sampling as iterative optimization on the canonical manifold, with the generative model itself as the optimization variable. Although Figure~\ref{fig:4} and our main development use the velocity representation, the same canonical model admits equivalent representations in terms of velocity, score, drift, data prediction, and noise prediction. Regardless of the chosen representation, restricting the search to the canonical manifold keeps the iterates within the class of models that retain the conditional-expectation structure of standard diffusion and flow training. The optimization objective is to minimize the reverse Kullback--Leibler (KL) divergence between the terminal density generated by the current model and the prescribed target density. Unlike optimization in Euclidean space, where the standard geometric choices are usually left implicit, optimization on a manifold requires its geometry to be specified. Given an objective, a first-order method requires a Riemannian metric to define the gradient, while a second-order method requires an affine connection to define the Hessian. We equip the density manifold with the Fisher--Rao metric and the mixture connection; then, we transport both structures to the canonical manifold through a diffeomorphism. For the reverse-KL objective, we prove a remarkable identity: its Hessian under the mixture connection equals the Fisher--Rao metric as a bilinear form on the tangent space.
Consequently, the same intrinsic optimization direction is both the Newton direction and the negative Fisher--Rao gradient.
Crucially, this direction is computationally accessible rather than merely a geometric abstraction. It is generated by the current \emph{regularized reward}, which combines the task reward with the density-ratio correction required by the prescribed target. The direction admits two equivalent conditional-expectation representations: a covariance form based on regularized-reward values and a gradient form based on their gradients. Either representation can be realized through matching-style sample-wise regression, without importance sampling or backpropagation through the entire generation trajectory.

Each Newton Matching stage consists of the two operations depicted in Figure~\ref{fig:4}. First, the \emph{tangential update} takes a damped or full step along the Newton direction, realized through either the forward or the reverse construction. Since the canonical manifold is generally non-affine, this finite tangential step need not remain on the manifold. The subsequent \emph{canonicalization} preserves the updated terminal density while replacing its dynamical representation with the unique canonical model associated with that density, thereby providing a valid starting point for the next stage. The composition of these two operations defines a \emph{canonical retraction} that realizes one damped or full Newton update on the canonical manifold and can be implemented through scalable matching-style regression losses. At the same time, the induced terminal-density update admits an exact finite-stepsize characterization. Thus, a single Newton Matching stage has three mutually consistent interpretations: intrinsically, it is a Newton step; computationally, it is realized through conditional-matching regression; and distributionally, its finite-stepsize effect is exactly analyzable.

The \emph{exact Newton Matching} framework admits multiple population-exact realizations, all of which preserve the scalable regression structure of diffusion and flow training. These realizations differ in the sampling constructions and auxiliary computations used to form their regression targets, yielding a range of computational options. To broaden the design space and make the trade-off between population exactness and computational efficiency explicit, we further develop \emph{approximate Newton Matching}, which replaces the exact tangential update with computationally cheaper surrogates while preserving the canonical-manifold iteration. To distinguish approximations that preserve the prescribed target at the level of stationarity from those that do not, we introduce \emph{critical-point consistency}: an approximate tangential update is critical-point consistent if its tangential displacement vanishes exactly when the current terminal density equals the target density \(\pi\). Together, exact and approximate Newton Matching organize the fragmented landscape of existing matching methods into a principled hierarchy, revealing the population-level update implemented by each method and clarifying its trade-offs among computational cost, stagewise exactness, and fidelity to the prescribed target.

Finally, we summarize the contributions of this paper as follows.

\begin{enumerate}[leftmargin=*]

\item \textbf{A paradigm shift to iterative optimization over canonical models.}
We introduce two fundamental shifts in how fine-tuning and sampling in generative modeling are conceptualized. First, rather than viewing a method as an isolated regression loss, we view learning as an iterative optimization process: the current model generates samples, each stage updates its terminal density, and the resulting density determines the model used to initialize the next stage. Second, we elevate the conditional-expectation structure of standard diffusion and flow training from a regression identity to a principle for selecting the optimization domain. Since every terminal density admits infinitely many dynamical representations, optimization over arbitrary models introduces redundant degrees of freedom and need not preserve the structure required for scalable matching. We thereby restrict attention to canonical models, each defined as the population minimizer of the standard conditional-matching objective for its terminal density. This new perspective disentangles three questions that existing methods often conflate: which distributional direction should govern the iteration, how a finite-stepsize update along that direction transforms the terminal density, and how that update can be realized through matching-style regression.

\item \textbf{The canonical manifold and its differential geometry.}
Under the compatible smooth-realization assumption, we identify canonical velocity fields with a split embedded Banach submanifold of the velocity space. The canonical-velocity map is a diffeomorphism from the density manifold onto this canonical manifold, with the terminal-density map restricted to canonical fields as its inverse. This correspondence induces a smooth, idempotent canonical projection that maps any admissible velocity field to the unique canonical representative of the same terminal density. We then characterize the tangent and cotangent spaces and derive canonical lifts of terminal-density perturbations, with equivalent representations as posterior-value gradients and posterior covariances. Finally, we transport the Fisher--Rao metric and the mixture connection from the density manifold to the canonical manifold, furnishing the first-order and second-order geometric structures required for optimization.

\item \textbf{Canonical retractions and finite-stepsize value ascent.}
Before specifying a particular objective or search direction, we develop an objective-agnostic mechanism for taking finite-stepsize updates on the canonical manifold. Given a canonical velocity model and a tangent direction, we first take a displacement in the velocity space and then canonicalize the result without changing its terminal density. This two-stage construction defines a canonical retraction, whose transport through the canonical diffeomorphism yields an equivalent retraction on the density manifold.  For tangent directions generated by admissible terminal observables, we derive an exact evolution identity along every characteristic of the updated flow, its density-level counterpart, and an explicit formula for the updated terminal density. These results yield a finite-stepsize value-ascent certificate: the expectation of every nonconstant terminal observable increases strictly, and the gain decomposes exactly into a stepsize-scaled KL divergence from the updated density to the current density and a nonnegative expected path-dissipation term.

\item \textbf{Newton Matching as Fisher--Rao gradient descent and Newton's method.}
For both fine-tuning and sampling, we specialize the canonical optimization framework to reverse-KL minimization. The negative Fisher--Rao gradient is generated by the current regularized reward, and its canonical lift admits equivalent posterior-value-gradient and conditional-covariance representations, making the direction directly learnable through matching-style regression. We then prove that the reverse-KL Hessian under the mixture connection equals the Fisher--Rao metric as a bilinear form, so the negative Fisher--Rao gradient coincides exactly with the Newton direction under the mixture connection. This Newton characterization is essential: it identifies \(\eta=\tau\) as the full step solving the linearized stationarity equation and underlies the local quadratic convergence established later. Combining this direction with the canonical retraction defines Newton Matching, identifies the prescribed target as the unique stationary terminal density, and yields an exact characterization of the induced terminal-density update.

\item \textbf{Finite-stepsize KL descent, global convergence, and local quadratic convergence.}
For a single Newton Matching stage, we derive a dissipation-corrected three-point identity and an equivalent KL-proximal characterization. These results establish strict reverse-KL descent for every stepsize \(\eta\in(0,\tau]\), including the full Newton step, unless the target has already been reached. For the resulting iteration, we prove the global convergence of the terminal densities and canonical models to the target under mild assumptions. Locally, full-step Newton Matching converges quadratically for both terminal densities and canonical models. As an optional complement, we further develop a continuation scheme along the inverse-temperature path that initializes each successive stage within the quadratic-convergence neighborhood of its target, thereby ensuring quadratic convergence throughout the continuation procedure.

\item \textbf{Exact realizations of Newton Matching.}
We convert both operations in an ideal Newton Matching stage, the tangential update and canonicalization, into population-exact sample-wise regression objectives. For the tangential update, the framework separates two independent design choices: how regression pairs are generated and how the Newton direction is represented. The \emph{forward construction} first samples an endpoint from the current model and then generates an intermediate state through the prescribed noising kernel. The \emph{reverse construction} first draws an intermediate state from a flexible proposal and then samples one or more endpoints from the corresponding canonical posterior. Within the same-marginal SDE family, we prove that this posterior requirement uniquely determines the \emph{posterior-preserving SDE}, whose terminal transition law is exactly the required canonical posterior. We further establish \emph{bridge universality}: conditioning on the states at any two times, the intervening path law of the posterior-preserving SDE is independent of the terminal density and admits an explicit Gaussian representation. This allows endpoint pairs produced by the forward construction to be augmented with exact posterior-compatible paths whenever pathwise targets are required.
The Newton direction itself admits two exact representations. The \emph{covariance form} uses regularized-reward values, for which we derive exact evaluations of the required terminal log densities and log-density ratios using either ODE or SDE calculus. To obtain the \emph{gradient form}, we introduce posterior Stein kernels and establish a covariance--gradient identity that rewrites the same posterior covariance as a conditional expectation involving regularized-reward gradients. We then construct the \emph{posterior sensitivity kernel}, a particular exact posterior Stein kernel obtained from the initial-state sensitivity of the posterior-preserving SDE. Its action can be evaluated without forming sensitivity matrices through adjoint calculus, yielding exact Mayer and Bolza realizations of the gradient-form target. In contrast to methods based on stochastic optimal control, which require deriving the Hamilton--Jacobi--Bellman (HJB) equation and associated optimality conditions, our adjoint equations arise naturally as matrix-free realizations of the initial-state sensitivity underlying the posterior Stein kernel. Stein control variates further unify the covariance and gradient forms into a broader family of population-exact targets.
We further show that exactness is preserved under multi-time supervision, allowing a single endpoint or posterior-compatible path and its associated computations to be shared across multiple regression pairs.
We prove that all resulting tangential-update losses share the same unique population minimizer and complete the stage with an explicit canonicalization loss based on conditional flow matching, together with an implicit alternative under suitable conditions.
A summary of losses for the tangential update in exact Newton Matching is provided in Table~\ref{tab:1}.

\item \textbf{Approximate realizations of Newton Matching and critical-point consistency.}
We broaden the computational design space by replacing the exact Newton tangential update with more efficient surrogates while retaining the canonical-manifold iteration. To formalize target correctness, we introduce \emph{critical-point consistency}: an approximate tangential update is critical-point consistent if it vanishes exactly when the current terminal density equals the prescribed target. As concrete examples, we derive four families satisfying this criterion: direct and split posterior-ratio linearizations in covariance form, together with reference-adjoint and Gaussian-kernel approximations in gradient form. These approximations generally differ from the exact Newton tangential update away from the target, while preserving the target as the unique density at which the update vanishes. We also study regularization trade-offs obtained by approximating the density-ratio correction or removing it altogether. Approximate regularization can reduce computation but may sacrifice critical-point consistency, whereas complete removal turns the update into direct reward ascent.
Complementary to the stage-by-stage construction, we develop a zero-displacement fixed-point perspective that translates the additive update rules underlying both exact and approximate tangential updates into stop-gradient objectives and precisely characterizes their population-stationary points.

\item \textbf{A unification of existing methods and a modular design space for new algorithms.}
We show that representative matching-based methods for fine-tuning and sampling can be recovered as particular configurations of exact or approximate Newton Matching, obtained through different choices of common building blocks, including the tangent representation, regression-pair construction, density-ratio regularization, posterior or sensitivity approximation, baselines, and stepsizes, among others \citep{potaptchik2026tilt,zheng2026diffusionnft,bergmeister2026reinforce,domingo2025adjoint,havens2026flow,havens2025adjoint}. As summarized in Table~\ref{tab:2}, this correspondence places algorithms motivated by seemingly distinct principles within a unified design space and reveals the population-level update implemented by each. It distinguishes methods that realize the exact Newton Matching update from those that retain the prescribed target only at the level of critical-point consistency or alter the underlying objective. Conversely, the same modular perspective turns this taxonomy into a constructive design principle: new exact and approximate matching algorithms can be derived systematically by recombining these building blocks.

\item \textbf{Analytic validation and broader extensions.}
We complement the general infinite-dimensional theory with an analytically tractable specialization to the isotropic Gaussian family. When the current and target densities are both Gaussian, one ideal Newton Matching stage produces another Gaussian; consequently, the entire iteration reduces exactly to a finite-dimensional recursion for the mean and variance. Analyzing this recursion directly, we prove that both forward and reverse KL divergences converge to zero and exhibit local quadratic convergence near the target. We also extend the finite-stepsize analysis by allowing time-dependent stepsize profiles. We derive their exact terminal-density formula and show that nondecreasing stepsize profiles preserve finite-stepsize value ascent and, for the regularized-reward case, reverse-KL descent. Finally, we show that Newton Matching is not tied to velocity prediction: common parameterizations represent the same canonical model, and we explicitly demonstrate Newton Matching in score and drift coordinates. We further extend the framework to one-sided interpolants with deterministic initial states.

\end{enumerate}

\section{Preliminaries}

\subsection{Diffusion and Flow}\label{subsec:1}

In this subsection, we review the basic concepts and constructions underlying diffusion and flow models.

\subsubsection{Probability Paths and Interpolants}

Let \(p_0\) and \(p_1\) be a source density and a terminal density on
\(\mathbb{R}^d\), respectively. In diffusion and flow, the goal is to learn a deterministic or stochastic evolution that transports a simple source density $p_0$ to a complex terminal density $p_1$. A family of intermediate densities $\left(p_t\right)_{t\in[0,1]}$ is called the marginal probability path. \((X_0,X_1)\sim\nu_{0,1}\) is called a coupling of \(p_0\) and \(p_1\). In this paper, we adopt the standard setting with Gaussian source \(p_0=\Normal(0,I)\) and independent coupling $\nu_{0,1}(x_0,x_1)=p_0(x_0)p_1(x_1)$.

The central idea for making training tractable is to introduce a conditional probability path $\left(p_{t| Z}\right)_{t\in[0,1]}$, whose marginalization recovers the desired probability path from $p_0$ to $p_1$. Here $Z$ denotes the conditioning variable. Common choices include the one-sided $Z=X_1$ and the two-sided $Z=(X_0,X_1)$.

The standard approach for constructing a conditional probability path is to introduce the linear interpolant $X_t=\alpha_tX_1+\beta_tX_0$, where $(\alpha_t,\beta_t)$ is referred to as the interpolant schedule. For one-sided conditioning and $t\in[0,1)$,
\[
p_{t| 1}(x_t| x_1)
=\frac{1}{\beta_t^d}p_0\left(\frac{x_t-\alpha_tx_1}{\beta_t}\right)
=\Normal\left(x_t;\alpha_tx_1, \beta_t^2 I\right).
\]
For two-sided conditioning,
\[
p_{t| 0,1}(x_t| x_0,x_1)
=\delta(x_t-(\alpha_tx_1+\beta_tx_0))
\]
is a Dirac measure concentrated on the interpolant.

For well-posedness of the flow matching framework, we impose the following regularity conditions on the schedule \((\alpha_t,\beta_t)\). We assume that
\[
(\alpha_t)_{t \in [0,1]},\ (\beta_t)_{t \in [0,1]}
\in C^1([0,1];[0,+\infty)),
\]
with \(\alpha_t>0\) and \(\beta_t>0\) for all \(t\in(0,1)\), and that they satisfy the boundary conditions
\[
\alpha_0=0,\qquad\beta_0=1,\qquad\alpha_1=1,\qquad\beta_1=0.
\]
For \(t\in(0,1)\), define
\begin{equation*}%
\kappa_t:=\beta_t^2\frac{\dd}{\dd t}\log\frac{\alpha_t}{\beta_t}=\frac{\beta_t}{\alpha_t}\left(\dot{\alpha}_t\beta_t-\alpha_t\dot{\beta}_t\right).
\end{equation*}
We further assume that the signal-to-noise ratio \(\frac{\alpha_t^2}{\beta_t^2}\) has strictly positive derivative on \((0,1)\), which implies \(\kappa_t> 0\) for all \(t\in(0,1)\). The boundary conditions may be slightly relaxed in practice to avoid endpoint singularities. We note that the above regularity conditions are satisfied by commonly used schedules in diffusion and flow models.

We use the following notation convention throughout the paper. The symbols \(X_0,X_t,X_1\) are reserved for the variables of the linear interpolant \(X_t=\alpha_t X_1+\beta_t X_0\), whereas \(Y_t\) denotes the state of an ordinary differential equation (ODE) or stochastic differential equation (SDE). The dynamics induce their own path-space distribution for \((Y_t)_{t\in[0,1]}\) and agree with the interpolant only at the level of one-time marginals:
\[
X_t\sim p_t,
\qquad
Y_t\sim p_t.
\]
However, this agreement does not imply that they have the same joint distribution across time. In particular, the conditional distribution of \(Y_1\) given \(Y_t\) need not coincide with that of \(X_1\) given \(X_t\). When a terminal state generated by a trajectory is subsequently used as an endpoint of the interpolant, we make this change of role explicit by writing \(X_1:=Y_1\).

\subsubsection{Deterministic Dynamics and Flow Maps}

When sampling with deterministic dynamics, the state \(Y_t\) evolves according to the ODE
\begin{equation}
\label{eq:1}
\frac{\dd Y_t}{\dd t}=v_t(Y_t), \qquad Y_0\sim p_0,
\end{equation}
where \(v_t:\mathbb{R}^d\to\mathbb{R}^d\) is the marginal velocity field. The probability transport is governed by the continuity equation
\begin{equation}
\partial_t p_t(x_t)+\nabla\cdot(p_t(x_t)v_t(x_t))=0, \qquad p_0=\Normal(0,I).
\label{eq:2}
\end{equation}

For ODE sampling, given a velocity field \(v_t\), it is convenient to define flow maps over the time interval \([0,1]\). For \(t_1,t_2\in[0,1]\), the flow \(\flow_{t_1\to t_2}^v:\R^d\to\R^d\) carries a state at time \(t_1\) to the corresponding state at time \(t_2\) along the ODE:
    \begin{align*}
        &\frac{\dd}{\dd s} \flow_{t_1 \to s}^v(x)=v_s\left(\flow_{t_1 \to s}^v(x)\right), \\
        &\flow_{t_1 \to t_1}^v(x)=x.
    \end{align*}
Both \(t_1\leq t_2\) and \(t_1>t_2\) are allowed, corresponding to traversing the ODE solution forward and backward in time, respectively. Under standard well-posedness assumptions, the flow maps satisfy \(\flow_{t_1\to t_2}^v=\left(\flow_{t_2\to t_1}^v\right)^{-1}\).

\subsubsection{Conditional Flow Matching}

Although infinitely many velocity fields can transport \(p_0\) to \(p_1\), they are generally not available in closed form. Flow matching provides a scalable way to learn one such field by making use of conditional velocity fields \(v_{t| Z}(x| Z)\). For two-sided conditioning,
\[
v_{t| 0,1}(x_t| x_0,x_1)=\dot{\alpha}_tx_1+\dot{\beta}_tx_0.
\]
For one-sided conditioning and $t\in[0,1)$,
\[
v_{t| 1}(x_t| x_1)
=\dot{\alpha}_tx_1+\dot{\beta}_t\frac{x_t-\alpha_tx_1}{\beta_t}
=\frac{\dot{\beta}_t}{\beta_t}x_t+\frac{\dot{\alpha}_t\beta_t-\dot{\beta}_t\alpha_t}{\beta_t}x_1.
\]
They lead to the same results, though in different contexts one may be more convenient than the other for ease of presentation.

Given a parameterized velocity field $v_t^\theta(x_t)$, the conditional flow matching (CFM) loss is defined as
\begin{align}
\mathcal{L}_{\mathrm{CFM}}(\theta)
=&\E_{\substack{t\sim\U(0,1),\,X_1\sim p_1,\\X_t\sim p_{t| 1}(\cdot| X_1)}}
\left[\norm[2]{v_t^\theta(X_t)-v_{t| 1}(X_t| X_1)}^2\right]\notag\\
=&\E_{\substack{t\sim\U(0,1),\,X_1\sim p_1,\,X_0\sim p_0,\\X_t\sim p_{t| 0,1}(\cdot| X_0,X_1)}}
\left[\norm[2]{v_t^\theta(X_t)-v_{t| 0,1}(X_t| X_0,X_1)}^2\right].\label{eq:3}
\end{align}
Its population minimizer is the conditional expectation
\begin{align}
v_t(x_t)
=&\E\left[v_{t| 0, 1}(X_t| X_0, X_1) \middle| X_t = x_t\right]=\E\left[\dot{\alpha}_tX_1+\dot{\beta}_tX_0 \middle| X_t = x_t\right],\label{eq:4}
\end{align}
which is one specific velocity field that can transport \(p_0\) to \(p_1\).

\subsubsection{Stochastic Dynamics}

When sampling with stochastic dynamics, we use the SDE
\begin{equation}
\label{eq:5}
\dd Y_t= \left(v_t(Y_t) + \frac{\sigma_t^2}{2}\nabla\log p_t(Y_t)\right)\dd t + \sigma_t \dd W_t, \qquad Y_0\sim p_0,
\end{equation}
where $W_t$ denotes standard Brownian motion and $\sigma_t$ is a scalar noise schedule. The evolution of the density is governed by the Fokker--Planck equation
\begin{equation*}
\partial_t p_t(x_t)+\nabla\cdot\left(p_t(x_t)b_t(x_t)\right)=\frac{\sigma_t^2}{2}\Delta p_t(x_t), \qquad p_0=\Normal(0,I),
\end{equation*}
where $b_t(x_t)=v_t(x_t)+\frac{\sigma_t^2}{2}\nabla_{x_t}\log p_t(x_t)$ denotes the drift. The SDE~\eqref{eq:5} and the ODE~\eqref{eq:1} produce the same marginal probability path $\{p_t\}_{t\in[0,1]}$ for any choice of $\sigma_t\ge 0$.

\subsubsection{Equivalent Predictions}

Diffusion and flow models admit equivalent predictions in terms of velocity, score, drift, data, and noise under standard conditions. The relation between velocity and score provides one example. For the Gaussian source and linear interpolant, the score $\nabla_{x_t}\log p_t(x_t)$ satisfies
\begin{equation*}
\nabla_{x_t}\log p_t(x_t) = -\frac{1}{\beta_t}\E\left[X_0\middle| X_t=x_t\right].
\end{equation*}
Together with the interpolant identity
\begin{equation*}
x_t=\alpha_t \E\left[X_1\middle| X_t=x_t\right]+\beta_t\E\left[X_0\middle| X_t=x_t\right],
\end{equation*}
it suffices to know either the velocity or the score, since one can be derived from the other. This observation extends to other prediction forms as well. In this paper, we focus on the standard flow models (i.e., predicting the velocity); extensions of our method to other prediction forms are provided in Appendix~\ref{app:8}.

\subsection{Differential Geometry on Banach Manifolds}

This subsection briefly reviews the differential geometry of Banach manifolds \citep{lang1999fundamentals}. Most notations used in this review are generic and should be understood as local to this subsection.

\subsubsection{Banach Space}

A real Banach space $\mathbb{B}$ is a normed vector space that is complete under its norm $\norm[\mathbb{B}]{\cdot}$. The Banach space $\mathbb{B}$ may be of finite or infinite dimension.

Consider an open subset $\mathcal{O}_\mathbb{B}\subset\mathbb{B}$ and another Banach space $\mathbb{A}$. A map $F:\mathcal{O}_\mathbb{B}\to\mathbb{A}$ is Fr\'echet differentiable at $x\in\mathcal{O}_\mathbb{B}$ if there exists a continuous linear map $\DD_x F:\mathbb{B}\to\mathbb{A}$ such that
\[
F(x+h)=F(x)+\DD_x F[h]+o(\norm[\mathbb{B}]{h})
\]
as $h\to0$. The map $\DD_x F$ is called the Fr\'echet derivative of $F$ at $x$.

Higher-order Fr\'echet derivatives are defined recursively by differentiating the derivative map $x\mapsto \DD_x F$. A map is called smooth if Fr\'echet derivatives of all orders exist and are continuous.

\subsubsection{Banach Manifold}

A smooth Banach manifold $\mathscr{M}$, modeled on a Banach space $\mathbb{B}$, is a Hausdorff topological space equipped with a collection of charts whose domains cover $\mathscr{M}$. A chart is a pair $(U,\varphi)$, where $U\subset\mathscr{M}$ is open and $\varphi:U\to\varphi(U)\subset\mathbb{B}$ is a homeomorphism onto an open subset $\varphi(U)\subset\mathbb{B}$. $\varphi(p)$ is the coordinate representation of $p$.

Whenever two charts $(U,\varphi)$ and $(V,\psi)$ overlap, their coordinate-change maps 
\[
\psi\circ\where{\varphi^{-1}}{\varphi(U\cap V)}:\varphi(U\cap V)\to\psi(U\cap V),\qquad
\varphi\circ\where{\psi^{-1}}{\psi(U\cap V)}:\psi(U\cap V)\to\varphi(U\cap V)
\]
are smooth in the Fr\'echet sense.

Let $\mathscr{M}$ and $\mathscr{N}$ be smooth Banach manifolds. A map $F:\mathscr{M}\to\mathscr{N}$ is called smooth if, for every $p\in\mathscr{M}$, there exist a chart
$(U,\varphi)$ of $\mathscr{M}$ around $p$ and a chart $(V,\psi)$ of
$\mathscr{N}$ around $F(p)$, with $F(U)\subset V$, such that the coordinate
representation
\[
\psi\circ F\circ\varphi^{-1}
:
\varphi(U)\to\psi(V)
\]
is smooth in the Fr\'echet sense.

A smooth bijection $F:\mathscr{M}\to\mathscr{N}$ whose inverse $F^{-1}:\mathscr{N}\to\mathscr{M}$ is also smooth is called a diffeomorphism, denoted by $F:\mathscr{M}\diffeomorphismto\mathscr{N}$ or $\mathscr{M}\cong\mathscr{N}$. Diffeomorphic manifolds have equivalent smooth structures. In particular, their tangent vectors and the geometric structures introduced below can be transported through the differential of $F$.

A subset $\mathscr{S}\subset\mathscr{M}$ is called a split embedded Banach submanifold if, for every $p\in\mathscr{S}$, there exist a chart $(U,\varphi)$ of $\mathscr{M}$ around $p$ and a closed complemented linear subspace $\mathbb{B}_0\subset\mathbb{B}$ such that
\[
\varphi(U\cap\mathscr{S})
=
\varphi(U)\cap\mathbb{B}_0.
\]

\subsubsection{Tangent Space}

A tangent vector at $p\in\mathscr{M}$ represents a first-order perturbation of $p$ that remains in the manifold. Consider a smooth curve
\[
(p^\varepsilon)_{\varepsilon\in(-\delta,\delta)}\subset\mathscr{M},
\qquad
p^0=p
\]
for some $\delta>0$. In a chart $(U,\varphi)$ around $p$, two smooth curves $(p^\varepsilon)_{\varepsilon\in(-\delta,\delta)}$ and $(q^\varepsilon)_{\varepsilon\in(-\delta,\delta)}$ through $p$ are said to represent the same tangent vector if
\[
\where{\frac{\dd}{\dd\varepsilon}\varphi(p^\varepsilon)}{\varepsilon=0}
=
\where{\frac{\dd}{\dd\varepsilon}\varphi(q^\varepsilon)}{\varepsilon=0}\in\mathbb{B}.
\]
This equivalence relation is independent of the selected chart.

A tangent vector is an equivalence class of such smooth curves. The tangent vector represented by $(p^\varepsilon)_{\varepsilon\in(-\delta,\delta)}$ is denoted by
\[
\xi=\where{\frac{\dd}{\dd\varepsilon}p^\varepsilon}{\varepsilon=0},
\]
where the derivative is understood through any local chart as above.

The tangent space of $\mathscr{M}$ at $p$ is defined as the collection of all tangent vectors at $p$, denoted by
\[
T_p\mathscr{M}=\left\{\where{\frac{\dd}{\dd\varepsilon}p^\varepsilon}{\varepsilon=0}:(p^\varepsilon)_{\varepsilon\in(-\delta,\delta)}\subset\mathscr{M}\text{ smooth},\,p^0=p,\,\delta>0\right\}.
\]
Here $p\in\mathscr{M}$ is called the base point. Every chart around $p$ identifies $T_p\mathscr{M}$ with the Banach model space $\mathbb{B}$. In particular, $T_p\mathscr{M}$ is itself a Banach space:
\[
T_p\mathscr{M}\cong\mathbb{B}.
\]

Let $F:\mathscr{M}\to\mathscr{N}$ be smooth. The differential of $F$ at $p\in\mathscr{M}$ is the continuous linear map
\[
\dd_p F:T_p\mathscr{M}\to T_{F(p)}\mathscr{N},\qquad
\dd_p F[\xi]
:=
\where{\frac{\dd}{\dd\varepsilon}F(p^\varepsilon)}{\varepsilon=0},
\]
where $(p^\varepsilon)_{\varepsilon\in(-\delta,\delta)}$ is any smooth curve satisfying $p^0=p$ and $\where{\frac{\dd}{\dd\varepsilon}p^\varepsilon}{\varepsilon=0}=\xi$. The value $\dd_p F[\xi]$ is independent of the selected path representing $\xi$.
Thus, the differential maps a first-order perturbation of $p$ to the corresponding first-order perturbation of $F(p)$.

Throughout this paper, lowercase $\dd_p F$ denotes the intrinsic differential of a smooth map between Banach manifolds, whereas uppercase $\DD_x F$ denotes the Fr\'echet derivative of a map between Banach spaces, or of a coordinate representation after charts have been selected. More precisely, for charts $(U,\varphi)$ around $p$ and $(V,\psi)$ around $F(p)$, the two notations are related by
\begin{equation*}
    \DD_{\varphi(p)}\left(\psi\circ F\circ\varphi^{-1}\right)
    =
    \dd_{F(p)}\psi
    \circ
    \dd_p F
    \circ
    \left(\dd_p\varphi\right)^{-1}.
\end{equation*}
Thus, $\dd_p F$ is intrinsic and independent of the selected charts, while the Fr\'echet derivative on the left is taken in the corresponding Banach coordinates. When the manifolds are open subsets of Banach spaces, the two are identified through the canonical identifications of their tangent spaces.

A tangent vector field $X$ on $\mathscr{M}$ is a smoothly varying assignment
\[
p\mapsto X_p\in T_p\mathscr{M}.
\]
The space of smooth tangent vector fields on $\mathscr{M}$ is denoted by $\mathfrak{X}(\mathscr{M})$. In summary, a tangent vector describes one perturbation direction at one point, whereas a tangent vector field specifies a perturbation direction at every point of the manifold.

\subsubsection{Cotangent Space}\label{sub2sec:1}

The cotangent space of $\mathscr{M}$ at $p$ is the continuous dual of the tangent space:
\[
T_p^*\mathscr{M}:=(T_p\mathscr{M})^*=\left\{\text{continuous linear functional }T_p\mathscr{M}\to\R\right\}.
\]

An element $\ell\in T_p^*\mathscr{M}$ is called a cotangent vector. Its action on a tangent vector $\xi\in T_p\mathscr{M}$ is denoted by $\ell[\xi]\in\mathbb{R}$.

Let $J:\mathscr{M}\to\mathbb{R}$ be a smooth objective. Its differential at $p$ is the cotangent vector
\[
\dd_p J\in T_p^*\mathscr{M},\qquad
\dd_p J[\xi]
:=
\left.
\frac{\dd}{\dd\varepsilon}J(p^\varepsilon)
\right|_{\varepsilon=0},
\]
where $(p^\varepsilon)_{\varepsilon\in(-\delta,\delta)}$ is any smooth curve satisfying $p^0=p$ and $\where{\frac{\dd}{\dd\varepsilon}p^\varepsilon}{\varepsilon=0}=\xi$. The value $\dd_p J[\xi]$ is independent of the selected path representing $\xi$.
It describes the first-order change of the objective $J$ along the tangent direction $\xi$.

For a tangent vector field $X\in\mathfrak{X}(\mathscr{M})$, the directional derivative of $J$ along $X$ is defined by
\[
X[J]:\mathscr{M}\to\R,\qquad X[J](p):=\dd_p J[X_p].
\]
Consider a smooth curve $(p^\varepsilon)_{\varepsilon\in(-\delta,\delta)}\subset\mathscr{M}$ which passes through $p$ at $\varepsilon=0$ and satisfies $X_p=\where{\frac{\dd}{\dd\varepsilon}p^\varepsilon}{\varepsilon=0}$. Then, we can also write
\[
X[J](p)=\dd_p J[X_p]=\where{\frac{\dd}{\dd\varepsilon}J(p^\varepsilon)}{\varepsilon=0}.
\]

\subsubsection{Riemannian Metric and Gradient}\label{sub2sec:2}

A Riemannian metric on $\mathscr{M}$ is a smoothly varying family of continuous, symmetric, and positive-definite bilinear forms
\[
g_p:T_p\mathscr{M}\times T_p\mathscr{M}\to\R,\qquad p\in\mathscr{M}.
\]
The metric induces a continuous linear map
\[
g_p^\flat:T_p\mathscr{M}\to T_p^*\mathscr{M},\qquad
g_p^\flat(\xi)[\zeta]:=g_p(\xi,\zeta),\qquad \xi,\zeta\in T_p\mathscr{M},
\]
which is called the flat map. Positive definiteness implies that \(g_p^\flat\) is injective.
Its inverse is defined as
\[
(g_p^\flat)^{-1}:
\operatorname{Im}(g_p^\flat)\to T_p\mathscr{M}.
\]
For a cotangent vector $\ell\in\mathrm{Im}(g_p^\flat)$, the unique tangent vector $(g_p^\flat)^{-1}(\ell)\in T_p\mathscr{M}$ is called the metric dual of $\ell$.
The metric is called strong if
\(
g_p^\flat:T_p\mathscr{M}\to T_p^*\mathscr{M}
\)
is a Banach-space isomorphism for every \(p\in\mathscr{M}\); otherwise, it is called weak. For a weak Riemannian metric, it may hold that
\(
\mathrm{Im}(g_p^\flat)
\subsetneq
T_p^*\mathscr{M};
\) in this case, not every cotangent vector has a metric-dual tangent representation.

Let $J:\mathscr{M}\to\R$ be a smooth objective. Whenever $\dd_p J\in g_p^\flat(T_p\mathscr{M})$, the gradient of $J$ at $p$ is the unique tangent vector $\grad_g J(p)\in T_p\mathscr{M}$ satisfying
\[
g_p\bigl(\grad_g J(p),\xi\bigr)=\dd_p J[\xi]
\]
for every $\xi\in T_p\mathscr{M}$.

Thus, the differential $\dd_p J$ contains the intrinsic first-order information of the objective, whereas the gradient is its tangent-vector representation under the selected metric. For a weak Riemannian metric, the gradient need not exist for every smooth objective, and its existence must be verified in the setting under consideration.

\subsubsection{Affine Connection}

Tangent vectors at different points belong to different tangent spaces and therefore cannot be directly subtracted. An affine connection provides a rule for differentiating one tangent vector field along another.

An affine connection on \(\mathscr{M}\) is a local covariant derivative on the tangent bundle. In global-section notation, it is written as a map
\[
\nabla:\mathfrak{X}(\mathscr{M})\times\mathfrak{X}(\mathscr{M})\to\mathfrak{X}(\mathscr{M}),\qquad (X,Y)\mapsto\nabla_XY,
\]
satisfying
\begin{subequations}\label{eq:6}
    \begin{align}
    &\nabla_{fX+hY}Z=f\nabla_XZ+h\nabla_YZ,\label{eq:7}\\
    &\nabla_X(aY+bZ)=a\nabla_XY+b\nabla_XZ,\label{eq:8}\\
    &\nabla_X(fY)=X[f]Y+f\nabla_XY,\label{eq:9}
    \end{align}
\end{subequations}
for all tangent vector fields $X,Y,Z\in\mathfrak{X}(\mathscr{M})$, smooth scalar functions $f,h:\mathscr{M}\to\R$, and constants $a,b\in\R$. The last equation is the Leibniz rule in its second argument. The same notation is used for vector fields defined on an open subset \(U\subset\mathscr{M}\), compatibly with restriction to smaller open subsets.

Note that, for consistency with the differential-geometry literature, we use \(\nabla\) to denote an affine connection. The same symbol denotes the usual gradient operator when applied to scalar functions on \(\mathbb{R}^d\); its meaning is clear from context.

\subsubsection{Covariant Hessian}\label{sub2sec:3}

Let $J:\mathscr{M}\to\R$ be a smooth objective, and let $\nabla$ be an affine connection on $\mathscr{M}$. The covariant Hessian of $J$ at $p\in\mathscr{M}$ is the bilinear form
\[
\operatorname{Hess}^{\nabla}J_p:T_p\mathscr{M}\times T_p\mathscr{M}\to\R,\qquad
\operatorname{Hess}^{\nabla}J_p(\xi,\zeta):=X[Y[J]](p)-\dd_p J\bigl[(\nabla_XY)_p\bigr],
\]
where $X,Y$ are arbitrary smooth local tangent vector fields satisfying $X_p=\xi$, $Y_p=\zeta$. The value $\operatorname{Hess}^{\nabla}J_p(\xi,\zeta)$ is independent of the selected extensions $X$ and $Y$. The covariant Hessian can equivalently be viewed as a continuous linear operator
\[
\operatorname{Hess}^{\nabla}J_p:T_p\mathscr{M}\to T_p^*\mathscr{M},\qquad
\bigl(\operatorname{Hess}^{\nabla}J_p[\xi]\bigr)[\zeta]:=\operatorname{Hess}^{\nabla}J_p(\xi,\zeta).
\]
For a fixed objective \(J\), the covariant Hessian depends on the selected affine connection, does not require a Riemannian metric, and need not be symmetric in general. For brevity, we refer to $\operatorname{Hess}^{\nabla}J$ simply as the Hessian of $J$ throughout this paper.

\section{Problem Formulation}

\subsection{Sampling and Fine-Tuning}
\label{subsec:2}

In this paper, we consider the following target density:
\begin{equation*}
    \pi_{\mu,\tau,r}(x)
    :=\frac{\mu(x)e^{\tau r(x)}}{\int_{\R^d}\mu(z)e^{\tau r(z)}\dd z}
    \propto \mu(x)e^{\tau r(x)},
\end{equation*}
where $\mu:\R^d\to(0,\infty)$ is a reference factor, $r:\R^d\to\R$ is a reward function, and $\tau>0$ is an inverse temperature. The normalizing constant satisfies
\[
0<\int_{\R^d}\mu(z)e^{\tau r(z)}\dd z<\infty.
\]
The additional regularity assumptions required by the subsequent analysis are stated in Appendix~\ref{app:1}. The reference factor $\mu$ has two cases of interest:
\begin{itemize}
    \item $\mu(x)=1$, which corresponds to sampling from the Boltzmann target density 
    \begin{equation*}%
        \pi_{1,\tau,r}(x)\propto \exp(\tau r(x)).
    \end{equation*}
    \item $\mu(x)=\rho^\base(x)$, where $\rho^\base$ is the terminal density of a base model $v_t^\base$ trained with standard diffusion or flow losses. This corresponds to fine-tuning the base model to sample from the tilted density 
    \begin{equation*}%
        \pi_{\rho^\base,\tau,r}(x)\propto \rho^\base(x)e^{\tau r(x)}.
    \end{equation*}
    Typically, we have access to the base model, but its terminal density $\rho^\base$ is implicit and not directly available. We can sample from $\rho^\base$ by solving the ODE or SDE induced by $v_t^\base$.
\end{itemize}

Our goal is to train a diffusion or flow model to sample from the target density \(\pi_{\mu,\tau,r}\), thereby covering both sampling from unnormalized densities and fine-tuning pretrained models. In both cases, the task is distinct from standard generative modeling, where samples from the target density are directly available, and the model can be trained via supervised learning. In contrast, here we only have access to the reward function $r(x)$, and possibly its gradient when $r$ is differentiable and the algorithm requires it. The model must therefore be trained by leveraging the reward signal, which places the problem in the reinforcement learning (RL) setting.

Existing works often treat the two cases $\mu=1$ and $\mu=\rho^\base$ separately, with different algorithms developed for each. In this paper, we propose a unified approach that handles both cases simultaneously. We present our method primarily in the standard velocity-prediction formulation of flow models, but the theory and algorithms also apply to other commonly used prediction forms. We illustrate extensions to score and drift forms in Appendix~\ref{app:8}.

\subsection{Optimization on Velocity Fields}\label{subsec:3}

Given a velocity field \(v\), the ODE~\eqref{eq:1} transports the source density \(p_0\) to a terminal density \(\rho^v:=p_1^v\). Specifically, $\rho^v$ is the pushforward of \(p_0\) under the flow map \(\flow_{0\to 1}^v\):
\begin{equation*}
    \rho^v=\left(\flow_{0\to 1}^v\right)_\#p_0.
\end{equation*}

To align the terminal density \(\rho^v\) with the target density \(\pi_{\mu,\tau,r}\), we introduce an objective that measures the discrepancy between them. Choosing the reverse KL divergence as the objective, we seek to solve
\begin{equation}
    \min_v\,\KL{\rho^v}{\pi_{\mu,\tau,r}}.
    \label{eq:10}
\end{equation}
Thus, the task can be viewed as a \emph{functional optimization} problem over the space of velocity fields. Any velocity field \(v^\star\) that transports \(p_0\) to \(\pi_{\mu,\tau,r}\) satisfies \(\rho^{v^\star}=\pi_{\mu,\tau,r}\) and achieves the global minimum \(\KL{\rho^{v^\star}}{\pi_{\mu,\tau,r}}=0\).

Although optimization problem~\eqref{eq:10} provides a natural formulation of our task, it is not directly amenable to scalable optimization. The dependence of the terminal density \(\rho^v\) on the velocity field \(v\) is specified only implicitly through the associated continuity equation~\eqref{eq:2}. Consequently, a local perturbation of \(v\) at an intermediate time can propagate through the entire flow and alter \(\rho^v\) globally, making the map \(v\mapsto \rho^v\) difficult to handle numerically. More importantly, the terminal KL objective does not expose the conditional-expectation structure that underlies the scalability of the flow-matching framework. Optimizing it naïvely would generally require lifting the terminal-density objective to path space, constructing a suitable surrogate objective, and, most likely, backpropagating through the entire sampling trajectory. This would forgo the central advantage of CFM, which transforms an intractable marginal objective into a tractable sample-wise regression loss. We therefore need to improve the formulation in~\eqref{eq:10} so that it admits CFM-like optimization and leads to scalable algorithms.

To develop such a formulation, we reinterpret the flow-matching construction as a map from terminal densities to velocity fields. We first observe that, once the source density \(p_0\) and the schedule \((\alpha_t,\beta_t)\) are fixed, two central ingredients are determined independently of the terminal density \(p_1\). The first is the conditional probability path
\begin{equation}
    p_{t| 1}(x_t| x_1)=\frac{1}{\beta_t^d}p_0\left(\frac{x_t-\alpha_t x_1}{\beta_t}\right),
    \label{eq:11}
\end{equation}
and the second is the conditional velocity field
\begin{equation}
    v_{t| 1}(x_t| x_1)=\frac{\dot{\beta}_t}{\beta_t}x_t+\frac{\dot{\alpha}_t\beta_t-\alpha_t\dot{\beta}_t}{\beta_t}x_1=\frac{\dot{\beta}_t}{\beta_t}x_t+\frac{\alpha_t\kappa_t}{\beta_t^2}x_1.
    \label{eq:12}
\end{equation}
Both depend only on the source density \(p_0\) and the selected schedule, not on the terminal density \(p_1\).

By contrast, several induced quantities depend on the terminal density. Let \(\rho\) be a terminal density and set \(p_1=\rho\) in the flow-matching construction. To make this dependence explicit, we attach \(\rho\) as a superscript to the corresponding quantities. The first is the marginal probability path
\begin{equation*}
    p_t^\rho(x_t)=\int_{\R^d}\rho(x_1)p_{t| 1}(x_t| x_1)\,\dd x_1.
\end{equation*}
The second is the posterior
\begin{equation}
    p_{1| t}^\rho(x_1| x_t)=\frac{\rho(x_1)p_{t| 1}(x_t| x_1)}{p_t^\rho(x_t)}.
    \label{eq:13}
\end{equation}
The third is the marginal velocity field, which is the population minimizer of the CFM loss:
\begin{equation}
    v_t^\rho(x_t)=\E_{X_1\sim p_{1| t}^\rho(\cdot| x_t)}\left[v_{t| 1}(x_t| X_1)\right].
    \label{eq:14}
\end{equation}
By construction, \(v^\rho\) transports \(p_0\) to \(\rho\). Equivalently, \(v^\rho\) and \(p_t^\rho\) satisfy the continuity equation
\begin{equation*}
    \partial_t p_t^\rho(x_t)+\nabla_{x_t}\cdot\left(p_t^\rho(x_t)v_t^\rho(x_t)\right)=0,\qquad p_0^\rho=p_0,\qquad p_1^\rho=\rho.
\end{equation*}

\begin{figure}[t]
    \centering
    \input{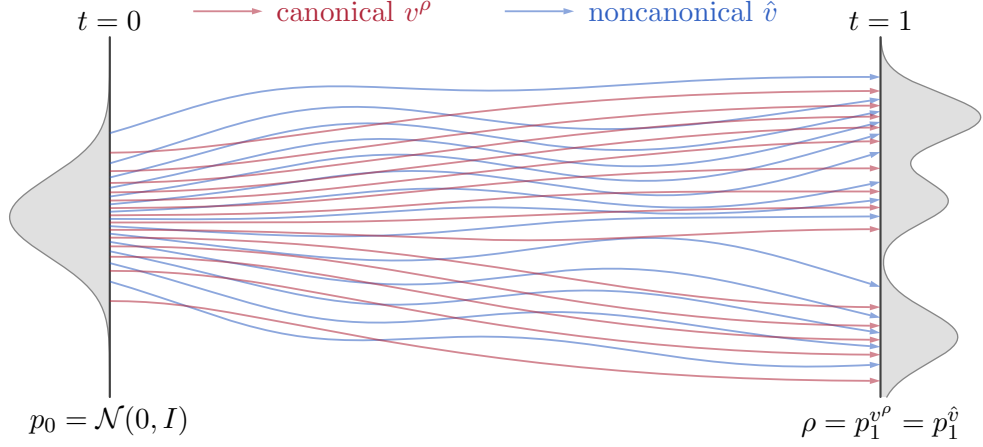}
    \caption{Nonuniqueness of velocity representations for terminal densities. The canonical velocity field $v^\rho$ and a distinct noncanonical velocity field $\hat{v}$ both transport $p_0=\Normal(0,I)$ to the same terminal density $\rho$.}
    \label{fig:5}
\end{figure}

Given a terminal density \(\rho\), there are generally infinitely many velocity fields that transport \(p_0\) to \(\rho\), as illustrated in Figure~\ref{fig:5}. The flow-matching construction selects the particular field \(v^\rho\) through the conditional-expectation representation~\eqref{eq:14}. We refer to velocity fields obtained in this way as \emph{canonical velocity fields}. Their conditional-expectation structure is especially valuable algorithmically: in the standard generative modeling setting, where samples from \(\rho\) are available, it allows \(v^\rho\) to be learned through sample-wise regression.

In the tasks considered in this paper, we do not have direct access to samples from the target density \(\pi_{\mu,\tau,r}\). Nevertheless, the target density determines a canonical velocity field \(v^{\pi_{\mu,\tau,r}}\). Our goal is not merely to find any velocity field that transports \(p_0\) to \(\pi_{\mu,\tau,r}\), but specifically to recover this canonical field. Moreover, throughout the iterative optimization process, we require all intermediate velocity fields to remain canonical. This leads to the constrained optimization problem
\begin{equation}
    \begin{aligned}
        \min_v\,\, &\KL{\rho^v}{\pi_{\mu,\tau,r}}\\
        \text{s.t.}\,\, &v\text{ is a canonical velocity field}.
    \end{aligned}
    \label{eq:15}
\end{equation}
We restate this problem precisely in Section~\ref{sec:3}, after introducing the necessary ingredients.
In contrast to~\eqref{eq:10}, the formulation~\eqref{eq:15} restricts the search to canonical velocity fields, each corresponding to a terminal density \(\rho\) through~\eqref{eq:14}. This choice is both natural and essential: a canonical velocity field is accompanied by a well-defined marginal probability path and posterior density, thereby preserving the conditional-expectation structure required for CFM-like optimization.

This viewpoint motivates us to characterize this special class of velocity fields. For fixed \(p_0\) and \((\alpha_t,\beta_t)\), each terminal density \(\rho\) induces a unique canonical velocity field \(v^\rho\), while each canonical velocity field determines its terminal density through its flow. Thus, canonical velocity fields and terminal densities are in one-to-one correspondence. Geometrically, the image of the map \(\rho\mapsto v^\rho\) can be viewed as a \emph{canonical manifold} embedded in the space of all velocity fields. We next develop this geometric perspective, which forms the foundation of our Newton Matching framework.

\section{Canonical Geometry}\label{sec:1}

Section~\ref{subsec:3} illustrates that a terminal density does not uniquely determine a velocity representation: the map from velocity fields to terminal densities is many-to-one. The canonical construction resolves this ambiguity by selecting, for each terminal density, a distinguished velocity representative with the conditional-expectation structure underlying scalable CFM training.

This section develops the geometry of this canonical correspondence and equips the canonical manifold with the differential structures required for first-order and second-order optimization. Section~\ref{subsec:4} establishes the global geometry: the canonical velocity fields form an embedded submanifold of the velocity space, and the canonical-velocity map is a diffeomorphism from the density manifold onto this submanifold. The first-order development in Section~\ref{subsec:5} characterizes the tangent and cotangent spaces, constructs canonical lifts of density perturbations, and transports the Fisher--Rao metric to the canonical manifold. These structures provide the admissible directions used by the canonical retraction in Section~\ref{sec:2} and the metric used to derive the Fisher--Rao gradient in Section~\ref{subsec:9}. The second-order development in Section~\ref{subsec:6} characterizes the mixture connection through variations of Fisher--Rao pairings against fixed terminal observables and transports this connection to the canonical manifold. It provides the covariant differentiation needed to define the Hessian and Newton direction in Section~\ref{subsec:10}. The Banach-manifold realization and the regularity assumptions underlying these constructions are deferred to Appendix~\ref{app:1}.

\subsection{Global Geometry}\label{subsec:4}

Throughout this paper, the source density $p_0$ and the interpolant schedule $(\alpha_t,\beta_t)_{t\in[0,1]}$ are fixed. Unless stated otherwise, Assumptions~\ref{assume:1} and~\ref{assume:2} are in force throughout the remainder of the paper. Stronger regularity assumptions are invoked only for the particular first-order or second-order calculations that require them.

\begin{figure}[t]
    \centering
    \input{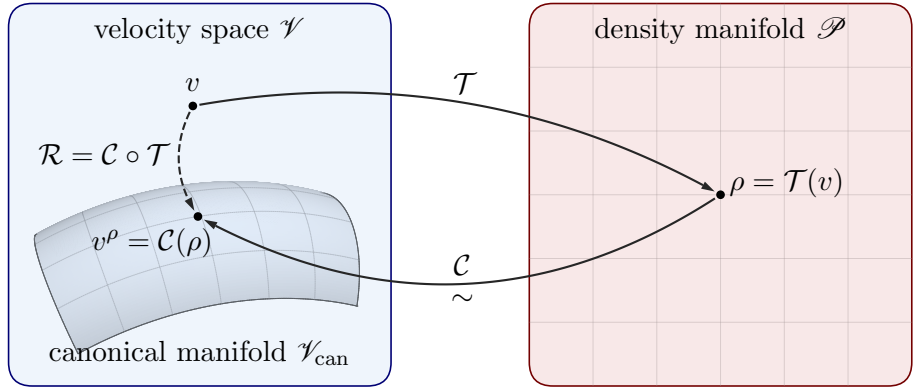}
    \caption{Global geometry of the canonical construction.}
    \label{fig:6}
\end{figure}

This subsection develops the global geometry of the canonical construction. We first formalize the relationship between velocity fields and terminal densities, and then use this relationship to define the canonical manifold and the canonical projection. The resulting structure is summarized in Figure~\ref{fig:6}.

As introduced in Section~\ref{subsec:3}, two objects are central to the present construction: the velocity field, which is the trainable object defining the sampling dynamics, and the terminal density induced by its flow. The following definition specifies the spaces in which these objects live and the maps that relate them.

\begin{definition}\label{def:1}
    The \textit{velocity space} $\velspace$ is a Banach space satisfying
    \begin{equation*}
        \velspace\subset\left\{v=\left(v_t\right)_{t\in[0,1]}:[0,1]\times \R^d\to\R^d\right\}.
    \end{equation*}

    The \textit{density manifold} $\pdfspace$ is a Banach manifold satisfying
    \begin{equation*}%
        \pdfspace\subset\left\{
        \rho\in C\left(\R^d;(0,\infty)\right): \int_{\R^d}\rho(x)\,\dd x=1\right\}.
    \end{equation*}

    The \textit{terminal-density map} $\terminalmap$ transforms a velocity field $v\in\velspace$ to the terminal density $\rho^v\in\pdfspace$, i.e.,
    \begin{equation*}
        \terminalmap:\velspace\to\pdfspace,\qquad v\mapsto \rho^v:=(\flow_{0\to1}^v)_\#p_0,
    \end{equation*}
    where $\#$ denotes the pushforward operator.

    The \textit{canonical-velocity map} $\canonicalmap$ is defined as
    \begin{equation*}
        \canonicalmap:\pdfspace\to\velspace,\qquad \rho\mapsto v^\rho,\qquad v_t^\rho(x_t)=\E_{X_1\sim p_{1|t}^\rho(\cdot| x_t)}\left[v_{t|1}(x_t| X_1)\right],
    \end{equation*}
    where the posterior density $p_{1|t}^\rho(\cdot|x_t)$ is given in~\eqref{eq:13}. $v^\rho\in\velspace$ is called the \textit{canonical velocity field} of the terminal density $\rho\in\pdfspace$.
\end{definition}

Recall the nonuniqueness illustrated in Figure~\ref{fig:5}: the terminal-density map $\terminalmap:\velspace\to\pdfspace$ forgets the particular velocity representation and retains only its terminal density through its flow. For a fixed density $\rho\in\pdfspace$, the preimage $\terminalmap^{-1}(\rho)$ generally consists of different velocity fields that transport $p_0$ to $\rho$. The canonical-velocity map $\canonicalmap:\pdfspace\to\velspace$ selects the CFM representative $v^\rho\in\velspace$ specified by the conditional-expectation formula above. The key compatibility property, as stated in Theorem~\ref{thm:1}, is that the canonical velocity field $v^\rho$ also realizes the terminal density $\rho$, i.e.,
\begin{equation*}
    \terminalmap(v^\rho)=\rho.
\end{equation*}

Having associated each terminal density with a distinguished canonical velocity field, we now regard the collection of all such fields as a geometric subset of the velocity space. This subset serves as the intrinsic search space for Newton Matching. We also introduce a map that replaces an arbitrary velocity field by the canonical representative associated with its terminal density.

\begin{definition}%
    The \textit{canonical manifold} $\velmancan$ is the collection of canonical velocity fields, i.e.,
    \begin{equation*}
        \velmancan:=\canonicalmap(\pdfspace)\subset\velspace.
    \end{equation*}

    The \textit{canonical projection} $\projection$ is defined as
    \begin{equation*}
        \projection:=\canonicalmap\circ\terminalmap:\velspace\to\velmancan,\qquad v\mapsto v^{\rho^v}=\canonicalmap(\terminalmap(v)).
    \end{equation*}
\end{definition}

By construction, $\velmancan$ consists of all canonical velocity fields. The composition $\projection=\canonicalmap\circ\terminalmap$ admits a simple two-step interpretation. Given an arbitrary velocity field $v\in\velspace$, the terminal-density map first extracts its terminal density $\rho^v=\terminalmap(v)$; the canonical-velocity map then reconstructs the corresponding canonical velocity field $v^{\rho^v}=\canonicalmap(\rho^v)$. Thus, once the identity $\terminalmap\circ\canonicalmap=\Id_{\pdfspace}$
is established, all velocity fields with the same terminal density are mapped by $\projection$ to the same canonical representative, without changing their terminal density. These relationships are summarized in Figure~\ref{fig:6}.

The definitions above describe the canonical construction at the set-theoretic level. The following theorem establishes its smooth geometric structure. In particular, it shows that $\velmancan$, which is diffeomorphic to $\pdfspace$, is a split embedded Banach submanifold of $\velspace$, whereas $\projection$ is a smooth idempotent projection that preserves terminal densities.

\begin{theorem}\label{thm:1}
    Under Assumption~\ref{assume:1}, the following conclusions hold.
    \begin{enumerate}
        \item The terminal-density map $\terminalmap:\velspace\to\pdfspace$ and the canonical-velocity map $\canonicalmap:\pdfspace\to\velspace$ are both smooth. Consequently, the canonical projection $\projection=\canonicalmap\circ\terminalmap:\velspace\to\velmancan$ is smooth.
        \item The canonical-velocity map $\canonicalmap:\pdfspace\to\velspace$ preserves the terminal density, i.e.,
        \[
        \terminalmap\circ\canonicalmap=\Id_\pdfspace.
        \]
        Consequently, $\terminalmap$ is surjective and $\canonicalmap$ is injective.
        \item The canonical manifold $\velmancan=\canonicalmap(\pdfspace)$ is a split embedded Banach submanifold of the velocity space $\velspace$, and
        \begin{equation*}
            \canonicalmap:\pdfspace\diffeomorphismto\velmancan
        \end{equation*}
        is a diffeomorphism with inverse
        \begin{equation*}
            \canonicalmap^{-1}=\terminalmap|_{\velmancan}.
        \end{equation*}
        \item The canonical projection $\projection:\velspace\to\velmancan$ is idempotent and preserves the terminal density, i.e.,
        \begin{equation*}
            \projection^2=\projection,\qquad\where{\projection}{\velmancan}=\mathrm{Id}_{\velmancan},\qquad \terminalmap\circ\projection=\terminalmap.
        \end{equation*}
    \end{enumerate}
\end{theorem}

\begin{proof}
    The detailed proof of this theorem is deferred to Appendix~\ref{app:1}. Under Assumption~\ref{assume:1},
    Definition~\ref{def:7} states that $\terminalmap:\velspace\to\pdfspace$ and $\canonicalmap:\pdfspace\to\velspace$ are smooth. In particular, Proposition~\ref{prop:22} proves that $\projection=\canonicalmap\circ\terminalmap$ is smooth, that $\velmancan=\canonicalmap(\pdfspace)$ is a split embedded Banach submanifold of $\velspace$, and that $\canonicalmap:\pdfspace\diffeomorphismto\velmancan$ is a diffeomorphism. Corollary~\ref{cor:5} proves that $\terminalmap\circ\canonicalmap=\Id_{\pdfspace}$, which further implies that $\terminalmap$ is surjective and $\canonicalmap$ is injective. Theorem~\ref{thm:10} proves that $\canonicalmap^{-1}=\where{\terminalmap}{\velmancan}$, $\projection^2=\projection$, $\where{\projection}{\velmancan}=\Id_{\velmancan}$, and $\terminalmap\circ\projection=\terminalmap$.
\end{proof}

Theorem~\ref{thm:1} provides two structural ingredients used in the remainder of the paper. First, the identity
\begin{equation*}
    \terminalmap\circ\projection=\terminalmap
\end{equation*}
allows an ambient velocity field to be canonicalized without changing its terminal density. This property underlies the canonical retraction introduced in Section~\ref{sec:2}.

Second, the diffeomorphism between the density manifold $\pdfspace$ and the canonical manifold $\velmancan$ allows us to equip the canonical optimization domain with a geometry that has a direct distributional interpretation. Since the optimization objective and the terminal effects of model updates are naturally expressed in terms of probability densities, Sections~\ref{subsec:5} and~\ref{subsec:6} first formulate the relevant first-order and second-order differential structures on $\pdfspace$, and then transport them through the diffeomorphism
\begin{equation*}
\canonicalmap:\pdfspace\diffeomorphismto\velmancan.
\end{equation*}
At the same time, the canonical representation is computationally natural, since training and model updates are carried out in velocity-field coordinates. Accordingly, Section~\ref{sec:2} defines the retraction directly on $\velmancan$, while the corresponding terminal-density update is obtained through the inverse diffeomorphism
\begin{equation*}
\canonicalmap^{-1}
=
\left.\terminalmap\right|_{\velmancan}
:
\velmancan\diffeomorphismto\pdfspace.
\end{equation*}
Thus, $\canonicalmap$ transports the selected density-space geometry to the canonical velocity representation, whereas $\canonicalmap^{-1}$ identifies the distributional effect of an update performed in velocity-field coordinates. These complementary roles are summarized in Figure~\ref{fig:7}.

\begin{figure}[t]
    \centering
    \input{figures/define_transport.tikz}
    \caption{Geometric structures and retraction operations under the canonical diffeomorphism. The tangent representation, Fisher--Rao metric, and mixture connection on $\pdfspace$ are transported to $\velmancan$ through $\canonicalmap:\pdfspace\diffeomorphismto\velmancan$. Conversely, the canonical retraction on $\velmancan$ corresponds through $\canonicalmap^{-1}=\where{\terminalmap}{\velmancan}:\velmancan\diffeomorphismto\pdfspace$ to the induced retraction on $\pdfspace$.}
    \label{fig:7}
\end{figure}

\subsection{First-Order Geometry}\label{subsec:5}

Having established the diffeomorphism $\canonicalmap:\pdfspace\diffeomorphismto\velmancan$, we develop the first-order structures needed to construct updates on the canonical manifold in this subsection. The canonical retraction introduced in Section~\ref{sec:2} takes a tangent direction at a canonical velocity field and converts it into a finite-stepsize velocity-field update. We therefore begin by characterizing tangent vectors on $\pdfspace$ and their canonical lifts to $\velmancan$ in Section~\ref{sub2sec:4}. An objective function, however, initially provides first-order information through its differential, which is a cotangent vector rather than an update direction. We identify the cotangent spaces of the two manifolds in Section~\ref{sub2sec:5}. Finally, we equip $\pdfspace$ with the Fisher--Rao metric, which provides a tangent-vector representation of a differential whenever its metric dual exists, and transport this metric to $\velmancan$. As summarized in Figure~\ref{fig:7}, all first-order geometric structures defined on the density manifold $\pdfspace$ are transported to the canonical manifold $\velmancan$.

\subsubsection{Tangent Spaces and Canonical Lifts}\label{sub2sec:4}

In the CFM framework, the trainable object is the velocity field, so an update should ultimately be represented as a velocity-field perturbation. The canonical-velocity update introduced in Section~\ref{sec:2} takes a tangent direction on $\velmancan$ as input. Accordingly, we characterize the relevant tangent spaces below.

Since the canonical-velocity map $\canonicalmap:\pdfspace\diffeomorphismto\velmancan$ is a diffeomorphism, it is sufficient to first characterize $T_\rho\pdfspace$. Every density tangent vector $\xi\in T_\rho\pdfspace$ has a unique canonical lift
\begin{equation}\label{eq:16}
    \dd_\rho\canonicalmap[\xi]\in T_{v^\rho}\velmancan.
\end{equation}
Since $\velmancan$ is an embedded submanifold of the Banach space $\velspace$, this canonical lift can be used directly as a velocity-field perturbation in Section~\ref{sec:2}.

We first characterize a tangent vector $\xi\in T_\rho\pdfspace$. Consider any smooth curve $(\rho^\varepsilon)_{\varepsilon\in(-\delta,\delta)}\subset\pdfspace$ which passes through $\rho$ at $\varepsilon=0$ and satisfies $\xi=\where{\frac{\dd}{\dd\varepsilon}\rho^\varepsilon}{\varepsilon=0}$. Since each $\rho^\varepsilon$ is normalized, we have
\[
\forall\varepsilon\in(-\delta,\delta),\,\int_{\R^d}\rho^\varepsilon(x)\,\dd x=1.
\]
As formalized in Theorem~\ref{thm:11}, we identify \(\xi\) with its density-function representation on \(\R^d\). Therefore,
\[
\int_{\R^d}\xi(x)\,\dd x=\where{\frac{\dd}{\dd\varepsilon}\int_{\R^d}\rho^\varepsilon(x)\,\dd x}{\varepsilon=0}=0.
\]
Writing $f=\frac{\xi}{\rho}+\const$, we obtain
\[
\xi=\xi^{\rho,f}:=\rho\cdot\left(f-\E_\rho[f]\right).
\]
Here, the term $f-\E_\rho[f]$ centers the variation of the density. More precisely, the function $f$ should belong to a regular function class. Evidently, for any constant $c\in\R$,
\[
\xi^{\rho,f+c}=\xi^{\rho,f}.
\]

\begin{proposition}\label{prop:1}
    Under Assumption~\ref{assume:3}, the tangent space of $\pdfspace$ at any density $\rho\in\pdfspace$ is a vector space satisfying
    \begin{equation*}
        T_\rho\pdfspace\subset\left\{\rho\cdot\left(f-\E_\rho[f]\right):f\in C(\R^d;\R),\E_\rho[\abs{f}]<\infty\right\}.
    \end{equation*}
    
\end{proposition}

\begin{proof}
    The proof is deferred to Theorem~\ref{thm:11} in Appendix~\ref{sub2sec:23}.
\end{proof}

The representation $\xi^{\rho,f}=\rho\cdot\left(f-\E_\rho[f]\right)$ describes an infinitesimal perturbation of the terminal density, but it is not yet a velocity-field update. We therefore transport it through $\dd_\rho\canonicalmap:T_\rho\pdfspace\to T_{v^\rho}\velmancan$ and define its canonical lift by~\eqref{eq:16}. To make this lifted direction $\dd_\rho\canonicalmap[\xi^{\rho,f}]$ usable in the flow-matching framework, we seek an explicit representation in terms of the posterior associated with the current canonical field. For this purpose, define the posterior value operator
\begin{equation}\label{eq:17}
    V_t^\rho[f]:\R^d\to\R,\qquad V_t^\rho[f](x_t):=\E_{X_1\sim p_{1|t}^{\rho}(\cdot|x_t)}\left[f(X_1)\right],
\end{equation}
where $f\in C(\R^d;\R)$ is called a terminal observable. At $t=1$, set $V_1^\rho[f](x_1):=f(x_1)$.

The following proposition shows that the canonical lift can be written equivalently as the gradient of a posterior value or as a posterior covariance. Both representations make its conditional-expectation structure explicit and yield a CFM-style learnable update in Section~\ref{subsec:14}.

\begin{proposition}\label{prop:2}
    Consider a terminal density $\rho\in\pdfspace$ and its canonical velocity field $v^\rho=\canonicalmap(\rho)\in\velmancan$. For any tangent vector $\xi^{\rho,f}=\rho\cdot(f-\E_{\rho}[f])\in T_\rho\pdfspace$, its canonical lift 
    \[\Gamma^{\rho,f}:=\dd_\rho\canonicalmap[\xi^{\rho,f}]\in T_{v^\rho}\velmancan\]
    can be represented, for every $t\in(0,1)$, by
    \begin{equation*}
        \Gamma^{\rho,f}_t(x_t)=\kappa_t\nabla V_t^\rho[f](x_t)=\Cov_{X_1\sim p_{1|t}^\rho(\cdot|x_t)}\left(v_{t|1}(x_t|X_1),f(X_1)\right).
    \end{equation*}
\end{proposition}

\begin{proof}
    First, we prove $\Gamma^{\rho,f}_t=\kappa_t\nabla V_t^\rho[f]$. Consider any smooth curve $(\rho^\varepsilon)_{\varepsilon\in[-\delta,\delta]}\subset\pdfspace$ which passes through $\rho$ at $\varepsilon=0$, where $\xi^{\rho,f}=\where{\frac{\dd}{\dd\varepsilon}\rho^\varepsilon}{\varepsilon=0}=\rho\cdot(f-\E_\rho[f])$. Note that
    \[
    \begin{aligned}
        V_t^{\rho}[f](x_t)
        =&\int_{\R^d}(f(x_1)-\E_\rho[f])p_{1|t}^{\rho}(x_1|x_t)\,\dd x_1+\E_\rho[f]\\
        =&\frac{\int_{\R^d}(f(x_1)-\E_\rho[f])\rho(x_1)p_{t|1}(x_t|x_1)\dd x_1}{p_{t}^{\rho}(x_t)}+\E_\rho[f]\\
        =&\frac{\int_{\R^d}\where{\frac{\dd}{\dd\varepsilon}\rho^\varepsilon(x_1)}{\varepsilon=0}p_{t|1}(x_t|x_1)\dd x_1}{p_{t}^{\rho}(x_t)}+\E_\rho[f]\\
        =&\frac{\where{\frac{\dd}{\dd\varepsilon}p_t^{\rho^\varepsilon}(x_t)}{\varepsilon=0}}{p_{t}^{\rho}(x_t)}+\E_\rho[f]\\
        =&\where{\frac{\dd}{\dd\varepsilon}\left(\log p_t^{\rho^\varepsilon}(x_t)\right)}{\varepsilon=0}+\E_\rho[f].
    \end{aligned}
    \]
    We also have
    \[
    \begin{aligned}
        \where{\frac{\dd}{\dd\varepsilon}v_t^{\rho^\varepsilon}(x_t)}{\varepsilon=0}
        =\where{\frac{\dd}{\dd\varepsilon}\left(\frac{\dot{\alpha}_t}{\alpha_t}x_t+\kappa_t\nabla_{x_t}\log p_t^{\rho^\varepsilon}(x_t)\right)}{\varepsilon=0}
        =\kappa_t\nabla_{x_t}\where{\frac{\dd}{\dd\varepsilon}\left(\log p_t^{\rho^\varepsilon}(x_t)\right)}{\varepsilon=0},
    \end{aligned}
    \]
    which gives $\Gamma^{\rho,f}_t=\kappa_t\nabla V_t^\rho[f]$.

    Next, we compute $\nabla V_t^\rho[f]$. By~\eqref{eq:11}, we have $\nabla_{x_t}p_{t|1}(x_t|x_1)=-\frac{x_t-\alpha_tx_1}{\beta_t^2}p_{t|1}(x_t|x_1)$. In particular,
    \[
    \begin{aligned}
        \nabla V_t^\rho[f](x_t)
        =&\nabla_{x_t}\left(\frac{\int_{\R^d}f(x_1)\rho(x_1)p_{t|1}(x_t|x_1)\dd x_1}{\int_{\R^d}\rho(x_1)p_{t|1}(x_t|x_1)\dd x_1}\right)\\
        =&\frac{\int_{\R^d}f(x_1)\rho(x_1)\nabla_{x_t}p_{t|1}(x_t|x_1)\dd x_1}{p_t^\rho(x_t)}-V_t^\rho[f](x_t)\frac{\int_{\R^d}\rho(x_1)\nabla_{x_t}p_{t|1}(x_t|x_1)\dd x_1}{p_t^\rho(x_t)}\\
        =&\int_{\R^d}\left(f(x_1)-V_t^\rho[f](x_t)\right)\frac{\rho(x_1)\nabla_{x_t}p_{t|1}(x_t|x_1)}{p_t^\rho(x_t)}\dd x_1\\
        =&\int_{\R^d}\left(f(x_1)-V_t^\rho[f](x_t)\right)\frac{\rho(x_1)p_{t|1}(x_t|x_1)}{p_t^\rho(x_t)}\frac{\alpha_tx_1-x_t}{\beta_t^2}\dd x_1\\
        =&\E_{X_1\sim p_{1|t}^\rho(\cdot|x_t)}\left[\left(f(X_1)-V_t^\rho[f](x_t)\right)\frac{\alpha_tX_1-x_t}{\beta_t^2}\right]\\
        =&\frac{\alpha_t}{\beta_t^2}\Cov_{X_1\sim p_{1|t}^\rho(\cdot|x_t)}\left(X_1,f(X_1)\right)
    \end{aligned}
    \]
    implies $\kappa_t\nabla V_t^\rho[f](x_t)=\Cov_{X_1\sim p_{1|t}^\rho(\cdot|x_t)}\left(v_{t|1}(x_t|X_1),f(X_1)\right)$.
\end{proof}

Based on Proposition~\ref{prop:2}, we obtain the tangent space of $\velmancan$ as follows.

\begin{corollary}\label{cor:1}
    The tangent space of $\velmancan$ at $v^\rho=\canonicalmap(\rho)\in\velmancan$ is
    \[
    T_{v^\rho}\velmancan=\left\{\Gamma^{\rho,f}:\xi^{\rho,f}\in T_\rho\pdfspace\right\}\cong T_\rho\pdfspace.
    \]
    Moreover, for any density $\rho\in\pdfspace$, regular terminal observables $f,g\in C(\R^d;\R)$, and constants $a,b,c\in\R$ such that $\xi^{\rho,f},\xi^{\rho,g}\in T_\rho\pdfspace$, we have
    \[
    \Gamma^{\rho,af+bg+c}=a\Gamma^{\rho,f}+b\Gamma^{\rho,g}.
    \]
\end{corollary}

\begin{proof}
    By Theorem~\ref{thm:1}, the canonical-velocity map $\canonicalmap:\pdfspace\diffeomorphismto\velmancan$ is a diffeomorphism. Therefore, $\dd_\rho\canonicalmap: T_\rho\pdfspace\diffeomorphismto T_{v^\rho}\velmancan$ is a linear isomorphism; hence,
    \[
    T_{v^\rho}\velmancan=\left\{\dd_\rho\canonicalmap[\xi]:\xi\in T_\rho\pdfspace\right\}=\left\{\Gamma^{\rho,f}:\xi^{\rho,f}\in T_\rho\pdfspace\right\}.
    \]
    Moreover,
    \[
    \Gamma^{\rho,af+bg+c}_t=\kappa_t\nabla V_t^\rho[af+bg+c]=a\kappa_t\nabla V_t^\rho[f]+b\kappa_t\nabla V_t^\rho[g]=a\Gamma^{\rho,f}_t+b\Gamma^{\rho,g}_t,
    \]
    which finishes the proof.
\end{proof}

We use the $\Gamma$-representation as the basic learnable update direction throughout the remainder of the paper. The next result establishes the interior-time identifiability of this representation. At any nondegenerate time slice, the lifted field determines the terminal observable up to an additive constant; equivalently, vanishing at a single interior time already forces the observable to be constant.

\begin{proposition}\label{prop:3}
    For a terminal density $\rho\in\pdfspace$ and a terminal observable $f\in C(\R^d;\R)$, consider the tangent vector $\Gamma^{\rho,f}$. The following three statements are equivalent:
    \[
    f\equiv\const\quad\Leftrightarrow\quad
    \forall t\in(0,1),\,V_t^\rho[f]\equiv\const\quad\Leftrightarrow\quad\exists t_0\in(0,1),\,V_{t_0}^\rho[f]\equiv\const.
    \]
    In other words, we have
    \[
    f\equiv\const\quad\Leftrightarrow\quad
    \forall t\in(0,1),\,\Gamma_t^{\rho,f}\equiv0\quad\Leftrightarrow\quad\exists t_0\in(0,1),\,\Gamma_{t_0}^{\rho,f}\equiv0.
    \]
\end{proposition}

\begin{proof}
    If $f\equiv c$ is constant, then $V_t^\rho[f](x_t)=\E_{X\sim p_{1|t}^\rho(\cdot|x_t)}[c]=c$ is constant for all $t\in(0,1)$, which is equivalent to $\Gamma_t^{\rho,f}\equiv0$ for all $t\in(0,1)$.

    If $V_t^\rho[f]\equiv c$ is constant for all $t\in(0,1)$, let $t_0\in(0,1)$ be arbitrary. Then, $V_{t_0}^\rho[f]\equiv c$ is constant, which is equivalent to $\Gamma_{t_0}^{\rho,f}\equiv0$.

    If $V_{t_0}^\rho[f]\equiv c$ is constant for some $t_0\in(0,1)$, then we have
    \[
        \int_{\R^d}(f(x_1)-c)\rho(x_1)p_{t_0|1}(x_{t_0}|x_1)\,\dd x_1=0
    \]
    for every $x_{t_0}\in\R^d$. Since the Gaussian convolution is injective and $f$, $\rho$ are continuous, we have $(f(x_1)-c)\rho(x_1)\equiv0$. Therefore, the strict positivity of $\rho$ implies $f\equiv\const$.
\end{proof}

Equivalently, for every fixed $t_0\in(0,1)$, the map $f\mapsto\Gamma_{t_0}^{\rho,f}$ is injective modulo additive constants. Thus, the $\Gamma$-representation is nondegenerate: a nonconstant terminal observable satisfies $\Gamma_t^{\rho,f}\not\equiv0$ at every interior time $t\in(0,1)$.

\subsubsection{Cotangent Spaces and Differentials}\label{sub2sec:5}

A tangent space describes the admissible infinitesimal perturbations of a density. To describe the first-order change of an objective along such perturbations, we need a continuous linear functional acting on
the tangent space. This motivates the cotangent space
\begin{equation*}
    T_\rho^*\pdfspace
    =
    \left\{
        \ell:T_\rho\pdfspace\to\R:
        \ell\text{ is continuous and linear}
    \right\}.
\end{equation*}
Its elements are called cotangent vectors. Similarly, for any $v\in\velmancan$, the cotangent space of $\velmancan$ at $v$ is
\[
T_v^*\velmancan=\left\{\ell^\mathrm{can}:T_v\velmancan\to\R:
        \ell^\mathrm{can}\text{ is continuous and linear}\right\}.
\]
These two cotangent spaces are linearly isomorphic when $v=\canonicalmap(\rho)$:
\[
T_{\canonicalmap(\rho)}^*\velmancan=\left\{\ell\circ\dd_{\canonicalmap(\rho)}\canonicalmap^{-1}:\ell\in T_\rho^*\pdfspace\right\}\cong T_\rho^*\pdfspace.
\]

As recalled in Section~\ref{sub2sec:1}, the differential of a smooth objective $J:\pdfspace\to\R$ at $\rho$ is the cotangent vector $\dd_\rho J\in T_\rho^*\pdfspace$ defined as
\begin{equation*}
    \dd_\rho J:T_\rho\pdfspace\to\R,\qquad \dd_\rho J[\xi]
    :=
    \where{
        \frac{\dd}{\dd\varepsilon}
        J(\rho^\varepsilon)
    }{\varepsilon=0},
\end{equation*}
where $(\rho^\varepsilon)_\varepsilon\subset\pdfspace$ is any smooth curve which passes through $\rho$ at $\varepsilon=0$ and satisfies $\where{\frac{\dd}{\dd\varepsilon}\rho^\varepsilon}{\varepsilon=0}=\xi$. Thus, $\dd_\rho J$ assigns to every admissible density perturbation its corresponding first-order change in the objective.

\subsubsection{Fisher--Rao Metric}\label{sub2sec:6}

A cotangent vector records the first-order variation of an objective, whereas the canonical retraction in Section~\ref{sec:2} requires a tangent direction as the first-order perturbation direction. A Riemannian metric provides the link between these two objects. Recall from Section~\ref{sub2sec:2} that a Riemannian metric determines a unique tangent-vector representation of a cotangent vector whenever such a representation exists. The choice of metric is therefore part of the optimization method: different metrics generally produce different tangent-vector representations of the same differential and hence different gradient directions.

To identify objective differentials with tangent vectors, we equip $\pdfspace$ with the \textit{Fisher--Rao metric} \citep{ay2017information} and transport this metric through the diffeomorphism $\canonicalmap$ to the canonical manifold $\velmancan$. The Fisher--Rao metric is well suited to a density manifold since it measures density perturbations through their relative changes $\frac{\xi}{\rho}$:
\begin{equation}\label{eq:18}
    g_\rho^\fisher:T_\rho\pdfspace\times T_\rho\pdfspace\to\R,\qquad g_\rho^\fisher(\xi_1,\xi_2):=\int_{\R^d}\frac{\xi_1(x)\xi_2(x)}{\rho(x)}\,\dd x.
\end{equation}
In particular, for any two terminal observables $f_1,f_2$, we have
\[
g_\rho^\fisher(\xi^{\rho,f_1},\xi^{\rho,f_2})=\E_{\rho}\left[(f_1-\E_\rho[f_1])(f_2-\E_\rho[f_2])\right].
\]
Under Assumption~\ref{assume:4}, the map $g^\fisher:(\rho,\xi_1,\xi_2)\mapsto g_\rho^\fisher(\xi_1,\xi_2)$ defines a smooth, possibly weak, Riemannian metric on the Banach manifold \(\pdfspace\). In other words, \(\pdfspace\) is a Riemannian manifold.

Since the Fisher--Rao metric may be weak, a cotangent vector $\ell\in T_\rho^*\pdfspace$ need not admit a metric-dual tangent representation. Section~\ref{subsec:9} verifies such a representation explicitly for the reverse-KL differential.

Consider any $v\in\velmancan$ and denote $\rho^v=\terminalmap(v)\in\pdfspace$. For any $\Gamma_1,\Gamma_2\in T_{v}\velmancan$, the transported metric is $g_{v}^{\mathrm{can}}:T_{v}\velmancan\times T_{v}\velmancan\to\R$,
\begin{align}
    g_{v}^{\mathrm{can}}\left(\Gamma_1,\Gamma_2\right)
    :=&g_{\rho^v}^\fisher\left(\dd_v\canonicalmap^{-1}\left[\Gamma_1\right],\dd_v\canonicalmap^{-1}\left[\Gamma_2\right]\right).\label{eq:19}
\end{align}
For any two terminal observables $f_1$, $f_2$, we have
\[
g_{v}^{\mathrm{can}}\left(\Gamma^{\rho^v,f_1},\Gamma^{\rho^v,f_2}\right)=\E_{\rho^v}\left[(f_1-\E_{\rho^v}[f_1])(f_2-\E_{\rho^v}[f_2])\right].
\]
By construction, $\canonicalmap$ is an isometry between $(\pdfspace,g^\fisher)$ and $(\velmancan,g^{\mathrm{can}})$. Thus, $\velmancan$ inherits the same Riemannian, possibly weak, structure as $\pdfspace$.

\subsection{Second-Order Geometry}\label{subsec:6}

The first-order structures in Section~\ref{subsec:5} are defined separately at each base point. To differentiate first-order objects as the density varies, and thereby define the Hessian used in Section~\ref{subsec:10}, we require an affine connection. The connection is part of the geometric specification of the optimization method: even for the same objective, different connections generally yield different Hessians and may therefore lead to different Newton directions.

On the density manifold, we compare first-order objects at neighboring densities by holding the terminal observable fixed as the base density varies. An admissible terminal observable thereby induces a tangent vector field across densities. Although the observable remains unchanged, the induced tangent vector and the Fisher--Rao metric both vary with the base density. The mixture connection captures the resulting first-order variation of the Fisher--Rao pairing. We adopt this connection on $\pdfspace$ and transport it to $\velmancan$ through the diffeomorphism $\canonicalmap$.

To make this characterization precise, consider a terminal observable $f$ for which
\[
\xi^{\cdot,f}\in\mathfrak{X}(\pdfspace),
\qquad
\xi^{q,f}:=q\cdot\left(f-\E_q[f]\right),
\qquad
q\in\pdfspace,
\]
defines a smooth tangent vector field on $\pdfspace$. We call
$\xi^{\cdot,f}$ the \textit{fixed-observable extension} of
$\xi^{\rho,f}\in T_\rho\pdfspace$. Along this extension, the terminal observable $f$ is held fixed, whereas its density-function representation varies with the base density through both $q$ and $\E_q[f]$. Under the smooth-realization assumptions of Appendix~\ref{subsec:34}, the values of admissible fixed-observable extensions exhaust each tangent space \(T_\rho\pdfspace\).

For any $\DensityTangentFieldSecond\in\mathfrak{X}(\pdfspace)$, define the real-valued function
\[
g^\fisher\left(\DensityTangentFieldSecond,\xi^{\cdot,f}\right):
\pdfspace\to\R,
\qquad 
q\mapsto
g_q^\fisher\left(\DensityTangentFieldSecond_q,\xi^{q,f}\right).
\]
The mixture connection records the first-order change of this pairing
when the base density varies. Specifically, for
$\DensityTangentFieldFirst,\DensityTangentFieldSecond\in\mathfrak{X}(\pdfspace)$, we define
$\left(\nabla_{\DensityTangentFieldFirst}^\mathrm{mix}\DensityTangentFieldSecond\right)_\rho\in T_\rho\pdfspace$ by the tangent vector satisfying
\begin{equation}\label{eq:20}
\begin{aligned}
    g_\rho^\fisher
    \left(
        \left(\nabla_{\DensityTangentFieldFirst}^\mathrm{mix}\DensityTangentFieldSecond\right)_\rho,
        \xi^{\rho,f}
    \right)
    &=
    \DensityTangentFieldFirst\left[
        g^\fisher\left(\DensityTangentFieldSecond,\xi^{\cdot,f}\right)
    \right](\rho)
    =
    \dd_\rho
    \left(
        g^\fisher\left(\DensityTangentFieldSecond,\xi^{\cdot,f}\right)
    \right)
    [\DensityTangentFieldFirst_\rho]
\end{aligned}
\end{equation}
for every admissible terminal observable $f$. Equivalently, for any smooth curve
$(\rho^\varepsilon)_{\varepsilon\in(-\delta,\delta)}\subset\pdfspace$
which passes through $\rho$ at $\varepsilon=0$ and satisfies $\where{\frac{\dd}{\dd\varepsilon}\rho^\varepsilon}{\varepsilon=0}=\DensityTangentFieldFirst_\rho$,
\[
g_\rho^\fisher
\left(
    \left(\nabla_{\DensityTangentFieldFirst}^\mathrm{mix}\DensityTangentFieldSecond\right)_\rho,
    \xi^{\rho,f}
\right)
=
\where{
    \frac{\dd}{\dd\varepsilon}
    g_{\rho^\varepsilon}^\fisher
    \left(
        \DensityTangentFieldSecond_{\rho^\varepsilon},
        \xi^{\rho^\varepsilon,f}
    \right)
}{\varepsilon=0}.
\]
Here both the tangent vector field
$\DensityTangentFieldSecond_{\rho^\varepsilon}$ and the base density $\rho^\varepsilon$ vary, while $f$ is fixed.
In particular, the variation of $\rho^\varepsilon$ changes both the
Fisher--Rao pairing and the centered density representation
$\xi^{\rho^\varepsilon,f}$. The proof of~\eqref{eq:20} is deferred to Theorem~\ref{thm:12}.

Under Assumption~\ref{assume:5}, the defining relation~\eqref{eq:20} uniquely determines $\left(\nabla^{\mathrm{mix}}_{\DensityTangentFieldFirst}\DensityTangentFieldSecond\right)_\rho \in T_\rho\pdfspace$ for every $(\DensityTangentFieldFirst,\DensityTangentFieldSecond,\rho)$, and the resulting assignment is smooth. Therefore,
\[
\nabla^\mathrm{mix}:
\mathfrak{X}(\pdfspace)
\times
\mathfrak{X}(\pdfspace)
\to
\mathfrak{X}(\pdfspace),
\qquad
(\DensityTangentFieldFirst,\DensityTangentFieldSecond)\mapsto\nabla_{\DensityTangentFieldFirst}^\mathrm{mix}\DensityTangentFieldSecond,
\]
is well-defined. Its construction, together with the proofs of uniqueness and smoothness, is deferred to Appendix~\ref{sub2sec:24}. The following proposition verifies the connection axioms~\eqref{eq:6}.

\begin{proposition}
    Under Assumption~\ref{assume:5}, the operator $\nabla^{\mathrm{mix}}$ in~\eqref{eq:20} satisfies the definition of a connection~\eqref{eq:6}. Consequently, $\nabla^\mathrm{mix}$ is an affine connection on $\pdfspace$.
\end{proposition}

\begin{proof}
    Under Assumption~\ref{assume:5}, $\nabla^\mathrm{mix}$ is well-defined and smooth. Consider any tangent vector fields $\DensityTangentFieldFirst,\DensityTangentFieldSecond,Z\in\mathfrak{X}(\pdfspace)$, smooth scalar functions $h_1,h_2:\pdfspace\to\R$, and constants $a,b\in\R$. First, for every $\xi^{\rho,f}\in T_\rho\pdfspace$,
    \begin{align*}
    g_\rho^\fisher\left((\nabla^{\mathrm{mix}}_{h_1\DensityTangentFieldFirst+h_2\DensityTangentFieldSecond}Z)_\rho,\xi^{\rho,f}\right)
    =&\dd_{\rho}\left(g^\fisher\left(Z,\xi^{\cdot,f}\right)\right)\left[h_1(\rho)\DensityTangentFieldFirst_\rho+h_2(\rho)\DensityTangentFieldSecond_\rho\right]\\
    =&h_1(\rho)\dd_{\rho}\left(g^\fisher\left(Z,\xi^{\cdot,f}\right)\right)\left[\DensityTangentFieldFirst_\rho\right]+h_2(\rho)\dd_{\rho}\left(g^\fisher\left(Z,\xi^{\cdot,f}\right)\right)\left[\DensityTangentFieldSecond_\rho\right]\\
    =&g_\rho^\fisher\left((h_1\nabla^{\mathrm{mix}}_{\DensityTangentFieldFirst}Z+h_2\nabla^{\mathrm{mix}}_{\DensityTangentFieldSecond}Z)_\rho,\xi^{\rho,f}\right).
    \end{align*}
    The positive definiteness of $g_\rho^\fisher$ gives~\eqref{eq:7}. Then, for every $\xi^{\rho,f}\in T_\rho\pdfspace$,
    \begin{align*}
    g_\rho^\fisher\left((\nabla^{\mathrm{mix}}_{\DensityTangentFieldFirst}(a\DensityTangentFieldSecond+bZ))_\rho,\xi^{\rho,f}\right)
    =&\dd_{\rho}\left(g^\fisher\left(a\DensityTangentFieldSecond+bZ,\xi^{\cdot,f}\right)\right)\left[\DensityTangentFieldFirst_\rho\right]\\
    =&a\dd_{\rho}\left(g^\fisher\left(\DensityTangentFieldSecond,\xi^{\cdot,f}\right)\right)\left[\DensityTangentFieldFirst_\rho\right]+b\dd_{\rho}\left(g^\fisher\left(Z,\xi^{\cdot,f}\right)\right)\left[\DensityTangentFieldFirst_\rho\right]\\
    =&g_\rho^\fisher\left((a\nabla^{\mathrm{mix}}_{\DensityTangentFieldFirst}\DensityTangentFieldSecond+b\nabla^{\mathrm{mix}}_{\DensityTangentFieldFirst}Z)_\rho,\xi^{\rho,f}\right);
    \end{align*}
    hence, we have~\eqref{eq:8}. Finally,~\eqref{eq:9} is implied by
    \begin{align*}
    g_\rho^\fisher\left((\nabla^{\mathrm{mix}}_{\DensityTangentFieldFirst}(h_1\DensityTangentFieldSecond))_\rho,\xi^{\rho,f}\right)
    =&\where{\frac{\dd}{\dd\varepsilon}g^\fisher_{\rho^\varepsilon}\left(h_1(\rho^\varepsilon)\DensityTangentFieldSecond_{\rho^\varepsilon},\xi^{\rho^\varepsilon,f}\right)}{\varepsilon=0}\\
    =&h_1(\rho)\where{\frac{\dd}{\dd\varepsilon}g^\fisher_{\rho^\varepsilon}\left(\DensityTangentFieldSecond_{\rho^\varepsilon},\xi^{\rho^\varepsilon,f}\right)}{\varepsilon=0}+\where{\frac{\dd}{\dd\varepsilon}h_1(\rho^\varepsilon)}{\varepsilon=0}g^\fisher_{\rho}\left(\DensityTangentFieldSecond_{\rho},\xi^{\rho,f}\right)\\
    =&h_1(\rho)g_\rho^\fisher\left((\nabla^{\mathrm{mix}}_{\DensityTangentFieldFirst}\DensityTangentFieldSecond)_\rho,\xi^{\rho,f}\right)+\DensityTangentFieldFirst[h_1](\rho)g^\fisher_{\rho}\left(\DensityTangentFieldSecond_{\rho},\xi^{\rho,f}\right)\\
    =&g_\rho^\fisher\left(h_1(\rho)(\nabla^{\mathrm{mix}}_{\DensityTangentFieldFirst}\DensityTangentFieldSecond)_\rho+\DensityTangentFieldFirst[h_1](\rho)\DensityTangentFieldSecond_{\rho},\xi^{\rho,f}\right).
    \end{align*}
    Therefore, $\nabla^\mathrm{mix}$ is an affine connection on $\pdfspace$.
\end{proof}

We also transport the mixture connection $\nabla^{\mathrm{mix}}$ to $\velmancan$. For every $\widetilde{\DensityTangentFieldFirst},\widetilde{\DensityTangentFieldSecond}\in\mathfrak{X}(\velmancan)$, define the tangent vector fields $\DensityTangentFieldFirst,\DensityTangentFieldSecond\in\mathfrak{X}(\pdfspace)$ by
\[
\DensityTangentFieldFirst_\rho=\dd_{\canonicalmap(\rho)}\canonicalmap^{-1}\left[\widetilde{\DensityTangentFieldFirst}_{\canonicalmap(\rho)}\right],\qquad \DensityTangentFieldSecond_\rho=\dd_{\canonicalmap(\rho)}\canonicalmap^{-1}\left[\widetilde{\DensityTangentFieldSecond}_{\canonicalmap(\rho)}\right].
\]
Then, we define the transported mixture connection $\nabla^\mathrm{can}$ on $\velmancan$ by
\[
\nabla^\mathrm{can}_{\widetilde{\DensityTangentFieldFirst}}\widetilde{\DensityTangentFieldSecond}\in\mathfrak{X}(\velmancan),\qquad \left(\nabla^\mathrm{can}_{\widetilde{\DensityTangentFieldFirst}}\widetilde{\DensityTangentFieldSecond}\right)_{\canonicalmap(\rho)}:=\dd_\rho\canonicalmap\left[\left(\nabla^{\mathrm{mix}}_{\DensityTangentFieldFirst}\DensityTangentFieldSecond\right)_\rho\right].
\]

\section{Canonical Retraction and Value Ascent}\label{sec:2}

As discussed in Section~\ref{sec:1}, the diffeomorphism \(\canonicalmap:\pdfspace\diffeomorphismto\velmancan\) allows the geometric structures defined on the density manifold $\pdfspace$ to be transported to the canonical manifold $\velmancan$. At the same time, $\velmancan$ is the natural computational domain, since training is carried out in velocity-field coordinates. Accordingly, this section constructs finite-stepsize updates directly on $\velmancan$. As summarized in Figure~\ref{fig:7}, through the inverse diffeomorphism \(\canonicalmap^{-1}:\velmancan\diffeomorphismto\pdfspace\), the resulting canonical retraction admits an equivalent density-coordinate representation on $\pdfspace$.

The purpose of this section is to construct an objective-agnostic finite-stepsize update rule on the canonical manifold $\velmancan$ that is compatible with the conditional-expectation structure underlying standard diffusion and flow training. At this stage, we do not yet specify whether the tangent direction is a gradient, a Newton direction, or another search direction. Instead, given a canonical velocity field \(v\in\velmancan\) and a tangent vector \(\Gamma\in T_v\velmancan\), we construct a finite-stepsize canonical-velocity update that returns another canonical velocity field \(v^+\in\velmancan\) using tractable operations in the velocity space $\velspace$. Beyond establishing geometric feasibility, we also seek to understand how this update changes the entire induced probability path and its terminal density. In general, this finite-stepsize effect is nonlinear and nonlocal, admitting only an implicit characterization through the continuity equation. However, by exploiting the posterior-value-gradient structure of canonical tangent directions, we obtain an exact characterization. Section~\ref{sec:3} instantiates this general objective-agnostic update rule with the reverse-KL Newton direction and thereby turns it into an optimization algorithm.

Section~\ref{subsec:7} constructs the canonical retraction through a tangential update followed by canonicalization, and proves that this update induces a retraction on the density manifold. Together, these retractions provide velocity-coordinate and density-coordinate representations of the same finite-stepsize update rule. Section~\ref{subsec:8} then establishes exact trajectory-wise and density-level identities that characterize the finite-stepsize effect on the induced probability path. These identities establish posterior-value improvement at every intermediate time and, at the terminal time, yield an explicit formula for the updated density together with a finite-stepsize value-ascent certificate.

\subsection{Canonical Retractions in Velocity and Density Coordinates}\label{subsec:7}

Fix a terminal density \(\rho\in\pdfspace\), its canonical velocity field \(v^\rho=\canonicalmap(\rho)\in\velmancan\), and a tangent vector \(\Gamma^{\rho,f}\in T_{v^\rho}\velmancan\). A tangent vector represents an infinitesimal first-order perturbation of the canonical velocity field, whereas a finite-stepsize update is required in practice. Since \(\velmancan\) is an embedded submanifold of the Banach space \(\velspace\), the tangent vector \(\Gamma^{\rho,f}\) can also be identified with a velocity-field displacement in \(\velspace\). As shown in Proposition~\ref{prop:2}, \(\Gamma^{\rho,f}:[0,1]\times\R^d\to\R^d\) admits the representation
\[
\Gamma_t^{\rho,f}(x_t)=\kappa_t\nabla V_t^\rho[f](x_t)=\Cov_{X_1\sim p_{1|t}^\rho(\cdot|x_t)}\left(v_{t|1}(x_t|X_1),f(X_1)\right).
\]
We may therefore form the finite-stepsize velocity-space update
\begin{equation}\label{eq:21}
    \bar v:=v^\rho+\Gamma^{\rho,f}\in\velspace,
\end{equation}
which is called the \textit{tangential update}.

The tangential update~\eqref{eq:21} is computationally useful since it preserves the conditional-expectation structure required by CFM. More precisely, the updated velocity field
\[
\bar{v}_t(x_t)=\E_{X_1\sim p_{1|t}^{\rho}(\cdot| x_t)}\left[v_{t|1}(x_t|X_1)+f(X_1)\left(v_{t|1}(x_t|X_1)-v_t^\rho(x_t)\right)\right]
\]
can be learned by shifting the sample-wise conditional regression target. Section~\ref{sec:5} generalizes this target and develops population-exact implementations of the resulting update.

Although the updated velocity field \(\bar{v}\) is a valid element of the velocity space \(\velspace\), it does not necessarily lie on the canonical manifold \(\velmancan\). The fact that \(\Gamma^{\rho,f}\in T_{v^\rho}\velmancan\) is tangent at the base point \(v^\rho\in\velmancan\) is only a first-order property; it does not imply that the obtained \(\bar{v}\) remains on the canonical manifold. Since \(\velmancan\) is generally non-affine,
\[
\bar v=v^\rho+\Gamma^{\rho,f}\notin\velmancan\text{ in general}.
\]
Appendix~\ref{subsec:35} gives an explicit Gaussian example for which this failure occurs.
Consequently, to continue the update loop, we need a new canonical velocity field. Although \(\bar v\) is a valid velocity field with a well-defined terminal density \(\terminalmap(\bar v)\in\pdfspace\), it is generally noncanonical and therefore cannot serve directly as the next iterate.

A retraction is the standard geometric device for converting such a tangent displacement into a feasible point on the manifold. At a base point \(v^\rho\in\velmancan\), a retraction is a smooth map 
\[
\retraction_{v^\rho}:T_{v^\rho}\velmancan\to\velmancan
\]
satisfying
\[
\retraction_{v^\rho}(0)=v^\rho,\qquad \dd_0\retraction_{v^\rho}=\Id_{T_{v^\rho}\velmancan}.
\]
Intuitively, the tangent vector specifies both an initial direction and a displacement magnitude, whereas the retraction specifies how that displacement is realized as a finite-stepsize update on the manifold. The first condition $\retraction_{v^\rho}(0)=v^\rho$ requires that a zero displacement leaves the current point unchanged. The second condition $\dd_0\retraction_{v^\rho}=\Id_{T_{v^\rho}\velmancan}$ indicates that the retraction preserves every prescribed tangent direction to first order. Equivalently, for every \(\Gamma\in T_{v^\rho}\velmancan\),
\[
\where{\frac{\dd}{\dd\varepsilon}\retraction_{v^\rho}(\varepsilon\Gamma)}{\varepsilon=0}=\Gamma.
\]
Thus, the curve \(\varepsilon\mapsto\retraction_{v^\rho}(\varepsilon\Gamma)\) starts at \({v^\rho}\) with initial velocity \(\Gamma\), even though its finite-stepsize trajectory need not coincide with the straight line \({v^\rho}+\varepsilon\Gamma\) in the velocity space $\velspace$.

When the exponential map of a selected affine connection is locally well-defined, it provides a natural geometric example of a retraction, i.e., one follows the geodesic starting from \({v^\rho}\) with initial velocity \(\Gamma\). On the canonical manifold, however, evaluating such an update would require solving a nonlinear infinite-dimensional geodesic equation. This does not expose the conditional-expectation structure needed for scalable regression and is generally computationally intractable.

\begin{figure}[t]
    \centering
    \input{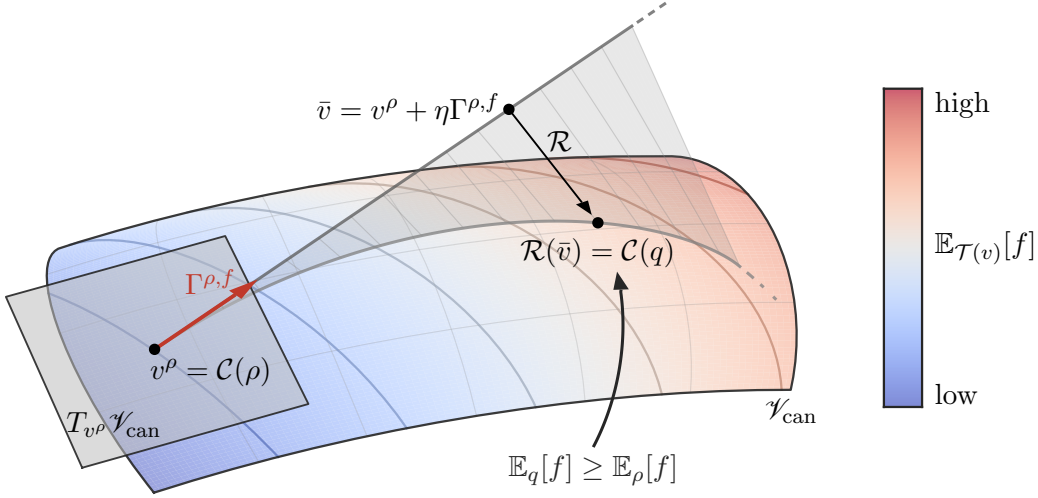}
    \caption{Canonical retraction and its value-ascent certificate. Starting from the canonical velocity field \(v^\rho=\canonicalmap(\rho)\), the finite-stepsize tangential update \(\bar v=v^\rho+\eta\Gamma^{\rho,f}\) generally leaves the canonical manifold $\velmancan$. Then, \(\bar v\) is projected back to the canonical manifold as $\projection(\bar{v})$ while preserving the terminal density \(q=\terminalmap(\bar{v})\). The induced update satisfies the value-ascent certificate \(\E_{X\sim q}[f(X)]\geq\E_{X\sim\rho}[f(X)]\).}
    \label{fig:8}
\end{figure}

We instead exploit two structures already available from Section~\ref{sec:1}, namely, the linear structure of \(\velspace\) and the canonical projection \(\projection:\velspace\to\velmancan\). Starting from \(v^\rho\in\velmancan\), we first perform the tangential update, obtaining the ambient field \(\bar v=v^\rho+\Gamma\), and then project the resulting field back onto the canonical manifold:
\[
v^+=\projection(\bar v)=\canonicalmap(\terminalmap(\bar v))\in\velmancan.
\]
We refer to this projection step as \textit{canonicalization}. Intuitively, the terminal-density map $\terminalmap$ first forgets the particular velocity representation and retains only its terminal density. The canonical-velocity map $\canonicalmap$ then reconstructs the canonical representative associated with the terminal density. Since \(\terminalmap\circ\projection=\terminalmap\), $v^+$ and $\bar{v}$ share the same terminal density $q=\terminalmap(\bar{v})=\terminalmap(v^+)$. Motivated by this two-stage construction, define the map
\begin{equation}\label{eq:22}
    \retraction_{v^\rho}^\mathrm{can}:T_{v^\rho}\velmancan\to\velmancan,\qquad \Gamma\mapsto\projection(v^\rho+\Gamma).
\end{equation}
This two-stage construction is illustrated in Figure~\ref{fig:8}.

The following proposition states that the map defined in~\eqref{eq:22} is a valid retraction.

\begin{proposition}\label{prop:4}
    Fix the terminal density $\rho\in\pdfspace$ and its canonical velocity field $v^\rho=\canonicalmap(\rho)\in\velmancan$. The map~\eqref{eq:22} is well-defined on $T_{v^\rho}\velmancan$ and smooth. Moreover, we have
    \[
        \retraction_{v^\rho}^\mathrm{can}(0)={v^\rho},\qquad \dd_0\retraction_{v^\rho}^\mathrm{can}=\Id_{T_{v^\rho}\velmancan}.
    \]
    Therefore, the map~\eqref{eq:22} is a retraction, which is called the canonical retraction.
\end{proposition}

\begin{proof}
    We first prove that the map~\eqref{eq:22} is well-defined on $T_{v^\rho}\velmancan$. According to Definition~\ref{def:1}, $\velspace$ is a Banach space. Since $\velmancan$ is a split embedded manifold of the Banach space $\velspace$, all tangent vectors in $T_{v^\rho}\velmancan$ can be identified as elements in $\velspace$. Therefore, for every $\Gamma\in T_{v^\rho}\velmancan$, ${v^\rho}+\Gamma\in\velspace$; hence, $\projection({v^\rho}+\Gamma)$ is well-defined.%

    According to Theorem~\ref{thm:1}, the canonical projection $\projection:\velspace\to\velmancan$ is smooth and satisfies $\where{\projection}{\velmancan}=\mathrm{Id}_{\velmancan}$. Therefore, $\retraction_{v^\rho}^\mathrm{can}$ is smooth and satisfies $\retraction_{v^\rho}^\mathrm{can}(0)=\projection({v^\rho}+0)={v^\rho}$.

    Finally, we consider $\dd_0\retraction_{v^\rho}^\mathrm{can}:T_{v^\rho}\velmancan\to T_{v^\rho}\velmancan$. To evaluate this differential, consider any path $(v^\varepsilon)_{\varepsilon\in(-\delta,\delta)}\subset\velmancan$ which passes through ${v^\rho}$ at $\varepsilon=0$. Then, we have $\projection(v^\varepsilon)\equiv v^\varepsilon$. For $\Gamma=\where{\frac{\dd}{\dd \varepsilon}v^\varepsilon}{\varepsilon=0}\in T_{v^\rho}\velmancan$, differentiating at $\varepsilon=0$ gives
    \[
    \dd_{v^\rho}\projection: T_{v^\rho}\velspace\to T_{v^\rho}\velmancan,\qquad \dd_{v^\rho}\projection[\Gamma]=\Gamma.
    \]
    The chain rule implies $\dd_0\retraction_{v^\rho}^\mathrm{can}=\Id_{T_{v^\rho}\velmancan}$.
\end{proof}

As shown in Figure~\ref{fig:7}, the canonical retraction $\retraction^\mathrm{can}$ can be transported back to the density manifold $\pdfspace$ through the inverse diffeomorphism \(\canonicalmap^{-1}=\where{\terminalmap}{\velmancan}:\velmancan\diffeomorphismto\pdfspace\). For $\rho\in\pdfspace$, \(v^\rho=\canonicalmap(\rho)\in\velmancan\), define 
\begin{equation}\label{eq:23}
    \retraction_\rho^\mathrm{den}:=\canonicalmap^{-1}\circ\retraction_{v^\rho}^\mathrm{can}\circ \dd_\rho\canonicalmap:T_\rho\pdfspace\to\pdfspace.
\end{equation}
The following proposition states that the map~\eqref{eq:23} is indeed a retraction on the density manifold $\pdfspace$.

\begin{proposition}\label{prop:5}
    The map~\eqref{eq:23} is a retraction on the density manifold $\pdfspace$. In particular, for a tangent vector $\xi^{\rho,f}=\rho\cdot(f-\E_\rho[f])\in T_\rho\pdfspace$, the updated density is
    \[
    \retraction_\rho^\mathrm{den}(\xi^{\rho,f})=\terminalmap(v^\rho+\Gamma^{\rho,f}).
    \]
\end{proposition}

\begin{proof}
    Since $\canonicalmap^{-1}$, $\retraction_{v^\rho}^\mathrm{can}$, and $\dd_\rho\canonicalmap$ are smooth, $\retraction_\rho^\mathrm{den}$ is smooth. For a tangent vector $\xi^{\rho,f}=\rho\cdot(f-\E_\rho[f])\in T_\rho\pdfspace$, we have
    \[
    \retraction_\rho^\mathrm{den}(\xi^{\rho,f})=\canonicalmap^{-1}\left(\retraction_{v^\rho}^\mathrm{can}(\dd_\rho\canonicalmap[\xi^{\rho,f}])\right)=\terminalmap\left(\retraction_{v^\rho}^\mathrm{can}(\Gamma^{\rho,f})\right)=\terminalmap(v^\rho+\Gamma^{\rho,f}).
    \]
    In particular, for $\xi^{\rho,f}=0$, i.e., $\Gamma^{\rho,f}=\dd_\rho\canonicalmap[\xi^{\rho,f}]=0$, we have
    \[
    \retraction_\rho^\mathrm{den}(0)=\terminalmap(v^\rho+0)=\rho.
    \]
    For the differential, the chain rule implies
    \[
    \dd_0\retraction_\rho^\mathrm{den}
    =\dd_{v^\rho}\canonicalmap^{-1}\circ \dd_0\retraction_{v^\rho}^\mathrm{can}\circ \dd_\rho\canonicalmap
    =\dd_{v^\rho}\canonicalmap^{-1}\circ \Id_{T_{v^\rho}\velmancan}\circ \dd_\rho\canonicalmap
    =\Id_{T_\rho\pdfspace}.
    \]
    Therefore, $\retraction_\rho^\mathrm{den}$ is a retraction on $\pdfspace$.
\end{proof}

For a stepsize \(\eta>0\), we evaluate the canonical and density retractions along the scaled tangent direction \(\eta\Gamma^{\rho,f}=\Gamma^{\rho,\eta f}\):
\begin{subequations}\label{eq:24}
    \begin{align}
        &\bar{v}=v^\rho+\eta\Gamma^{\rho,f}\in\velspace,\label{eq:25}\\
        &q=\terminalmap(\bar{v})=\terminalmap(v^\rho+\eta\Gamma^{\rho,f})\in\pdfspace,\label{eq:26}\\
        &v^q=\canonicalmap(q)=\projection(v^\rho+\eta\Gamma^{\rho,f})\in\velmancan.\label{eq:27}
    \end{align}
\end{subequations}
The stepsize $\eta$ controls the displacement magnitude. As summarized in Figure~\ref{fig:8}, the canonical retraction consists of two steps: tangential update~\eqref{eq:25} and canonicalization~\eqref{eq:26}--\eqref{eq:27}.

\subsection{Value-Ascent Certificate}\label{subsec:8}

Section~\ref{subsec:7} constructs a geometrically feasible and CFM-compatible finite-stepsize update rule. The retraction properties determine where the updated iterate lives, but they do not yet characterize how the finite-stepsize velocity-space perturbation changes the induced probability path or terminal density. In this subsection, we show that canonical tangent directions generated by posterior values admit an exact dynamical characterization.

Let \((\nu_t)_{t\in[0,1]}\) denote the density path generated by \(\bar v\), and recall that \((p_t^\rho)_{t\in[0,1]}\) is the density path generated by \(v^\rho\). These two paths satisfy
\[
\nu_0=p_0^\rho=p_0,
\qquad
\nu_1=q,
\qquad
p_1^\rho=\rho.
\]
The following proposition gives the central dynamics governing their difference.

\begin{proposition}\label{prop:6}
    Consider the canonical retraction~\eqref{eq:24} with a stepsize $\eta>0$. Along every characteristic 
    \[
    \frac{\dd Y_t}{\dd t}=\bar{v}_t(Y_t),
    \]
    we have the trajectory-wise identity
    \begin{equation}\label{eq:28}
    \frac{\dd}{\dd t}\left[\log\frac{p_t^\rho(Y_t)}{\nu_t(Y_t)}+\eta\left(V_t^\rho[f](Y_t)-\E_\rho[f]\right)\right]=\eta^2\kappa_t\norm[2]{\nabla V_t^\rho[f](Y_t)}^2\geq0.
    \end{equation}
    Consequently, we have the density-level identity
    \begin{equation}\label{eq:29}
        \frac{\dd}{\dd t}\left(\E_{X\sim\nu_t}\left[V_t^\rho[f](X)\right]-\E_\rho[f]-\frac{1}{\eta}\KL{\nu_t}{p_t^\rho}\right)=\eta\kappa_t\E_{X\sim\nu_t}\left[\norm[2]{\nabla V_t^\rho[f](X)}^2\right]\geq0.
    \end{equation}
    In other words,
    \begin{equation}\label{eq:30}
        \E_{X\sim\nu_t}\left[V_t^\rho[f](X)\right]-\E_\rho[f]=\frac{1}{\eta}\KL{\nu_t}{p_t^\rho}+\eta\int_{0}^{t}\kappa_s\E_{X\sim\nu_s}\left[\norm[2]{\nabla V_s^\rho[f](X)}^2\right]\dd s\geq0.
    \end{equation}
\end{proposition}

\begin{proof}
    Along the characteristic $\frac{\dd Y_t}{\dd t}=\bar{v}_t(Y_t)$, the continuity equation of $(\nu_t,\bar{v}_t)$ gives
    \[
    \frac{\dd}{\dd t}\log\nu_t(Y_t)=\partial_t\log\nu_t(Y_t)+\bar{v}_t(Y_t)\cdot\nabla\log\nu_t(Y_t)=-\nabla\cdot\bar{v}_t(Y_t).
    \]
    The continuity equation of $(p_t^\rho,v_t^\rho)$ gives
    \begin{align*}
        \frac{\dd}{\dd t}\log p_t^\rho(Y_t)
        =&\partial_t\log p_t^\rho(Y_t)+\left(v_t^\rho(Y_t)+\eta\kappa_t\nabla V_t^\rho[f](Y_t)\right)\cdot\nabla\log p_t^\rho(Y_t)\\
        =&-\nabla\cdot v_t^\rho(Y_t)+\eta\kappa_t\nabla V_t^\rho[f](Y_t)\cdot\nabla\log p_t^\rho(Y_t).
    \end{align*}
    Therefore,
    \[
    \frac{\dd}{\dd t}\log\frac{p_t^\rho(Y_t)}{\nu_t(Y_t)}=\eta\kappa_t\nabla V_t^\rho[f](Y_t)\cdot\nabla\log p_t^\rho(Y_t)+\eta\kappa_t\Delta V_t^\rho[f](Y_t).
    \]

    As provided in Appendix~\ref{subsec:41}, the backward equation~\eqref{eq:196} for the posterior value is
    \begin{equation*}
        \partial_tV_t^\rho[f](Y_t)
        +
        \left(v_t^\rho(Y_t)+\kappa_t\nabla\log p_t^\rho(Y_t)\right)\cdot\nabla V_t^\rho[f](Y_t)
        +
        \kappa_t\Delta V_t^\rho[f](Y_t)
        =0;
    \end{equation*}
    hence, along the characteristic $\frac{\dd Y_t}{\dd t}=\bar{v}_t(Y_t)$, the posterior value satisfies
    \begin{align*}
        \frac{\dd}{\dd t}V_t^\rho[f](Y_t)
        =&\partial_tV_t^\rho[f](Y_t)+\left(v_t^\rho(Y_t)+\eta\kappa_t\nabla V_t^\rho[f](Y_t)\right)\cdot\nabla V_t^\rho[f](Y_t)\\
        =&\partial_tV_t^\rho[f](Y_t)+v_t^\rho(Y_t)\cdot\nabla V_t^\rho[f](Y_t)+\eta\kappa_t\norm[2]{\nabla V_t^\rho[f](Y_t)}^2\\
        =&-\kappa_t\nabla\log p_t^\rho(Y_t)\cdot\nabla V_t^\rho[f](Y_t)-\kappa_t\Delta V_t^\rho[f](Y_t)+\eta\kappa_t\norm[2]{\nabla V_t^\rho[f](Y_t)}^2.
    \end{align*}
    Therefore,~\eqref{eq:28} holds. Taking expectation over $Y_0\sim p_0$, which implies $Y_t\sim\nu_t$, gives~\eqref{eq:29}. Integrating from $0$ to $t$ gives~\eqref{eq:30}.
\end{proof}

The density-level identity~\eqref{eq:29} provides an intuitive dynamical picture of how posterior-value improvement and marginal KL divergence evolve along the updated probability path. By the definition of \(V_t^\rho[f]\) and the law of total expectation, its expectation under the marginal \(p_t^\rho\) equals \(\E_\rho[f]\) at every time \(t\):
\[
\E_{X\sim p_t^\rho}\left[V_t^\rho[f](X)\right]=\E_{X_t\sim p_t^\rho}\left[\E_{X_1\sim p_{1|t}^\rho(\cdot|X_t)}\left[f(X_1)\right]\right]=\E_\rho[f].
\]
Therefore,
\[
\E_{X\sim\nu_t}\left[V_t^\rho[f](X)\right]-\E_\rho[f]
=
\E_{X\sim\nu_t}\left[V_t^\rho[f](X)\right]
-
\E_{X\sim p_t^\rho}\left[V_t^\rho[f](X)\right]
\]
measures the expected posterior-value gain of the updated marginal \(\nu_t\) relative to the marginal \(p_t^\rho\) at the same time \(t\). This gain starts from zero,
\[
\left.
\left(
\E_{X\sim\nu_t}\left[V_t^\rho[f](X)\right]-\E_\rho[f]
\right)
\right|_{t=0}
=
0,
\]
and terminates at the terminal-value improvement
\[
\left.
\left(
\E_{X\sim\nu_t}\left[V_t^\rho[f](X)\right]-\E_\rho[f]
\right)
\right|_{t=1}
=
\E_{X\sim q}[f(X)]-\E_{X\sim\rho}[f(X)].
\]
Equation~\eqref{eq:30} decomposes this posterior-value gain into two nonnegative terms: the scaled marginal KL divergence \(\frac{1}{\eta}\KL{\nu_t}{p_t^\rho}\) and the integral term
\[
\eta\int_0^t\kappa_s\E_{X\sim\nu_s}\left[\norm[2]{\nabla V_s^\rho[f](X)}^2\right]\dd s.
\]

\begin{figure}[t]
    \centering
    \input{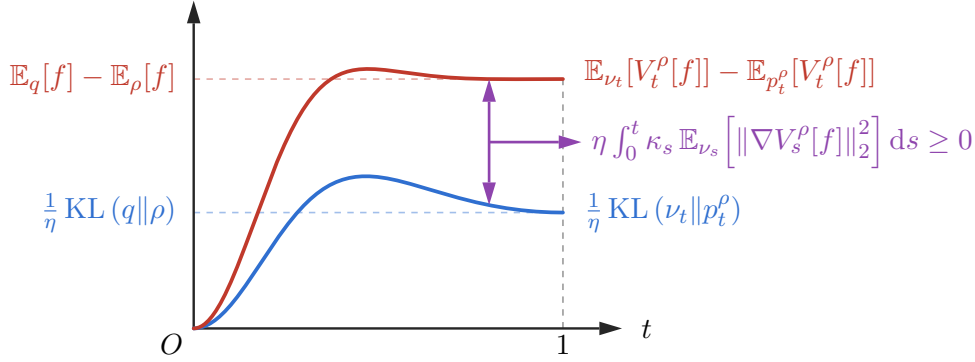}
    \caption{Mechanism of finite-stepsize value ascent. Along the updated density path \((\nu_t)_{t\in[0,1]}\) generated by \(\bar{v}=v^\rho+\eta\Gamma^{\rho,f}\), the expected posterior-value improvement relative to the density path \((p_t^\rho)_{t\in[0,1]}\) decomposes into the scaled marginal divergence \(\frac{1}{\eta}\KL{\nu_t}{p_t^\rho}\) and an accumulated nonnegative quadratic term $\eta\int_0^t\kappa_s\E_{X\sim\nu_s}\left[\norm[2]{\nabla V_s^\rho[f](X)}^2\right]\dd s$. At \(t=1\), this decomposition yields the terminal value improvement \(\E_{X\sim q}[f(X)]-\E_{X\sim\rho}[f(X)]\geq0\).}
    \label{fig:9}
\end{figure}

As illustrated in Figure~\ref{fig:9}, neither the posterior-value gain nor the scaled marginal KL divergence is required to be monotone in \(t\). Their difference, however, is exactly the integral term and is therefore nondecreasing. Equivalently, at every time $t\in[0,1]$,
\[
\E_{X\sim\nu_t}\left[V_t^\rho[f](X)\right]-\E_\rho[f]
\geq
\frac{1}{\eta}\KL{\nu_t}{p_t^\rho}
\geq0,
\]
and the excess of the posterior-value gain over the scaled KL divergence accumulates monotonically along the updated probability path. At the terminal time \(t=1\), the fully accumulated gain becomes \(\E_q[f]-\E_\rho[f]\).

The subsequent canonicalization step~\eqref{eq:27} forgets both the particular noncanonical velocity representation \(\bar{v}\) and its intermediate marginal path \((\nu_t)_{t\in(0,1)}\), while retaining the terminal density $q=\terminalmap(\bar{v})$. Consequently, the complete terminal-value improvement accumulated along the updated path is preserved, since \(\E_q[f]-\E_\rho[f]\) depends only on the retained terminal density \(q\). At the same time, the projection replaces \(\bar{v}\) by the canonical representative \(v^q=\canonicalmap(q)\), thereby providing a new canonical marginal path \((p_t^q)_{t\in[0,1]}\) and the corresponding posterior-value structures required by the next update. In general, the new canonical path \(p_t^q\) need not coincide with the intermediate updated path \(\nu_t\) for \(t\in(0,1)\), even though both share the same source density \(p_0\) and terminal density \(q\). The above results are summarized in the following theorem.

\begin{theorem}%
    \label{thm:2}
    Define the path dissipation $\Diss_{\rho,\eta,f}:\R^d\to[0,\infty)$ by
    \begin{equation}\label{eq:31}
        \Diss_{\rho,\eta,f}(x):=\eta^2\int_{0}^{1}\kappa_t\norm[2]{\nabla V_t^\rho[f]\left(\flow_{1\to t}^{\rho,\eta,f}(x)\right)}^2\dd t\geq0,
    \end{equation}
    where $\flow^{\rho,\eta,f}$ is the flow map generated by the velocity field $\bar{v}=v^\rho+\eta\Gamma^{\rho,f}$. The updated terminal density~\eqref{eq:26} is
    \begin{equation}\label{eq:32}
        q(x)=\rho(x)\exp\left(\eta\left(f(x)-\E_{X\sim\rho}[f(X)]\right)-\Diss_{\rho,\eta,f}(x)\right),
    \end{equation}
    which, for $\eta>0$, enjoys the value-ascent certificate
    \begin{equation}\label{eq:33}
    \E_{X\sim q}[f(X)]-\E_{X\sim\rho}[f(X)]=\frac{1}{\eta}\KL{q}{\rho}+\frac{1}{\eta}\E_{X\sim q}\left[\Diss_{\rho,\eta,f}(X)\right]\geq 0.
    \end{equation}
    Furthermore, we have
    \[
    \E_{X\sim q}[f(X)]=\E_{X\sim\rho}[f(X)]\quad\Leftrightarrow\quad\rho= q
    \quad\Leftrightarrow\quad f\equiv\const.
    \]
\end{theorem}

\begin{proof}
    Integrating~\eqref{eq:28} from $t=0$ to $t=1$ gives~\eqref{eq:32}. Substituting $t=1$ into~\eqref{eq:30} gives~\eqref{eq:33}. The remaining task is to prove the necessity and sufficiency of the equality. First, if $\E_{X\sim q}[f(X)]=\E_{X\sim\rho}[f(X)]$ holds, then~\eqref{eq:33} implies $\KL{q}{\rho}=0$. By the continuity of $\rho,q$, we have $\rho= q$. Second, $\rho= q$ implies $\E_{X\sim q}\left[\Diss_{\rho,\eta,f}(X)\right]=0$. By~\eqref{eq:31}, we have $\nabla V_t^\rho[f]\left(\flow_{1\to t}^{\rho,\eta,f}(x)\right)\aeq0$ in $\R^d$. Since the flow map is a diffeomorphism, we have $\nabla V_t^\rho[f](x)=0$ for $t\in(0,1)$ and $x\in\R^d$ almost everywhere (a.e.). Since $\E_{\rho}[\abs{f}]<\infty$ and the convolution kernel is a nondegenerate Gaussian, $V_t^\rho[f]$ is smooth; hence, $V_t^\rho[f]\equiv c_t$ is constant. By Proposition~\ref{prop:3}, $f\equiv\const$. Third, if $f\equiv\const$, then $\E_{X\sim q}[f(X)]=\E_{X\sim\rho}[f(X)]$.
\end{proof}

Remarkably, Theorem~\ref{thm:2} provides a finite-stepsize guarantee, rather than merely a first-order ascent result for infinitesimal perturbations. The increase in the expectation of $f$ is decomposed exactly into a KL displacement of $q$ from $\rho$ and a nonnegative path-dissipation term. This decomposition provides the stability mechanism behind the finite-stepsize KL descent of Newton Matching in Section~\ref{sub2sec:7}. Extensions with a time-dependent step schedule $\eta_t$ and a non-tangent update direction are provided in Appendix~\ref{app:7}.

\begin{remark}\label{remark:1}
    Theorem~\ref{thm:2} can be read as a general value-ascent certificate. The terminal observable \(f\) plays two roles simultaneously: it parametrizes a density tangent vector \(\xi^{\rho,f}=\rho\cdot(f-\E_\rho[f])\) as well as its lifted canonical tangent vector $\Gamma^{\rho,f}$, and it is the value whose expectation is improved by the corresponding canonical retraction. Therefore, different choices of \(f\) lead to different update semantics.
    \begin{itemize}
        \item We can choose a fixed nonconstant terminal observable $f$. Then, the expectation of $f$ strictly increases if we perform the canonical retraction $\rho_{k+1}=\terminalmap(v^{\rho_k}+\eta\Gamma^{\rho_k,f})$ iteratively. As discussed in Section~\ref{sub2sec:17}, choosing $f=r$ gives direct reward ascent whenever $r$ is nonconstant.
        \item We can choose a density-dependent terminal observable \(f^\rho\). Although \(f^\rho\) need not itself be an objective, the density tangent vector \(\rho\cdot\left(f^\rho-\E_\rho[f^\rho]\right)\in T_\rho\pdfspace\) may encode a local descent direction for an objective on $\pdfspace$. Theorem~\ref{thm:2} guarantees finite-stepsize ascent of the current observable \(f^\rho\), i.e., $\E_q[f^\rho]\geq\E_\rho[f^\rho]$; establishing finite-stepsize descent of the underlying objective requires additional structures. For the reverse-KL objective, \(f^\rho\) is the regularized reward, Section~\ref{sec:3} identifies the corresponding canonical tangent as both the negative Fisher--Rao gradient and the Newton direction, and Section~\ref{sub2sec:7} proves finite-stepsize KL descent.\qedhere
    \end{itemize}
\end{remark}

\section{Newton Matching on the Canonical Manifold}
\label{sec:3}

Section~\ref{subsec:3} motivates restricting the search to canonical velocity fields. With the canonical manifold and its update mechanism now established in Sections~\ref{sec:1} and~\ref{sec:2}, we can state this optimization problem rigorously. The optimization problem on the density manifold $\pdfspace$ is
\[
    \min_{\rho\in\pdfspace}J(\rho),
    \qquad
    J(\rho):=\KL{\rho}{\pi_{\mu,\tau,r}}.
\]
By Theorem~\ref{thm:1}, the canonical-velocity map $\canonicalmap:\pdfspace\diffeomorphismto\velmancan$ is a diffeomorphism. Therefore, the density-coordinate problem is equivalent to the velocity-coordinate problem
\begin{equation*}%
    \min_{v\in\velmancan}J^{\mathrm{can}}(v),
    \qquad
    J^{\mathrm{can}}(v):=\KL{\terminalmap(v)}{\pi_{\mu,\tau,r}}.
\end{equation*}
Under the assumption $\pi_{\mu,\tau,r}\in\pdfspace$, the unique minimizers are
\[
    \rho^\star=\pi_{\mu,\tau,r},
    \qquad
    v^\star=\canonicalmap(\pi_{\mu,\tau,r}).
\]

These equivalent optimization problems specify their respective objectives and minimizers, but they do not yet specify the tangent direction used to approach them. Recall from Section~\ref{sec:2} that, at a current canonical velocity field \(v^\rho=\canonicalmap(\rho)\), the canonical retraction maps any tangent vector \(\Gamma\in T_{v^\rho}\velmancan\) to a finite-stepsize feasible update:
\[
    \retraction_{v^\rho}^{\mathrm{can}}(\Gamma)
    =
    \projection(v^\rho+\Gamma)
    \in\velmancan.
\]
The remaining question is therefore which tangent vector $\Gamma$ should be supplied to this retraction for the reverse-KL objective. By Proposition~\ref{prop:2}, every canonical tangent vector can be represented as \(\Gamma^{\rho,f}\) for a terminal observable \(f\), with \(\Gamma_t^{\rho,f}=\kappa_t\nabla V_t^\rho[f]\). Thus, we need to identify the density-dependent terminal observable \(f^\rho\) whose canonical lift \(\Gamma^{\rho,f^\rho}\) gives the appropriate search direction.

Since the problem is formulated as an optimization problem on a Riemannian manifold, two standard choices are the negative Riemannian gradient, which uses first-order information, and the Newton direction, which uses the Hessian defined by an affine connection. For the Fisher--Rao metric and mixture connection introduced in Section~\ref{sec:1}, we will prove that these two directions coincide. This tangent direction then instantiates the objective-agnostic canonical retraction of Section~\ref{sec:2}.

Section~\ref{subsec:9} derives the negative Fisher--Rao gradient and transports it from the density manifold $\pdfspace$ to the canonical manifold $\velmancan$. Section~\ref{subsec:10} derives the Newton direction under the mixture connection and proves that it coincides with the negative Fisher--Rao gradient. Finally, Section~\ref{subsec:11} applies the canonical retraction to this tangent direction, thereby defining the Newton Matching iteration and deriving the exact formula for the terminal-density update, which is important for the convergence analysis in Section~\ref{sec:4}.

\subsection{Fisher--Rao Gradient}\label{subsec:9}

We first derive the first-order choice of the tangent direction for the canonical retraction. Since the reverse-KL objective \(J\) is naturally defined on the density manifold \(\pdfspace\), it is convenient to derive its gradient in density coordinates and then transport the resulting tangent vector to the canonical manifold through \(\dd_\rho\canonicalmap\). The derivation proceeds in three steps. We first compute the differential \(\dd_\rho J\in T_\rho^*\pdfspace\). Then, we identify its tangent-vector representation under the Fisher--Rao metric. Finally, we canonically lift this density tangent vector to \(T_{v^\rho}\velmancan\), obtaining the negative-gradient direction that can be supplied to the canonical retraction.

For the first step, we compute the differential $\dd_\rho J$. Consider a smooth curve $(\rho^\varepsilon)_{\varepsilon\in(-\delta,\delta)}\subset\pdfspace$ which passes through $\rho$ at $\varepsilon=0$, where $\xi=\where{\frac{\dd}{\dd\varepsilon}\rho^\varepsilon}{\varepsilon=0}\in T_\rho\pdfspace$ denotes the tangent vector. According to Proposition~\ref{prop:1}, we have
\[
    \int_{\R^d}\xi(x)\,\dd x=0.
\]
Along the tangent vector $\xi\in T_\rho\pdfspace$, the first variation of $J$ is
\begin{equation*}
    \where{\frac{\dd}{\dd \varepsilon}J(\rho^\varepsilon)}{\varepsilon=0}=\int_{\R^d}\left(\log\frac{\rho(x)}{\pi_{\mu,\tau,r}(x)}+1\right)\xi(x)\,\dd x=\int_{\R^d}\log\frac{\rho(x)}{\pi_{\mu,\tau,r}(x)}\xi(x)\,\dd x.
\end{equation*}
By definition, the differential of $J$ at $\rho$ is a linear functional of $\xi$:
\begin{equation*}
    \dd_\rho J\in T_\rho^*\pdfspace,\qquad \dd_\rho J: T_\rho\pdfspace\to\R,\qquad \xi\mapsto \dd_\rho J[\xi]:=\int_{\R^d}\log\frac{\rho(x)}{\pi_{\mu,\tau,r}(x)}\xi(x)\,\dd x.
\end{equation*}

For the second step, to obtain an admissible update direction in $T_\rho\pdfspace$, we compute the metric dual of $\dd_\rho J$ under the Fisher--Rao metric, i.e., the Fisher--Rao gradient. Recall from Section~\ref{sub2sec:6} that $\pdfspace$ is equipped with the Fisher--Rao metric $g_\rho^\fisher$ in~\eqref{eq:18}. The Fisher--Rao gradient $\gradFR J_\rho\in T_\rho\pdfspace$ of $J$ at $\rho$ is a tangent vector which satisfies
\[
\forall \xi\in T_\rho\pdfspace\text{, }g_\rho^\fisher(\gradFR J_\rho,\xi)=\dd_\rho J[\xi].
\]
In other words, we need to find the unique tangent vector $\gradFR J_\rho\in T_\rho\pdfspace$ such that
\begin{equation}\label{eq:34}
\forall \xi\in T_\rho\pdfspace,\,\int_{\R^d}\left(\frac{\gradFR J_\rho(x)}{\rho(x)}-\log\frac{\rho(x)}{\pi_{\mu,\tau,r}(x)}\right)\xi(x)\,\dd x=0.
\end{equation}
Corollary~\ref{cor:6} shows that
\[
\xi^{\rho,\log(\rho/\pi_{\mu,\tau,r})}=\rho\cdot\left(\log\frac{\rho}{\pi_{\mu,\tau,r}}-\E_{X\sim\rho}\left[\log\frac{\rho(X)}{\pi_{\mu,\tau,r}(X)}\right]\right)
\]
is a tangent vector in $T_\rho\pdfspace$. The detailed proof is deferred to Appendix~\ref{sub2sec:23}. This tangent vector satisfies condition~\eqref{eq:34}. Therefore, the Fisher--Rao gradient is
\begin{equation}\label{eq:35}
\gradFR J_\rho(x)=\rho(x)\cdot\left(\log\frac{\rho(x)}{\pi_{\mu,\tau,r}(x)}-\E_{X\sim\rho}\left[\log\frac{\rho(X)}{\pi_{\mu,\tau,r}(X)}\right]\right).
\end{equation}

Finally, we compute the canonical lift $\dd_\rho\canonicalmap[\gradFR J_\rho]\in T_{v^\rho}\velmancan$ to obtain a directly learnable direction. Since \(\pi_{\mu,\tau,r}(x)\propto\mu(x)e^{\tau r(x)}\), we define the regularized reward as
\[
\tilde{r}^\rho(x):=r(x)-\frac{1}{\tau}\log\frac{\rho(x)}{\mu(x)},
\]
which satisfies
\[
\log\frac{\rho(x)}{\pi_{\mu,\tau,r}(x)}=-\tau\tilde{r}^\rho(x)+\const.
\]
In other words,
\[
\gradFR J_\rho(x)=\xi^{\rho,-\tau\tilde{r}^\rho}=-\tau\xi^{\rho,\tilde{r}^\rho}\in T_\rho\pdfspace.
\]
According to Proposition~\ref{prop:2}, the canonical lift is
\[
\dd_\rho\canonicalmap[\gradFR J_\rho]=\Gamma^{\rho,-\tau\tilde{r}^\rho}=-\tau\Gamma^{\rho,\tilde{r}^\rho}\in T_{v^\rho}\velmancan,
\]
where
\begin{equation}\label{eq:36}
\Gamma_t^{\rho,\tilde{r}^\rho}(x_t):=\kappa_t\nabla V_t^\rho[\tilde{r}^\rho](x_t)=\Cov_{X_1\sim p_{1|t}^\rho(\cdot|x_t)}\left(v_{t|1}(x_t|X_1),\tilde{r}^\rho(X_1)\right).
\end{equation}
Thus, the density-dependent signal $\tilde r^\rho$ specializes the objective-agnostic canonical retraction of Section~\ref{sec:2} to the first-order descent direction of the reverse-KL objective.

\begin{remark}\label{remark:2}
    In Euclidean space $\R^d$, the distinction between differentials and gradients is often hidden by the Euclidean metric. Consider a smooth objective function $J:\R^d\to\R$. The tangent space at $x\in\R^d$ is exactly $T_x\R^d\cong\R^d$. Consider a smooth curve $(x^\varepsilon)_{\varepsilon\in(-\delta,\delta)}\subset\R^d$ which passes through $x$ at $\varepsilon=0$. Denote $\xi=\where{\frac{\dd}{\dd\varepsilon}x^\varepsilon}{\varepsilon=0}\in T_x\R^d$; equivalently, denote $\xi=(\xi_i)_{i=1}^d\in T_x\R^d\cong\R^d$. Then, the first variation of $J$ along the direction $\xi$ is
    \[
    \where{\frac{\dd}{\dd\varepsilon}J(x^\varepsilon)}{\varepsilon=0}=\frac{\partial J}{\partial x^\top}\xi,
    \]
    where
    \[
    \frac{\partial J}{\partial x}=\left(\frac{\partial J}{\partial x_i}\right)_{i=1}^d\in T_x\R^d\cong\R^d.
    \]
    In other words, the differential of $J$ at $x$ is
    \[
    \dd_x J\in T_x^*\R^d,\qquad \xi\mapsto \dd_x J[\xi]=\frac{\partial J}{\partial x^\top}\xi.
    \]
    The standard Euclidean metric is defined as
    \[
        g^{\mathrm{Ed}}_x:T_x\R^d\times T_x\R^d\to\R,\qquad g^{\mathrm{Ed}}_x(\xi,\zeta):=\xi^\top\zeta.
    \]
    Therefore, the gradient under the Euclidean metric is
    \[
    \grad_{\mathrm{Ed}} J_x=\frac{\partial J}{\partial x}\in T_x\R^d\cong\R^d,
    \]
    since
    \[
    \forall \xi\in T_x\R^d\cong\R^d,\,\dd_x J[\xi]=\left(\grad_{\mathrm{Ed}} J_x\right)^\top\xi=g^{\mathrm{Ed}}_x(\grad_{\mathrm{Ed}} J_x,\xi).\qedhere
    \]
\end{remark}

\begin{remark}\label{remark:3}
    The tangent vector $\dd_\rho\canonicalmap[\gradFR J_\rho]\in T_{v^\rho}\velmancan$ is also the gradient of the objective $J\circ\canonicalmap^{-1}$ at $\canonicalmap(\rho)\in\velmancan$ under the transported Fisher--Rao metric. In other words, the diffeomorphism $\canonicalmap:\pdfspace\diffeomorphismto\velmancan$ preserves the gradient when the metric is transported accordingly. Let
    \[
    J^{\mathrm{can}}=J\circ\canonicalmap^{-1}: \velmancan\to[0,\infty),\qquad v\mapsto\KL{\terminalmap(v)}{\pi_{\mu,\tau,r}}.
    \]
    Consider any canonical velocity field $v\in\velmancan$ and its terminal density $\rho^v=\terminalmap(v)\in\pdfspace$. Then, we have
    \[
    \dd_v J^{\mathrm{can}}\in T_{v}^*\velmancan,\qquad \dd_v J^{\mathrm{can}}:T_{v}\velmancan\to \R,\qquad \Gamma\mapsto \dd_{\rho^v} J\left[\dd_v\canonicalmap^{-1}[\Gamma]\right].
    \]
    
    Consider the transported Fisher--Rao metric $g^\mathrm{can}$ in~\eqref{eq:19}. Then, $\forall\Gamma\in T_v\velmancan$, we have
    \[
    \begin{aligned}
    g^\mathrm{can}_{v}\left(\dd_{\rho^v}\canonicalmap[\gradFR J_{\rho^v}],\Gamma\right)
    =&g^\mathrm{FR}_{\rho^v}\left(\gradFR J_{\rho^v},\dd_v\canonicalmap^{-1}\left[\Gamma\right]\right)
    =\dd_{\rho^v} J\left[\dd_v\canonicalmap^{-1}\left[\Gamma\right]\right]=\dd_v J^{\mathrm{can}}\left[\Gamma\right].
    \end{aligned}
    \]
    In other words, the gradient of $J^{\mathrm{can}}$ under the transported Fisher--Rao metric $g^\mathrm{can}_{v}$ is
    \[
    \grad_{\mathrm{can}}J_v=\dd_{\rho^v}\canonicalmap[\gradFR J_{\rho^v}]=-\tau\Gamma^{\rho^v,\tilde{r}^{\rho^v}}\in T_{v}\velmancan.\qedhere
    \]
\end{remark}

\subsection{Newton Direction}\label{subsec:10}

Section~\ref{subsec:9} identifies the first-order choice for the input to the canonical retraction:
\[
    -\grad_{\mathrm{can}}J^{\mathrm{can}}_{v^\rho}
    =
    \tau\Gamma^{\rho,\tilde r^\rho}.
\]
We now derive the corresponding second-order choice. The Newton direction is obtained by linearizing the stationarity condition \(\dd_\rho J=0\), where the Hessian is defined using the mixture connection introduced in Section~\ref{subsec:6}. The central result is the bilinear-form identity
\[
    \mathrm{Hess}^{\mathrm{mix}}J_\rho
    =
    g_\rho^\fisher.
\]
Consequently, after the Fisher--Rao metric identifies tangent and cotangent vectors, the Hessian acts as the identity, and the Newton direction coincides with the negative gradient.

For the first step, we compute the Hessian of $J$. Consider the mixture connection defined in~\eqref{eq:20} and recall the definition of the Hessian in Section~\ref{sub2sec:3}. For any $\rho\in\pdfspace$ and two tangent vectors $\xi_1,\xi_2\in T_\rho\pdfspace$, we extend them to smooth tangent vector fields $\DensityTangentFieldFirst,\DensityTangentFieldSecond\in\mathfrak{X}(\pdfspace)$ which satisfy $\DensityTangentFieldFirst_\rho=\xi_1$, $\DensityTangentFieldSecond_\rho=\xi_2$. By definition, the Hessian takes a bilinear form
\[
\mathrm{Hess}^{\mathrm{mix}}J_\rho: T_\rho\pdfspace\times T_\rho\pdfspace\to\R,\qquad \mathrm{Hess}^{\mathrm{mix}}J_\rho[\xi_1,\xi_2]:=\DensityTangentFieldFirst[\DensityTangentFieldSecond[J]](\rho)-\dd_\rho J\left[\left(\nabla_{\DensityTangentFieldFirst}^\mathrm{mix}\DensityTangentFieldSecond\right)_\rho\right].
\]
Consider a smooth curve $(\rho^\varepsilon)_{\varepsilon\in(-\delta,\delta)}\subset\pdfspace$ which passes through $\rho$ at $\varepsilon=0$ and satisfies $\xi_1=\DensityTangentFieldFirst_\rho=\where{\frac{\dd}{\dd\varepsilon}\rho^\varepsilon}{\varepsilon=0}$. According to~\eqref{eq:20}, we have
\[
\dd_\rho J\left[\left(\nabla_{\DensityTangentFieldFirst}^\mathrm{mix}\DensityTangentFieldSecond\right)_\rho\right]=g_\rho^\fisher\left(\left(\nabla^\mathrm{mix}_{\DensityTangentFieldFirst}\DensityTangentFieldSecond\right)_\rho,\xi^{\rho,\log(\rho/\pi_{\mu,\tau,r})}\right)=\DensityTangentFieldFirst\left[g^\fisher\left(\DensityTangentFieldSecond,\xi^{\cdot,\log(\rho/\pi_{\mu,\tau,r})}\right)\right](\rho).
\]
Here $\xi^{\cdot,\log(\rho/\pi_{\mu,\tau,r})}\in\mathfrak{X}(\pdfspace)$ is the fixed-observable extension,
\[
\xi^{\cdot,\log(\rho/\pi_{\mu,\tau,r})}:q\mapsto \xi^{q,\log(\rho/\pi_{\mu,\tau,r})}:=q\cdot\left(\log\frac{\rho}{\pi_{\mu,\tau,r}}-\E_{q}\left[\log\frac{\rho}{\pi_{\mu,\tau,r}}\right]\right).
\]
Therefore, we have
\begin{align*}
    \mathrm{Hess}^{\mathrm{mix}}J_\rho[\xi_1,\xi_2]
    =&\where{\frac{\dd}{\dd\varepsilon}\int_{\R^d}\log\frac{\rho^\varepsilon(x)}{\pi_{\mu,\tau,r}(x)}\DensityTangentFieldSecond_{\rho^\varepsilon}(x)\,\dd x}{\varepsilon=0}-\where{\frac{\dd}{\dd\varepsilon}\int_{\R^d}\log\frac{\rho(x)}{\pi_{\mu,\tau,r}(x)}\DensityTangentFieldSecond_{\rho^\varepsilon}(x)\,\dd x}{\varepsilon=0}\\
    =&\int_{\R^d}\where{\frac{\dd}{\dd\varepsilon}\log\frac{\rho^\varepsilon(x)}{\pi_{\mu,\tau,r}(x)}}{\varepsilon=0}\DensityTangentFieldSecond_{\rho}(x)\,\dd x\\
    =&\int_{\R^d}\frac{\xi_1(x)\xi_2(x)}{\rho(x)}\,\dd x.
\end{align*}
In other words, we have the bilinear-form identity
\begin{equation}\label{eq:37}
    \mathrm{Hess}^{\mathrm{mix}}J_\rho=g_\rho^\fisher.
\end{equation}
Thus, the Hessian is exactly the same bilinear form that represents the differential as a Fisher--Rao gradient. As illustrated in Remark~\ref{remark:4}, this identity corresponds to the Euclidean case in which the classical Hessian matrix is the identity matrix.

For the second step, we state the Newton equation and solve the Newton direction $n_\rho^{\mathrm{mix}}\in T_\rho\pdfspace$. We regard the Hessian as a map from the tangent space $T_\rho\pdfspace$ to the cotangent space $T_\rho^*\pdfspace$:
\[
\mathrm{Hess}^{\mathrm{mix}}J_\rho: T_\rho\pdfspace\to T_\rho^*\pdfspace,\qquad \left(\mathrm{Hess}^{\mathrm{mix}}J_\rho[\xi_1]\right)[\xi_2]:=\mathrm{Hess}^{\mathrm{mix}}J_\rho[\xi_1,\xi_2].
\]
The Newton equation is the linearized equation of $\dd_\rho J=0$, i.e.,
\[
\dd_\rho J+\mathrm{Hess}^{\mathrm{mix}}J_\rho[n_\rho^{\mathrm{mix}}]=0\in T_\rho^*\pdfspace.
\]
In this equation, the right-hand side denotes the zero functional. In other words, $\forall\xi\in T_\rho\pdfspace$,
\[
\dd_\rho J[\xi]+\mathrm{Hess}^{\mathrm{mix}}J_\rho[n_\rho^{\mathrm{mix}},\xi]=g_\rho^\fisher(\gradFR J_\rho+n_\rho^{\mathrm{mix}},\xi)=0.
\]
Here the Fisher--Rao gradient is provided in~\eqref{eq:35}. Take $\xi=\gradFR J_\rho+n_\rho^{\mathrm{mix}}$. By the positive definiteness of the Fisher--Rao metric, the Newton direction is the unique solution
\[
n_\rho^{\mathrm{mix}}=-\gradFR J_\rho=\tau\xi^{\rho,\tilde{r}^\rho}.
\]

Finally, the canonical lift of the Newton direction $n_\rho^{\mathrm{mix}}$ is
\[
n_\rho^{\mathrm{can}}:=\dd_\rho\canonicalmap[n_\rho^{\mathrm{mix}}]=-\dd_\rho\canonicalmap[\gradFR J_\rho]=\tau\Gamma^{\rho,\tilde{r}^\rho}.
\]
By the same argument as in Remark~\ref{remark:3}, $n_\rho^{\mathrm{can}}$ is the Newton direction of $J^{\mathrm{can}}=J\circ\where{\terminalmap}{\velmancan}$ under the transported Fisher--Rao metric $g^\mathrm{can}$ and the transported connection $\nabla^{\mathrm{can}}$.

\begin{remark}\label{remark:4}
    Use the notation of Remark~\ref{remark:2}. In Euclidean space $\R^d$, the standard Euclidean affine connection is defined as
    \[
    \nabla^{\mathrm{Ed}}:\mathfrak{X}(\R^d)\times \mathfrak{X}(\R^d)\to\mathfrak{X}(\R^d),\qquad
    (X,Y)\mapsto\nabla^{\mathrm{Ed}}_XY,\qquad
    (\nabla^{\mathrm{Ed}}_XY)_x:=\DD_x Y[X_x].
    \]
    Consider a smooth curve $(x^\varepsilon)_{\varepsilon\in(-\delta,\delta)}\subset\R^d$ which passes through $x$ at $\varepsilon=0$ and satisfies $\xi:=X_x=\where{\frac{\dd}{\dd\varepsilon}x^\varepsilon}{\varepsilon=0}\in T_x\R^d$. Then, the derivative can be written as
    \[
    \DD_x Y[X_x]:=\where{\frac{\dd}{\dd\varepsilon}Y_{x^\varepsilon}}{\varepsilon=0}.
    \]
    For a smooth objective $J:\R^d\to\R$, a direct calculation gives the Hessian
    \[
    \mathrm{Hess}^\mathrm{Ed}J_x[\xi,\zeta]=\DD_x^2J[\xi,\zeta]=\xi^\top\frac{\partial^2 J_x}{\partial x\partial x^\top}\zeta,
    \]
    where
    \[
    \frac{\partial^2 J_x}{\partial x\partial x^\top}=\left(\frac{\partial^2 J_x}{\partial x_i\partial x_j}\right)_{i,j}\in\R^{d\times d}
    \]
    is the classical Hessian matrix of $J$ at $x$. Therefore,
    \[
    \forall \xi,\zeta\in T_x\R^d\cong\R^d,\,\mathrm{Hess}^{\mathrm{Ed}}J_x[\xi,\zeta]= g^{\mathrm{Ed}}_x(\xi,\zeta)\quad\Leftrightarrow\quad\frac{\partial^2 J_x}{\partial x\partial x^\top}=I.
    \]
    In other words, the Hessian bilinear form is equal to the Euclidean metric if and only if the Hessian matrix is the identity matrix.
\end{remark}

\begin{remark}
    The identity~\eqref{eq:37} is specific to the combination considered in this paper: the reverse-KL objective \(J(\rho)=\mathrm{KL}(\rho\|\pi_{\mu,\tau,r})\), the Fisher--Rao metric $g^\fisher$, and the mixture connection \(\nabla^{\mathrm{mix}}\). Consequently, under other choices, the Newton direction need not coincide with the negative gradient. We retain the name ``Newton Matching'' for our method to emphasize the special coincidence under the considered setting.
\end{remark}

We now turn the interior-time identifiability of the canonical lift in Proposition~\ref{prop:3} into an exact stationarity criterion for the Newton direction.

\begin{proposition}
\label{prop:7}
    For any terminal density $\rho\in\pdfspace$, the following statements are equivalent:
    \[
    \rho=\pi_{\mu,\tau,r}\quad\Leftrightarrow\quad\tilde{r}^\rho\equiv\const\quad\Leftrightarrow\quad\forall t\in(0,1),\,\Gamma^{\rho,\tilde{r}^\rho}\equiv0\quad\Leftrightarrow\quad\exists t_0\in(0,1),\,\Gamma_{t_0}^{\rho,\tilde{r}^\rho}\equiv0.
    \]
\end{proposition}

\begin{proof}
    Since $\rho$ and $\pi_{\mu,\tau,r}$ are both normalized densities, we have $\rho=\pi_{\mu,\tau,r}\,\Leftrightarrow\,\tilde{r}^\rho\equiv\const$. Substituting $f=\tilde{r}^\rho$ into Proposition~\ref{prop:3} finishes the proof.
\end{proof}

Thus, vanishing at any single interior time characterizes the target density: whenever $\rho\neq\pi_{\mu,\tau,r}$,
\[
\Gamma_t^{\rho,\tilde r^\rho}\not\equiv0
\qquad\text{for every }t\in(0,1).
\]
Having identified the direction and its stationarity criterion, we next insert it into the canonical retraction to define the Newton Matching iteration.

\subsection{Newton's Method with Canonical Retraction}
\label{subsec:11}

\begin{figure}[t]
    \centering
    \input{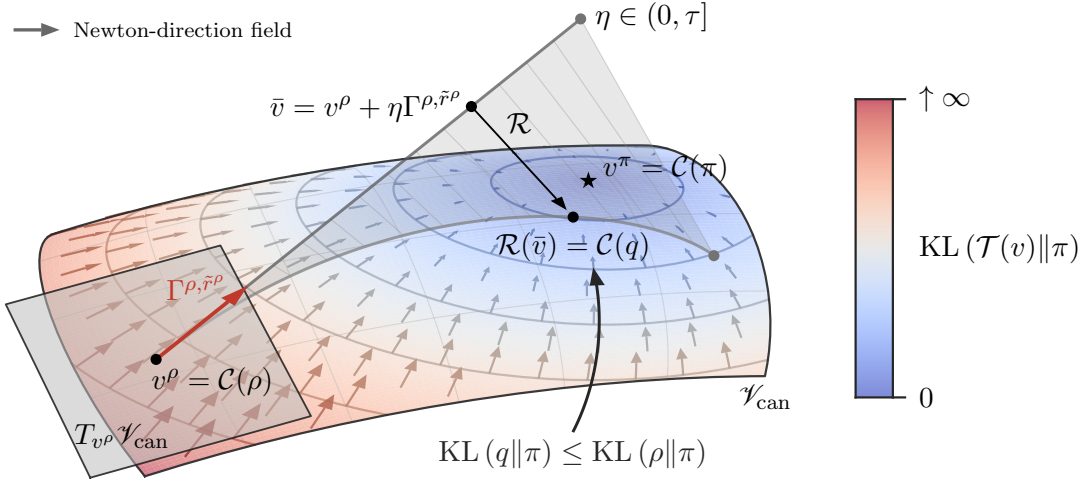}
    \caption{One Newton Matching step. The negative Fisher--Rao gradient and the mixture-connection Newton direction coincide. The canonical retraction with this direction produces a canonical-velocity update that decreases the reverse KL divergence.}
    \label{fig:10}
\end{figure}

We now substitute the tangent vector~\eqref{eq:36} into the canonical retraction~\eqref{eq:24} of Section~\ref{sec:2}. Let $v^k=v^{\rho_k}=\canonicalmap(\rho_k)\in\velmancan$ be the current canonical iterate, with terminal density \(\rho_k=\terminalmap(v^k)\).
Since the full Newton direction is
\[
    n_{\rho_k}^{\mathrm{can}}
    =
    \tau\Gamma^{\rho_k,\tilde r^{\rho_k}},
\]
a stepsize \(\eta_k\in(0,\tau]\) corresponds to the standard Newton damping factor \(\frac{\eta_k}{\tau}\in(0,1]\). Applying the canonical retraction~\eqref{eq:24} gives
\begin{subequations}\label{eq:38}
\begin{align}
    &\bar{v}^{k+1}:=v^k+\eta_k\Gamma^{\rho_k,\tilde r^{\rho_k}}\in\velspace,\label{eq:39}\\
    &\rho_{k+1}
    :=
    \terminalmap
    \left(
    \bar{v}^{k+1}
    \right)\in\pdfspace,\label{eq:40}
    \\
    &v^{k+1}
    :=
    \canonicalmap(\rho_{k+1})=\projection(\bar{v}^{k+1})\in\velmancan.\label{eq:41}
\end{align}
\end{subequations}
In particular, the regularized reward at the current iterate is
\begin{equation*}%
    \tilde r^{\rho_k}(X_1)
    :=
    r(X_1)-\frac{1}{\tau}\log\frac{\rho_k(X_1)}{\mu(X_1)},
\end{equation*}
and the associated tangent vector is given by
\begin{equation*}
    \Gamma_t^{\rho_k,\tilde r^{\rho_k}}(x_t)
    =\kappa_t\nabla V_t^{\rho_k}[\tilde{r}^{\rho_k}](x_t)=
    \Cov_{X_1\sim p_{1|t}^{\rho_k}(\cdot|x_t)}
    \left(
    v_{t|1}(x_t|X_1),
    \tilde r^{\rho_k}(X_1)
    \right).
\end{equation*}
When $\eta_k=\tau$,~\eqref{eq:38} corresponds to a full Newton step. When \(\eta_k\in(0,\tau)\),~\eqref{eq:38} is a damped Newton step. For later use, one generic Newton Matching stage~\eqref{eq:38} can be written compactly as
\begin{equation}\label{eq:42}
    q=\terminalmap\left(v^\rho+\eta\Gamma^{\rho,\tilde{r}^\rho}\right).
\end{equation}
The setting $q=\rho_{k+1}$, $\rho=\rho_{k}$, and $\eta=\eta_k$ corresponds to an ideal Newton update~\eqref{eq:38} at iteration $k$. An ideal Newton Matching step~\eqref{eq:38} is illustrated in Figure~\ref{fig:10}. The Newton Matching iteration is provided in Algorithm~\ref{alg:1}.

\begin{algorithm}[!t]
    \caption{Newton Matching.}
    \label{alg:1}
    \begin{algorithmic}[1]
    \REQUIRE A reference factor $\mu=\rho^\base$ for fine-tuning (with the pretrained model $v^\base$) or $\mu=1$ for sampling; a reward function $r:\R^d\to\R$; an inverse temperature $\tau>0$; a stepsize sequence $\{\eta_k\}_{k=0}^\infty\subset(0,\tau]$; and a convergence criterion.
    \ENSURE The target canonical velocity field $v^\pi=\canonicalmap(\pi_{\mu,\tau,r})\in\velmancan$.
    \STATE Initialize $k\gets0$ and choose an arbitrary canonical velocity $v^{\rho_0}\in\velmancan$.\\
    \COMMENT{For fine-tuning, it is convenient to set $v^{\rho_0}$ as $v^\base$; for sampling, it is convenient to set $v^{\rho_0}$ as the analytical canonical velocity associated with a Gaussian $\rho_0$.}
    \REPEAT
        \STATE Perform the tangential update $\bar{v}^{k+1}\gets v^{\rho_k}+\eta_k\Gamma^{\rho_k,\tilde{r}^{\rho_k}}$. 
        \STATE Perform the canonicalization $v^{\rho_{k+1}}\gets \projection(\bar{v}^{k+1})$.
        \STATE Set $k\gets k+1$.
    \UNTIL{the convergence criterion is satisfied.}
    \STATE Return $v^{\rho_k}$.
\end{algorithmic}
\end{algorithm}

To analyze objective descent and convergence in density coordinates, we specialize Theorem~\ref{thm:2} to the density-dependent observable \(f=\tilde r^\rho\). This yields the following exact formula for the terminal-density update of one ideal Newton Matching stage.

\begin{proposition}\label{prop:8}
    For a terminal density $\rho\in\pdfspace$ and a stepsize $\eta\in(0,\tau]$, we define the path dissipation $\Diss_{\rho,\eta,\tilde r^\rho}:\R^d\to[0,\infty)$ by
    \[
        \Diss_{\rho,\eta,\tilde r^\rho}(x):=\eta^2\int_{0}^{1}\kappa_t\norm[2]{\nabla V_t^\rho[\tilde r^\rho]\left(\flow_{1\to t}^{\rho,\eta,\tilde r^\rho}(x)\right)}^2\dd t,
    \]
    where $\flow^{\rho,\eta,\tilde r^\rho}$ is the flow map generated by the velocity field $v^\rho+\eta\Gamma^{\rho,\tilde r^\rho}$. Then, for an ideal Newton step~\eqref{eq:42}, we have
    \begin{align}
        q(x)
        =&
        \rho(x)^{1-\eta/\tau}
        \pi_{\mu,\tau,r}(x)^{\eta/\tau}
        \exp
        \left(
        \frac{\eta}{\tau}
        \KL{\rho}{\pi_{\mu,\tau,r}}
        -
        \Diss_{\rho,\eta,\tilde r^\rho}(x)
        \right).
        \label{eq:43}
    \end{align}
\end{proposition}
\begin{proof}
    By Theorem~\ref{thm:2}, letting $f=\tilde{r}^\rho$ gives
    \[
    \begin{aligned}
        q(x)
        =&\rho(x)\exp\left(\eta\left(\tilde{r}^\rho(x)-\E_{X\sim\rho}[\tilde{r}^\rho(X)]\right)-\Diss_{\rho,\eta,\tilde{r}^\rho}(x)\right)\\
        =&\rho(x)\exp\left(\frac{\eta}{\tau}\left(\log\frac{\pi_{\mu,\tau,r}(x)}{\rho(x)}-\E_{X\sim\rho}\left[\log\frac{\pi_{\mu,\tau,r}(X)}{\rho(X)}\right]\right)-\Diss_{\rho,\eta,\tilde{r}^\rho}(x)\right),
    \end{aligned}
    \]
    which proves~\eqref{eq:43}.
\end{proof}

In~\eqref{eq:43}, the factor $\rho^{1-\eta/\tau}\pi^{\eta/\tau}$ is the linear interpolation between the current and target densities in log-density coordinates, while the path-dissipation term \(\Diss_{\rho,\eta,\tilde r^\rho}\) is the state-dependent correction induced by the finite-stepsize update in velocity space. In particular, when \(\eta=\tau\), the path-dissipation term is the only state-dependent deviation of the updated terminal density from the target density.

\begin{remark}\label{remark:5}
    The update identity~\eqref{eq:43} takes a form similar to composite objective mirror descent (COMID). In particular, $q(x)\propto\rho(x)\exp\left(\eta\tilde r^{\rho}(x)-\Diss_{\rho,\eta,\tilde r^\rho}(x)\right)$ admits the following equivalent variational characterization
    \[
    q\in\argmin_{p\in\pdfspace}\left\{\KL{p}{\rho}+\eta\left<p,-\tilde{r}^\rho\right>+\left<p,\Diss_{\rho,\eta,\tilde r^\rho}\right>\right\},
    \]
    as follows from the first-order optimality conditions. Here,
    \[
    \left<p,f\right>:=\E_{p}[f]=\int_{\R^d}p(x)f(x)\,\dd x.
    \]
    This variational problem has the same term-by-term structure as the classical COMID update \cite[Equation (3)]{duchi2010composite}. Motivated by this observation, we introduce a COMID-type condition~\eqref{eq:51} in Section~\ref{sub2sec:9} for the global convergence of Newton Matching.
\end{remark}

The geometric construction of Newton Matching is now complete. At the intrinsic level, one stage follows the Newton direction on the canonical manifold. At the algorithmic level, this direction is implemented by a velocity-space correction followed by canonicalization. Finally, at the density level, the same stage satisfies the exact terminal-density formula~\eqref{eq:43}. The next section makes use of this density-level representation to analyze the convergence properties of Newton Matching.

\section{Convergence Guarantees of Newton Matching}\label{sec:4}

Section~\ref{sec:3} identifies the canonical tangent $\tau\Gamma^{\rho,\tilde r^\rho}$ as both the negative Fisher--Rao gradient and the mixture-connection Newton direction, and combines it with the canonical retraction to define Newton Matching. This section studies the convergence guarantees of the Newton Matching iteration. Section~\ref{subsec:12} first establishes finite-stepsize reverse-KL descent for every stepsize $\eta\in(0,\tau]$, and then proves global convergence to the target in reverse KL under mild conditions. Furthermore, reflecting the characteristic local behavior of Newton's method, Section~\ref{subsec:13} establishes local quadratic convergence of the full-step iteration (\(\eta=\tau\)) for both terminal densities and canonical velocity fields. Motivated by this local result, we further provide an optional continuation scheme along the inverse-temperature path as a means of exploiting the quadratic-convergence regime. Under a sufficiently fine temperature grid, each continuation stage is initialized within the local quadratic-convergence neighborhood of its target and hence converges quadratically. Detailed proofs of the global and local convergence results are collected in Appendix~\ref{app:2}.

A time-dependent extension with a stepsize schedule $\eta_t$ is developed in Appendix~\ref{sub2sec:37}. To complement the general infinite-dimensional theory with a concrete, analytically tractable validation, Appendix~\ref{app:3} studies Newton Matching on the isotropic Gaussian family. This family is finite-dimensional and closed under Newton Matching, so the iteration reduces to an explicit parameter recursion. Direct analysis of this recursion shows that both forward and reverse KL divergences converge to zero and, when the initial density is sufficiently close to the target, exhibit local quadratic convergence.

\subsection{KL Descent and Global Convergence}
\label{subsec:12}

This subsection is organized into three parts. Section~\ref{sub2sec:7} first records the infinitesimal descent property of the Newton direction and then strengthens it to a finite-stepsize result through a dissipation-corrected three-point identity. Consequently, the reverse KL decreases at every ideal Newton Matching stage for any finite stepsize $\eta\in(0,\tau]$, while a separate sufficient condition yields forward-KL descent. Section~\ref{sub2sec:8} studies the asymptotic consequences of this stagewise monotonicity. Under non-summable stepsizes, the reverse-KL values converge, and the log-density residual $\log\frac{\rho_k}{\pi}$ becomes asymptotically constant in $L^1(\rho_k)$. Since this on-policy property does not by itself rule out loss of target mass or identify the limiting reverse KL value as zero, Section~\ref{sub2sec:9} introduces two additional mild conditions, under either of which convergence to the target is established. All proofs in Sections~\ref{sub2sec:8} and~\ref{sub2sec:9} are deferred to Appendix~\ref{subsec:36}.

For notational simplicity, we write $\pi:=\pi_{\mu,\tau,r}$ throughout this subsection.

\subsubsection{Finite-Stepsize KL Descent}
\label{sub2sec:7}

Section~\ref{sec:3} identifies the tangent vector used in the canonical retraction as both the negative Fisher--Rao gradient and the Newton direction. The remaining question is finite-stepsize stability: a Newton direction is defined through local differential information, whereas Newton Matching applies a prescribed damped or full canonical retraction. We first record the infinitesimal descent property as a baseline.

\begin{proposition}%
\label{prop:9}
Denote $q^\eta:=\terminalmap\left(v^\rho+\eta\Gamma^{\rho,\tilde{r}^\rho}\right)$. Assume that for every $x\in\R^d$, the map
\[
\eta\mapsto\int_{0}^{1}\kappa_t\norm[2]{\nabla V_t^\rho[\tilde r^\rho]\left(\flow_{1\to t}^{\rho,\eta,\tilde r^\rho}(x)\right)}^2\dd t
\]
is continuous in a neighborhood of $0$. Then, we have
\begin{equation*}%
    \left.
    \frac{\dd}{\dd\eta}
    \KL{q^\eta}{\pi}
    \right|_{\eta=0}
    =
    -
    \tau\Var_{X\sim\rho}\left(\tilde{r}^\rho(X)\right)
    \le0.
\end{equation*}
In particular,
\[
\Var_{X\sim\rho}\left(\tilde{r}^\rho(X)\right)=0\quad\Leftrightarrow\quad\rho=\pi.
\]
\end{proposition}

\begin{proof}
According to Proposition~\ref{prop:8}, we have
\[
    q^\eta(x)
    =
    \rho(x)
    \left(
        1
        +
        \eta
        \left(
            \tilde{r}^\rho(x)-\E_\rho[\tilde{r}^\rho]
        \right)
        +
        O(\eta^2)
    \right).
\]
Therefore,
\begin{align*}
    \where{\frac{\dd}{\dd\eta}\KL{q^\eta}{\pi}}{\eta=0}
    =&\int_{\R^d}\left(\log\frac{\rho(x)}{\pi(x)}+1\right)\where{\frac{\dd}{\dd\eta}q^\eta(x)}{\eta=0}\dd x\\
    =&\int_{\R^d}\left(\log\frac{\rho(x)}{\pi(x)}+1\right)\rho(x)\left(\tilde{r}^\rho(x)-\E_\rho[\tilde{r}^\rho]\right)\dd x\\
    =&-\tau\Var_{X\sim\rho}\left(\tilde{r}^\rho(X)\right)\leq0.
\end{align*}
In particular, $\Var_{X\sim\rho}\left(\tilde{r}^\rho(X)\right)=0$ if and only if $\tilde{r}^\rho=\const$, i.e., $\rho=\pi$.
\end{proof}

Proposition~\ref{prop:9} records the guarantee available from generic first-order analysis. Away from the target, the directional derivative at \(\eta=0\) is strictly negative and therefore ensures descent for sufficiently small positive steps. Without quantitative control of the nonlinear remainder along the retraction, however, this result neither specifies how small the stepsize must be nor certifies any prescribed finite-stepsize update.

The following theorem goes beyond this infinitesimal result by exploiting the specific structure of Newton Matching. The posterior-value-gradient representation of the canonical tangent makes the density path generated by the velocity-space update exactly analyzable, while canonicalization preserves its terminal density. This structure yields a dissipation-corrected three-point identity that controls the complete nonlinear update rather than only its linearization at \(\eta=0\). Consequently, the reverse KL decreases throughout the full admissible stepsize range \(\eta\in(0,\tau]\).

\begin{theorem}%
\label{thm:3}
\begin{subequations}
Let $q=\terminalmap(v^\rho+\eta\Gamma^{\rho,\tilde r^\rho})$ be the terminal density generated by one ideal Newton Matching stage~\eqref{eq:42}. For any density $p\in\pdfspace$ such that $\KL{p}{q}$, $\KL{p}{\rho}$, $\KL{p}{\pi}$, and $\E_{X\sim p}[\Diss_{\rho,\eta,\tilde r^\rho}(X)]$ are finite, we have
\begin{equation}\label{eq:44}
    \KL{p}{q}=\left(1-\frac{\eta}{\tau}\right)\KL{p}{\rho}+\frac{\eta}{\tau}\left(\KL{p}{\pi}-\KL{\rho}{\pi}\right)+\E_{X\sim p}\left[\Diss_{\rho,\eta,\tilde{r}^\rho}(X)\right].
\end{equation}
Consequently, applying $p=q$ gives
\begin{equation}\label{eq:45}
\KL{\rho}{\pi}-\KL{q}{\pi}=\left(\frac{\tau}{\eta}-1\right)\KL{q}{\rho}+\frac{\tau}{\eta}\E_{X\sim q}\left[\Diss_{\rho,\eta,\tilde{r}^\rho}(X)\right].
\end{equation}
Applying $p=\pi$ gives
\begin{equation}\label{eq:46}
    \KL{\pi}{\rho}-\KL{\pi}{q}=\frac{\eta}{\tau}(\KL{\pi}{\rho}+\KL{\rho}{\pi})-\E_{X\sim \pi}\left[\Diss_{\rho,\eta,\tilde{r}^\rho}(X)\right].
\end{equation}
Applying $p=\rho$ gives
\begin{equation}%
    \KL{\rho}{q}=\E_{X\sim \rho}\left[\Diss_{\rho,\eta,\tilde{r}^\rho}(X)\right].
\end{equation}
\end{subequations}
\end{theorem}

\begin{proof}
    By Proposition~\ref{prop:8}, we have
    \[
    \log q(x)=\left(1-\frac{\eta}{\tau}\right)\log\rho(x)+\frac{\eta}{\tau}\log\pi(x)+\frac{\eta}{\tau}\KL{\rho}{\pi}-\Diss_{\rho,\eta,\tilde r^\rho}(x).
    \]
    For any $p\in\pdfspace$, taking the expectation gives
    \begin{align*}
        &\KL{p}{q}\\
        =&\E_{X\sim p}\left[\log\frac{p(X)}{q(X)}\right]\\
        =&\left(1-\frac{\eta}{\tau}\right)\E_{X\sim p}\left[\log\frac{p(X)}{\rho(X)}\right]+\frac{\eta}{\tau}\E_{X\sim p}\left[\log\frac{p(X)}{\pi(X)}\right]-\frac{\eta}{\tau}\KL{\rho}{\pi}+\E_{X\sim p}\left[\Diss_{\rho,\eta,\tilde r^\rho}(X)\right],
    \end{align*}
    which gives~\eqref{eq:44}. Applying $p=\rho,\pi,q$ finishes the proof.
\end{proof}

\begin{remark}
    The dissipation-corrected three-point identity in Theorem~\ref{thm:3} admits an equivalent KL-proximal variational interpretation. In particular, \(q\) is the unique minimizer of the KL-proximal objective
    \[
    q
    =
    \argmin_{p\in\pdfspace}
    \left\{
    \KL{p}{\pi}
    +
    \left(\frac{\tau}{\eta}-1\right)\KL{p}{\rho}
    +
    \frac{\tau}{\eta}\left<p,\Diss_{\rho,\eta,\tilde r^\rho}\right>
    \right\}.
    \]
    This KL-proximal form is equivalent to the COMID-type variational representation discussed in Remark~\ref{remark:5}. The COMID form emphasizes the mirror-descent structure of the update, whereas the KL-proximal form emphasizes its finite-stepsize objective stability. Together, the two interpretations show that Newton Matching couples an intrinsic Newton direction with a KL-proximal stabilization induced by the canonical retraction.
\end{remark}

The finite-stepsize descent theorem below follows directly from~\eqref{eq:45}.
\begin{theorem}%
\label{thm:4}
Let $q=\terminalmap(v^\rho+\eta\Gamma^{\rho,\tilde r^\rho})$ be the terminal density generated by one ideal Newton Matching stage~\eqref{eq:42}. Then, for any stepsize $\eta\in(0,\tau]$, the reverse KL decreases:
\begin{equation*}%
    \KL{q}{\pi} \le \KL{\rho}{\pi}.
\end{equation*}
Furthermore, since $\pi,\rho,q$ are positive densities on $\R^d$, we have
\[
\KL{q}{\pi}=\KL{\rho}{\pi}\quad\Leftrightarrow\quad\rho=q=\pi.%
\]
\end{theorem}

\begin{proof}
Since $\E_{X\sim q}\left[\Diss_{\rho,\eta,\tilde{r}^\rho}(X)\right]\geq0$ and $\eta\in(0,\tau]$,~\eqref{eq:45} implies $\KL{q}{\pi}\le\KL{\rho}{\pi}$.

$\rho=q=\pi$ evidently implies $\KL{q}{\pi}=\KL{\rho}{\pi}$. For the opposite direction, assume that $\KL{q}{\pi}=\KL{\rho}{\pi}$. By~\eqref{eq:45}, we have $\left(\frac{\tau}{\eta}-1\right)\KL{q}{\rho}=0$ and $\E_{X\sim q}\left[\Diss_{\rho,\eta,\tilde{r}^\rho}(X)\right]=0$; hence, $\Diss_{\rho,\eta,\tilde{r}^\rho}\aeq0$. By Proposition~\ref{prop:8}, we have $\nabla V_t^\rho[\tilde{r}^\rho]\aeq0$. Since $\E_{\rho}[\abs{\tilde{r}^\rho}]<\infty$ and the convolution kernel is a nondegenerate Gaussian, $V_t^\rho[\tilde{r}^\rho]$ is smooth; hence, $V_t^\rho[\tilde{r}^\rho]\equiv\const$ for a.e. $t$. According to Proposition~\ref{prop:7}, we have $\rho=\pi$ and $\Gamma^{\rho,\tilde{r}^\rho}=0$; hence, $q=\rho=\pi$.
\end{proof}

Note that the reverse-KL descent in Theorem~\ref{thm:4} is stronger than the value ascent for general tangent vectors in Theorem~\ref{thm:2}. According to Theorem~\ref{thm:2}, the value-ascent certificate~\eqref{eq:33} gives
\[
\E_{X\sim q}[\tilde{r}^\rho(X)]\ge\E_{X\sim\rho}[\tilde{r}^\rho(X)]\quad\Leftrightarrow\quad\KL{q}{\pi}-\KL{q}{\rho}\le\KL{\rho}{\pi}.
\]
In contrast, Theorem~\ref{thm:4} directly gives
\[
\KL{q}{\pi}\le\KL{\rho}{\pi}.
\]

Therefore, for the ideal iteration
\begin{equation*}%
    \rho_{k+1}
    =
    \terminalmap
    \left(
    v^{\rho_k}+\eta_k\Gamma^{\rho_k,\tilde r^{\rho_k}}
    \right),
    \qquad
    \eta_k\in (0,\tau],
\end{equation*}
the sequence
$\{\KL{\rho_k}{\pi}\}_{k=0}^\infty$
is nonincreasing. Furthermore, the reverse KL strictly decreases at each Newton Matching stage, unless the current density $\rho_k$ reaches the target density $\pi$. This establishes the finite-stepsize descent depicted in Figure~\ref{fig:10} throughout the full admissible stepsize range, including the full Newton step.

Before proceeding to the global convergence of Newton Matching, we state an auxiliary forward-KL criterion. This result is not needed for reverse KL descent, but it later provides an ideal anti-collapse safeguard in Remark~\ref{remark:6}.

\begin{corollary}\label{cor:2}
    Let $q=\terminalmap(v^\rho+\eta\Gamma^{\rho,\tilde r^\rho})$ be the terminal density generated by one ideal Newton Matching stage~\eqref{eq:42}. Consider a stepsize $\eta>0$. If
    \begin{equation}\label{eq:47}
        \tau\eta\E_{X\sim \pi}\left[\int_{0}^{1}\kappa_t\norm[2]{\nabla V_t^\rho[\tilde r^\rho]\left(\flow_{1\to t}^{\rho,\eta,\tilde r^\rho}(X)\right)}^2\dd t\right]\le \KL{\pi}{\rho}+\KL{\rho}{\pi},
    \end{equation}
    then the forward KL decreases:
    \[
        \KL{\pi}{q} \le \KL{\pi}{\rho}.
    \]
\end{corollary}

\begin{proof}
    By~\eqref{eq:46}, $\KL{\pi}{q} \le \KL{\pi}{\rho}$ holds if and only if
    \[
        \frac{\tau}{\eta}\E_{X\sim \pi}\left[\Diss_{\rho,\eta,\tilde{r}^\rho}(X)\right]\le \KL{\pi}{\rho}+\KL{\rho}{\pi},
    \]
    which finishes the proof.
\end{proof}

In Corollary~\ref{cor:2}, if the map 
\[
\eta\mapsto\E_{X\sim \pi}\left[\int_{0}^{1}\kappa_t\norm[2]{\nabla V_t^\rho[\tilde r^\rho]\left(\flow_{1\to t}^{\rho,\eta,\tilde r^\rho}(X)\right)}^2\dd t\right]
\] is continuous near $0$, the left-hand side of~\eqref{eq:47} is $O(\eta)$ as $\eta\to0$. In this case, the forward KL also decreases with a sufficiently small stepsize $\eta>0$.

\subsubsection{Self-Calibration}\label{sub2sec:8}

Finite-stepsize descent guarantees a monotone sequence of objective values, but it does not yet describe the asymptotic shape of the density residual. The following theorem identifies the stronger, unconditional statement supplied by the exact Newton Matching iteration: under non-summable stepsizes, the log-density ratio $\log\frac{\rho_k}{\pi}$ becomes asymptotically flat under the current density $\rho_k$.

\begin{restatable}%
    {theorem}{selfcalibration}\label{thm:5}
    Consider the ideal Newton Matching stage~\eqref{eq:38}, where the sequences of densities and stepsizes are $\{\rho_k\}_{k=0}^\infty\subset\pdfspace$ and $\{\eta_k\}_{k=0}^\infty\subset(0,\tau]$, respectively. Then, the reverse KL converges:
    \begin{equation}\label{eq:48}
        K_\infty:=\lim_{k\to\infty}\KL{\rho_k}{\pi}\in[0,\KL{\rho_0}{\pi}].
    \end{equation}
    If the stepsizes are non-summable, i.e., $\sum_{k=0}^{\infty}\eta_k=\infty$, then, as $k\to\infty$,
    \begin{equation}\label{eq:49}
        \E_{X\sim\rho_k}\left[\abs{\log\frac{\rho_k(X)}{\pi(X)}-K_\infty}\right]\to0.
    \end{equation}
    Equivalently,
    \[
        \log\frac{\rho_k}{\pi}\xlongrightarrow{L^1(\rho_k)}K_\infty.
    \]
\end{restatable}

Theorem~\ref{thm:5} shows more than convergence of the scalar reverse-KL values. It states that Newton Matching becomes asymptotically self-consistent on the regions visited by the current sampler. This is a genuine stationarity property, but it is on-policy: it does not by itself certify that the current density continues to cover every region carrying target mass. The following remark separates these two notions.

\begin{remark}
    If a positive density $\rho\in\pdfspace$ satisfies
    \[
        \log\frac{\rho}{\pi}\equiv\const,
    \]
    then normalization forces $\rho=\pi$. Exact flatness of the log-density ratio therefore identifies the target. Theorem~\ref{thm:5}, however, establishes asymptotic flatness in the changing on-policy space $L^1(\rho_k)$. A sequence may consequently approach the boundary of $\pdfspace$ while becoming calibrated only on the mass that it continues to retain. In particular, its limiting behavior may approach the conditional target
    \[
        \pi_A(x)
        :=
        \begin{cases}
            \dfrac{\pi(x)}{\int_A\pi(x)\,\dd x},
            &x\in A,\\[2mm]
            0,
            &\text{otherwise},
        \end{cases}
    \]
    for a measurable set $A\subset\R^d$ with $\int_A\pi(x)\,\dd x>0$. On the covered region $A$, the residual is perfectly flat; nevertheless, all target mass on $A^{\mathrm c}$ is missed. Although every iterate $\rho_k$ belongs to $\pdfspace$, the limiting density $\pi_A$ lies outside $\pdfspace$ because it is not strictly positive.

    The distinction above is familiar in convergence analyses of iterative algorithms whose progress is measured by KL divergence. Monotone KL descent controls the scalar objective values, but it does not by itself imply that the density iterates converge to the target: one must also rule out loss of target mass and ensure that every accumulation point belongs to the solution set. Classical global-convergence theory makes this separation explicit by combining descent with precompactness of the iterates, closedness of the update map, and conditions identifying accumulation points as solutions \citep[Section 4.5]{zangwill1969nonlinear}. These additional properties are not automatic on the infinite-dimensional density manifold considered here. Similarly, classical $I$-divergence projection theory derives convergence from KL identities only under accompanying feasibility, support, or absolute-continuity conditions \citep{csiszar1975divergence,csiszar1984information}; analogous assumptions appear in modern analyses of Sinkhorn and Schr\"odinger-bridge iterations \citep[Propositions 4 and 5]{de2021diffusion}.
\end{remark}

\subsubsection{Global KL Convergence under Mild Conditions}\label{sub2sec:9}

Self-calibration identifies the unconditional asymptotic stationarity of Newton Matching under the current density. To identify this stationary limit with the target, it remains to rule out loss of target mass. We give two complementary mild conditions. The first is an anti-collapse condition expressed through bounded forward KL. The second requires the target-averaged path dissipation to vanish asymptotically. Either condition forces the limiting reverse KL value to be zero, while the second additionally yields convergence of the forward KL. Before stating these results, we summarize the corresponding notions of convergence for terminal densities and canonical velocity fields.

Consider two densities $\rho,q\in\pdfspace$. Their total variation distance is
\[
    \TV{\rho}{q}:=\frac12\int_{\R^d}\abs{\rho(x)-q(x)}\,\dd x,
\]
and Pinsker's inequality \citep{kullback1967lower} gives
\[
    \TV{\rho}{q}\leq\sqrt{\frac12\min\left\{\KL{\rho}{q},\KL{q}{\rho}\right\}}.
\]

Now consider a sequence $\{\rho_k\}_{k=0}^\infty\subset\pdfspace$ and a target $\pi\in\pdfspace$. Since convergence in total variation implies weak convergence, convergence of either the reverse or forward KL divergence to zero already implies $\rho_k\weakto\pi$. Moreover, Corollary~\ref{cor:8} in Appendix~\ref{subsec:44} identifies the following properties:
\[
    \KL{\rho_k}{\pi}
    =
    \int_{0}^{1}\frac{1}{\kappa_t}\E_{X_t\sim p_t^{\rho_k}}\left[\norm[2]{v_t^{\rho_k}(X_t)-v_t^{\pi}(X_t)}^2\right]\dd t,
\]
and
\[
    \KL{\pi}{\rho_k}
    =
    \int_{0}^{1}\frac{1}{\kappa_t}\E_{X_t\sim p_t^{\pi}}\left[\norm[2]{v_t^{\rho_k}(X_t)-v_t^{\pi}(X_t)}^2\right]\dd t.
\]
Thus, reverse-KL convergence yields vanishing velocity discrepancy averaged along the current marginal paths $p_t^{\rho_k}$, whereas forward-KL convergence yields the analogous statement along the target marginal path $p_t^\pi$. In either case, it also implies convergence of the terminal densities in total variation and hence weakly. The KL divergences therefore simultaneously control distributional convergence and a corresponding integrated notion of canonical-velocity convergence.

In the following, we consider the ideal Newton Matching iteration~\eqref{eq:38}, with density sequence $\{\rho_k\}_{k=0}^\infty\subset\pdfspace$ and stepsizes $\{\eta_k\}_{k=0}^\infty\subset(0,\tau]$. We assume that the stepsizes are non-summable:
\[
    \sum_{k=0}^{\infty}\eta_k=\infty.
\]
Accordingly, we use the forward and reverse KL divergences as the primary convergence quantities; whenever either is shown to converge to zero, the corresponding velocity-field and distributional convergence follow from the identities above.

The following theorem proves convergence of the reverse KL under the anti-collapse condition~\eqref{eq:50}.

\begin{restatable}
    {theorem}{klconvergenceanticollapse}\label{thm:6}
    Using the notation and setting of Theorem~\ref{thm:5}, assume that the forward KL is bounded, i.e.,
    \begin{equation}\label{eq:50}
        \sup_{k}\KL{\pi}{\rho_k}<\infty.
    \end{equation}
    Then, as $k\to\infty$, both the terminal densities and the canonical velocity fields converge:
    \[
        \KL{\rho_k}{\pi}=\int_{0}^{1}\frac{1}{\kappa_t}\E_{X_t\sim p_t^{\rho_k}}\left[\norm[2]{v_t^{\rho_k}(X_t)-v_t^{\pi}(X_t)}^2\right]\dd t\to0.
    \]
\end{restatable}

\begin{remark}
    \label{remark:6}
    Theorem~\ref{thm:6} also suggests an ideal forward-KL safeguard. At the population level, Corollary~\ref{cor:2} shows that a sufficiently small stepsize $\eta_k$ makes $\KL{\pi}{\rho_k}$ nonincreasing. Starting from finite forward KL, such a choice maintains
    \[
        \sup_k\KL{\pi}{\rho_k}
        \leq
        \KL{\pi}{\rho_0}
        <
        \infty.
    \]
    To retain the self-calibration conclusion of Theorem~\ref{thm:5}, however, the accepted stepsizes must remain non-summable.
\end{remark}

Inspired by Remark~\ref{remark:5}, we develop the second condition~\eqref{eq:51} to control the nonlinear path dissipation introduced by the finite-stepsize canonical retraction. It is complementary to the anti-collapse condition~\eqref{eq:50}: rather than assuming a uniform bound on the forward KL sequence, it assumes finite initial forward KL and requires the target-averaged dissipation to vanish asymptotically. Under this condition, both forward and reverse KL divergences converge to zero.

\begin{restatable}
    {theorem}{klconvergencedissipationvanishing}\label{thm:7}
    Using the notation and setting of Theorem~\ref{thm:5}, assume that $\KL{\pi}{\rho_0}<\infty$ and the dissipation vanishes, i.e.,
    \begin{equation}\label{eq:51}
        \lim_{k\to\infty}\eta_k\E_{X\sim \pi}\left[\int_{0}^{1}\kappa_t\norm[2]{\nabla V_t^{\rho_k}[\tilde r^{\rho_k}]\left(\flow_{1\to t}^{{\rho_k},\eta_k,\tilde r^{\rho_k}}(X)\right)}^2\dd t\right]=0.
    \end{equation}
    Then, as $k\to\infty$, both the terminal densities and the canonical velocity fields converge:
    \[
        \begin{aligned}
            &\KL{\rho_k}{\pi}=\int_{0}^{1}\frac{1}{\kappa_t}\E_{X_t\sim p_t^{\rho_k}}\left[\norm[2]{v_t^{\rho_k}(X_t)-v_t^{\pi}(X_t)}^2\right]\dd t\to0,\\
            &\KL{\pi}{\rho_k}=\int_{0}^{1}\frac{1}{\kappa_t}\E_{X_t\sim p_t^{\pi}}\left[\norm[2]{v_t^{\rho_k}(X_t)-v_t^{\pi}(X_t)}^2\right]\dd t\to0.
        \end{aligned}
    \]
\end{restatable}

In summary, Theorems~\ref{thm:6} and~\ref{thm:7} provide two complementary routes from self-calibration to KL convergence. Bounded forward KL rules out collapse and yields reverse-KL convergence, whereas vanishing target-averaged dissipation yields convergence in both forward and reverse KL.

\subsection{Local Quadratic Convergence and Continuation Method}
\label{subsec:13}

The global convergence results above describe the long-run behavior of Newton Matching from general initial points under mild target-identification conditions. We now turn to the complementary local regime. As expected for Newton's method, the full-step iteration converges quadratically once the current density lies sufficiently close to the target. Under the compatible smooth realization in Assumption~\ref{assume:1}, Section~\ref{sub2sec:10} states the local quadratic convergence rate for both terminal densities and canonical velocity fields. As an optional way to exploit this rapid local convergence, Section~\ref{sub2sec:11} develops a continuation scheme designed to keep successive stages within the quadratic-convergence regime as the inverse temperature increases.

\subsubsection{Local Quadratic Convergence}\label{sub2sec:10}

Here, we denote $\pi:=\pi_{\mu,\tau,r}$ and take the full stepsize $\eta_{k}=\tau$ in~\eqref{eq:43}.
All proofs are deferred to Appendix~\ref{sub2sec:25}.

For the local convergence analysis, we equip the density manifold $\pdfspace$ with a metric 
\[
d_\pdfspace:\pdfspace\times\pdfspace\to[0,\infty)
\] induced by the norm of its model Banach space through a global chart. The detailed construction is deferred to Appendix~\ref{sub2sec:25}. We also transport $d_\pdfspace$ to the canonical manifold $\velmancan$ by
\begin{equation}\label{eq:52}
d_{\velmancan}:\velmancan\times\velmancan\to[0,\infty),\qquad
d_{\velmancan}(v^{(1)},v^{(2)}):=d_\pdfspace(\terminalmap(v^{(1)}),\terminalmap(v^{(2)})).
\end{equation}
In particular, we verify that $d_\pdfspace$ and $d_{\velmancan}$ are valid metrics on $\pdfspace$ and $\velmancan$, respectively.

\begin{restatable}
    {proposition}{metricondensitymanifold}\label{prop:10}
    The map $d_\pdfspace:\pdfspace\times\pdfspace\to[0,\infty)$ is symmetric and definite, and satisfies the triangle inequality. In other words, $d_\pdfspace$ is a metric on the density manifold $\pdfspace$. Furthermore, $d_\pdfspace$ induces exactly the Banach-manifold topology on $\pdfspace$.
    
    Consequently, $d_{\velmancan}:\velmancan\times\velmancan\to[0,\infty)$ is a metric on the canonical manifold $\velmancan$ and induces exactly the Banach-manifold topology on $\velmancan$.
\end{restatable}

Under these metrics, full-step Newton Matching exhibits the characteristic local behavior of Newton's method: once the current density lies within a sufficiently small neighborhood of the target, a single ideal stage reduces the error to at most a fixed constant times its square. Since the metric on $\velmancan$ is transported from $\pdfspace$ through the canonical diffeomorphism, the same quadratic rate holds for the corresponding canonical velocity fields.

\begin{restatable}
    {theorem}{localquadraticconvergence}\label{thm:8}
    We take the full stepsize $\eta=\tau$. Then, there exist a radius $R_\pi\in(0,\infty)$ and a constant $C_\pi\in(0,\infty)$ such that for all $\rho\in\pdfspace$ satisfying $d_{\pdfspace}(\rho,\pi)<R_\pi$, we have
    \begin{equation}\label{eq:53}
        d_{\pdfspace}(q,\pi)\leq C_\pi d_{\pdfspace}(\rho,\pi)^2,
    \end{equation}
    where $q=\terminalmap\left(v^\rho+\tau\Gamma^{\rho,\tilde{r}^\rho}\right)\in\pdfspace$ is the updated density obtained from $\rho$ after one step with the full stepsize $\eta=\tau$.
    
    Consequently, if the initial point $\rho_0$ satisfies $d_{\pdfspace}(\rho_0,\pi)<\min\left\{R_\pi,\frac{1}{C_\pi}\right\}$, then under the full stepsize $\eta_k=\tau$ for all $k$, we always have
    \begin{equation}\label{eq:54}
        d_{\pdfspace}(\rho_k,\pi)<\min\left\{R_\pi,\frac{1}{C_\pi}\right\},\qquad d_{\pdfspace}(\rho_{k+1},\pi)\leq C_\pi d_{\pdfspace}(\rho_k,\pi)^2.
    \end{equation}
    Therefore, both the terminal densities $\rho_k$ and the canonical velocity fields $v^{\rho_k}:=\canonicalmap(\rho_k)$ converge quadratically to their targets, i.e., $\pi$ and $v^\pi:=\canonicalmap(\pi)$, respectively:
    \[
        C_\pi d_{\pdfspace}(\rho_k,\pi)=C_\pi d_{\velmancan}(v^{\rho_k},v^\pi)\leq \left(C_\pi d_{\pdfspace}(\rho_0,\pi)\right)^{2^k}\to0.
    \]
\end{restatable}

\subsubsection{Continuation Method}\label{sub2sec:11}

Motivated by the local quadratic convergence established above, we next develop an optional continuation method along the inverse-temperature path. Note that, under the mild target-identification conditions of Section~\ref{subsec:12}, Newton Matching can be applied directly to the final target $\pi_{\mu,\tau,r}$ and converges globally without introducing intermediate temperatures. The continuation method provides an optional way to exploit this rapid local convergence by choosing successive targets close enough that each stage begins within the quadratic-convergence neighborhood of its target.

Let
\[
    \pi_s(x)\propto\mu(x)e^{s r(x)}
\]
and choose a finite grid of the inverse temperature
\[
    0\leq\tau_0<\tau_1<\cdots<\tau_n=\tau.
\]
Rather than solving directly for $\pi_\tau$ from an arbitrary initialization, the $i$th continuation stage uses the previous target $\pi_{\tau_{i-1}}$ as a warm start for the nearby target $\pi_{\tau_i}$. Since
\[
    \pi_{\tau_i}(x)
    \propto
    \pi_{\tau_{i-1}}(x)
    e^{(\tau_i-\tau_{i-1})r(x)},
\]
this stage is itself a Newton Matching problem with reference density $\pi_{\tau_{i-1}}$, target $\pi_{\tau_i}$, inverse-temperature increment
\[
    \Delta_i:=\tau_i-\tau_{i-1},
\]
and full stepsize $\eta_k^{(i)}=\Delta_i$. It is expected that a sufficiently fine grid ensures that each warm start lies inside the local quadratic-convergence neighborhood of the next target. If $\mu$ is a density and $\pi_0=\mu\in\pdfspace$, we may take $\tau_0=0$. If $\mu=1$, we instead choose a positive $\tau_0$ such that $\pi_s\in\pdfspace$ for every $s\in[\tau_0,\tau]$.

The following proposition shows that a single quadratic-convergence constant and a single upper bound on the grid spacing suffice for all continuation stages along the inverse-temperature interval. The proof is deferred to Appendix~\ref{sub2sec:26}.

\begin{restatable}
    {proposition}{continuationmethod}\label{prop:11}
    Assume that $\forall s\in[\tau_0,\tau]$, $\pi_s:=\pi_{\mu,s,r}\in\pdfspace$, where $\tau_0=0$ is allowed when $\mu$ is a normalized density and $\tau_0>0$ is required when $\mu=1$. There exist a positive constant $C\in(0,\infty)$ and an admissible grid size $\Delta_{\max}\in(0,\infty)$, both depending on $\mu,\tau_0,\tau,r$, such that for any finite grid $0\leq\tau_0<\tau_1<\cdots<\tau_n=\tau$ satisfying $\Delta_i:=\tau_i-\tau_{i-1}\leq\Delta_{\max}$ for all $i$, Newton Matching with full stepsize $\eta_k^{(i)}=\Delta_i$ satisfies
    \[
        d_{\pdfspace}\left(\rho_{k+1}^{(i)},\pi^{(i)}\right)
        \leq
        C d_{\pdfspace}\left(\rho_{k}^{(i)},\pi^{(i)}\right)^2,
        \qquad
        d_{\pdfspace}\left(\rho_{0}^{(i)},\pi^{(i)}\right)<\frac{1}{C}.
    \]
    Here, the target is $\pi^{(i)}:=\pi_{\tau_i}$ and the initial density is $\rho_0^{(i)}:=\pi^{(i-1)}$.
    
    Consequently, for every continuation stage $i$, both the terminal densities $\rho_k^{(i)}$ and the canonical velocity fields $v^{\rho_k^{(i)}}:=\canonicalmap\left(\rho_k^{(i)}\right)$ converge quadratically to their targets, i.e., $\pi^{(i)}$ and $v^{\pi^{(i)}}:=\canonicalmap\left(\pi^{(i)}\right)$, respectively:
    \[
    Cd_{\pdfspace}\left(\rho_{k}^{(i)},\pi^{(i)}\right)=Cd_{\velmancan}\left(v^{\rho_k^{(i)}},v^{\pi^{(i)}}\right)\leq \left(Cd_{\pdfspace}\left(\rho_{0}^{(i)},\pi^{(i)}\right)\right)^{2^k}\to0.
    \]
\end{restatable}

Together, Sections~\ref{subsec:12} and~\ref{subsec:13} establish global and local convergence guarantees for Newton Matching. Section~\ref{sec:5} next turns from the abstract iteration~\eqref{eq:38} to its computational realization by deriving population-exact regression objectives for the tangential update and canonicalization that retain the scalable matching structure of diffusion and flow training.

\section{Exact Newton Matching}\label{sec:5}

The preceding sections characterize the abstract Newton Matching iteration and establish its convergence properties. Given a canonical velocity field \(v^\rho=\canonicalmap(\rho)\in\velmancan\) and a stepsize \(\eta\in(0,\tau]\), one exact stage of Newton Matching consists of two operations. First, the tangential update constructs the ambient velocity field
\begin{equation}
    \bar v_t(x_t)
    =
    v_t^\rho(x_t)+\eta\Gamma_t^{\rho,\tilde r^\rho}(x_t),
    \label{eq:55}
\end{equation}
where
\begin{equation*}
    \tilde r^\rho(x)
    =
    r(x)-\frac{1}{\tau}\log\frac{\rho(x)}{\mu(x)},
    \qquad
    \Gamma_t^{\rho,\tilde r^\rho}(x_t)
    =
    \Cov_{X_1\sim p_{1|t}^\rho(\cdot|x_t)}
    \left(
        v_{t|1}(x_t|X_1),
        \tilde r^\rho(X_1)
    \right).
\end{equation*}
Second, the canonicalization produces a new canonical velocity field \(v^q=\canonicalmap(q)\in\velmancan\), whose terminal density coincides with that induced by \(\bar v\), i.e., \(q=\terminalmap(v^q)=\terminalmap(\bar v)\). The next Newton Matching stage then proceeds with \(v^q\).

In this section, we present population-exact implementations of both the tangential update and canonicalization.
We denote the neural velocity field being trained by \(v^\theta\), where \(\theta\) represents its network parameters. To distinguish it from the trainable field, we refer to \(v^\rho\) as the \emph{anchor}, emphasizing that \(v^\rho\) remains fixed throughout each tangential-update stage and is used to construct the training objectives, including generating samples, estimating the required quantities, and evaluating the relevant gradients. Consequently, all anchor-dependent objects, such as \(p_t^\rho\) and \(p_{1|t}^\rho\), are also fixed during the training of \(v^\theta\) at one tangential-update stage. In practice, the anchor \(v^\rho\) is also represented by a neural network. It is therefore natural to initialize \(v^\theta\) with the parameters of \(v^\rho\) and subsequently optimize it using the matching losses introduced later. Throughout this optimization, the losses are differentiated only with respect to the trainable parameters \(\theta\).

Sections~\ref{subsec:14}--\ref{subsec:18} develop population-exact realizations of the tangential update. First, Section~\ref{subsec:14} presents two equivalent conditional-expectation representations of the tangent vector \(\Gamma^{\rho,\tilde r^\rho}\): a covariance form and a gradient form. These representations form the basis of the regression targets used in the subsequent losses. Then, Section~\ref{subsec:15} studies the sampling order used to construct the regression pair \((X_t,X_1)\). The forward construction first samples \(X_1\) from the anchor density and then applies the noising kernel to obtain \(X_t\), whereas the reverse construction first samples \(X_t\) and then draws \(X_1\) from the corresponding anchor posterior. Both constructions satisfy the posterior requirement \(X_1|X_t\sim p_{1|t}^\rho(\cdot|X_t)\) and therefore support objectives for scalable regression with the same minimizer \(\bar v=v^\rho+\eta\Gamma^{\rho,\tilde r^\rho}\). Together, Sections~\ref{subsec:14} and~\ref{subsec:15} separate two independent design choices: the tangent representation, which may be covariance or gradient, and the sampling construction, which may be forward or reverse. Sections~\ref{subsec:16} and~\ref{subsec:17} then develop the covariance-form and gradient-form losses under both constructions, respectively, with particular attention to the nontrivial computations required to evaluate \(\tilde r^\rho\) or its gradient \(\nabla\tilde r^\rho\). Section~\ref{subsec:18} unifies these forms through Stein control variates: all resulting targets have the same conditional mean and hence the same population minimizer, while differing in their sample-wise variance. Table~\ref{tab:1} summarizes representative losses for the tangential update.
To explicitly illustrate the usage of multi-time supervision for enhancing sample efficiency, Section~\ref{subsec:19} shows how regression pairs at multiple supervision times can be constructed from a single endpoint or path sample, with the associated endpoint-level or path-level computations shared across those pairs.
Finally, Section~\ref{subsec:21} presents the training procedures for canonicalization, completing one exact Newton Matching stage.

\subsection{Tangent Representations}\label{subsec:14}

The tangential update in~\eqref{eq:55} is determined by the tangent vector
\begin{equation*}
    \Gamma_t^{\rho,\tilde r^\rho}(x_t)
    =
    \Cov_{X_1\sim p_{1|t}^\rho(\cdot|x_t)}
    \left(
    v_{t|1}(x_t|X_1),
    \tilde r^\rho(X_1)
    \right).
\end{equation*}
Since the canonical anchor satisfies \(v_t^\rho(x_t)=\E_{X_1\sim p_{1|t}^\rho(\cdot|x_t)}[v_{t|1}(x_t|X_1)]\), we have
\begin{equation}
    \Gamma_t^{\rho,\tilde r^\rho}(x_t)
    = \E_{X_1\sim p_{1|t}^\rho(\cdot|x_t)}\left[
    \left(v_{t|1}(x_t|X_1)-v_t^\rho(x_t)\right)
    \left(\tilde r^\rho(X_1)-B_t(x_t)\right)\right],
    \label{eq:56}
\end{equation}
where \(B_t(x_t)\in\R\) is any regular baseline that depends only on \((t,x_t)\). This conditional-expectation structure naturally suggests a CFM-style loss. As we show later, exploiting this structure yields the \emph{covariance form} of the tangential-update loss for exact Newton Matching.

On the other hand, recall from Proposition~\ref{prop:2} that the tangent vector \(\Gamma^{\rho,\tilde r^\rho}\) can also be written as the gradient of a posterior value:
\begin{equation*}
\Gamma_t^{\rho,\tilde r^\rho}(x_t)
=
\kappa_t \nabla_{x_t} V_t^\rho[\tilde r^\rho](x_t),
\end{equation*}
where
\begin{equation*}
V_t^\rho[\tilde r^\rho](x_t)
=
\E_{X_1 \sim p_{1|t}^\rho(\cdot|x_t)}
\bigl[\tilde r^\rho(X_1)\bigr]
\end{equation*}
denotes the expected regularized reward under the anchor posterior. However, the gradient lies outside the conditional expectation and is therefore not directly amenable to a CFM-style objective.
Ideally, we would move the gradient inside the expectation, thereby recovering the conditional-expectation structure needed to construct such an objective.
This interchange is more subtle than it may initially appear, since the anchor posterior itself depends on \(x_t\).
In particular, differentiating the conditional expectation with respect to \(x_t\) does not simply yield the conditional expectation of the regularized-reward gradient:
\begin{equation*}
\nabla_{x_t}
\E_{X_1 \sim p_{1|t}^\rho(\cdot|x_t)}
\bigl[\tilde r^\rho(X_1)\bigr]
\neq
\E_{X_1 \sim p_{1|t}^\rho(\cdot|x_t)}
\bigl[\nabla_{X_1}\tilde r^\rho(X_1)\bigr]\text{ in general}.
\end{equation*}

Fortunately, it is still possible to move the gradient operator inside the conditional expectation through a careful reformulation. To motivate the construction, we first recall the Gaussian Stein identity. Let \(X \sim \Normal(m,\Sigma)\) be a $d$-dimensional Gaussian random variable. Then, for every sufficiently regular test function \(f\),
\begin{equation*}
    \Cov_{X\sim\Normal(m,\Sigma)}
    \left(X,f(X)\right)
    =
    \E_{X\sim\Normal(m,\Sigma)}
    \left[\Sigma\nabla f(X)\right].
\end{equation*}
This identity can be obtained via integration by parts.

Our setting is more involved, as the anchor posterior
\(p_{1|t}^\rho(\cdot|x_t)\) is generally non-Gaussian. Nevertheless, an analogous integration-by-parts identity can be obtained via the \emph{Langevin Stein operator}. To tailor our construction to the posterior-covariance structure from \(\Gamma_t^{\rho,\tilde r^\rho}\), we further introduce a \emph{posterior Stein kernel} based on Langevin Stein operators. For fixed \((t,x_t)\), let the matrix field
\[
    \Lambda_t^\rho(\cdot|x_t)
    \colon \mathbb{R}^d \to \mathbb{R}^{d\times d}
\]
be a posterior Stein kernel associated with
\(p_{1|t}^\rho(\cdot|x_t)\). Then, for every sufficiently regular scalar-valued function \(f\), it satisfies the covariance--gradient identity
\begin{equation}
    \Cov_{X_1 \sim p_{1|t}^\rho(\cdot|x_t)}
    \left(
        X_1,
        f(X_1)
    \right)
    =
    \E_{X_1 \sim p_{1|t}^\rho(\cdot|x_t)}
    \left[
        \Lambda_t^\rho(X_1|x_t)^\top
        \nabla f(X_1)
    \right].
    \label{eq:57}
\end{equation}
Detailed results on Langevin Stein operators and posterior Stein kernels are provided in Appendix~\ref{app:5}.

To apply this identity to the tangent vector \(\Gamma^{\rho,\tilde r^\rho}\), first observe that the conditional velocity
\(v_{t|1}(x_t|x_1)\) is affine in the endpoint \(x_1\):
\begin{equation*}
    v_{t|1}(x_t|x_1)
    =
    \frac{\dot{\beta}_t}{\beta_t}x_t
    +
    \frac{\alpha_t\kappa_t}{\beta_t^2}x_1.
\end{equation*}
Conditioning on \(x_t\), the first term is independent of \(x_1\) and therefore does not contribute to the covariance. It follows that
\begin{align}
    \Gamma_t^{\rho,\tilde r^\rho}(x_t)
    &=
    \frac{\alpha_t\kappa_t}{\beta_t^2}
    \Cov_{X_1 \sim p_{1|t}^\rho(\cdot|x_t)}
    \left(
        X_1,
        \tilde r^\rho(X_1)
    \right).
    \label{eq:58}
\end{align}
Applying~\eqref{eq:57} with
\(f=\tilde r^\rho\) to~\eqref{eq:58} yields
\begin{align}
    \Gamma_t^{\rho,\tilde r^\rho}(x_t)
    &=
    \frac{\alpha_t\kappa_t}{\beta_t^2}
    \E_{X_1 \sim p_{1|t}^\rho(\cdot|x_t)}
    \left[
        \Lambda_t^\rho(X_1|x_t)^\top
        \nabla \tilde r^\rho(X_1)
    \right].
    \label{eq:59}
\end{align}
This recovers the desired conditional-expectation structure in terms of the regularized-reward gradient. As we show later, this representation leads to the \emph{gradient form} of the tangential-update loss for exact Newton Matching.

Equations~\eqref{eq:56} and~\eqref{eq:59} provide two exact representations of the same tangent vector $\Gamma^{\rho,\tilde r^\rho}$. The former requires a scalar evaluation of \(\tilde r^\rho(X_1)\), whereas the latter requires evaluating the action of a posterior Stein kernel on \(\nabla \tilde r^\rho(X_1)\). As the following subsections show, these two representations give rise to distinct loss functions for tangential-update training and require different computational techniques for their implementation.

\subsection{Forward and Reverse Constructions}\label{subsec:15}

The conditional-expectation representations of the tangent vector \(\Gamma^{\rho,\tilde r^\rho}\) are particularly useful for scalable training, since squared-loss regression allows conditional expectations to be recovered from sample-wise targets. Specifically, let \((X,Y)\) be a pair of random variables, and suppose that the goal is to learn a function \(f^\theta\) approximating \(\E_{Y\sim p_{Y|X}(\cdot|x)}\!\left[g(x,Y)\right]\). Rather than evaluating this conditional expectation explicitly, we can use \(g(X,Y)\) directly as a regression target and minimize
\begin{equation*}
    \E_{(X,Y)\sim p_{X,Y}}\!\left[\left\|f^\theta(X)-g(X,Y)\right\|_2^2\right],
\end{equation*}
provided that the joint density \(p_{X,Y}\) used to generate the training samples induces the required conditional density \(p_{Y|X}(y|x)\).

In our case, \(X=X_t\) and \(Y=X_1\). We therefore do not need to evaluate the tangent vector explicitly during training, provided that we can generate pairs \((X_t,X_1)\) from a joint distribution \(\Xi_{t,1}^\rho\) satisfying
\begin{equation}
(X_t,X_1)\sim\Xi_{t,1}^\rho
\quad\Rightarrow\quad
X_1|X_t\sim p_{1|t}^\rho(\cdot|X_t).
\label{eq:60}
\end{equation}
We use \(\Xi_{t,1}^\rho\) as the generic notation for any joint distribution satisfying~\eqref{eq:60}. Its \(X_t\)-marginal is positive, and may be implicitly defined by the chosen construction.

Before presenting the matching losses for the tangential update, we describe two complementary constructions of \(\Xi_{t,1}^\rho\). The \emph{forward construction} first generates the endpoint \(X_1\) and then produces the noisy state \(X_t\), following the order used in the standard CFM loss. In contrast, the \emph{reverse construction} first generates \(X_t\) and then samples \(X_1\) conditionally. This construction has a Bayesian flavor: \(X_t\) is treated as observed evidence, and \(X_1\) is inferred from the anchor posterior.

\subsubsection{Forward Construction}
The forward approach proceeds by
\begin{equation}
    X_1 \sim \rho,
    \qquad
    X_0 \sim p_0,
    \qquad
    X_t = \alpha_t X_1 + \beta_t X_0,
    \label{eq:61}
\end{equation}
where \(X_0\) and \(X_1\) are sampled independently. 

Conditioning on \(X_1=x_1\), the interpolant in~\eqref{eq:61} follows the conditional probability path
\begin{equation*}
    X_t| X_1
    \sim
    p_{t|1}(\cdot|X_1).
\end{equation*}
Combining this conditional density with the endpoint marginal \(X_1\sim\rho\), the joint density of \((X_t,X_1)\) is
\begin{equation*}
    p_{t,1}^\rho(x_t,x_1)
    =
    \rho(x_1)p_{t|1}(x_t|x_1).
\end{equation*}
Marginalizing over \(x_1\) and applying Bayes' rule therefore gives
\begin{equation}
    X_t\sim p_t^\rho,
    \qquad
    X_1| X_t
    \sim
    p_{1|t}^\rho(\cdot|X_t).
    \label{eq:62}
\end{equation}
Thus, \(\Xi_{t,1}^\rho\) satisfies~\eqref{eq:60} and is given by
\[
\Xi_{t,1}^\rho(x_t,x_1)=p_t^\rho(x_t) p_{1|t}^\rho(x_1|x_t)\qquad\text{under the forward construction}.
\]

The endpoint \(X_1\) may be generated with any exact sampler whose terminal density is \(\rho\). A broad family of such samplers is given by
\begin{equation*}
    \dd Y_t
    =
    \left(
        v_t^\rho(Y_t)
        +
        \frac{\sigma_t^2}{2}
        \nabla \log p_t^\rho(Y_t)
    \right)\dd t
    +
    \sigma_t \dd W_t,
    \qquad
    Y_0 \sim p_0,
\end{equation*}
which has marginal density \(Y_t\sim p_t^\rho\) for every \(t\in[0,1]\), and hence produces \(X_1:=Y_1\sim\rho\) for any noise schedule \(\sigma_t\geq0\). Using the score--velocity relation
\begin{equation*}
    \nabla \log p_t^\rho(x)
    =
    \frac{1}{\kappa_t}
    \left(
        v_t^\rho(x)
        -
        \frac{\dot\alpha_t}{\alpha_t}x
    \right),
\end{equation*}
we can equivalently write
\begin{equation}
    \dd Y_t
    =
    \left[
        \left(
            1+\frac{\sigma_t^2}{2\kappa_t}
        \right)
        v_t^\rho(Y_t)
        -
        \frac{\sigma_t^2}{2\kappa_t}
        \frac{\dot\alpha_t}{\alpha_t}Y_t
    \right]\dd t
    +
    \sigma_t\dd W_t,
    \qquad
    Y_0\sim p_0.
    \label{eq:63}
\end{equation}
When \(\sigma_t=0\), this SDE reduces to the standard probability flow ODE. Therefore, one key advantage of the forward construction is its compatibility with a wide range of samplers. For example, one may use an ODE sampler to potentially reduce the number of function evaluations (NFEs). Alternatively, since the neural-network anchor \(v^\rho\) is generally imperfect in practice, one may use an SDE sampler with an empirically tuned noise schedule \(\sigma_t\) to encourage better exploration of the state space.

\subsubsection{Reverse Construction}
The reverse approach is motivated by the observation that, for training, the only requirement on the pair \((X_t,X_1)\) is that it satisfies~\eqref{eq:60}; the marginal density of \(X_t\) does not affect the global minimizer of the squared-loss regression. Therefore, we can sample in the reverse order:
\begin{equation*}
    X_t \sim \hat p_t, \qquad X_1 \sim p_{1|t}^\rho(\cdot|X_t),
\end{equation*}
where \(\hat p_t\) is any positive proposal density on \(\mathbb{R}^d\).

The key remaining challenge is to determine how, given a noisy state \(X_t\), to sample \(X_1 \sim p_{1|t}^\rho(\cdot|X_t)\) in practice. To highlight the subtlety, we first note that the SDE~\eqref{eq:63}, with an arbitrary noise schedule \(\sigma_t\geq 0\), does not in general preserve the anchor posterior. Specifically, if we run the SDE over the time interval \([t,1]\),
\begin{align*}
    &\dd Y_s = \left[\left(1+\frac{\sigma_s^2}{2\kappa_s}\right)v_s^\rho(Y_s)-\frac{\sigma_s^2}{2\kappa_s}\frac{\dot{\alpha}_s}{\alpha_s}Y_s\right]\dd s+\sigma_s\dd W_s, \qquad s\in[t,1],\\
    &Y_t=x_t,
\end{align*}
the resulting terminal state \(Y_1\) does not generally satisfy \(p_{Y_1|Y_t}(\cdot|x_t)=p_{1|t}^\rho(\cdot|x_t)\). The SDE preserves the prescribed marginal densities only when initialized with \(Y_0\sim p_0\), or equivalently, when initialized at time \(t\) with \(Y_t\sim p_t^\rho\). It does not, however, preserve the posterior density when started from an arbitrary fixed state \(Y_t=x_t\).

Fortunately, within the family of SDEs in~\eqref{eq:63}, there is a unique SDE that preserves the posterior \(p_{Y_1|Y_t}(\cdot|x_t)=p_{1|t}^\rho(\cdot|x_t)\) for $t\in(0,1)$. It corresponds to the specific choice of noise schedule \(\sigma_t=\sqrt{2\kappa_t}\). Given a canonical model \(v^\rho\), we define the \emph{posterior-preserving SDE} by
\begin{equation}
    \dd Y_s = b_s^\rho(Y_s)\dd s + \sqrt{2\kappa_s}\dd W_s, \qquad b_s^\rho(x) := 2v_s^\rho(x) - \frac{\dot\alpha_s}{\alpha_s}x, \qquad s\in[t,1].
    \label{eq:64}
\end{equation}
With this SDE, the resulting \(\Xi_{t,1}^\rho\) satisfies~\eqref{eq:60} and is given by
\[
\Xi_{t,1}^\rho(x_t,x_1)=\hat{p}_t(x_t) p_{1|t}^\rho(x_1|x_t)\qquad\text{under the reverse construction}.
\]
The posterior-preserving SDE is an important tool for constructing Newton Matching losses and has several properties useful for both theoretical analysis and computational implementation. We summarize these results in Appendix~\ref{subsec:42}.

The forward construction produces \(X_t\sim p_t^\rho\) by design. In contrast, the reverse construction does not have to choose \(\hat{p}_t = p_t^\rho\). Changing the proposal \(\hat{p}_t\) affects the weighting of the regression problem but not its pointwise conditional target. This allows the proposal to be chosen for practical convenience. For example, one may use an ``off-policy'' approach to obtain \(X_t\): sample \(X_1\) from a buffer, draw \(X_0\sim p_0\), and set \(X_t=\alpha_tX_1+\beta_tX_0\). Whenever the sampling order and the \(X_t\)-marginal are immaterial, we use the generic notation \((X_t,X_1)\sim\Xi_{t,1}^\rho\) to indicate only the posterior requirement \(X_1|X_t\sim p_{1|t}^\rho(\cdot|X_t)\).

Another advantage of the reverse construction is that multiple endpoints can be generated from the same noisy state. By running~\eqref{eq:64} multiple times at a fixed \(X_t\), we obtain
\begin{equation*}
    X_1^{(1)},\cdots,X_1^{(N)} \iidsim p_{1|t}^\rho(\cdot|X_t),
\end{equation*}
where \(\mathrm{i.i.d.}\) denotes independent and identically distributed. Averaging the corresponding targets potentially reduces the variance and improves training stability. In contrast, the forward construction cannot achieve this, since \(X_1\) is sampled first and \(X_t\) is subsequently constructed from it.

\subsection{Covariance Form}\label{subsec:16}

The covariance form is derived from the covariance representation of the tangent vector \(\Gamma^{\rho,\tilde r^\rho}\) in~\eqref{eq:56}. In particular, define the sample-wise target
\begin{align}
    \target_{\rho,\eta,B}^{\mathrm{cov}}(t,X_t,X_1)
    :=&
    v_{t|1}(X_t|X_1)
    +
    \eta
    \left(
        \tilde r^\rho(X_1)-B_t(X_t)
    \right)
    \left(
        v_{t|1}(X_t|X_1)-v_t^\rho(X_t)
    \right),
    \label{eq:65}
\end{align}
where \(B_t(x_t)\) is any scalar baseline that depends only on \((t,x_t)\). Since
\begin{equation*}
    \E_{X_1\sim p_{1|t}^\rho(\cdot|x_t)}
    \left[
        v_{t|1}(x_t|X_1)
    \right]
    =
    v_t^\rho(x_t),
\end{equation*}
the target satisfies
\begin{equation}
    \E_{X_1\sim p_{1|t}^\rho(\cdot|x_t)}
    \left[
        \target_{\rho,\eta,B}^{\mathrm{cov}}(t,x_t,X_1)
    \right]
    =
    v_t^\rho(x_t)
    +
    \eta\Gamma_t^{\rho,\tilde r^\rho}(x_t).
    \label{eq:66}
\end{equation}

Therefore, under the forward construction, the covariance form of the tangential-update loss is given by
\begin{align}
    \calL_{\rho,\eta,B}^{\mathrm{cov},\mathrm{fwd}}(\theta)
    :=&
    \E_{\substack{
        t\sim\U(0,1),\,X_1\sim\rho,\,X_0\sim p_0,\,
        X_t=\alpha_tX_1+\beta_tX_0}}
    \left[
        \norm[2]{
            v_t^\theta(X_t)
            -
            \target_{\rho,\eta,B}^{\mathrm{cov}}(t,X_t,X_1)
        }^2
    \right],
    \label{eq:67}
\end{align}
and under the reverse construction,
\begin{subequations}
    \label{eq:68}
\begin{align}
    \hspace{-1mm}
    \calL_{\rho,\eta,B}^{\mathrm{cov},\mathrm{rev}}(\theta)
    :=&
    \E_{\substack{
        t\sim\U(0,1),\,X_t\sim\hat p_t,\,
        X_1\sim p_{1|t}^\rho(\cdot|X_t)}}
    \left[
        \norm[2]{
            v_t^\theta(X_t)
            -
            \target_{\rho,\eta,B}^{\mathrm{cov}}(t,X_t,X_1)
        }^2
    \right]\\
    =&\E_{\substack{
        t\sim\U(0,1),\,X_t\sim\hat p_t}}
    \left[
        \norm[2]{
            v_t^\theta(X_t)
            -
            \E_{X_1\sim p_{1|t}^\rho(\cdot|X_t)}\left[\target_{\rho,\eta,B}^{\mathrm{cov}}(t,X_t,X_1)\right]
        }^2
    \right] + \const.
    \label{eq:69}
\end{align}
\end{subequations}
In~\eqref{eq:69}, we push the expectation over \(X_1\) inside the squared loss. Since the reverse construction can generate multiple i.i.d. posterior samples of \(X_1\) for a fixed \(X_t\), this expectation can be estimated by averaging over these samples, thereby reducing the variance.

The remaining task is to specify how to compute the regularized reward
\begin{equation*}
    \tilde r^\rho(X_1)
    =
    r(X_1)
    -
    \frac{1}{\tau}
    \log\frac{\rho(X_1)}{\mu(X_1)},
\end{equation*}
where the \(\log\frac{\rho(X_1)}{\mu(X_1)}\) term is generally nontrivial to evaluate.
In the sampling setting, where \(\mu=1\), this term reduces to the log density \(\log\rho(X_1)\). In the fine-tuning setting, where \(\mu=\rho^\base\), it is the log-density ratio \(\log\frac{\rho(X_1)}{\rho^\base(X_1)}\).

We present two methods for computing it exactly: an ODE-based approach that leverages the instantaneous change-of-variables formula for the probability flow ODE, and an SDE-based approach that exploits the stochastic path associated with the posterior-preserving SDE.
Note that ``ODE-based'' and ``SDE-based'' refer to the calculus used to compute the term \(\log\frac{\rho(X_1)}{\mu(X_1)}\), not to the sampling method used to generate the endpoint \(X_1\). Each calculus is compatible with both forward and reverse constructions, thereby yielding a \(2\times 2\) matrix of possible combinations.

\subsubsection{ODE-Based Estimation}
The key idea is to evaluate the log density or log-density ratio at the sampled endpoint via the instantaneous change-of-variables formula. Let \(\flow^{v^\rho}\) denote the ODE flow generated by the anchor \(v^\rho\). We have
\begin{equation}
    \label{eq:70}
    \frac{\dd}{\dd s}
    \log p_s^\rho
    \left(
        \flow_{t\to s}^{v^\rho}(x_t)
    \right)
    =
    -
    \nabla\cdot v_s^\rho
    \left(
        \flow_{t\to s}^{v^\rho}(x_t)
    \right).
\end{equation}
For simplicity, in the case of \(\mu=\rho^\base\), let \(v_t^\base\) denote the base canonical velocity and \(p_t^\base\) its associated marginal probability path. The identity~\eqref{eq:70} holds for the base model upon replacing \(p_t^\rho\) and \(v_t^\rho\) with \(p_t^\base\) and \(v_t^\base\), respectively.

If \(X_1\) is sampled using an ODE (under forward construction), then \(\log\frac{\rho(X_1)}{\mu(X_1)}\) can be computed alongside the forward integration by augmenting the ODE with an additional equation for the log density or log-density ratio. If \(X_1\) is sampled using an SDE, under either the forward or reverse construction, then \(\log\frac{\rho(X_1)}{\mu(X_1)}\) can be computed by integrating a backward ODE for the log density or log-density ratio from the endpoint \(X_1\) to the initial time \(t=0\).

\paragraph{Augmented forward integration.}
For \(\mu=1\), the sampling ODE
\[
\frac{\dd Y_t}{\dd t}=v_t^\rho(Y_t),
\]
can be augmented directly with
\begin{equation}
    \frac{\dd}{\dd t}
    \log p_t^\rho(Y_t)
    =
    -
    \nabla\cdot v_t^\rho(Y_t),
    \qquad
    \log p_0^\rho(Y_0)
    =
    \log p_0(Y_0).
    \label{eq:71}
\end{equation}
Let $X_1:=Y_1$. Then, the terminal value is 
\[
\log\rho(X_1)=\log p_1^\rho(Y_1).
\]

For \(\mu=\rho^\base\), the sampling ODE can be augmented with
\begin{align}
    &\frac{\dd}{\dd t}
    \log
    \frac{
        p_t^\rho(Y_t)
    }{
        p_t^\base(Y_t)
    }\notag\\
    =&
    \nabla\cdot v_t^\base(Y_t)
    -
    \nabla\cdot v_t^\rho(Y_t)
    -
    \left(
        v_t^\rho(Y_t)-v_t^\base(Y_t)
    \right)^\top
    \nabla
    \log p_t^\base(Y_t)
    \notag\\
    =&
    \nabla\cdot
    \left(
        v_t^\base-v_t^\rho
    \right)(Y_t)
    -
    \frac{1}{\kappa_t}
    \left(
        v_t^\rho(Y_t)-v_t^\base(Y_t)
    \right)^\top
    \left(
        v_t^\base(Y_t)
        -
        \frac{\dot\alpha_t}{\alpha_t}Y_t
    \right),
    \label{eq:72}
\end{align}
with initial condition
\begin{equation*}
    \log
    \frac{
        p_0^\rho(Y_0)
    }{
        p_0^\base(Y_0)
    }
    =0.
\end{equation*}
At \(t=1\), setting $X_1:=Y_1$ gives
\[
    \log\frac{\rho(X_1)}{\rho^\base(X_1)}
    =
    \log\frac{p_1^\rho(Y_1)}{p_1^\base(Y_1)}.
\]

\paragraph{Backward integration from a given endpoint.}
If \(X_1=x_1\) has already been generated by an SDE, \(\log\rho(x_1)\) or \(\log \frac{\rho(x_1)}{\rho^\base(x_1)}\) can be evaluated by integrating backward from \(x_1\).
For \(\mu=1\), we have
\begin{equation}
    \log\rho(x_1)
    =
    \log p_0
    \left(
        \flow_{1\to0}^{v^\rho}(x_1)
    \right)
    -
    \int_0^1
    \nabla\cdot v_s^\rho
    \left(
        \flow_{1\to s}^{v^\rho}(x_1)
    \right)
    \dd s.
    \label{eq:73}
\end{equation}

For simplicity, in the case of $\mu=\rho^\base$, we use the shorthand $\Phi^\base:=\Phi^{v^{\base}}$ for the ODE flow generated by $v^\base$. In this case, we have
\begin{align}
    \log\frac{\rho(x_1)}{\rho^\base(x_1)}=\log\frac{p_0\left(\flow_{1\to0}^{v^\rho}(x_1)\right)}{p_0\left(\flow_{1\to0}^{\base}(x_1)\right)}-\int_0^1\left[\nabla\cdot v_s^\rho\left(\flow_{1\to s}^{v^\rho}(x_1)\right)-\nabla\cdot v_s^\base\left(\flow_{1\to s}^{\base}(x_1)\right)\right]\dd s.
    \label{eq:74}
\end{align}
Unlike the augmented forward integration in~\eqref{eq:72}, the above evaluation follows two backward trajectories, one under the anchor ODE and one under the base ODE.

The required divergences may be evaluated exactly or estimated using Hutchinson's trace estimator \citep{hutchinson1990stochastic}:
\begin{equation*}
    \nabla\cdot v_s(x)=\E_{\varepsilon}\left[\varepsilon^\top\nabla v_s(x)\varepsilon\right],
\end{equation*}
where the probe vector \(\varepsilon\) is a random variable in $\R^d$ and satisfies
\begin{equation*}
\E_\varepsilon\left[\varepsilon\varepsilon^\top\right]=I.
\end{equation*}
For example, \(\varepsilon \sim \Normal(0, I)\) is a valid choice.

\subsubsection{SDE-Based Estimation}

The SDE-based method exploits two special properties of the posterior-preserving SDE. First, its generator yields exact pathwise representations of the terminal log density and log-density ratio through It\^o's formula. Second, conditioning on the two endpoints \(X_t\) and \(X_1\), the intervening stochastic path has a universal distribution independent of the drift, allowing the same representations to be used regardless of the sampler employed to generate \(X_1\). The pathwise equalities below use the convention of Appendix~\ref{subsec:40}.

Fix \(t\in(0,1)\) and \(x_t\in\R^d\). Recall from~\eqref{eq:64} that the posterior-preserving SDE initialized at \(Y_t=x_t\) is
\begin{equation*}
    \dd Y_s
    =
    b_s^\rho(Y_s)\dd s
    +
    \sqrt{2\kappa_s}\dd W_s,
    \qquad
    b_s^\rho(x)
    =
    2v_s^\rho(x)
    -
    \frac{\dot\alpha_s}{\alpha_s}x,
    \qquad
    s\in[t,1].
\end{equation*}
Let
\[
    \BY_{[t,1]}
    =
    (Y_s)_{s\in[t,1]}
\]
denote the resulting random path. We write
\begin{equation}
    \BY_{[t,1]}
    \sim
    \mathbb P_{[t,1]|t}^{\rho}
    (\cdot|x_t)
    \label{eq:75}
\end{equation}
to indicate that \(\BY_{[t,1]}\) is generated by the posterior-preserving SDE starting from \(Y_t=x_t\). In particular, its endpoint satisfies
\[
    Y_1
    \sim
    p_{1|t}^\rho(\cdot|x_t).
\]

For a $C^2$ function \(f:\R^d\to\R\), the generator of the posterior-preserving SDE is
\begin{equation*}
    \left(
        \calL_s^\rho f
    \right)(x)
    :=
    b_s^\rho(x)^\top\nabla f(x)
    +
    \kappa_s\Delta f(x).
\end{equation*}
Accordingly, for a sufficiently regular time-dependent function
\(h_s:\R^d\to\R\), It\^o's formula gives
\begin{equation}
    \dd h_s(Y_s)
    =
    \left[
        \partial_s h_s
        +
        \calL_s^\rho h_s
    \right](Y_s)\dd s
    +
    \sqrt{2\kappa_s}
    \nabla h_s(Y_s)^\top
    \dd W_s.
    \label{eq:76}
\end{equation}
We now apply this formula separately to the two cases considered in this paper.

When \(\mu=1\), consider
\begin{equation*}
    h_s(x)
    =
    \log\left(
        \alpha_s^d p_s^\rho(x)
    \right).
\end{equation*}
Since \(\alpha_1=1\) and \(p_1^\rho=\rho\), its terminal value is
\[
    h_1(x)=\log\rho(x).
\]
The generator of the posterior-preserving SDE for \(h_s=\log(\alpha_s^d p_s^\rho)\) admits an analytic expression:
\begin{align*}
    \left[
        \partial_s h_s
        +
        \calL_s^\rho h_s
    \right](x)
    &=
    \frac{1}{\kappa_s}
    \norm[2]{
        v_s^\rho(x)
        -
        \frac{\dot\alpha_s}{\alpha_s}x
    }^2.
\end{align*}
Also, the canonical score--velocity relation gives
\begin{align*}
    \nabla h_s(x)
    &=
    \frac{1}{\kappa_s}
    \left(
        v_s^\rho(x)
        -
        \frac{\dot\alpha_s}{\alpha_s}x
    \right).
\end{align*}
Substituting these identities into~\eqref{eq:76} and integrating from \(t\) to \(1\) gives
\begin{align}
    \log\rho(Y_1)
    =&
    \log\left(\alpha_t^d p_t^\rho(x_t)\right)
    +
    \int_t^1
    \frac{1}{\kappa_s}
    \norm[2]{
        v_s^\rho(Y_s)
        -
        \frac{\dot\alpha_s}{\alpha_s}Y_s
    }^2
    \dd s
    +
    \int_t^1
    \sqrt{
        \frac{2}{\kappa_s}
    }
    \left(
        v_s^\rho(Y_s)
        -
        \frac{\dot\alpha_s}{\alpha_s}Y_s
    \right)^\top
    \dd W_s.
    \label{eq:77}
\end{align}
The detailed derivation is given in Proposition~\ref{prop:37}.

When \(\mu=\rho^\base\), consider
\begin{equation*}
    h_s(x)
    =
    \log
    \frac{
        p_s^\rho(x)
    }{
        p_s^\base(x)
    }.
\end{equation*}
Its terminal value is
\[
    h_1(x)
    =
    \log
    \frac{
        \rho(x)
    }{
        \rho^\base(x)
    }.
\]
Using the canonical score--velocity relations for the anchor and base models, we have
\begin{align*}
    \nabla h_s(x)
    &=
    \frac{1}{\kappa_s}
    \left(
        v_s^\rho(x)-v_s^\base(x)
    \right).
\end{align*}
The generator of the posterior-preserving SDE for \(h_s=\log\frac{p_s^\rho}{p_s^\base}\) admits an analytic expression:
\begin{align*}
    \left(
        \partial_s h_s
        +
        \calL_s^\rho h_s
    \right)(x)
    &=
    \frac{1}{\kappa_s}
    \norm[2]{
        v_s^\rho(x)-v_s^\base(x)
    }^2.
\end{align*}
Applying~\eqref{eq:76} and integrating from \(t\) to \(1\) therefore yields
\begin{align}
    \log
    \frac{
        \rho(Y_1)
    }{
        \rho^\base(Y_1)
    }
    =&
    \log
    \frac{
        p_t^\rho(x_t)
    }{
        p_t^\base(x_t)
    }
    +
    \int_t^1
    \frac{
        \norm[2]{
            v_s^\rho(Y_s)-v_s^\base(Y_s)
        }^2
    }{
        \kappa_s
    }
    \dd s
    \notag\\
    &+
    \int_t^1
    \sqrt{
        \frac{2}{\kappa_s}
    }
    \left(
        v_s^\rho(Y_s)-v_s^\base(Y_s)
    \right)^\top
    \dd W_s.
    \label{eq:78}
\end{align}
The detailed derivation is given in Proposition~\ref{prop:35}.

For simplicity, we write
\[
v_{t}^\mu(x_t):=\begin{cases}
    \frac{\dot\alpha_t}{\alpha_t}x_t,&\mu=1,\\
    v_t^\base(x_t),&\mu=\rho^\base,
\end{cases}
\qquad
p_{t}^\mu(x_t):=\begin{cases}
    \alpha_t^{-d},&\mu=1,\\
    p_{t}^\base(x_t),&\mu=\rho^\base.
\end{cases}
\]
Then,~\eqref{eq:77} and~\eqref{eq:78} can be unified as
\begin{align}
    \log
    \frac{
        \rho(Y_1)
    }{
        \mu(Y_1)
    }
    =&
    \log
    \frac{
        p_t^\rho(x_t)
    }{
        p_t^\mu(x_t)
    }
    +
    \int_t^1
    \frac{
        \norm[2]{
            v_s^\rho(Y_s)-v_s^\mu(Y_s)
        }^2
    }{
        \kappa_s
    }
    \dd s
    +
    \int_t^1
    \sqrt{
        \frac{2}{\kappa_s}
    }
    \left(
        v_s^\rho(Y_s)-v_s^\mu(Y_s)
    \right)^\top
    \dd W_s.
    \label{eq:79}
\end{align}

Observe that the \(\log \frac{p_t^\rho(x_t)}{p_t^\mu(x_t)}\) term in~\eqref{eq:79} depends only on \((t,x_t)\). Since the covariance representation is unchanged when any function of  \((t,x_t)\) is added to the baseline, these terms do not need to be evaluated. After absorbing
\[
    \frac{1}{\tau}\log\frac{p_t^\rho(x_t)}{p_t^\mu(x_t)}
\]
into \(B_t(x_t)\), the same scalar factor can be evaluated as
\begin{align}
    r(Y_1)
    -
    B_t(x_t)
    -
    \frac{1}{\tau}
    \int_t^1
    \frac{
        \norm[2]{
            v_s^\rho(Y_s)-v_s^\mu(Y_s)
        }^2
    }{
        \kappa_s
    }
    \dd s -
    \frac{1}{\tau}
    \int_t^1
    \sqrt{
        \frac{2}{\kappa_s}
    }
    \left(
        v_s^\rho(Y_s)-v_s^\mu(Y_s)
    \right)^\top
    \dd W_s.
    \label{eq:80}
\end{align}

It remains to demonstrate how the stochastic path~\eqref{eq:75} is generated for the forward and reverse constructions. Under the reverse construction, this path is obtained directly. Starting from the given noisy state \(X_t\), the posterior-preserving SDE directly generates
\begin{equation*}
    \BY_{[t,1]}
    \sim
    \mathbb P_{[t,1]|t}^{\rho}
    (\cdot|X_t),
    \qquad
    Y_t= X_t.
\end{equation*}
The path and Brownian increments required by~\eqref{eq:80} are therefore obtained from the same simulation.

\begin{algorithm}[!p]
    \caption{Tangential Update: Exact Realization with Covariance Form.}
    \label{alg:2}
    \begin{algorithmic}[1]
    \REQUIRE At stage $k$, the current canonical model $v^{\rho_k}\in\velmancan$; a stepsize $\eta_k\in(0,\tau]$; a reference factor $\mu=\rho^\base$ for fine-tuning (with the pretrained model $v^\base$) or $\mu=1$ for sampling; a reward function $r:\R^d\to\R$; an inverse temperature $\tau>0$; and a batch size $N_\mathrm{batch}$.
    \ENSURE The updated velocity model $\bar{v}^{k+1}=v^{\rho_k}+\eta_k\Gamma^{\rho_k,\tilde{r}^{\rho_k}}$.
    \STATE Warm-start $\theta$ with the network parameters of $v^{\rho_k}$.
    \FOR{each gradient step}
        \STATE \COMMENT{The following operations are applied to a batch.}
            \IF{forward construction is used}
                \IF{ODE-based sampling is used}
                    \STATE Sample $X_1\sim\rho_k$ through ODE \eqref{eq:1} of $v^{\rho_k}$.
                    \IF{ODE-based estimation is used}
                        \STATE Compute $\log\frac{\rho_k(X_1)}{\mu(X_1)}$ by augmenting \eqref{eq:1} with \eqref{eq:71} or \eqref{eq:72}.\label{line:1}
                    \ENDIF
                \ELSIF{SDE-based sampling is used}
                    \STATE Sample $X_1\sim\rho_k$ through SDE \eqref{eq:5} of $v^{\rho_k}$.
                    \IF{ODE-based estimation is used}
                        \STATE Compute $\log\frac{\rho_k(X_1)}{\mu(X_1)}$ by the backward integration \eqref{eq:73} or \eqref{eq:74}.\label{line:2}
                    \ENDIF
                \ENDIF
                \STATE Sample $t\sim\U(0,1)$ and $X_0\sim\Normal(0,I)$, and set $X_t\gets\alpha_tX_1+\beta_tX_0$.
                \IF{SDE-based estimation is used}
                    \STATE Sample $\BY_{[t,1]}\sim\mathbb{P}_{[t,1]}^\mathrm{uni}(\cdot|X_t,X_1)$ based on Theorem \ref{thm:14}.
                \ENDIF
            \ELSIF{reverse construction is used}
                \STATE Sample $t\sim\U(0,1)$ and $X_t\sim \hat{p}_t$.
                \STATE Sample $\BY_{[t,1]}\sim \mathbb{P}_{[t,1]|t}^{\rho_k}(\cdot|X_t)$ through the posterior-preserving SDE \eqref{eq:64} of $\rho_k$.
                \STATE Set $X_1\gets Y_1$.
                \IF{ODE-based estimation is used}
                    \STATE Compute $\log\frac{\rho_k(X_1)}{\mu(X_1)}$ by the backward integration \eqref{eq:73} or \eqref{eq:74}.\label{line:3}
                \ENDIF
            \ENDIF
            \IF{SDE-based estimation is used}
                \STATE Compute $\log\frac{\rho_k(X_1)}{\rho^\mu(X_1)}-\log\frac{p_t^{\rho_k}(X_t)}{p_t^{\mu}(X_t)}$ by \eqref{eq:79}.\label{line:4}
            \ENDIF
            \STATE Compute the target $\target$ by \eqref{eq:65}, where $\log\frac{p_t^{\rho_k}(X_t)}{p_t^{\mu}(X_t)}$ is merged into $B_t(X_t)$ in SDE-based estimation.\label{line:5}
        \STATE Update $\theta$ based on the batch loss $\mathcal{L}\gets\frac{1}{N_{\mathrm{batch}}}\sum_{(t,X_t,\target)}\norm[2]{v_t^\theta(X_t)-\target}^2$.
    \ENDFOR
    \STATE Set $\bar{v}^{k+1}\gets v^\theta$.
\end{algorithmic}

\end{algorithm}

Under the forward construction, the pair \((X_t,X_1)\) is constructed and satisfies
\[
    X_1| X_t
    \sim
    p_{1|t}^\rho(\cdot|X_t),
\]
but this construction does not provide a bridge of the posterior-preserving SDE between the two states. For fixed endpoints \((x_t,x_s)\), $0< t<s\leq 1$, let
\[
    \mathbb P_{[t,s]|t,s}^{\rho}
    (\cdot|x_t,x_s)
\]
denote the conditional path distribution of the posterior-preserving SDE given
\(Y_t=x_t\) and \(Y_s=x_s\). In particular, for $s=1$, the full posterior path distribution can be reproduced by the hierarchical sampling rule
\begin{equation}
    X_1
    \sim
    p_{1|t}^\rho(\cdot|x_t),\quad
    \BY_{[t,1]}
    \sim
    \mathbb P_{[t,1]|t,1}^{\rho}
    (\cdot|x_t,X_1)
    \quad\Rightarrow\quad
    \BY_{[t,1]}
    \sim
    \mathbb P_{[t,1]|t}^{\rho}
    (\cdot|x_t).
    \label{eq:81}
\end{equation}
Since the forward construction has already produced the first sample in~\eqref{eq:81}, it remains only to draw
\begin{equation*}
    \BY_{[t,1]}
    \sim
    \mathbb P_{[t,1]|t,1}^{\rho}
    (\cdot|X_t,X_1).
\end{equation*}

An interesting property of the posterior-preserving SDE is that its two-sided conditional distribution \(\mathbb P_{[t,s]|t,s}^{\rho}(\cdot|x_t,x_s)\) is independent of the terminal density \(\rho\) and depends only on the schedules \(\alpha_t\) and \(\beta_t\). We refer to this property as \emph{bridge universality}, and denote the bridge distribution by
\[
\mathbb P_{[t,s]}^{\mathrm{uni}}(\cdot|x_t,x_s).
\]
It enables efficient generation of a stochastic path \(\BY_{[t,1]}\) that is consistent with the posterior-preserving SDE conditioning on the two endpoints \((X_t,X_1)\), thereby allowing the hierarchical sampling scheme in~\eqref{eq:81} to be realized efficiently. A detailed derivation is provided in Appendix~\ref{subsec:43}. Specifically, for any two terminal densities \(\rho_1,\rho_2\) and any \(0< t<s\leq1\),
\begin{equation*}
    \mathbb P_{[t,s]|t,s}^{\rho_1}
    (\cdot|x_t,x_s)
    =\mathbb P_{[t,s]}^{\mathrm{uni}}(\cdot|x_t,x_s)
    =
    \mathbb P_{[t,s]|t,s}^{\rho_2}
    (\cdot|x_t,x_s).
\end{equation*}

To give the explicit construction, define \(\gamma_s:=\frac{\beta_s^2}{\alpha_s^2}\).
For \(t<u<s\), conditioning on the two endpoints, the state at time \(u\) follows
\begin{equation}
    Y_u
    |
    \left(
        Y_t,
        Y_s
    \right)
    \sim
    \Normal
    \left(
        m_{u|t,s}(Y_t,Y_s),
        \Sigma_{u|t,s}
    \right),
    \label{eq:82}
\end{equation}
where
\begin{align*}
    m_{u|t,s}(x_t,x_s)
    &:=
    \alpha_u
    \left[
        \frac{\gamma_u-\gamma_s}
             {\gamma_t-\gamma_s}
        \frac{x_t}{\alpha_t}
        +
        \frac{\gamma_t-\gamma_u}
             {\gamma_t-\gamma_s}
        \frac{x_s}{\alpha_s}
    \right],
    \qquad
    \Sigma_{u|t,s}
    :=
    \alpha_u^2
    \frac{
        \left(\gamma_t-\gamma_u\right)
        \left(\gamma_u-\gamma_s\right)
    }{
        \gamma_t-\gamma_s
    }
    I.
\end{align*}
Equation~\eqref{eq:82} can be applied recursively to sample the entire path on a discrete time grid.

The universal bridge directly generates the state path, whereas the stochastic integral in~\eqref{eq:80} is written in terms of the Brownian motion driving the posterior-preserving SDE. Since the diffusion coefficient is state-independent, the corresponding increments can be recovered from the sampled state path through
\begin{equation*}
    \dd W_s
    =
    \frac{
        \dd Y_s
        -
        b_s^\rho(Y_s)\dd s
    }{
        \sqrt{2\kappa_s}
    }.
\end{equation*}

The exact tangential update under the covariance form is summarized in Algorithm~\ref{alg:2}.

\subsection{Gradient Form}\label{subsec:17}

Whereas~\eqref{eq:56} leads to the covariance form,~\eqref{eq:59} expresses the tangent vector through the action of a posterior Stein kernel on the regularized-reward gradient. We now present the derivation of the gradient form based on the posterior Stein kernel \(\Lambda_t^\rho\).

Suppose first that an exact posterior Stein kernel \(\Lambda_t^\rho(\cdot|x_t)\) is available. The gradient representation~\eqref{eq:59} then suggests the following target
\begin{align}
    \target_{\rho,\eta}^{\mathrm{grad,ker}}(t,X_t,X_1):=v_{t|1}(X_t|X_1)+\eta\frac{\alpha_t\kappa_t}{\beta_t^2}\Lambda_t^\rho(X_1|X_t)^\top\nabla\tilde r^\rho(X_1).
    \label{eq:83}
\end{align}
By~\eqref{eq:59},
\begin{equation}
    \E_{X_1\sim p_{1|t}^\rho(\cdot|x_t)}
    \left[
        \target_{\rho,\eta}^{\mathrm{grad,ker}}(t,x_t,X_1)
    \right]
    =
    v_t^\rho(x_t)
    +
    \eta\Gamma_t^{\rho,\tilde r^\rho}(x_t).
    \label{eq:84}
\end{equation}

Therefore, under the forward construction, the ideal kernel-based gradient loss is given by
\begin{align}
    \calL_{\rho,\eta}^{\mathrm{grad,ker},\mathrm{fwd}}(\theta)
    :=&
    \E_{\substack{
        t\sim\U(0,1),\,X_1\sim\rho,\,X_0\sim p_0,\\
        X_t=\alpha_tX_1+\beta_tX_0}}
    \left[
        \norm[2]{
            v_t^\theta(X_t)
            -
            \target_{\rho,\eta}^{\mathrm{grad,ker}}(t,X_t,X_1)
        }^2
    \right],
    \label{eq:85}
\end{align}
and under the reverse construction,
\begin{subequations}
    \label{eq:86}
\begin{align}
    &\calL_{\rho,\eta}^{\mathrm{grad,ker},\mathrm{rev}}(\theta)\notag\\
    :=&
    \E_{\substack{
        t\sim\U(0,1),\,X_t\sim\hat p_t,\,
        X_1\sim p_{1|t}^\rho(\cdot|X_t)}}
    \left[
        \norm[2]{
            v_t^\theta(X_t)
            -
            \target_{\rho,\eta}^{\mathrm{grad,ker}}(t,X_t,X_1)
        }^2
    \right]
    \\
    =&\E_{\substack{
            t\sim\U(0,1),\,X_t\sim\hat p_t}}
        \left[
            \norm[2]{
                v_t^\theta(X_t)
                -
                \E_{X_1\sim p_{1|t}^\rho(\cdot|X_t)}
                \left[
                    \target_{\rho,\eta}^{\mathrm{grad,ker}}(t,X_t,X_1)
                \right]
            }^2
        \right]
        + \const.
    \label{eq:87}
\end{align}
\end{subequations}
These losses exactly parallel the covariance-form losses in~\eqref{eq:67} and~\eqref{eq:69}. The key question is how to construct a posterior Stein kernel and evaluate \(\Lambda_t^\rho(X_1|X_t)^\top\nabla\tilde r^\rho(X_1)\).

For a general non-Gaussian posterior, obtaining such a kernel is a nontrivial problem. For fixed \((t,x_t)\), a posterior Stein kernel must satisfy
\begin{equation}
    \nabla_{x_1}\cdot
    \left(
        p_{1|t}^\rho(x_1|x_t)
        \Lambda_t^\rho(x_1|x_t)
    \right)
    =
    -
    p_{1|t}^\rho(x_1|x_t)
    \left(
        x_1-M_t^\rho(x_t)
    \right),
    \label{eq:88}
\end{equation}
where \(M_t^\rho(x_t)=\E_{X_1\sim p_{1|t}^\rho(\cdot|x_t)}[X_1]\) denotes the posterior mean. The details are presented in Appendix~\ref{subsec:48}. Solving~\eqref{eq:88} directly is in general intractable.

Fortunately, the posterior-preserving SDE~\eqref{eq:64} provides a constructive way to obtain a particular posterior Stein kernel. Its terminal transition density is exactly \(p_{1|t}^\rho(\cdot|x_t)\). Hence, for every sufficiently regular \(f\),
\begin{equation*}
    \E_{\BY_{[t,1]}\sim\mathbb P_{[t,1]|t}^{\rho}(\cdot|x_t)}
    \left[
        f(Y_1)
    \right]
    =
    \E_{X_1\sim p_{1|t}^\rho(\cdot|x_t)}
    \left[
        f(X_1)
    \right]
    =
    V_t^\rho[f](x_t).
\end{equation*}
For \(f=\tilde r^\rho\), the initial-state sensitivity of this SDE therefore yields the posterior-value gradient \(\nabla_{x_t}V_t^\rho[\tilde r^\rho](x_t)\). This observation motivates using sensitivity analysis of the posterior-preserving SDE to construct a posterior Stein kernel. Indeed, conditioning the resulting state-transition matrix on the endpoints \((x_t,x_1)\) and applying the schedule-dependent rescaling below yields an exact posterior Stein kernel, while a backward adjoint ODE provides a matrix-free Monte Carlo evaluation of its action on a vector. We call this special kernel the \emph{posterior sensitivity kernel}.

\paragraph{Posterior sensitivity kernel.} Fix \(t\in(0,1)\) and \(x_t\in\R^d\), and recall the posterior-preserving SDE
\begin{subequations}
\label{eq:89}
    \begin{align}
        &\dd Y_s
        =
        b_s^\rho(Y_s)\dd s
        +
        \sqrt{2\kappa_s}\dd W_s,
        \qquad s\in[t,1],\\
        &Y_t=x_t.
    \end{align}
\end{subequations}
Its path distribution is \(\mathbb P_{[t,1]|t}^\rho(\cdot|x_t)\), and its endpoint satisfies \(Y_1\sim p_{1|t}^\rho(\cdot|x_t)\).

For a realized continuous path \(\BY_{[t,1]}=(Y_s)_{s\in[t,1]}\), define the pathwise state-transition matrix of the SDE~\eqref{eq:89} by
\begin{subequations}\label{eq:90}
    \begin{align}
        &\frac{\dd}{\dd s}G_{t\to s}^{b^\rho}(\BY_{[t,1]}) = \nabla b_s^\rho(Y_s)G_{t\to s}^{b^\rho}(\BY_{[t,1]}), \qquad s\in[t,1],\\
        &G_{t\to t}^{b^\rho}(\BY_{[t,1]}) = I.
    \end{align}
\end{subequations}
When the path $\BY_{[t,1]}$ is generated by the same SDE~\eqref{eq:89}, \(G_{t\to s}^{b^\rho}\) coincides with the Jacobian of \(Y_s\) with respect to the initial state \(x_t\). Throughout this paper, whenever derivatives with respect to the initial state \(x_t\) are considered, the same Brownian motion $(W_s)_{s\in[t,1]}$ is used for all initial states.

A posterior Stein kernel must depend only on the endpoint pair \((x_t,x_1)\), not on the intervening realized path. We therefore define the endpoint-conditioned state-transition matrix
\begin{equation}
    K_{1|t}^{\rho}(x_1|x_t)
    :=
    \E_{\BY_{[t,1]}\sim
    \mathbb P_{[t,1]|t,1}^{\rho}(\cdot|x_t,x_1)}
    \left[
        G_{t\to1}^{b^\rho}(\BY_{[t,1]})
    \right]
    =
    \E_{\BY_{[t,1]}\sim
    \mathbb P_{[t,1]}^{\mathrm{uni}}(\cdot|x_t,x_1)}
    \left[
        G_{t\to1}^{b^\rho}(\BY_{[t,1]})
    \right].
    \label{eq:91}
\end{equation}
In Appendix~\ref{subsec:49}, we show that
\begin{equation}
    \Lambda_t^\rho(x_1|x_t)
    =
    \frac{\beta_t^2}{\alpha_t}
    K_{1|t}^{\rho}(x_1|x_t)
    \label{eq:92}
\end{equation}
is a posterior Stein kernel for \(p_{1|t}^\rho(\cdot|x_t)\), called the posterior sensitivity kernel. Therefore, sensitivity analysis of the posterior-preserving SDE provides both a construction of a posterior Stein kernel and a practical representation in terms of the state-transition matrix dynamics in~\eqref{eq:90}.

However, explicitly forming the matrix \(G_{t\to s}^{b^\rho}\in\R^{d\times d}\) can be computationally expensive for high-dimensional data such as images. A standard technique from optimal control, known as the adjoint method, allows the relevant matrix-vector products to be computed without explicitly forming the matrix. Next, we state the adjoint formulation needed for the posterior-preserving SDE~\eqref{eq:89}. In Appendix~\ref{subsec:45}, we develop the pathwise sensitivity and adjoint identities for a general SDE; thus, the adjoint construction itself is not specific to the posterior-preserving SDE~\eqref{eq:89}.

\paragraph{Matrix-free representation via adjoint.} For a terminal objective \(\varphi:\R^d\to\R\) and a running objective \(l_s:\R^d\to\R\), define the pathwise objective
\begin{equation*}
    J_t^{\varphi,l}(\BY_{[t,1]})
    :=
    \varphi(Y_1)
    +
    \int_t^1 l_s(Y_s)\dd s.
\end{equation*}
Its pathwise adjoint is defined by
    \begin{align*}
        &\frac{\dd}{\dd s}
        \lambda_s^{b^\rho,\varphi,l}(\BY_{[t,1]})
        =
        -\nabla b_s^\rho(Y_s)^\top
        \lambda_s^{b^\rho,\varphi,l}(\BY_{[t,1]})
        -
        \nabla l_s(Y_s),
        \qquad s\in[t,1],\\
        &\lambda_1^{b^\rho,\varphi,l}(\BY_{[t,1]})
        =
        \nabla\varphi(Y_1).
    \end{align*}
In Appendix~\ref{sub2sec:32}, we show that the adjoint and the state-transition matrix are related by
\begin{align}
    \lambda_t^{b^\rho,\varphi,l}(\BY_{[t,1]})
    =
    G_{t\to1}^{b^\rho}(\BY_{[t,1]})^\top
    \nabla\varphi(Y_1)
    +
    \int_t^1
    G_{t\to s}^{b^\rho}(\BY_{[t,1]})^\top
    \nabla l_s(Y_s)\dd s.
    \label{eq:93}
\end{align}
For a terminal observable \(f\), take \(\varphi=f\) and \(l_s=0\).
Equations~\eqref{eq:91}, \eqref{eq:92}, and~\eqref{eq:93} imply
\begin{align}
    \frac{\alpha_t\kappa_t}{\beta_t^2}
    \Lambda_t^\rho(x_1|x_t)^\top
    \nabla f(x_1)
    =
    \kappa_t
    \E_{\BY_{[t,1]}\sim
    \mathbb P_{[t,1]}^{\mathrm{uni}}(\cdot|x_t,x_1)}
    \left[
        \lambda_t^{b^\rho,f,0}(\BY_{[t,1]})
    \right].
    \label{eq:94}
\end{align}
Thus, a single pathwise adjoint provides a matrix-free Monte Carlo estimator of the kernel-based target.

The covariance-form target itself is defined only by the endpoint pair \((X_t,X_1)\), with any ODE or SDE path used only to evaluate \(\tilde r^\rho(X_1)\). In contrast, the adjoint-based targets below are intrinsically path-dependent and therefore require a path distribution. Whenever the specific path construction and the \(Y_t\)-marginal are immaterial, we use the generic notation
\begin{equation}\label{eq:95}
    \BY_{[t,1]}\sim\Xi_{[t,1]}^\rho
    \quad\Rightarrow\quad
    \BY_{[t,1]}|Y_t
    \sim
    \mathbb P_{[t,1]|t}^\rho(\cdot|Y_t).
\end{equation}
The \(Y_t\)-marginal is required to have a positive density.

We next present two exact adjoint-based realizations for the gradient form. The first retains the full regularized reward in the terminal objective and includes no running objective, corresponding to the Mayer form in optimal-control terminology. The second expresses the same posterior value using the terminal reward $r$ together with a running regularization objective, corresponding to the Bolza form. Both yield targets whose conditional mean, given $Y_t=x_t$, is \(v_t^\rho(x_t)+\eta\Gamma_t^{\rho,\tilde r^\rho}(x_t)\).

\paragraph{Mayer realization.} In~\eqref{eq:94}, we can directly set \(f=\tilde r^\rho\). The adjoint satisfies
\begin{subequations}
\label{eq:96}
    \begin{align}
        &\frac{\dd}{\dd s}
        \lambda_s^{b^\rho,\tilde r^\rho,0}(\BY_{[t,1]})
        =
        -\left(2\nabla v_s^\rho(Y_s)-\frac{\dot\alpha_s}{\alpha_s}I\right)^\top
        \lambda_s^{b^\rho,\tilde r^\rho,0}(\BY_{[t,1]}),\qquad s\in[t,1],\\
        &\lambda_1^{b^\rho,\tilde r^\rho,0}(\BY_{[t,1]})
        =
        \nabla\tilde r^\rho(Y_1).
    \end{align}
\end{subequations}
Conditioning on an endpoint pair \((X_t,X_1)\), let \(\BY_{[t,1]}\) denote a draw from the endpoint-conditioned path distribution \(\mathbb P_{[t,1]|t,1}^{\rho}(\cdot|X_t,X_1)=\mathbb P_{[t,1]}^{\mathrm{uni}}(\cdot|X_t,X_1)\). The corresponding pathwise Mayer target is written as
\begin{align}
    \target_{\rho,\eta}^{\mathrm{grad,Mayer}}
    (t,\BY_{[t,1]})
    :=&
    v_{t|1}(Y_t|Y_1)
    +
    \eta\kappa_t
    \lambda_t^{b^\rho,\tilde r^\rho,0}(\BY_{[t,1]}).
    \label{eq:97}
\end{align}
By~\eqref{eq:94},
\begin{equation*}
    \E_{\BY_{[t,1]}\sim\mathbb P_{[t,1]}^{\mathrm{uni}}(\cdot|x_t,x_1)}
    \left[
        \target_{\rho,\eta}^{\mathrm{grad,Mayer}}
        (t,\BY_{[t,1]})
    \right]
    =
    \target_{\rho,\eta}^{\mathrm{grad,ker}}(t,x_t,x_1).
\end{equation*}
Thus, the Mayer target is a bridge-level Monte Carlo estimation of the ideal kernel-based target.

When \(\mu=1\), the boundary condition is
\begin{equation*}
    \lambda_1^{b^\rho,\tilde r^\rho,0}(\BY_{[t,1]})
    =
    \nabla r(Y_1)
    -
    \frac{1}{\tau}\nabla\log\rho(Y_1).
\end{equation*}
When \(\mu=\rho^\base\), it is
\begin{equation*}
    \lambda_1^{b^\rho,\tilde r^\rho,0}(\BY_{[t,1]})
    =
    \nabla r(Y_1)
    -
    \frac{1}{\tau}
    \nabla\log\frac{\rho(Y_1)}{\rho^\base(Y_1)}.
\end{equation*}
The Mayer realization involves no running objective, but its terminal condition requires the exact gradient of the log density or log-density ratio. Computing this terminal condition can be challenging in practice when \(\rho\) and \(\rho^\base\) are terminal densities induced by neural-network velocity fields and may therefore require careful numerical treatment.

Under the forward construction, the Mayer realization of the gradient-form loss is
\begin{align}
    \calL_{\rho,\eta}^{\mathrm{grad,Mayer},\mathrm{fwd}}(\theta)
    :=&
    \E_{\substack{
        t\sim\U(0,1),\,X_1\sim\rho,\,X_0\sim p_0,\\
        X_t=\alpha_tX_1+\beta_tX_0,\\
        \BY_{[t,1]}\sim\mathbb P_{[t,1]}^{\mathrm{uni}}(\cdot|X_t,X_1)}}
    \left[
        \norm[2]{
            v_t^\theta(X_t)
            -
            \target_{\rho,\eta}^{\mathrm{grad,Mayer}}(t,\BY_{[t,1]})
        }^2
    \right],
    \label{eq:98}
\end{align}
and under the reverse construction,
\begin{subequations}
    \label{eq:99}
\begin{align}
    &\calL_{\rho,\eta}^{\mathrm{grad,Mayer},\mathrm{rev}}(\theta)\notag\\
    :=&
    \E_{\substack{
        t\sim\U(0,1),\,X_t\sim\hat p_t,\,
        \BY_{[t,1]}\sim\mathbb P_{[t,1]|t}^{\rho}(\cdot|X_t)}}
    \left[
        \norm[2]{
            v_t^\theta(X_t)
            -
            \target_{\rho,\eta}^{\mathrm{grad,Mayer}}(t,\BY_{[t,1]})
        }^2
    \right]
    \\
    =&
        \E_{\substack{
            t\sim\U(0,1),\,X_t\sim\hat p_t}}
        \left[
            \norm[2]{
                v_t^\theta(X_t)
                -
                \E_{\BY_{[t,1]}\sim\mathbb P_{[t,1]|t}^{\rho}(\cdot|X_t)
                }\left[
                    \target_{\rho,\eta}^{\mathrm{grad,Mayer}}(t,\BY_{[t,1]})
                \right]
            }^2
        \right]
        + \const.
    \label{eq:100}
\end{align}
\end{subequations}

\paragraph{Bolza realization.} The density regularization can instead be moved from the terminal objective to a running objective along the paths compatible with the posterior-preserving SDE. This gives a Bolza realization with terminal objective \(r\).

For \(\mu\in\{1,\rho^\base\}\), define the running objective
\begin{equation}
    l_s^{\rho,\mu}(x)
    :=
    -
    \frac{1}{\tau\kappa_s}\norm[2]{v_s^\rho(x)-v_s^{\mu}(x)}^2.
    \label{eq:101}
\end{equation}
When \(\mu=1\), \(v_s^{1}(x):=\frac{\dot\alpha_s}{\alpha_s}x\). When \(\mu=\rho^\base\), \(v_s^\base(x)\) denotes the canonical velocity field associated with the base density \(\rho^\base\).
For either case, let \(\lambda^{b^\rho,r,l^{\rho,\mu}}\) denote the adjoint associated with the terminal objective \(r\) and the running objective \(l^{\rho,\mu}\):
\begin{subequations}
\label{eq:102}
    \begin{align}
        &\frac{\dd}{\dd s}
        \lambda_s^{b^\rho,r,l^{\rho,\mu}}(\BY_{[t,1]})
        =
        -\left(2\nabla v_s^\rho(Y_s)-\frac{\dot\alpha_s}{\alpha_s}I\right)^\top
        \lambda_s^{b^\rho,r,l^{\rho,\mu}}(\BY_{[t,1]})
        -
        \nabla l_s^{\rho,\mu}(Y_s),
        \quad s\in[t,1],
        \label{eq:103}\\
        &\lambda_1^{b^\rho,r,l^{\rho,\mu}}(\BY_{[t,1]})
        =
        \nabla r(Y_1).
    \end{align}
\end{subequations}

For the \(\mu=1\) case, we can derive
\begin{align}
    V_t^\rho[\tilde r^\rho](x_t)
    =
    \E_{\BY_{[t,1]}\sim
    \mathbb P_{[t,1]|t}^{\rho}(\cdot|x_t)}
    \left[
        r(Y_1)
        +
        \int_t^1 l_s^{\rho,1}(Y_s)\dd s
    \right]
    -
    \frac{1}{\tau}
    \log\left(
        \alpha_t^d p_t^\rho(x_t)
    \right).
    \label{eq:104}
\end{align}
The proof is provided in Appendix~\ref{subsec:44}.
We also have
\begin{equation*}
    \nabla_{x_t}
    \log\left(
        \alpha_t^d p_t^\rho(x_t)
    \right)
    =
    \frac{1}{\kappa_t}
    \left(
        v_t^\rho(x_t)
        -
        v_t^{1}(x_t)
    \right).
\end{equation*}
Differentiating~\eqref{eq:104} and applying the adjoint representation therefore gives
\begin{align}
    \Gamma_t^{\rho,\tilde r^\rho}(x_t)
    =
    \kappa_t
    \E_{\BY_{[t,1]}\sim
    \mathbb P_{[t,1]|t}^{\rho}(\cdot|x_t)}
    \left[
        \lambda_t^{b^\rho,r,l^{\rho,1}}(\BY_{[t,1]})
    \right]
    -
    \frac{1}{\tau}
    \left(
        v_t^\rho(x_t)
        -
        v_t^{1}(x_t)
    \right).
    \label{eq:105}
\end{align}

\begin{algorithm}[!t]
    \caption{Tangential Update: Exact Realization with Gradient Form.}
    \label{alg:3}
    \begin{algorithmic}[1]
    \REQUIRE At stage $k$, the current canonical model $v^{\rho_k}\in\velmancan$; a stepsize $\eta_k\in(0,\tau]$; a reference factor $\mu=\rho^\base$ for fine-tuning (with the pretrained model $v^\base$) or $\mu=1$ for sampling; a reward function $r:\R^d\to\R$; an inverse temperature $\tau>0$; and a batch size $N_\mathrm{batch}$.
    \ENSURE The updated velocity model $\bar{v}^{k+1}=v^{\rho_k}+\eta_k\Gamma^{\rho_k,\tilde{r}^{\rho_k}}$.
    \STATE Warm-start $\theta$ with the network parameters of $v^{\rho_k}$.
    \FOR{each gradient step}
        \STATE \COMMENT{The following operations are applied to a batch.}
            \IF{forward construction is used}
                \IF{ODE-based sampling is used}
                    \STATE Sample $X_1\sim\rho_k$ through ODE \eqref{eq:1} of $v^{\rho_k}$.
                \ELSIF{SDE-based sampling is used}
                    \STATE Sample $X_1\sim\rho_k$ through SDE \eqref{eq:5} of $v^{\rho_k}$.
                \ENDIF
                \STATE Sample $t\sim\U(0,1)$ and $X_0\sim\Normal(0,I)$, and set $X_t\gets\alpha_tX_1+\beta_tX_0$.
                \STATE Sample $\BY_{[t,1]}\sim\mathbb{P}_{[t,1]}^\mathrm{uni}(\cdot|X_t,X_1)$ based on Theorem \ref{thm:14}.
            \ELSIF{reverse construction is used}
                \STATE Sample $t\sim\U(0,1)$ and $X_t\sim \hat{p}_t$.
                \STATE Sample $\BY_{[t,1]}\sim \mathbb{P}_{[t,1]|t}^{\rho_k}(\cdot|X_t)$ through the posterior-preserving SDE \eqref{eq:64} of $\rho_k$.
            \ENDIF
            \IF{Mayer realization is used}\label{line:6}
                \STATE Compute the pathwise adjoint $\lambda_t^{b^{\rho_k},\tilde r^{\rho_k},0}(\BY_{[t,1]})$ by ODE \eqref{eq:96}.
                \STATE Compute the target $\target$ by \eqref{eq:97}.
            \ELSIF{Bolza realization is used}
                \STATE Compute the pathwise adjoint $\lambda_t^{b^{\rho_k},r,l^{\rho_k,\mu}}(\BY_{[t,1]})$ by ODE \eqref{eq:102}.
                \STATE Compute the target $\target$ by \eqref{eq:108}.
            \ENDIF\label{line:7}
        \STATE Update $\theta$ based on the batch loss $\mathcal{L}\gets\frac{1}{N_{\mathrm{batch}}}\sum_{(t,X_t,\target)}\norm[2]{v_t^\theta(X_t)-\target}^2$.
    \ENDFOR
    \STATE Set $\bar{v}^{k+1}\gets v^\theta$.
\end{algorithmic}
\end{algorithm}

For the \(\mu=\rho^\base\) case, we can derive
\begin{align}
    V_t^\rho[\tilde r^\rho](x_t)
    =
    \E_{\BY_{[t,1]}\sim
    \mathbb P_{[t,1]|t}^{\rho}(\cdot|x_t)}
    \left[
        r(Y_1)
        +
        \int_t^1 l_s^{\rho,\rho^\base}(Y_s)\dd s
    \right]
    -
    \frac{1}{\tau}
    \log\frac{p_t^\rho(x_t)}{p_t^\base(x_t)}.
    \label{eq:106}
\end{align}
The proof is provided in Appendix~\ref{subsec:44}.
We also have
\begin{equation*}
    \nabla_{x_t}
    \log\frac{p_t^\rho(x_t)}{p_t^\base(x_t)}
    =
    \frac{1}{\kappa_t}
    \left(
        v_t^\rho(x_t)
        -
        v_t^{\base}(x_t)
    \right).
\end{equation*}
Differentiating~\eqref{eq:106} and applying the adjoint representation therefore gives
\begin{align}
    \Gamma_t^{\rho,\tilde r^\rho}(x_t)
    =
    \kappa_t
    \E_{\BY_{[t,1]}\sim
    \mathbb P_{[t,1]|t}^{\rho}(\cdot|x_t)}
    \left[
        \lambda_t^{b^\rho,r,l^{\rho,\rho^\base}}(\BY_{[t,1]})
    \right]
    -
    \frac{1}{\tau}
    \left(
        v_t^\rho(x_t)
        -
        v_t^{\base}(x_t)
    \right).
    \label{eq:107}
\end{align}

For either \(\mu\in\{1,\rho^\base\}\), the pathwise Bolza target is written as
\begin{align}
    \target_{\rho,\eta}^{\mathrm{grad,Bolza}}
    (t,\BY_{[t,1]})
    :=
    \left(1-\frac{\eta}{\tau}\right)v_{t|1}(Y_t|Y_1)+\frac{\eta}{\tau}v_t^{\mu}(Y_t)+\eta\kappa_t\lambda_t^{b^\rho,r,l^{\rho,\mu}}(\BY_{[t,1]}).
    \label{eq:108}
\end{align}
Equations~\eqref{eq:105} and~\eqref{eq:107} imply
\begin{equation*}
    \E_{\BY_{[t,1]}\sim
    \mathbb P_{[t,1]|t}^{\rho}(\cdot|x_t)}
    \left[
        \target_{\rho,\eta}^{\mathrm{grad,Bolza}}
        (t,\BY_{[t,1]})
    \right]
    =
    v_t^\rho(x_t)
    +
    \eta\Gamma_t^{\rho,\tilde r^\rho}(x_t).
\end{equation*}

Under the forward construction, the Bolza realization of the gradient-form loss is
\begin{align}
    \calL_{\rho,\eta}^{\mathrm{grad,Bolza},\mathrm{fwd}}(\theta)
    :=&
    \E_{\substack{
        t\sim\U(0,1),\,X_1\sim\rho,\,X_0\sim p_0,\\
        X_t=\alpha_tX_1+\beta_tX_0,\\
        \BY_{[t,1]}\sim\mathbb P_{[t,1]}^{\mathrm{uni}}(\cdot|X_t,X_1)}}
    \left[
        \norm[2]{
            v_t^\theta(X_t)
            -
            \target_{\rho,\eta}^{\mathrm{grad,Bolza}}(t,\BY_{[t,1]})
        }^2
    \right],
    \label{eq:109}
\end{align}
and under the reverse construction,
\begin{subequations}
    \label{eq:110}
\begin{align}
    &\calL_{\rho,\eta}^{\mathrm{grad,Bolza},\mathrm{rev}}(\theta)\notag\\
    :=&
    \E_{\substack{
        t\sim\U(0,1),\,X_t\sim\hat p_t,\,
        \BY_{[t,1]}\sim\mathbb P_{[t,1]|t}^{\rho}(\cdot|X_t)}}
    \left[
        \norm[2]{
            v_t^\theta(X_t)
            -
            \target_{\rho,\eta}^{\mathrm{grad,Bolza}}(t,\BY_{[t,1]})
        }^2
    \right]
    \\
    =
        &\E_{\substack{
            t\sim\U(0,1),\,X_t\sim\hat p_t}}
        \left[
            \norm[2]{
                v_t^\theta(X_t)
                -
                \E_{\BY_{[t,1]}\sim\mathbb P_{[t,1]|t}^{\rho}(\cdot|X_t)
                }
                \left[
                    \target_{\rho,\eta}^{\mathrm{grad,Bolza}}(t,\BY_{[t,1]})
                \right]
            }^2
        \right]
        + \const.
\end{align}
\end{subequations}

The exact tangential update under the gradient form is summarized in Algorithm~\ref{alg:3}.

\begin{sidewaystable}[p]
    \centering
    \caption{Exact Newton Matching: summary of losses for the tangential update $v_t^\theta(x_t)=v_t^\rho(x_t)+\eta\Gamma_t^{\rho,\tilde{r}^\rho}(x_t)$.}
    \label{tab:1}
    \begingroup
\renewcommand{\arraystretch}{1.25}

\newlength{\exactnewtontablewidth}
\newlength{\exactnewtoncolone}
\newlength{\exactnewtoncoltwo}
\newlength{\exactnewtoncolthree}
\setlength{\exactnewtontablewidth}{\linewidth}
\addtolength{\exactnewtontablewidth}{-6\tabcolsep}
\addtolength{\exactnewtontablewidth}{-4\arrayrulewidth}
\setlength{\exactnewtoncoltwo}{0.47\exactnewtontablewidth}
\setlength{\exactnewtoncolthree}{0.45\exactnewtontablewidth}
\setlength{\exactnewtoncolone}{\exactnewtontablewidth}
\addtolength{\exactnewtoncolone}{-\exactnewtoncoltwo}
\addtolength{\exactnewtoncolone}{-\exactnewtoncolthree}

\newlength{\exactnewtoncellpadding}
\setlength{\exactnewtoncellpadding}{0.3em}
\newcommand{\exactnewtoncell}[1]{%
    \kern\exactnewtoncellpadding
    {\centering
    #1\par}%
    \kern\exactnewtoncellpadding
}
\newcolumntype{N}[1]{%
    >{\begin{minipage}[c]{#1}\ignorespaces}c<{\end{minipage}}%
}

\footnotesize
\fontsize{9pt}{10.5pt}\selectfont
\setlength{\abovedisplayskip}{0.4em}
\setlength{\belowdisplayskip}{0.4em}
\setlength{\abovedisplayshortskip}{0.4em}
\setlength{\belowdisplayshortskip}{0.4em}
\begin{tabular}{
    |N{\exactnewtoncolone}
    |N{\exactnewtoncoltwo}
    |N{\exactnewtoncolthree}|}
    \hline
    &
    \exactnewtoncell{Covariance Form ($\Omega_t(x_t)\equiv 0$)
        $$
        \Gamma_t^{\rho,\tilde{r}^\rho}(x_t)=\E_{X_1\sim p_{1|t}^\rho(\cdot|x_t)}\left[\left(\tilde r^\rho(X_1)-B_t(x_t)\right)\left(v_{t|1}(x_t|X_1)-v_t^\rho(x_t)\right)\right]
        $$}
    &
    \exactnewtoncell{Gradient Form ($\Omega_t(x_t)\equiv I$)
        $$
        \Gamma_t^{\rho,\tilde{r}^\rho}(x_t)=\frac{\alpha_t\kappa_t}{\beta_t^2}\E_{X_1\sim p_{1|t}^\rho(\cdot|x_t)}\left[\Lambda_t^\rho(X_1|x_t)^\top\nabla\tilde r^\rho(X_1)\right]
        $$
        }
    \tabularnewline
    \hline
    \exactnewtoncell{Target\\Realization}
    &
    \exactnewtoncell{$$
        \target^{\mathrm{cov}}%
        =v_{t|1}(x_t|x_1)+\eta\left(\tilde r^\rho(x_1)-B_t(x_t)\right)\left(v_{t|1}(x_t|x_1)-v_t^\rho(x_t)\right)
        $$
        {\raggedright
        \,\,Compute $\log\frac{\rho(x_1)}{\mu(x_1)}$ through:\par}
        \begin{itemize}[
            leftmargin=1.4em,
            labelwidth=0.6em,
            labelsep=0.4em,
            itemsep=0pt,
            parsep=0pt,
            topsep=0pt
        ]
            \raggedright
            \item ODE-based estimation: \eqref{eq:71}, \eqref{eq:72}, \eqref{eq:73} or \eqref{eq:74}.
            \item SDE-based estimation: \eqref{eq:77} or \eqref{eq:78}.
        \end{itemize}
        }
    &
    \exactnewtoncell{
        $$
        \target^{\mathrm{grad}}=v_{t|1}(x_t|x_1)+\eta\frac{\alpha_t\kappa_t}{\beta_t^2}\Lambda_t^\rho(x_1|x_t)^\top\nabla\tilde r^\rho(x_1)
        $$
        {\raggedright
        \,\,Estimate $\Lambda_t^\rho(x_1|x_t)^\top\nabla\tilde r^\rho(x_1)$ through:\par}
        \begin{itemize}[
            leftmargin=1.4em,
            labelwidth=0.6em,
            labelsep=0.4em,
            itemsep=0pt,
            parsep=0pt,
            topsep=0pt
        ]
            \raggedright
            \item Mayer realization \eqref{eq:97}, with adjoint ODE \eqref{eq:96}.
            \item Bolza realization \eqref{eq:108}, with adjoint ODE \eqref{eq:102}.
        \end{itemize}
        }
    \tabularnewline
    \hline
    \exactnewtoncell{Forward\\\makebox[\linewidth][c]{Construction}}
    &
    \exactnewtoncell{
        \[
        \E_{\substack{t\sim\U(0,1),\,X_0\sim p_0,\,X_1\sim\rho,\,X_t=\alpha_tX_1+\beta_tX_0}}\left[\norm[2]{v_t^\theta(X_t)-\target^{\mathrm{cov}}(t,X_t,X_1)}^2\right]
        \]
    }
    &
    \exactnewtoncell{
        \[
        \E_{\substack{
        t\sim\U(0,1),\,X_1\sim\rho,\,X_0\sim p_0,\,
        X_t=\alpha_tX_1+\beta_tX_0,\\
        \BY_{[t,1]}\sim\mathbb P_{[t,1]}^{\mathrm{uni}}(\cdot|X_t,X_1)}}\left[\norm[2]{v_t^\theta(X_t)-\target^{\mathrm{grad}}(t,\BY_{[t,1]})}^2\right]
        \]
    }
    \tabularnewline
    \hline
    \exactnewtoncell{Reverse\\\makebox[\linewidth][c]{Construction}}
    &
    \exactnewtoncell{
        \[
        \E_{\substack{t\sim\U(0,1),\,X_t\sim \hat{p}_t}}\left[\norm[2]{v_t^\theta(X_t)-\E_{X_1\sim p_{1|t}^\rho(\cdot|X_t)}\left[\target^{\mathrm{cov}}(t,X_t,X_1)\right]}^2\right]
        \]
    }
    &
    \exactnewtoncell{
        \[
        \E_{\substack{t\sim\U(0,1),\,X_t\sim \hat{p}_t}}\left[\norm[2]{v_t^\theta(X_t)-\E_{\BY_{[t,1]}\sim\mathbb P_{[t,1]|t}^{\rho}(\cdot|X_t)
            }\left[\target^{\mathrm{grad}}(t,\BY_{[t,1]})\right]}^2\right]
        \]
    }
    \tabularnewline
    \hline
\end{tabular}
\endgroup

\end{sidewaystable}

\subsection{Unification via Control Variates}\label{subsec:18}

The covariance and gradient forms developed earlier involve different sample-wise quantities, but both recover the same tangent vector \(\Gamma^{\rho,\tilde r^\rho}\) in conditional expectation. This equivalence can be understood more deeply through zero-mean control variates induced by Langevin Stein operators. In Appendix~\ref{app:5}, we present the corresponding Stein identity and the product rule for the Langevin Stein operator. We now specialize these results to the regularized reward, thereby constructing a unified family of targets that interpolates between the covariance and gradient forms.

By the definitions of the covariance target in~\eqref{eq:65} and the kernel-based gradient target in~\eqref{eq:83}, their sample-wise difference is
\begin{align*}
    &\target_{\rho,\eta}^{\mathrm{grad,ker}}(t,x_t,X_1)
    -
    \target_{\rho,\eta,B}^{\mathrm{cov}}(t,x_t,X_1)
    \notag\\
    =&
    \eta
    \left[
        \frac{\alpha_t\kappa_t}{\beta_t^2}
        \Lambda_t^\rho(X_1|x_t)^\top
        \nabla\tilde r^\rho(X_1)
        -
        \left(
            \tilde r^\rho(X_1)-B_t(x_t)
        \right)
        \left(
            v_{t|1}(x_t|X_1)-v_t^\rho(x_t)
        \right)
    \right].
\end{align*}
By~\eqref{eq:66} and~\eqref{eq:84}, the posterior means of both targets are $v_t^\rho(x_t)+\eta\Gamma_t^{\rho,\tilde r^\rho}(x_t)$; hence,
\begin{equation}
    \E_{X_1\sim p_{1|t}^\rho(\cdot|x_t)}
    \left[
        \target_{\rho,\eta}^{\mathrm{grad,ker}}(t,x_t,X_1)
        -
        \target_{\rho,\eta,B}^{\mathrm{cov}}(t,x_t,X_1)
    \right]
    =0.
    \label{eq:111}
\end{equation}
This demonstrates that the term \(\target_{\rho,\eta}^{\mathrm{grad,ker}}(t,x_t,X_1)-\target_{\rho,\eta,B}^{\mathrm{cov}}(t,x_t,X_1)\) is a zero-mean control variate obtained from the Stein identity, with the corresponding matrix test function given by \(\eta \frac{\alpha_t \kappa_t}{\beta_t^2}(\tilde r^\rho(x_1)-B_t(x_t))\Lambda_t^\rho(x_1|x_t)\).
The proof is provided in Appendix~\ref{app:5}.

Let \(\Omega_t(x_t)\in\R^{d\times d}\) be any matrix field that depends only on \((t,x_t)\). Left-multiplying~\eqref{eq:111} by \(\Omega_t(x_t)\) preserves its zero conditional mean. Starting from the covariance target, we therefore have
\begin{align*}
    &\E_{X_1\sim p_{1|t}^\rho(\cdot|x_t)}
    \left[
        \target_{\rho,\eta,B}^{\mathrm{cov}}(t,x_t,X_1)
    \right]
    \notag\\
    =&
    \E_{X_1\sim p_{1|t}^\rho(\cdot|x_t)}
    \left[
        \target_{\rho,\eta,B}^{\mathrm{cov}}(t,x_t,X_1)
        +
        \Omega_t(x_t)
        \left(
            \target_{\rho,\eta}^{\mathrm{grad,ker}}(t,x_t,X_1)
            -
            \target_{\rho,\eta,B}^{\mathrm{cov}}(t,x_t,X_1)
        \right)
    \right]
    \notag\\
    =&
    \E_{X_1\sim p_{1|t}^\rho(\cdot|x_t)}
    \left[
        \left(
            I-\Omega_t(x_t)
        \right)
        \target_{\rho,\eta,B}^{\mathrm{cov}}(t,x_t,X_1)
        +
        \Omega_t(x_t)
        \target_{\rho,\eta}^{\mathrm{grad,ker}}(t,x_t,X_1)
    \right].
\end{align*}
This identity motivates the unified target
\begin{align}
    \target_{\rho,\eta,B,\Omega}^{\mathrm{uni}}(t,X_t,X_1)
    :=
    \left(
        I-\Omega_t(X_t)
    \right)
    \target_{\rho,\eta,B}^{\mathrm{cov}}(t,X_t,X_1)
    +
    \Omega_t(X_t)
    \target_{\rho,\eta}^{\mathrm{grad,ker}}(t,X_t,X_1).
    \label{eq:112}
\end{align}
Substituting the definitions of the targets gives
\begin{align*}
    \target_{\rho,\eta,B,\Omega}^{\mathrm{uni}}(t,X_t,X_1)
    =&
    v_{t|1}(X_t|X_1)
    +
    \eta
    \Omega_t(X_t)
    \frac{\alpha_t\kappa_t}{\beta_t^2}
    \Lambda_t^\rho(X_1|X_t)^\top
    \nabla\tilde r^\rho(X_1)
    \notag\\
    &+
    \eta
    \left(
        I-\Omega_t(X_t)
    \right)
    \left(
        \tilde r^\rho(X_1)-B_t(X_t)
    \right)
    \left(
        v_{t|1}(X_t|X_1)-v_t^\rho(X_t)
    \right).
\end{align*}
The choices \(\Omega_t\equiv0\) and \(\Omega_t\equiv I\) recover the covariance target and kernel-based gradient target, respectively. We also have the unified representation of the tangent vector
\begin{equation*}
    \E_{X_1\sim p_{1|t}^\rho(\cdot|x_t)}
    \left[
        \target_{\rho,\eta,B,\Omega}^{\mathrm{uni}}(t,x_t,X_1)
    \right]
    =
    v_t^\rho(x_t)
    +
    \eta\Gamma_t^{\rho,\tilde r^\rho}(x_t).
\end{equation*}

The construction above uses the kernel-based gradient target since it makes the role of the posterior Stein kernel explicit. The preceding subsection also develops the Mayer and Bolza targets as exact pathwise realizations of the same gradient form. We have
\begin{align*}
    v_t^\rho(x_t)
    +
    \eta\Gamma_t^{\rho,\tilde r^\rho}(x_t)
    =&\E_{\BY_{[t,1]}\sim\mathbb P_{[t,1]|t}^{\rho}(\cdot|x_t)}
    \left[
        \target_{\rho,\eta}^{\mathrm{grad,Mayer}}(t,\BY_{[t,1]})
    \right]
    \\
    =&
    \E_{\BY_{[t,1]}\sim\mathbb P_{[t,1]|t}^{\rho}(\cdot|x_t)}
    \left[
        \target_{\rho,\eta}^{\mathrm{grad,Bolza}}(t,\BY_{[t,1]})
    \right].
\end{align*}
Therefore, the kernel-based component \(\target_{\rho,\eta}^{\mathrm{grad,ker}}\) in~\eqref{eq:112} may be replaced by either pathwise target. For simplicity, we omit these versions here.

Under the forward construction, the unified tangential-update loss is
\begin{align}
    \calL_{\rho,\eta,B,\Omega}^{\mathrm{uni},\mathrm{fwd}}(\theta)
    :=&
    \E_{\substack{
        t\sim\U(0,1),\,X_1\sim\rho,\,X_0\sim p_0,\\
        X_t=\alpha_tX_1+\beta_tX_0}}
    \left[
        \norm[2]{
            v_t^\theta(X_t)
            -
            \target_{\rho,\eta,B,\Omega}^{\mathrm{uni}}(t,X_t,X_1)
        }^2
    \right],
\label{eq:113}
\end{align}
and under the reverse construction,
\begin{subequations}
\label{eq:114}
\begin{align}
    &\calL_{\rho,\eta,B,\Omega}^{\mathrm{uni},\mathrm{rev}}(\theta)\notag\\
    :=&
    \E_{\substack{
        t\sim\U(0,1),\,X_t\sim\hat p_t,\,
        X_1\sim p_{1|t}^\rho(\cdot|X_t)}}
    \left[
        \norm[2]{
            v_t^\theta(X_t)
            -
            \target_{\rho,\eta,B,\Omega}^{\mathrm{uni}}(t,X_t,X_1)
        }^2
    \right]
    \\
    =&
    \E_{\substack{
        t\sim\U(0,1),\,X_t\sim\hat p_t}}
    \left[
        \norm[2]{
            v_t^\theta(X_t)
            -
            \E_{X_1\sim p_{1|t}^\rho(\cdot|X_t)}
            \left[
                \target_{\rho,\eta,B,\Omega}^{\mathrm{uni}}(t,X_t,X_1)
            \right]
        }^2
    \right]
    +
    \const.
\end{align}
\end{subequations}
In principle, \(\Omega_t(x_t)\) can be chosen according to an optimality criterion, such as minimizing the variance of the unified tangential-update loss. We provide a relevant result in Appendix~\ref{subsec:47}.

The preceding conditional-mean identities yield the following theorem. Throughout this paper, population minimizers and stationary points are understood up to a.e. equality under the corresponding regression distribution. The displayed formulas are taken as representatives.

\begin{theorem}%
    \label{thm:9}
    Fix a canonical anchor \(v^\rho\), a scalar stepsize \(\eta\in(0,\tau]\), and a positive proposal $\hat{p}_t$ wherever a reverse construction is used. Assume that the sample-wise targets are square-integrable. Viewed as regression over square-integrable velocity fields, all exact tangential-update losses introduced in Sections~\ref{subsec:16}--\ref{subsec:18}---whether based on the forward or reverse construction and whether expressed in covariance, gradient, or unified form---share the same population minimizer,
    \[
    v^\rho_t(x_t)+\eta\Gamma_t^{\rho,\tilde r^\rho}(x_t),
    \]
    unique up to a.e. equality.
    This statement applies to the losses defined in~\eqref{eq:67}, \eqref{eq:68}, \eqref{eq:85}, \eqref{eq:86}, \eqref{eq:98}, \eqref{eq:99}, \eqref{eq:109}, \eqref{eq:110}, \eqref{eq:113}, and~\eqref{eq:114}.
\end{theorem}

\begin{proof}
    For every loss covered by this theorem, the conditional mean of its sample-wise target, including any auxiliary path randomness, is $v^\rho_t(x)+\eta\Gamma_t^{\rho,\tilde r^\rho}(x)$ by the identities established in Sections~\ref{subsec:16}--\ref{subsec:18}. Since an expected squared-error loss is uniquely minimized up to a.e. equality by the conditional mean of its target, the stated conclusion follows.
\end{proof}

Theorem~\ref{thm:9} shows that the population minimizer is invariant across several implementation choices. These choices can be organized along three distinct design dimensions: 
\begin{itemize}
    \item Target realization, which may take the covariance form, the gradient form, or a control-variate combination of the two;
    \item Sampling construction, which may be forward or reverse and determines how the noisy state and posterior endpoint are generated;
    \item Representation-specific computational realization, namely ODE-based or SDE-based evaluation of $\log\frac{\rho}{\mu}$ for the covariance form, and kernel-based or adjoint-based (Mayer or Bolza) calculation for the gradient form.
\end{itemize}
Representative exact Newton Matching configurations are summarized in Table~\ref{tab:1}.

\subsection{Multi-Time Supervision from Shared Samples}\label{subsec:19}

Sections~\ref{subsec:16}--\ref{subsec:18} primarily develop the exact tangential-update losses from the population perspective. Accordingly, both the text and the algorithms describe each loss in terms of the basic sample-wise objects associated with a given \(X_t\): an endpoint pair \((X_t,X_1)\), or, when the target is path-dependent, a path pair \((X_t,\BY_{[t,1]})\). This sample-wise presentation does not imply that a separate endpoint or path must be generated for each supervision time \(t\). In practice, both endpoint generation and target computation can be relatively expensive. Under the forward construction, sampling \(X_1\sim\rho\) requires solving the ODE~\eqref{eq:1} or SDE~\eqref{eq:5} associated with \(\rho\). Under the reverse construction, sampling \(X_1\sim p_{1|t}^\rho(\cdot|X_t)\) requires solving the posterior-preserving SDE~\eqref{eq:64}. The covariance form further requires ODE-based or SDE-based estimation of the regularized reward \(\tilde{r}^\rho(X_1)\), whereas the gradient form requires solving an adjoint ODE. Repeating these operations independently for every supervision time \(t\) and its corresponding state \(X_t\) would discard information already available from the same endpoint or path.

Given the practical importance of sample efficiency, we now make explicit how a single endpoint \(X_1\), or a single posterior-compatible path ending at \(X_1\), can provide supervision at multiple times. Consider an ordered collection of supervision times
\begin{equation}
    0<t_0<t_1<\cdots<t_{N_{\mathrm{time}}-1}<t_{N_{\mathrm{time}}}:=1.
\label{eq:115}
\end{equation}
For an endpoint-based target, it is sufficient that, for every \(j=0,1,\ldots,N_{\mathrm{time}}-1\),
\begin{equation}\label{eq:116}
    X_1| X_{t_j}
    \sim
    p_{1| t_j}^\rho(\cdot| X_{t_j}).
\end{equation}
For a pathwise target, it is sufficient that
\begin{equation}\label{eq:117}
    \BY_{[t_j,1]}| Y_{t_j}=X_{t_j}
    \sim
    \mathbb P_{[t_j,1]| t_j}^\rho(\cdot| X_{t_j}).
\end{equation}
These are precisely the posterior and conditional-path requirements in~\eqref{eq:60} and~\eqref{eq:95}, imposed separately at each supervision time. Sharing an endpoint or path generally induces correlation among the resulting regression pairs across supervision times, but this correlation does not affect their validity: exactness requires the appropriate conditional law at each time, not independence across times.

This subsection develops representative shared-sample constructions tailored to the sampling and target-computation procedures of the covariance and gradient forms. Each construction produces regression pairs at multiple supervision times from a single endpoint or path sample, with the associated endpoint-level or path-level computations shared across those pairs.

\subsubsection{Multi-Time Supervision in Covariance Form}

\paragraph{ODE-based estimation.}
Suppose first that the ODE-based estimation method in Section~\ref{subsec:16} is used to evaluate \(\log\frac{\rho(X_1)}{\mu(X_1)}\). If \(X_1\sim\rho\) is generated by the anchor ODE under the forward construction, the log density or log-density ratio can be computed together with the endpoint through the augmented integrations~\eqref{eq:71} or~\eqref{eq:72}. If \(X_1\) is generated by an SDE, either under the forward construction or by the posterior-preserving SDE under the reverse construction, the same quantity can be computed from the obtained endpoint through the backward integrations~\eqref{eq:73} or~\eqref{eq:74}. In either case, this gives one exact endpoint value
\begin{equation*}
    \tilde r^\rho(X_1)
    =
    r(X_1)
    -
    \frac{1}{\tau}
    \log\frac{\rho(X_1)}{\mu(X_1)}
\end{equation*}
for the covariance target~\eqref{eq:65}. The remaining task is therefore to construct as many inexpensive noisy states $(t,X_t)$ as possible, each of which can reuse the same value \(\tilde r^\rho(X_1)\) for supervision.

Under the forward construction, one direct choice is to sample \(X_0\sim p_0\) once and set
\begin{equation*}
    X_{t_j}
    =
    \alpha_{t_j}X_1+\beta_{t_j}X_0.
\end{equation*}
Alternatively, one may sample several source variables \(X_{0}^{(j)}\iidsim p_0\), and set
\begin{equation*}
    X_{t_j}
    =
    \alpha_{t_j}X_1+\beta_{t_j}X_{0}^{(j)}.
\end{equation*}
For every fixed \(j\), both choices reproduce~\eqref{eq:62}, so the posterior requirement~\eqref{eq:116} is satisfied. The supervision times can be sampled independently
\[
t_0,\,t_1,\,\cdots,\,t_{N_\mathrm{time}-1}\iidsim \U(0,1),
\]
or 
\begin{equation}\label{eq:118}
    t_j
    \sim
    \U\left(
        \frac{j}{N_{\mathrm{time}}},
        \frac{j+1}{N_{\mathrm{time}}}
    \right),
    \qquad
    j=0,1,\ldots,N_{\mathrm{time}}-1.
\end{equation}
In the former case, the notation means that \(N_{\mathrm{time}}\) times are first sampled i.i.d. from \(\U(0,1)\) and then relabeled to comply with~\eqref{eq:115}.

\begin{algorithm}[!p]
    \caption{Tangential Update: Multi-Time Supervision with Shared Samples.}
    \label{alg:4}
    \begin{algorithmic}[1]
    \REQUIRE At stage $k$, the current canonical model $v^{\rho_k}\in\velmancan$; a stepsize $\eta_k\in(0,\tau]$; a reference factor $\mu=\rho^\base$ for fine-tuning (with the pretrained model $v^\base$) or $\mu=1$ for sampling; a reward function $r:\R^d\to\R$; an inverse temperature $\tau>0$; a batch size $N_\mathrm{batch}$; and a number of supervision times $N_\mathrm{time}$.
    \ENSURE The updated velocity model $\bar{v}^{k+1}=v^{\rho_k}+\eta_k\Gamma^{\rho_k,\tilde{r}^{\rho_k}}$.
    \STATE Warm-start $\theta$ with the network parameters of $v^{\rho_k}$.
    \FOR{each gradient step}
        \STATE \COMMENT{The following operations are applied to a batch.}
        \STATE Sample $0<t_0<t_1<\cdots<t_{N_\mathrm{time}}=1$.
        \IF{forward construction is used}
            \STATE Sample $X_1\sim\rho_k$ through ODE \eqref{eq:1} or SDE \eqref{eq:5} of $v^{\rho_k}$.
            \STATE Sample $X_0\sim\Normal(0,I)$.
            \IF{covariance form and ODE-based estimation are used}
                \STATE For $j\gets 0,1,\cdots,N_\mathrm{time}-1$, set $X_{t_j}\gets\alpha_{t_j}X_1+\beta_{t_j}X_0$.
            \ELSE
                \STATE Set $X_{t_0}\gets\alpha_{t_0}X_1+\beta_{t_0}X_0$.
                \STATE Sample $\BY_{[t_0,1]}\sim\mathbb{P}_{[t_0,1]}^\mathrm{uni}(\cdot|X_{t_0},X_1)$ based on Theorem \ref{thm:14}.
                \STATE For $j\gets 1,2,\cdots,N_\mathrm{time}-1$, set $X_{t_j}\gets Y_{t_j}$.
            \ENDIF
        \ELSIF{reverse construction is used}
            \STATE Sample $X_{t_0}\sim \hat{p}_{t_0}$, and sample $\BY_{[{t_0},1]}\sim \mathbb{P}_{[{t_0},1]|{t_0}}^{\rho_k}(\cdot|X_{t_0})$ through the posterior-preserving SDE \eqref{eq:64} of $\rho_k$.
            \STATE For $j\gets 1,2,\cdots,N_\mathrm{time}$, set $X_{t_j}\gets Y_{t_j}$.
        \ENDIF
        \IF{covariance form is used}
            \IF{ODE-based estimation is used}
                \STATE Compute $\log\frac{\rho_k(X_1)}{\mu(X_1)}$ using applicable branches in Lines \ref{line:1}, \ref{line:2}, and \ref{line:3} of Algorithm \ref{alg:2}.
                \STATE Based on the same value $\log\frac{\rho_k(X_1)}{\mu(X_1)}$, for $j\gets 0,1,\cdots,N_\mathrm{time}-1$, compute the target $\target_j$ at $X_{t_j}$ using Line \ref{line:5} of Algorithm \ref{alg:2}.
            \ELSIF{SDE-based estimation is used}
                \STATE For $j\gets N_\mathrm{time}-1,N_\mathrm{time}-2,\cdots,0$, recursively compute $\log\frac{\rho_k(X_1)}{\rho^\mu(X_1)}-\log\frac{p_t^{\rho_k}(X_t)}{p_t^{\mu}(X_t)}$ and the target $\target_j$ at $X_{t_j}$ backward along $\BY_{[{t_0},1]}$ using Lines \ref{line:4} and \ref{line:5} of Algorithm \ref{alg:2}.
            \ENDIF
        \ELSIF{gradient form is used}
            \STATE For $j\gets N_\mathrm{time}-1,N_\mathrm{time}-2,\cdots,0$, recursively compute the pathwise adjoint $\lambda_{t_j}(\BY_{[{t_0},1]})$ and the target $\target_j$ at $X_{t_j}$ backward along $\BY_{[{t_0},1]}$ using applicable branches in Lines \ref{line:6}--\ref{line:7} of Algorithm \ref{alg:3}.
        \ENDIF
        \STATE Update $\theta$ based on the batch loss $\mathcal{L}\gets\frac{1}{N_{\mathrm{batch}}N_{\mathrm{time}}}\sum_{(t_j,X_{t_j},\target_j)}\norm[2]{v_{t_j}^\theta(X_{t_j})-\target_j}^2$.%
    \ENDFOR
    \STATE Set $\bar{v}^{k+1}\gets v^\theta$.
\end{algorithmic}
\end{algorithm}

Under the reverse construction, we first sample \(t_0\sim\U(0,1)\) and \(X_{t_0}\sim\hat p_{t_0}\). We then generate
\begin{equation*}
    \BY_{[t_0,1]}
    \sim
    \mathbb P_{[t_0,1]| t_0}^\rho(\cdot| X_{t_0})
\end{equation*}
using the posterior-preserving SDE~\eqref{eq:64}, and set \(X_1:=Y_1\).

To avoid restricting supervision to a fixed time grid, the intermediate supervision times \(t_j\) may be sampled according to~\eqref{eq:118}, with the SDE solution evaluated at the resulting times. In practice, we may impose a small lower cutoff \(\varepsilon>0\) and require \(t_0\geq\varepsilon\) to avoid the singularity of the posterior-preserving SDE at \(t=0\).

The endpoint value \(\tilde r^\rho(X_1)\) is computed once using either backward integration~\eqref{eq:73} or~\eqref{eq:74}. Since the complete path \(\BY_{[t_0,1]}\) has already been generated in obtaining \(X_1\), the states at the supervision times can be taken directly from this path:
\begin{equation*}
    X_{t_j}:=Y_{t_j}.
\end{equation*}
By the Markov property, each tail path \(\BY_{[t_j,1]}\) satisfies~\eqref{eq:117}, which in turn implies~\eqref{eq:116}. Thus, a single posterior-preserving SDE trajectory \(\BY_{[t_0,1]}\), together with a single backward ODE-based evaluation of \(\tilde r^\rho(X_1)\), provides multiple exact covariance targets~\eqref{eq:65}, one at each supervision time.

\paragraph{SDE-based estimation.}
Although the regularized reward \(\tilde r^\rho(X_1)\) in the covariance target~\eqref{eq:65} depends only on the endpoint \(X_1\), its SDE-based estimation requires a path. Specifically, in~\eqref{eq:80}, the quantity
\[
    \tilde r^\rho(X_1)
    +
    \frac{1}{\tau}
    \log\frac{p_t^\rho(X_t)}{p_t^\mu(X_t)}
\]
is evaluated along a path \(\BY_{[t,1]}\) satisfying \(Y_t=X_t\) and \(Y_1=X_1\). Thus, independently generating multiple endpoint pairs \((X_t,X_1)\) does not allow the pathwise computation to be shared across supervision times. Instead, the states \(X_{t_j}\) should be taken from a single posterior-compatible path. The supervision times \(t_j\) should be sampled according to~\eqref{eq:118}.

Under the forward construction, after sampling \(X_1\sim\rho\), \(X_0\sim p_0\), and the earliest supervision time \(t_0\), we construct the initial state \(X_{t_0}\) and then sample a single universal bridge using~\eqref{eq:82}:
\begin{equation}\label{eq:119}
    X_{t_0}
    =
    \alpha_{t_0}X_1+\beta_{t_0}X_0,
    \qquad
    \BY_{[t_0,1]}
    \sim
    \mathbb P_{[t_0,1]}^{\mathrm{uni}}
    (\cdot| X_{t_0},X_1).
\end{equation}
The forward construction of \((X_{t_0},X_1)\), together with the sampling decomposition~\eqref{eq:81}, ensures that the resulting path satisfies the requirement~\eqref{eq:117} at \(t_0\). Under the reverse construction, a posterior-compatible path is obtained directly by sampling
\begin{equation}\label{eq:120}
    X_{t_0}\sim\hat p_{t_0},
    \qquad
    \BY_{[t_0,1]}
    \sim
    \mathbb P_{[t_0,1]| t_0}^\rho
    (\cdot| X_{t_0})
\end{equation}
using the posterior-preserving SDE. Under either construction, we set
\begin{equation*}
    X_{t_j}:=Y_{t_j},
    \qquad
    j=0,1,\ldots,N_{\mathrm{time}}-1.
\end{equation*}
The Markov property then guarantees~\eqref{eq:117}, and hence~\eqref{eq:116}, at every supervision time \(t_j\). Again, in practice, we may impose a lower cutoff \(\varepsilon>0\) by requiring \(t_0\geq\varepsilon\).

The SDE-based evaluation in~\eqref{eq:80} can be carried out at all supervision times through a single backward recursion along the shared path. At the terminal time \(t_{N_{\mathrm{time}}}=1\), the recursion is initialized with
\[
    r(X_1)-B_1(X_1).
\]
Proceeding backward, for \(j=N_{\mathrm{time}}-1,\ldots,0\), the value at \(t_j\) is obtained from its value at \(t_{j+1}\) by adding
\[
    B_{t_{j+1}}(X_{t_{j+1}})
    -B_{t_j}(X_{t_j})
    -
    \frac{1}{\tau}
    \int_{t_j}^{t_{j+1}}
    \frac{
        \norm[2]{
            v_s^\rho(Y_s)-v_s^\mu(Y_s)
        }^2
    }{
        \kappa_s
    }
    \dd s
    -
    \frac{1}{\tau}
    \int_{t_j}^{t_{j+1}}
    \sqrt{
        \frac{2}{\kappa_s}
    }
    \left(
        v_s^\rho(Y_s)-v_s^\mu(Y_s)
    \right)^\top
    \dd W_s.
\]
Consequently, the covariance targets at all supervision times can be evaluated through a single backward recursion along \(\BY_{[t_0,1]}\).

\subsubsection{Multi-Time Supervision in Gradient Form}\label{subsec:20}

Under both the Mayer and Bolza realizations, the gradient form requires a pathwise adjoint to evaluate the sample-wise target. For both forward and reverse constructions, the supervision times \(t_j\), the states \(X_{t_j}\), and the shared posterior-compatible path \(\BY_{[t_0,1]}\) should be sampled and constructed in the same manner as in the covariance form with SDE-based estimation. Notably, a single backward solution of the Mayer adjoint ODE~\eqref{eq:96} or the Bolza adjoint ODE~\eqref{eq:102} along the full trajectory, from \(t=1\) to \(t=t_0\), yields the adjoint value at every supervision time \(t_j\). More precisely, the value obtained at \(t_j\) from the full backward solve coincides with that obtained by solving the adjoint separately along the tail path \(\BY_{[t_j,1]}\):
\begin{align*}
    \lambda_{t_j}^{b^\rho,\tilde r^\rho,0}
    \left(
        \BY_{[t_0,1]}
    \right)
    =
    \lambda_{t_j}^{b^\rho,\tilde r^\rho,0}
    \left(
        \BY_{[t_j,1]}
    \right),
    \qquad
    \lambda_{t_j}^{b^\rho,r,l^{\rho,\mu}}
    \left(
        \BY_{[t_0,1]}
    \right)
    =
    \lambda_{t_j}^{b^\rho,r,l^{\rho,\mu}}
    \left(
        \BY_{[t_j,1]}
    \right).
\end{align*}
Indeed, for each \(j\), the restriction of the full adjoint solution to \([t_j,1]\) satisfies the same terminal condition and the same backward ODE along the same tail path as the adjoint solved separately on \([t_j,1]\). The identities above therefore follow from uniqueness of the terminal-value problem. Hence, the adjoint values obtained from a single full-trajectory solve can be inserted directly into the Mayer target~\eqref{eq:97} or the Bolza target~\eqref{eq:108} at every supervision time, without solving a separate adjoint ODE for each \(t_j\). Consequently, the gradient-form targets at all supervision times can be evaluated from a single backward solution of the corresponding adjoint ODE along \(\BY_{[t_0,1]}\), with the pathwise adjoint computation shared across the resulting regression pairs.

\begin{remark}
A notable consequence of bridge universality is that, under the forward construction, the Mayer and Bolza realizations of the gradient form do not require simulation of the posterior-preserving SDE. Although the adjoint ODEs~\eqref{eq:96} and~\eqref{eq:102} are derived from the initial-state sensitivity of that SDE, the required posterior-compatible path can be generated by a different procedure. Specifically, one may sample \(X_1\sim\rho\) using the anchor ODE, construct \(X_{t_0}\) from the forward noising kernel, sample
\[
    \BY_{[t_0,1]}
    \sim
    \mathbb P_{[t_0,1]}^{\mathrm{uni}}
    (\cdot| X_{t_0},X_1)
\]
as in~\eqref{eq:119} using the analytic Gaussian transitions in~\eqref{eq:82}, and then solve the corresponding adjoint ODE backward along this path. Since the universal bridge is exactly the endpoint-conditioned path law of the posterior-preserving SDE, this procedure yields the same Mayer and Bolza targets as those in~\eqref{eq:98} and~\eqref{eq:109}. Thus, all endpoint samples from the anchor model may be generated by ODE sampling, while the auxiliary stochastic path is sampled analytically rather than through a numerical rollout of the posterior-preserving SDE. Together with multi-time supervision, one such path and one backward adjoint solve provide the adjoint values at all supervision times. This yields a particularly interesting algorithm: although adjoints are used in the training targets, all samples used for training can be generated solely by ODE sampling, so the entire training procedure requires no SDE rollout of the trained model.

This observation separates the origin of the adjoint from the mechanism used to generate the path along which it is evaluated. Adjoint Matching~\citep{domingo2025adjoint} introduces adjoints through a stochastic-optimal-control formulation and evaluates them along paths induced by the SDE associated with the trained model. In our derivation, by contrast, the adjoint provides a matrix-free representation of the initial-state sensitivity underlying the posterior sensitivity kernel, as expressed in~\eqref{eq:93} and~\eqref{eq:94}. The pathwise adjoint ODE may be evaluated along any prescribed path, while exactness of the regression target is ensured by sampling that path from the required posterior-compatible law. Thus, the use of an adjoint in the training target is not intrinsically tied either to a stochastic-optimal-control derivation or to the simulation of SDE paths from a trained model. Moreover, the required sensitivity cannot generally be recovered by directly differentiating through the sampling procedure used here. Under the forward construction, \(X_1\) is sampled before \(X_{t_0}\), after which the universal bridge is sampled conditioning on both endpoints. Consequently, the implemented sampling graph contains no model-generated map from \(X_{t_0}\) to \(X_1\) whose derivative coincides with the state-transition matrix of the posterior-preserving SDE.

The same separation applies to the critical-point-consistent reference-adjoint approximation in Section~\ref{sub2sec:14}. Under the forward construction, the current posterior-compatible path required in~\eqref{eq:136} and~\eqref{eq:138} may again be generated from an ODE-sampled endpoint and the universal bridge. The reference adjoint \(\lambda^{b^\mu,r,0}\) is then solved backward along this path according to~\eqref{eq:134}. Consequently, the reference-adjoint approximation can likewise be implemented without a numerical SDE rollout of the trained model, while retaining the critical-point consistency established in Proposition~\ref{prop:14}.
\end{remark}

The multi-time supervision schemes developed in this subsection are representative rather than exhaustive and provide natural ways to improve sample efficiency. Other valid schemes may also exist. Algorithm~\ref{alg:4} summarizes one representative scheme for each configuration considered above.

\subsection{Canonicalization}\label{subsec:21}

The preceding subsections present several exact realizations of the tangential update \(\bar v=v^\rho+\eta\Gamma^{\rho,\tilde r^\rho}\) at a canonical anchor \(v^\rho\). However, \(\bar v\) is not canonical in general. To return to the canonical manifold and proceed to the next stage of Newton Matching, we need to canonicalize \(\bar v\), that is, to find a canonical velocity field \(v^q\) whose terminal density matches that induced by \(\bar v\). In this subsection, we first present a straightforward canonicalization procedure based on the standard CFM loss, which we call \emph{explicit canonicalization}. We then discuss an interesting idea, called \emph{implicit canonicalization}, which sidesteps this need by carefully constructing the tangential-update loss so that the noncanonical field \(\bar v\) can serve as the anchor.

\paragraph{Explicit canonicalization via CFM.}
Let \(q=\terminalmap(\bar v)\) denote the terminal density induced by \(\bar v\). We can generate endpoint samples \(X_1\sim q\) using the ODE driven by \(\bar v\). To construct a canonical velocity field \(v^q\) with terminal density \(q\), we can utilize the standard CFM loss. Specifically, after sampling \(X_1\sim q\), we independently draw \(X_0\sim p_0\), set \(X_t=\alpha_tX_1+\beta_tX_0\), and train a parameterized velocity field \(v^\theta\) using the loss
\begin{equation}
    \calL^{\mathrm{can}}(\theta)
    :=
    \E_{\substack{
        t\sim\U(0,1),\,X_1\sim q,\,X_0\sim p_0,\,
        X_t=\alpha_tX_1+\beta_tX_0}}
    \left[
        \norm[2]{
            v_t^\theta(X_t)
            -
            v_{t| 1}(X_t| X_1)
        }^2
    \right].
    \label{eq:121}
\end{equation}
In practice, \(v^\theta\) can be warm-started from the parameters representing \(\bar v\). The population minimizer of~\eqref{eq:121} is the canonical velocity field \(v^q=\canonicalmap(q)=\projection(\bar v)\). Once training is complete, we replace the current anchor \(v^\rho\) with \(v^q\) and proceed to the next stage of Newton Matching. The explicit canonicalization is summarized in Algorithm~\ref{alg:5}.

\begin{algorithm}[!t]
    \caption{Explicit Canonicalization.}
    \label{alg:5}
    \begin{algorithmic}[1]
    \REQUIRE At stage $k$, the updated velocity model $\bar{v}^{k+1}\in\velspace$; and a batch size $N_\mathrm{batch}$.%
    \ENSURE The canonical velocity model $v^{\rho_{k+1}}=\projection\left(\bar{v}^{k+1}\right)$.
    \STATE Warm-start $\theta$ with the network parameters of $\bar{v}^{k+1}$.
    \FOR{each gradient step}
        \STATE \COMMENT{The following operations are applied to a batch.}
        \STATE Sample $X_1\sim\rho_{k+1}=\terminalmap\left(\bar{v}^{k+1}\right)$ through ODE \eqref{eq:1} of $\bar{v}^{k+1}$.
        \STATE Sample $t\sim\U(0,1)$ and $X_0\sim\Normal(0,I)$, and set $X_t\gets\alpha_tX_1+\beta_tX_0$.
        \STATE Update $\theta$ based on the batch loss $\mathcal{L}\gets\frac{1}{N_{\mathrm{batch}}}\sum_{(t,X_t,X_1)}\norm[2]{v_t^\theta(X_t)-v_{t|1}(X_t|X_1)}^2$.
    \ENDFOR
    \STATE Set $v^{\rho_{k+1}}\gets v^\theta$.
\end{algorithmic}
\end{algorithm}

\paragraph{Implicit canonicalization.}
Sometimes, we may prefer to avoid the explicit canonicalization procedure. Suppose that \(\bar v\) is a noncanonical velocity field with terminal density \(q=\terminalmap(\bar v)\). At the next stage of Newton Matching, our goal is to perform a tangential update at the canonical anchor \(v^q\), yielding the updated velocity field \(v^q+\eta\Gamma^{q,\tilde r^q}\). The question is whether this goal can be achieved using \(\bar v\) as the endpoint-generation field, without first explicitly constructing \(v^q\). The answer is yes, subject to additional restrictions.

The forward construction provides a way to sidestep \(v^q\). We first sample \(X_1\sim q\) using the ODE driven by \(\bar v\). Applying~\eqref{eq:62} with \(\rho=q\) yields
\begin{equation*}
    X_1| X_t
    \sim
    p_{1| t}^q(\cdot| X_t).
\end{equation*}
Thus, even though \(\bar v\) may be noncanonical, the \((X_t,X_1)\) pair satisfies the posterior requirement.

For the covariance form, replace the canonical anchor \(v_t^q\) in~\eqref{eq:65} by \(\bar v_t\), giving
\begin{align*}
    &v_{t|1}(X_t|X_1)
    +
    \eta
    \left(
        \tilde r^q(X_1)-B_t(X_t)
    \right)
    \left(
        v_{t|1}(X_t|X_1)-\bar v_t(X_t)
    \right).
\end{align*}
For every fixed \((t,x_t)\), its conditional mean is
\begin{align*}
    &\E_{X_1\sim p_{1|t}^q(\cdot|x_t)}
    \left[
        v_{t|1}(x_t|X_1)
        +
        \eta
        \left(
            \tilde r^q(X_1)-B_t(x_t)
        \right)
        \left(
            v_{t|1}(x_t|X_1)-\bar v_t(x_t)
        \right)
    \right]
    \notag\\
    =&
    v_t^q(x_t)
    +
    \eta\Gamma_t^{q,\tilde r^q}(x_t)
    +
    \eta
    \left(
        V_t^q[\tilde r^q](x_t)-B_t(x_t)
    \right)
    \left(
        v_t^q(x_t)-\bar v_t(x_t)
    \right).
\end{align*}
In particular, the exact value baseline
\begin{equation*}
    B_t(x_t)
    =
    V_t^q[\tilde r^q](x_t)
\end{equation*}
removes the dependence of the conditional mean on the noncanonical \(\bar v_t\), yielding \(v_t^q(x_t)+\eta\Gamma_t^{q,\tilde r^q}(x_t)\). Moreover, the term \(\log\frac{q(X_1)}{\mu(X_1)}\) in \(\tilde r^q\) can be evaluated using the ODE-based method, which does not require \(\bar v\) to be canonical.

We note that obtaining the exact value baseline may be challenging and may require training a separate neural network.
With an approximate baseline, the conditional-mean error is
\begin{equation*}
    \eta
    \left(
        V_t^q[\tilde r^q](x_t)-B_t(x_t)
    \right)
    \left(
        v_t^q(x_t)-\bar v_t(x_t)
    \right).
\end{equation*}
Thus, the error in the tangential update arising from the use of the noncanonical \(\bar v\) is given by the product of the value-function error and the canonicality defect.

\section{Approximate Newton Matching}\label{sec:6}

At the population level, the exact realizations developed in Section~\ref{sec:5} recover the ideal Newton Matching stage introduced in Section~\ref{subsec:11}. As established in Section~\ref{sec:4}, they retain the finite-stepsize reverse-KL descent, global convergence under mild conditions, and local quadratic convergence of the full-step iteration. These exact realizations, however, involve several nontrivial computations. The regularized reward
\[
\tilde{r}^\rho(x)
=
r(x)-\frac{1}{\tau}\log\frac{\rho(x)}{\mu(x)}
\]
contains either the log-density term $\log\rho$ when $\mu=1$, or the log-density-ratio term $\log\frac{\rho}{\rho^\base}$ when $\mu=\rho^\base$. In the covariance form, this correction term must be evaluated at the sampled endpoints, whereas in the gradient form, its gradient is computed through adjoint equations. When computational resources are limited, it is therefore natural to trade population-level exactness at each stage for reduced computational cost.

In this section, we develop approximate Newton Matching, making this trade-off explicit. Rather than requiring the matching objective to recover the exact tangent vector \(\Gamma^{\rho,\tilde r^\rho}\), we replace it with a computationally cheaper surrogate 
\[
    \widehat{\Gamma}^\rho\approx\Gamma^{\rho,\tilde r^\rho}.
\]
Throughout Sections~\ref{subsec:22}--\ref{subsec:24}, we develop principled choices of \(\widehat{\Gamma}^\rho\).
We define the resulting approximate Newton Matching stage by
\begin{subequations}\label{eq:122}
    \begin{align}
        &\hat v
        :=
        v^\rho+\eta\widehat{\Gamma}^\rho,
        \label{eq:123}\\
        &q
        =
        \terminalmap(\hat v),
        \,
        v^q
        =
        \projection(\hat v)
        =
        \canonicalmap(q).
        \label{eq:124}
    \end{align}
\end{subequations}
Thus, only the tangential update is approximated, while canonicalization remains exact. Exact Newton Matching is recovered by setting \(\widehat{\Gamma}^\rho=\Gamma^{\rho,\tilde r^\rho}\).
For notational simplicity, we write \(\pi:=\pi_{\mu,\tau,r}\). At a given stage, we fix a canonical anchor \(v^\rho=\canonicalmap(\rho)\in\velmancan\) and choose a scalar stepsize \(\eta\in(0,\tau]\).

As expected, an approximate direction need not coincide with the exact Newton tangent vector at a generic anchor. As a target-correctness requirement, however, it should preserve the intended critical point. This motivates the following definition.
\begin{definition}\label{def:2}
A method \eqref{eq:122} is \emph{critical-point consistent} for $\pi$ if
\[
    \widehat{\Gamma}^\rho=0
    \quad\Leftrightarrow\quad
    \rho=\pi.
\]
\end{definition}
Since $\eta>0$, critical-point consistency is equivalently expressed as $\hat v=v^\rho$ if and only if $\rho=\pi$ for the ambient update~\eqref{eq:123}. We impose this condition before canonicalization. Requiring the displacement \(\widehat{\Gamma}^\rho\) itself to vanish at the target density $\pi$ is stronger than requiring the complete stage~\eqref{eq:122} to preserve the terminal density there, since the terminal-density map $\terminalmap$ is many-to-one. According to Proposition~\ref{prop:7}, the exact Newton Matching is critical-point consistent.

Sections~\ref{subsec:22} and~\ref{subsec:23} develop two covariance approximations and two gradient approximations, respectively. These four methods preserve critical-point consistency. Since the density-ratio correction in the regularized reward $\tilde{r}^\rho$ is a major source of computational challenge, Section~\ref{subsec:24} examines the trade-offs that arise when this term is approximated or removed entirely. In the fully unregularized case, the update is driven by the raw reward $r$ and admits an exact finite-stepsize reward-ascent certificate. Throughout these three subsections, the prescribed update direction is realized stagewise and followed by exact canonicalization. Method-specific proofs are collected in Appendix~\ref{app:6}.

Section~\ref{subsec:25} then develops a complementary viewpoint. Rather than realizing the additive update~\eqref{eq:123} stage by stage, we consider its fixed-point condition directly. Forcing the displacement to vanish gives
\[
    v^\rho
    =
    v^\rho+\eta\widehat{\Gamma}^\rho.
\]
Since the right-hand side admits a conditional-expectation representation, we can construct a matching-style stop-gradient objective whose population-stationary points are exactly the fixed points of the additive update rule.

The constructions in this section are representative rather than exhaustive. The four critical-point-consistent examples illustrate only a subset of the possible choices. Additionally, if computational efficiency is strongly prioritized, the ideas in this section can be taken further to develop more aggressive approximations.

Some approximate constructions require $\mu$ to be a normalized reference density, since they draw samples from the corresponding posterior $p_{1|t}^\mu(\cdot|x_t)$. When $\mu=1$, this requirement can be satisfied through the following \textit{auxiliary-reference factorization}. Choose any normalized auxiliary reference density $\hat\mu\in\pdfspace$, such as a Gaussian. Define the auxiliary reward
\begin{equation}\label{eq:125}
    \hat r(x)
    :=
    r(x)-\frac{1}{\tau}\log\hat\mu(x).
\end{equation}
The unnormalized target factor is preserved pointwise:
\[
    \hat\mu(x)e^{\tau\hat r(x)}
    =
    e^{\tau r(x)}.
\]
Consequently, the pairs $(1,r)$ and $(\hat\mu,\hat r)$ define exactly the same target density
\[
    \pi_{1,\tau,r}
    =
    \pi_{\hat\mu,\tau,\hat r},
\]
while the latter provides a normalized reference density whose posterior can be readily sampled from. Any construction requiring such a reference density may therefore be applied to $(\hat\mu,\tau,\hat r)$ without changing the sampling target. If $\hat\mu$ is Gaussian, Proposition~\ref{prop:26} gives a closed-form expression for the canonical velocity field $v^{\hat\mu}$, which can be computed efficiently. Whenever this auxiliary-reference factorization is used later, we relabel $(\hat\mu,\hat r)$ as $(\mu,r)$ for notational simplicity.

\subsection{Approximate Covariance Forms with Critical-Point Consistency}\label{subsec:22}

The exact covariance form represents the Newton tangential update as
\[
    \bar{v}_t(x_t)
    =
    v_t^\rho(x_t)
    +
    \eta\Cov_{X_1\sim p_{1|t}^{\rho}(\cdot|x_t)}
    \left(
    v_{t|1}(x_t|X_1),
    \tilde{r}^\rho(X_1)
    \right).
\]
In this representation, the computationally challenging quantity is the logarithmic component $\log\frac{\rho(X_1)}{\mu(X_1)}$ in the regularized reward $\tilde{r}^\rho(X_1)$. Both approximations below first rewrite this logarithmic component through posterior density ratios and then replace it by a linearized form. Both constructions preserve critical-point consistency. Their proofs are given in Appendix~\ref{subsec:51}.

\subsubsection{Direct Linearization}\label{sub2sec:12}

The first construction rewrites the exact Newton tangent vector through the posterior log-density ratio between the target and the current anchor. Since
\[
    \tilde r^\rho(x)
    =
    \frac{1}{\tau}\log\frac{\pi(x)}{\rho(x)}+\const,
\]
the additive constant does not contribute to the posterior covariance. Moreover, by Bayes' rule,
\[
    \frac{p_{1|t}^{\pi}(x_1|x_t)}{p_{1|t}^{\rho}(x_1|x_t)}
    =
    \frac{\pi(x_1)p_{t|1}(x_t|x_1)}{p_t^\pi(x_t)}
    \frac{p_t^\rho(x_t)}{\rho(x_1)p_{t|1}(x_t|x_1)}
    =
    \frac{p_t^\rho(x_t)\pi(x_1)}{p_t^\pi(x_t)\rho(x_1)}.
\]
The factor depending only on $(t,x_t)$ disappears inside the posterior covariance. Consequently, we have
\begin{align}
    \Gamma_t^{\rho,\tilde r^\rho}(x_t)
    =&
    \frac{1}{\tau}
    \Cov_{X_1\sim p_{1|t}^{\rho}(\cdot|x_t)}
    \left(
        v_{t|1}(x_t|X_1),
        \log\frac{\pi(X_1)}{\rho(X_1)}
    \right)
    \notag\\
    =&
    \frac{1}{\tau}
    \Cov_{X_1\sim p_{1|t}^{\rho}(\cdot|x_t)}
    \left(
        v_{t|1}(x_t|X_1),
        \log\frac{p_{1|t}^{\pi}(X_1|x_t)}{p_{1|t}^{\rho}(X_1|x_t)}
    \right).
    \label{eq:126}
\end{align}

The posterior ratio equals one when $\rho=\pi$. We linearize the logarithmic term:
\[
    \log\frac{p_{1|t}^{\pi}(X_1|x_t)}{p_{1|t}^{\rho}(X_1|x_t)}
    \approx
    \frac{p_{1|t}^{\pi}(X_1|x_t)}{p_{1|t}^{\rho}(X_1|x_t)}-1.
\]
This linearization is exact at the target and retains the first-order variation of the posterior log-density ratio nearby. Substituting it into~\eqref{eq:126} gives
\begin{align*}
    \bar v_t(x_t)
    \approx&
    v_t^\rho(x_t)
    +
    \frac{\eta}{\tau}
    \Cov_{X_1\sim p_{1|t}^{\rho}(\cdot|x_t)}
    \left(
        v_{t|1}(x_t|X_1),
        \frac{p_{1|t}^{\pi}(X_1|x_t)}{p_{1|t}^{\rho}(X_1|x_t)}
    \right)
    \\
    =&
    v_t^\rho(x_t)
    +
    \frac{\eta}{\tau}
    \E_{X_1\sim p_{1|t}^{\rho}(\cdot|x_t)}
    \left[
        \frac{p_{1|t}^{\pi}(X_1|x_t)}{p_{1|t}^{\rho}(X_1|x_t)}
        \left(
            v_{t|1}(x_t|X_1)-v_t^\rho(x_t)
        \right)
    \right]
    \\
    =&
    v_t^\rho(x_t)
    +
    \frac{\eta}{\tau}
    \E_{X_1\sim p_{1|t}^{\pi}(\cdot|x_t)}
    \left[
        v_{t|1}(x_t|X_1)-v_t^\rho(x_t)
    \right]
    \\
    =&
    \left(1-\frac{\eta}{\tau}\right)v_t^\rho(x_t)
    +
    \frac{\eta}{\tau}v_t^\pi(x_t).
\end{align*}
Therefore, we approximate the tangential update through direct linearization:
\begin{subequations}
    \label{eq:127}
    \begin{align}
        &\widehat{\Gamma}_t^{\rho,\mathrm{dir}\text{-}\mathrm{lin}}(x_t)
        :=
        \frac{1}{\tau}
        \left(
            v_t^\pi(x_t)-v_t^\rho(x_t)
        \right),\\
        &\hat v_t^{\mathrm{dir}\text{-}\mathrm{lin}}(x_t)
        :=
        v_t^\rho(x_t)
        +
        \eta\widehat{\Gamma}_t^{\rho,\mathrm{dir}\text{-}\mathrm{lin}}(x_t)
        =
        \left(1-\frac{\eta}{\tau}\right)v_t^\rho(x_t)
        +
        \frac{\eta}{\tau}v_t^\pi(x_t).
    \end{align}
\end{subequations}
Thus, the posterior-ratio linearization produces a convex interpolation in canonical-velocity coordinates. It changes the exact Newton tangent vector at a generic anchor, while the full step $\eta=\tau$ directly targets $v^\pi$.

\begin{algorithm}[!t]
    \caption{Tangential Update: Approximate Realization with Direct Linearization.}
    \label{alg:6}
    \begin{algorithmic}[1]
    \REQUIRE At stage $k$, the current canonical model $v^{\rho_k}\in\velmancan$; a stepsize $\eta_k\in(0,\tau]$; a reference factor $\mu=\rho^\base$ for fine-tuning (with the pretrained model $v^\base$) or $\mu=1$ for sampling; a reward function $r:\R^d\to\R$; an inverse temperature $\tau>0$; and a batch size $N_\mathrm{batch}$.
    \ENSURE The updated velocity model $\hat{v}^{k+1}\approx v^{\rho_k}+\eta_k\Gamma^{\rho_k,\tilde{r}^{\rho_k}}$.
    \IF{$\mu=1$}
        \STATE Set $\rho^\base\gets\Normal(m,\sigma^2I)$ and its analytical velocity field $v^\base$ based on the auxiliary-reference factorization \eqref{eq:125}.
        \STATE Set $r\gets r-\frac{1}{\tau}\log\rho^\base$.
    \ENDIF
    \STATE Warm-start $\theta$ with the network parameters of $v^{\rho_k}$.
    \FOR{each gradient step}
        \STATE \COMMENT{The following operations are applied to a batch.}
            \IF{forward construction is used}
                \IF{$\mu=1$}
                    \STATE Sample $X_1\sim\rho^\base=\Normal(m,\sigma^2I)$.
                \ELSIF{ODE-based sampling is used}
                    \STATE Sample $X_1\sim\rho^\base$ through ODE \eqref{eq:1} of $v^{\base}$.
                \ELSIF{SDE-based sampling is used}
                    \STATE Sample $X_1\sim\rho^\base$ through SDE \eqref{eq:5} of $v^{\base}$.
                \ENDIF
                \STATE Sample $t\sim\U(0,1)$ and $X_0\sim\Normal(0,I)$, and set $X_t\gets\alpha_tX_1+\beta_tX_0$.
            \ELSIF{reverse construction is used}
                \STATE Sample $t\sim\U(0,1)$ and $X_t\sim \hat{p}_t$.
                \IF{$\mu=1$}
                    \STATE Sample $X_1\sim p_{1|t}^\base(\cdot|X_t)=\Normal\left(\frac{\alpha_t\sigma^2}{\alpha_t^2\sigma^2+\beta_t^2}X_t+\frac{\beta_t^2}{\alpha_t^2\sigma^2+\beta_t^2}m,\frac{\beta_t^2\sigma^2}{\alpha_t^2\sigma^2+\beta_t^2}I\right)$.
                \ELSIF{$\mu=\rho^\base$}
                    \STATE Sample $X_1\sim p_{1|t}^\base(\cdot|X_t)$ through the posterior-preserving SDE \eqref{eq:64} of $\rho^\base$.
                \ENDIF
            \ENDIF
        \STATE Compute the target $\target$ by \eqref{eq:129}.
        \STATE Update $\theta$ based on the batch loss $\mathcal{L}\gets\frac{1}{N_{\mathrm{batch}}}\sum_{(t,X_t,\target)}\norm[2]{v_t^\theta(X_t)-\target}^2$.
    \ENDFOR
    \STATE Set $\hat{v}^{k+1}\gets v^\theta$.
\end{algorithmic}
\end{algorithm}

The field $v^\pi$ is not directly available, so the remaining task is to realize~\eqref{eq:127} through matching-style regression. Assume that $\mu=\rho^\base\in\pdfspace$ is normalized; when the original reference is $\mu=1$, we apply the auxiliary-reference factorization~\eqref{eq:125}. The target posterior is then an exponential tilt of the reference posterior:
\begin{equation}
    p_{1|t}^{\pi}(x_1|x_t)
    \propto
    e^{\tau r(x_1)}p_{1|t}^{\base}(x_1|x_t).
    \label{eq:128}
\end{equation}
The conditional normalizing constant in~\eqref{eq:128} depends on $(t,x_t)$ and is generally unavailable. Nevertheless, the approximate tangential update in~\eqref{eq:127} admits the following stop-gradient realization under the joint distribution from the reference:
\begin{align}
    &\calL_{\rho,\eta,B}^{\mathrm{dir}\text{-}\mathrm{lin}}(\theta)
    :=
    \E_{\substack{
    t\sim\U(0,1),\,(X_t,X_1)\sim\Xi_{t,1}^\base}}
    \bigg[
    \Big\|
    v_t^\theta(X_t)
    -
    \left[
        \left(1-\frac{\eta}{\tau}\right)v_t^\rho(X_t)
        +
        \frac{\eta}{\tau}v_t^\base(X_t)
    \right]
    \notag\\
    &\quad
    -
    \left(
        e^{\tau(r(X_1)-B_t(X_t))}-1
    \right)
    \Big[
        \left(1-\frac{\eta}{\tau}\right)v_t^\rho(X_t)
        +
        \frac{\eta}{\tau}v_{t|1}(X_t|X_1)
        -
        \sg\left(v_t^\theta(X_t)\right)
    \Big]
    \Big\|_2^2
    \bigg].
    \label{eq:129}
\end{align}
Here, $\sg$ denotes the stop-gradient operator, and $B_t(X_t)$ is an arbitrary scalar baseline. We adopt the notation in~\eqref{eq:60} and use the shorthand $\Xi_{t,1}^\base:=\Xi_{t,1}^{\rho^\base}$ for simplicity. At a fixed noisy state, subtracting $B_t(X_t)$ rescales all exponential weights by the same factor and therefore does not change the population-stationary point. It can nevertheless improve numerical conditioning and reduce variance. A useful choice is
\[
    B_t(x_t)
    =
    \E_{X_1\sim p_{1|t}^\base(\cdot|x_t)}[r(X_1)].
\]

The computation of approximate tangential update with direct linearization is summarized in Algorithm~\ref{alg:6}. Section~\ref{sub2sec:19} presents a general formula for deriving the population-stationary points of stop-gradient objectives, and Appendix~\ref{subsec:51} applies the formula to prove the following proposition.

\begin{restatable}
    {proposition}{targetPosteriorConsistency}
\label{prop:12}
Assume that $\mu=\rho^\base\in\pdfspace$ is a normalized reference density and the loss~\eqref{eq:129} is integrable. The unique population-stationary point of the loss~\eqref{eq:129}, up to a.e. equality, is $\hat v^{\mathrm{dir}\text{-}\mathrm{lin}}$ in~\eqref{eq:127}. Furthermore, the direct linearization is critical-point consistent for $\pi$, i.e., $\widehat{\Gamma}^{\rho,\mathrm{dir}\text{-}\mathrm{lin}}=0$ if and only if $\rho=\pi$.
\end{restatable}

\subsubsection{Split Linearization}\label{sub2sec:13}

The preceding construction realizes the target posterior through a positive exponential tilt of the reference posterior. A complementary construction starts from the current posterior and introduces the normalized negative-tilt density
\[
    \rho^-(x)
    :=
    \frac{\rho(x)e^{-\tau r(x)}}
    {\int_{\R^d}\rho(z)e^{-\tau r(z)}\dd z}.
\]
We assume that \(\rho\) is a normalizable density in $\pdfspace$, satisfying
\[
    \int_{\R^d}\rho(z)e^{-\tau r(z)}\dd z
    <
    \infty.
\]
We also assume that $\mu=\rho^\base\in\pdfspace$ is normalized, applying the auxiliary-reference factorization~\eqref{eq:125} when the original reference is $\mu=1$.

The role of $\rho^-$ is to express the target-to-current posterior ratio through two ratios centered at the current posterior. By Bayes' rule,
\begin{subequations}
    \begin{align}
        &\frac{p_{1|t}^{\pi}(X_1|x_t)}{p_{1|t}^{\base}(X_1|x_t)}=\frac{\pi(x_1)p_{t|1}(x_t|x_1)}{p_t^\pi(x_t)}\frac{p_t^\base(x_t)}{\rho^\base(x_1)p_{t|1}(x_t|x_1)}=\frac{p_t^\base(x_t)e^{\tau r(x_1)}}{p_t^\pi(x_t)\E_{X_1\sim\rho^\base}\left[e^{\tau r(X_1)}\right]},\label{eq:130}\\
        &\frac{p_{1|t}^{\rho}(X_1|x_t)}{p_{1|t}^{\rho^-}(X_1|x_t)}=\frac{\rho(x_1)p_{t|1}(x_t|x_1)}{p_t^\rho(x_t)}\frac{p_t^{\rho^-}(x_t)}{\rho^-(x_1)p_{t|1}(x_t|x_1)}=\frac{p_t^{\rho^-}(x_t)e^{\tau r(x_1)}}{p_t^\rho(x_t)\E_{X_1\sim\rho^-}\left[e^{\tau r(X_1)}\right]}.\label{eq:131}
    \end{align}
\end{subequations}
The factors depending only on $(t,x_t)$ disappear inside the posterior covariance. Combining~\eqref{eq:130} and~\eqref{eq:131} with~\eqref{eq:126} gives
\begin{align*}
    \Gamma_t^{\rho,\tilde r^\rho}(x_t)
    =&
    \frac{1}{\tau}
    \Cov_{X_1\sim p_{1|t}^{\rho}(\cdot|x_t)}
    \left(
        v_{t|1}(x_t|X_1),
        \log\frac{p_{1|t}^{\base}(X_1|x_t)}{p_{1|t}^{\rho^-}(X_1|x_t)}
    \right)
    \\
    =&
    \frac{1}{\tau}
    \Cov_{X_1\sim p_{1|t}^{\rho}(\cdot|x_t)}
    \left(
        v_{t|1}(x_t|X_1),
        \log\frac{p_{1|t}^{\base}(X_1|x_t)}{p_{1|t}^{\rho}(X_1|x_t)}
        -
        \log\frac{p_{1|t}^{\rho^-}(X_1|x_t)}{p_{1|t}^{\rho}(X_1|x_t)}
    \right).
\end{align*}

\begin{algorithm}[!t]
    \caption{Tangential Update: Approximate Realization with Split Linearization.}
    \label{alg:7}
    \begin{algorithmic}[1]
    \REQUIRE At stage $k$, the current canonical model $v^{\rho_k}\in\velmancan$; a stepsize $\eta_k\in(0,\tau]$; a reference factor $\mu=\rho^\base$ for fine-tuning (with the pretrained model $v^\base$) or $\mu=1$ for sampling; a reward function $r:\R^d\to\R$; an inverse temperature $\tau>0$; and a batch size $N_\mathrm{batch}$.
    \ENSURE The updated velocity model $\hat{v}^{k+1}\approx v^{\rho_k}+\eta_k\Gamma^{\rho_k,\tilde{r}^{\rho_k}}$.
    \IF{$\mu=1$}
        \STATE Set $\rho^\base\gets\Normal(m,\sigma^2I)$ and its analytical velocity field $v^\base$ based on the auxiliary-reference factorization \eqref{eq:125}.
        \STATE Set $r\gets r-\frac{1}{\tau}\log\rho^\base$.
    \ENDIF
    \STATE Warm-start $\theta$ with the network parameters of $v^{\rho_k}$.
    \FOR{each gradient step}
        \STATE \COMMENT{The following operations are applied to a batch.}
            \IF{forward construction is used}
                \IF{ODE-based sampling is used}
                    \STATE Sample $X_1\sim\rho_k$ through ODE \eqref{eq:1} of $v^{\rho_k}$.
                \ELSIF{SDE-based sampling is used}
                    \STATE Sample $X_1\sim\rho_k$ through SDE \eqref{eq:5} of $v^{\rho_k}$.
                \ENDIF
                \STATE Sample $t\sim\U(0,1)$ and $X_0\sim\Normal(0,I)$, and set $X_t\gets\alpha_tX_1+\beta_tX_0$.
            \ELSIF{reverse construction is used}
                \STATE Sample $t\sim\U(0,1)$ and $X_t\sim \hat{p}_t$.
                \STATE Sample $X_1\sim p_{1|t}^{\rho_k}(\cdot|X_t)$ through the posterior-preserving SDE \eqref{eq:64} of $\rho_k$.
            \ENDIF
        \STATE Compute the target $\target$ by \eqref{eq:133}.
        \STATE Update $\theta$ based on the batch loss $\mathcal{L}\gets\frac{1}{N_{\mathrm{batch}}}\sum_{(t,X_t,\target)}\norm[2]{v_t^\theta(X_t)-\target}^2$.
    \ENDFOR
    \STATE Set $\hat{v}^{k+1}\gets v^\theta$.
\end{algorithmic}
\end{algorithm}

We linearize both logarithmic terms:
\[
    \log\frac{p_{1|t}^{\base}(X_1|x_t)}{p_{1|t}^{\rho}(X_1|x_t)}
    \approx
    \frac{p_{1|t}^{\base}(X_1|x_t)}{p_{1|t}^{\rho}(X_1|x_t)}-1,
    \qquad
    \log\frac{p_{1|t}^{\rho^-}(X_1|x_t)}{p_{1|t}^{\rho}(X_1|x_t)}
    \approx
    \frac{p_{1|t}^{\rho^-}(X_1|x_t)}{p_{1|t}^{\rho}(X_1|x_t)}-1.
\]
Substituting the linearizations into the covariance representation yields
\begin{align*}
    \bar v_t(x_t)
    \approx&
    v_t^\rho(x_t)
    +
    \frac{\eta}{\tau}
    \Cov_{X_1\sim p_{1|t}^{\rho}(\cdot|x_t)}
    \left(
        v_{t|1}(x_t|X_1),
        \frac{p_{1|t}^{\base}(X_1|x_t)}{p_{1|t}^{\rho}(X_1|x_t)}
        -
        \frac{p_{1|t}^{\rho^-}(X_1|x_t)}{p_{1|t}^{\rho}(X_1|x_t)}
    \right)
    \\
    =&
    v_t^\rho(x_t)
    +
    \frac{\eta}{\tau}
    \left(
        \E_{X_1\sim p_{1|t}^{\base}(\cdot|x_t)}
        [v_{t|1}(x_t|X_1)-v_t^\rho(x_t)]
        -
        \E_{X_1\sim p_{1|t}^{\rho^-}(\cdot|x_t)}
        [v_{t|1}(x_t|X_1)-v_t^\rho(x_t)]
    \right)
    \\
    =&
    v_t^\rho(x_t)
    +
    \frac{\eta}{\tau}
    \left(
        v_t^\base(x_t)-v_t^{\rho^-}(x_t)
    \right).
\end{align*}
Accordingly, we approximate the tangential update through split linearization:
\begin{subequations}
    \label{eq:132}
    \begin{align}
        &\widehat{\Gamma}_t^{\rho,\mathrm{spl}\text{-}\mathrm{lin}}(x_t)
        :=
        \frac{1}{\tau}
        \left(
            v_t^\base(x_t)-v_t^{\rho^-}(x_t)
        \right),\\
        &\hat v_t^{\mathrm{spl}\text{-}\mathrm{lin}}(x_t)
        :=
        v_t^\rho(x_t)
        +
        \eta\widehat{\Gamma}_t^{\rho,\mathrm{spl}\text{-}\mathrm{lin}}(x_t)
        =
        v_t^\rho(x_t)
        +
        \frac{\eta}{\tau}
        \left(
            v_t^\base(x_t)-v_t^{\rho^-}(x_t)
        \right).
    \end{align}
\end{subequations}
Unlike~\eqref{eq:127}, this approximation is expressed as a correction to the current anchor rather than an interpolation with $v^\pi$. It involves the canonical velocity of the negative-tilt density.

The approximate tangential update in~\eqref{eq:132} admits the following stop-gradient realization under the joint distribution from the current anchor:
\begin{align}
    &\calL_{\rho,\eta,B}^{\mathrm{spl}\text{-}\mathrm{lin}}(\theta)
    :=
    \E_{\substack{
    t\sim\U(0,1),\,(X_t,X_1)\sim\Xi_{t,1}^\rho}}
    \bigg[
    \Big\|
    v_t^\theta(X_t)
    -
    \left[
        \left(1-\frac{\eta}{\tau}\right)v_t^\rho(X_t)
        +
        \frac{\eta}{\tau}v_t^\base(X_t)
    \right]
    \notag\\
    &\quad
    -
    \left(
        e^{-\tau(r(X_1)-B_t(X_t))}-1
    \right)
    \Big[
        v_t^\rho(X_t)
        -
        \frac{\eta}{\tau}
        \left(
            v_{t|1}(X_t|X_1)-v_t^\base(X_t)
        \right)
        -
        \sg\left(v_t^\theta(X_t)\right)
    \Big]
    \Big\|_2^2
    \bigg].
    \label{eq:133}
\end{align}
Similar to the direct-linearization case, we adopt the notation in~\eqref{eq:60}.
$B_t(X_t)$ is an arbitrary scalar baseline, with a useful choice
\[
    B_t(X_t)
    =
    \E_{X_1\sim p_{1|t}^\rho(\cdot|X_t)}[r(X_1)].
\]

The computation of the approximate tangential update with split linearization is summarized in Algorithm~\ref{alg:7}. Section~\ref{sub2sec:19} presents a general formula for deriving the population-stationary points of stop-gradient objectives, and Appendix~\ref{subsec:51} applies the formula to prove the following proposition.

\begin{restatable}
    {proposition}{negativeTiltConsistency}
\label{prop:13}
Assume that $\rho^-\in\pdfspace$ is normalizable, $\mu=\rho^\base\in\pdfspace$ is a normalized reference density, and the loss~\eqref{eq:133} is integrable. The unique population-stationary point of the loss~\eqref{eq:133}, up to a.e. equality, is $\hat v^{\mathrm{spl}\text{-}\mathrm{lin}}$ in~\eqref{eq:132}. Furthermore, the split linearization is critical-point consistent for $\pi$, i.e., $\widehat{\Gamma}^{\rho,\mathrm{spl}\text{-}\mathrm{lin}}=0$ if and only if $\rho=\pi$.
\end{restatable}

The two covariance approximations therefore use complementary exponential reweightings. Direct linearization samples from the reference posterior and applies a positive reward tilt, whereas split linearization samples from the current posterior and applies a negative reward tilt. Their updates may differ given the same anchor \(v^\rho\), but they are both critical-point consistent.

\subsection{Approximate Gradient Forms with Critical-Point Consistency}\label{subsec:23}

The exact gradient form represents the Newton tangential update as
\[
    \bar v_t(x_t)
    =
    v_t^\rho(x_t)
    +
    \eta\frac{\alpha_t\kappa_t}{\beta_t^2}
    \E_{X_1\sim p_{1|t}^{\rho}(\cdot|x_t)}
    \left[
        \Lambda_t^\rho(X_1|x_t)^\top
        \nabla\tilde r^\rho(X_1)
    \right].
\]
This form avoids direct evaluation of the scalar log-density ratio $\log\frac{\rho}{\mu}$, but it requires the action of an exact posterior Stein kernel on the regularized-reward gradient. Section~\ref{sec:5} provides exact realizations through adjoint calculus of the posterior-preserving SDE. The approximations below reduce the computational cost in two different ways. Section~\ref{sub2sec:14} uses the adjoint ODE associated with the reference in place of that associated with the current anchor. Section~\ref{sub2sec:15} approximates the posterior Stein kernel by a positive-definite matrix field, which corresponds to treating the anchor posterior as Gaussian. These two approximations use samples from the current density $\rho$, which is essential to the critical-point consistency.

\subsubsection{Reference-Adjoint Approximation}\label{sub2sec:14}

The exact Bolza realization in Section~\ref{subsec:17} uses the path distribution $\mathbb{P}_{[t,1]|t}^\rho(\cdot|x_t)$ of the posterior-preserving SDE associated with the current anchor, together with the pathwise adjoint $\lambda_s^{b^\rho,r,l^{\rho,\mu}}$. Its adjoint ODE \eqref{eq:103} contains two terms: the first arises from the Jacobian of the drift in the posterior-preserving SDE associated with the current anchor and therefore involves \(\nabla v_s^\rho\); the second involves the gradient \(\nabla l_s^{\rho,\mu}\) of the running regularization objective.

In the reference-adjoint approximation, every occurrence of the current anchor \(v_s^\rho\) in the adjoint ODE is replaced by the reference velocity field \(v_s^\mu\). Recall that, when \(\mu=1\), \(v_s^1(x)=\frac{\dot\alpha_s}{\alpha_s}x\), whereas, when \(\mu=\rho^\base\), \(v_s^\mu=v_s^\base\) is the pretrained velocity field. The adjoint calculation is therefore frozen at the reference model:
\[
    \lambda_s^{b^\rho,r,l^{\rho,\mu}}
    \approx
    \lambda_s^{b^\mu,r,l^{\mu,\mu}}
    =
    \lambda_s^{b^\mu,r,0},
\]
where
\[
    b_s^\mu(x)
    =
    2v_s^\mu(x)-\frac{\dot\alpha_s}{\alpha_s}x
\]
is the reference drift. Under this approximation, the running-objective term vanishes since \(l_s^{\mu,\mu}\equiv0\). Moreover, the adjoint ODE no longer depends on the current anchor and therefore remains the same across stages; only the path along which it is evaluated changes.

For a prescribed path $\BY_{[t,1]}$, the reference adjoint is solved backward through
\begin{subequations}\label{eq:134}
    \begin{align}
        &\frac{\dd}{\dd s}
        \lambda_s^{b^\mu,r,0}(\BY_{[t,1]})
        =
        -\nabla b_s^\mu(Y_s)^\top
        \lambda_s^{b^\mu,r,0}(\BY_{[t,1]}),
        \qquad s\in[t,1],
        \label{eq:135}\\
        &\lambda_1^{b^\mu,r,0}(\BY_{[t,1]})
        =
        \nabla r(Y_1).
    \end{align}
\end{subequations}
The adjoint ODE is defined along the prescribed path and does not determine how that path is sampled. It is important to retain the path distribution $\mathbb P_{[t,1]|t}^{\rho}(\cdot|x_t)$ of the posterior-preserving SDE associated with the current anchor and replace only the adjoint dynamics.

Substituting the reference-adjoint approximation into the exact Bolza realization~\eqref{eq:105} and~\eqref{eq:107}, we obtain
\begin{align*}
    \Gamma_t^{\rho,\tilde r^\rho}(x_t)
    \approx
    \kappa_t
    \E_{\BY_{[t,1]}\sim
    \mathbb P_{[t,1]|t}^{\rho}(\cdot|x_t)}
    \left[
        \lambda_t^{b^\mu,r,0}(\BY_{[t,1]})
    \right]
    -
    \frac{1}{\tau}
    \left(
        v_t^\rho(x_t)-v_t^\mu(x_t)
    \right).
\end{align*}
Accordingly, the reference-adjoint approximation for the tangential update is given by:
\begin{subequations}
    \label{eq:136}
\begin{align}
    &\widehat{\Gamma}_t^{\rho,\mathrm{ref}\text{-}\mathrm{adj}}(x_t)
    :=
    \kappa_t
    \E_{\BY_{[t,1]}\sim
    \mathbb P_{[t,1]|t}^{\rho}(\cdot|x_t)}
    \left[
        \lambda_t^{b^\mu,r,0}(\BY_{[t,1]})
    \right]
    -
    \frac{1}{\tau}
    \left(
        v_t^\rho(x_t)-v_t^\mu(x_t)
    \right),\\
    &\hat v_t^{\mathrm{ref}\text{-}\mathrm{adj}}(x_t)
    :=
    v_t^\rho(x_t)
    +
    \eta\widehat{\Gamma}_t^{\rho,\mathrm{ref}\text{-}\mathrm{adj}}(x_t)\notag\\
    &\hphantom{\hat v_t^{\mathrm{ref}\text{-}\mathrm{adj}}(x_t)}\,
    =
    \left(1-\frac{\eta}{\tau}\right)v_t^\rho(x_t)
    +
    \frac{\eta}{\tau}v_t^\mu(x_t)
    +
    \eta\kappa_t
    \E_{\BY_{[t,1]}\sim
    \mathbb P_{[t,1]|t}^{\rho}(\cdot|x_t)}
    \left[
        \lambda_t^{b^\mu,r,0}(\BY_{[t,1]})
    \right].\label{eq:137}
\end{align}
\end{subequations}
The corresponding loss is
\begin{align}
    \calL_{\rho,\eta}^{\mathrm{ref}\text{-}\mathrm{adj}}(\theta)
    :=
    \E_{\substack{
    t\sim\U(0,1),\,
    \BY_{[t,1]}\sim\Xi_{[t,1]}^{\rho}}}
    \bigg[
        \Big\|&
        v_t^\theta(Y_t)
        -
        \left(1-\frac{\eta}{\tau}\right)v_{t|1}(Y_t|Y_1)
        \notag\\
        &
        -
        \frac{\eta}{\tau}v_t^\mu(Y_t)
        -
        \eta\kappa_t
        \lambda_t^{b^\mu,r,0}(\BY_{[t,1]})
        \Big\|_2^2
    \bigg],
    \label{eq:138}
\end{align}
where the path distribution $\Xi_{[t,1]}^{\rho}$ satisfies~\eqref{eq:95}.
The reference-adjoint approximation deliberately pairs the path distribution associated with the current anchor with adjoint dynamics determined by the reference velocity field. This reduces the computational cost of the adjoint calculation, but, when the current terminal density differs from the target, the resulting population minimizer generally differs from that of exact Newton Matching. The following proposition shows that this discrepancy vanishes at the target and that the reference-adjoint approximation is critical-point consistent.

\begin{algorithm}[!t]
    \caption{Tangential Update: Approximate Realization with Reference Adjoint.}
    \label{alg:8}
    \begin{algorithmic}[1]
    \REQUIRE At stage $k$, the current canonical model $v^{\rho_k}\in\velmancan$; a stepsize $\eta_k\in(0,\tau]$; a reference factor $\mu=\rho^\base$ for fine-tuning (with the pretrained model $v^\base$) or $\mu=1$ for sampling; a reward function $r:\R^d\to\R$; an inverse temperature $\tau>0$; and a batch size $N_\mathrm{batch}$.
    \ENSURE The updated velocity model $\hat{v}^{k+1}\approx v^{\rho_k}+\eta_k\Gamma^{\rho_k,\tilde{r}^{\rho_k}}$.
    \STATE Warm-start $\theta$ with the network parameters of $v^{\rho_k}$.
    \FOR{each gradient step}
        \STATE \COMMENT{The following operations are applied to a batch.}
            \IF{forward construction is used}
                \IF{ODE-based sampling is used}
                    \STATE Sample $X_1\sim\rho_k$ through ODE \eqref{eq:1} of $v^{\rho_k}$.
                \ELSIF{SDE-based sampling is used}
                    \STATE Sample $X_1\sim\rho_k$ through SDE \eqref{eq:5} of $v^{\rho_k}$.
                \ENDIF
                \STATE Sample $t\sim\U(0,1)$ and $X_0\sim\Normal(0,I)$, and set $X_t\gets\alpha_tX_1+\beta_tX_0$.
                \IF{$\mu=\rho^\base$}
                    \STATE Sample $\BY_{[t,1]}\sim\mathbb{P}_{[t,1]}^\mathrm{uni}(\cdot|X_t,X_1)$ based on Theorem \ref{thm:14}.
                \ENDIF
            \ELSIF{reverse construction is used}
                \STATE Sample $t\sim\U(0,1)$ and $X_t\sim \hat{p}_t$.
                \STATE Sample $\BY_{[t,1]}\sim \mathbb{P}_{[t,1]|t}^{\rho_k}(\cdot|X_t)$ through the posterior-preserving SDE \eqref{eq:64} of $\rho_k$.
                \STATE Set $X_1\gets Y_1$.
            \ENDIF
        \IF{$\mu=\rho^\base$}
            \STATE Compute the pathwise adjoint $\lambda_t^{b^\mu,r,0}(\BY_{[t,1]})$ by ODE \eqref{eq:134}.
            \STATE Compute the target $\target$ by \eqref{eq:138}.
        \ELSIF{$\mu=1$}
            \STATE Compute the target $\target$ by \eqref{eq:139}.
        \ENDIF
        \STATE Update $\theta$ based on the batch loss $\mathcal{L}\gets\frac{1}{N_{\mathrm{batch}}}\sum_{(t,X_t,\target)}\norm[2]{v_t^\theta(X_t)-\target}^2$.
    \ENDFOR
    \STATE Set $\hat{v}^{k+1}\gets v^\theta$.
\end{algorithmic}
\end{algorithm}

\begin{restatable}
    {proposition}{referenceSensitivityConsistency}
\label{prop:14}
Assume that the loss~\eqref{eq:138} is integrable. The unique population minimizer of the loss~\eqref{eq:138}, up to a.e. equality, is $\hat v^{\mathrm{ref}\text{-}\mathrm{adj}}$ in~\eqref{eq:136}. Furthermore, the reference-adjoint approximation is critical-point consistent for $\pi$, i.e., $\widehat{\Gamma}^{\rho,\mathrm{ref}\text{-}\mathrm{adj}}=0$ if and only if $\rho=\pi$.
\end{restatable}

The reference-adjoint approximation applies to both \(\mu=\rho^\base\) and \(\mu=1\). The case \(\mu=1\) admits a further closed-form simplification. Since
\[
    v_s^{1}(x)
    =
    \frac{\dot\alpha_s}{\alpha_s}x,
    \qquad
    b_s^1(x)
    =
    \frac{\dot\alpha_s}{\alpha_s}x,
\]
the reference-adjoint equation~\eqref{eq:135} reduces to
\[
    \frac{\dd}{\dd s}
    \lambda_s^{b^1,r,0}(\BY_{[t,1]})
    =
    -\frac{\dot\alpha_s}{\alpha_s}
    \lambda_s^{b^1,r,0}(\BY_{[t,1]})
    \quad\Leftrightarrow\quad
    \frac{\dd}{\dd s}
    \left(
        \alpha_s\lambda_s^{b^1,r,0}(\BY_{[t,1]})
    \right)
    \equiv
    0.
\]
Using \(\alpha_1=1\) and the terminal condition
\(\lambda_1^{b^1,r,0}(\BY_{[t,1]})=\nabla r(Y_1)\), the pathwise adjoint therefore admits the closed-form expression
\[
    \lambda_t^{b^1,r,0}(\BY_{[t,1]})
    =
    \frac{1}{\alpha_t}\nabla r(Y_1).
\]
Consequently, no backward adjoint integration is required in this case. Substituting this expression into~\eqref{eq:138} yields
\begin{align}
    \calL_{\rho,\eta}^{\mathrm{ref}\text{-}\mathrm{adj}}(\theta)
    :=
    \E_{\substack{
        t\sim\U(0,1),\\
        (X_t,X_1)\sim\Xi_{t,1}^{\rho}}}
    \left[
        \norm[2]{
            v_t^\theta(X_t)
            -
            \left(1-\frac{\eta}{\tau}\right)
            v_{t|1}(X_t|X_1)
            -
            \frac{\eta\dot{\alpha}_t}{\tau\alpha_t}X_t
            -
            \frac{\eta\kappa_t}{\alpha_t}\nabla r(X_1)
        }^2
    \right].
    \label{eq:139}
\end{align}

The reference-adjoint approximation is summarized in Algorithm~\ref{alg:8}.

\subsubsection{Gaussian-Kernel Approximation}\label{sub2sec:15}

The Mayer and Bolza realizations of the gradient form in Section~\ref{subsec:17} arise from the posterior sensitivity kernel, a particular exact posterior Stein kernel \(\Lambda_t^\rho(x_1|x_t)\) constructed from the initial-state sensitivity of the posterior-preserving SDE. The reference-adjoint approximation modifies the adjoint dynamics used to evaluate this particular kernel. A complementary approach is to work directly with the general posterior Stein kernel and approximate it without relying on the sensitivity-based construction.

For each fixed \((t,x_t)\), an exact posterior Stein kernel must satisfy condition~\eqref{eq:88}. Solving this equation is generally intractable for a generic, potentially complex posterior density \(p_{1|t}^\rho(\cdot|x_t)\). A convenient simplification is to approximate this posterior by a Gaussian distribution:
\begin{equation}
    \hat p_{1|t}^\rho(\cdot|x_t)
    :=
    \Normal
    \left(
        \widehat M_t^\rho(x_t),
        \widehat\Sigma_t^\rho(x_t)
    \right).
    \label{eq:140}
\end{equation}
For this Gaussian surrogate, replacing \(p_{1|t}^\rho\) with \(\hat p_{1|t}^\rho\) in~\eqref{eq:88} yields an equation with an explicit solution \(\widehat\Sigma_t^\rho(x_t)\). This motivates the direct approximation
\[
    \Lambda_t^\rho(x_1|x_t)
    \approx
    \widehat\Sigma_t^\rho(x_t),
\]
where \(\widehat\Sigma_t^\rho(x_t)\in\S_{++}^d\) is the covariance matrix of the Gaussian surrogate. Properties related to Gaussian Stein kernels and posterior moments are collected in Appendix~\ref{subsec:50}.

The matrix field \(\widehat\Sigma_t^\rho(x_t)\) may be learned by a separate model or prescribed analytically. Since it does not depend on \(x_1\), it can be pulled out of the conditional expectation over $X_1$:
\[
\E_{X_1\sim p_{1|t}^{\rho}(\cdot|x_t)}
    \left[
        \widehat\Sigma_t^\rho(x_t)
        \nabla\tilde r^\rho(X_1)
    \right]=
   \widehat\Sigma_t^\rho(x_t)\E_{X_1\sim p_{1|t}^{\rho}(\cdot|x_t)}
    \left[
        \nabla\tilde r^\rho(X_1)
    \right].
\]
Therefore, the tangential update with the Gaussian-kernel approximation is given by:
\begin{subequations}
    \begin{align}
        &\widehat{\Gamma}_t^{\rho,\mathrm{Gau}\text{-}\mathrm{ker}}(x_t)
        :=
        \frac{\alpha_t\kappa_t}{\beta_t^2}
        \widehat\Sigma_t^\rho(x_t)
        \E_{X_1\sim p_{1|t}^\rho(\cdot|x_t)}
        \left[
            \nabla\tilde r^\rho(X_1)
        \right],\\
        &\hat v_t^{\mathrm{Gau}\text{-}\mathrm{ker}}(x_t)
        :=
        v_t^\rho(x_t)
        +
        \eta
        \widehat{\Gamma}_t^{\rho,\mathrm{Gau}\text{-}\mathrm{ker}}(x_t).
        \label{eq:141}
    \end{align}
\end{subequations}

\begin{remark}
The Gaussian surrogate~\eqref{eq:140} is used only to approximate the posterior Stein kernel. Solving condition~\eqref{eq:88} for this surrogate yields the tractable matrix field \(\widehat\Sigma_t^\rho(x_t)\), which is then used in place of the exact kernel in the gradient form. The endpoint \(X_1\), however, should still be sampled from the exact posterior of the current anchor, \(p_{1|t}^\rho(\cdot| x_t)\), to ensure the critical-point consistency established below.

A cruder approximation is to also sample the endpoint \(X_1\) from the Gaussian surrogate~\eqref{eq:140}. Although this may further reduce computational cost and could be useful empirically, it will generally break critical-point consistency.
\end{remark}

Expanding the regularized-reward gradient in~\eqref{eq:141} gives
\begin{align*}
    \hat v_t^{\mathrm{Gau}\text{-}\mathrm{ker}}(x_t)
    =&
    v_t^\rho(x_t)
    +
    \eta
    \frac{\alpha_t\kappa_t}{\beta_t^2}
    \widehat\Sigma_t^\rho(x_t)
    \E_{X_1\sim p_{1|t}^\rho(\cdot|x_t)}
    \left[
        \nabla r(X_1)
        -
        \frac{1}{\tau}\nabla\log\frac{\rho(X_1)}{\mu(X_1)}
    \right]
    \\
    =&
    v_t^\rho(x_t)
    +
    \eta
    \frac{\alpha_t\kappa_t}{\beta_t^2}
    \widehat\Sigma_t^\rho(x_t)
    \E_{X_1\sim p_{1|t}^\rho(\cdot|x_t)}
    \left[
        \nabla r(X_1)
        +
        \frac{1}{\tau}\nabla\log\mu(X_1)
        -
        \frac{1}{\tau}\nabla\log\rho(X_1)
    \right].
\end{align*}
The endpoint score $\nabla\log\rho(X_1)$ need not be evaluated. The posterior score has zero conditional mean, which yields the endpoint-score identity
\[
    \E_{X_1\sim p_{1|t}^\rho(\cdot|x_t)}
    \left[
        \nabla\log\rho(X_1)
        +
        \frac{\alpha_t}{\beta_t^2}
        \left(
            x_t-\alpha_tX_1
        \right)
    \right]
    =
    0.
\]
Then the approximate update can be rewritten as
\begin{align*}
    &\hat v_t^{\mathrm{Gau}\text{-}\mathrm{ker}}(x_t)\\
    =&
    v_t^\rho(x_t)
    +
    \eta
    \frac{\alpha_t\kappa_t}{\beta_t^2}
    \widehat\Sigma_t^\rho(x_t)
    \E_{X_1\sim p_{1|t}^\rho(\cdot|x_t)}
    \left[
        \nabla r(X_1)
        +
        \frac{1}{\tau}\nabla\log\mu(X_1)
        +
        \frac{\alpha_t}{\tau\beta_t^2}
        \left(
            x_t-\alpha_tX_1
        \right)
    \right].
\end{align*}
Consequently,~\eqref{eq:141} is realized by the regression loss
\begin{align}
    \calL_{\rho,\eta,\widehat{\Sigma}}^{\mathrm{Gau}\text{-}\mathrm{ker}}(\theta)
    :=&
    \E_{\substack{
    t\sim\U(0,1),\,(X_t,X_1)\sim \Xi_{t,1}^{\rho}}}
    \Bigg[
        \bigg\|
        v_t^\theta(X_t)
        -
        v_{t|1}(X_t|X_1)
        \notag\\
    &
        -
        \eta\frac{\alpha_t\kappa_t}{\beta_t^2}
        \widehat\Sigma_t^\rho(X_t)
        \left(
            \nabla r(X_1)
            +
            \frac{1}{\tau}\nabla\log\mu(X_1)
            +
            \frac{\alpha_t}{\tau\beta_t^2}
            \left(
                X_t-\alpha_tX_1
            \right)
        \right)
        \bigg\|_2^2
    \Bigg].
    \label{eq:142}
\end{align}
The following proposition shows that the Gaussian-kernel approximation is critical-point consistent. Its procedure is summarized in Algorithm~\ref{alg:9}.

\begin{algorithm}[!t]
    \caption{Tangential Update: Approximate Realization with Gaussian Kernel.}
    \label{alg:9}
    \begin{algorithmic}[1]
    \REQUIRE At stage $k$, the current canonical model $v^{\rho_k}\in\velmancan$; a stepsize $\eta_k\in(0,\tau]$; a reference factor $\mu=\rho^\base$ for fine-tuning (with the pretrained model $v^\base$) or $\mu=1$ for sampling; a reward function $r:\R^d\to\R$; an inverse temperature $\tau>0$; a matrix field $\widehat{\Sigma}^\rho:(0,1)\times\R^d\to\S_{++}^d$; and a batch size $N_\mathrm{batch}$.
    \ENSURE The updated velocity model $\hat{v}^{k+1}\approx v^{\rho_k}+\eta_k\Gamma^{\rho_k,\tilde{r}^{\rho_k}}$.
    \STATE Warm-start $\theta$ with the network parameters of $v^{\rho_k}$.
    \FOR{each gradient step}
        \STATE \COMMENT{The following operations are applied to a batch.}
            \IF{forward construction is used}
                \IF{ODE-based sampling is used}
                    \STATE Sample $X_1\sim\rho_k$ through ODE \eqref{eq:1} of $v^{\rho_k}$.
                \ELSIF{SDE-based sampling is used}
                    \STATE Sample $X_1\sim\rho_k$ through SDE \eqref{eq:5} of $v^{\rho_k}$.
                \ENDIF
                \STATE Sample $t\sim\U(0,1)$ and $X_0\sim\Normal(0,I)$, and set $X_t\gets\alpha_tX_1+\beta_tX_0$.
            \ELSIF{reverse construction is used}
                \STATE Sample $t\sim\U(0,1)$ and $X_t\sim \hat{p}_t$.
                \STATE Sample $X_1\sim p_{1|t}^{\rho_k}(\cdot|X_t)$ through the posterior-preserving SDE \eqref{eq:64} of $\rho_k$.
            \ENDIF
        \STATE Compute the target $\target$ by \eqref{eq:142}.
        \STATE Update $\theta$ based on the batch loss $\mathcal{L}\gets\frac{1}{N_{\mathrm{batch}}}\sum_{(t,X_t,\target)}\norm[2]{v_t^\theta(X_t)-\target}^2$.
    \ENDFOR
    \STATE Set $\hat{v}^{k+1}\gets v^\theta$.
\end{algorithmic}
\end{algorithm}

\begin{restatable}
    {proposition}{gaussianKernelConsistency}
\label{prop:15}
Assume that $\widehat\Sigma_t^\rho(x_t)$ is positive definite for every interior $(t,x_t)$ and the loss~\eqref{eq:142} is integrable. The unique population minimizer of the loss~\eqref{eq:142}, up to a.e. equality, is $\hat v^{\mathrm{Gau}\text{-}\mathrm{ker}}$ in~\eqref{eq:141}. Furthermore, the Gaussian-kernel approximation is critical-point consistent for $\pi$, i.e., $\widehat{\Gamma}^{\rho,\mathrm{Gau}\text{-}\mathrm{ker}}=0$ if and only if $\rho=\pi$.
\end{restatable}

The posterior covariance provides a natural choice for the surrogate \(\widehat\Sigma_t^\rho(x_t)\). For the Gaussian source and linear interpolant, we have
\[
    \Var_{X_1\sim p_{1|t}^\rho(\cdot|x_t)}[X_1]
    =
    \frac{\beta_t^2}{\alpha_t^2}I
    +
    \frac{\beta_t^4}{\alpha_t^2}
    \nabla_{x_t}^2\log p_t^\rho(x_t).
\]
This expression motivates the efficient choice
\[
    \widehat\Sigma_t^\rho(x_t)=\frac{\beta_t^2}{\alpha_t^2}I.
\]
In this case, the resulting update simplifies to
\begin{align*}
    &\where{\hat{v}_t^{\mathrm{Gau}\text{-}\mathrm{ker}}(x_t)}{\widehat\Sigma_t^\rho\equiv\frac{\beta_t^2}{\alpha_t^2}I}\\
    =&
    \E_{X_1\sim p_{1|t}^\rho(\cdot|x_t)}
    \left[
        v_{t|1}(x_t|x_1)+
    \frac{\eta\kappa_t}{\alpha_t}\left(
    \nabla r(X_1)+\frac{1}{\tau}\nabla\log\mu(X_1)+\frac{\alpha_t}{\tau\beta_t^2}\left(x_t-\alpha_tX_1\right)\right)
    \right]\\
    =&
    \E_{X_1\sim p_{1|t}^\rho(\cdot|x_t)}
    \left[
        \frac{\dot\beta_t}{\beta_t}x_t+\frac{\alpha_t\kappa_t}{\beta_t^2}X_1+
    \frac{\eta\kappa_t}{\alpha_t}\left(
    \nabla r(X_1)+\frac{1}{\tau}\nabla\log\mu(X_1)+\frac{\alpha_t}{\tau\beta_t^2}\left(x_t-\alpha_tX_1\right)\right)
    \right]\\
    =&
    \E_{X_1\sim p_{1|t}^\rho(\cdot|x_t)}
    \left[
        \left(\frac{\dot\beta_t}{\beta_t}+\frac{\eta\kappa_t}{\tau\beta_t^2}\right)x_t+\left(1-\frac{\eta}{\tau}\right)\frac{\alpha_t\kappa_t}{\beta_t^2}X_1+
    \frac{\eta\kappa_t}{\alpha_t}\left(
    \nabla r(X_1)+\frac{1}{\tau}\nabla\log\mu(X_1)\right)
    \right].
\end{align*}
In particular, when $\mu=1$,
\[
    \where{\hat{v}_t^{\mathrm{Gau}\text{-}\mathrm{ker}}(x_t)}{\mu=1,\,\widehat\Sigma_t^\rho\equiv\frac{\beta_t^2}{\alpha_t^2}I}
    =
    \E_{X_1\sim p_{1|t}^\rho(\cdot|x_t)}
    \left[
        \left(
            \frac{\dot\beta_t}{\beta_t}
            +
            \frac{\eta\kappa_t}{\tau\beta_t^2}
        \right)x_t
        +
        \left(1-\frac{\eta}{\tau}\right)
        \frac{\alpha_t\kappa_t}{\beta_t^2}X_1
        +
        \frac{\eta\kappa_t}{\alpha_t}\nabla r(X_1)
    \right].
\]
The corresponding loss is
\begin{align*}
    &\where{\calL_{\rho,\eta,\widehat{\Sigma}}^{\mathrm{Gau}\text{-}\mathrm{ker}}(\theta)}{\mu=1,\,\widehat\Sigma_t^\rho\equiv\frac{\beta_t^2}{\alpha_t^2}I}\\
    =&
    \E_{\substack{
    t\sim\U(0,1),\,(X_t,X_1)\sim \Xi_{t,1}^{\rho}}}
    \left[
        \norm[2]{
        v_t^\theta(X_t)
    -
        \left(\frac{\dot\beta_t}{\beta_t}+\frac{\eta\kappa_t}{\tau\beta_t^2}\right)X_t-\left(1-\frac{\eta}{\tau}\right)\frac{\alpha_t\kappa_t}{\beta_t^2}X_1-
        \frac{\eta\kappa_t}{\alpha_t}\nabla r(X_1)
        }^2
    \right].
\end{align*}
Using $\kappa_t=\beta_t^2\left(\frac{\dot\alpha_t}{\alpha_t}-\frac{\dot\beta_t}{\beta_t}\right)$, the loss above is sample-wise identical to the reference-adjoint approximation loss~\eqref{eq:139}. Therefore, when \( \mu=1 \), the reference-adjoint approximation is a special case of the Gaussian-kernel approximation.

\subsection{Regularization Trade-Offs}\label{subsec:24}

In the previous two subsections, we provide several approximate tangential updates that preserve critical-point consistency. We now take a different perspective by focusing directly on the regularization itself.
The term \(\log\frac{\rho}{\mu}\) is the key component that makes the iteration follow the Newton direction for the reverse-KL objective. It is also a major source of computational difficulty in exact Newton Matching, requiring nontrivial estimation procedures.

In this subsection, we directly analyze the influence of regularization on the update rules and consider approximating or removing it entirely.
Section~\ref{sub2sec:16} first decomposes the exact regularization into a marginal velocity correction and a posterior-KL gradient. The former is easy to compute exactly, whereas the latter admits various approximation options. Section~\ref{sub2sec:17} removes the regularization entirely and develops the corresponding algorithms. The resulting update performs direct raw-reward ascent.

\subsubsection{Approximate Regularization}\label{sub2sec:16}

Throughout the analysis below, assume that $\mu=\rho^\base\in\pdfspace$ is a normalized reference density. When the original reference is $\mu=1$, we apply the auxiliary-reference factorization~\eqref{eq:125}. The canonical tangent vector $\Gamma^{\rho,\tilde{r}^\rho}$ admits the following decomposition.

\begin{restatable}
    {proposition}{partialRegularizationDecomposition}
\label{prop:16}
Assume that $\mu=\rho^\base\in\pdfspace$. For every canonical anchor $v^\rho$ and stepsize $\eta>0$, we have
\begin{align*}
    \bar v_t(x_t)
    =&
    v_t^\rho(x_t)
    +
    \eta\Gamma_t^{\rho,\tilde r^\rho}(x_t)
    \\
    =&
    \left(1-\frac{\eta}{\tau}\right)v_t^\rho(x_t)
    +
    \frac{\eta}{\tau}v_t^\base(x_t)
    +
    \eta\Gamma_t^{\rho,r}(x_t)
    -
    \frac{\eta}{\tau}\kappa_t
    \nabla_{x_t}
    \KL{p_{1|t}^{\rho}(\cdot|x_t)}{p_{1|t}^{\base}(\cdot|x_t)}.
\end{align*}
\end{restatable}

The decomposition consists of three conceptually distinct contributions. The first, represented by the first two terms, interpolates between the current and reference canonical velocities. The second is the raw-reward tangent. The third is the posterior-KL gradient, which can be computationally challenging since it depends on how the current and reference posteriors vary with \(x_t\).

We may learn this gradient in a separate model or replace it with a prescribed approximation:
\[
    \nabla_{x_t}\KL{p_{1|t}^{\rho}(\cdot|x_t)}{p_{1|t}^{\base}(\cdot|x_t)}
    \approx
    \widehat K_t(x_t).
\]
This gives the approximate tangential update
\begin{subequations}
    \label{eq:143}
\begin{align}
    &\widehat{\Gamma}_t^{\rho,\mathrm{appr}\text{-}\mathrm{reg}}(x_t)
    :=
    -
    \frac{1}{\tau}
    \left(
        v_t^\rho(x_t)-v_t^\base(x_t)
    \right)
    +
    \Gamma_t^{\rho,r}(x_t)
    -
    \frac{1}{\tau}\kappa_t\widehat K_t(x_t),\\
    &\hat v_t^{\mathrm{appr}\text{-}\mathrm{reg}}(x_t)
    :=
    v_t^\rho(x_t)
    +
    \eta\widehat{\Gamma}_t^{\rho,\mathrm{appr}\text{-}\mathrm{reg}}(x_t)\notag\\
    &\phantom{\hat v_t^{\mathrm{appr}\text{-}\mathrm{reg}}(x_t)}\,\,=
    \left(1-\frac{\eta}{\tau}\right)v_t^\rho(x_t)
    +
    \frac{\eta}{\tau}v_t^\base(x_t)
    +
    \eta\Gamma_t^{\rho,r}(x_t)
    -
    \frac{\eta}{\tau}\kappa_t\widehat K_t(x_t).
\end{align}
\end{subequations}
The covariance-form loss is given by
\begin{align}
    &\calL_{\rho,\eta,B}^{\mathrm{appr}\text{-}\mathrm{reg}\text{-}\mathrm{cov}}(\theta)
    :=
    \E_{\substack{
    t\sim\U(0,1),\,(X_t,X_1)\sim \Xi_{t,1}^{\rho}}}
    \bigg[
        \Big\|
        v_t^\theta(X_t)
        -
        \left(1-\frac{\eta}{\tau}\right)v_{t|1}(X_t|X_1)
        \notag\\
    &\qquad
        -
        \frac{\eta}{\tau}v_t^\base(X_t)
        -
        \eta\left(r(X_1)-B_t(X_t)\right)
        \left(
            v_{t|1}(X_t|X_1)-v_t^\rho(X_t)
        \right)
        +
        \frac{\eta}{\tau}\kappa_t\widehat K_t(X_t)
        \Big\|_2^2
    \bigg],
    \label{eq:144}
\end{align}
where $B_t(X_t)$ is an arbitrary scalar baseline. The same approximate update can be realized in gradient form by using the raw-reward adjoint associated with the current anchor:
\begin{align}
    \calL_{\rho,\eta}^{\mathrm{appr}\text{-}\mathrm{reg}\text{-}\mathrm{grad}}(\theta)
    :=&
    \E_{\substack{
    t\sim\U(0,1),\,X_t\sim\hat p_t,\,
    \BY_{[t,1]}\sim\mathbb P_{[t,1]|t}^\rho(\cdot|X_t)}}
    \bigg[
        \Big\|
        v_t^\theta(X_t)
        -
        \left(1-\frac{\eta}{\tau}\right)v_{t|1}(X_t|Y_1)
        \notag\\
    &\qquad\qquad\qquad
        -
        \frac{\eta}{\tau}v_t^\base(X_t)
        -
        \eta\kappa_t
        \lambda_t^{b^\rho,r,0}(\BY_{[t,1]})
        +
        \frac{\eta}{\tau}\kappa_t\widehat K_t(X_t)
        \Big\|_2^2
    \bigg].
    \label{eq:145}
\end{align}
The pathwise adjoint is computed through
\begin{subequations}\label{eq:146}
    \begin{align}
        &\frac{\dd}{\dd s}
        \lambda_s^{b^\rho,r,0}(\BY_{[t,1]})
        =
        -\nabla b_s^\rho(Y_s)^\top
        \lambda_s^{b^\rho,r,0}(\BY_{[t,1]}),
        \qquad s\in[t,1],\\
        &\lambda_1^{b^\rho,r,0}(\BY_{[t,1]})
        =
        \nabla r(Y_1).
    \end{align}
\end{subequations}
For a generic approximation \(\widehat K_t\), there is no guarantee that the correction \(\widehat{\Gamma}^{\rho,\mathrm{appr}\text{-}\mathrm{reg}}\) vanishes exactly when \(\rho=\pi\). Consequently, the resulting construction need not be critical-point consistent. The tangential update with approximate regularization is summarized in Algorithm~\ref{alg:10}.

The most economical choice is to drop the posterior-KL gradient entirely:
\[
    \nabla_{x_t}
    \KL{p_{1|t}^{\rho}(\cdot|x_t)}{p_{1|t}^{\base}(\cdot|x_t)}
    \approx
    0.
\]
The corresponding update is
\[
    \hat v_t^{\mathrm{appr}\text{-}\mathrm{reg}}(x_t)
    =
    \left(1-\frac{\eta}{\tau}\right)v_t^\rho(x_t)
    +
    \frac{\eta}{\tau}v_t^\base(x_t)
    +
    \eta\Gamma_t^{\rho,r}(x_t).
\]
The covariance-form and gradient-form losses are obtained by substituting $\widehat K_t=0$ into~\eqref{eq:144} and~\eqref{eq:145}.%

\begin{algorithm}[!t]
    \caption{Tangential Update: Approximate Regularization.}
    \label{alg:10}
    \begin{algorithmic}[1]
    \REQUIRE At stage $k$, the current canonical model $v^{\rho_k}\in\velmancan$; a stepsize $\eta_k\in(0,\tau]$; a reference factor $\mu=\rho^\base$ for fine-tuning (with the pretrained model $v^\base$) or $\mu=1$ for sampling; a reward function $r:\R^d\to\R$; an inverse temperature $\tau>0$; a vector field $\widehat{K}:(0,1)\times\R^d\to\R^{d}$; and a batch size $N_\mathrm{batch}$.
    \ENSURE The updated velocity model $\hat{v}^{k+1}\approx v^{\rho_k}+\eta_k\Gamma^{\rho_k,\tilde{r}^{\rho_k}}$.
    \IF{$\mu=1$}
        \STATE Set $\rho^\base\gets\Normal(m,\sigma^2I)$ and its analytical velocity field $v^\base$ based on the auxiliary-reference factorization \eqref{eq:125}.
        \STATE Set $r\gets r-\frac{1}{\tau}\log\rho^\base$.%
    \ENDIF
    \STATE Warm-start $\theta$ with the network parameters of $v^{\rho_k}$.
    \FOR{each gradient step}
        \STATE \COMMENT{The following operations are applied to a batch.}
            \IF{forward construction is used}
                \IF{ODE-based sampling is used}
                    \STATE Sample $X_1\sim\rho_k$ through ODE \eqref{eq:1} of $v^{\rho_k}$.
                \ELSIF{SDE-based sampling is used}
                    \STATE Sample $X_1\sim\rho_k$ through SDE \eqref{eq:5} of $v^{\rho_k}$.
                \ENDIF
                \STATE Sample $t\sim\U(0,1)$ and $X_0\sim\Normal(0,I)$, and set $X_t\gets\alpha_tX_1+\beta_tX_0$.
                \IF{gradient form is used}
                    \STATE Sample $\BY_{[t,1]}\sim\mathbb{P}_{[t,1]}^\mathrm{uni}(\cdot|X_t,X_1)$ based on Theorem \ref{thm:14}.
                \ENDIF
            \ELSIF{reverse construction is used}
                \STATE Sample $t\sim\U(0,1)$ and $X_t\sim \hat{p}_t$.
                \STATE Sample $\BY_{[t,1]}\sim \mathbb{P}_{[t,1]|t}^{\rho_k}(\cdot|X_t)$ through the posterior-preserving SDE \eqref{eq:64} of $\rho_k$.
                \STATE Set $X_1\gets Y_1$.
            \ENDIF
        \IF{covariance form is used}
            \STATE Compute the target $\target$ by \eqref{eq:144}.
        \ELSIF{gradient form is used}
            \STATE Compute the pathwise adjoint $\lambda_t^{b^{\rho_k},r,0}(\BY_{[t,1]})$ by ODE \eqref{eq:146}.
            \STATE Compute the target $\target$ by \eqref{eq:145}.
        \ENDIF
        \STATE Update $\theta$ based on the batch loss $\mathcal{L}\gets\frac{1}{N_{\mathrm{batch}}}\sum_{(t,X_t,\target)}\norm[2]{v_t^\theta(X_t)-\target}^2$.
    \ENDFOR
    \STATE Set $\hat{v}^{k+1}\gets v^\theta$.
\end{algorithmic}
\end{algorithm}

A less aggressive alternative approximates both posteriors by Gaussian distributions with a shared isotropic covariance:
\[
    p_{1|t}^{\rho}(\cdot|x_t)
    \approx
    \Normal\left(M_t^\rho(x_t);\frac{\beta_t^2}{\alpha_t^2}I\right),
    \qquad
    p_{1|t}^{\base}(\cdot|x_t)
    \approx
    \Normal\left(M_t^\base(x_t);\frac{\beta_t^2}{\alpha_t^2}I\right),
\]
where the posterior means are exact:
\begin{align*}
    &M_t^\rho(x_t)
    :=
    \E_{X_1\sim p_{1|t}^{\rho}(\cdot|x_t)}[X_1]
    =
    \frac{\beta_t^2}{\alpha_t\kappa_t}
    \left(
        v_t^\rho(x_t)-\frac{\dot\beta_t}{\beta_t}x_t
    \right),\\
    &M_t^\base(x_t)
    :=
    \E_{X_1\sim p_{1|t}^{\base}(\cdot|x_t)}[X_1]
    =
    \frac{\beta_t^2}{\alpha_t\kappa_t}
    \left(
        v_t^\base(x_t)-\frac{\dot\beta_t}{\beta_t}x_t
    \right).
\end{align*}
For two Gaussians with the same covariance, we have
\begin{align*}
    \nabla_{x_t}\KL{p_{1|t}^{\rho}(\cdot|x_t)}{p_{1|t}^{\base}(\cdot|x_t)}
    \approx&\nabla_{x_t}\KL{\Normal\left(M_t^\rho(x_t);\frac{\beta_t^2}{\alpha_t^2}I\right)}{\Normal\left(M_t^\base(x_t);\frac{\beta_t^2}{\alpha_t^2}I\right)}\\
    =&\frac{\alpha_t^2}{2\beta_t^2}\nabla_{x_t}\norm[2]{M_t^\rho(x_t)-M_t^\base(x_t)}^2\\
    =&\frac{\beta_t^2}{2\kappa_t^2}\nabla_{x_t}\norm[2]{v_t^\rho(x_t)-v_t^\base(x_t)}^2.
\end{align*}
The Gaussian KL calculation is given in Lemma~\ref{lem:7} of Appendix~\ref{subsec:38}. The corresponding approximate tangential update is
\[
    \hat v_t^{\mathrm{appr}\text{-}\mathrm{reg}}(x_t)
    =
    \left(1-\frac{\eta}{\tau}\right)v_t^\rho(x_t)
    +
    \frac{\eta}{\tau}v_t^\base(x_t)
    +
    \eta\Gamma_t^{\rho,r}(x_t)
    -
    \frac{\eta\beta_t^2}{2\tau\kappa_t}
    \nabla_{x_t}
    \left\|v_t^\rho(x_t)-v_t^\base(x_t)\right\|_2^2.
\]
The corresponding losses are obtained by setting
\[
    \widehat K_t(x_t)
    =
    \frac{\beta_t^2}{2\kappa_t^2}
    \nabla_{x_t}
    \left\|v_t^\rho(x_t)-v_t^\base(x_t)\right\|_2^2
\]
in~\eqref{eq:144} and~\eqref{eq:145}.

\subsubsection{No Regularization}\label{sub2sec:17}

The most aggressive approximation is to remove the regularization entirely:
\[
    \tilde r^\rho
    \approx
    r.
\]
The resulting tangential update is
\begin{equation}
    \hat v_t^{\mathrm{no}\text{-}\mathrm{reg}}(x_t)
    =
    v_t^\rho(x_t)
    +
    \eta\Gamma_t^{\rho,r}(x_t).
    \label{eq:147}
\end{equation}
At this level of approximation, the update direction is governed by the raw reward $r$ and therefore admits a different optimization interpretation, namely, direct ascent of the expected reward.

The corresponding covariance-form loss is
\begin{align}
    \calL_{\rho,\eta,B}^{\mathrm{no}\text{-}\mathrm{reg}\text{-}\mathrm{cov}}(\theta)
    :=&
    \E_{\substack{
    t\sim\U(0,1),\,(X_t,X_1)\sim \Xi_{t,1}^{\rho}}}
    \Big[
        \big\|
        v_t^\theta(X_t)
        -
        v_{t|1}(X_t|X_1)
        \notag\\
    &\qquad
        -
        \eta
        \left(
        r(X_1)-B_t(X_t)
        \right)
        \left(
        v_{t|1}(X_t|X_1)-v_t^\rho(X_t)
        \right)
        \big\|_2^2
    \Big],
    \label{eq:148}
\end{align}
where $B_t(X_t)$ is an arbitrary scalar baseline. We can also realize the unregularized tangential update through the following gradient-form loss:
\begin{align}
    \calL_{\rho,\eta}^{\mathrm{no}\text{-}\mathrm{reg}\text{-}\mathrm{grad}}(\theta)
    :=&
    \E_{\substack{
        t\sim\U(0,1),\,\BY_{[t,1]}\sim \Xi_{[t,1]}^{\rho}}}
    \left[ \norm[2]{ v_t^\theta(Y_t)-v_{t|1}(Y_t|Y_1) -\eta\kappa_t\lambda_t^{b^\rho,r,0}(\BY_{[t,1]})}^2 \right],
    \label{eq:149}
\end{align}
where the pathwise adjoint
$\lambda_t^{b^\rho,r,0}(\BY_{[t,1]})$
can be computed through the ODE~\eqref{eq:146}.

\begin{algorithm}[!t]
    \caption{Tangential Update: No Regularization (Reward Ascent).}
    \label{alg:11}
    \begin{algorithmic}[1]
    \REQUIRE At stage $k$, the current canonical model $v^{\rho_k}\in\velmancan$; a stepsize $\eta_k>0$; a reward function $r:\R^d\to\R$; and a batch size $N_\mathrm{batch}$.
    \ENSURE The updated velocity model $\hat{v}^{k+1}=v^{\rho_k}+\eta_k\Gamma^{\rho_k,r}$.
    \STATE Warm-start $\theta$ with the network parameters of $v^{\rho_k}$.
    \FOR{each gradient step}
        \STATE \COMMENT{The following operations are applied to a batch.}
            \IF{forward construction is used}
                \IF{ODE-based sampling is used}
                    \STATE Sample $X_1\sim\rho_k$ through ODE \eqref{eq:1} of $v^{\rho_k}$.
                \ELSIF{SDE-based sampling is used}
                    \STATE Sample $X_1\sim\rho_k$ through SDE \eqref{eq:5} of $v^{\rho_k}$.
                \ENDIF
                \STATE Sample $t\sim\U(0,1)$ and $X_0\sim\Normal(0,I)$, and set $X_t\gets\alpha_tX_1+\beta_tX_0$.
                \IF{gradient form is used}
                    \STATE Sample $\BY_{[t,1]}\sim\mathbb{P}_{[t,1]}^\mathrm{uni}(\cdot|X_t,X_1)$ based on Theorem \ref{thm:14}.
                \ENDIF
            \ELSIF{reverse construction is used}
                \STATE Sample $t\sim\U(0,1)$ and $X_t\sim \hat{p}_t$.
                \STATE Sample $\BY_{[t,1]}\sim \mathbb{P}_{[t,1]|t}^{\rho_k}(\cdot|X_t)$ through the posterior-preserving SDE \eqref{eq:64} of $\rho_k$.
                \STATE Set $X_1\gets Y_1$.
            \ENDIF
        \IF{covariance form is used}
            \STATE Compute the target $\target$ by \eqref{eq:148}.
        \ELSIF{gradient form is used}
            \STATE Compute the pathwise adjoint $\lambda_t^{b^{\rho_k},r,0}(\BY_{[t,1]})$ by ODE \eqref{eq:146}.
            \STATE Compute the target $\target$ by \eqref{eq:149}.
        \ENDIF
        \STATE Update $\theta$ based on the batch loss $\mathcal{L}\gets\frac{1}{N_{\mathrm{batch}}}\sum_{(t,X_t,\target)}\norm[2]{v_t^\theta(X_t)-\target}^2$.
    \ENDFOR
    \STATE Set $\hat{v}^{k+1}\gets v^\theta$.
\end{algorithmic}
\end{algorithm}

The update~\eqref{eq:147} is exactly the canonical retraction of Section~\ref{subsec:7} with terminal observable $f=r$. The computation process is summarized in Algorithm~\ref{alg:11}. The value-ascent theorem therefore yields the following finite-stepsize reward-ascent guarantee.

\begin{restatable}
    {proposition}{noRegularizationRewardImprovementNotCPC}
\label{prop:17}
Consider the terminal density induced by~\eqref{eq:147}, i.e.,
\[
    q
    =
    \terminalmap
    \left(
    v^\rho+\eta\Gamma^{\rho,r}
    \right).
\]
We have the reward-ascent certificate
\[
    \E_{X\sim q}
    \left[
    r(X)
    \right]
    \geq
    \E_{X\sim\rho}
    \left[
    r(X)
    \right].
\]
Furthermore, whenever
\[
    \eta>0,
    \qquad
    r\not\equiv\const,
\]
the reward increases strictly:
\[
    \E_{X\sim q}
    \left[
    r(X)
    \right]
    >
    \E_{X\sim\rho}
    \left[
    r(X)
    \right].
\]
Consequently, if the reward is nonconstant, then the correction $\Gamma^{\rho,r}$ is not critical-point consistent for $\pi$.
\end{restatable}

Naturally, repeated unregularized updates with finite stepsizes continue to increase the expected reward and therefore favor the maximal-reward region rather than a finite-temperature target. The following result makes this concentration behavior precise.

\begin{restatable}
    {proposition}{noRegularizationConvergeToArgmax}
    \label{prop:18}
Let $\eta_k>0$ be the stepsize. Define the unregularized iteration
\[
    \rho_{k+1}
    =
    \terminalmap
    \left(
    v^{\rho_k}
    +
    \eta_k\Gamma^{\rho_k,r}
    \right).
\]
Assume that the stepsizes are non-summable:
\[
    \sum_{k=0}^{\infty}\eta_k
    =
    \infty,
\]
that the reward is continuous with a finite upper bound:
\[
    r_{\max}
    :=
    \sup_{x\in\R^d}r(x)
    <
    \infty,
\]
and that the dissipation vanishes:
\begin{equation}
    \lim_{k\to\infty}
    \frac{1}{\eta_k}
    \E_{X\sim\rho_0}
    \left[
    \Diss_{\rho_k,\eta_k,r}(X)
    \right]
    =
    0.
    \label{eq:150}
\end{equation}
Then, for any $\varepsilon>0$, we have
\[
    \lim_{k\to\infty}
    \rho_k
    \left(
    \left\{
    x\in\R^d:
    r(x)\geq r_{\max}-\varepsilon
    \right\}
    \right)
    =
    1.
\]
\end{restatable}

Proposition~\ref{prop:18} characterizes the limit behavior of the unregularized updates. 
The following proposition provides a complementary result. It quantifies the terminal-density error of a finite sequence of unregularized updates relative to $\pi$ in both directions of KL divergence.

\begin{restatable}
    {proposition}{noRegularizationConvergeToPi}
    \label{prop:19}
    Assume that $\mu\in\pdfspace$ and the reward $r$ is bounded:
    \[
    \norm[\infty]{r}:=\sup_{x\in\R^d}\abs{r(x)}<\infty.
    \]
    Consider a sequence of stepsizes $\left\{\eta_k\right\}_{k=0}^{K-1}\subset(0,\infty)$ and the unregularized iteration
    \[
        \rho_0
        =
        \mu,
        \qquad
        \rho_{k+1}
        =
        \terminalmap
        \left(
        v^{\rho_k}
        +
        \eta_k
        \Gamma^{\rho_k,r}
        \right).
    \]
    Assume that there exists a constant $C>0$ such that for every $k$,
    \begin{equation}\label{eq:151}
        \E_{X\sim\pi}\left[\Diss_{\rho_k,\eta_k,r}(X)\right]\leq C\eta_k^2.
    \end{equation}
    Denote
    \[
    \hat{\tau}:=\sum_{k=0}^{K-1}\eta_k,\qquad
    \eta_\mathrm{max}:=\max_{0\leq k<K}\eta_k.
    \]
    Then, we have
    \[
    \max\left\{\KL{\rho_K}{\pi},\KL{\pi}{\rho_K}\right\}\leq\frac{1}{2}\norm[\infty]{r}^2(\tau-\hat{\tau})^2+C\hat{\tau}\eta_\mathrm{max}.
    \]
\end{restatable}

The assumption in~\eqref{eq:151} is natural since
\[
\E_{X\sim\pi}\left[\Diss_{\rho_k,\eta_k,r}(X)\right]=\eta_k^2\,\E_{X\sim\pi}\left[\int_{0}^{1}\kappa_t\norm[2]{\nabla V_t^{\rho_k}[r]\left(\flow_{1\to t}^{\rho_k,\eta_k,r}(X)\right)}^2\dd t\right],
\]
where $\flow^{\rho_k,\eta_k,r}$ is the flow map generated by the velocity field $v^{\rho_k}+\eta_k\Gamma^{\rho_k,r}$.

\subsection{Fixed-Point Conditions via Zero Displacement}\label{subsec:25}

The preceding subsections explicitly realize a velocity field \(\hat v=v^\rho+\eta\widehat{\Gamma}^\rho\) and then canonicalize it. The resulting canonical field is then used for the next stage of the update.

Complementary to this stage-by-stage construction, we can take a fixed-point perspective on the update rule \(\hat v=v^\rho+\eta\widehat{\Gamma}^\rho\) itself. Rather than performing the iterative update, we focus on when the prescribed correction produces zero displacement and hence leaves \(v^\rho\) unchanged. This perspective applies to both exact and approximate tangential updates. As shown below, since $v^\rho+\eta\widehat{\Gamma}^\rho$ often admits a conditional-expectation representation, the corresponding fixed-point conditions can be converted into matching-style regression objectives using the \textit{stop-gradient trick}. However, this trick inevitably introduces approximations, even when applied to the exact tangential-update formula. We therefore place this construction in the present subsection under approximate Newton Matching.

This subsection develops the fixed-point viewpoint in three steps. Section~\ref{sub2sec:18} first characterizes fixed points of an ambient additive update through the vanishing of its displacement. Section~\ref{sub2sec:19} then formulates a generic stop-gradient objective and gives conditions under which its population-stationary points coincide with the fixed points of the additive update rule. Finally, Section~\ref{sub2sec:20} illustrates this general method with several examples drawn from exact and approximate Newton Matching. These examples make concrete how exact and approximate tangential updates give rise to self-anchored stop-gradient objectives and pave the way for demystifying existing methods in Section~\ref{sec:7}.

\subsubsection{Fixed Points of Tangential Updates}\label{sub2sec:18}

Both exact and approximate tangential updates considered in this paper have the ambient additive form
\[
    \hat v
    =
    v^\rho+\eta\widehat{\Gamma}^\rho,
\]
where the exact Newton Matching case corresponds to $\widehat{\Gamma}^\rho=\Gamma^{\rho,\tilde r^\rho}$.

Since $\eta>0$, a canonical velocity field $v^\rho$ is a fixed point of the ambient additive update if and only if applying the update results in zero displacement:
\begin{equation}
    \hat v=v^\rho
    \quad\Leftrightarrow\quad
    v^\rho+\eta\widehat{\Gamma}^\rho=v^\rho
    \quad\Leftrightarrow\quad
    \widehat{\Gamma}^\rho=0.
    \label{eq:152}
\end{equation}
If $\widehat{\Gamma}^\rho$ is critical-point consistent for $\pi$, then~\eqref{eq:152} holds if and only if $\rho=\pi$. By the injectivity of the canonical-velocity map, $v^\pi$ is therefore the unique canonical fixed point.

The fixed-point condition in~\eqref{eq:152} is imposed before canonicalization. It is stronger than the fixed-point condition of the complete canonicalized stage,
\[
    v^\rho
    =
    \projection
    \left(
        v^\rho+\eta\widehat{\Gamma}^\rho
    \right),
\]
since the canonical projection, like the terminal-density map, is many-to-one on the ambient velocity space. A nonzero ambient displacement can therefore return to the same canonical field after projection. Zero displacement rules out this ambiguity and is in the same spirit as the definition of critical-point consistency.

\subsubsection{Population-Stationary Points of Stop-Gradient Objectives}\label{sub2sec:19}

We next formalize how to enforce the fixed-point condition in~\eqref{eq:152} through a stop-gradient objective. Let \(Z\) collect all auxiliary randomness beyond \(t\) and \(X_t\). Given a terminal density \(\hat\rho\in\pdfspace\), let \(\Xi_{t,Z}^{\hat\rho}\) denote the joint distribution of \((X_t,Z)\), and let \(\Xi_{Z|t}^{\hat\rho}(\cdot| X_t)\) denote the conditional distribution of \(Z\) given \(X_t\). The particular sampling construction and its \(X_t\)-marginal are left implicit. For endpoint sampling associated with \(\hat\rho\), we write
\[
Z=X_1,\qquad
(X_t,X_1)\sim \Xi_{t,Z}^{\hat\rho}:=\Xi_{t,1}^{\hat\rho},\qquad
X_1| X_t\sim \Xi_{Z|t}^{\hat\rho}(\cdot| X_t):=p_{1|t}^{\hat\rho}(\cdot| X_t).
\]
For path sampling associated with \(\hat\rho\), we write
\[
Z=\BY_{[t,1]},\qquad
(X_t,\BY_{[t,1]})\sim \Xi_{t,Z}^{\hat\rho}:=\Xi_{[t,1]}^{\hat\rho},\qquad
\BY_{[t,1]}| X_t\sim \Xi_{Z|t}^{\hat\rho}(\cdot| X_t):=\mathbb{P}_{[t,1]|t}^{\hat\rho}(\cdot| X_t).
\]
Here, \(\Xi_{t,Z}^{\hat\rho}\) provides unified notation for the posterior requirements in~\eqref{eq:60} and~\eqref{eq:95}, since the fixed-point condition may involve either endpoint-based or path-based sampling. This notation specifies the required posterior-compatible law, which may be realized through either a forward or a reverse construction. The superscript \(\hat\rho\) emphasizes that the notation applies to a general terminal density, not necessarily the terminal density \(\rho\) associated with the current anchor \(v^\rho\).

For a canonical velocity model $v^\rho$, let
\[
    \target(t,x_t,Z;v^\rho)
\]
denote the corresponding sample-wise target, which may represent the targets presented in either exact or approximate Newton Matching. Consider the fixed-point condition whose right-hand side can be expressed as a conditional expectation:
\begin{equation}
    v_t^\rho(x_t)
    =
    \E_{Z\sim\Xi_{Z|t}^{\hat\rho}(\cdot|x_t)}
    \left[
        \target(t,x_t,Z;v^\rho)
    \right]
    \qquad\text{for a.e. }(t,x_t),
    \label{eq:153}
\end{equation}
where \(\hat{\rho}\) is typically the current terminal density \(\rho\), or, in some cases, the base terminal density \(\rho^\base\). A parameterized stop-gradient objective associated with~\eqref{eq:153} takes the form
\begin{align}
    \calL^{\mathrm{sg}}(\theta)
    :=&
    \E_{\substack{
        t\sim\U(0,1),\,
        (X_t,Z)\sim\Xi_{t,Z}^{\sg(\theta)}}}
        \left[
            \norm[2]{v_t^\theta(X_t)-
            \target
            \left(
                t,X_t,Z;\sg(v^\theta)
            \right)}^2
        \right].
    \label{eq:154}
\end{align}
Here \(\Xi_{t,Z}^{\sg(\theta)}\) denotes the selected sampling law \(\Xi_{t,Z}^{\hat\rho}\), with any dependence of its construction on \(\theta\) placed under stop-gradient. Thus, the superscript \(\sg(\theta)\) specifies the differentiation convention rather than the model used for sampling: if \(\hat\rho=\rho^\base\), the sampling law is simply the usual \(\Xi_{t,Z}^{\base}\); if \(\hat\rho=\rho\), we treat the current model \(v^\theta\) as canonical and substitute it for \(v^\rho\) in any sampling or target construction that requires a canonical model. All \(\theta\)-dependent quantities in the target are likewise held fixed during differentiation. Gradients flow only through the prediction \(v_t^\theta(X_t)\). Further discussion of this implicit canonicality assumption is provided in Remark~\ref{remark:7}.

To characterize precisely what a stop-gradient loss represents, it is important to formulate its stationarity condition directly in function space rather than through a particular finite-dimensional parameterization. The following definition introduces the directional derivative and the stationarity condition of the stop-gradient loss~\eqref{eq:154}.

\begin{definition}\label{def:3}
    Consider the stop-gradient loss in the function space
    \begin{align}
    \label{eq:155}
        \calL^{\mathrm{sg}\text{-}\mathrm{func}}(v)
        :=&
        \E_{\substack{t\sim\U(0,1),\,(X_t,Z)\sim\Xi_{t,Z}^{\sg(v)}}}\left[\norm[2]{v_t(X_t)-\target\left(t,X_t,Z;\sg(v)\right)}^2\right].
    \end{align}
    Given an admissible perturbation direction $h:(0,1)\times\R^d\to\R^d$, the \emph{directional derivative} of $\calL^{\mathrm{sg}\text{-}\mathrm{func}}$ at $v^\rho$ along $h$ is
    \begin{equation}
    \label{eq:156}
    \where{\frac{\dd}{\dd\varepsilon}\E_{\substack{t\sim\U(0,1),\,(X_t,Z)\sim\Xi_{t,Z}^{\hat\rho}}}\left[\norm[2]{v_t^\rho(X_t)+\varepsilon h_t(X_t)-\target\left(t,X_t,Z;v^\rho\right)}^2\right]}{\varepsilon=0}.
    \end{equation}
    A canonical velocity field $v^\rho$ is a \emph{population-stationary point} of the stop-gradient loss if, for every admissible perturbation direction $h$,
    \[
    \where{\frac{\dd}{\dd\varepsilon}\E_{\substack{t\sim\U(0,1),\,(X_t,Z)\sim\Xi_{t,Z}^{\hat\rho}}}\left[\norm[2]{v_t^\rho(X_t)+\varepsilon h_t(X_t)-\target\left(t,X_t,Z;v^\rho\right)}^2\right]}{\varepsilon=0}=0.
    \]
\end{definition}

\begin{proposition}\label{prop:20}
Assume that the $X_t$-marginal of $\Xi_{t,Z}^{\hat\rho}$ admits a positive density $\hat{p}_t$, and that the quantities in~\eqref{eq:156} are square-integrable for the perturbation directions considered below. The directional derivative in~\eqref{eq:156} is equal to
\begin{equation}\label{eq:157}
2\E_{\substack{t\sim\U(0,1),\,X_t\sim\hat{p}_t}}\left[h_t(X_t)^\top\left(v_t^\rho(X_t)-\E_{Z\sim\Xi_{Z|t}^{\hat\rho}(\cdot|X_t)}\left[\target\left(t,X_t,Z;v^\rho\right)\right]\right)\right].
\end{equation}

Assume further that there exists a function $\omega:(0,1)\times\R^d\to(0,\infty)$ such that
\[
\begin{aligned}
    h^\omega:(t,x_t)
    \mapsto
    \omega_t(x_t)\left(v_t^\rho(x_t)-\E_{Z\sim\Xi_{Z|t}^{\hat\rho}(\cdot|x_t)}\left[\target(t,x_t,Z;v^\rho)\right]\right)
\end{aligned}
\]
is an admissible perturbation direction. Then, $v^\rho$ is a population-stationary point of the stop-gradient loss~\eqref{eq:155} if and only if it satisfies the fixed-point condition~\eqref{eq:153}, up to a.e. equality.
\end{proposition}

\begin{proof}
    The directional derivative in \eqref{eq:156} is equal to \eqref{eq:157} since
    \begin{align*}
    &\where{\frac{\dd}{\dd\varepsilon}\E_{\substack{t\sim\U(0,1),\,(X_t,Z)\sim\Xi_{t,Z}^{\hat\rho}}}\left[\norm[2]{v_t^\rho(X_t)+\varepsilon h_t(X_t)-\target\left(t,X_t,Z;v^\rho\right)}^2\right]}{\varepsilon=0}\\
    =&\where{\frac{\dd}{\dd\varepsilon}\E_{\substack{t\sim\U(0,1),\,X_t\sim\hat{p}_t}}\left[\norm[2]{v_t^\rho(X_t)+\varepsilon h_t(X_t)-\E_{Z\sim\Xi_{Z|t}^{\hat\rho}(\cdot|X_t)}\left[\target\left(t,X_t,Z;v^\rho\right)\right]}^2\right]}{\varepsilon=0}\\
    =&2\E_{\substack{t\sim\U(0,1),\,X_t\sim\hat{p}_t}}\left[h_t(X_t)^\top\left(v_t^\rho(X_t)-\E_{Z\sim\Xi_{Z|t}^{\hat\rho}(\cdot|X_t)}\left[\target\left(t,X_t,Z;v^\rho\right)\right]\right)\right].
    \end{align*}

    Consequently, the directional derivative of $\calL^{\mathrm{sg}\text{-}\mathrm{func}}$ at $v^\rho$ along $h^\omega$ is
    \begin{align*}
    &\where{\frac{\dd}{\dd\varepsilon}\E_{\substack{t\sim\U(0,1),\,(X_t,Z)\sim\Xi_{t,Z}^{\hat\rho}}}\left[\norm[2]{v_t^\rho(X_t)+\varepsilon h_t^\omega(X_t)-\target\left(t,X_t,Z;v^\rho\right)}^2\right]}{\varepsilon=0}\\
    =&2\E_{\substack{t\sim\U(0,1),\,X_t\sim\hat{p}_t}}\left[h_t^\omega(X_t)^\top\left(v_t^\rho(X_t)-\E_{Z\sim\Xi_{Z|t}^{\hat\rho}(\cdot|X_t)}\left[\target\left(t,X_t,Z;v^\rho\right)\right]\right)\right]\\
    =&2\E_{\substack{t\sim\U(0,1),\,X_t\sim\hat{p}_t}}\left[\omega_t(X_t)\norm[2]{v_t^\rho(X_t)-\E_{Z\sim\Xi_{Z|t}^{\hat\rho}(\cdot|X_t)}\left[\target\left(t,X_t,Z;v^\rho\right)\right]}^2\right].
    \end{align*}
    If $v^\rho$ is a population-stationary point up to a.e. equality, Definition~\ref{def:3} implies that the above directional derivative is $0$ since $h^\omega$ is an admissible perturbation direction. Conversely, assume that the directional derivative along $h^\omega$ is $0$. The positivity of $\hat{p}_t$ and $\omega_t(x_t)$ implies that $v^\rho$ satisfies \eqref{eq:153} a.e. Substituting \eqref{eq:153} into \eqref{eq:157} proves that the directional derivative along every admissible perturbation direction $h$ is $0$.
\end{proof}

\begin{proposition}\label{prop:21}
Let $\eta>0$, and let $\widehat{\Gamma}^\rho$ be an exact or approximate correction at the canonical field $v^\rho$. Suppose that the sample-wise target in~\eqref{eq:155} has conditional mean
\begin{equation}
    \E_{Z\sim\Xi_{Z|t}^{\hat\rho}(\cdot|x_t)}
    \left[
        \target(t,x_t,Z;v^\rho)
    \right]
    =
    v_t^\rho(x_t)
    +
    \eta\widehat{\Gamma}_t^\rho(x_t)
    \qquad
    \text{for a.e. }(t,x_t).
    \label{eq:158}
\end{equation}
Then, among canonical velocity fields,
\[
    v^\rho
    \text{ is a population-stationary point}
    \quad\Leftrightarrow\quad
    \widehat{\Gamma}^\rho=0.
\]
Consequently, if the method is critical-point consistent for $\pi$, then
\[
    v^\rho
    \text{ is a population-stationary point}
    \quad\Leftrightarrow\quad
    \rho=\pi.
\]
\end{proposition}

\begin{proof}
By~\eqref{eq:158} and Proposition~\ref{prop:20}, $v^\rho$ is population-stationary if and only if
\[
    v^\rho
    =
    v^\rho+\eta\widehat{\Gamma}^\rho.
\]
Since $\eta>0$, the displayed equality is equivalent to $\widehat{\Gamma}^\rho=0$, which is the zero-displacement condition~\eqref{eq:152}. Critical-point consistency gives the second equivalence.
\end{proof}

Proposition~\ref{prop:20} characterizes the population-stationary points of the stop-gradient loss by showing that they coincide with the fixed points of the corresponding additive update. It does not, however, by itself establish uniqueness or convergence. If the fixed-point condition~\eqref{eq:153} has a unique solution among canonical velocity fields, then this solution is also the unique population-stationary point of the stop-gradient loss within the same class. The proposition does not guarantee that optimizing a stop-gradient loss will converge to a population-stationary point.

\begin{remark}
Stop-gradient objectives arise from the fixed-point perspective of Newton Matching and are also widely used in related methods~\citep{bergmeister2026reinforce,domingo2025adjoint,havens2026flow,havens2025adjoint}. To the best of our knowledge, however, prior work has typically treated stop-gradient as an implementation trick and has rarely characterized either the optimization semantics of the resulting objective or its population-stationary points. Propositions~\ref{prop:20} and~\ref{prop:21} provide such a characterization.
\end{remark}

\begin{remark}\label{remark:7}
The fixed-point viewpoint and its corresponding stop-gradient implementation blur the stage-by-stage nature of the exact Newton Matching implementations in Section~\ref{sec:5} and the earlier approximate constructions in Sections~\ref{subsec:22}--\ref{subsec:24}. The stop-gradient objective has a self-anchored flavor: the current trainable model is reused, under stop-gradient, to obtain any model-dependent quantities, including the samples. These constructions are held fixed only during each gradient evaluation and are refreshed as the parameters change, or, in practice, when the exponential moving average (EMA) model used for these constructions is updated. Consequently, there is no explicit notion of a stage and no clear criterion for when canonicalization should be applied. Furthermore, to justify the sample-wise regression target, the right-hand side of the fixed-point condition must admit a conditional-expectation representation as~\eqref{eq:158}, which generally requires the model to be canonical. Thus, in practice, the stop-gradient implementation is heuristic: the current model is treated as canonical when constructing the sample-wise target.
\end{remark}

\begin{remark}
The notions of a \emph{fixed point} and a \emph{population-stationary point} should not be conflated. A fixed point is defined relative to an update rule: applying the rule leaves the point unchanged. A population-stationary point is defined relative to a stop-gradient objective: the directional derivative vanishes for every admissible function-space perturbation. The two notions coincide only when the population-stationarity condition of the stop-gradient objective is shown to be equivalent to the fixed-point condition of the update rule.
\end{remark}

\subsubsection{Examples of Stop-Gradient Objectives for Newton Matching}\label{sub2sec:20}

The previous analysis is agnostic to both the choice of the auxiliary variable \(Z\) and whether the additive update rule arises from exact or approximate Newton Matching. We now instantiate this general construction with three representative gradient-form objectives. These stop-gradient objectives also serve as templates in Section~\ref{sec:7}, where several existing methods are recovered as special cases of exact or approximate Newton Matching.

We begin with the Bolza realization of the gradient-form update in
\eqref{eq:109}--\eqref{eq:110}.
Replacing the canonical anchor \(v^\rho\) by the current model \(v^\theta\) and placing its occurrences in the sampling and adjoint constructions under stop-gradient, we have
\begin{align}
    \calL_{\eta}^{\mathrm{grad,Bolza},\mathrm{sg}}(\theta)
    :=&
    \E_{\substack{
        t\sim\U(0,1),\,\BY_{[t,1]}\sim\Xi_{[t,1]}^{\sg(\theta)}}}
    \bigg[
        \Big\|
            v_t^\theta(Y_t)
            -
            \left(1-\frac{\eta}{\tau}\right)v_{t|1}(Y_t|Y_1)\notag\\
            &\qquad\qquad\qquad\qquad
            -\frac{\eta}{\tau}v_t^{\mu}(Y_t)-\eta\kappa_t\sg\left(\lambda_t^{\mathrm{Bolza},\theta}(\BY_{[t,1]})\right)
        \Big\|_2^2
    \bigg].
    \label{eq:159}
\end{align}
Here, \(\Xi_{[t,1]}^{\sg(\theta)}\) denotes the same posterior requirement as \(\Xi_{[t,1]}^\rho\) in~\eqref{eq:95}, but with the current model \(v^\theta\) treated as canonical in place of \(v^\rho\) and all resulting dependence on \(\theta\) stopped during differentiation. The pathwise adjoint in the target is computed along each sampled path through
\begin{subequations}
\label{eq:160}
    \begin{align}
        &\frac{\dd}{\dd s}
        \lambda_s^{\mathrm{Bolza},\theta}(\BY_{[t,1]})
        =
        -\left(2\nabla v_s^\theta(Y_s)-\frac{\dot\alpha_s}{\alpha_s}I\right)^\top
        \lambda_s^{\mathrm{Bolza},\theta}(\BY_{[t,1]})\notag\\
        &\qquad\qquad\qquad\qquad\qquad\qquad\qquad\quad
        +\nabla_{Y_s}\left(\frac{\norm[2]{v_s^\theta(Y_s)-v_s^{\mu}(Y_s)}^2}{\tau\kappa_s}\right),
        \qquad s\in[t,1],\\
        &\lambda_1^{\mathrm{Bolza},\theta}(\BY_{[t,1]})
        =
        \nabla r(Y_1).
    \end{align}
\end{subequations}

The same fixed-point perspective and stop-gradient template apply to the reference-adjoint approximation developed in Section~\ref{sub2sec:14}. In this construction, the posterior-compatible path is still generated from the current anchor \(v^\rho\), while the adjoint ODE is evaluated with the reference velocity \(v^\mu\). Applying the stop-gradient trick to~\eqref{eq:138} gives
\begin{align}
    \calL_{\eta}^{\mathrm{ref}\text{-}\mathrm{adj},\mathrm{sg}}(\theta)
    :=
    \E_{\substack{
    t\sim\U(0,1),\,
    \BY_{[t,1]}\sim\Xi_{[t,1]}^{\sg(\theta)}}}
    \bigg[
        \Big\|&
        v_t^\theta(Y_t)
        -
        \left(1-\frac{\eta}{\tau}\right)v_{t|1}(Y_t|Y_1)
        \notag\\
        &
        -
        \frac{\eta}{\tau}v_t^\mu(Y_t)
        -
        \eta\kappa_t
        \lambda_t^{b^\mu,r,0}(\BY_{[t,1]})
        \Big\|_2^2
    \bigg].
    \label{eq:161}
\end{align}
The notation \(\Xi_{[t,1]}^{\sg(\theta)}\) has the same interpretation as in the stop-gradient loss for the Bolza realization.

The Gaussian-kernel approximation \eqref{eq:142} developed in Section~\ref{sub2sec:15} removes the pathwise adjoint altogether and replaces the exact posterior Stein kernel with a positive-definite matrix field. This construction requires only an endpoint pair \((X_t,X_1)\). Applying the stop-gradient trick to~\eqref{eq:142} gives
\begin{align}
    &\calL_{\eta,\widehat{\Sigma}}^{\mathrm{Gau}\text{-}\mathrm{ker},\mathrm{sg}}(\theta)
    :=
    \E_{\substack{
    t\sim\U(0,1),\,(X_t,X_1)\sim \Xi_{t,1}^{\sg(\theta)}}}
    \Bigg[
        \bigg\|
        v_t^\theta(X_t)
        -
        v_{t|1}(X_t|X_1)
        \notag\\
    &\qquad\qquad
        -
        \eta\frac{\alpha_t\kappa_t}{\beta_t^2}
        \widehat\Sigma_t(X_t)
        \left(
            \nabla r(X_1)
            +
            \frac{1}{\tau}\nabla\log\mu(X_1)
            +
            \frac{\alpha_t}{\tau\beta_t^2}
            \left(
                X_t-\alpha_tX_1
            \right)
        \right)
        \bigg\|_2^2
    \Bigg].
    \label{eq:162}
\end{align}
Here, \(\Xi_{t,1}^{\sg(\theta)}\) represents the same posterior requirement as \(\Xi_{t,1}^\rho\) in~\eqref{eq:60}, but with the current model \(v^\theta\) treated as canonical in place of \(v^\rho\), and with all resulting dependence on \(\theta\) stopped during differentiation.

Since the Bolza realization from exact Newton Matching and the reference-adjoint and Gaussian-kernel approximations from approximate Newton Matching are all critical-point consistent, Proposition~\ref{prop:21} implies that each corresponding stop-gradient objective has \(v^\pi\) as its unique canonical population-stationary point.

\begin{sidewaystable}[p]
    \centering
    \caption{A Newton Matching taxonomy for fine-tuning and sampling, with representative existing methods recovered as special cases.}
    \label{tab:2}
    \begingroup
\renewcommand{\arraystretch}{1.25}

\newlength{\nmmethodtablewidth}
\newlength{\nmmethodcolone}
\newlength{\nmmethodcoltwo}
\newlength{\nmmethodcolthree}
\newlength{\nmmethodcolfour}
\setlength{\nmmethodtablewidth}{\linewidth}
\addtolength{\nmmethodtablewidth}{-8\tabcolsep}
\addtolength{\nmmethodtablewidth}{-5\arrayrulewidth}
\setlength{\nmmethodcoltwo}{0.2\nmmethodtablewidth}
\setlength{\nmmethodcolthree}{0.35\nmmethodtablewidth}
\setlength{\nmmethodcolfour}{0.35\nmmethodtablewidth}
\setlength{\nmmethodcolone}{\nmmethodtablewidth}
\addtolength{\nmmethodcolone}{-\nmmethodcoltwo}
\addtolength{\nmmethodcolone}{-\nmmethodcolthree}
\addtolength{\nmmethodcolone}{-\nmmethodcolfour}

\newlength{\nmmethodcellpadding}
\setlength{\nmmethodcellpadding}{0.3em}
\newcommand{\nmmethodcell}[1]{%
    \kern\nmmethodcellpadding
    {\centering
    #1\par}%
    \kern\nmmethodcellpadding
}
\newenvironment{nmmethoditems}{%
    \begin{itemize}[
        leftmargin=1.4em,
        labelwidth=0.6em,
        labelsep=0.4em,
        itemsep=0pt,
        parsep=0pt,
        topsep=0pt
    ]
    \raggedright
}{%
    \end{itemize}
}
\setlist[itemize,2]{
    label=\textopenbullet,
    leftmargin=1.25em,
    labelwidth=0.5em,
    labelsep=0.35em,
    itemsep=0pt,
    parsep=0pt,
    topsep=0.15em,
    partopsep=0pt
}
\newcommand{\nmmethodlistcell}[1]{%
    \nmmethodcell{%
        \kern0.2em
        \begin{nmmethoditems}
            #1
        \end{nmmethoditems}%
        \kern-0.2em
    }%
}
\newcommand{\nmmethodjustify}{%
    \leftskip=0pt\relax
    \rightskip=0pt\relax
    \parfillskip=0pt plus 1fil\relax
}
\newenvironment{nmmethoditemsjust}{%
    \begin{itemize}[
        leftmargin=1.4em,
        labelwidth=0.6em,
        labelsep=0.4em,
        itemsep=0pt,
        parsep=0pt,
        topsep=0pt
    ]
    \nmmethodjustify
}{%
    \end{itemize}
}
\newcommand{\nmmethodlistcelljust}[1]{%
    \nmmethodcell{%
        \kern0.2em
        \begin{nmmethoditemsjust}
            #1
        \end{nmmethoditemsjust}%
        \kern-0.2em
    }%
}
\newcolumntype{N}[1]{%
    >{\begin{minipage}[c]{#1}\ignorespaces}c<{\end{minipage}}%
}

\footnotesize
\fontsize{9pt}{10.5pt}\selectfont
\setlength{\abovedisplayskip}{0.4em}
\setlength{\belowdisplayskip}{0.4em}
\setlength{\abovedisplayshortskip}{0.4em}
\setlength{\belowdisplayshortskip}{0.4em}
\begin{tabular}{
    |N{\nmmethodcolone}
    |N{\nmmethodcoltwo}
    |N{\nmmethodcolthree}
    |N{\nmmethodcolfour}|}
    \hline
    \nmmethodcell{}
    &
    \nmmethodcell{}
    &
    \nmmethodcell{Covariance Form}
    &
    \nmmethodcell{Gradient Form}
    \tabularnewline
    \hline
    \nmmethodcell{}
    &
    \nmmethodcell{Exact Realizations}
    &
    \nmmethodlistcell{%
        \item Exact covariance form
        \eqref{eq:67}--%
        \eqref{eq:68}.
    }
    &
    \nmmethodlistcell{%
        \item Exact Mayer realization
        \eqref{eq:98}--%
        \eqref{eq:99}.
        \item Exact Bolza realization
        \eqref{eq:109}--%
        \eqref{eq:110}.
        \begin{itemize}
            \item Basic Adjoint Matching
            \citep{domingo2025adjoint}, recovered by specialization \eqref{eq:168}.
        \end{itemize}
    }
    \tabularnewline
    \cline{2-4}
    \nmmethodcell{%
        \smash{\raisebox{-2.3em}{\shortstack{Fine-Tuning\\$\mu=\rho^\base$}}}%
    }
    &
    \nmmethodcell{Approximations with Critical-Point Consistency}
    &
    \nmmethodlistcell{%
        \item Direct linearization
        \eqref{eq:129}.
        \begin{itemize}
            \item Implicit Tilt Matching
            \citep{potaptchik2026tilt}, recovered by specialization \eqref{eq:164}.
        \end{itemize}
        \item Split linearization
        \eqref{eq:133}.
    }
    &
    \nmmethodlistcell{%
        \item Reference-adjoint approximation
        \eqref{eq:138}.
        \begin{itemize}
            \item Lean Adjoint Matching
            \citep{domingo2025adjoint}, recovered by specialization \eqref{eq:169}.
        \end{itemize}
        \item Gaussian-kernel approximation
        \eqref{eq:142}.
    }
    \tabularnewline
    \cline{2-4}
    \nmmethodcell{}
    &
    \nmmethodcell{Approximations without Critical-Point Consistency}
    &
    \nmmethodlistcell{%
        \item Approximate regularization
        \eqref{eq:144}.
        \begin{itemize}
            \item Reinforce Adjoint Matching \citep{bergmeister2026reinforce}, recovered by specialization \eqref{eq:167}.
        \end{itemize}
        \item No regularization
        \eqref{eq:148}.
        \begin{itemize}
            \item Explicit Tilt Matching
            \citep{potaptchik2026tilt}, recovered by specialization \eqref{eq:163}.
            \item DiffusionNFT
            \citep{zheng2026diffusionnft}, recovered by specialization \eqref{eq:165}.
        \end{itemize}
    }
    &
    \nmmethodlistcell{%
        \item Approximate regularization
        \eqref{eq:145}.
        \item No regularization
        \eqref{eq:149}.
    }
    \tabularnewline
    \hline
    \nmmethodcell{}
    &
    \nmmethodcell{Exact Realizations}
    &
    \nmmethodlistcell{%
        \item Exact covariance form
        \eqref{eq:67}--%
        \eqref{eq:68}.
    }
    &
    \nmmethodlistcell{%
        \item Exact Mayer realization
        \eqref{eq:98}--%
        \eqref{eq:99}.
        \item Exact Bolza realization
        \eqref{eq:109}--%
        \eqref{eq:110}.
    }
    \tabularnewline
    \cline{2-4}
    \nmmethodcell{%
        \smash{\raisebox{-1.4em}{\shortstack{Sampling\\$\mu=1$}}}%
    }
    &
    \nmmethodcell{Approximations with Critical-Point Consistency}
    &
    \nmmethodlistcell{%
        \item Direct linearization \eqref{eq:129} via auxiliary-reference factorization \eqref{eq:125}.
        \begin{itemize}
            \item Implicit Tilt Matching
            \citep{potaptchik2026tilt}, recovered by specialization \eqref{eq:164}.
        \end{itemize}
        \item Split linearization \eqref{eq:133} via auxiliary-reference factorization \eqref{eq:125}.
    }
    &
    \nmmethodlistcell{%
        \item Gaussian-kernel
        approximation \eqref{eq:142}.
        \begin{itemize}
            \item Flow Sampling
            \citep{havens2026flow}, recovered by specialization \eqref{eq:170}.
            \item Adjoint Sampling
            \citep{havens2025adjoint}, recovered by specialization \eqref{eq:171}.
        \end{itemize}
        \item Reference-adjoint
        approximation \eqref{eq:139} as a specialization of the Gaussian-kernel approximation \eqref{eq:142}.
    }
    \tabularnewline
    \cline{2-4}
    \nmmethodcell{}
    &
    \nmmethodcell{Approximations without Critical-Point Consistency}
    &
    \nmmethodlistcell{%
        \item Approximate regularization
        \eqref{eq:144} via auxiliary-reference factorization \eqref{eq:125}.
        \item No regularization
        \eqref{eq:148}.
    }
    &
    \nmmethodlistcell{%
        \item Approximate regularization
        \eqref{eq:145} via auxiliary-reference factorization \eqref{eq:125}.
        \item No regularization
        \eqref{eq:149}.
    }
    \tabularnewline
    \hline
\end{tabular}
\endgroup

\end{sidewaystable}

\section{Demystifying Existing Works}\label{sec:7}

Sections~\ref{sec:5} and~\ref{sec:6} develop exact Newton Matching and a family of approximate schemes, offering different trade-offs between update exactness and computational cost. Building on these developments, this section demystifies representative recent methods for fine-tuning and sampling. These works approach either of the two tasks from seemingly distinct perspectives, such as stochastic optimal control (SOC), contrastive learning, fixed-point iteration, and velocity evolution under exponential tilting. By translating their notation, sampler, and parameterization into the language of Newton Matching, we show that they can be recovered as special cases of exact or approximate Newton Matching algorithms. This unification is summarized in Table~\ref{tab:2}, with detailed derivations deferred to Appendix~\ref{app:9}.

An important subtlety is canonicality. As developed throughout the paper, both the matching-style regression objectives and the scalable methods used to compute their targets rely on the conditional-expectation structure of a canonical model and the associated posterior calculus. In many of the works discussed below, this requirement is left implicit: their derivations treat the current model as canonical, even though optimizing the proposed losses does not automatically preserve canonicality.

Many of the existing works make use of stop-gradient losses. As shown in Section~\ref{subsec:25}, when an additive update of the form \(\hat v=v^\rho+\eta\widehat{\Gamma}^\rho\) admits a conditional-expectation representation, its fixed-point condition can be encoded by a stop-gradient loss. Two points are worth emphasizing. First, one must identify the particular additive update, and hence the particular fixed-point condition, encoded by the loss. Second, even for the same update rule, the stop-gradient formulation is only one possible implementation and may not be the best choice: it encodes an additive update rule through its fixed-point condition, while implicitly treating the current model \(v^\theta\) as canonical, as discussed in Remark~\ref{remark:7}. By contrast, Newton Matching also provides the option of implementing the update \(\hat v=v^\rho+\eta\widehat{\Gamma}^\rho\) stage by stage, with a natural opportunity to perform canonicalization after each stage.

\subsection{Tilt Matching}\label{subsec:26}

Explicit Tilt Matching (ETM) and Implicit Tilt Matching (ITM)~\citep{potaptchik2026tilt} consider fine-tuning from a normalized reference density \(\mu=\rho^\base\in\pdfspace\) with associated canonical velocity \(v^\base=\mathcal C(\rho^\base)\in\velmancan\). They also apply to the sampling case \(\mu=1\) through the auxiliary-reference factorization~\eqref{eq:125}.
Both methods are organized along the exponential-tilt continuation path
\[
    \pi_s(x)
    :=
    \frac{\rho^\base(x)e^{s r(x)}}
    {\int_{\mathbb R^d}\rho^\base(z)e^{s r(z)}\,\mathrm dz}.
\]
For a continuation increment \(s\to s+h\), the consecutive targets satisfy
\[
    \pi_{s+h}(x)\propto \pi_s(x)e^{h r(x)}.
\]
Assuming that the increment \(h\) is sufficiently small, each continuation stage can therefore be viewed as a local fine-tuning problem with reference density \(\pi_s\), canonical anchor \(v^{\pi_s}\), and inverse-temperature \(h\). For notational simplicity, throughout the remainder of this subsection, we relabel the current continuation density \(\pi_s\), its canonical velocity \(v^{\pi_s}\), and the increment \(h\) as \(\rho^\base\), \(v^\base\), and \(\tau\), respectively. It then suffices to analyze a generic stage with reference density \(\mu=\rho^\base\in\pdfspace\) and target density \(\pi_{\rho^\base,\tau,r}\in\pdfspace\). Detailed derivations, together with an analysis of the discretization error defined in \cite[Section 3.1]{potaptchik2026tilt}, are deferred to Appendix~\ref{subsec:58}.

\subsubsection{Explicit Tilt Matching}

At a canonical base anchor \(v^\base\), the ETM loss can be written as
\begin{align*}
    \mathcal{L}^{\mathrm{ETM}}(\theta)
    =&\E_{\substack{t\sim\U(0,1),\,X_1\sim\rho^\base,\,X_0\sim p_0,\,X_t=\alpha_tX_1+\beta_tX_0}}
    \Bigg[\bigg\|v_t^\theta(X_t)-v_{t|1}(X_t|X_1)\\
    &\qquad\qquad\qquad
    -\tau\left(r(X_1)-\frac{1}{\tau}\right)\left(v_{t|1}(X_t|X_1)-v_t^\base(X_t)\right)\bigg\|_2^2\Bigg].
\end{align*}
Its unique population minimizer is
\begin{equation*}
    \bar v^{\mathrm{ETM}}
    =
    v^\base
    +
    \tau\Gamma^{\rho^\base,r}.
\end{equation*}

The ETM loss $\mathcal{L}^{\mathrm{ETM}}$ can be recovered directly from Newton Matching by making the following choices. We take the no-regularization approximation from Section~\ref{sub2sec:17}, namely,
\[
    \tilde r^\rho\approx r,
\]
together with the covariance form via forward construction. Then, we set
\[
    \rho=\rho^\base,
    \qquad
    \eta=\tau,
    \qquad
    B_t\equiv\frac{1}{\tau}.
\]
Comparing the two losses yields the exact sample-wise correspondence
\begin{equation}
\mathcal{L}^{\mathrm{ETM}}(\theta)=\where{\calL_{\rho,\eta,B}^{\mathrm{no}\text{-}\mathrm{reg}\text{-}\mathrm{cov}}(\theta)}{
    \rho=\rho^\base,\,\eta=\tau,\,B_t\equiv\frac{1}{\tau}
}
\qquad\text{under the forward construction.}
\label{eq:163}
\end{equation}
Therefore, ETM is the covariance form of the full-step unregularized update at the base anchor with the forward construction and a constant baseline. It is not critical-point consistent whenever $r$ is nonconstant.

\subsubsection{Implicit Tilt Matching}

ITM uses a stop-gradient objective. With a positive control-variate coefficient $c(x_t)>0$, its loss is
\begin{align*}
    \mathcal{L}_c^{\mathrm{ITM}}(\theta)
    :=&\E_{\substack{t\sim\U(0,1),\,X_1\sim\rho^\base,\,X_0\sim p_0,\,X_t=\alpha_tX_1+\beta_tX_0}}
    \bigg[\Big\|c(X_t)\left(v_t^\theta(X_t)-v_t^\base(X_t)\right)\notag\\
    &\qquad\quad
    +\left(e^{\tau r(X_1)}-c(X_t)\right)\left(\sg\left(v_t^\theta(X_t)\right)-v_{t|1}(X_t|X_1)\right)\Big\|_2^2\bigg].
\end{align*}
The unique population-stationary point is the canonical velocity field
\begin{equation*}
    v^{\mathrm{ITM}}
    =
    v^{\pi_{\rho^\base,\tau,r}}.
\end{equation*}

The ITM loss $\mathcal{L}_c^{\mathrm{ITM}}$ is recovered from Newton Matching by choosing the direct-linearization loss~\eqref{eq:275}. Comparing the two losses yields the exact sample-wise correspondence
\begin{equation}
    \label{eq:164}
\mathcal{L}_c^{\mathrm{ITM}}(\theta)=\where{\calL_{\rho,\eta,B,\omega}^{\mathrm{dir}\text{-}\mathrm{lin}}(\theta)}{\rho=\rho^\base,\,\eta=\tau,\,B_t=\frac{1}{\tau}\log c,\,\omega_t=c^2}
\quad\text{under the forward construction.}
\end{equation}
Therefore, ITM is equivalent to the full-step update of the direct-linearization approximation under the forward construction, with specific choices of the baseline and weighting. It is critical-point consistent.

\subsection{DiffusionNFT}\label{subsec:27}

DiffusionNFT \citep{zheng2026diffusionnft} approaches the fine-tuning task from a contrastive-learning perspective.
At each data-collection stage, DiffusionNFT transforms the raw reward into an optimality probability by subtracting its prompt-wise mean, dividing by a positive normalizing factor, clipping the result to \([-1,1]\), and then mapping it affinely to \([0,1]\). For our analysis, we absorb this preprocessing into the definition of the reward model, suppress the prompt dependence, and denote the resulting bounded reward by \(r:\R^d\to[0,1]\).

At a canonical anchor $v^\rho$, its loss can be written in our notation as
\begin{align*}
    \mathcal{L}_{\rho,c}^{\mathrm{NFT}}(\theta)
    =&c^2\E_{\substack{t\sim\U(0,1),\,X_1\sim\rho,\,X_0\sim p_0,\,X_t=\alpha_tX_1+\beta_tX_0}}
    \Bigg[
    \bigg\|v_t^\theta(X_t)-v_{t|1}(X_t|X_1)\\
    &\qquad
    +\frac{2}{c}\left(r(X_1)-\frac{c+1}{2}\right)\left(v_t^\rho(X_t)-v_{t|1}(X_t|X_1)\right)\bigg\|_2^2
    \Bigg]+\const,
\end{align*}
where $c>0$ is a scalar hyperparameter and the additive constant is independent of $\theta$. The unique population minimizer is
\[
\bar{v}^{\mathrm{NFT}}=v^\rho+\frac{2}{c}\Gamma^{\rho,r}.
\]

The DiffusionNFT loss $\mathcal{L}_{\rho,c}^{\mathrm{NFT}}(\theta)$ is a special case of the covariance form from the no-regularization variant in approximate Newton Matching. Specifically, we drop the regularization term and set
\[
    \tilde r^\rho\approx r.
\]
We use the forward construction and the covariance form, and then set
\[
    \eta=\frac{2}{c},\qquad
    B_t\equiv \frac{c+1}{2}.
\]
Under these choices, comparing the two losses yields the exact sample-wise correspondence up to a $\theta$-independent additive term:
\begin{equation}
\mathcal{L}_{\rho,c}^{\mathrm{NFT}}(\theta)=c^2\cdot\where{\calL_{\rho,\eta,B}^{\mathrm{no}\text{-}\mathrm{reg}\text{-}\mathrm{cov}}(\theta)}{\eta=\frac{2}{c},\,B_t\equiv \frac{c+1}{2}}+\const
\qquad\text{under the forward construction.}
\label{eq:165}
\end{equation}
Therefore, DiffusionNFT is the covariance form of the unregularized update under the forward construction, stepsize $\eta=\frac2c$, and constant baseline $B_t\equiv\frac{c+1}2$. It is not critical-point consistent whenever $r$ is nonconstant.

\subsection{Reinforce Adjoint Matching}\label{subsec:28}

Reinforce Adjoint Matching (RAM) \citep{bergmeister2026reinforce} considers the fine-tuning case $\mu=\rho^\base$ with a canonical base velocity $v^\base$. Using our notation, the RAM loss can be written as
\begin{align*}
\mathcal{L}^{\mathrm{RAM}}(\theta)
=&\E_{\substack{t\sim\U(0,1),\,X_1\sim \rho^{\sg(\theta)},\,X_0\sim p_0,\,X_t=\alpha_tX_1+\beta_tX_0}}\bigg[\Big\|v_t^\theta(X_t)-v_t^\base(X_t)\\
&\qquad\qquad\qquad\qquad\qquad
-r(X_1)\left(v_{t|1}(X_t|X_1)-\sg\left(v_t^\theta(X_t)\right)\right)\Big\|_2^2\bigg],
\end{align*}
where \(X_1\sim\rho^{\sg(\theta)}\) denotes an endpoint generated by the ODE of the current model \(v^\theta\), with the dependence of the sampling procedure on \(\theta\) stopped during differentiation.

To derive the RAM loss from Newton Matching, we begin with the approximate-regularization construction of Section~\ref{sub2sec:16} and make the crude approximation of dropping the posterior-KL term:
\[
\nabla_{x_t}\KL{p_{1|t}^{\rho}(\cdot|x_t)}{p_{1|t}^{\base}(\cdot|x_t)}
\approx
0.
\]
This corresponds to setting \(\widehat K_t\equiv0\) in~\eqref{eq:143}. We then take the full-step choice \(\eta=\tau\), with \(\tau\) implicitly fixed at \(1\) in RAM, and set the baseline to zero:
\[
\eta=\tau=1,\qquad
\widehat K_t\equiv0,\qquad
B_t\equiv0.
\]
Under these choices, the additive update in~\eqref{eq:143} reduces to
\[
\bar v_t(x_t)
=
v_t^\rho(x_t)
+
\where{\widehat{\Gamma}_t^{\rho,\mathrm{appr}\text{-}\mathrm{reg}}(x_t)}{\tau=1,\,\widehat{K}_t\equiv0}
=
v_t^\base(x_t)
+
\Gamma_t^{\rho,r}(x_t).
\]
At a canonical anchor, its fixed-point condition is therefore
\begin{equation}\label{eq:166}
v^\rho=v^\base+\Gamma^{\rho,r}.
\end{equation}
By the stop-gradient construction of Section~\ref{subsec:25}, this fixed-point condition can be realized by the following loss:
\begin{align*}
&\where{\calL_{\rho,\eta,B}^{\mathrm{appr}\text{-}\mathrm{reg}\text{-}\mathrm{cov},\mathrm{sg}}(\theta)}{\eta=\tau=1,\,\widehat{K}_t\equiv 0,\,B_t\equiv 0}\\
=&
\E_{\substack{t\sim\U(0,1),\,(X_t,X_1)\sim \Xi_{t,1}^{\sg(\theta)}}}
\bigg[
\Big\|
v_t^\theta(X_t)
-v_t^\base(X_t)
-r(X_1)
\left(
v_{t|1}(X_t|X_1)-\sg\left(v_t^\theta(X_t)\right)
\right)
\Big\|_2^2
\bigg].
\end{align*}

Comparing this loss with \(\mathcal{L}^{\mathrm{RAM}}\) yields the exact sample-wise correspondence
\begin{equation}\label{eq:167}
\mathcal{L}^{\mathrm{RAM}}(\theta)
=
\where{\calL_{\rho,\eta,B}^{\mathrm{appr}\text{-}\mathrm{reg}\text{-}\mathrm{cov},\mathrm{sg}}(\theta)}{\eta=\tau=1,\,\widehat{K}_t\equiv 0,\,B_t\equiv 0}
\qquad
\text{under the forward construction.}
\end{equation}
Therefore, RAM is the stop-gradient covariance-form realization of the fixed-point condition~\eqref{eq:166} associated with the full-step approximate-regularization update, using the forward construction and a zero baseline.
RAM is not critical-point consistent in general.

\subsection{Adjoint Matching}\label{subsec:29}

Basic Adjoint Matching (BAM) and Lean Adjoint Matching (LAM) \citep{domingo2025adjoint} investigate the fine-tuning problem through SOC. They take a normalized reference density $\mu=\rho^\base\in\pdfspace$ and assume access to its canonical velocity field $v^\base=\mathcal C(\rho^\base)\in\velmancan$. The inverse temperature $\tau$ is set to $1$, corresponding to the target density $\pi_{\rho^\base,1,r}$.

For a current model $v^\theta$, the SOC problem can be written in our notation as
\begin{align*}
    \min_{\theta}\,\,&\E\left[\int_{0}^{1}\frac{1}{\kappa_t}\norm[2]{v_t^\theta(Y_t)-v_t^\base(Y_t)}^2\dd t-r\left(Y_1\right)\right],\\
    \mathrm{s.t.}\,\,&\dd Y_t=\left(2v_t^\theta(Y_t)-\frac{\dot{\alpha}_t}{\alpha_t}Y_t\right)\dd t+\sqrt{2\kappa_t}\dd W_t,\qquad t\in(0,1),\\
    &Y_0\sim p_0=\Normal(0,I).
\end{align*}
Both BAM and LAM generate the full path \(\BY_{[0,1]}\) by simulating the SDE above from \(Y_0\sim\Normal(0,I)\), and evaluate their losses along this shared trajectory. This corresponds to the multi-time supervision scheme for the gradient-form (under the reverse construction) introduced in Section~\ref{subsec:20}. For simplicity, we present the losses equivalently in single-time form. We let \(\Xi_{[t,1]}^\theta\) denote the same posterior requirement as \(\Xi_{[t,1]}^\rho\) in~\eqref{eq:95}, but with the current model \(v^\theta\) treated as canonical in place of \(v^\rho\). See also Remark~\ref{remark:7} for the discussion of canonicality in stop-gradient objectives.

\subsubsection{Basic Adjoint Matching}

Using our notation, the BAM loss can be written as
\[
\mathcal{L}^{\mathrm{BAM}}(\theta)=\E_{\substack{t\sim\U(0,1),\\\BY_{[t,1]}\sim\Xi^{\sg(\theta)}_{[t,1]}}}\left[\frac{1}{\kappa_t}\norm[2]{v_t^\theta(Y_t)-v_t^\base(Y_t)-\kappa_t\sg\left(\where{\lambda_t^{\mathrm{Bolza},\theta}(\BY_{[t,1]})}{\tau=1,\,\mu=\rho^\base}\right)}^2\right].
\]
Here, the pathwise adjoint $\lambda_t^{\mathrm{Bolza},\theta}(\BY_{[t,1]})$ satisfies the adjoint ODE~\eqref{eq:160}.

The BAM loss $\mathcal{L}^{\mathrm{BAM}}$ is recovered from the Bolza realization of the gradient form in exact Newton Matching under the reverse construction~\eqref{eq:110}. For the full-step choice $\eta=\tau=1$, the associated fixed-point condition at a canonical anchor is
\begin{equation*}
v^\rho=v^\rho+\Gamma^{\rho,\tilde{r}^\rho}.
\end{equation*}
By the stop-gradient construction of Section~\ref{subsec:25}, this fixed-point condition admits the stop-gradient loss $\calL_{\eta,\omega}^{\mathrm{grad,Bolza},\mathrm{sg}}$ in~\eqref{eq:279}. We then set
\[
\tau=\eta=1,\qquad\omega_t\equiv\frac{1}{\kappa_t},\qquad\mu=\rho^\base.
\]
With these choices, we obtain the exact sample-wise correspondence
\begin{equation}\label{eq:168}
\mathcal{L}^{\mathrm{BAM}}(\theta)=\where{\calL_{\eta,\omega}^{\mathrm{grad,Bolza},\mathrm{sg}}(\theta)}{\tau=\eta=1,\,\omega_t\equiv\frac{1}{\kappa_t},\,\mu=\rho^\base}
\qquad\text{under the reverse construction}.
\end{equation}
Therefore, BAM is the stop-gradient Bolza realization of the fixed-point condition associated with the full-step exact tangential update, using the reverse construction and a specific weighting.
According to Propositions~\ref{prop:7} and~\ref{prop:21}, the stop-gradient loss has $\canonicalmap(\pi_{\rho^\base,1,r})$ as its unique canonical population-stationary point.

\subsubsection{Lean Adjoint Matching}

LAM uses the same path distribution as BAM but replaces all occurrences of the current model \(v_t^\theta(Y_t)\) with the reference model \(v_t^\base(Y_t)\) in the adjoint ODE. Using our notation, the LAM loss can be written as
\[
\mathcal{L}^{\mathrm{LAM}}(\theta)=\E_{t\sim\U(0,1),\,\BY_{[t,1]}\sim\Xi^{\sg(\theta)}_{[t,1]}}\left[\frac{1}{\kappa_t}\norm[2]{v_t^\theta(Y_t)-v_t^\base(Y_t)-\where{\kappa_t\lambda_t^{b^\mu,r,0}(\BY_{[t,1]})}{\mu=\rho^\base}}^2\right].
\]
Here, the pathwise adjoint $\lambda_t^{b^\mu,r,0}(\BY_{[t,1]})$ is obtained via the reference-adjoint ODE~\eqref{eq:134}.

The LAM loss $\mathcal{L}^{\mathrm{LAM}}$ is recovered from Newton Matching by adopting the reference-adjoint approximation~\eqref{eq:138}. Following Section~\ref{sub2sec:14}, we approximate the pathwise adjoint by
\[
    \lambda_t^{b^\rho,r,l^{\rho,\mu}}
    \approx
    \lambda_t^{b^\mu,r,l^{\mu,\mu}}
    =
    \lambda_t^{b^\mu,r,0}.
\]
For the full-step choice $\eta=\tau=1$, the associated fixed-point condition at a canonical anchor is
\begin{equation*}
v^\rho=v^\rho+\widehat{\Gamma}^{\rho,\mathrm{ref}\text{-}\mathrm{adj}}.
\end{equation*}
By the stop-gradient construction of Section~\ref{subsec:25}, this fixed-point condition admits the stop-gradient loss $\calL_{\eta,\omega}^{\mathrm{ref}\text{-}\mathrm{adj},\mathrm{sg}}$ in~\eqref{eq:280}. We then set
\[
\tau=\eta=1,\qquad\omega_t\equiv\frac{1}{\kappa_t},\qquad\mu=\rho^\base.
\]
Consequently, we obtain the exact sample-wise correspondence
\begin{equation}\label{eq:169}
\mathcal{L}^{\mathrm{LAM}}(\theta)=\where{\calL_{\eta,\omega}^{\mathrm{ref}\text{-}\mathrm{adj},\mathrm{sg}}(\theta)}{\tau=\eta=1,\,\omega_t\equiv\frac{1}{\kappa_t},\,\mu=\rho^\base}
\qquad\text{under the reverse construction}.
\end{equation}
Therefore, LAM is the stop-gradient realization of the full-step reference-adjoint approximation update, using the reverse construction and a specific weighting.
According to Propositions~\ref{prop:14} and~\ref{prop:21}, the stop-gradient loss has $\canonicalmap(\pi_{\rho^\base,1,r})$ as its unique canonical population-stationary point.

\subsection{Flow Sampling}\label{subsec:30}

Flow Sampling (FS) \citep{havens2026flow} considers the sampling task $\mu=1$. FS is motivated by fixed-point iterations. It trains a drift model $b^\theta$ and uses SDE-based sampling. The inverse temperature \(\tau\) is set to \(1\).

In our notation, the FS loss can be written as
\begin{align*}
    \mathcal{L}^{\mathrm{FS}}(\theta)
    :=\E_{\substack{t\sim\U(0,1),\,X_1\sim\rho^{\sg(\theta)},\,X_0\sim p_0,\,X_t=\alpha_tX_1+\beta_tX_0}}&\Bigg[
        \bigg\|
    b_t^\theta(X_t)-b_{t|1}(X_t|X_1)\\
    &-\frac{\kappa_t}{\alpha_t}\left(\nabla r(X_1)+\frac{\alpha_t}{\beta_t^2}\left(X_t-\alpha_tX_1\right)\right)
    \bigg\|_2^2
    \Bigg].
\end{align*}
Here, \(\rho^{\sg(\theta)}\) denotes the terminal density induced by the current model through~\eqref{eq:281}, with the dependence of the sampling construction on $\theta$ placed under stop-gradient.

FS can be recovered from the Gaussian-kernel approximation of Newton Matching. We approximate the posterior Stein kernel by
\begin{equation*}
    \Lambda_t^\rho(x_1|x_t)
    \approx
    \widehat\Sigma_t^\rho(x_t),
\end{equation*}
and use the drift-coordinate extension of the approximation loss~\eqref{eq:142}. This extension is denoted by $\calL_{\rho,\eta,\widehat{\Sigma}}^{\mathrm{Gau}\text{-}\mathrm{ker},\mathrm{drift}}$ and is provided in~\eqref{eq:282}. The associated fixed-point condition at a canonical anchor is
\[
b_t^\rho(x_t)
=
b_t^\rho(x_t)
+
2\eta\frac{\alpha_t\kappa_t}{\beta_t^2}
\widehat\Sigma_t^\rho(x_t)
\E_{X_1\sim p_{1|t}^\rho(\cdot|x_t)}
\left[
\nabla r(X_1)
+
\frac{1}{\tau}\nabla\log\mu(X_1)
+
\frac{\alpha_t}{\tau\beta_t^2}
\left(
x_t-\alpha_tX_1
\right)
\right].
\]
By the stop-gradient construction of Section~\ref{subsec:25}, this fixed-point condition admits the stop-gradient loss $\calL_{\eta,\widehat{\Sigma}}^{\mathrm{Gau}\text{-}\mathrm{ker},\mathrm{drift},\mathrm{sg}}$ in~\eqref{eq:283}.

We then use the forward construction with
\[
\tau=1,\qquad
\eta=\frac\tau2=\frac12,\qquad
\widehat\Sigma_t\equiv\frac{\beta_t^2}{\alpha_t^2}I.
\]
Under these choices, comparing the two objectives yields the exact sample-wise correspondence
\begin{equation}\label{eq:170}
\mathcal{L}^{\mathrm{FS}}(\theta)
=
\where{\calL_{\eta,\widehat{\Sigma}}^{\mathrm{Gau}\text{-}\mathrm{ker},\mathrm{drift},\mathrm{sg}}(\theta)}{\mu=1,\,\tau=1,\,\eta=\frac12,\,\widehat\Sigma_t\equiv \frac{\beta_t^2}{\alpha_t^2}I}
\quad
\text{under the forward construction}.
\end{equation}
Therefore, FS is the stop-gradient realization of the damped Gaussian-kernel approximation update with stepsize \(\eta=\frac12\), using the drift coordinate and the forward construction with SDE-based sampling. According to Propositions~\ref{prop:15} and~\ref{prop:21}, the stop-gradient loss has $\canonicalmap(\pi_{1,1,r})$ as its unique canonical population-stationary point.

\subsection{Adjoint Sampling}\label{subsec:31}

Adjoint Sampling (AS) \citep{havens2025adjoint} considers the sampling task $\mu=1$. AS is motivated by the SOC perspective. It trains a drift model, and during sampling, the SDE is initialized at the deterministic state \(Y_0=0\).

We show that, by extending Newton Matching to the one-sided interpolant, we can sidestep all SOC-related elements, such as the reference processes, and specify the schedule directly from the one-sided interpolant. This extension is introduced in Appendix~\ref{subsec:57}, and the detailed derivation of the correspondence between AS and Newton Matching is provided in Appendix~\ref{subsec:63}.

With a noise schedule $\sigma_t$, AS adopts
\begin{equation*}
\widetilde\gamma_t:=\int_{0}^{t}\sigma_u^2\,\dd u,\quad
\widetilde\alpha_t:=\frac{\widetilde\gamma_t}{\widetilde\gamma_1},\quad
\widetilde\beta_t:=\sqrt{\frac{\widetilde\gamma_t(\widetilde\gamma_1-\widetilde\gamma_t)}{\widetilde\gamma_1}},\quad
\widetilde{\kappa}_t:=\frac{\widetilde\beta_t}{\widetilde\alpha_t}\left(\dot{\widetilde\alpha}_t\widetilde\beta_t-\widetilde\alpha_t\dot{\widetilde\beta}_t\right)=\frac12\sigma_t^2.
\end{equation*}
The detailed derivation is deferred to Appendix~\ref{subsec:63}.

The AS loss can be written as
\begin{align*}
    \mathcal{L}^{\mathrm{AS}}(\theta)
    =&\E_{\substack{
        t\sim\U(0,1),\,X_1\sim\rho^{\sg(\theta)},\,Z\sim\Normal(0,I),\,
        X_t=\widetilde\alpha_tX_1+\widetilde\beta_tZ}}
        \Bigg[\frac{1}{8\widetilde\kappa_t^2}\bigg\|b_t^\theta(X_t)-\widetilde{b}_{t|1}(X_t|X_1)\notag\\
        &\qquad\qquad\qquad\qquad\qquad\qquad\qquad
        -2\widetilde\kappa_t\tau\bigg(\nabla r(X_1)+\frac{\widetilde{\alpha}_t}{\tau\widetilde{\beta}_t^2}\left(X_t-\widetilde{\alpha}_t X_1\right)\bigg)\bigg\|_2^2\Bigg].
\end{align*}
Here, $X_1\sim\rho^{\sg(\theta)}$ denotes the terminal density induced by the current model through~\eqref{eq:284}, with the dependence of the sampling construction on $\theta$ placed under stop-gradient.

AS can be recovered from the Gaussian-kernel approximation of Newton Matching. At a canonical anchor, we approximate the posterior Stein kernel by
\begin{equation*}
    \Lambda_t^\rho(x_1|x_t)
    \approx
    \widehat\Sigma_t^\rho(x_t),
\end{equation*}
and extend the approximation loss~\eqref{eq:142} to the drift coordinate and the one-sided interpolant. This loss is denoted by $\widetilde{\calL}_{\rho,\eta,\widehat{\Sigma},\omega}^{\mathrm{Gau}\text{-}\mathrm{ker},\mathrm{drift}}$ and is provided in~\eqref{eq:287}. The associated fixed-point condition at a canonical anchor is
\begin{align*}
    \widetilde b_t^\rho(x_t)
    =\widetilde b_t^\rho(x_t)+
        2\eta\frac{\widetilde\alpha_t\widetilde\kappa_t}{\widetilde\beta_t^2}
        \widehat\Sigma_t^\rho(x_t)
    \E_{X_1\sim \widetilde{p}_{1|t}^\rho(\cdot|x_t)}
    \left[
            \nabla r(X_1)
            +
            \frac{1}{\tau}\nabla\log\mu(X_1)
            +
            \frac{\widetilde\alpha_t}{\tau\widetilde\beta_t^2}
            \left(
                x_t-\widetilde\alpha_tX_1
            \right)
    \right].
\end{align*}
By the stop-gradient construction of Section~\ref{subsec:25}, this fixed-point condition admits the stop-gradient loss $\widetilde{\calL}_{\rho,\eta,\widehat{\Sigma},\omega}^{\mathrm{Gau}\text{-}\mathrm{ker},\mathrm{drift},\mathrm{sg}}$ in~\eqref{eq:288}.

We use the forward construction and set
\[
\eta=\tau,\qquad
\omega_t\equiv\frac{1}{8\widetilde\kappa_t^2},\qquad
\widehat\Sigma_t\equiv \frac{\widetilde\beta_t^2}{\widetilde\alpha_t}I.
\]
Under these choices, comparing the two objectives yields the exact sample-wise correspondence
\begin{align}
    \mathcal{L}^{\mathrm{AS}}(\theta)
    =
    \where{\widetilde{\calL}_{\rho,\eta,\widehat{\Sigma},\omega}^{\mathrm{Gau}\text{-}\mathrm{ker},\mathrm{drift},\mathrm{sg}}(\theta)}{\mu=1,\,\widetilde\alpha_t=\frac{\widetilde\gamma_t}{\widetilde\gamma_1},\,
    \widetilde\beta_t=\sqrt{\frac{\widetilde\gamma_t(\widetilde\gamma_1-\widetilde\gamma_t)}{\widetilde\gamma_1}},\,\eta=\tau,\,\widehat\Sigma_t\equiv \frac{\widetilde\beta_t^2}{\widetilde\alpha_t}I,\,\omega_t\equiv\frac{1}{8\widetilde\kappa_t^2}}
    \label{eq:171}
\end{align}
under the forward construction.

Therefore, AS is the stop-gradient realization of the full-step Gaussian-kernel approximation update under the forward construction with SDE-based sampling, using the drift coordinate and the one-sided interpolant. The stop-gradient loss has $\canonicalmap(\pi_{1,\tau,r})$ as its unique canonical population-stationary point.

\section{Conclusion}

We developed Newton Matching, a unified framework for fine-tuning and sampling when the target distribution is specified through its defining factors rather than direct samples. The central conceptual shift is from treating methods as isolated regression losses to viewing learning as iterative optimization over canonical models---the population minimizers that standard conditional matching associates with terminal densities. This restriction removes representational redundancy while retaining the conditional-expectation structure that makes diffusion and flow training scalable. Within this general formulation, we specialized to reverse-KL minimization and proved that the mixture-connection Newton direction coincides with the negative Fisher--Rao gradient. The canonical retraction realizes finite steps along this direction through a tangential update followed by terminal-density-preserving canonicalization. Its exact terminal-density characterization yields strict reverse-KL descent for every damped or full step, global convergence under mild conditions, and local quadratic convergence of the full-step iteration. We further derived multiple population-exact realizations and approximate variants organized by critical-point consistency. By recovering representative existing methods as configurations of shared building blocks, Newton Matching demystifies their relationships and distinguishes exact Newton updates, target-preserving approximations, and variants that alter the underlying objective. More broadly, Newton Matching lays a foundation for principled post-training of generative models from reward feedback, providing a common language for designing and analyzing distributional objectives, finite-stepsize updates, and scalable regression realizations. In doing so, it advances both theoretical understanding and principled design of reinforcement learning algorithms for generative models.

\newpage

\begin{appendices}
    
\section*{Appendices}

The appendices are organized into five groups.

Appendices~\ref{app:1}--\ref{app:3} provide the geometric and analytic theory behind Newton Matching. Appendix~\ref{app:1} establishes the mathematical foundations underlying the canonical geometry, the finite-stepsize canonical retraction, and the Newton Matching construction in Sections~\ref{sec:1}--\ref{sec:3}. It specifies a compatible smooth Banach-manifold realization and states the regularity assumptions under which the global, first-order, and second-order geometric arguments in the main text are well-defined. It also proves that the canonical manifold $\velmancan$ is generally non-affine, thereby motivating the canonical retraction used for finite-stepsize updates. Appendix~\ref{app:2} provides the technical convergence analysis underlying Section~\ref{sec:4}. It turns the one-stage descent properties established in the main text into global convergence guarantees, identifies the topology-compatible metrics in which local quadratic convergence is measured on the density and canonical manifolds, and establishes the guarantee for the continuation scheme. Appendix~\ref{app:3} complements this general theory with an analytically solvable Gaussian model, for which the Newton Matching iteration admits an explicit finite-dimensional characterization and both global convergence and local quadratic convergence can be verified directly.

Appendices~\ref{app:4}--\ref{app:5} provide the mathematical and computational tools underlying the exact and approximate Newton Matching developed in Sections~\ref{sec:5} and~\ref{sec:6}. Appendix~\ref{app:4} develops stochastic representations, posterior-preserving dynamics, universal bridges, log-density and log-density-ratio identities, initial-state sensitivity, and adjoint calculus. These results make the posterior quantities and density-ratio correction required by Newton Matching accessible through sampling and pathwise computation. Appendix~\ref{app:5} complements these tools by developing Stein control variates and posterior Stein kernels. In particular, it establishes the covariance--gradient identity that converts the
covariance-form target~\eqref{eq:56} into the gradient-form target~\eqref{eq:59}.%

Appendix~\ref{app:6} provides the technical analysis behind approximate Newton Matching. It collects the proofs deferred from Section~\ref{sec:6}, identifies the population-stationary points of the approximate regression objectives, establishes the critical-point-consistency guarantees stated for the covariance and gradient approximations, and proves the conclusions associated with the regularization trade-offs.

Appendices~\ref{app:7}--\ref{app:8} provide two complementary extensions of the Newton Matching framework. Appendix~\ref{app:7} replaces the constant stepsize $\eta$ in the tangential update by a time-dependent modulation $\eta_t$. Although the resulting modulated field need not remain tangent to the canonical manifold $\velmancan$, the appendix derives the exact formula for its terminal-density update and shows that nondecreasing modulation schedules preserve the value-ascent certificate in Section~\ref{subsec:8} and the reverse-KL descent in Section~\ref{sub2sec:7}. Appendix~\ref{app:8} instead changes the representation used to realize the update. It demonstrates how the exact and approximate regression objectives can be realized in score and drift coordinates. It also illustrates a representative alternative CFM construction: the one-sided interpolant with deterministic initial state \(X_0=0\).

Finally, Appendix~\ref{app:9} provides the detailed method-specific derivations deferred from Section~\ref{sec:7}. By translating the notation, sampler, and parameterization of the original works into the language of this paper, we establish the stated correspondences and show how these methods arise as special cases of exact or approximate Newton Matching algorithms. The derivations for Flow Sampling and Adjoint Sampling additionally use the drift-coordinate and one-sided-interpolant constructions developed in Appendix~\ref{app:8}.

\section{Geometric Foundation}\label{app:1}

This appendix provides a rigorous mathematical setting for the canonical geometry, the finite-stepsize canonical retraction, and the Newton Matching construction presented in Sections~\ref{sec:1}--\ref{sec:3}. It defines the relevant spaces and maps, equips the density and velocity classes with compatible Banach-manifold structures, and states the regularity assumptions under which the global, first-order, and second-order arguments in Newton Matching, as well as the local convergence analysis of Section~\ref{subsec:13}, are well-defined.

First, Appendix~\ref{subsec:32} introduces ambient velocity and density classes on which ODE flows, terminal densities, and velocities from the CFM construction are well-defined. Then, Appendix~\ref{subsec:33} selects admissible subclasses and constructs the canonical subclass and projection at the set-theoretic level. Appendix~\ref{subsec:34} equips these classes with compatible Banach-manifold structures and develops the associated first-order and second-order differential geometry used in Newton Matching. Finally, Appendix~\ref{subsec:35} gives a Gaussian counterexample showing that $\velmancan$ need not be affine, thereby motivating the canonical retraction used for finite-stepsize updates.

\subsection{Ambient Classes}\label{subsec:32}

The ambient classes provide a common regularity setting for the constructions that follow. They are chosen so that the ODE flow is globally well posed, the densities transported by the flow have finite second moments, and the canonical velocity field associated with each terminal density is pointwise well defined. No manifold structure is imposed at this stage.

\subsubsection{Ambient Velocity Space}\label{sub2sec:21}

Here, we define a Banach space of time-dependent velocity fields $v=(v_t)_{t\in[0,1]}$ that are integrable in time and globally Lipschitz in space. These conditions are sufficient for the existence of a unique global Carath\'eodory flow and provide the growth estimate needed to control pushforward moments.

\begin{definition}\label{def:4}
    The \textit{ambient velocity space} is defined as the Banach space
    \begin{equation*}
        \widehat{\velspace}:=\left\{v=(v_t)_{t\in[0,1]}\in L^1\left([0,1];C_{\mathrm{lin}}^1\right): \norm[\widehat{\velspace}]{v}:=\int_{0}^{1}\norm[C_{\mathrm{lin}}^1]{v_t}\dd t<\infty\right\},
    \end{equation*}
    where the Lipschitz function space $C_{\mathrm{lin}}^1$ is the Banach space
    \begin{equation*}%
        C_{\mathrm{lin}}^1:=\left\{v\in C^1(\R^d;\R^d):\norm[C_{\mathrm{lin}}^1]{v}:=\norm[2]{v(0)}+\norm[\infty]{\nabla v}<\infty\right\}.
    \end{equation*}
    In other words, $t\mapsto\norm[C_{\mathrm{lin}}^1]{v_t}$ is integrable on $[0,1]$. For a.e. $t\in[0,1]$, the spatial vector field $v_t$ is $C^1$ and globally Lipschitz continuous.
\end{definition}

Throughout the ambient construction, the ODE $\dot Y_t=v_t(Y_t)$ is understood in the Carath\'eodory sense: $t\mapsto Y_t$ is absolutely continuous and satisfies
\[
Y_t=Y_s+\int_s^t v_u(Y_u)\,\dd u.
\]
Equivalently, $\dot Y_t=v_t(Y_t)$ holds for a.e. $t\in[0,1]$. The following lemma records the resulting global well-posedness and the quantitative growth bound used below.

\begin{lemma}\label{lem:1}
    For any velocity field $v\in\widehat{\velspace}$, we denote its Carath\'eodory ODE flow by $\Phi_{0\to t}^v:\R^d\to\R^d$. Then, for any initial point $x_0\in\R^d$, the ODE admits a unique nonexplosive solution $x_t=\Phi_{0\to t}^v(x_0)$ that satisfies
    \begin{equation}\label{eq:172}
        \norm[2]{x_t}\le e^{\norm[\widehat{\velspace}]{v}}(\norm[2]{x_0}+\norm[\widehat{\velspace}]{v}).
    \end{equation}
\end{lemma}

\begin{proof}
    The existence of the ODE flow is a classical result \cite[Chapter 6]{jafarpour2014time}. By Gr\"onwall's inequality, for any initial point $x_0$, the ODE solution satisfies the pathwise estimate
    \begin{equation*}
        \norm[2]{x_t}\le\left(\norm[2]{x_0}+\int_{0}^{t}\norm[2]{v_s(0)}\,\dd s\right)\exp\left(\int_{0}^{t}\norm[\infty]{\nabla v_s}\,\dd s\right)\le e^{\norm[\widehat{\velspace}]{v}}(\norm[2]{x_0}+\norm[\widehat{\velspace}]{v}).
    \end{equation*}
    Therefore, the regularity condition $\norm[\widehat{\velspace}]{v}<\infty$ implies that the ODE admits a unique nonexplosive solution.
\end{proof}

\subsubsection{Ambient Density Class}

Here, we identify a broad class of terminal densities compatible with the ambient velocity space. The first result shows that the flow of every $v\in\widehat{\velspace}$ transports the Gaussian source to a positive continuous density with finite second moment.

\begin{lemma}\label{lem:2}
    Consider the source density $p_0=\Normal(0,I)$ and the probability path $(\mathbb{P}_t^v)_{t\in[0,1]}$ generated by the pushforward
    \begin{equation*}
        \mathbb{P}_t^v:=(\flow_{0\to t}^v)_\#(p_0\,\dd x).
    \end{equation*}
    Then, $\mathbb{P}_t^v=p_t^v\,\dd x$ has finite second moment, where $p_t^v$ is a positive, normalized, continuous density. The density $p_t^v$ also satisfies the continuity equation
    \begin{equation*}
        \partial_t p_t^v(x_t)+\nabla_{x_t}\cdot\left(p_t^v(x_t)v_t(x_t)\right)=0,
    \qquad
    p_0^v=p_0
    \end{equation*}
    in the weak sense.
\end{lemma}

\begin{proof}
    According to \cite[Chapter 6]{jafarpour2014time}, $\Phi_{t\to0}^v$ is a $C^1$ diffeomorphism. Then, we have the explicit expression
    \begin{equation*}
        p_t^v(y)=p_0(\Phi_{t\to0}^v(y))\abs{\det\nabla\Phi_{t\to0}^v(y)}.
    \end{equation*}
    Hence, $p_t^v$ is a positive, normalized, continuous density, rather than merely a probability measure. Furthermore,~\eqref{eq:172} implies that
    \begin{equation*}
        \E_{X_t\sim p_t^v}\left[\norm[2]{X_t}^2\right]\le\E_{X_0\sim \Normal(0,I)}\left[e^{2\norm[\widehat{\velspace}]{v}}(\norm[2]{X_0}+\norm[\widehat{\velspace}]{v})^2\right]<\infty.
    \end{equation*}
    In other words, $p_t^v$ has finite second moment. The continuity equation is a classical result \cite[Section 8.1]{ambrosio2005gradient}.
\end{proof}

Throughout this paper, we identify a density $p$ with its associated probability measure $p\,\dd x$. Therefore, we can write $p_t^v:=(\flow_{0\to t}^v)_\#p_0$.

Lemma~\ref{lem:2} shows that every ambient velocity field generates a density of the following form. This motivates a density class that is broad enough to contain all ambient terminal densities while retaining the moment and positivity assumptions needed later.

\begin{definition}\label{def:5}
    The \textit{ambient density class} is defined by
    \begin{equation*}%
        \widehat{\pdfspace}:=\left\{
            \rho\in C\left(\R^d;(0,+\infty)\right):\int_{\R^d}\rho(x)\,\dd x=1,\,\E_{X\sim \rho}\left[\norm[2]{X}^2\right]<\infty
        \right\}.
    \end{equation*}

    Define the \textit{ambient terminal-density map} by
    \begin{equation*}%
        \widehat{\terminalmap}:\widehat{\velspace}\to\widehat{\pdfspace},\qquad
        v\mapsto \rho^v:=(\flow_{0\to1}^v)_\#p_0,
    \end{equation*}
    where $\flow_{s\to t}^v:\R^d\to\R^d$ is the Carath\'eodory ODE flow generated by $v$.
\end{definition}

The finite-second-moment condition $\E_{X_1\sim \rho}\left[\norm[2]{X_1}^2\right]<\infty$, together with the boundedness of $\dot\alpha$ and $\dot\beta$ implied by their continuity on the compact interval $[0,1]$, is precisely what is needed for square integrability of the CFM target. For every $\rho\in\widehat{\pdfspace}$, we have
\[
\E_{t\sim\U(0,1),\,X_0\sim p_0,\,X_1\sim\rho}\left[\norm[2]{\dot\alpha_tX_1+\dot\beta_tX_0}^2\right]<\infty.
\]
Consequently, the CFM loss~\eqref{eq:3} and the conditional-expectation velocity are integrable whenever
\[
\E_{t\sim\U(0,1),\,X_0\sim p_0,\,X_1\sim\rho}\left[\norm[2]{v_t^\theta(X_t)}^2\right]<\infty.
\]
Here $\theta$ is the parameter of the CFM neural network. Strict positivity $\rho(x)>0$ is imposed to avoid singularities in the log-density ratios, the KL divergence, and the Fisher--Rao metric used later.

\subsubsection{Ambient Canonical-Velocity Map}

Fix the interpolant schedule $(\alpha_t,\beta_t)_{t\in(0,1)}$. For each terminal density, the population minimizer of the CFM loss~\eqref{eq:3} defines a pointwise velocity field through conditional expectation~\eqref{eq:4}. We first establish that this assignment and its marginal probability path are well-defined. The resulting velocity need not belong to $\widehat{\velspace}$ for every $\rho\in\widehat{\pdfspace}$; terminal correctness is asserted whenever the required ambient velocity regularity holds.

\begin{lemma}\label{lem:3}
    Consider any terminal density $\rho\in\widehat{\pdfspace}$. For all $t\in[0,1)$ and $x_t\in\R^d$, the computations
    \[
    p_{t}^\rho(x_t):=\int_{\R^d}p_{t|1}(x_t|x_1)\rho(x_1)\,\dd x_1>0,\qquad v_t^\rho(x_t):=\frac{\int_{\R^d}v_{t|1}(x_t|x_1)p_{t|1}(x_t|x_1)\rho(x_1)\,\dd x_1}{p_{t}^\rho(x_t)}
    \]
    are well-defined. For $t=1$, we define
    \[
    p_1^\rho(x_1):=\rho(x_1),\qquad v_1^\rho(x_1):=\dot{\alpha}_1x_1.
    \]
    In particular, $(v_t^\rho,p_t^\rho)$ satisfies the continuity equation weakly:
    \begin{equation*}
        \partial_t p_t^\rho(x_t)+\nabla_{x_t}\cdot\left(p_t^\rho(x_t)v_t^\rho(x_t)\right)=0,\qquad p_0^\rho=p_0,\qquad p_1^\rho=\rho.
    \end{equation*}
    
    Moreover, if $v^\rho\in\widehat{\velspace}$, then
    \[
    \rho=\widehat{\terminalmap}(v^\rho).
    \]
\end{lemma}

\begin{proof}
    According to~\eqref{eq:11}, $p_{t|1}(x_t|\cdot)$ is bounded and positive if $t\in[0,1)$; hence, the integral $p_{t}^\rho(x_t)$ is finite and positive. If $t=1$, then $p_{1}^\rho(x_1)=\rho(x_1)$ is also positive. By~\eqref{eq:12}, $v_{t|1}(x_t|\cdot)p_{t|1}(x_t|\cdot)$ has at most linear growth if $t\in[0,1)$; hence, the numerator of $v_t^\rho(x_t)$ is absolutely convergent.

    According to \cite[Theorem 1]{lipman2023flow}, $v_t^\rho$ and $p_t^\rho$ satisfy the continuity equation.
    
    By Jensen's inequality, we have
    \[
        \begin{aligned}
            \int_{0}^{1}\E_{X_t\sim p_t^\rho}\left[\norm[2]{v_t^\rho(X_t)}^2\right]\dd t
            =&\int_{0}^{1}\E_{X_t\sim p_t^\rho}\left[\norm[2]{\E_{X_1\sim p_{1|t}^\rho(\cdot|X_t)}\left[v_{t|1}(X_t|X_1)\right]}^2\right]\dd t\\
            \leq&\int_{0}^{1}\E_{X_t\sim p_t^\rho,\,X_1\sim p_{1|t}^\rho(\cdot|X_t)}\left[\norm[2]{v_{t|1}(X_t|X_1)}^2\right]\dd t\\
            =&\int_{0}^{1}\E_{X_0\sim p_0,\,X_1\sim\rho}\left[\norm[2]{\dot\alpha_tX_1+\dot\beta_tX_0}^2\right]\dd t\\
            =&\E_{X_0\sim p_0}\left[\norm[2]{X_0}^2\right]\int_{0}^{1}\dot{\beta}_t^2\dd t+\E_{X_1\sim \rho}\left[\norm[2]{X_1}^2\right]\int_{0}^{1}\dot{\alpha}_t^2\dd t\\
            <&\infty,
        \end{aligned}
    \]
    which implies the regularity condition \cite[Equation (8.1.2)]{ambrosio2005gradient}. If $v^\rho\in\widehat{\velspace}$, then the integrability condition \cite[Equation (8.1.7)]{ambrosio2005gradient} holds, and the ODE flow $\Phi^{v^\rho}$ is well-defined. By the representation formula for the continuity equation \cite[Proposition 8.1.8]{ambrosio2005gradient}, $\rho$ is the terminal density of $v$. In this case, we have $\rho=\widehat{\terminalmap}(v^\rho)$.
\end{proof}

\begin{definition}
    For any terminal density $\rho\in\widehat\pdfspace$, its \textit{canonical velocity field} $v^\rho=(v_t^\rho)_{t\in[0,1]}$ is defined as
    \begin{equation*}
        v_t^\rho(x_t)=
        \begin{dcases}
        \E_{X_1\sim p_{1|t}^\rho(\cdot| x_t)}\left[v_{t|1}(x_t| X_1)\right],&t\in[0,1),\\
        \dot{\alpha}_1x_1,&t=1,
        \end{dcases}
    \end{equation*}
    which is pointwise well-defined and serves as the population minimizer of the CFM loss~\eqref{eq:3}. We define the \textit{ambient canonical-velocity map} as
    \[
    \widehat{\canonicalmap}:\widehat{\pdfspace}\to\left\{v:[0,1]\times\R^d\to\R^d\right\},\qquad
    \rho\mapsto v^\rho=(v_t^\rho)_{t\in[0,1]}.
    \]
\end{definition}

\begin{corollary}\label{cor:3}
    On the set of densities whose canonical velocity belongs to $\widehat{\velspace}$, the ambient canonical-velocity map is the inverse of the restricted ambient terminal-density map:
    \begin{equation*}
        \where{\widehat{\terminalmap}}{\widehat{\canonicalmap}\left(\widehat{\pdfspace}\right)\cap\widehat{\velspace}}=\left(\where{\widehat{\canonicalmap}}{\widehat{\canonicalmap}^{-1}\left(\widehat{\velspace}\right)}\right)^{-1}.
    \end{equation*}
\end{corollary}

\begin{proof}
    For every $\rho\in\widehat{\canonicalmap}^{-1}(\widehat{\velspace})$, Lemma~\ref{lem:3} gives $\widehat{\terminalmap}(\widehat{\canonicalmap}(\rho))=\rho$, which finishes the proof.
\end{proof}

\begin{remark}\label{remark:8}
    The ambient terminal-density map $\widehat\terminalmap:\widehat\velspace\to\widehat\pdfspace$ is generally not injective. For example, suppose $w=(w_t)_{t\in[0,1]}$ is sufficiently regular, $v+w\in\widehat{\velspace}$, and $\nabla\cdot(p_t^v w_t)=0$ for a.e.\ $t\in[0,1]$. Then $p_t^v$ satisfies the continuity equation for both $v$ and $v+w$, so the two velocity fields generate the same density path and, in particular, the same terminal density. Therefore, the ambient canonical-velocity map $\widehat{\canonicalmap}$ selects one velocity field $v^\rho$ among many possible representations of a terminal density $\rho$.
\end{remark}

The relations among these ambient classes and their maps are summarized in Figure~\ref{fig:11}.

\subsection{Set-Theoretic Definitions}\label{subsec:33}

The ambient classes in Appendix~\ref{subsec:32} are deliberately broad and need not be closed under the operations required by Newton Matching. We therefore select a density class and a velocity class on which both the terminal-density map and the canonical-velocity map take values in the desired classes. At this stage, the density and velocity classes are treated simply as sets, and the associated maps as set maps; no topology or smooth manifold structure is introduced.

\subsubsection{Admissible Subclasses and Restricted Maps}

\begin{figure}[t]
    \centering
    \input{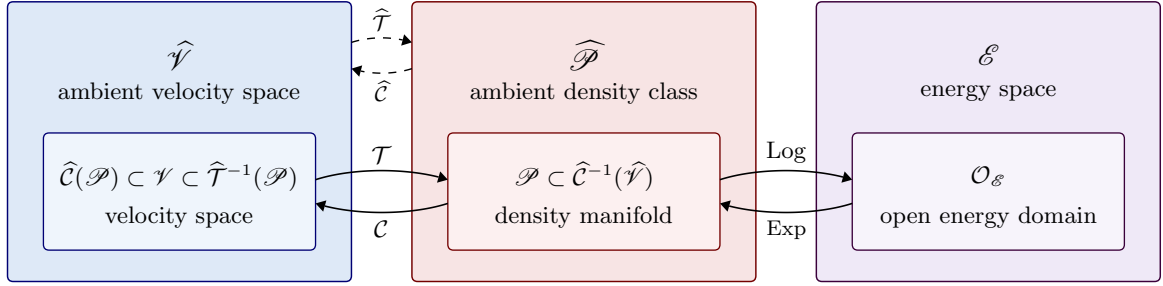}
    \caption{Construction of ambient classes, admissible subclasses, and smooth structures.}
    \label{fig:11}
\end{figure}

We first select terminal densities whose canonical velocity fields satisfy the ambient velocity regularity. This condition guarantees that the canonical assignment is compatible with the ambient terminal-density map.

\begin{definition}\label{def:6}
    A \textit{density class} $\pdfspace$ is a non-empty subset of the ambient density class $\widehat{\pdfspace}$ satisfying
    \[
    \pdfspace\subset \widehat{\canonicalmap}^{-1}\left(\widehat{\velspace}\right)\subset\widehat{\pdfspace}.
    \]
\end{definition}

Definition~\ref{def:6} ensures that every selected density has a canonical velocity in $\widehat{\velspace}$. Ambient terminal correctness then places that canonical velocity in the corresponding terminal-density preimage.

\begin{corollary}\label{cor:4}
    $\widehat\canonicalmap(\pdfspace)\subset\widehat{\terminalmap}^{-1}(\pdfspace)$.
\end{corollary}

\begin{proof}
    By Definition~\ref{def:6}, we have $\pdfspace\subset \widehat{\canonicalmap}^{-1}\left(\widehat{\velspace}\right)$. Then,
    \[
        \widehat{\canonicalmap}(\pdfspace)=\where{\widehat{\canonicalmap}}{\widehat{\canonicalmap}^{-1}\left(\widehat{\velspace}\right)}(\pdfspace)=\left(\where{\widehat{\terminalmap}}{\widehat{\canonicalmap}\left(\widehat{\pdfspace}\right)\cap\widehat{\velspace}}\right)^{-1}(\pdfspace)\subset\widehat{\terminalmap}^{-1}(\pdfspace)
    \]
    is given by Corollary~\ref{cor:3}.
\end{proof}

Based on Corollary~\ref{cor:4}, we can define the admissible velocity subclasses as follows. The selected velocity class may be strictly smaller than the full terminal-density preimage, which allows additional regularity to be imposed without excluding the canonical velocities. We also define the maps between these admissible subclasses as the restricted maps in Appendix~\ref{subsec:32}.

\begin{definition}
    A \textit{velocity class} $\velspace$ compatible with $\pdfspace$ is defined as a subset of the ambient velocity space $\widehat{\velspace}$ satisfying
    \[
    \widehat{\canonicalmap}(\pdfspace)\subset\velspace\subset\widehat{\terminalmap}^{-1}(\pdfspace)\subset\widehat{\velspace}.
    \]
    The \textit{terminal-density map} is defined as
    \[
        \terminalmap:=\where{\widehat{\terminalmap}}{\velspace}:\velspace\to\pdfspace,\qquad
        v\mapsto \rho^v:=(\flow_{0\to1}^v)_\#p_0.
    \]
    The \textit{canonical-velocity map} is defined as
    \[
        \canonicalmap:=\where{\widehat{\canonicalmap}}{\pdfspace}:\pdfspace\to\canonicalmap(\pdfspace)\subset\velspace,\qquad
        \rho\mapsto v^\rho,\qquad
        v_t^\rho(x_t)=\E_{X_1\sim p_{1|t}^\rho(\cdot| x_t)}\left[v_{t|1}(x_t| X_1)\right].
    \]
\end{definition}

The defining restrictions preserve the ambient terminal-correctness identity.

\begin{corollary}\label{cor:5}
    The canonical-velocity map $\canonicalmap$ is a right inverse of the terminal-density map $\terminalmap$:
    \[
    \terminalmap\circ\canonicalmap=\Id_\pdfspace.
    \]
\end{corollary}

\begin{proof}
    This corollary is the restriction of Corollary~\ref{cor:3} to $\pdfspace$.
\end{proof}

The relations among the two admissible subclasses and their maps are summarized in Figure~\ref{fig:11}.

\subsubsection{Canonical Subclass}

The following definition singles out the canonical velocity field $v^\rho$ for each density $\rho$. Its image forms the canonical subclass.

\begin{definition}
    The \textit{canonical subclass} $\velmancan$ consists of the canonical velocity field of densities in $\pdfspace$, i.e.,
    \[
    \velmancan:=\canonicalmap(\pdfspace)\subset\velspace.
    \]
    The \textit{canonical projection} is defined as
    \[
        \projection:=\canonicalmap\circ\terminalmap:\velspace\to\velmancan.
    \]
\end{definition}

\begin{theorem}\label{thm:10}
    The canonical-velocity map $\canonicalmap$ is the inverse of the restricted terminal-density map, i.e.,
    \[
    \where{\terminalmap}{\velmancan}=\canonicalmap^{-1}.
    \]
    In other words, if $v^\rho=\canonicalmap(\rho)\in\velspace$ is a canonical velocity field of a density $\rho\in\pdfspace$, then $\rho$ is also the terminal density $\terminalmap(v^\rho)$ of $v^\rho$. Therefore, the canonical-velocity map $\canonicalmap$ induces a set-theoretic identification, i.e.,
    \[
    \pdfspace\cong_\mathrm{set}\velmancan.
    \]

    Consequently, the canonical projection $\projection$ preserves the terminal density, i.e.,
    \[
    \terminalmap\circ\projection=\terminalmap.
    \]
    In other words, for any $v\in\velspace$, its canonical velocity $\projection(v)\in\velspace$ has the same terminal density as $v$, i.e., $\terminalmap(v)=\terminalmap(\projection(v))\in\pdfspace$.

    Furthermore, the canonical projection $\projection$ is idempotent, i.e.,
    \[
    \projection^2=\projection,\qquad\where{\projection}{\velmancan}=\Id_{\velmancan}.
    \]
    In other words, the projection of a canonical velocity field $v^\rho\in\velmancan$ is itself, i.e., $\projection(v^\rho)=v^\rho$.
\end{theorem}

\begin{proof}
    According to Corollary~\ref{cor:5}, we have $\where{\terminalmap}{\velmancan}=\canonicalmap^{-1}$. Consequently, we have
    \[
    \terminalmap\circ\projection=\terminalmap\circ\canonicalmap\circ\terminalmap=\where{\terminalmap}{\velmancan}\circ\canonicalmap\circ\terminalmap=\canonicalmap^{-1}\circ\canonicalmap\circ\terminalmap=\terminalmap.
    \]
    Furthermore,
    \[
    \projection^2=\canonicalmap\circ\terminalmap\circ\projection=\canonicalmap\circ\terminalmap=\projection, \qquad
    \where{\projection}{\velmancan}=\canonicalmap\circ\where{\terminalmap}{\velmancan}=\canonicalmap\circ\canonicalmap^{-1}=\Id_{\velmancan},
    \]
    which finishes the proof.
\end{proof}

The construction of the canonical subclass is shown in Figure~\ref{fig:6}.

\subsection{Smooth Geometric Realization}\label{subsec:34}

Appendix~\ref{subsec:33} establishes the canonical correspondence only at the level of sets. Those identities identify a unique canonical velocity representative for each terminal density, but they do not yet support the differential operations used in Sections~\ref{subsec:5} and~\ref{subsec:6}. Tangent and cotangent spaces, metric duality, covariant differentiation, and smooth retractions require compatible Banach-manifold structures. In this subsection, we therefore equip $\pdfspace$ with a global centered-energy chart and $\velspace$ with the Banach structure inherited from its realization as a closed linear subspace of $\widehat{\velspace}$. Assuming that $\terminalmap$ and $\canonicalmap$ are smooth, the set-theoretic identities from Appendix~\ref{subsec:33} then become smooth geometric statements: $\velmancan$ is a split embedded Banach submanifold of $\velspace$, $\canonicalmap:\pdfspace\diffeomorphismto\velmancan$ is a diffeomorphism, and $\projection$ is a smooth terminal-density-preserving projection.

Appendix~\ref{sub2sec:22} specifies this compatible smooth realization and establishes the global manifold structure. Appendix~\ref{sub2sec:23} then identifies the tangent and cotangent spaces in centered-energy coordinates and states the additional smoothness condition required by the Fisher--Rao metric. Finally, Appendix~\ref{sub2sec:24} pulls the Fisher--Rao metric back to the global energy chart, constructs the flat energy connection and the mixture connection, and proves the fixed-observable characterization used in Section~\ref{subsec:6}.

\subsubsection{Compatible Smooth Structures}\label{sub2sec:22}

We use a global centered log-density coordinate on $\pdfspace$. This choice is in the spirit of exponential-coordinate constructions in information geometry, while the assumptions below are stated directly for the selected density class and need not coincide with a maximal nonparametric statistical manifold \citep{ay2017information}.

\begin{definition}\label{def:7}
    A \textit{compatible smooth realization} of $(\pdfspace,\velspace)$ consists of the following structures.
    \begin{enumerate}
        \item There exists a Banach function space $\mathbb{B}\subset C(\R^d;\R)$ which contains the constant functions and induces an energy space
        \[
            \energyspace:=\Setmod{\mathbb{B}}{\R}=\left\{[f]:=\left\{f+c:c\in\R\right\}:f\in\mathbb{B}\right\},
        \]
        such that the energy space $\energyspace$ contains a non-empty open subset 
        \[
        \mathcal{O}_\energyspace\subset\left\{[f]\in\energyspace:\int_{\R^d}e^{f(x)}\,\dd x<\infty\right\},
        \]
        and the centered-energy map
        \[
        \mathrm{Log}:\pdfspace\to\mathcal{O}_\energyspace,\qquad
        \rho\mapsto [\log\rho]
        \]
        is bijective.
        \item The density class $\pdfspace$ is equipped with the global Banach chart $(\pdfspace,\mathrm{Log})$. Equivalently, this is the unique smooth Banach-manifold structure on $\pdfspace$ for which $\mathrm{Log}:\pdfspace\diffeomorphismto\mathcal{O}_\energyspace$ is a diffeomorphism.
        \item The velocity class $\velspace$ is a closed linear subspace of $\widehat{\velspace}$; hence, $\velspace$ is a Banach space under the restricted norm. We equip $\velspace$ with the induced smooth Banach-manifold structure.%
        \item The terminal-density map $\terminalmap:\velspace\to\pdfspace$ and the canonical-velocity map $\canonicalmap:\pdfspace\to\velspace$ are smooth.
    \end{enumerate}
\end{definition}

In Definition~\ref{def:7}, $\energyspace=\Setmod{\mathbb{B}}{\R}$ is also a Banach space. Let $\norm[\mathbb{B}]{\cdot}$ denote the norm on $\mathbb{B}$. Then, the norm of $\energyspace$ is
\[
\norm[\energyspace]{[f]}:=\inf_{c\in\R}\norm[\mathbb{B}]{f+c}.
\]
Since $\left\{x\mapsto c:c\in\mathbb{R}\right\}$ is a closed one-dimensional subspace of $\mathbb{B}$, the energy space $\energyspace$ is a Banach space under the norm $\norm[\energyspace]{\cdot}$.

Note that for any $c\in\R$,
\[
\int_{\R^d}e^{f(x)}\,\dd x<\infty\quad\Leftrightarrow\quad\int_{\R^d}e^{f(x)+c}\,\dd x=e^c\int_{\R^d}e^{f(x)}\,\dd x<\infty.
\]
Therefore, the open subset $\mathcal{O}_\energyspace$ is well-defined.

We assume the existence of compatible smooth structures as follows and fix them throughout this paper.

\begin{assumption}\label{assume:1}
    The density class $\pdfspace$ and the velocity class $\velspace$ are chosen so that they admit at least one pair of compatible smooth structures in Definition~\ref{def:7}.
\end{assumption}

\begin{proposition}\label{prop:22}
    The canonical projection $\projection$ is smooth. The canonical-velocity map embeds $\pdfspace$ as a split submanifold $\velmancan$ of $\velspace$, i.e.,
    \[
    \canonicalmap:\pdfspace\diffeomorphismto\velmancan\subset\velspace.
    \]
    In particular, $\velmancan$ is a split embedded Banach submanifold of $\velspace$. The inverse of $\mathrm{Log}$ is the normalized exponential map
    \[
    \mathrm{Exp}:=\mathrm{Log}^{-1}:\mathcal{O}_\energyspace\diffeomorphismto\pdfspace,\qquad
    [f]\mapsto\frac{e^f}{\int_{\R^d}e^{f(x)}\,\dd x}.
    \]
\end{proposition}

\begin{proof}
    Since $\canonicalmap$ and $\terminalmap$ are both smooth, the canonical projection $\projection=\canonicalmap\circ\terminalmap:\velspace\to\velspace$ is smooth under the topology of $\velspace$. Since $\projection^2=\projection$, \cite[Lemma 9.46]{meyer2015groupoid} implies that the image of $\projection$, i.e., the canonical manifold $\velmancan$, is a split embedded Banach submanifold of $\velspace$.
    
    Since $\projection\circ\canonicalmap=\canonicalmap$, the corestriction
    $
    \canonicalmap:\pdfspace\to\velmancan
    $
    is smooth. Since $\velmancan$ is an embedded submanifold of $\velspace$ and $\terminalmap:\velspace\to\pdfspace$ is smooth, the restriction
    \(
    \where{\terminalmap}{\velmancan}:\velmancan\to\pdfspace
    \)
    is also smooth. By Theorem~\ref{thm:10},
    \(
    \canonicalmap^{-1}
    =
    \where{\terminalmap}{\velmancan}.
    \)
    Therefore,
    \(
    \canonicalmap:\pdfspace\diffeomorphismto\velmancan
    \)
    is a diffeomorphism.

    For every $[f]\in\mathcal{O}_\energyspace$, denote $\rho=\mathrm{Log}^{-1}([f])\in\pdfspace$. Then, $f=\log\rho+\const$. Since $\rho$ is a normalized positive density, we have $\rho=\frac{e^f}{\int_{\R^d}e^{f(x)}\,\dd x}$, which gives the explicit expression of $\mathrm{Exp}=\mathrm{Log}^{-1}$.
\end{proof}

In Proposition~\ref{prop:22},
    $
        \mathrm{Exp}=\mathrm{Log}^{-1}
    $
    denotes the normalized exponential parametrization associated with the global energy chart. It should not be conflated with the Riemannian exponential map of the density manifold.

From now on, $\pdfspace$ and $\velspace$ carry the smooth structures fixed above, and $\velmancan$ carries its split embedded submanifold structure. We refer to $\pdfspace$ as the density manifold, to $\velspace$ as the velocity space, and to $\velmancan$ as the canonical manifold. The open set $\mathcal O_\energyspace$ is the global energy-coordinate domain, whereas $\energyspace$ is its Banach model space. We suppress the chosen smooth structures from the notation.

For convenience of computation, we apply the following assumption in this paper.

\begin{assumption}\label{assume:2}
    Unless stated otherwise, all functions, curves, and vector fields considered throughout this paper are assumed sufficiently integrable and smooth for the displayed quantities and derivatives to be well defined. Differentiation may be interchanged with integration, and repeated integrals may be reordered.
\end{assumption}

\subsubsection{First-Order Geometry}\label{sub2sec:23}

The global chart $\mathrm{Log}:\pdfspace\diffeomorphismto\mathcal{O}_\energyspace$ reduces the first-order geometry of $\pdfspace$ to differentiation in the Banach energy space $\energyspace$. We first differentiate the inverse chart $\mathrm{Exp}:\mathcal{O}_\energyspace\diffeomorphismto\pdfspace$, which converts additive energy perturbations into centered density perturbations and thereby identifies the tangent and cotangent spaces of $\pdfspace$. We then specialize this identification to the reverse-KL log-density residual and transport the resulting first-order structures to $\velmancan$ through $\canonicalmap$. Finally, we impose the additional smoothness condition required by the Fisher--Rao metric in Section~\ref{sub2sec:6}.

The differential of $\mathrm{Exp}$ involves expectations of representatives in $\mathbb{B}$. We therefore impose the following integrability condition.

\begin{assumption}\label{assume:3}
    Assume that for every $f\in\mathbb{B}$, and $\rho\in\pdfspace$, we have $\E_{\rho}[\abs{f}]<\infty$.
\end{assumption}

\begin{proposition}\label{prop:23}
    Since $\mathcal{O}_\energyspace$ is an open subset of the Banach space $\energyspace$, we use the canonical identification
    \[
        T_{[f]}\mathcal{O}_\energyspace\cong\energyspace,
        \qquad
        [f]\in\mathcal{O}_\energyspace.
    \]
    Consequently, every smooth tangent vector field
    $E\in\mathfrak{X}(\mathcal{O}_\energyspace)$ can be identified
    with a smooth map
    \[
        E:\mathcal{O}_\energyspace\to\energyspace.
    \]

    Let $\rho=\mathrm{Exp}([f])\in\pdfspace$. Then, for every $[h]\in T_{[f]}\energyspace\cong\energyspace$, the differential of $\mathrm{Exp}$ has the density-function representation
    \[
        \dd_{[f]}\mathrm{Exp}[[h]]
        =
        \rho\left(h-\E_\rho[h]\right).
    \]
\end{proposition}

\begin{proof}
    Since $\mathcal{O}_\energyspace$ is an open set of the Banach space $\energyspace$, we have $T_{[f]}\mathcal{O}_\energyspace\cong\energyspace$, and every smooth tangent vector field can be identified with a smooth map.

    Consider a smooth curve $([f^\varepsilon])_{\varepsilon\in(-\delta,\delta)}$ passing through $[f]$ at $\varepsilon=0$. Let $\rho^\varepsilon=\mathrm{Exp}([f^\varepsilon])$ and $h=\where{\frac{\dd}{\dd\varepsilon}f^\varepsilon}{\varepsilon=0}$, with $[h]\in T_{[f]}\mathcal{O}_\energyspace$. By Assumption~\ref{assume:2}, differentiation may be interchanged with integration, and hence
    \[
    \dd_{[f]}\mathrm{Exp}[h]=\where{\frac{\dd}{\dd\varepsilon}\rho^\varepsilon}{\varepsilon=0}=\where{\frac{\dd}{\dd\varepsilon}\frac{e^{f^\varepsilon}}{\int_{\R^d}e^{f^\varepsilon(x)}\,\dd x}}{\varepsilon=0}=\rho\left(h-\E_\rho[h]\right),
    \]
    which finishes the proof.
\end{proof}

The differential formula above converts an additive energy perturbation into its density-function representation. Consequently, the global energy chart $\mathrm{Log}$ identifies every density tangent space with the common Banach model space $\energyspace$, and every density cotangent space with its continuous dual $\energyspace^*$.

\begin{theorem}\label{thm:11}
    For any density $\rho\in\pdfspace$, the tangent space of the density manifold $\pdfspace$ at $\rho$ is
    \[
    T_\rho\pdfspace=\left\{\rho\cdot\left(f-\E_\rho[f]\right):[f]\in\energyspace\right\}\cong\energyspace.
    \]
   We can also relax the quotient and write
    \[
    T_\rho\pdfspace=\left\{\rho\cdot\left(f-\E_\rho[f]\right):f\in\mathbb{B}\right\}\cong\Setmod{\mathbb{B}}{\R}\cong\energyspace.
    \]
    The cotangent space of $\pdfspace$ at $\rho$ is
    \[
    T_\rho^*\pdfspace=\left\{\ell\circ \dd_\rho\mathrm{Log}:\ell\in\energyspace^*\right\}\cong\energyspace^*.
    \]
\end{theorem}

\begin{proof}
    By Proposition~\ref{prop:23}, we have 
    \[T_\rho\pdfspace=\dd_{\mathrm{Log}(\rho)}\mathrm{Exp}(T_{\mathrm{Log}(\rho)}\energyspace)=\left\{\rho\cdot\left(f-\E_\rho[f]\right):[f]\in\energyspace\right\}.\]
    Evidently, we have
    \[
    \left\{\rho\cdot\left(f-\E_\rho[f]\right):[f]\in\energyspace\right\}=\left\{\rho\cdot\left(f-\E_\rho[f]\right):f\in\mathbb{B}\right\}.
    \]
    The expression for $T_\rho^*\pdfspace$ is straightforward.
\end{proof}

Theorem~\ref{thm:11} provides a common Banach model for all density tangent and cotangent spaces. We next record the two consequences used directly by Newton Matching.

\begin{corollary}\label{cor:6}
    For any $\rho,\pi\in\pdfspace$, we have
    \[
    \left[\log\frac{\rho}{\pi}\right]\in\energyspace,\qquad\rho\cdot\left(\log\frac{\rho}{\pi}-\KL{\rho}{\pi}\right)\in T_\rho\pdfspace.
    \]
\end{corollary}

\begin{proof}
    Note that $\left[\log\frac{\rho}{\pi}\right]=\mathrm{Log}(\rho)-\mathrm{Log}(\pi)\in\energyspace$. According to Theorem~\ref{thm:11}, we have $\rho\cdot\left(\log\frac{\rho}{\pi}-\KL{\rho}{\pi}\right)\in T_\rho\pdfspace$.
\end{proof}

Corollary~\ref{cor:6} shows that the centered reverse-KL log-density residual defines an admissible tangent vector on $\pdfspace$. Newton Matching, however, performs its updates in velocity-field coordinates. We therefore differentiate the canonical diffeomorphism to transport the tangent and cotangent structures from $\pdfspace$ to the canonical manifold $\velmancan$.

\begin{corollary}\label{cor:7}
    Let $v^\rho=\canonicalmap(\rho)\in\velmancan$. Then
    \[
    \dd_\rho\canonicalmap:T_\rho\pdfspace\diffeomorphismto T_{v^\rho}\velmancan
    \]
    is a Banach-space isomorphism, and
    \[
    T_{v^\rho}\velmancan=\dd_\rho\canonicalmap(T_\rho\pdfspace).
    \]
    Dually,
    \[
    \left((\dd_\rho\canonicalmap)^{-1}\right)^*:T_\rho^*\pdfspace\diffeomorphismto T_{v^\rho}^*\velmancan
    \]
    is a Banach-space isomorphism.
\end{corollary}

\begin{proof}
    This follows by differentiating the diffeomorphism $\canonicalmap:\pdfspace\diffeomorphismto\velmancan$ and taking continuous duals.
\end{proof}

The preceding results determine the admissible first-order perturbations on both $\pdfspace$ and $\velmancan$. To represent the relevant objective differentials by tangent update directions whenever their metric duals exist, we additionally equip $\pdfspace$ with the Fisher--Rao metric introduced in Section~\ref{sub2sec:6}. The Banach-manifold structure alone does not guarantee that this bilinear form varies smoothly with the base point and tangent vectors, so we impose the following condition.

\begin{assumption}\label{assume:4}
For every \(\rho\in\pdfspace\), 
\[
g_\rho^\fisher(\xi_1,\xi_2)
=
\int_{\R^d}
\frac{\xi_1(x)\xi_2(x)}{\rho(x)}\,\dd x
\]
defines a finite, continuous, symmetric, and positive-definite bilinear form on
\(T_\rho\pdfspace\). Moreover, the map
\[
g^\fisher:
(\rho,\xi_1,\xi_2)
\mapsto
g_\rho^\fisher(\xi_1,\xi_2),
\]
is smooth.
\end{assumption}

Under Assumption~\ref{assume:4}, the Fisher--Rao metric defines the smooth, possibly weak, Riemannian structure used on $\pdfspace$, and its transport through $\canonicalmap$ defines the corresponding smooth metric on $\velmancan$. This completes the first-order smooth realization of $\pdfspace$ and $\velmancan$.

To proceed to the second-order geometry, we pull back the Fisher--Rao metric to $\mathcal{O}_\energyspace$ by defining at every $[f]\in\mathcal{O}_\energyspace$:
\[
g^{\energyspace}_{[f]}:\energyspace\times\energyspace\to\R,\qquad
([h_1],[h_2])\mapsto g^{\energyspace}_{[f]}([h_1],[h_2]):=g^\fisher_{\mathrm{Exp}([f])}\left(\dd_{[f]}\mathrm{Exp}[h_1],\dd_{[f]}\mathrm{Exp}[h_2]\right).
\]
In particular, denote $\rho=\mathrm{Exp}([f])$, and we have
\[
g^{\energyspace}_{[f]}([h_1],[h_2])=g^\fisher_{\rho}\left(\xi^{\rho,h_1},\xi^{\rho,h_2}\right)=\E_{\rho}\left[(h_1-\E_{\rho}\left[h_1\right])(h_2-\E_{\rho}\left[h_2\right])\right].
\]

The Fisher--Rao flat map on $\mathcal{O}_\energyspace$ at $[f]\in\mathcal{O}_\energyspace$ is defined as
\[
G_{[f]}^\flat:\energyspace\to \energyspace^*,\qquad
G_{[f]}^\flat([h_1])([h_2]):=g^{\energyspace}_{[f]}([h_1],[h_2]).
\]
Due to the positive definiteness of $g^{\energyspace}_{[f]}$, the Fisher--Rao flat map $G_{[f]}^\flat$ is injective. Since $g^{\energyspace}$ is possibly weak, $G_{[f]}^\flat$ need not be surjective.

\subsubsection{Second-Order Geometry}\label{sub2sec:24}

The first-order construction identifies tangent and cotangent vectors at each fixed density and equips the density manifold with the Fisher--Rao metric. To define covariant derivatives and Hessians, we need to additionally compare tangent vector fields as the base point varies, which requires an affine connection.

We first define the energy flat connection and then construct its dual connection with respect to the Fisher--Rao metric, which we call the energy mixture connection. We subsequently transport the latter to $\pdfspace$. This construction is objective-agnostic; the coincidence between the Newton direction and the negative Fisher--Rao gradient for the reverse-KL objective is derived later as a consequence in Section~\ref{subsec:10}.

\begin{proposition}%
    For two smooth tangent vector fields
    $E,H\in\mathfrak{X}(\mathcal{O}_\energyspace)$, define
    \[
        \left(
            \nabla^{\energyspace,\mathrm{flat}}_E H
        \right)_{[f]}
        :=
        \DD_{[f]}H\left[E_{[f]}\right]
        \in
        \energyspace
        \cong
        T_{[f]}\mathcal{O}_\energyspace.
    \]
    Equivalently, consider a smooth curve $([f^\varepsilon])_{\varepsilon\in(-\delta,\delta)}
    \subset\mathcal{O}_\energyspace$ which passes through $[f]$ at $\varepsilon=0$ and satisfies $E_{[f]}=\where{\frac{\dd}{\dd\varepsilon}[f^\varepsilon]}{\varepsilon=0}$. Then, we have
    \[
        \DD_{[f]}H\left[E_{[f]}\right]
        =
        \where{
            \frac{\dd}{\dd\varepsilon}
            H_{[f^\varepsilon]}
        }{\varepsilon=0},
    \]
    where the derivative is taken under the canonical identification
    $
        T_{[f^\varepsilon]}\mathcal{O}_\energyspace
        \cong
        \energyspace
    $.

    Then, the map
    \[
        \nabla^{\energyspace,\mathrm{flat}}
        :
        \mathfrak{X}(\mathcal{O}_\energyspace)
        \times
        \mathfrak{X}(\mathcal{O}_\energyspace)
        \to
        \mathfrak{X}(\mathcal{O}_\energyspace),
        \qquad
        (E,H)\mapsto
        \nabla^{\energyspace,\mathrm{flat}}_E H,
    \]
    is an affine connection on $\mathcal{O}_\energyspace$, which we
    call the energy flat connection.
\end{proposition}

\begin{proof}
    The map $\nabla^{\energyspace,\mathrm{flat}}$ is well-defined and smooth. We now verify the definition of a connection~\eqref{eq:6}. Consider any tangent vector fields $E,F,H\in\mathfrak{X}(\mathcal{O}_\energyspace)$, smooth scalar functions ${h_1},{h_2}:\mathcal{O}_\energyspace\to\R$, and constants $a,b\in\R$. First,
    \[
    \left(\nabla^{\energyspace,\mathrm{flat}}_{{h_1}E+{h_2}F}H\right)_{[f]}
    ={h_1}({[f]})\DD_{[f]} H[E_{[f]}]+{h_2}({[f]})\DD_{[f]} H[F_{[f]}]=\left({h_1}\nabla^{\energyspace,\mathrm{flat}}_EH+{h_2}\nabla^{\energyspace,\mathrm{flat}}_FH\right)_{[f]}
    \]
    gives~\eqref{eq:7}. Then,
    \[
    \left(\nabla^{\energyspace,\mathrm{flat}}_E(aF+bH)\right)_{[f]}
    =a\DD_{[f]} F[E_{[f]}]+b\DD_{[f]} H[E_{[f]}]=\left(a\nabla^{\energyspace,\mathrm{flat}}_EF+b\nabla^{\energyspace,\mathrm{flat}}_EH\right)_{[f]}
    \]
    gives~\eqref{eq:8}. Finally,~\eqref{eq:9} is implied by
    \[
    \left(\nabla^{\energyspace,\mathrm{flat}}_E({h_1}F)\right)_{[f]}
    =\where{\frac{\dd}{\dd\varepsilon}h_1({[f^\varepsilon]})F_{{[f^\varepsilon]}}}{\varepsilon=0}={h_1}({[f]})\left(\nabla^{\energyspace,\mathrm{flat}}_E(F)\right)_{[f]}+E[h_1]({[f]})F_{[f]}.
    \]
    Therefore, $\nabla^{\energyspace,\mathrm{flat}}$ is an affine connection on $\mathcal{O}_\energyspace$.
\end{proof}

The energy flat connection accounts for the variation of tangent vector fields in the fixed Banach model space $\energyspace$. To construct its dual connection with respect to the Fisher--Rao metric, we additionally account for the variation of the metric itself with the base point. The next proposition computes this variation as a trilinear form.

\begin{proposition}
    For every $[f]\in\mathcal{O}_\energyspace$, $\rho:=\mathrm{Exp}([f])$, and $[h_1],[h_2],[h_3]\in T_{[f]}\mathcal{O}_\energyspace\cong \energyspace$, consider the tensor
    \[
    C_{[f]}([h_1],[h_2],[h_3]):=\DD_{[f]}\left(g^\energyspace\left([h_1],[h_2]\right)\right)[[h_3]].
    \]
    Here
    \[
    g^\energyspace\left([h_1],[h_2]\right):[f]\mapsto g_{[f]}^\energyspace\left([h_1],[h_2]\right)
    \]
    is a real-valued map. Then, the map
    \[
    ([f],[h_1],[h_2],[h_3])\mapsto C_{[f]}([h_1],[h_2],[h_3])
    \]
    is smooth. Furthermore, we have
    \[
    C_{[f]}([h_1],[h_2],[h_3])=\E_{\rho}\left[\left(h_1-\E_{\rho}\left[h_1\right]\right)\left(h_2-\E_{\rho}\left[h_2\right]\right)\left(h_3-\E_{\rho}\left[h_3\right]\right)\right].
    \]
\end{proposition}

\begin{proof}
    By Assumption~\ref{assume:4}, the map $([f],[h_1],[h_2])\mapsto g_{[f]}^\energyspace\left([h_1],[h_2]\right)$ is smooth; hence, $([f],[h_1],[h_2],[h_3])\mapsto C_{[f]}([h_1],[h_2],[h_3])=\DD_{[f]}\left(g^\energyspace\left([h_1],[h_2]\right)\right)[[h_3]]$ is smooth. Consider a smooth curve
    $([f^\varepsilon])_{\varepsilon\in(-\delta,\delta)}\subset\mathcal{O}_\energyspace$ which passes through $[f]$ at $\varepsilon=0$ and satisfies $[h_3]=\where{\frac{\dd}{\dd\varepsilon}[f^\varepsilon]}{\varepsilon=0}$. Let $\rho^\varepsilon=\mathrm{Exp}([f^\varepsilon])$. Then, we have
    \begin{align*}
        &\DD_{[f]}\left(g^\energyspace\left([h_1],[h_2]\right)\right)[[h_3]]\\
        =&\where{\frac{\dd}{\dd\varepsilon}\E_{\rho^\varepsilon}\left[\left(h_1-\E_{\rho^\varepsilon}\left[h_1\right]\right)\left(h_2-\E_{\rho^\varepsilon}\left[h_2\right]\right)\right]}{\varepsilon=0}\\
        =&\where{\frac{\dd}{\dd\varepsilon}\E_{\rho}\left[\left(h_1-\E_{\rho^\varepsilon}\left[h_1\right]\right)\left(h_2-\E_{\rho^\varepsilon}\left[h_2\right]\right)\right]}{\varepsilon=0}+\where{\frac{\dd}{\dd\varepsilon}\E_{\rho^\varepsilon}\left[\left(h_1-\E_{\rho}\left[h_1\right]\right)\left(h_2-\E_{\rho}\left[h_2\right]\right)\right]}{\varepsilon=0}\\
        =&0+\E_{\rho}\left[\left(h_1-\E_{\rho}\left[h_1\right]\right)\left(h_2-\E_{\rho}\left[h_2\right]\right)\left(h_3-\E_{\rho}\left[h_3\right]\right)\right],
    \end{align*}
    which finishes the proof.
\end{proof}

The trilinear form $C_{[f]}$ measures the base-point variation of the Fisher--Rao metric when its two tangent arguments are held fixed. For every $[f]\in\mathcal{O}_\energyspace$ and $[h_1],[h_2]\in\energyspace$, it determines the continuous linear functional $C_{[f]}([h_1],[h_2],\cdot)\in\energyspace^*$. To incorporate this metric variation into an affine connection, this functional must admit a tangent-vector representation under the Fisher--Rao metric. Since the metric may be weak, the flat map $G_{[f]}^\flat$ need not be surjective, and the existence of such a representation is therefore an additional second-order requirement.

\begin{assumption}\label{assume:5}
    For every $[f]\in\mathcal{O}_\energyspace$ and $[h_1],[h_2]\in T_{[f]}\mathcal{O}_\energyspace\cong \energyspace$, assume that $$C_{[f]}([h_1],[h_2],\cdot)\in\mathrm{Im}\left(G_{[f]}^\flat\right).$$
    Since $G_{[f]}^\flat$ is injective, there exists a unique $R_{[f]}([h_1],[h_2])\in T_{[f]}\mathcal{O}_\energyspace\cong \energyspace$ such that for every $[h_3]\in T_{[f]}\mathcal{O}_\energyspace\cong \energyspace$,
    \[
    C_{[f]}([h_1],[h_2],[h_3])=g^{\energyspace}_{[f]}\left(R_{[f]}([h_1],[h_2]),[h_3]\right).
    \]
    We additionally assume that the map
    \[
    ([f],[h_1],[h_2])\mapsto R_{[f]}([h_1],[h_2])
    \]
    is smooth.
\end{assumption}

Assumption~\ref{assume:5} supplies the tangent correction that represents the base-point variation of the Fisher--Rao metric. Adding this correction to the energy flat connection gives the candidate connection dual to
$\nabla^{\energyspace,\mathrm{flat}}$. The next proposition verifies that this construction indeed defines an affine connection and then transports it to the density manifold.

\begin{proposition}
    The map
    \[
    \nabla^{\energyspace,\mathrm{mix}}
    :
    \mathfrak{X}(\mathcal{O}_\energyspace)
    \times
    \mathfrak{X}(\mathcal{O}_\energyspace)
    \to
    \mathfrak{X}(\mathcal{O}_\energyspace),
    \qquad
    \left(\nabla^{\energyspace,\mathrm{mix}}_{E}H\right)_{[f]}:=\left(\nabla^{\energyspace,\mathrm{flat}}_{E}H\right)_{[f]}+R_{[f]}\left(E_{[f]},H_{[f]}\right)
    \]
    is an affine connection on $\mathcal{O}_\energyspace$, which we call the energy mixture connection. Then, we transport it to the density manifold $\pdfspace$ through the diffeomorphism $\mathrm{Log}:\pdfspace\diffeomorphismto\mathcal{O}_\energyspace$. Specifically, for $\DensityTangentFieldFirst,\DensityTangentFieldSecond\in\mathfrak{X}(\pdfspace)$, define $E,H\in\mathfrak{X}(\mathcal{O}_\energyspace)$ by
    \[
    E_{[h]}
    :=
    \dd_{\mathrm{Exp}([h])}\mathrm{Log}
    \left[
        \DensityTangentFieldFirst_{\mathrm{Exp}([h])}
    \right],
    \qquad
    H_{[h]}
    :=
    \dd_{\mathrm{Exp}([h])}\mathrm{Log}
    \left[
        \DensityTangentFieldSecond_{\mathrm{Exp}([h])}
    \right].
    \]
    Then, the transported mixture connection on $\pdfspace$ is defined by
    \[
    \nabla^{\mathrm{mix}}
    :
    \mathfrak{X}(\pdfspace)
    \times
    \mathfrak{X}(\pdfspace)
    \to
    \mathfrak{X}(\pdfspace),
    \qquad
    \left(\nabla^{\mathrm{mix}}_{\DensityTangentFieldFirst}\DensityTangentFieldSecond\right)_\rho
    :=
    \dd_{\mathrm{Log}(\rho)}\mathrm{Exp}
    \left[
        \left(
            \nabla^{\energyspace,\mathrm{mix}}_{E}H
        \right)_{\mathrm{Log}(\rho)}
    \right].
    \]
\end{proposition}

\begin{proof}
    By Assumption~\ref{assume:5}, $\nabla^{\energyspace,\mathrm{mix}}$ is well-defined and smooth. We now verify the definition of a connection~\eqref{eq:6}. Consider any energy $[f]\in\mathcal{O}_\energyspace$, tangent vector fields $E,F,H\in\mathfrak{X}(\mathcal{O}_\energyspace)$, smooth scalar functions ${h_1},{h_2}:\mathcal{O}_\energyspace\to\R$, and constants $a,b\in\R$.
    
    First, for every $[h_3]\in\energyspace$,
    \begin{align*}
        &g^{\energyspace}_{[f]}\left(R_{[f]}\left(h_1([f])E_{[f]}+h_2([f])F_{[f]},H_{[f]}\right),[h_3]\right)\\
        =&C_{[f]}(h_1([f])E_{[f]}+h_2([f])F_{[f]},H_{[f]},[h_3])\\
        =&g^{\energyspace}_{[f]}\left(h_1([f])R_{[f]}\left(E_{[f]},H_{[f]}\right)+h_2([f])R_{[f]}\left(F_{[f]},H_{[f]}\right),[h_3]\right).
    \end{align*}
    By the positive definiteness of $g^{\energyspace}_{[f]}$, we have
    \[
    R\left(h_1E+h_2F,H\right)=h_1R\left(E,H\right)+h_2R\left(F,H\right).
    \]
    Then,~\eqref{eq:7} is given by
    \begin{align*}
        \nabla^{\energyspace,\mathrm{mix}}_{{h_1}E+{h_2}F}H
        =&\nabla^{\energyspace,\mathrm{flat}}_{{h_1}E+{h_2}F}H+R\left(h_1E+h_2F,H\right)\\
        =&{h_1}\nabla^{\energyspace,\mathrm{flat}}_EH+{h_2}\nabla^{\energyspace,\mathrm{flat}}_FH+h_1R\left(E,H\right)+h_2R\left(F,H\right)\\
        =&{h_1}\nabla^{\energyspace,\mathrm{mix}}_EH+{h_2}\nabla^{\energyspace,\mathrm{mix}}_FH.
    \end{align*}
    
    Second, for every $[h_3]\in\energyspace$,
    \begin{align*}
        g^{\energyspace}_{[f]}\left(R_{[f]}\left(E_{[f]},aF_{[f]}+bH_{[f]}\right),[h_3]\right)
        =&C_{[f]}(E_{[f]},aF_{[f]}+bH_{[f]},[h_3])\\
        =&g^{\energyspace}_{[f]}\left(aR_{[f]}\left(E_{[f]},F_{[f]}\right)+bR_{[f]}\left(E_{[f]},H_{[f]}\right),[h_3]\right).
    \end{align*}
    By the positive definiteness of $g^{\energyspace}_{[f]}$, we have
    \[
    R\left(E,aF+bH\right)=aR\left(E,F\right)+bR\left(E,H\right).
    \]
    Then,~\eqref{eq:8} is given by
    \begin{align*}
        \nabla^{\energyspace,\mathrm{mix}}_E(aF+bH)
        =&\nabla^{\energyspace,\mathrm{flat}}_E(aF+bH)+R\left(E,aF+bH\right)\\
        =&a\nabla^{\energyspace,\mathrm{flat}}_EF+b\nabla^{\energyspace,\mathrm{flat}}_EH+aR\left(E,F\right)+bR\left(E,H\right)\\
        =&a\nabla^{\energyspace,\mathrm{mix}}_EF+b\nabla^{\energyspace,\mathrm{mix}}_EH.
    \end{align*}

    Finally, applying a similar analysis gives 
    \[
    R(E,h_1F)=h_1R(E,F).
    \]
    Then,~\eqref{eq:9} is given by
    \begin{align*}
        \nabla^{\energyspace,\mathrm{mix}}_E({h_1}F)
        =&\nabla^{\energyspace,\mathrm{flat}}_E({h_1}F)+R(E,h_1F)\\
        =&{h_1}({[f]})\left(\nabla^{\energyspace,\mathrm{flat}}_E(F)\right)_{[f]}+E[h_1]({[f]})F_{[f]}+h_1R(E,F)\\
        =&{h_1}({[f]})\left(\nabla^{\energyspace,\mathrm{mix}}_E(F)\right)_{[f]}+E[h_1]({[f]})F_{[f]}.
    \end{align*}
    Therefore, $\nabla^{\energyspace,\mathrm{mix}}$ is an affine connection on $\mathcal{O}_\energyspace$. Since affine connections are preserved under transport through a diffeomorphism, $\nabla^{\mathrm{mix}}$ is an affine connection on $\pdfspace$.
\end{proof}

The above proposition establishes the connection axioms, but it does not yet identify the metric relation that motivates the correction $R$. We next prove that the energy mixture connection is dual to the energy flat connection with respect to the Fisher--Rao metric. Transporting this duality identity to $\pdfspace$ then yields the fixed-observable characterization used in Section~\ref{subsec:6}.

\begin{theorem}\label{thm:12}
    For every $E,F,H\in\mathfrak{X}(\mathcal{O}_\energyspace)$, the energy flat connection and the energy mixture connection satisfy the following duality identity:
    \[
    E\left[g^\energyspace(F,H)\right]=g^\energyspace\left(\nabla^{\energyspace,\mathrm{flat}}_EF,H\right)+g^\energyspace\left(F,\nabla^{\energyspace,\mathrm{mix}}_EH\right).
    \]
    Consequently, for every $\DensityTangentFieldFirst,\DensityTangentFieldSecond\in\mathfrak{X}(\pdfspace)$ and $\rho\in\pdfspace$, $\left(\nabla_{\DensityTangentFieldFirst}^{\mathrm{mix}}\DensityTangentFieldSecond\right)_\rho$ is the unique tangent vector in $T_\rho\pdfspace$ which satisfies
    \begin{equation}\label{eq:173}
        g_\rho^{\fisher}\left(\left(\nabla_{\DensityTangentFieldFirst}^{\mathrm{mix}}\DensityTangentFieldSecond\right)_\rho,\xi^{\rho,h}\right)=\DensityTangentFieldFirst\left[g^\fisher\left(\DensityTangentFieldSecond,\xi^{\cdot,h}\right)\right](\rho)
    \end{equation}
    for every $[h]\in\energyspace$. Here we consider
    \[
    g^\fisher\left(\DensityTangentFieldSecond,\xi^{\cdot,h}\right):\pdfspace\to\R,\qquad
    q\mapsto g^\fisher_q\left(\DensityTangentFieldSecond_q,\xi^{q,h}\right)
    \]
    as a real-valued function on $\pdfspace$, whereas $\xi^{\cdot,h}\in\mathfrak{X}(\pdfspace)$ is a tangent vector field whose terminal observable is always $h$.
\end{theorem}

\begin{proof}
    For every $[f]\in\mathcal{O}_{\energyspace}$, consider a smooth curve $([f^\varepsilon])_{\varepsilon\in(-\delta,\delta)}\subset\mathcal{O}_\energyspace$ which passes through $[f]$ at $\varepsilon=0$ and satisfies $E_{[f]}=\where{\frac{\dd}{\dd\varepsilon}[f^\varepsilon]}{\varepsilon=0}$. Then, the duality identity is given by
    \begin{align*}
        &E\left[g^\energyspace(F,H)\right]([f])\\
        =&\where{\frac{\dd}{\dd\varepsilon}g_{[f^\varepsilon]}^\energyspace\left(F_{[f^\varepsilon]},H_{[f^\varepsilon]}\right)}{\varepsilon=0}\\
        =&\where{\frac{\dd}{\dd\varepsilon}g_{[f^\varepsilon]}^\energyspace\left(F_{[f]},H_{[f]}\right)}{\varepsilon=0}+\where{\frac{\dd}{\dd\varepsilon}g_{[f]}^\energyspace\left(F_{[f^\varepsilon]},H_{[f]}\right)}{\varepsilon=0}+\where{\frac{\dd}{\dd\varepsilon}g_{[f]}^\energyspace\left(F_{[f]},H_{[f^\varepsilon]}\right)}{\varepsilon=0}\\
        =&C_{[f]}\left(F_{[f]},H_{[f]},E_{[f]}\right)+g_{[f]}^\energyspace\left(\left(\nabla_E^{\energyspace,\mathrm{flat}}F\right)_{[f]},H_{[f]}\right)+g_{[f]}^\energyspace\left(F_{[f]},\left(\nabla_E^{\energyspace,\mathrm{flat}}H\right)_{[f]}\right)\\
        =&g^\energyspace_{[f]}\left(\left(\nabla_E^{\energyspace,\mathrm{flat}}F\right)_{[f]},H_{[f]}\right)+g^\energyspace_{[f]}\left(F_{[f]},\left(\nabla_E^{\energyspace,\mathrm{mix}}H\right)_{[f]}\right).
    \end{align*}
    Consequently, we have
    \begin{align*}
        g^\energyspace_{[f]}\left(F_{[f]},\left(\nabla_E^{\energyspace,\mathrm{mix}}H\right)_{[f]}\right)
        =&E\left[g^\energyspace(F,H)\right]([f])-g^\energyspace_{[f]}\left(\left(\nabla_E^{\energyspace,\mathrm{flat}}F\right)_{[f]},H_{[f]}\right)\\
        =&\where{\frac{\dd}{\dd\varepsilon}g_{[f^\varepsilon]}^\energyspace\left(F_{[f]},H_{[f]}\right)}{\varepsilon=0}+\where{\frac{\dd}{\dd\varepsilon}g_{[f]}^\energyspace\left(F_{[f]},H_{[f^\varepsilon]}\right)}{\varepsilon=0}\\
        =&\where{\frac{\dd}{\dd\varepsilon}g_{[f^\varepsilon]}^\energyspace\left(F_{[f]},H_{[f^\varepsilon]}\right)}{\varepsilon=0}\\
        =&\DD_{[f]}\left(g^\energyspace\left(F_{[f]},H\right)\right)\left[E_{[f]}\right].
    \end{align*}
    Transporting the above equation to $\pdfspace$ implies~\eqref{eq:173}.
\end{proof}

Equation~\eqref{eq:173} also admits a geometric interpretation in density coordinates. For every fixed terminal observable $f$ and tangent vector field $\DensityTangentFieldSecond\in\mathfrak{X}(\pdfspace)$, since $\DensityTangentFieldSecond_q$ has zero total mass for every $q\in\pdfspace$, we have
\[
g_q^\fisher
\left(
    \DensityTangentFieldSecond_q,
    \xi^{q,f}
\right)
=
\int_{\R^d}
f(x)\DensityTangentFieldSecond_q(x)\,\dd x.
\]
Therefore,
along a smooth curve
$(\rho^\varepsilon)_{\varepsilon\in(-\delta,\delta)}
\subset\pdfspace$ which passes through $\rho$ at $\varepsilon=0$,
\[
\where{
    \frac{\dd}{\dd\varepsilon}
    g_{\rho^\varepsilon}^\fisher
    \left(
        \DensityTangentFieldSecond_{\rho^\varepsilon},
        \xi^{\rho^\varepsilon,f}
    \right)
}{\varepsilon=0}
=
\int_{\R^d}
f(x)
\where{
    \frac{\dd}{\dd\varepsilon}
    \DensityTangentFieldSecond_{\rho^\varepsilon}(x)
}{\varepsilon=0}
\dd x.
\]
In this variation, the terminal observable $f$ is held fixed, whereas
the base point $\rho^\varepsilon$, the tangent vector
$\DensityTangentFieldSecond_{\rho^\varepsilon}$, the Fisher--Rao metric
$g_{\rho^\varepsilon}^\fisher$, and the corresponding tangent vector
$\xi^{\rho^\varepsilon,f}$ all vary with $\varepsilon$. Thus, the
pairing compares the tangent vector field $\DensityTangentFieldSecond$ at neighboring densities
against the same terminal observable. The first identity shows that
the explicit density dependence of the Fisher--Rao metric and the
representation $\xi^{q,f}$ cancels in this pairing; the remaining dependence on $q$ is entirely through $\DensityTangentFieldSecond_q$. Consequently,~\eqref{eq:173} characterizes
$\left(\nabla_{\DensityTangentFieldFirst}^\mathrm{mix}\DensityTangentFieldSecond\right)_\rho$ as the unique tangent vector
whose Fisher--Rao pairing with every $\xi^{\rho,f}$ reproduces this
first-order variation along the tangent vector $\DensityTangentFieldFirst_\rho$. In
this sense, the mixture connection differentiates $\DensityTangentFieldSecond$ while keeping
the terminal observable $f$ fixed.

\subsection{Non-Affineness of the Canonical Manifold $\velmancan$}\label{subsec:35}

In this subsection, we show that the canonical manifold, despite being a split embedded Banach submanifold of $\velspace$, need not be affine, which explains why the canonical retraction must apply $\projection$ after ambient addition. Assume that the density class $\pdfspace$ contains the centered Gaussian family and that the function space $\mathbb{B}$ contains all quadratic forms. In this setting, we construct a canonical velocity field and a canonical tangent direction such that every nonzero finite ambient step along that direction leaves $\velmancan$.

Throughout this subsection, we arbitrarily select and fix a nonzero symmetric matrix $A\in\S^d$. Fix the density $\rho=\Normal(0,I)$ and its canonical velocity field $v^\rho=\canonicalmap(\rho)\in\velmancan$. Let $f(x):=\frac{1}{2}x^\top Ax$. According to the assumption, $f\in\mathbb{B}$. Based on Proposition~\ref{prop:2}, the field $\Gamma^{\rho,f}:=(\kappa_t\nabla V_t^\rho[f])_{t\in[0,1]}\in T_{v^\rho}\velmancan$ should be a tangent vector at $v^\rho$. We then prove that for all scalar stepsizes $\eta\neq0$,
\begin{equation}\label{eq:174}
    \bar{v}^\eta\neq\projection(\bar{v}^\eta),\qquad \bar{v}^\eta:=v^\rho+\eta\Gamma^{\rho,f}.
\end{equation}
According to Theorem~\ref{thm:1}, we have $\where{\projection}{\velmancan}=\mathrm{Id}_{\velmancan}$. Therefore, Equation~\eqref{eq:174} implies $\bar{v}^\eta\notin\velmancan$ for all $\eta\neq0$.

For the first step, we compute the canonical velocity field $v^\rho\in\velmancan$. Since $\rho=\Normal(0,I)$, we have
\[
\E_{X_1\sim p_{1|t}^\rho(\cdot|x_t)}\left[X_1\right]=\frac{\alpha_t}{\alpha_t^2+\beta_t^2}x_t,\qquad \E_{X_0\sim p_{0|t}^\rho(\cdot|x_t)}\left[X_0\right]=\frac{\beta_t}{\alpha_t^2+\beta_t^2}x_t.
\]
Therefore,
\[
v_t^\rho(x_t)=\dot\alpha_t\E_{X_1\sim p_{1|t}^\rho(\cdot|x_t)}\left[X_1\right]+\dot\beta_t\E_{X_0\sim p_{0|t}^\rho(\cdot|x_t)}\left[X_0\right]=\frac{\alpha_t\dot{\alpha}_t+\beta_t\dot{\beta}_t}{\alpha_t^2+\beta_t^2}x_t.
\]

For the second step, we compute the tangent vector $\Gamma^{\rho,f}_t=\kappa_t\nabla V_t^\rho[f]$, where $f(x)=\frac{1}{2}x^\top Ax$. Since $p_{1|t}^\rho(\cdot|x_t)=\Normal\left(\frac{\alpha_t}{\alpha_t^2+\beta_t^2}x_t,\frac{\beta_t^2}{\alpha_t^2+\beta_t^2}I\right)$, we have
\[
V_t^\rho[f](x_t)=\E_{X_1\sim p_{1|t}^\rho(\cdot|x_t)}\left[\frac12X_1^\top AX_1\right]=\frac{\alpha_t^2}{2(\alpha_t^2+\beta_t^2)^2}x_t^\top Ax_t+\frac{\beta_t^2}{2(\alpha_t^2+\beta_t^2)}\trace(A).
\]
Since $A\in\S^d$, we have
\[
\Gamma^{\rho,f}_t(x_t)=\kappa_t\nabla V_t^\rho[f](x_t)=\frac{\kappa_t\alpha_t^2}{(\alpha_t^2+\beta_t^2)^2}Ax_t.
\]

For the third step, we compute the updated velocity field $\bar{v}^\eta=v^\rho+\eta\Gamma^{\rho,f}\in\velspace$ and its terminal density $q^\eta:=\terminalmap(\bar{v}^\eta)\in\pdfspace$. Note that
\[
\bar{v}_t^\eta(x_t)=v_t^\rho(x_t)+\eta\Gamma^{\rho,f}_t(x_t)=\frac12\frac{\dd}{\dd t}\left(\log\left(\alpha_t^2+\beta_t^2\right)I+\frac{\eta\alpha_t^2}{\alpha_t^2+\beta_t^2}A\right)x_t.
\]
Denote the ODE flow of $\bar{v}^\eta$ by $\Phi_{0\to t}^{\bar{v}^\eta}$. Solving the ODE gives
\[
\Phi_{0\to t}^{\bar{v}^\eta}(x_0)=\exp\left(\frac12\left(\log\left(\alpha_t^2+\beta_t^2\right)I+\frac{\eta\alpha_t^2}{\alpha_t^2+\beta_t^2}A\right)\right)x_0=\sqrt{\alpha_t^2+\beta_t^2}\exp\left(\frac{\eta\alpha_t^2}{2(\alpha_t^2+\beta_t^2)}A\right)x_0.
\]
Then, the time-$t$ marginal density induced by the ODE flow $\Phi_{0\to t}^{\bar{v}^\eta}$ is
\[
(\Phi_{0\to t}^{\bar{v}^\eta})_\#p_0=\Normal\left(0,\left(\alpha_t^2+\beta_t^2\right)\exp\left(\frac{\eta\alpha_t^2}{\alpha_t^2+\beta_t^2}A\right)\right).
\]
In particular, the terminal density is
\[
q^\eta=(\Phi_{0\to 1}^{\bar{v}^\eta})_\#p_0=\Normal\left(0,\exp\left(\eta A\right)\right).
\]

For the last step, we compare the canonical velocity $v^{q^\eta}:=\canonicalmap(q^\eta)\in\velmancan$ with $\bar{v}^\eta$. For simplicity, we compare their time-$t$ marginal densities. Since
\[
p_t^{q^\eta}=\Normal\left(0,\alpha_t^2 e^{\eta A}+\beta_t^2I\right)\neq (\Phi_{0\to t}^{\bar{v}^\eta})_\#p_0,
\]
we have
\[
\frac{\alpha_t^2 e^{\eta A}+\beta_t^2I}{\alpha_t^2+\beta_t^2}-e^{\frac{\eta\alpha_t^2}{\alpha_t^2+\beta_t^2}A}\in\Setminus{\S_{+}^d}{\{0\}}
\]
for all $\eta\neq0$, $A\neq0$. That means $p_t^{q^\eta}\neq(\Phi_{0\to t}^{\bar{v}^\eta})_\#p_0$ for $t\in(0,1)$; hence,
\[
\bar{v}^\eta\neq v^{q^\eta}=\canonicalmap(q^\eta)=\canonicalmap(\terminalmap(\bar{v}^\eta))=\projection(\bar{v}^\eta).
\]
Therefore, $\bar{v}^\eta\notin\velmancan$.

In summary, once the density class $\pdfspace$ contains the centered Gaussian family $\{\Normal(0,\Sigma):\Sigma\in\S_{++}^d\}$, the canonical manifold $\velmancan$ is non-affine. This establishes the geometric necessity of canonicalization in the canonical retraction of Section~\ref{subsec:7}.

\section{Convergence Proofs for Newton Matching}\label{app:2}

This appendix contains the proofs of the convergence results stated in Section~\ref{sec:4}. Appendix~\ref{subsec:36} proves self-calibration and global KL convergence from the exact terminal-density recursion, whereas Appendix~\ref{subsec:37} uses centered-energy coordinates to prove local quadratic convergence and the continuation result.

\subsection{Proofs of Global Convergence}\label{subsec:36}

This subsection proves the global convergence theorems in Sections~\ref{sub2sec:8} and~\ref{sub2sec:9}. We first record the required Toeplitz lemma \citep{toeplitz1911allgemeine}.

\begin{lemma}[Toeplitz averaging]\label{lem:4}
    Let $\left\{a_i\right\}_{i=0}^\infty\subset\R$ satisfy
    \[
    \lim_{i\to\infty}a_i=a.
    \]
    Let $\left\{w_{k,i}\right\}_{0\leq i<k}\subset\R$ satisfy
    \[
    \sup_{k\geq1}\sum_{i=0}^{k-1}\abs{w_{k,i}}<\infty,\qquad
    \lim_{k\to\infty}\sum_{i=0}^{k-1}w_{k,i}=1,
    \]
    and for every $i$,
    \[
    \lim_{k\to\infty}w_{k,i}=0.
    \]
    Then, we have
    \[
    \lim_{k\to\infty}\sum_{i=0}^{k-1}w_{k,i}a_i=a.
    \]
\end{lemma}

\begin{proof}
    Denote
    \[
    M:=\sup_{k\geq1}\sum_{i=0}^{k-1}\abs{w_{k,i}}<\infty.
    \]
    Then, we have
    \[
    \sum_{i=0}^{k-1}w_{k,i}a_i-a=\sum_{i=0}^{k-1}w_{k,i}(a_i-a)+a\left(\sum_{i=0}^{k-1}w_{k,i}-1\right).
    \]
    It remains to show that both terms on the right-hand side converge to $0$ as $k\to\infty$.

    For any $\varepsilon>0$, there exists a sufficiently large $I_\varepsilon$ such that for every $i\geq I_\varepsilon$, we have $\abs{a_i-a}<\varepsilon$. Therefore, for $k>I_\varepsilon$,
    \begin{align*}
        \abs{\sum_{i=0}^{k-1}w_{k,i}(a_i-a)}
        \leq\sum_{i=0}^{k-1}\abs{w_{k,i}}\abs{a_i-a}
        \leq\sum_{i=0}^{I_\varepsilon-1}\abs{w_{k,i}}\abs{a_i-a}+\varepsilon\sum_{i=I_\varepsilon}^{k-1}\abs{w_{k,i}}.
    \end{align*}
    Taking $k\to\infty$ gives
    \[
    \forall\varepsilon>0,\,\limsup_{k\to\infty}\abs{\sum_{i=0}^{k-1}w_{k,i}(a_i-a)} \leq M\varepsilon.
    \]
    Taking $\varepsilon\to0$ gives
    \[
    \lim_{k\to\infty}\abs{\sum_{i=0}^{k-1}w_{k,i}(a_i-a)}=0.
    \]
    Since $\lim_{k\to\infty}\sum_{i=0}^{k-1}w_{k,i}=1$, we have $\lim_{k\to\infty}\sum_{i=0}^{k-1}w_{k,i}a_i=a$.
\end{proof}

\selfcalibration*

\begin{proof}
    By Theorem~\ref{thm:4}, the sequence $\left\{\KL{\rho_k}{\pi}\right\}_{k=0}^\infty$ is nonincreasing and nonnegative. Therefore, it converges to some $K_\infty\in[0,\KL{\rho_0}{\pi}]$.

    Denote $\lambda_k:=\frac{\eta_k}{\tau}\in(0,1]$. By Proposition~\ref{prop:8}, it holds pointwise that
    \begin{equation}\label{eq:175}
        \log\frac{\rho_k}{\pi}=B_{0,k}\log\frac{\rho_0}{\pi}+\sum_{i=0}^{k-1}B_{i+1,k}\left(\lambda_i\KL{\rho_i}{\pi}-\Diss_{\rho_i,\eta_i,\tilde r^{\rho_i}}\right),
    \end{equation}
    where the coefficient $B_{i,k}$ is defined as
    \[
    B_{i,k}:=\begin{cases}
        \prod_{j=i}^{k-1}(1-\lambda_j),&i<k,\\
        1,&i=k.
    \end{cases}
    \]
    Note that $\Diss_{\rho_k,\eta_k,\tilde r^{\rho_k}}$ is nonnegative for all $k$. Therefore, we have
    \[
        \begin{aligned}
            \E_{X\sim\rho_k}\left[\abs{\log\frac{\rho_k(X)}{\pi(X)}-K_\infty}\right]
            \leq&\abs{\sum_{i=0}^{k-1}B_{i+1,k}\lambda_i\KL{\rho_i}{\pi}-K_\infty}+\E_{X\sim\rho_k}\left[B_{0,k}\abs{\log\frac{\rho_0(X)}{\pi(X)}}\right]\\
            &+\E_{X\sim\rho_k}\left[\sum_{i=0}^{k-1}B_{i+1,k}\Diss_{\rho_i,\eta_i,\tilde r^{\rho_i}}(X)\right].
        \end{aligned}
    \]
    The remaining task is to prove that the three terms on the right-hand side converge to zero as $k\to\infty$.

    For the first step, we verify the conditions of the Toeplitz averaging lemma in Lemma~\ref{lem:4} for $\left\{\KL{\rho_i}{\pi}\right\}_{i=0}^\infty$ and $\left\{B_{i+1,k}\lambda_i\right\}_{0\leq i<k}$. Since $\sum_{i=0}^{\infty}\lambda_i=\infty$, we have 
    \begin{subequations}\label{eq:176}
    \begin{equation}\label{eq:177}
    \lim_{k\to\infty}B_{i,k}=0
    \end{equation}
    for every $i$. Note that $B_{i+1,k}\lambda_i=B_{i+1,k}-B_{i,k}$; hence,
    \begin{equation}\label{eq:178}
        \lim_{k\to\infty}\sum_{i=0}^{k-1}B_{i+1,k}\lambda_i=\lim_{k\to\infty}1-B_{0,k}=1.
    \end{equation}
    Since $\lambda_i>0$, $B_{i+1,k}\geq0$, and $\lim_{k\to\infty}\sum_{i=0}^{k-1}B_{i+1,k}\lambda_i=1$, we have
    \begin{equation}\label{eq:179}
    \sup_{k\geq 1}\sum_{i=0}^{k-1}\abs{B_{i+1,k}\lambda_i}=\sup_{k\geq 1}\sum_{i=0}^{k-1}B_{i+1,k}\lambda_i<\infty.
    \end{equation}
    \end{subequations}
    Since $\lim_{i\to\infty}\KL{\rho_i}{\pi}=K_\infty$, the Toeplitz averaging lemma in Lemma~\ref{lem:4} gives
    \begin{equation*}
    \lim_{k\to\infty}\sum_{i=0}^{k-1}B_{i+1,k}\lambda_i\KL{\rho_i}{\pi}=K_\infty.
    \end{equation*}

    For the second step,~\eqref{eq:175}, $\KL{\rho_i}{\pi}\leq\KL{\rho_0}{\pi}$, and $\Diss_{\rho_i,\eta_i,\tilde r^{\rho_i}}(x)\geq0$ imply
    \[
    \log\frac{\rho_k}{\pi}
    \leq B_{0,k}\log\frac{\rho_0}{\pi}+\KL{\rho_0}{\pi}\sum_{i=0}^{k-1}B_{i+1,k}\lambda_i
    \leq B_{0,k}\log\frac{\rho_0}{\pi}+\KL{\rho_0}{\pi}.
    \]
    Denote the set $A=\left\{x\in\R^d:\rho_0(x)\geq\pi(x)\right\}$. Then, we have
    \[
        \begin{aligned}
            \E_{X\sim\rho_k}\left[B_{0,k}\abs{\log\frac{\rho_0(X)}{\pi(X)}}\right]
            =&\E_{X\sim\pi}\left[\frac{\rho_k(X)}{\pi(X)}B_{0,k}\abs{\log\frac{\rho_0(X)}{\pi(X)}}\right]\\
            \leq&e^{\KL{\rho_0}{\pi}}\E_{X\sim\pi}\left[B_{0,k}\abs{\log\frac{\rho_0(X)}{\pi(X)}}\exp\left(B_{0,k}\log\frac{\rho_0(X)}{\pi(X)}\right)\right]\\
            =&e^{\KL{\rho_0}{\pi}}\int_{A}\pi(x)B_{0,k}\log\frac{\rho_0(x)}{\pi(x)}\exp\left(B_{0,k}\log\frac{\rho_0(x)}{\pi(x)}\right)\,\dd x\\
            &-e^{\KL{\rho_0}{\pi}}\int_{A^\mathrm{c}}\pi(x)B_{0,k}\log\frac{\rho_0(x)}{\pi(x)}\exp\left(B_{0,k}\log\frac{\rho_0(x)}{\pi(x)}\right)\,\dd x
        \end{aligned}
    \]
    By the elementary inequality $xe^{-x}\leq\frac{1}{e}$ for $x\geq0$, we bound the integral over $A$ by
    \[
        \begin{aligned}
            &\int_{A}\pi(x)B_{0,k}\log\frac{\rho_0(x)}{\pi(x)}\exp\left(B_{0,k}\log\frac{\rho_0(x)}{\pi(x)}\right)\,\dd x\\
            =&\frac{B_{0,k}}{1-B_{0,k}}\int_{A}\pi(x)\exp\left(\log\frac{\rho_0(x)}{\pi(x)}\right)\cdot(1-B_{0,k})\log\frac{\rho_0(x)}{\pi(x)}\exp\left(-(1-B_{0,k})\log\frac{\rho_0(x)}{\pi(x)}\right)\,\dd x\\
            \leq&\frac{B_{0,k}}{1-B_{0,k}}\int_{A}\rho_0(x)\frac{1}{e}\,\dd x\\
            \leq&\frac{B_{0,k}}{e(1-B_{0,k})}.
        \end{aligned}
    \]
    Letting $k\to\infty$ gives
    \[
        \int_{A}\pi(x)B_{0,k}\log\frac{\rho_0(x)}{\pi(x)}\exp\left(B_{0,k}\log\frac{\rho_0(x)}{\pi(x)}\right)\,\dd x\to0.
    \]
    Next, we bound the integral over $A^\mathrm{c}=\left\{x\in\R^d:\rho_0(x)<\pi(x)\right\}$. By the bound
    \[
    B_{0,k}\abs{\log\frac{\rho_0(x)}{\pi(x)}}\exp\left(B_{0,k}\log\frac{\rho_0(x)}{\pi(x)}\right)\leq\frac{1}{e}
    \]
    and the pointwise convergence
    \[
    B_{0,k}\abs{\log\frac{\rho_0(x)}{\pi(x)}}\exp\left(B_{0,k}\log\frac{\rho_0(x)}{\pi(x)}\right)\to 0,\,\forall x\in A^\mathrm{c},
    \]
    the dominated convergence theorem gives
    \[
        \int_{A^\mathrm{c}}\pi(x)B_{0,k}\abs{\log\frac{\rho_0(x)}{\pi(x)}}\exp\left(B_{0,k}\log\frac{\rho_0(x)}{\pi(x)}\right)\,\dd x\to0.
    \]
    Therefore, we have
    \[
    \lim_{k\to\infty}\E_{X\sim\rho_k}\left[B_{0,k}\abs{\log\frac{\rho_0(X)}{\pi(X)}}\right]=0.
    \]
    
    For the third step, we have
    \[
        \begin{aligned}
            &\E_{X\sim\rho_k}\left[\sum_{i=0}^{k-1}B_{i+1,k}\Diss_{\rho_i,\eta_i,\tilde r^{\rho_i}}(X)\right]\\
            =&\E_{X\sim\rho_k}\left[B_{0,k}\log\frac{\rho_0(X)}{\pi(X)}-\log\frac{\rho_k(X)}{\pi(X)}+\sum_{i=0}^{k-1}B_{i+1,k}\lambda_i\KL{\rho_i}{\pi}\right]\\
            =&\E_{X\sim\rho_k}\left[B_{0,k}\log\frac{\rho_0(X)}{\pi(X)}\right]-\KL{\rho_k}{\pi}+\sum_{i=0}^{k-1}B_{i+1,k}\lambda_i\KL{\rho_i}{\pi}.
        \end{aligned}
    \]
    Based on the conclusion in the first two steps, we have
    \[
    \limsup_{k\to\infty}\E_{X\sim\rho_k}\left[\sum_{i=0}^{k-1}B_{i+1,k}\Diss_{\rho_i,\eta_i,\tilde r^{\rho_i}}(X)\right]\leq 0-K_\infty+K_\infty=0.
    \]
    
    Together, these three steps prove~\eqref{eq:49}.
\end{proof}

We next prove the two global-convergence theorems. We first record the following lemma.

\begin{lemma}\label{lem:5}
    Consider two positive normalized densities $\rho,\pi$ on $\R^d$. Let $A\subset\R^d$ be measurable. Both $A$ and $A^\mathrm{c}$ are of positive measure. Then, we have
    \[
        \KL{\pi}{\rho}\geq \pi(A^\mathrm{c})\log\frac{\pi(A^\mathrm{c})}{\rho(A^\mathrm{c})}+\pi(A)\log\frac{\pi(A)}{\rho(A)}.
    \]
\end{lemma}

\begin{proof}
    Let $\rho_A:=\frac{\rho}{\rho(A)}$ and $\pi_A:=\frac{\pi}{\pi(A)}$ be the conditional densities on $A$. Then, we have
    \begin{align*}
        \int_{A}\pi(x)\log\frac{\pi(x)}{\rho(x)}\,\dd x
        =&\pi(A)\int_{A}\pi_A(x)\log\frac{\pi_A(x)\pi(A)}{\rho_A(x)\rho(A)}\,\dd x
        =\pi(A)\left(\log\frac{\pi(A)}{\rho(A)}+\KL{\pi_A}{\rho_A}\right).
    \end{align*}
    Thus,
    \[
    \int_{A}\pi(x)\log\frac{\pi(x)}{\rho(x)}\,\dd x\geq \pi(A)\log\frac{\pi(A)}{\rho(A)}.
    \]
    For the same reason,
    \[
    \int_{A^\mathrm{c}}\pi(x)\log\frac{\pi(x)}{\rho(x)}\,\dd x\geq \pi(A^\mathrm{c})\log\frac{\pi(A^\mathrm{c})}{\rho(A^\mathrm{c})}.
    \]
    Therefore,
    \[
    \KL{\pi}{\rho}=\int_{A}\pi(x)\log\frac{\pi(x)}{\rho(x)}\,\dd x+\int_{A^\mathrm{c}}\pi(x)\log\frac{\pi(x)}{\rho(x)}\,\dd x
    \geq \pi(A)\log\frac{\pi(A)}{\rho(A)}+\pi(A^\mathrm{c})\log\frac{\pi(A^\mathrm{c})}{\rho(A^\mathrm{c})},
    \]
    which finishes the proof.
\end{proof}

\klconvergenceanticollapse*

\begin{proof}
    Assume that~\eqref{eq:50} holds but $K_\infty>0$, where $K_\infty$ is defined in~\eqref{eq:48}. We consider the set $A_k:=\left\{x\in\R^d:\abs{\log\frac{\rho_k(x)}{\pi(x)}-K_\infty}\leq\frac{K_\infty}{2}\right\}$. According to Lemma~\ref{lem:5}, we have
    \begin{equation*}
        \KL{\pi}{\rho_k}\geq \pi(A_k^\mathrm{c})\log\frac{\pi(A_k^\mathrm{c})}{\rho_k(A_k^\mathrm{c})}+\pi(A_k)\log\frac{\pi(A_k)}{\rho_k(A_k)}.
    \end{equation*}
    Applying the inequality $x\log x\geq-\frac{1}{e}$ for $x>0$ gives
    \[
        \KL{\pi}{\rho_k}\geq -\pi(A_k^\mathrm{c})\log\rho_k(A_k^\mathrm{c})-\pi(A_k)\log\rho_k(A_k)-\frac{2}{e}.
    \]
    We can bound $\pi(A_k)$ by
    \[
        \pi(A_k)=\int_{A_k}\pi(x)\,\dd x
        \leq \int_{A_k}\rho_k(x)e^{-K_\infty/2}\,\dd x
        \leq e^{-K_\infty/2}<1.
    \]
    Therefore,
    \[
        \KL{\pi}{\rho_k}\geq -(1-e^{-K_\infty/2})\log\rho_k(A_k^\mathrm{c})-\frac{2}{e}.
    \]
    Note that
    \[
        \int_{A_k^\mathrm{c}}\rho_k(x)\,\dd x
        \leq\int_{A_k^\mathrm{c}}\rho_k(x)\frac{2}{K_\infty}\abs{\log\frac{\rho_k(x)}{\pi(x)}-K_\infty}\,\dd x
        \leq\frac{2}{K_\infty}\E_{X\sim\rho_k}\left[\abs{\log\frac{\rho_k(X)}{\pi(X)}-K_\infty}\right].
    \]
    By Theorem~\ref{thm:5}, we have 
    \[
    \lim_{k\to\infty}\rho_k(A_k^\mathrm{c})=0.
    \]
    Therefore, as $k\to\infty$,
    \[
        \KL{\pi}{\rho_k}\geq -(1-e^{-K_\infty/2})\log\rho_k(A_k^\mathrm{c})-\frac{2}{e}\to+\infty,
    \]
    which leads to a contradiction with~\eqref{eq:50}.
\end{proof}

\klconvergencedissipationvanishing*

\begin{proof}
    Denote $\Diss_i:=\frac{1}{\eta_i^2}\Diss_{\rho_i,\eta_i,\tilde r^{\rho_i}}$. Taking the expectation of~\eqref{eq:175} over $\pi$ gives
    \[
        \KL{\pi}{\rho_k}=B_{0,k}\KL{\pi}{\rho_0}-\sum_{i=0}^{k-1}B_{i+1,k}\lambda_i\left(\KL{\rho_i}{\pi}-\tau\eta_i\E_{X\sim\pi}\left[\Diss_i(X)\right]\right).
    \]
    By~\eqref{eq:48} and condition~\eqref{eq:51}, we have
    \[
    \lim_{i\to\infty}\KL{\rho_i}{\pi}-\tau\eta_i\E_{X\sim\pi}\left[\Diss_i(X)\right]=K_\infty.
    \]
    By~\eqref{eq:176}, the Toeplitz averaging lemma in Lemma~\ref{lem:4} gives
    \[
    \lim_{k\to\infty}\sum_{i=0}^{k-1}B_{i+1,k}\lambda_i\left(\KL{\rho_i}{\pi}-\tau\eta_i\E_{X\sim\pi}\left[\Diss_i(X)\right]\right)=K_\infty.
    \]
    Therefore, we have
    \[
    \lim_{k\to\infty}\KL{\pi}{\rho_k}=-K_\infty\geq0;
    \]
    hence, $K_\infty=0$.
\end{proof}

\subsection{Proofs of Local Convergence}\label{subsec:37}

The global-convergence proofs above use the exact terminal-density recursion and asymptotic averaging. The local analysis instead represents the full-step update in centered-energy coordinates, where Newton cancellation appears as a vanishing first derivative at the target. Appendix~\ref{sub2sec:25} proves the resulting quadratic estimate, and Appendix~\ref{sub2sec:26} makes this estimate uniform along the inverse-temperature path used by the continuation method.

\subsubsection{Local Quadratic Convergence}\label{sub2sec:25}

We adopt the notation in Section~\ref{sub2sec:10}. Taking the full stepsize $\eta_{k}=\tau$ in~\eqref{eq:43} gives
\begin{equation}\label{eq:180}
    \log\frac{\rho_{k+1}}{\pi}
    =\KL{\rho_k}{\pi}-\Diss_{\rho_k,1,\log(\pi/\rho_k)},
\end{equation}
where
\begin{equation*}
    \Diss_{\rho,1,\log(\pi/\rho)}(x)
    =\Diss_{\rho,\tau,\tilde r^\rho}(x)
    =
    \int_0^1
    \kappa_t
    \left\|
    \nabla
    V_t^\rho\left[\log\frac{\pi}{\rho}\right]
    \left(
    \flow_{1\to t}^{\rho,1,\log(\pi/\rho)}(x)
    \right)
    \right\|_2^2
    \dd t
\end{equation*}
is independent of $\tau$ for a fixed $\pi$.

Recall the centered-energy map $\mathrm{Log}:\pdfspace\to\mathcal{O}_\energyspace\subset\energyspace$ and the equivalence class in Definition~\ref{def:7}. We have
\[
    \left[\log\frac{\rho}{\pi}\right]
    =\mathrm{Log}(\rho)-\mathrm{Log}(\pi)
    \in\energyspace.
\]
Taking equivalence classes in~\eqref{eq:180} removes the constant term $\KL{\rho_k}{\pi}$ and gives
\[
    \left[\log\frac{\rho_{k+1}}{\pi}\right]
    =-\left[\Diss_{\rho_k,1,\log(\pi/\rho_k)}\right].
\]
We therefore represent the full-step iteration by the centered-energy update map
\[
    \left[\log\frac{\rho_{k+1}}{\pi}\right]
    =\iteration_{\pi}\left(\left[\log\frac{\rho_{k}}{\pi}\right]\right)
    :=-\left[\Diss_{\rho_k,1,\log(\pi/\rho_k)}\right].
\]
The following proposition provides its smooth local realization and the full-step Newton cancellation, which contributes to the local quadratic convergence.

\begin{proposition}\label{prop:24}
    We define
    \[
    \chi_\pi:\pdfspace\to\energyspace,\qquad \rho\mapsto\left[\log\frac{\rho}{\pi}\right]=\mathrm{Log}(\rho)-\mathrm{Log}(\pi).
    \]
    Then, $\chi_\pi$ is a diffeomorphism onto the open set
    \[
        \mathcal{O}_\pi:=\mathrm{Im}(\chi_\pi)=\mathcal{O}_\energyspace-\mathrm{Log}(\pi)\subset\energyspace,
    \]
    and its inverse
    \[
    \chi_\pi^{-1}:\mathcal{O}_\pi\to\pdfspace,\qquad \left(\chi_\pi^{-1}([f])\right)(x):=\frac{\pi(x) e^{f(x)}}{\E_{X\sim\pi}\left[e^{f(X)}\right]}
    \]
    is well-defined. The map
    \[
    \widehat{\iteration}_\pi:\pdfspace\to\pdfspace,\qquad \rho\mapsto\terminalmap\left(\canonicalmap(\rho)+\Gamma^{\rho,\log(\pi/\rho)}\right)
    \]
    is well-defined and smooth. Moreover,
    \[
    \iteration_\pi:=\chi_\pi\circ\widehat{\iteration}_\pi\circ\chi_\pi^{-1}:\mathcal{O}_\pi\to\mathcal{O}_\pi
    \]
    is smooth and satisfies
    \begin{equation}\label{eq:181}
        \iteration_{\pi}(0)=0,\qquad \DD_0 \iteration_{\pi}=0.
    \end{equation}
\end{proposition}

\begin{proof}
    Since $\mathrm{Log}:\pdfspace\diffeomorphismto\mathcal{O}_\energyspace$ is a diffeomorphism,
    $
        \chi_\pi(\rho)=\mathrm{Log}(\rho)-\mathrm{Log}(\pi)
    $
    is a diffeomorphism from $\pdfspace$ onto the open set $\mathcal{O}_\energyspace-\mathrm{Log}(\pi)$. Its inverse is
    \[
        \chi_\pi^{-1}([f])=\mathrm{Exp}\left(\mathrm{Log}(\pi)+[f]\right),\qquad \left(\chi_\pi^{-1}([f])\right)(x)
        =\frac{\pi(x)e^{f(x)}}{\E_{X\sim\pi}\left[e^{f(X)}\right]},
    \]
    which is independent of the selected representative of $[f]$.

    Consider $[f]\in\mathcal{O}_\energyspace-\mathrm{Log}(\pi)$ and denote
    $
        \rho=\chi_\pi^{-1}([f])
    $. Since $[f]=-\left[\log\frac{\pi}{\rho}\right]$, Proposition~\ref{prop:23} gives
    \[
        \dd_{\mathrm{Log}(\pi)+[f]}\mathrm{Exp}[-[f]]
        =\rho\left(\log\frac{\pi}{\rho}-\E_\rho\left[\log\frac{\pi}{\rho}\right]\right).
    \]
    Therefore, by Proposition~\ref{prop:2},
    \[
        \DD_{\mathrm{Log}(\pi)+[f]}\left(\canonicalmap\circ\mathrm{Exp}\right)[-[f]]
        =\dd_\rho\canonicalmap\left[\dd_{\mathrm{Log}(\pi)+[f]}\mathrm{Exp}[-[f]]\right]
        =\Gamma^{\rho,\log(\pi/\rho)}.
    \]

    Hence, the ambient full-step update in the energy coordinates is
    \[
        \left(\canonicalmap\circ\mathrm{Exp}\right)\left(\mathrm{Log}(\pi)+[f]\right)
        +\DD_{\mathrm{Log}(\pi)+[f]}\left(\canonicalmap\circ\mathrm{Exp}\right)[-[f]].
    \]
    This expression depends smoothly on $[f]$. At $[f]=0$, it is equal to $\canonicalmap(\pi)\in\velspace$. Since $\velspace$ is a Banach space, $\widehat{\iteration}_\pi$ is well-defined for every $[f]\in\mathcal{O}_\pi$. Moreover,
    \[
        \iteration_\pi([f])
        =\mathrm{Log}\circ\terminalmap\left(
        \left(\canonicalmap\circ\mathrm{Exp}\right)\left(\mathrm{Log}(\pi)+[f]\right)
        +\DD_{\mathrm{Log}(\pi)+[f]}\left(\canonicalmap\circ\mathrm{Exp}\right)[-[f]]
        \right)
        -\mathrm{Log}(\pi).
    \]
    Therefore, $\widehat{\iteration}_\pi$ and $\iteration_\pi$ are smooth.

    At $[f]=0$, we have
    \[
        \iteration_\pi(0)
        =\mathrm{Log}\circ\terminalmap\left(\canonicalmap(\pi)\right)-\mathrm{Log}(\pi)
        =0.
    \]
    For any $h\in\energyspace$, differentiating the ambient update at $[f]=0$ gives
    \[
    \begin{aligned}
        &\DD_{\mathrm{Log}(\pi)}\left(\canonicalmap\circ\mathrm{Exp}\right)[h]
        +\DD^2_{\mathrm{Log}(\pi)}\left(\canonicalmap\circ\mathrm{Exp}\right)[h,0]
        -\DD_{\mathrm{Log}(\pi)}\left(\canonicalmap\circ\mathrm{Exp}\right)[h]
        =0.
    \end{aligned}
    \]
    Applying the chain rule to the coordinate expression of $\iteration_\pi$ gives $\DD_0\iteration_\pi[h]=0$. Therefore, we have
    $
        \DD_0\iteration_\pi=0
    $.
\end{proof}

To express the centered-energy estimate intrinsically on the density and canonical manifolds, define
\[
    d_{\pdfspace}(\rho_1,\rho_2):=\norm[\energyspace]{\mathrm{Log}(\rho_1)-\mathrm{Log}(\rho_2)},
\]
which is the pullback under the global energy chart $\mathrm{Log}:\pdfspace\to\mathcal{O}_{\energyspace}$ of the metric on $\mathcal{O}_{\energyspace}$ induced by the norm $\norm[\energyspace]{\cdot}$. The following proposition verifies that $d_{\pdfspace}$ is a metric on $\pdfspace$ and that the transported metric $d_{\velmancan}$ in~\eqref{eq:52} is a metric on $\velmancan$.

\restateproposition{\metricondensitymanifold}

\begin{proof}
    The map
    \[
    \energyspace\times\energyspace\to[0,\infty),\qquad
    ([f_1],[f_2])\mapsto\norm[\energyspace]{[f_1-f_2]}
    \]
    is a metric on $\energyspace$. Since $\mathrm{Log}$ is a diffeomorphism between $\pdfspace$ and the open subset $\mathcal{O}_\energyspace\subset\energyspace$, $d_{\pdfspace}$ is a metric on $\pdfspace$. As the pullback of the norm metric on $\mathcal{O}_\energyspace$, $d_\pdfspace$ induces exactly the Banach-manifold topology on $\pdfspace$. Consequently, since $\where{\terminalmap}{\velmancan}$ is a diffeomorphism between $\velmancan$ and $\pdfspace$, $d_{\velmancan}$ is a metric on $\velmancan$ and induces exactly the Banach-manifold topology on $\velmancan$.
\end{proof}

With the centered-energy update and the induced metrics in place, local quadratic convergence follows from a second-order Taylor expansion of $\iteration_\pi$ at the fixed point $0\in\energyspace$.

\localquadraticconvergence*

\begin{proof}
    Since $\iteration_{\pi}$ is $C^2$ on the open neighborhood $\mathcal{O}_\pi$ of $0\in\energyspace$, there exist a constant $C_\pi>0$ and a sufficiently small radius $R_\pi>0$ such that, for any $\hat\rho\in\pdfspace$ satisfying $d_{\pdfspace}(\hat\rho,\pi)=\norm[\energyspace]{\left[\log\frac{\hat\rho}{\pi}\right]}<R_\pi$, we have 
    \[
        \left[\log\frac{\hat{\rho}}{\pi}\right]\in\mathcal{O}_\pi,\qquad \norm[\mathrm{op}]{\DD_{\left[\log\frac{\hat\rho}{\pi}\right]}^2\iteration_{\pi}}\leq 2C_\pi.
    \]
    Here, the norm $\norm[\mathrm{op}]{\cdot}$ is the operator norm with respect to $\norm[\energyspace]{\cdot}$. According to Proposition~\ref{prop:24}, taking the Taylor expansion at $0$ gives
    \[
        \iteration_{\pi}\left(\left[\log\frac{\rho}{\pi}\right]\right)=\int_{0}^{1}(1-\zeta)\DD_{\left[\zeta\log\frac{\rho}{\pi}\right]}^2\iteration_{\pi}\left[\left[\log\frac{\rho}{\pi}\right],\left[\log\frac{\rho}{\pi}\right]\right]\dd\zeta.
    \]
    If $\norm[\energyspace]{\left[\log\frac{\rho}{\pi}\right]}<R_\pi$, then we have $\norm[\energyspace]{\left[\zeta\log\frac{\rho}{\pi}\right]}<R_\pi$ for all $\zeta\in[0,1]$. By~\eqref{eq:181},
    \[
        \norm[\energyspace]{\iteration_{\pi}\left(\left[\log\frac{\rho}{\pi}\right]\right)}\leq \int_{0}^{1}(1-\zeta)\norm[\mathrm{op}]{\DD_{\left[\zeta\log\frac{\rho}{\pi}\right]}^2\iteration_{\pi}}\norm[\energyspace]{\left[\log\frac{\rho}{\pi}\right]}^2\dd\zeta \leq C_\pi\norm[\energyspace]{\left[\log\frac{\rho}{\pi}\right]}^2,
    \]
    which gives~\eqref{eq:53}.

    Consequently, suppose that the initial point $\rho_0$ satisfies $d_{\pdfspace}(\rho_0,\pi)<\min\left\{R_\pi,\frac{1}{C_\pi}\right\}$. By~\eqref{eq:53}, we have
    \[
        d_{\pdfspace}(\rho_1,\pi)\leq C_\pi d_{\pdfspace}(\rho_0,\pi)^2\leq d_{\pdfspace}(\rho_0,\pi)<\min\left\{R_\pi,\frac{1}{C_\pi}\right\}.
    \]
    By induction,~\eqref{eq:54} holds for all $k$. Therefore, $\{\rho_k\}_{k=0}^\infty$ converges quadratically to $\pi$ under the metric $d_{\pdfspace}$, which also gives the local quadratic convergence of the canonical velocity fields.
\end{proof}

\subsubsection{Proof of Continuation Method}\label{sub2sec:26}

We adopt the notation in Section~\ref{sub2sec:11}. The local estimate above has constants that may depend on the fixed target. To obtain uniform constants over the compact family $\{\pi_s:s\in[\tau_0,\tau]\}$, we first establish joint smoothness of the centered-energy update in $(s,[f])$.

\begin{proposition}\label{prop:25}
    If $\pi_s:=\pi_{\mu,s,r}\in\pdfspace$ for all $s\in[\tau_0,\tau]$, then the map
    \[
        (s,[f])\mapsto\iteration_{\pi_s}([f])
    \]
    is smooth on an open neighborhood of $[\tau_0,\tau]\times\{0\}$ in $\R\times\energyspace$.
\end{proposition}

\begin{proof}
    Since
    $
        \mathrm{Log}(\pi_s)-\mathrm{Log}(\pi_{\tau_0})=(s-\tau_0)[r]\in\energyspace
    $, we have $[r]\in\energyspace$. Therefore, $(s,[f])\mapsto\iteration_{\pi_s}([f])$ is smooth wherever it is defined. Note that
    \[
    \left\{(s,[f]):\mathrm{Log}(\pi_s)+[f]\in\mathcal{O}_\energyspace\right\}\supset[\tau_0,\tau]\times\{0\}
    \]
    is open; hence, the map is smooth on an open neighborhood of $[\tau_0,\tau]\times\{0\}$.
\end{proof}

Joint smoothness and compactness now yield a common coordinate neighborhood and a uniform second-derivative bound over the entire inverse-temperature interval $[\tau_0,\tau]$.

\continuationmethod*

\begin{proof}
    Let
    \[
    C:=\sup_{s\in[\tau_0,\tau]}\norm[\mathrm{op}]{\DD_0 ^2\iteration_{\pi_{s}}}+1<\infty,\qquad M:=\norm[\energyspace]{\left[r\right]}<\infty.
    \]
    Since the domain of $(s,[f])\mapsto\iteration_{\pi_s}([f])$ is an open neighborhood of the compact set $[\tau_0,\tau]\times\{0\}$, there exists $R>0$ such that
    \[
        s\in[\tau_0,\tau],\,\norm[\energyspace]{[f]}<R
        \,\Rightarrow\,
        [f]\in\mathcal{O}_{\pi_s}.
    \]

    For the first step, we prove that, after decreasing $R$ if necessary, for any $s\in[\tau_0,\tau]$ and $\rho\in\pdfspace$ which satisfies $d_{\pdfspace}(\rho,\pi_s)<R$, we have 
    \[
    \norm[\mathrm{op}]{\DD_{\left[\log\frac{\rho}{\pi_s}\right]}^2\iteration_{\pi_{s}}}<C.
    \]
    Otherwise, for every $j>0$, there exist $s_j\in[\tau_0,\tau]$ and $\rho_j\in\pdfspace$ satisfying $d_{\pdfspace}(\rho_j,\pi_{s_j})<\min\left\{R,\frac{1}{j+1}\right\}$ such that
    \[
    \norm[\mathrm{op}]{\DD_{\left[\log\frac{\rho_j}{\pi_{s_j}}\right]}^2\iteration_{\pi_{s_j}}}\geq C.
    \]
    Note that $\left\{s_j\right\}_{j=0}^\infty$ has a convergent subsequence. Without loss of generality, we assume that the sequence $\left\{s_j\right\}_{j=0}^\infty$ converges to $s^\star$ as $j\to\infty$. Then, we have 
    \[
    d_{\pdfspace}(\pi_{s_j},\pi_{s^\star})=\abs{s_j-s^\star}\norm[\energyspace]{\left[r\right]}\leq \abs{s_j-s^\star}M\to0,
    \]
    and
    \[
    d_{\pdfspace}(\rho_j,\pi_{s^\star})\leq d_{\pdfspace}(\rho_j,\pi_{s_j})+d_{\pdfspace}(\pi_{s^\star},\pi_{s_j})\to0.
    \]
    Moreover,
    \[
        \norm[\energyspace]{\left[\log\frac{\rho_j}{\pi_{s_j}}\right]}
        =d_{\pdfspace}(\rho_j,\pi_{s_j})\to0.
    \]
    Therefore, Proposition~\ref{prop:25} gives
    \[
    \norm[\mathrm{op}]{\DD_{\left[\log\frac{\rho_j}{\pi_{s_j}}\right]}^2\iteration_{\pi_{s_j}}}\to \norm[\mathrm{op}]{\DD _0^2\iteration_{\pi_{s^\star}}},
    \]
    which leads to a contradiction since $\norm[\mathrm{op}]{\DD_0^2\iteration_{\pi_{s^\star}}}\leq C-1$. This completes the first step.

    For the second step, we define $\Delta_{\max}=\frac1{M+1}\min\left\{R,\frac{1}{C}\right\}>0$. Then, for any $s\in[\tau_0,\tau]$ and $\Delta\in[0,\min\{s-\tau_0,\Delta_{\max}\}]$, we have
    \[
    d_{\pdfspace}(\pi_{s-\Delta},\pi_s)=\Delta\norm[\energyspace]{\left[r\right]}\leq \Delta_{\max}M<\min\left\{R,\frac{1}{C}\right\}.
    \]
    
    For the last step, consider any finite grid $\tau_0<\tau_1<\cdots<\tau_n=\tau$ satisfying $\Delta_i:=\tau_i-\tau_{i-1}\leq\Delta_{\max}$ for all $i$. The same Taylor expansion as in the proof of Theorem~\ref{thm:8}, together with the uniform bound in the first step, gives
    \[
        d_{\pdfspace}\left(\rho_{k+1}^{(i)},\pi^{(i)}\right)
        \leq C d_{\pdfspace}\left(\rho_k^{(i)},\pi^{(i)}\right)^2
    \]
    whenever $d_{\pdfspace}(\rho_k^{(i)},\pi^{(i)})<R$. The second step gives
    \[
        d_{\pdfspace}\left(\rho_0^{(i)},\pi^{(i)}\right)<\min\left\{R,\frac1C\right\}.
    \]
    By induction, the quadratic estimate holds for all $k$, and $\rho_k^{(i)}$ converges quadratically to $\pi^{(i)}$ for every $i$. The local quadratic convergence of the canonical velocity fields follows immediately.
\end{proof}

\section{Newton Matching for the Isotropic Gaussian Family}\label{app:3}

This appendix presents an analytically tractable instance of Newton Matching on the isotropic Gaussian family
\[
\left\{
\Normal(m,\sigma^2I):
m\in\R^d,\ \sigma>0
\right\}.
\]
Appendix~\ref{subsec:38} shows that this family is closed under the ideal Newton Matching update and derives the exact one-step parameter update. Appendix~\ref{subsec:39} then establishes global convergence of both forward and reverse KL divergences, together with local quadratic rates.

In this appendix, it suffices to analyze the normalized case $\tau=1$, as we can equivalently apply $\hat{r}:=\tau r$ and $\hat\eta:=\frac\eta\tau$ for general $\tau>0$.

\subsection{Gaussian Closure}\label{subsec:38}

This subsection derives the closure property of Newton Matching in the isotropic Gaussian family. In other words, if the current density $\rho$ and the target density $\pi$ are both isotropic Gaussian, then the updated density $q=\terminalmap(v^\rho+\eta\Gamma^{\rho,\tilde{r}^\rho})$ is also isotropic Gaussian.

We first derive the canonical velocity field of a Gaussian distribution.

\begin{proposition}\label{prop:26}
    Consider a Gaussian distribution $\rho=\Normal(m,\Sigma)$, $m\in\R^d$, $\Sigma\in\S_{++}^d$. Its canonical velocity field is
    \begin{equation}\label{eq:182}
        v_t^\rho(x_t)=\dot{\alpha}_tm+\left(\alpha_t\dot{\alpha}_t\Sigma+\beta_t\dot{\beta}_tI\right)\left(\alpha_t^2\Sigma+\beta_t^2I\right)^{-1}(x_t-\alpha_tm).
    \end{equation}
    In particular, if $\rho=\Normal(m,\sigma^2I)$ is isotropic, then its canonical velocity field is
    \begin{equation}\label{eq:183}
        v_t^\rho(x_t)=\dot{\alpha}_t m+\frac{\dot{\beta}_t\beta_t+\dot{\alpha}_t\alpha_t\sigma^2}{\beta_t^2+\alpha_t^2\sigma^2}(x_t-\alpha_t m).
    \end{equation}
\end{proposition}

\begin{proof}
    Consider a Gaussian distribution $\rho=\Normal(m,\Sigma)$. Under independent coupling, the interpolant $X_t=\alpha_tX_1+\beta_tX_0$ satisfies
    \[
    X_t\sim\Normal\left(\alpha_tm,\alpha_t^2\Sigma+\beta_t^2I\right),\qquad
    \left[\begin{array}{c}
        X_1\\X_t
    \end{array}\right]
    \sim
    \Normal\left(\left[\begin{array}{c}
        m\\\alpha_t m
    \end{array}\right],\left[\begin{array}{cc}
        \Sigma&\alpha_t\Sigma\\ \alpha_t\Sigma & \alpha_t^2\Sigma+\beta_t^2I
    \end{array}\right]\right),
    \]
    which gives
    \[
    p_{1|t}^\rho(\cdot|x_t)=\Normal\left(m+\alpha_t\Sigma\left(\alpha_t^2\Sigma+\beta_t^2I\right)^{-1}(x_t-\alpha_tm),\beta_t^2\Sigma\left(\alpha_t^2\Sigma+\beta_t^2I\right)^{-1}\right).
    \]
    Then,
    \[
    v_t^\rho(x_t)=\E_{X_1\sim p_{1|t}^\rho(\cdot|x_t)}\left[\frac{\dot\beta_t}{\beta_t}x_t+\frac{\alpha_t\kappa_t}{\beta_t^2}X_1\right]
    \]
    gives~\eqref{eq:182}. In particular, substituting $\Sigma=\sigma^2I$ into~\eqref{eq:182} implies~\eqref{eq:183}.
\end{proof}

The following proposition provides the one-step updated density for the isotropic Gaussian, which is a special case of Proposition~\ref{prop:8}. The full Gaussian family 
$$\left\{\Normal(m,\Sigma):m\in\R^d,\Sigma\in\S_{++}^d\right\}$$
is also closed under the ideal Newton Matching update; however, the resulting matrix-valued parameter recursion is considerably more involved. For simplicity, we restrict our analysis to the isotropic case in this appendix.

\begin{proposition}\label{prop:27}
    Consider the target density $\pi=\Normal(m_\pi,\sigma_\pi^2I)$ and the current density $\rho=\Normal(m_\rho,\sigma_\rho^2I)$. The regularized reward satisfies $\tilde{r}^\rho(x)=\log\frac{\pi(x)}{\rho(x)}+\const$. Consider a stepsize $\eta\in\R$. Let $q=\terminalmap(v^\rho+\eta\Gamma^{\rho,\tilde{r}^\rho})$. Then, the updated density is also Gaussian:
    \[
    q=\Normal(m_q,\sigma_q^2I),\quad m_q:=m_\rho+(m_\pi-m_\rho)\cdot w_\eta\left(\frac{\sigma_\rho^2}{\sigma_\pi^2}\right),\quad \sigma_q:=\sigma_\rho\exp\left(\frac{\eta}{2}\left(1-\frac{\sigma_\rho^2}{\sigma_\pi^2}\right)\right),
    \]
    where
    \[
    w_\eta(z):=\eta z\int_{0}^{1}\exp\left(\frac{\eta}{2}(1-z)(1-s^2)\right)\,\dd s.
    \]
\end{proposition}

\begin{proof}
Since
\[
\tilde{r}^\rho(x_1)
=
\log\frac{\pi(x_1)}{\rho(x_1)}
+\const
=
\frac{1}{2\sigma_\rho^2}
\norm[2]{x_1-m_\rho}^2
-
\frac{1}{2\sigma_\pi^2}
\norm[2]{x_1-m_\pi}^2
+\const,
\]
we have
\[
\nabla\tilde{r}^\rho(x_1)
=
\left(\frac{1}{\sigma_\rho^2}-\frac{1}{\sigma_\pi^2}\right)x_1
+\frac{m_\pi}{\sigma_\pi^2}-\frac{m_\rho}{\sigma_\rho^2}.
\]

According to Appendix~\ref{subsec:50}, the covariance matrix of $p_{1|t}^\rho(\cdot|x_t)$ is a posterior Stein kernel. Therefore,
\[
\begin{aligned}
    &\Cov_{X_1\sim p_{1|t}^\rho(\cdot|x_t)}\left(X_1,\tilde{r}^\rho(X_1)\right)\\
    =&\frac{\beta_t^2\sigma_\rho^2}{\beta_t^2+\alpha_t^2\sigma_\rho^2}\E_{X_1\sim p_{1|t}^\rho(\cdot|x_t)}\left[\nabla\tilde{r}^\rho(X_1)\right]\\
    =&\frac{\beta_t^2\sigma_\rho^2}{\beta_t^2+\alpha_t^2\sigma_\rho^2}\left(\left(\frac{1}{\sigma_\rho^2}-\frac{1}{\sigma_\pi^2}\right)\left(m_\rho+\frac{\alpha_t\sigma_\rho^2}{\beta_t^2+\alpha_t^2\sigma_\rho^2}(x_t-\alpha_t m_\rho)\right)+\frac{m_\pi}{\sigma_\pi^2}-\frac{m_\rho}{\sigma_\rho^2}\right)\\
    =&
    \frac{\beta_t^2\sigma_\rho^2}{\beta_t^2+\alpha_t^2\sigma_\rho^2}\left(\frac{m_\pi-m_\rho}{\sigma_\pi^2}+\frac{\alpha_t(\sigma_\pi^2-\sigma_\rho^2)}{\sigma_\pi^2\left(\beta_t^2+\alpha_t^2\sigma_\rho^2\right)}(x_t-\alpha_t m_\rho)\right).
\end{aligned}
\]
Consequently, we have
\[
\begin{aligned}
\Gamma_t^{\rho,\tilde{r}^\rho}(x_t)
=&
\frac{\alpha_t\kappa_t}{\beta_t^2}
\Cov_{X_1\sim p_{1|t}^\rho(\cdot|x_t)}
\left(
X_1,\tilde{r}^\rho(X_1)
\right)\\
=&\frac{\alpha_t\kappa_t\sigma_\rho^2}{(\beta_t^2+\alpha_t^2\sigma_\rho^2)\sigma_\pi^2}\left(m_\pi-m_\rho+\frac{\alpha_t(\sigma_\pi^2-\sigma_\rho^2)}{\beta_t^2+\alpha_t^2\sigma_\rho^2}(x_t-\alpha_t m_\rho)\right).
\end{aligned}
\]

Let $Y_t$ solve
\[
\frac{\dd Y_t}{\dd t}
=
v_t^\rho(Y_t)+\eta\Gamma_t^{\rho,\tilde{r}^\rho}(Y_t).
\]
Then, $Y_0\sim p_0$ implies $Y_1\sim q$. The remaining task is to solve the above ODE. By Proposition~\ref{prop:26}, a direct calculation gives
\[
    \frac{\dd}{\dd t}\left(\frac{Y_t-\alpha_t m_\rho}{\sqrt{\beta_t^2+\alpha_t^2\sigma_\rho^2}}\right)
    =\frac{\eta\Gamma_t^{\rho,\tilde{r}^\rho}(Y_t)}{\sqrt{\beta_t^2+\alpha_t^2\sigma_\rho^2}},\qquad
    \frac{\dd}{\dd t}\left(\frac{\sigma_\rho\alpha_t}{\sqrt{\beta_t^2+\alpha_t^2\sigma_\rho^2}}\right)=\frac{\sigma_\rho\alpha_t\kappa_t}{\left(\beta_t^2+\alpha_t^2\sigma_\rho^2\right)^{3/2}}.
\]
Then, we have
\[
\begin{aligned}
    \frac{\dd}{\dd t}\left(\frac{Y_t-\alpha_t m_\rho}{\sqrt{\beta_t^2+\alpha_t^2\sigma_\rho^2}}\right)=\frac{\eta\sigma_\rho}{\sigma_\pi^2}\left(m_\pi-m_\rho+\frac{\alpha_t(\sigma_\pi^2-\sigma_\rho^2)}{\sqrt{\beta_t^2+\alpha_t^2\sigma_\rho^2}}\frac{Y_t-\alpha_t m_\rho}{\sqrt{\beta_t^2+\alpha_t^2\sigma_\rho^2}}\right)\frac{\dd}{\dd t}\left(\frac{\sigma_\rho\alpha_t}{\sqrt{\beta_t^2+\alpha_t^2\sigma_\rho^2}}\right).
\end{aligned}
\]
Let $Z_t=\frac{Y_t-\alpha_t m_\rho}{\sqrt{\beta_t^2+\alpha_t^2\sigma_\rho^2}}$ and $\gamma_t=\frac{\sigma_\rho\alpha_t}{\sqrt{\beta_t^2+\alpha_t^2\sigma_\rho^2}}$. Equivalently,
\[
\dot{Z}_t=\frac{\eta}{\sigma_\pi^2}\left(\sigma_\rho(m_\pi-m_\rho)+(\sigma_\pi^2-\sigma_\rho^2)\gamma_tZ_t\right)\dot{\gamma}_t.
\]
In other words,
\[
\begin{aligned}
    \frac{\dd}{\dd t}\left(Z_t\exp\left(-\frac{\eta(\sigma_\pi^2-\sigma_\rho^2)}{2\sigma_\pi^2}\gamma_t^2\right)\right)
    =&\frac{\eta\sigma_\rho(m_\pi-m_\rho)}{\sigma_\pi^2}\exp\left(-\frac{\eta(\sigma_\pi^2-\sigma_\rho^2)}{2\sigma_\pi^2}\gamma_t^2\right)\dot\gamma_t.
\end{aligned}
\]
Integrating from $t=0$ to $t=1$ gives
\[
\begin{aligned}
    &\frac{Y_1-m_\rho}{\sigma_\rho}\exp\left(-\frac{\eta(\sigma_\pi^2-\sigma_\rho^2)}{2\sigma_\pi^2}\right)-X_0%
    =\frac{\eta\sigma_\rho(m_\pi-m_\rho)}{\sigma_\pi^2}\int_{0}^{1}\exp\left(-\frac{\eta(\sigma_\pi^2-\sigma_\rho^2)}{2\sigma_\pi^2}s^2\right)\,\dd s,
\end{aligned}
\]
which implies $q=\Normal(m_q,\sigma_q^2I)$.
\end{proof}

\subsection{Convergence Guarantee}\label{subsec:39}

This subsection investigates the convergence properties of Newton Matching in the isotropic Gaussian family. In Appendix~\ref{sub2sec:27}, we prove that the reverse KL decreases whenever the stepsize $\eta\in(0,1]$. The forward and reverse KL divergences both converge to zero. In Appendix~\ref{sub2sec:28}, we further prove the local quadratic convergence of both forward and reverse KL divergences.

\subsubsection{KL Descent and Global Convergence}\label{sub2sec:27}

Here, we investigate the KL descent and global convergence of Newton Matching. The following lemma estimates $w_\eta$ when the stepsize is not larger than 1.

\begin{lemma}\label{lem:6}
    Use the notation of Proposition~\ref{prop:27}. If $\eta\in(0,1]$, then for any $z>0$, we have $w_\eta(z)\in(0,2)$.
\end{lemma}

\begin{proof}
    Evidently, $w_\eta(z)>0$ holds for all $z>0$. For those $z\leq 1$, we have $w_\eta(z)\leq z\exp\left(\frac{1-z}{2}\right)\leq 1$. For those $z>1$, denote $\zeta=\frac{\eta}{2}(z-1)>0$, and we have
\[
w_\eta(z)=(2\zeta+\eta)\int_{0}^{1}e^{-\zeta(1-s^2)}\,\dd s\leq (2\zeta+1)\int_{0}^{1}e^{-\zeta(1-s^2)}\,\dd s.
\]
Integrating by parts gives
\[
\int_{0}^{1}e^{-\zeta(1-s^2)}\,\dd s = 1-2\zeta\int_{0}^{1}s^2e^{-\zeta(1-s^2)}\,\dd s.
\]
Hence,
\[
w_\eta(z)\leq 1+2\int_{0}^{1}\zeta(1-s^2)e^{-\zeta(1-s^2)}\,\dd s\leq 1+2e^{-1}<2.
\]
Therefore, for all $z>0$, we have $w_\eta(z)\in(0,2)$.
\end{proof}

We next give the KL divergence between two isotropic Gaussian distributions.

\begin{lemma}\label{lem:7}
    Consider two Gaussian distributions $\rho=\Normal(m_\rho,\sigma_\rho^2I)$ and $\pi=\Normal(m_\pi,\sigma_\pi^2I)$. Then, we have
    \begin{equation}\label{eq:184}
        \KL{\rho}{\pi}=\frac{\norm[2]{m_\rho-m_\pi}^2}{2\sigma_\pi^2}+\frac{d}{2}\left(\frac{\sigma_\rho^2}{\sigma_\pi^2}-1-\log\frac{\sigma_\rho^2}{\sigma_\pi^2}\right).
    \end{equation}
\end{lemma}

\begin{proof}
    Taking the expectation of
\[
\log\frac{\rho(x)}{\pi(x)}=\frac{d}{2}\log\frac{\sigma_\pi^2}{\sigma_\rho^2}-\frac{\norm[2]{x-m_\rho}^2}{2\sigma_\rho^2}+\frac{\norm[2]{x-m_\pi}^2}{2\sigma_\pi^2}
\]
under $\rho$ finishes the proof.
\end{proof}

Based on Lemmas~\ref{lem:6} and~\ref{lem:7}, we prove the descent property of the reverse KL. The following result serves as a specialization of Theorem~\ref{thm:4}.

\begin{proposition}\label{prop:28}
    Use the notation of Proposition~\ref{prop:27}. Denote
    \[
    g:\R_{++}\to\R,\qquad z\mapsto z-1-\log z.
    \]
    Then, for any $\eta\in\R$, the reverse KL of $q=\terminalmap(v^\rho+\eta\Gamma^{\rho,\tilde{r}^\rho})$ is
    \begin{equation}\label{eq:185}
    \KL{q}{\pi}=\frac{\norm[2]{m_\rho-m_\pi}^2}{2\sigma_\pi^2}\left(1-w_\eta\left(\frac{\sigma_\rho^2}{\sigma_\pi^2}\right)\right)^2+\frac{d}{2}g\left(\frac{\sigma_\rho^2}{\sigma_\pi^2}\exp\left(\eta\left(1-\frac{\sigma_\rho^2}{\sigma_\pi^2}\right)\right)\right).
    \end{equation}
    If $\eta\in(0,1]$, the reverse KL decreases:
    \[
    \KL{q}{\pi}\leq \KL{\rho}{\pi}.
    \]
    Furthermore,
    \[
    \KL{q}{\pi}=\KL{\rho}{\pi}\quad\Leftrightarrow\quad q=\rho=\pi.
    \]
\end{proposition}

\begin{proof}
    By Lemma~\ref{lem:7} and Proposition~\ref{prop:27}, we obtain the expression for $\KL{q}{\pi}$ in~\eqref{eq:185}. Lemma~\ref{lem:6} implies that, for $\eta\in(0,1]$, $w_\eta\left(\frac{\sigma_\rho^2}{\sigma_\pi^2}\right)\in(0,2)$; hence,
    \[
    \frac{\norm[2]{m_q-m_\pi}^2}{2\sigma_\pi^2}=\frac{\norm[2]{m_\rho-m_\pi}^2}{2\sigma_\pi^2}\left(1-w_\eta\left(\frac{\sigma_\rho^2}{\sigma_\pi^2}\right)\right)^2\leq \frac{\norm[2]{m_\rho-m_\pi}^2}{2\sigma_\pi^2}.
    \]
    Let $h_z(\eta):=g(ze^{\eta(1-z)})-g(z)=ze^{\eta(1-z)}-\eta(1-z)-z$, $z>0$, $\eta\in(0,1]$. We aim to prove $h_z(\eta)\leq0$. Note that $h_z''(\eta)=z(1-z)^2e^{\eta(1-z)}\geq0$; hence, $h_z$ is convex w.r.t. $\eta$. Therefore, we have $h_z(\eta)\leq(1-\eta)h_z(0)+\eta h_z(1)=\eta(ze^{1-z}-1)\leq0$. Then, $h_{\sigma_\rho^2/\sigma_\pi^2}(\eta)\leq0$ implies
    \[
    g\left(\frac{\sigma_q^2}{\sigma_\pi^2}\right)\leq g\left(\frac{\sigma_\rho^2}{\sigma_\pi^2}\right),
    \]
    and
    \[
    g\left(\frac{\sigma_q^2}{\sigma_\pi^2}\right)=g\left(\frac{\sigma_\rho^2}{\sigma_\pi^2}\right)\quad\Leftrightarrow\quad\sigma_\rho=\sigma_q=\sigma_\pi.
    \]
    Therefore, $\KL{q}{\pi}\leq \KL{\rho}{\pi}$.

    Assume that $\KL{q}{\pi}=\KL{\rho}{\pi}$. Then, we have
    \[
    \abs{1-w_\eta\left(\frac{\sigma_\rho^2}{\sigma_\pi^2}\right)}\norm[2]{m_\rho-m_\pi}=\norm[2]{m_\rho-m_\pi},\qquad g\left(\frac{\sigma_q^2}{\sigma_\pi^2}\right)=g\left(\frac{\sigma_\rho^2}{\sigma_\pi^2}\right).
    \]
    Hence, $\sigma_\rho=\sigma_q=\sigma_\pi$. Then, $w_\eta\left(\frac{\sigma_\rho^2}{\sigma_\pi^2}\right)=\eta>0$, which implies $q=\rho=\pi$.

    Assume that $q=\rho=\pi$. Then, $\KL{q}{\pi}=0=\KL{\rho}{\pi}$ evidently holds.
\end{proof}

In contrast to the case in Section~\ref{subsec:12}, any isotropic Gaussian distribution $\Normal(m,\sigma^2I)$ can be represented by $(d+1)$-dimensional parameters $m,\sigma$. Therefore, global convergence can be proved directly from the explicit parameter recursions, without the additional anti-collapse or dissipation-vanishing conditions in~\eqref{eq:50} and~\eqref{eq:51}.

\begin{proposition}\label{prop:29}
    Use the notation of Proposition~\ref{prop:27} and apply Newton Matching iteratively. For the $k$th step, the density is denoted by $\rho_k=\Normal(m_k,\sigma_k^2I)$. Let $\rho_{k+1}=\terminalmap(v^{\rho_k}+\eta_k\Gamma^{\rho_k,\tilde{r}^{\rho_k}})$ where the stepsize $\eta_k\in (0,1]$. If $\sum_{k=0}^{\infty}\eta_k=\infty$, then as $k\to\infty$,
    \[
    m_k\to m_\pi,\qquad \sigma_k\to \sigma_\pi,\qquad \KL{\rho_k}{\pi}\to0,\qquad \KL{\pi}{\rho_k}\to0.
    \]
    Consequently, the velocity field also converges:
    \[
    \int_{0}^{1}\frac{1}{\kappa_t}\E_{X_t\sim p_t^{\rho_k}}\left[\norm[2]{v_t^{\rho_k}(X_t)-v_t^{\pi}(X_t)}^2\right]\dd t\to0,\,
    \int_{0}^{1}\frac{1}{\kappa_t}\E_{X_t\sim p_t^{\pi}}\left[\norm[2]{v_t^{\rho_k}(X_t)-v_t^{\pi}(X_t)}^2\right]\dd t\to0.
    \]
\end{proposition}

\begin{proof}
    For the first step, we prove $\lim_{k\to\infty}\sigma_k=\sigma_\pi$. Denote $z_k:=\frac{\sigma_k^2}{\sigma_\pi^2}>0$. Consider the case where there exists a $k_0$, such that $z_{k_0}\leq1$. Then, for all $k\geq k_0$, we have $z_k\leq z_{k+1}=z_k e^{\eta_k(1-z_k)}\leq1$. Denote $z_\star:=\lim_{k\to\infty}z_k\in(0,1]$. Then, for all $k\geq k_0$, $z_k\leq\frac{1+z_\star}{2}$; hence, $\log z_{k+1}-\log z_k=\eta_k(1-z_k)\geq\frac{1-z_\star}{2}\eta_k$. Then, we have $\lim_{k\to\infty}\log z_k\geq \log z_{k_0}+\frac{1-z_\star}{2}\sum_{k=k_0}^{\infty}\eta_k=\infty$, which leads to a contradiction. Therefore, we must have $z_\star=1$. In other words,
    \[
    \exists k_0>0\text{, s.t. }\sigma_{k_0}\leq\sigma_\pi\,\Rightarrow\,\lim_{k\to\infty}\sigma_k=\sigma_\pi.
    \]
    Consider another case where for all $k$, $z_k>1$. Then, $1<z_{k+1}=z_ke^{\eta_k(1-z_k)}<z_k$. Denote $z_\star:=\lim_{k\to\infty}z_k\in[1,\infty)$. Assume that $z_\star>1$. Then, $z_k>\frac{1+z_\star}{2}$; hence, $\log z_{k+1}-\log z_k=\eta_k(1-z_k)\leq\frac{1-z_\star}{2}\eta_k<0$. Then, we have $\lim_{k\to\infty}\log z_k\leq \log z_{0}+\frac{1-z_\star}{2}\sum_{k=0}^{\infty}\eta_k=-\infty$, which leads to a contradiction. Therefore, we must have $z_\star=1$. In other words,
    \[
    \forall k,\,\sigma_{k}>\sigma_\pi\,\Rightarrow\,\lim_{k\to\infty}\sigma_k=\sigma_\pi.
    \]
    In summary, we have proved unconditionally that
    \[
    \lim_{k\to\infty}\sigma_k=\sigma_\pi.
    \]

    For the second step, we prove $\lim_{k\to\infty}m_k=m_\pi$. Denote $z_k:=\frac{\sigma_k^2}{\sigma_\pi^2}$ and $e_k=\norm[2]{m_k-m_\pi}$. We have proved in the first step that $\lim_{k\to\infty}z_k=1$. Hence, there exists a sufficiently large $k_0$, such that for all $k\geq k_0$, $z_k\in\left[\frac12,\frac32\right]$. Then, we have
    \[
    \frac{\eta_k}{2e^{1/4}}\leq w_k=\eta_kz_k\int_{0}^{1}e^{\frac{\eta_k}2(1-z_k)(1-s^2)}\,\dd s\leq\frac{3}{2}.
    \]
    Hence,
    \[
    \norm[2]{m_{k+1}-m_\pi}^2=(1-w_k(2-w_k))\norm[2]{m_k-m_\pi}^2\leq\left(1-\frac{\eta_k}{4e^{1/4}}\right)\norm[2]{m_k-m_\pi}^2.
    \]
    By $\sum_{k=k_0}^{\infty}\eta_k=\infty$, we have
    \[
    \lim_{k\to\infty}\norm[2]{m_k-m_\pi}=0.
    \]

    By~\eqref{eq:184}, we have $\KL{\rho_k}{\pi}\to0$ and $\KL{\pi}{\rho_k}\to0$. The convergence of the canonical velocity field is further given by Corollary~\ref{cor:8}.
\end{proof}

\subsubsection{Local Quadratic Convergence}\label{sub2sec:28}

Proposition~\ref{prop:29} establishes global convergence of Newton Matching in the isotropic Gaussian family. Here, we establish the local quadratic convergence rate. The following proposition serves as a specialization of Theorem~\ref{thm:8}.

\begin{proposition}
    Use the notation of Proposition~\ref{prop:28} and let $\eta_k=1$ for all $k$. If the reverse KL satisfies
    \[
    \KL{\rho_0}{\pi}<\frac{d}{2}\left(\frac12-\log\frac32\right),
    \]
    then we have
    \[
        \KL{\rho_{k+1}}{\pi}\leq \frac{39}{2d}\KL{\rho_k}{\pi}^2.
    \]
    Therefore, the reverse KL and the canonical velocity field exhibit local quadratic convergence:
    \begin{align*}
        &\frac{39}{2d}\KL{\rho_k}{\pi}=\frac{39}{2d}\int_{0}^{1}\frac{1}{\kappa_t}\E_{X_t\sim p_t^{\rho_k}}\left[\norm[2]{v_t^{\rho_k}(X_t)-v_t^{\pi}(X_t)}^2\right]\dd t\leq \left(\frac{39}{2d}\KL{\rho_0}{\pi}\right)^{2^k}\to 0.
    \end{align*}
    
    If the forward KL satisfies
    \[
    \KL{\pi}{\rho_0}<\frac{d}{2}\left(\frac12-\log\frac32\right),
    \]
    then we have
    \[
        \KL{\pi}{\rho_{k+1}}\leq \frac{875}{48d}\KL{\pi}{\rho_k}^2.
    \]
    Therefore, the forward KL and the canonical velocity field exhibit local quadratic convergence:
    \begin{align*}
        &\frac{875}{48d}\KL{\pi}{\rho_k}=\frac{875}{48d}\int_{0}^{1}\frac{1}{\kappa_t}\E_{X_t\sim p_t^{\pi}}\left[\norm[2]{v_t^{\rho_k}(X_t)-v_t^{\pi}(X_t)}^2\right]\dd t\leq \left(\frac{875}{48d}\KL{\pi}{\rho_0}\right)^{2^k}\to 0.
    \end{align*}
\end{proposition}

\begin{proof}
    Denote $z_k=\frac{\sigma_k^2}{\sigma_\pi^2}$. By Proposition~\ref{prop:27}, we have $z_{k+1}=z_k\exp(1-z_k)$. Then, we have
    \[
    \KL{\rho_k}{\pi}=\frac{\norm[2]{m_k-m_\pi}^2}{2\sigma_\pi^2}+\frac{d}{2}g(z_k),\qquad \KL{\pi}{\rho_k}=\frac{\norm[2]{m_k-m_\pi}^2}{2\sigma_k^2}+\frac{d}{2}g(z_k^{-1}).
    \]
    Denote
    \[
    w(z)=z\exp\left(\frac{1-z}{2}\right)\int_{0}^{1}\exp\left(-\frac{1-z}{2}s^2\right)\,\dd s=z\int_{0}^{1}\exp\left(\frac{1-z}{2}\left(1-s^2\right)\right)\,\dd s.
    \]
    Then, we have
    \[
    m_{k+1}-m_\pi=(1-w(z_k))(m_k-m_\pi).
    \]
    
    We first estimate $\KL{\rho_k}{\pi}$. By $\KL{\rho_k}{\pi}\leq \KL{\rho_0}{\pi}<\frac{d}{2}\left(\frac12-\log\frac32\right)$, we have $z_k\in\left(\frac12,\frac32\right)$. We have $w(1)=1$ and $0\leq w'(z_k)\leq 1$; hence, \[\abs{w(z_k)-1}\leq\abs{z_k-1}.\] Elementary calculus on $z\in\left(\frac12,\frac32\right)$ also gives
    \[
    g(z_k)\geq \frac{1}{3}(z_k-1)^2,\qquad g(z_{k+1})\leq(z_k-1)^4.
    \]
    Since
    \[
        \frac{\norm[2]{m_{k+1}-m_\pi}^2}{2\sigma_\pi^2}
        =(w(z_k)-1)^2\frac{\norm[2]{m_k-m_\pi}^2}{2\sigma_\pi^2},\qquad(w(z_k)-1)^2\leq(z_k-1)^2\leq 3g(z_k),
    \]
    we have
    \[
    \frac{\norm[2]{m_{k+1}-m_\pi}^2}{2\sigma_\pi^2}\leq \frac{6}{d}\cdot\frac{\norm[2]{m_k-m_\pi}^2}{2\sigma_\pi^2}\cdot\frac{d}{2}g(z_k)\leq \frac{3}{2d}\KL{\rho_k}{\pi}^2.
    \]
    Since
    \[
    \frac{d}{2}g(z_{k+1})\leq\frac{d}{2}(z_{k}-1)^4\leq \frac{9d}{2}g(z_k)^2\leq\frac{18}{d}\KL{\rho_k}{\pi}^2,
    \]
    we have
    \[
    \KL{\rho_{k+1}}{\pi}=\frac{\norm[2]{m_{k+1}-m_\pi}^2}{2\sigma_\pi^2}+\frac{d}{2}g(z_{k+1})\leq\frac{39}{2d}\KL{\rho_k}{\pi}^2.
    \]
    Therefore, $\frac{39}{2d}\cdot \frac{d}{2}\left(\frac12-\log\frac32\right)=\frac{39}{4}\left(\frac12-\log\frac32\right)<0.93<1$ proves the local quadratic convergence of $\KL{\rho_k}{\pi}$.

    Next, we estimate $\KL{\pi}{\rho_k}$. Since $\KL{\pi}{\rho_0}<\frac{d}{2}g\left(\frac32\right)$, we have $z_0\in\left(\frac23,\frac53\right)$. A direct calculation shows that $z_1=z_0e^{1-z_0}\in\left(\frac56,1\right]\subset\left(\frac23,\frac53\right)$. Consequently, for all $k$, we have $z_k\in\left(\frac23,\frac53\right)$. On $z\in\left(\frac23,\frac53\right)$, elementary calculus gives
    \[
    \abs{w(z_k)-1}\leq\abs{z_k-1},\qquad g(z_k^{-1})\geq \frac{6}{25}(z_k-1)^2,\qquad g(z_{k+1}^{-1})\leq\frac{81}{200}(z_k-1)^4.
    \]
    Since
    \[
    \frac{\norm[2]{m_{k+1}-m_\pi}^2}{2\sigma_{k+1}^2}\leq \frac{z_k}{z_{k+1}}(w(z_k)-1)^2\frac{\norm[2]{m_k-m_\pi}^2}{2\sigma_k^2}\leq\frac{25}{3}g(z_k^{-1})\frac{\norm[2]{m_k-m_\pi}^2}{2\sigma_k^2},
    \]
    we obtain
    \[
    \frac{\norm[2]{m_{k+1}-m_\pi}^2}{2\sigma_{k+1}^2}\leq \frac{25}{6d}\KL{\pi}{\rho_k}^2.
    \]
    We also have
    \[
    \frac{d}{2}g(z_{k+1}^{-1})\leq\frac{81d}{400}(z_{k}-1)^4\leq \frac{225d}{64}g(z_k^{-1})^2\leq\frac{225}{16d}\KL{\pi}{\rho_k}^2.
    \]
    Thus,
    \[
    \KL{\pi}{\rho_{k+1}}=\frac{\norm[2]{m_{k+1}-m_\pi}^2}{2\sigma_{k+1}^2}+\frac{d}{2}g(z_{k+1}^{-1})\leq\frac{875}{48d}\KL{\pi}{\rho_k}^2.
    \]
    Since $\frac{875}{48d}\cdot \frac{d}{2}\left(\frac12-\log\frac32\right)=\frac{875}{96}\left(\frac12-\log\frac32\right)<0.87<1$, the recurrence establishes local quadratic convergence of $\KL{\pi}{\rho_k}$.
\end{proof}

By the explicit Gaussian KL formula, both the forward and reverse KL divergences are locally comparable to the squared Euclidean error in the parameters $(m,\sigma)$ near the target $(m_\pi,\sigma_\pi)$. Since all norms on this finite-dimensional parameter space are equivalent, the one-step KL bounds above imply quadratic convergence of $(m_k,\sigma_k)$ under any fixed norm. We therefore do not state the corresponding norm-based bounds separately.

\section{Stochastic Foundations and Computational Tools}\label{app:4}

This appendix develops stochastic foundations and computational tools used primarily in Sections~\ref{sec:5} and~\ref{sec:6}; some results are also used in the convergence analysis of Section~\ref{sub2sec:9}. All results depend on the terminal condition
\[
\alpha_1=1,\qquad \beta_1=0,
\]
and are independent of the initial values $\alpha_0$, $\beta_0$. Therefore, conclusions in this appendix apply both to the standard flow matching construction, where
\[
\alpha_0=0,\qquad \beta_0=1,\qquad \alpha_1=1,\qquad \beta_1=0,
\]
and to the one-sided interpolant considered in Appendix~\ref{subsec:57}, where
\[
\widetilde\alpha_0=0,\qquad \widetilde\beta_0=0,\qquad \widetilde\alpha_1=1,\qquad \widetilde\beta_1=0.
\]
Here, $\widetilde{\alpha}_t$ and $\widetilde{\beta}_t$ denote the schedule in one-sided interpolants. The extension of Newton Matching to one-sided interpolants is discussed in Appendix~\ref{subsec:57}.

Appendix~\ref{subsec:40} presents the SDE setup and assumptions used throughout the appendix. Appendix~\ref{subsec:41} then derives the backward equation satisfied by the posterior value $V_t^\rho[f]$ defined in~\eqref{eq:17}. This equation contributes to the value-ascent certificate in Section~\ref{subsec:8}.

The remaining subsections of this appendix provide stochastic computational tools for Newton Matching.
Appendix~\ref{subsec:42} constructs the posterior-preserving SDE, whose terminal transition kernel is exactly $p_{1|t}^\rho(\cdot| x_t)$ in~\eqref{eq:13}. This SDE supplies the exact posterior sampler $X_1\sim p_{1|t}^\rho(\cdot| x_t)$ used by the reverse construction in Section~\ref{subsec:15}. Its path distribution also supports the SDE-based estimation in Section~\ref{subsec:16}, the gradient form in Section~\ref{subsec:17}, and the reference-adjoint approximation in Section~\ref{sub2sec:14}.

Appendix~\ref{subsec:43} establishes the bridge universality of the posterior-preserving SDE: the law of the intervening path conditioning on its two endpoints is independent of the terminal density $\rho$ and admits an explicit Gaussian representation. This property enables path-dependent computations under the forward construction by separating the generation of the endpoint pair $(X_t,X_1)$ from the generation of a path with the required posterior-compatible law. Once the endpoint pair has been obtained, a universal bridge supplies the path needed for the SDE-based covariance calculation in Section~\ref{subsec:16} and the pathwise adjoint calculation in Section~\ref{subsec:17}.

Appendix~\ref{subsec:44} develops SDE-based representations of the log density $\log\rho(x)$ and the log-density ratio $\log\frac{\rho_2(x)}{\rho_1(x)}$. In Section~\ref{subsec:16}, the pathwise identities provide exact evaluations of the regularized reward appearing in the covariance-form targets. In Section~\ref{subsec:17}, the conditional-expectation identities move the density-ratio correction from the terminal objective to a running objective, while retaining an explicit marginal correction. This decomposition yields the Bolza realization of the gradient form. Section~\ref{subsec:24} uses a related decomposition to isolate the posterior-specific correction that may be approximated.

Appendix~\ref{subsec:45} develops initial-state sensitivity and adjoint calculus for generic SDEs. The pathwise and endpoint-conditioned identities underlie the posterior sensitivity kernel in Appendix~\ref{subsec:49}, while the conditional-mean adjoint equation is used to prove critical-point consistency in Appendix~\ref{subsec:52}.

\subsection{Common SDE Setup and Basic Identities}\label{subsec:40}

This subsection provides the common SDE setup, standing assumptions, and stochastic-calculus tools used throughout the appendix. Appendix~\ref{sub2sec:29} introduces a generic SDE with a scalar state-independent diffusion coefficient, together with its solution, path laws, transition densities, and generator. Appendix~\ref{sub2sec:30} states the required regularity assumptions, while Appendix~\ref{sub2sec:31} records some lemmas, including the backward Kolmogorov equation, It\^o's formula, Dynkin's formula, and the Doob \(h\)-transform.

Throughout this appendix, identities involving realized SDE paths are understood to hold almost surely (a.s.) under the indicated path distribution, unless stated otherwise.

\subsubsection{SDE with a State-Independent Diffusion Coefficient}\label{sub2sec:29}

Fix an initial time \(t\in[0,1)\) and an initial state \(x_t\in\R^d\). We consider the SDE
\begin{subequations}\label{eq:186}
    \begin{align}
        &\dd Y_{s}^{b,\sigma}(t,x_t)
        =
        b_s\left(Y_{s}^{b,\sigma}(t,x_t)\right)\dd s
        +
        \sigma_s\dd W_s,
        \qquad s\in[t,1],\\
        \label{eq:187}
        &Y_{t}^{b,\sigma}(t,x_t)=x_t.
    \end{align}
\end{subequations}
Here, \(b_s:\R^d\to\R^d\) is the drift, \(\sigma_s\in\R\) is a scalar state-independent diffusion coefficient, and \(W\) is a standard
\(d\)-dimensional Brownian motion. The integral form of SDE~\eqref{eq:186} is
\begin{equation}
    Y_{s}^{b,\sigma}(t,x_t)
    =
    x_t
    +
    \int_t^s b_u\left(Y_{u}^{b,\sigma}(t,x_t)\right)\dd u
    +
    \int_t^s\sigma_u\dd W_u.
    \label{eq:188}
\end{equation}
Whenever derivatives with respect to the initial state \(x_t\) are considered, the same Brownian motion $(W_s)_{s\in[t,1]}$ is used for all initial states. 

For \(t\le s\le1\), we denote the random path by
\begin{equation*}
    \BY_{[t,s]}^{b,\sigma}(t,x_t)
    :=
    \left(Y_{u}^{b,\sigma}(t,x_t)\right)_{u\in[t,s]}.
\end{equation*}
We denote the path distribution of \(\BY_{[t,s]}^{b,\sigma}(t,x_t)\) by
\begin{equation*}
    \mathbb P_{[t,s]|t}^{b,\sigma}(\cdot|x_t).
\end{equation*}
The transition density is denoted by
\[
q_{s|t}^{b,\sigma}(\cdot|x_t).
\]

For every function \(f\in C^2(\R^d)\) and $s\in[t,1]$, define the generator of SDE~\eqref{eq:186} by
\begin{equation}
    \mathcal A_s^{b,\sigma} f(x)
    :=
    b_s(x)\cdot\nabla f(x)
    +
    \frac{\sigma_s^2}{2}\Delta f(x).
    \label{eq:189}
\end{equation}

\subsubsection{Regularity Assumptions}\label{sub2sec:30}

To avoid repeating technical qualifications in every result, we impose the following standing conditions throughout this appendix.

For every \(t\in(0,1)\), assume that the map \((s,x)\mapsto b_s(x)\) is jointly Borel measurable on $[t,1]\times\R^d$, that 
\(b_s\) is $C^1$ for a.e. \(s\in[t,1]\), and that $\sigma$ is a deterministic Borel-measurable scalar function on $[t,1]$. We assume
\begin{equation*}
    \int_t^1
    \left(
        \norm[2]{b_s(0)}
        +
        \norm[\infty]{\nabla b_s}
        +
        \sigma_s^2
    \right)\dd s
    <
    \infty.
\end{equation*}
Consequently, the SDE~\eqref{eq:186} admits a unique strong solution on \([t,1]\). Whenever initial-state sensitivities are invoked, solutions from different initial states are driven by the same Brownian motion, and the resulting solution map $x_t\mapsto Y_s^{b,\sigma}(t,x_t)$
is assumed to be continuously differentiable. Since some coefficients used below---notably those of the posterior-preserving SDE---may be singular at \(t=0\), the preceding well-posedness assumptions are imposed only on intervals \([t,1]\) with \(t>0\). Whenever an SDE is written with an initial law at \(t=0\), it is understood as a solution on \((0,1]\) with the stated initial distribution as its zero-time entrance law; existence and uniqueness in law are assumed.

Unless stated otherwise, the densities, conditional densities, test functions, and path functionals appearing below are assumed to have sufficient regularity and integrability for all displayed expressions, derivatives, conditional expectations, and PDE identities to be well defined. We assume that all required interchanges of differentiation, integration, and conditional expectation, together with all applications of Tonelli's theorem and spatial integration by parts, are justified. Whenever a martingale property is invoked, the relevant stochastic integral or stochastic exponential is assumed to be a true martingale. We further assume that the required transition densities and endpoint-conditioned path distributions exist and that the backward equations invoked below admit unique classical solutions in the relevant function classes.

\subsubsection{Basic Identities}\label{sub2sec:31}

We next record several standard results \citep{oksendal2003stochastic}. The backward Kolmogorov equation is used in Appendix~\ref{subsec:42} to identify the terminal transition kernel of the posterior-preserving SDE. The Doob $h$-transform is used in Appendix~\ref{subsec:43} to compare posterior path distributions and establish bridge universality. It\^o's formula and Dynkin's formula are used in Appendix~\ref{subsec:44} to derive pathwise and conditional-expectation representations of terminal log densities and log-density ratios. Dynkin's formula is also used in Appendix~\ref{subsec:45} to derive the generator-form sensitivity identity and the conditional-mean adjoint equation.

In the following lemmas, $Y_s^{b,\sigma}$ denotes $Y_s^{b,\sigma}(t,x_t)$ generated by SDE~\eqref{eq:186} for simplicity.

\begin{lemma}[Backward Kolmogorov equation for terminal observables]
\label{lem:8}
Consider SDE~\eqref{eq:186} and an integrable terminal observable $\varphi:\R^d\to\R$ satisfying $\forall s\in[t,1]$, $x\in\R^d$,
\begin{equation*}
    \E_{Y_1\sim q_{1|s}^{b,\sigma}(\cdot|x)}\left[
        \abs{\varphi(Y_1)}
    \right]<\infty .
\end{equation*}
For every $s\in[t,1]$, define
\[
\varphi_s(x):=\E_{Y_1\sim q_{1|s}^{b,\sigma}(\cdot|x)}\left[\varphi(Y_1)\right].
\]
Then, $\forall s\in[t,1]$, $x\in\R^d$, we have the backward Kolmogorov equation:
\begin{equation*}%
    (\partial_s+\mathcal A_s^{b,\sigma})\varphi_s(x)=0,
    \qquad
    \varphi_1(x)=\varphi(x).
\end{equation*}
\end{lemma}

\begin{proof}
For \(s\in[t,1)\) and sufficiently small \(\varepsilon>0\), the Markov property gives
\begin{align*}
    \varphi_s(x)
    =&
    \E_{Y_1\sim q_{1|s}^{b,\sigma}(\cdot|x)}\left[
        \varphi(Y_1)
    \right]\\
    =&
    \E_{Y_{s+\varepsilon}\sim q_{s+\varepsilon|s}^{b,\sigma}(\cdot|x)}\left[
        \E_{Y_1\sim q_{1|s+\varepsilon}^{b,\sigma}(\cdot|Y_{s+\varepsilon})}
        \left[\varphi(Y_1)\right]
    \right]\\
    =&
    \E_{Y_{s+\varepsilon}\sim q_{s+\varepsilon|s}^{b,\sigma}(\cdot|x)}\left[
        \varphi_{s+\varepsilon}(Y_{s+\varepsilon})
    \right].
\end{align*}
The short-time expansion associated with the backward operator $\partial_s+\mathcal A_s^{b,\sigma}$ gives
\[
    0
    =
    \E_{Y_{s+\varepsilon}\sim q_{s+\varepsilon|s}^{b,\sigma}(\cdot|x)}\left[
        \varphi_{s+\varepsilon}(Y_{s+\varepsilon})
        -
        \varphi_s(x)
    \right]
    =
    \varepsilon(\partial_s+\mathcal A_s^{b,\sigma})\varphi_s(x)
    +
    o(\varepsilon).
\]
Letting \(\varepsilon\downarrow0\) proves the backward equation. The terminal condition follows directly from the definition.
\end{proof}

\begin{lemma}[It\^o's formula]
\label{lem:9}
Let \(F_s(x)\) be a regular function that is $C^1$ in $s$ and $C^2$ in $x$. For the SDE~\eqref{eq:186}, \(\forall t\le s\le 1\), we have It\^o's formula
\begin{align}
    F_s(Y_s^{b,\sigma})
    =
    F_t(Y_t^{b,\sigma})
    +
    \int_t^s
    (\partial_u+\mathcal A_u^{b,\sigma})F_u(Y_u^{b,\sigma})\dd u
    +
    \int_t^s
    \sigma_u\nabla F_u(Y_u^{b,\sigma})^\top\dd W_u.
    \label{eq:190}
\end{align}
\end{lemma}

\begin{proof}
This is the standard form of It\^o's formula applied to~\eqref{eq:186}.
\end{proof}

\begin{lemma}[Dynkin's formula]
\label{lem:10}
Let \(F_s(x)\) be \(C^{1,2}\) and satisfy the usual integrability assumptions. For the SDE~\eqref{eq:186}, \(t\le s\le 1\), we have Dynkin's formula
\begin{equation*}
    \E_{Y_{s}\sim q_{s|t}^{b,\sigma}(\cdot|x)}\left[
        F_s(Y_s)
    \right]
    =
    F_t(x)
    +
    \E_{\BY_{[t,s]}\sim \mathbb{P}_{[t,s]|t}^{b,\sigma}(\cdot|x)}\left[
        \int_t^s
        (\partial_u+\mathcal A_u^{b,\sigma})F_u(Y_u)\dd u
    \right].
\end{equation*}
\end{lemma}

\begin{proof}
Taking conditional expectation in~\eqref{eq:190} and using that the stochastic integral has conditional-mean zero gives the result.
\end{proof}

\begin{lemma}[Doob \(h\)-transform]
\label{lem:11}
Consider SDE~\eqref{eq:186}. Assume \(h_s(x)>0\) is \(C^{1}\) in $s$ and $C^2$ in $x$, satisfying
\begin{equation}
    (\partial_s+\mathcal A_s^{b,\sigma})h_s(x)=0.
    \label{eq:191}
\end{equation}
Assume that, for the solution started from \(Y_t^{b,\sigma}=x_t\), the process \(\left(\frac{h_u(Y_u^{b,\sigma})}{h_t(x_t)}\right)_{u\in[t,1]}\) is a true martingale. Consider another SDE
\begin{equation}\label{eq:192}
    \dd Z_s=\left[b_s(Z_s)+\sigma_s^2\nabla\log h_s(Z_s)\right]\dd s+\sigma_s\dd W_s
\end{equation}
which also starts from $Z_t=x_t$. Denote the conditional path distribution of $\BZ_{[t,s]}=(Z_u)_{u\in[t,s]}$ by $\mathbb{P}^Z_{[t,s]|t}(\cdot|x_t)$. Then,
\begin{equation}
    \mathbb{P}^Z_{[t,s]| t}(\cdot| x_t)
    \ll
    \mathbb{P}^{b,\sigma}_{[t,s]| t}(\cdot| x_t),\qquad
    \frac{\dd \mathbb{P}^Z_{[t,s]|t}}{\dd \mathbb{P}^{b,\sigma}_{[t,s]|t}}=\frac{h_s(Y_s^{b,\sigma})}{h_t(x_t)}.
    \label{eq:193}
\end{equation}
Let \(q^Z_{r|u}\) denote the transition densities from $u$ to $r$ of SDE~\eqref{eq:192}. Then, we have
\begin{equation}
    q^Z_{r|u}(x_r|x_u)=q^{b,\sigma}_{r|u}(x_r|x_u)\frac{h_r(x_r)}{h_u(x_u)},
    \qquad
    t\le u\le r\le s.
    \label{eq:194}
\end{equation}
\end{lemma}

\begin{proof}
By~\eqref{eq:191}, It\^o's formula in Lemma~\ref{lem:9} implies
\[
    \dd h_s(Y_s^{b,\sigma})=\sigma_s\nabla h_s(Y_s^{b,\sigma})^\top\dd W_s.
\]
So \(\frac{h_s(Y_s^{b,\sigma})}{h_t(x_t)}\) is a positive local martingale and, by assumption, a true martingale. Therefore, we have
\[
    \E_{Y_s\sim q_{s|t}^{b,\sigma}(\cdot|x_t)}\left[\frac{h_s(Y_s)}{h_t(x_t)}\right]=\E_{Y_t\sim q_{t|t}^{b,\sigma}(\cdot|x_t)}\left[\frac{h_t(Y_t)}{h_t(x_t)}\right]=\frac{h_t(x_t)}{h_t(x_t)}=1.
\]
Therefore, $\mathbb{P}^Z_{[t,s]|t}$ in~\eqref{eq:193} is a normalized path distribution. Then, by Bayes' rule,~\eqref{eq:194} holds. In the following, we prove that $\mathbb{P}^Z_{[t,s]|t}$ and $q^Z_{r|u}$ are the conditional path distribution and the transition density of SDE~\eqref{eq:192}, respectively.

Fix \(u\in[t,s)\), and let \(\varepsilon>0\) be sufficiently small
that \(u+\varepsilon\le s\). Then
\begin{align*}
\E_{Z_{u+\varepsilon}\sim q^Z_{u+\varepsilon|u}(\cdot|x)}
\left[
    \varphi(Z_{u+\varepsilon})
\right]
&=
\frac{1}{h_u(x)}
\E_{Y_{u+\varepsilon}\sim q_{u+\varepsilon|u}^{b,\sigma}(\cdot|x)}
\left[
    h_{u+\varepsilon}(Y_{u+\varepsilon})
    \varphi(Y_{u+\varepsilon})
\right].
\end{align*}
Denote $b_s^h(x):=b_s(x)+\sigma_s^2\nabla\log h_s(x)$. Consequently,
\[
    \mathcal A_u^{b^h,\sigma}\varphi(x)
    =
    \frac{1}{h_u(x)}
    (\partial_u+\mathcal A_u^{b,\sigma})(h_u\varphi)(x)
    =
    \mathcal A_u^{b,\sigma}\varphi(x)
    +\sigma_u^2\nabla\log h_u(x)\cdot\nabla\varphi(x).
\]
By weak uniqueness for~\eqref{eq:192}, the path measure
\(\mathbb{P}^Z_{[t,s]| t}(\cdot| x_t)\) coincides with the path distribution of \(\BZ_{[t,s]}\); hence, $\mathbb{P}^Z_{[t,s]|t}$ is the conditional path distribution of SDE~\eqref{eq:192}.
\end{proof}

\subsection{Backward Equation for Posterior Value}\label{subsec:41}

The posterior value is defined as a conditional expectation under \(p_{1|t}^\rho(\cdot|x_t)\). In this appendix, we call this conditional law the \emph{interpolant posterior} to distinguish its interpolant-based definition from the terminal transition kernel of an SDE. To obtain a dynamic characterization, we derive the backward equation for the posterior value, which is the key analytic link for the constructions that follow. In Appendix~\ref{subsec:42}, we construct an SDE whose conditional expectations of terminal observables solve the same backward terminal-value problem; uniqueness then implies that its terminal transition kernel coincides with the interpolant posterior. In Appendix~\ref{subsec:43}, taking the terminal observable to be \(\rho_2/\rho_1\) shows that the marginal density ratio \(p_t^{\rho_2}/p_t^{\rho_1}\) satisfies the backward equation associated with the \(\rho_1\) posterior-preserving SDE. This relation enables the Doob \(h\)-transform between posterior-preserving SDEs and leads to the universal-bridge result.

For a terminal observable $f:\R^d\to\R$, recall the posterior value defined in~\eqref{eq:17}:
\begin{equation*}
    V_t^\rho[f](x_t)
    =
    \E_{X_1\sim p_{1|t}^\rho(\cdot|x_t)}[f(X_1)].
\end{equation*}
Define the weighted marginal
\begin{equation*}
    U_t^\rho[f](x_t)
    :=
    p_t^\rho(x_t)V_t^\rho[f](x_t)
    =
    \int_{\R^d}f(x_1)\rho(x_1)p_{t|1}(x_t|x_1)\dd x_1.
\end{equation*}
At terminal time, we have
\begin{equation*}
    V_1^\rho[f](x_1)=f(x_1),
    \qquad
    U_1^\rho[f](x_1)=f(x_1)\rho(x_1).
\end{equation*}
At initial time, the independent coupling gives
\begin{equation*}
    V_0^\rho[f](x_0)=\E_{X_1\sim\rho}[f(X_1)],
\end{equation*}
which is independent of \(x_0\).

\begin{proposition}%
\label{prop:30}
For every sufficiently regular terminal observable $f$ and every $t\in(0,1)$,
\begin{equation}
    \partial_tU_t^\rho[f](x)
    +
    \nabla\cdot
    \left(
    \frac{\dot\alpha_t}{\alpha_t}xU_t^\rho[f](x)
    +
    \kappa_t\nabla U_t^\rho[f](x)
    \right)
    =0.
    \label{eq:195}
\end{equation}
Consequently, we have the backward equation for the posterior value:
\begin{equation}
    \partial_tV_t^\rho[f](x)
    +
    \left(
    \frac{\dot\alpha_t}{\alpha_t}x
    +
    2\kappa_t\nabla\log p_t^\rho(x)
    \right)\cdot\nabla V_t^\rho[f](x)
    +
    \kappa_t\Delta V_t^\rho[f](x)
    =0.
    \label{eq:196}
\end{equation}
\end{proposition}

\begin{proof}
For fixed $x_1$, the conditional density satisfies
\begin{equation*}
    \partial_tp_{t|1}(x_t|x_1)
    +
    \nabla_{x_t}\cdot
    \left(
    p_{t|1}(x_t|x_1)v_{t|1}(x_t|x_1)
    \right)
    =0.
\end{equation*}
Multiplying by $f(x_1)\rho(x_1)$ and integrating gives
\begin{equation}
    \partial_tU_t^\rho[f](x_t)
    +
    \nabla_{x_t}\cdot
    \left(
    \int_{\R^d}
    f(x_1)\rho(x_1)p_{t|1}(x_t|x_1)v_{t|1}(x_t|x_1)
    \dd x_1
    \right)
    =0.
    \label{eq:197}
\end{equation}
Since
\begin{equation*}
    \nabla_{x_t}p_{t|1}(x_t|x_1)
    =
    -\frac{x_t-\alpha_tx_1}{\beta_t^2}p_{t|1}(x_t|x_1),
\end{equation*}
we have
\begin{align*}
    \int_{\R^d}x_1p_{t|1}(x_t|x_1)f(x_1)\rho(x_1)\dd x_1
    =
    \frac{\beta_t^2}{\alpha_t}\nabla U_t^\rho[f](x_t)
    +
    \frac{x_t}{\alpha_t}U_t^\rho[f](x_t).
\end{align*}
Substituting~\eqref{eq:12} and the above equation into~\eqref{eq:197} proves~\eqref{eq:195}. Taking $f\equiv1$ gives
\begin{equation}
    \partial_tp_t^\rho(x)
    +
    \nabla\cdot
    \left(
    \frac{\dot\alpha_t}{\alpha_t}xp_t^\rho(x)
    +
    \kappa_t\nabla p_t^\rho(x)
    \right)
    =0.
    \label{eq:198}
\end{equation}
For general \(f\), substitute \(U_t^\rho[f]=p_t^\rho V_t^\rho[f]\) into~\eqref{eq:195}. Expanding the derivatives and dividing by \(p_t^\rho(x)\) gives
\[
\begin{aligned}
    &\partial_t V_t^\rho[f](x)
    +
    \left(
        \frac{\dot\alpha_t}{\alpha_t}x
        +
        2\kappa_t\nabla\log p_t^\rho(x)
    \right)\cdot\nabla V_t^\rho[f](x)
    +
    \kappa_t\Delta V_t^\rho[f](x)
    \\
    =&
    -\frac{V_t^\rho[f](x)}{p_t^\rho(x)}
    \left[
        \partial_t p_t^\rho(x)
        +
        \nabla\cdot
        \left(
            \frac{\dot\alpha_t}{\alpha_t}x p_t^\rho(x)
            +
            \kappa_t\nabla p_t^\rho(x)
        \right)
    \right]
    =0,
\end{aligned}
\]
which proves~\eqref{eq:196}.
\end{proof}

\subsection{Posterior-Preserving SDE}\label{subsec:42}

The backward equation derived in Appendix~\ref{subsec:41} suggests an SDE realization of the interpolant posterior. We define an SDE whose generator is the spatial differential operator appearing in~\eqref{eq:196}. The main result of this subsection shows that, when initialized at $Y_t=x_t$, its terminal state $Y_1$ has density $p_{1|t}^\rho(\cdot|x_t)$. Thus, the SDE provides an exact sampler from the interpolant posterior.

\begin{definition}
\label{def:8}
For a terminal density \(\rho\) and $t\in(0,1]$, consider the drift
\begin{equation}
    b_t^\rho(x)
    :=
    v_t^\rho(x)+\kappa_t\nabla\log p_t^\rho(x)
    =
    \frac{\dot\alpha_t}{\alpha_t}x
    +
    2\kappa_t\nabla\log p_t^\rho(x)
    =
    2v_t^\rho(x)-\frac{\dot\alpha_t}{\alpha_t}x.
    \label{eq:199}
\end{equation}
The posterior-preserving SDE is defined as
\begin{equation}
    \dd Y_t^\rho
    =
    b_t^\rho(Y_t^\rho)\dd t
    +
    \sqrt{2\kappa_t}\dd W_t.
    \label{eq:200}
\end{equation}
By definition~\eqref{eq:189}, the generator of the posterior-preserving SDE~\eqref{eq:200} is
\begin{equation*}%
    \calL_t^\rho f(x)
    =
    b_t^\rho(x)\cdot\nabla f(x)
    +
    \kappa_t\Delta f(x).
\end{equation*}
For \(0< t< s\le1\), denote its transition density by \(q_{s|t}^\rho(x_s|x_t)\), and denote its conditional path distribution by \(\mathbb P_{[t,s]|t}^{\rho}(\cdot|x_t)\) for simplicity.
\end{definition}

\begin{theorem}%
\label{thm:13}
For every sufficiently regular terminal observable $f$,
\begin{subequations}
    \label{eq:201}
    \begin{align}
        &(\partial_t+\calL_t^\rho)V_t^\rho[f](x)=0,
        \qquad t\in(0,1),\\
        &V_1^\rho[f](x)=f(x).
    \end{align}
\end{subequations}
Consequently, the terminal transition density of the posterior-preserving SDE, denoted by \(q_{1|t}^\rho(\cdot|x_t)\), satisfies
\begin{equation}
    q_{1|t}^\rho(\cdot|x_t)
    =
    p_{1|t}^\rho(\cdot|x_t).
    \label{eq:202}
\end{equation}

Moreover, consider the marginal-preserving SDE family
\begin{equation}
    \label{eq:203}
    \dd Y_t^{\rho,\sigma}
    =
    \left[
    v_t^\rho(Y_t^{\rho,\sigma})
    +
    \frac{\sigma_t^2}{2}\nabla\log p_t^\rho(Y_t^{\rho,\sigma})
    \right]\dd t
    +
    \sigma_t\dd W_t,
\end{equation}
whose transition density from $t$ to $s$ is denoted by \(q_{s|t}^{\rho,\sigma}(\cdot|x_t)\). Assume that, for every $t\in(0,1)$, $x_t\in\R^d$, and sufficiently regular terminal observable \(f\),
\begin{equation*}
    \E_{Y_1\sim q_{1|t}^{\rho,\sigma}(\cdot|x_t)}\left[
        f\left(Y_1\right)
    \right]
    =
    V_t^\rho[f](x_t)
\end{equation*}
Assume additionally that, for every $t\in(0,1)$, there exists a sufficiently regular terminal observable $f$ such that
\begin{equation}\label{eq:204}
    \Delta V_t^\rho[f]
    +
    \nabla\log p_t^\rho
    \cdot
    \nabla V_t^\rho[f]
    \not\equiv
    0.
\end{equation}
Then, for a.e. \(t\in(0,1)\), we have
\begin{equation*}
    \sigma_t^2
    =
    2\kappa_t.
\end{equation*}
In other words, the schedule $\sigma_t^2=2\kappa_t$
is the unique one whose conditional values agree with the interpolant posterior $p_{1|t}^\rho(\cdot|x_t)$ for every sufficiently regular terminal observable.
\end{theorem}

\begin{proof}
Equation~\eqref{eq:196} is exactly~\eqref{eq:201}. Under the posterior-preserving SDE~\eqref{eq:200}, the conditional value
\begin{equation*}
    F_t(x_t)
    :=
    \E_{Y_1\sim q_{1|t}^{\rho}(\cdot|x_t)}\left[f(Y_1)\right]
\end{equation*}
solves the same backward equation and terminal condition by Lemma~\ref{lem:8}. Uniqueness of the classical backward problem gives \(F_t=V_t^\rho[f]\). Since this holds for every sufficiently regular terminal observable, the corresponding terminal transition kernels agree, proving~\eqref{eq:202}.

Every SDE in~\eqref{eq:203} has marginal density \(p_t^\rho\), since its Fokker--Planck equation reduces to the continuity equation. Its generator is
\[
    \mathcal{A}_t^{b^\rho,\sigma}g
    =
    \left(
        v_t^\rho
        +
        \frac{\sigma_t^2}{2}\nabla\log p_t^\rho
    \right)\cdot\nabla g
    +
    \frac{\sigma_t^2}{2}\Delta g.
\]
For \(\sigma_t^2=2\kappa_t\), we have \(\mathcal{A}_t^{b^\rho,\sigma}=\calL_t^\rho\), so the preceding posterior-preservation argument applies.

Conversely, suppose a noise schedule \(\sigma_t\) in~\eqref{eq:203} has conditional values equal to \(V_t^\rho[f]\) for every sufficiently regular terminal observable \(f\). Then \(V_t^\rho[f]\) solves the backward equations associated with both \(\mathcal{A}_t^{b^\rho,\sigma}\) and \(\calL_t^\rho\). Subtracting them gives
\[
    0
    =
    \left(
        \frac{\sigma_t^2}{2}-\kappa_t
    \right)
    \left(
        \Delta V_t^\rho[f]
        +
        \nabla\log p_t^\rho\cdot\nabla V_t^\rho[f]
    \right)
\]
for every such \(f\). Under the condition~\eqref{eq:204}, the second factor is nonzero for at least one pair \((f,x)\) at a.e. $t\in(0,1)$. Therefore, \(\sigma_t^2=2\kappa_t\) for a.e. $t\in(0,1)$, proving the uniqueness claim.
\end{proof}

\subsection{Universal Bridges}
\label{subsec:43}

This subsection establishes the universal bridge property of the posterior-preserving SDE~\eqref{eq:200}: when conditioned on two endpoints, the bridge distribution depends only on the interpolant schedule \((\alpha_t,\beta_t)_{t\in[0,1]}\) and not on the terminal density \(\rho\). Moreover, the bridge can be sampled on any chosen time grid from an explicit Gaussian bridge distribution.

Let $\rho_1,\rho_2\in\pdfspace$ be positive terminal densities. Define the marginal ratio as
\[
    h_t^{\rho_1,\rho_2}(x)
    =
    \frac{p_t^{\rho_2}(x)}{p_t^{\rho_1}(x)}.
\]

\begin{proposition}
\label{prop:31}
For every $t\in(0,1)$,
\begin{equation*}
    h_t^{\rho_1,\rho_2}(x_t)
    =
    V_t^{\rho_1}
    \left[
    \frac{\rho_2}{\rho_1}
    \right](x_t).
\end{equation*}
Consequently,
\begin{equation}
    (\partial_t+\calL_t^{\rho_1})h_t^{\rho_1,\rho_2}=0.
    \label{eq:205}
\end{equation}
\end{proposition}

\begin{proof}
Direct expansion gives
\begin{equation*}
    V_t^{\rho_1}\left[\frac{\rho_2}{\rho_1}\right](x_t)
    =
    \int_{\R^d}
    \frac{\rho_2(x_1)}{\rho_1(x_1)}
    \frac{\rho_1(x_1)p_{t|1}(x_t|x_1)}{p_t^{\rho_1}(x_t)}\dd x_1
    =
    \frac{p_t^{\rho_2}(x_t)}{p_t^{\rho_1}(x_t)}=h_t^{\rho_1,\rho_2}(x_t).
\end{equation*}
Equation~\eqref{eq:205} follows from Theorem~\ref{thm:13}.
\end{proof}

\begin{proposition}%
\label{prop:32}
For $t\in(0,1)$,
\begin{subequations}
    \begin{align}
        &
        v_t^{\rho_2}(x)-v_t^{\rho_1}(x)
        =
        \kappa_t\nabla\log h_t^{\rho_1,\rho_2}(x),
        \label{eq:206}\\
        &b_t^{\rho_2}(x)-b_t^{\rho_1}(x)
        =
        2\kappa_t\nabla\log h_t^{\rho_1,\rho_2}(x).
        \label{eq:207}
    \end{align}
\end{subequations}
\end{proposition}

\begin{proof}
    Substituting
    $v_t^{\rho_i}(x)
    =
    \frac{\dot\alpha_t}{\alpha_t}x
    +
    \kappa_t\nabla\log p_t^{\rho_i}(x)$
    into~\eqref{eq:199} proves the result.
\end{proof}

The following proposition gives the Radon--Nikodym derivative between the path distributions of two posterior-preserving SDEs.

\begin{proposition}
\label{prop:33}
For \(0<t<s\le1\), let $\mathbb P_{[t,s]|t}^{\rho_i}(\cdot|x_t)$ be the path distribution of the posterior-preserving SDE with terminal density $\rho_i$, started from $Y_t=x_t$. Then
\begin{equation}
    \frac{\dd\mathbb P_{[t,s]|t}^{\rho_2}}{\dd\mathbb P_{[t,s]|t}^{\rho_1}}(\BY_{[t,s]})
    =
    \frac{h_s^{\rho_1,\rho_2}(Y_s)}{h_t^{\rho_1,\rho_2}(x_t)}.
    \label{eq:208}
\end{equation}
\end{proposition}

\begin{proof}
    Apply the Doob \(h\)-transform in Lemma~\ref{lem:11} to the posterior-preserving SDE~\eqref{eq:200} of \(\rho_1\) with the positive function \(h_u^{\rho_1,\rho_2}\). By Proposition~\ref{prop:31}, \(h_u^{\rho_1,\rho_2}\) satisfies the required backward equation~\eqref{eq:191}. Moreover, posterior preservation and the Markov property imply that \(\left(\frac{h_u^{\rho_1,\rho_2}(Y_u)}{h_t^{\rho_1,\rho_2}(x_t)}\right)_{u\in[t,s]}\) is a true martingale. By~\eqref{eq:207}, the drift by Doob \(h\)-transform is \(b_u^{\rho_1}+2\kappa_u\nabla\log h_u^{\rho_1,\rho_2}=b_u^{\rho_2}\). Hence, the transformed path distribution is \(\mathbb P_{[t,s]|t}^{\rho_2}(\cdot|x_t)\), and the path-likelihood ratio~\eqref{eq:193} gives~\eqref{eq:208}.
\end{proof}

The Radon--Nikodym derivative between path distributions in~\eqref{eq:208} depends on a realized path only through its two endpoints. This observation yields the following bridge universality.

\begin{theorem}%
\label{thm:14}
For $0<t<s\le1$, let $\mathbb P_{[t,s]|t,s}^{\rho_i}(\cdot|x_t,x_s)$ denote the conditional path distribution of the posterior-preserving SDE of $\rho_i$, given $Y_t=x_t$ and $Y_s=x_s$. Then, we have
\begin{equation}
    \mathbb P_{[t,s]|t,s}^{\rho_2}(\cdot|x_t,x_s)
    =
    \mathbb P_{[t,s]|t,s}^{\rho_1}(\cdot|x_t,x_s).
    \label{eq:209}
\end{equation}
We denote the above universal conditional path distribution by
\[
\mathbb P_{[t,s]}^{\mathrm{uni}}(\cdot|x_t,x_s),
\]
which is independent of the terminal density $\rho$.

Assume that $\gamma_t:=\frac{\beta_t^2}{\alpha_t^2}$ is strictly decreasing on the interval under consideration. For $0<t<u<s\le1$, the universal bridge conditioned on $Y_t=x_t$ and $Y_s=x_s$ satisfies
\begin{equation*}
    Y_u|(Y_t=x_t,Y_s=x_s)
    \sim
    \Normal\left(
    m_{u|t,s}(x_t,x_s),
    \Sigma_{u|t,s}
    \right),
\end{equation*}
where
\begin{align*}
    m_{u|t,s}(x_t,x_s)
    &:=
    \alpha_u
    \left[
    \frac{\gamma_u-\gamma_s}{\gamma_t-\gamma_s}
    \frac{x_t}{\alpha_t}
    +
    \frac{\gamma_t-\gamma_u}{\gamma_t-\gamma_s}
    \frac{x_s}{\alpha_s}
    \right],\qquad
    \Sigma_{u|t,s}
    :=
    \alpha_u^2
    \frac{(\gamma_t-\gamma_u)(\gamma_u-\gamma_s)}{\gamma_t-\gamma_s}I.
\end{align*}
\end{theorem}

\begin{proof}
The Radon--Nikodym derivative in~\eqref{eq:208} is constant after conditioning on both endpoints. It therefore cancels in Bayes' rule for every bounded path functional, which proves~\eqref{eq:209}.

Consider the bridge under the following linear reference process:
\begin{equation*}
    \dd Y_r
    =
    \frac{\dot\alpha_r}{\alpha_r}Y_r\dd r
    +
    \sqrt{2\kappa_r}\dd W_r.
\end{equation*}
Set $Z_r=\frac{Y_r}{\alpha_r}$. Then
\begin{equation*}
    \dd Z_r
    =
    \frac{\sqrt{2\kappa_r}}{\alpha_r}\dd W_r,
    \qquad
    \dot\gamma_r
    =
    -\frac{2\kappa_r}{\alpha_r^2}.
\end{equation*}
Thus $Z_r$ is a Brownian motion in the decreasing clock $\gamma_r$. The result is the standard Brownian-bridge conditional distribution in that clock, multiplied by $\alpha_u$.
\end{proof}

According to Theorem~\ref{thm:14}, once an endpoint pair \((x_t,x_s)\) has been obtained---either from a rollout of the posterior-preserving SDE or from the forward construction---the same endpoints can be reused to sample arbitrarily many conditionally independent bridge trajectories $\BY_{[t,s]}=(Y_{u})_{u\in[t,s]}$ without rerunning the model-dependent SDE.

The universal bridge provides the state path, whereas the stochastic-integral terms in the SDE-based estimation of the covariance form are expressed in terms of the Brownian motion driving the posterior-preserving SDE before endpoint conditioning. Since the diffusion coefficient is state-independent, the required noise increments can be reconstructed from a bridge $\BY_{[t,s]}=(Y_u)_{u\in[t,s]}$
relative to the drift $b^\rho$:
\[
    \dd\widetilde W_u
    :=
    \frac{\dd Y_u-b_u^\rho(Y_u)\dd u}{\sqrt{2\kappa_u}},
    \qquad u\in[t,s].
\]
Under the universal bridge law
$\mathbb P_{[t,s]}^{\mathrm{uni}}(\cdot|x_t,x_s)$,
the reconstructed process $\widetilde W$ has the conditional law of the original Brownian driver given the two endpoints. In general, it is not a Brownian motion under the bridge measure.

\subsection{Log Density and Log-Density Ratio}
\label{subsec:44}

This subsection develops the SDE-based computational approaches for the log density \(\log\rho\) and the log-density ratio \(\log\frac{\rho_2}{\rho_1}\), which are used in the exact Newton Matching.

\begin{proposition}%
\label{prop:34}
The marginal log-density ratio satisfies
\begin{subequations}
    \begin{align}
    &
    (\partial_t+\calL_t^{\rho_1})
    \log h_t^{\rho_1,\rho_2}(x)
    =
    -\frac{1}{\kappa_t}
    \norm[2]{v_t^{\rho_2}(x)-v_t^{\rho_1}(x)}^2,
    \label{eq:210}\\
    &
    (\partial_t+\calL_t^{\rho_2})
    \log h_t^{\rho_1,\rho_2}(x)
    =
    \frac{1}{\kappa_t}
    \norm[2]{v_t^{\rho_2}(x)-v_t^{\rho_1}(x)}^2.
    \label{eq:211}
    \end{align}
\end{subequations}
\end{proposition}

\begin{proof}
The logarithmic chain rule for \(\calL_t^{\rho_1}=b_t^{\rho_1}\cdot\nabla+\kappa_t\Delta\), together with~\eqref{eq:205}, gives
\[
\begin{aligned}
    (\partial_t+\calL_t^{\rho_1})\log h_t^{\rho_1,\rho_2}
    &=
    \frac{1}{h_t^{\rho_1,\rho_2}}
    (\partial_t+\calL_t^{\rho_1})h_t^{\rho_1,\rho_2}
    -
    \kappa_t\norm[2]{\nabla\log h_t^{\rho_1,\rho_2}}^2
    =
    -\kappa_t\norm[2]{\nabla\log h_t^{\rho_1,\rho_2}}^2.
\end{aligned}
\]
Then,~\eqref{eq:206} implies~\eqref{eq:210}. Moreover, substituting \(f=\log h_t^{\rho_1,\rho_2}\) into
\[
    \calL_t^{\rho_2}f
    =
    \calL_t^{\rho_1}f
    +
    2\kappa_t\nabla\log h_t^{\rho_1,\rho_2}\cdot\nabla f
\]
and using~\eqref{eq:210} give~\eqref{eq:211}.
\end{proof}

\begin{proposition}%
\label{prop:35}
Let $(Y_s^{\rho_i})_{s\in[t,1]}$ follow the $\rho_i$ posterior-preserving SDE, started from $Y_t^{\rho_i}=x_t$ and driven by $W^{\rho_i}$, for $i\in\{1,2\}$. Along the $\rho_2$ path,
\begin{align}
    \log\frac{\rho_2(Y_1^{\rho_2})}{\rho_1(Y_1^{\rho_2})}
    =&
    \log h_t^{\rho_1,\rho_2}(x_t)
    +
    \int_t^1
    \frac{\norm[2]{v_s^{\rho_2}(Y_s^{\rho_2})-v_s^{\rho_1}(Y_s^{\rho_2})}^2}{\kappa_s}\dd s
    \notag\\
    &+
    \int_t^1
    \sqrt{\frac{2}{\kappa_s}}
    \left(
    v_s^{\rho_2}(Y_s^{\rho_2})-v_s^{\rho_1}(Y_s^{\rho_2})
    \right)^\top\dd W_s^{\rho_2}.
    \label{eq:212}
\end{align}
Along the $\rho_1$ path,
\begin{align}
    \log\frac{\rho_2(Y_1^{\rho_1})}{\rho_1(Y_1^{\rho_1})}
    =&
    \log h_t^{\rho_1,\rho_2}(x_t)
    -
    \int_t^1
    \frac{\norm[2]{v_s^{\rho_2}(Y_s^{\rho_1})-v_s^{\rho_1}(Y_s^{\rho_1})}^2}{\kappa_s}\dd s
    \notag\\
    &+
    \int_t^1
    \sqrt{\frac{2}{\kappa_s}}
    \left(
    v_s^{\rho_2}(Y_s^{\rho_1})-v_s^{\rho_1}(Y_s^{\rho_1})
    \right)^\top\dd W_s^{\rho_1}.
    \label{eq:213}
\end{align}
\end{proposition}

\begin{proof}
Apply It\^o's formula in Lemma~\ref{lem:9} to \(F_s(x)=\log h_s^{\rho_1,\rho_2}(x)\). Along the posterior-preserving SDE of \(\rho_2\),~\eqref{eq:211} gives
\[
\begin{aligned}
    \dd\log h_s^{\rho_1,\rho_2}(Y_s^{\rho_2})
    =
    \frac{1}{\kappa_s}
    \norm[2]{
        v_s^{\rho_2}(Y_s^{\rho_2})
        -
        v_s^{\rho_1}(Y_s^{\rho_2})
    }^2\dd s
    +
    \sqrt{2\kappa_s}
    \nabla\log h_s^{\rho_1,\rho_2}(Y_s^{\rho_2})^\top\dd W_s^{\rho_2}.
\end{aligned}
\]
By~\eqref{eq:206}, the integrand in the stochastic term can be written as
\[
    \sqrt{2\kappa_s}
    \nabla\log h_s^{\rho_1,\rho_2}(Y_s^{\rho_2})
    =
    \sqrt{\frac{2}{\kappa_s}}
    \left(
        v_s^{\rho_2}(Y_s^{\rho_2})
        -
        v_s^{\rho_1}(Y_s^{\rho_2})
    \right).
\]
Substituting this identity, integrating from \(t\) to \(1\), and using \(h_1^{\rho_1,\rho_2}=\frac{\rho_2}{\rho_1}\) proves~\eqref{eq:212}.

Along the posterior-preserving SDE of \(\rho_1\), the same calculation uses~\eqref{eq:210} and gives
\[
\begin{aligned}
    \dd\log h_s^{\rho_1,\rho_2}(Y_s^{\rho_1})
    =&
    -\frac{1}{\kappa_s}
    \norm[2]{
        v_s^{\rho_2}(Y_s^{\rho_1})
        -
        v_s^{\rho_1}(Y_s^{\rho_1})
    }^2\dd s
    +
    \sqrt{\frac{2}{\kappa_s}}
    \left(
        v_s^{\rho_2}(Y_s^{\rho_1})
        -
        v_s^{\rho_1}(Y_s^{\rho_1})
    \right)^\top\dd W_s^{\rho_1}.
\end{aligned}
\]
Integration proves~\eqref{eq:213}.
\end{proof}

\begin{proposition}%
\label{prop:36}
In the setting of Proposition~\ref{prop:35}, we have
\begin{align}
    V_t^{\rho_2}
    \left[
    \log\frac{\rho_2}{\rho_1}
    \right](x_t)
    =&
    \log h_t^{\rho_1,\rho_2}(x_t)
    +
    \E_{\BY_{[t,1]}\sim
    \mathbb P_{[t,1]|t}^{{\rho_2}}(\cdot|x_t)}
    \left[
    \int_t^1
    \frac{\norm[2]{v_s^{\rho_2}(Y_s)-v_s^{\rho_1}(Y_s)}^2}{\kappa_s}\dd s
    \right],
    \label{eq:214}\\
    V_t^{\rho_1}
    \left[
    \log\frac{\rho_2}{\rho_1}
    \right](x_t)
    =&
    \log h_t^{\rho_1,\rho_2}(x_t)
    -
    \E_{\BY_{[t,1]}\sim
    \mathbb P_{[t,1]|t}^{{\rho_1}}(\cdot|x_t)}
    \left[
    \int_t^1
    \frac{\norm[2]{v_s^{\rho_2}(Y_s)-v_s^{\rho_1}(Y_s)}^2}{\kappa_s}\dd s
    \right].\notag
\end{align}
\end{proposition}

\begin{proof}
Take conditional expectations in~\eqref{eq:212} and~\eqref{eq:213}. By the martingale assumptions, the stochastic integrals have conditional-mean zero. Theorem~\ref{thm:13} identifies the endpoint expectations with the corresponding posterior values, which finishes the proof.
\end{proof}

Recall from Definitions~\ref{def:5} and~\ref{def:6} that every density $\rho\in\pdfspace$ has finite second moment. In this class, we can represent the KL divergence between marginal densities by the difference of canonical velocity fields. Lemma~\ref{lem:12} and Corollary~\ref{cor:8} below are stated for the standard flow-matching setting; in the one-sided interpolant setting, the condition $\lim_{\varepsilon\downarrow0}\KL{p_\varepsilon^{\rho_2}}{p_\varepsilon^{\rho_1}}=0$ in Corollary~\ref{cor:8} holds automatically.

\begin{lemma}\label{lem:12}
    For every $\rho_1,\rho_2\in\pdfspace$, the $W_2$ distance is finite
    \begin{equation*}
        W_2(\rho_1,\rho_2):=\sqrt{\inf_{\gamma\in\Pi(\rho_1,\rho_2)}\E_{(X,Y)\sim\gamma}\left[\norm[2]{X-Y}^2\right]}<\infty,
    \end{equation*}
    where $\Pi(\rho_1,\rho_2)$ is the set of couplings of $\rho_1$ and $\rho_2$. 
    
    Moreover, for every $t\in[0,1)$, we have
    \begin{equation}\label{eq:215}
    \KL{p_t^{\rho_1}}{p_t^{\rho_2}}\leq\frac{\alpha_t^2}{2\beta_t^2}W_2(\rho_1,\rho_2)^2.
    \end{equation}
\end{lemma}

\begin{proof}
    Applying the independent coupling $\rho_1(x)\rho_2(y)\,\dd x\,\dd y$ gives
    \begin{align*}
        W_2(\rho_1,\rho_2)^2
        \leq\E_{X\sim\rho_1,Y\sim\rho_2}\left[\norm[2]{X-Y}^2\right]
        \leq2\E_{X\sim\rho_1,Y\sim\rho_2}\left[\norm[2]{X}^2+\norm[2]{Y}^2\right]<\infty.
    \end{align*}

    For every $\gamma\in\Pi(\rho_1,\rho_2)$ and $t\in[0,1)$, the joint convexity of relative entropy gives
    \begin{align*}
        \KL{p_t^{\rho_1}}{p_t^{\rho_2}}
        =&\KL{\E_{(X,Y)\sim\gamma}\left[p_{t|1}(\cdot|X)\right]}{\E_{(X,Y)\sim\gamma}\left[p_{t|1}(\cdot|Y)\right]}\\
        \leq&\E_{(X,Y)\sim\gamma}\left[\KL{p_{t|1}(\cdot|X)}{p_{t|1}(\cdot|Y)}\right]\\
        =&\E_{(X,Y)\sim\gamma}\left[\KL{\Normal(\alpha_tX,\beta_t^2I)}{\Normal(\alpha_tY,\beta_t^2I)}\right]\\
        =&\frac{\alpha_t^2}{2\beta_t^2}\E_{(X,Y)\sim\gamma}\left[\norm[2]{X-Y}^2\right].
    \end{align*}
    Therefore,~\eqref{eq:215} holds.
\end{proof}

\begin{corollary}\label{cor:8}
    For every $\rho_1,\rho_2\in\pdfspace$, we have
\begin{equation}
    \KL{\rho_2}{\rho_1}
    =
    \int_0^1
    \frac{1}{\kappa_s}
    \E_{Y_s\sim p_s^{\rho_2}}
    \left[
    \norm[2]{v_s^{\rho_2}(Y_s)-v_s^{\rho_1}(Y_s)}^2
    \right]\dd s.
    \label{eq:216}
\end{equation}
\end{corollary}

\begin{proof}
To derive the endpoint KL identity under a potentially singular schedule, fix \(\varepsilon\in(0,1)\), apply~\eqref{eq:214} at \(t=\varepsilon\), and average over \(X_\varepsilon\sim p_\varepsilon^{\rho_2}\). Posterior preservation gives
\[
    \E_{X_\varepsilon\sim p_\varepsilon^{\rho_2}}
    \left[
        V_\varepsilon^{\rho_2}
        \left[
            \log\frac{\rho_2}{\rho_1}
        \right](X_\varepsilon)
    \right]
    =
    \KL{\rho_2}{\rho_1},
\]
whereas
\[
    \E_{X_\varepsilon\sim p_\varepsilon^{\rho_2}}
    \left[
        \log h_\varepsilon^{\rho_1,\rho_2}(X_\varepsilon)
    \right]
    =
    \KL{p_\varepsilon^{\rho_2}}{p_\varepsilon^{\rho_1}}.
\]
Using Tonelli's theorem and the fact that the posterior-preserving SDE of \(\rho_2\) has marginal \(p_s^{\rho_2}\), we obtain
\[
\begin{aligned}
    \KL{\rho_2}{\rho_1}
    =&
    \KL{p_\varepsilon^{\rho_2}}{p_\varepsilon^{\rho_1}}
    +
    \int_\varepsilon^1
    \frac{1}{\kappa_s}
    \E_{Y_s\sim p_s^{\rho_2}}
    \left[
        \norm[2]{v_s^{\rho_2}(Y_s)-v_s^{\rho_1}(Y_s)}^2
    \right]\dd s.
\end{aligned}
\]
By Lemma~\ref{lem:12}, \(\KL{p_\varepsilon^{\rho_2}}{p_\varepsilon^{\rho_1}}\to0\) as \(\varepsilon\downarrow0\). Since the integrand in the energy term is nonnegative, the monotone convergence theorem allows the lower integration limit to pass to zero. Taking \(\varepsilon\downarrow0\) in the preceding identity therefore proves~\eqref{eq:216}. If the coefficients extend regularly to \(t=0\), the same argument can be written directly by setting \(t=0\) and averaging over \(X_0\sim p_0\).
\end{proof}

\begin{proposition}%
\label{prop:37}
Let $(Y_s^\rho)_{s\in[t,1]}$ follow the posterior-preserving SDE with terminal density $\rho$, started from $Y_t^\rho=x_t$. Then
\begin{align}
    \log\rho(Y_1^\rho)
    =&
    \log\left(\alpha_t^d p_t^\rho(x_t)\right)
    +
    \int_t^1
    \frac{1}{\kappa_s}
    \norm[2]{v_s^\rho(Y_s^\rho)-\frac{\dot\alpha_s}{\alpha_s}Y_s^\rho}^2\dd s
    \notag\\
    &+
    \int_t^1
    \sqrt{\frac{2}{\kappa_s}}
    \left(
    v_s^\rho(Y_s^\rho)-\frac{\dot\alpha_s}{\alpha_s}Y_s^\rho
    \right)^\top\dd W_s^\rho.
    \label{eq:217}
\end{align}
\end{proposition}

\begin{proof}
Apply It\^o's formula to \(F_s(x)=\log p_s^\rho(x)\) under the posterior-preserving SDE of \(\rho\). Its generator is
\[
    \calL_s^\rho f
    =
    \left(
        \frac{\dot\alpha_s}{\alpha_s}x
        +
        2\kappa_s\nabla\log p_s^\rho
    \right)\cdot\nabla f
    +
    \kappa_s\Delta f.
\]
By~\eqref{eq:198},
\[
    \partial_s p_s^\rho
    +
    \nabla\cdot\left(
        p_s^\rho\frac{\dot\alpha_s}{\alpha_s}x
    \right)
    +
    \kappa_s\Delta p_s^\rho
    =0.
\]
Dividing by \(p_s^\rho\) and using
\[
    \frac{\Delta p_s^\rho}{p_s^\rho}
    =
    \Delta\log p_s^\rho
    +
    \norm[2]{\nabla\log p_s^\rho}^2
\]
gives
\[
    (\partial_s+\calL_s^\rho)\log p_s^\rho(x)
    =
    -d\frac{\dot\alpha_s}{\alpha_s}
    +
    \kappa_s\norm[2]{\nabla\log p_s^\rho(x)}^2.
\]
Since \(p_1^\rho=\rho\), It\^o's formula from \(t\) to \(1\) yields the score-form identity
\[
\begin{aligned}
    \log\rho(Y_1^\rho)
    =&
    \log p_t^\rho(x_t)
    +
    \int_t^1
    \left(
        \kappa_s\norm[2]{\nabla\log p_s^\rho(Y_s^\rho)}^2
        -
        d\frac{\dot\alpha_s}{\alpha_s}
    \right)\dd s+
    \int_t^1
    \sqrt{2\kappa_s}
    \nabla\log p_s^\rho(Y_s^\rho)^\top\dd W_s^\rho.
\end{aligned}
\]
Finally,
$
    v_s^\rho(x)-\frac{\dot\alpha_s}{\alpha_s}x
    =
    \kappa_s\nabla\log p_s^\rho(x)
$
and
$
    \int_t^1
    -d\frac{\dot\alpha_s}{\alpha_s}\dd s
    =
    d\log\alpha_t
$
give~\eqref{eq:217}.
\end{proof}

\begin{proposition}%
\label{prop:38}
In the setting of Proposition~\ref{prop:37},
\begin{align}
    V_t^\rho[\log\rho](x_t)
    =&
    \log\left(\alpha_t^d p_t^\rho(x_t)\right)
    +
    \E_{\BY_{[t,1]}\sim
    \mathbb P_{[t,1]|t}^{\rho}(\cdot|x_t)}
    \left[
    \int_t^1
    \frac{1}{\kappa_s}
    \norm[2]{v_s^\rho(Y_s)-\frac{\dot\alpha_s}{\alpha_s}Y_s}^2\dd s
    \right].
    \label{eq:218}
\end{align}
\end{proposition}

\begin{proof}
Take conditional expectation in~\eqref{eq:217}. The stochastic integral has conditional-mean zero by the standing assumptions, and Theorem~\ref{thm:13} identifies
\[
    \E_{Y_1\sim q_{1|t}^{\rho}(\cdot|x_t)}\left[
        \log\rho(Y_1)
    \right]=
    \E_{Y_1\sim p_{1|t}^{\rho}(\cdot|x_t)}\left[
        \log\rho(Y_1)
    \right]
    =
    V_t^\rho[\log\rho](x_t),
\]
which gives~\eqref{eq:218}.
\end{proof}

\subsection{Initial-State Sensitivity and Adjoint Calculus}\label{subsec:45}

This subsection develops pathwise and endpoint-conditioned sensitivity for generic SDEs, together with pathwise and conditional-mean adjoint identities.

\subsubsection{Pathwise Adjoint}\label{sub2sec:32}

We first develop the sensitivity calculus for a generic SDE \citep{yong1999stochastic}. The pathwise state-transition matrix propagates infinitesimal perturbations of the initial state along a realized path. The transposed state-transition matrices pull the state derivatives of the terminal and running objectives back to the initial time and aggregate their contributions; the same quantity can equivalently be computed by a backward adjoint. These pathwise constructions are well-defined for any sufficiently regular path and drift; they coincide with actual initial-state derivatives when the path is generated by the same drift used in the sensitivity equation.

Fix $t\in[0,1)$ and an initial state $x_t\in\R^d$. Recall the SDE in~\eqref{eq:186}:
\begin{align*}
    &\dd Y_{s}^{b,\sigma}(t,x_t)
    =
    b_s\left(Y_{s}^{b,\sigma}(t,x_t)\right)\dd s
    +
    \sigma_s\dd W_s,
    \qquad s\in[t,1],\\
    &Y_{t}^{b,\sigma}(t,x_t)=x_t.
\end{align*}

\begin{definition}
\label{def:9}
For a continuous path $\BY_{[t,1]}=(Y_s)_{s\in[t,1]}$, define the \textit{pathwise state-transition matrix} $G_{t\to s}^b(\BY_{[t,1]})\in\R^{d\times d}$ by the forward variational ODE
\begin{subequations}
\label{eq:219}
    \begin{align}
        &\frac{\dd}{\dd s}G_{t\to s}^b(\BY_{[t,1]})
        =
        \nabla b_s(Y_s)G_{t\to s}^b(\BY_{[t,1]}),
        \qquad s\in[t,1],\\
        &G_{t\to t}^b(\BY_{[t,1]})=I.\label{eq:220}
    \end{align}
\end{subequations}
\end{definition}

Definition~\ref{def:9} does not require that the path be generated by~\eqref{eq:186}. The matrix $G_{t\to s}^b(\BY_{[t,1]})$ is defined for any sufficiently regular path $\BY_{[t,1]}$ and drift $b$. The following lemma shows that, when the path is generated by~\eqref{eq:186} with the same drift, the pathwise state-transition matrix is the actual derivative of the SDE path with respect to its initial state.

\begin{lemma}
The following identity holds a.s. simultaneously for every $s\in[t,1]$:
\begin{equation}
    \nabla_{x_t}Y_{s}^{b,\sigma}(t,x_t)
    =
    G_{t\to s}^b
    \left(
    \BY_{[t,1]}^{b,\sigma}(t,x_t)
    \right).
    \label{eq:221}
\end{equation}
\end{lemma}

\begin{proof}
The integral form of~\eqref{eq:186} has been given in~\eqref{eq:188}, i.e.,
\begin{equation*}
    Y_{s}^{b,\sigma}(t,x_t)
    =
    x_t
    +
    \int_t^s b_u\left(Y_{u}^{b,\sigma}(t,x_t)\right)\dd u
    +
    \int_t^s\sigma_u\dd W_u.
\end{equation*}
Recall that, when considering initial-state sensitivities, the same Brownian motion is used. The stochastic integral $\int_t^s\sigma_u\dd W_u$ is independent of $x_t$. Differentiating~\eqref{eq:188} with respect to $x_t$ gives
\begin{equation*}
    \nabla_{x_t}Y_{s}^{b,\sigma}(t,x_t)
    =
    I
    +
    \int_t^s
    \nabla b_u\left(Y_{u}^{b,\sigma}(t,x_t)\right)
    \nabla_{x_t}Y_{u}^{b,\sigma}(t,x_t)
    \dd u\quad
    \mathrm{a.s.}
\end{equation*}
This is exactly the integral form of~\eqref{eq:219} along the path $\BY_{[t,1]}^{b,\sigma}(t,x_t)$, which proves~\eqref{eq:221}.
\end{proof}

Consider the pathwise objective
\begin{equation}
    J_t^{\varphi,l}(\BY_{[t,1]})
    :=
    \varphi(Y_1)
    +
    \int_t^1 l_s(Y_s)\dd s,
    \label{eq:222}
\end{equation}
where the terminal objective $\varphi:\R^d\to\R$ and the running objective $l_s:\R^d\to\R$ are $C^1$ and satisfy the regularity conditions in Appendix~\ref{sub2sec:30}.

When applying the objective~\eqref{eq:222} to SDE~\eqref{eq:186}, the initial-state sensitivity of the objective $J_t^{\varphi,l}$ can be evaluated by the adjoint.

\begin{lemma}
\label{lem:13}
For a continuous path $\BY_{[t,1]}=(Y_s)_{s\in[t,1]}$, define the pathwise adjoint by
\begin{subequations}
\label{eq:223}
    \begin{align}
        &\frac{\dd}{\dd s}\lambda_s^{b,\varphi,l}(\BY_{[t,1]})
        =
        -\nabla b_s(Y_s)^\top
        \lambda_s^{b,\varphi,l}(\BY_{[t,1]})
        -
        \nabla l_s(Y_s),
        \qquad s\in[t,1],\\
        &\lambda_1^{b,\varphi,l}(\BY_{[t,1]})
        =
        \nabla\varphi(Y_1).\label{eq:224}
    \end{align}
\end{subequations}
Then, we have
\begin{align}
    \lambda_t^{b,\varphi,l}(\BY_{[t,1]})
    =&
    G_{t\to1}^b(\BY_{[t,1]})^\top\nabla\varphi(Y_1)
    +
    \int_t^1
    G_{t\to s}^b(\BY_{[t,1]})^\top
    \nabla l_s(Y_s)\dd s.
    \label{eq:225}
\end{align}
If the path is generated by~\eqref{eq:186}, i.e., $\BY_{[t,1]}=\BY_{[t,1]}^{b,\sigma}(t,x_t)$, then the pathwise adjoint is the initial-state sensitivity of the objective $J_t^{\varphi,l}$:
\begin{align}
    \lambda_t^{b,\varphi,l}
    \left(
    \BY_{[t,1]}^{b,\sigma}(t,x_t)
    \right)
    &=
    \nabla_{x_t}
    J_t^{\varphi,l}
    \left(
    \BY_{[t,1]}^{b,\sigma}(t,x_t)
    \right).
    \label{eq:226}
\end{align}
\end{lemma}

\begin{proof}
By~\eqref{eq:219} and~\eqref{eq:223}, we have
\begin{align*}
    &\frac{\dd}{\dd s}
    \left(
    G_{t\to s}^b(\BY_{[t,1]})^\top
    \lambda_s^{b,\varphi,l}(\BY_{[t,1]})
    \right)
    \\
    =&
    \left(
    \nabla b_s(Y_s)G_{t\to s}^b(\BY_{[t,1]})
    \right)^\top
    \lambda_s^{b,\varphi,l}(\BY_{[t,1]})
    -
    G_{t\to s}^b(\BY_{[t,1]})^\top
    \left(
    \nabla b_s(Y_s)^\top
    \lambda_s^{b,\varphi,l}(\BY_{[t,1]})
    +
    \nabla l_s(Y_s)
    \right)
    \\
    =&
    -G_{t\to s}^b(\BY_{[t,1]})^\top\nabla l_s(Y_s).
\end{align*}
Integrating from $t$ to $1$ gives
\begin{align*}
    &G_{t\to1}^b(\BY_{[t,1]})^\top
    \lambda_1^{b,\varphi,l}(\BY_{[t,1]})
    -
    G_{t\to t}^b(\BY_{[t,1]})^\top
    \lambda_t^{b,\varphi,l}(\BY_{[t,1]})
    =
    -\int_t^1
    G_{t\to s}^b(\BY_{[t,1]})^\top
    \nabla l_s(Y_s)\dd s.
\end{align*}
Then,~\eqref{eq:224} and~\eqref{eq:220} prove~\eqref{eq:225}. If the path is generated by~\eqref{eq:186}, the chain rule together with~\eqref{eq:221} gives~\eqref{eq:226}.
\end{proof}

We next characterize the conditional mean of a pathwise adjoint under a possibly different path-generating SDE.

\begin{proposition}\label{prop:39}
    Consider another SDE
    \begin{subequations}\label{eq:227}
        \begin{align}
            &\dd Y_{s}^{\hat{b},\hat\sigma}(t,x_t)
            =
            {\hat{b}}_s\left(Y_{s}^{\hat{b},\hat\sigma}(t,x_t)\right)\dd s
            +
            \hat\sigma_s\dd W_s,
            \qquad s\in[t,1],
            \\
            &Y_{t}^{\hat{b},\hat\sigma}(t,x_t)=x_t.
        \end{align}
    \end{subequations}
    whose conditional path distribution and generator are $\mathbb P_{[t,1]|t}^{\hat{b},\hat{\sigma}}(\cdot|x_t)$ and $\mathcal{A}_t^{\hat{b},\hat\sigma}$, respectively.
    Define the conditional-mean adjoint by
    \[
    \bar\lambda_t(x_t)
    :=
    \E_{\BY_{[t,1]}\sim
    \mathbb P_{[t,1]|t}^{\hat{b},\hat{\sigma}}(\cdot|x_t)}
    \left[
    \lambda_t^{b,\varphi,l}(\BY_{[t,1]})
    \right],
    \]
    Under the standing regularity and uniqueness assumptions, $\bar\lambda_t$ is the unique classical solution of
    \begin{align*}
        &\left(\partial_t+\mathcal{A}_t^{\hat{b},\hat\sigma}\right)\bar\lambda_t(x)+\nabla b_t(x)^\top\bar\lambda_t(x)+\nabla l_t(x)=0,\qquad t\in(0,1),\\
        &\bar\lambda_1(x)=\nabla\varphi(x).
    \end{align*}
\end{proposition}

\begin{proof}
    For simplicity, denote $\lambda_t:=\lambda_t^{b,\varphi,l}$. Then, we have the pathwise identity
    \[
    \lambda_t(\BY_{[t,1]})=\lambda_{s}(\BY_{[t,1]})+\int_{t}^{s}\left(\nabla b_u(Y_u)^\top\lambda_u(\BY_{[t,1]})+\nabla l_u(Y_u)\right)\dd u.
    \]
    Let $Y^{\hat{b},\hat\sigma}$ be the trajectory of the SDE~\eqref{eq:227}. By the Markov property of
    $Y^{\hat{b},\hat\sigma}$, taking the expectation gives
    \begin{align*}
    \bar{\lambda}_t(x)
    =&\E_{Y_s\sim q_{s|t}^{\hat{b},\hat\sigma}(\cdot|x)}\left[\bar\lambda_{s}\left(Y_{s}\right)\right]
    +\E_{\BY_{[t,s]}\sim\mathbb{P}_{[t,s]|t}^{\hat{b},\hat{\sigma}}(\cdot|x)}\left[\int_{t}^{s}\left(\nabla b_u\left(Y_u\right)^\top\bar\lambda_{u}\left(Y_u\right)+\nabla l_u\left(Y_u\right)\right)\dd u\right].
    \end{align*}
    Applying Dynkin's formula in Lemma~\ref{lem:10} componentwise to the vector-valued function $\bar{\lambda}_s$ gives
    \[
    \E_{Y_s\sim q_{s|t}^{\hat{b},\hat\sigma}(\cdot|x)}\left[\bar\lambda_{s}\left(Y_{s}\right)\right]=\bar{\lambda}_t(x)
    +\E_{\BY_{[t,s]}\sim\mathbb{P}_{[t,s]|t}^{\hat{b},\hat{\sigma}}(\cdot|x)}\left[\int_{t}^{s}\left(\partial_u+\mathcal{A}_u^{\hat{b},\hat\sigma}\right)\bar{\lambda}_u\left(Y_u\right)\dd u\right].
    \]
    Therefore, for every $0<t<s<1$ and $x\in\R^d$, we have
    \[
    \E_{\BY_{[t,s]}\sim\mathbb{P}_{[t,s]|t}^{\hat{b},\hat{\sigma}}(\cdot|x)}\left[\int_{t}^{s}\left(\left(\partial_u+\mathcal{A}_u^{\hat{b},\hat\sigma}\right)\bar{\lambda}_u\left(Y_u\right)+\nabla b_u\left(Y_u\right)^\top\bar\lambda_{u}\left(Y_u\right)+\nabla l_u\left(Y_u\right)\right)\dd u\right]=0.
    \]
    Taking $s\downarrow t$ finishes the proof.
\end{proof}

\subsubsection{Endpoint-Conditioned Sensitivity}\label{sub2sec:33}

The preceding sensitivity objects depend on the entire realized path. For later endpoint-based construction of sensitivity Stein kernels in Appendix~\ref{subsec:49}, we instead need a deterministic sensitivity operator that depends only on the endpoint pair $(x_t,x_1)$. We therefore average the terminal state-transition matrix over the conditional path distribution given the two endpoints.

\begin{definition}
\label{def:10}
For the SDE~\eqref{eq:186}, let $\mathbb P_{[t,1]|t,1}^{b,\sigma}(\cdot|x_t,x_1)$ denote the conditional path distribution given $Y_t=x_t$ and $Y_1=x_1$. Define the \textit{endpoint-conditioned state-transition matrix} by
\begin{equation*}
    K_{1|t}^{b,\sigma}(x_1|x_t)
    :=
    \E_{\BY_{[t,1]}\sim
    \mathbb P_{[t,1]|t,1}^{b,\sigma}(\cdot|x_t,x_1)}
    \left[
    G_{t\to1}^b(\BY_{[t,1]})
    \right].
\end{equation*}
\end{definition}

The following proposition converts pathwise sensitivity into an endpoint-conditioned expectation. Its terminal-only specialization is used in Appendix~\ref{subsec:49} to construct the posterior sensitivity kernel, whereas the generator-form extension supports the Bolza realization with running objectives in Section~\ref{subsec:17}.

\begin{proposition}
\label{prop:40}
We have
\begin{align}
    &\nabla_{x_t}
    \E_{\BY_{[t,1]}\sim
    \mathbb P_{[t,1]|t}^{b,\sigma}(\cdot|x_t)}
    \left[
    J_t^{\varphi,l}
    \left(
    \BY_{[t,1]}
    \right)
    \right]
    =
    \E_{\BY_{[t,1]}\sim
    \mathbb P_{[t,1]|t}^{b,\sigma}(\cdot|x_t)}
    \left[
    \lambda_t^{b,\varphi,l}
    \left(
    \BY_{[t,1]}
    \right)
    \right].
    \label{eq:228}
\end{align}
Suppose additionally that the running objective takes the form
\begin{equation*}
    l_s=(\partial_s+\mathcal A_s^{b,\sigma})h_s,
\end{equation*}
where $\mathcal A_s^{b,\sigma}$ is the generator of~\eqref{eq:186}, defined in~\eqref{eq:189}, and $h:[t,1]\times\R^d\to\R$ satisfies the regularity assumptions in Appendix~\ref{sub2sec:30}. Then,
\begin{align}
    &\nabla_{x_t}
    \E_{\BY_{[t,1]}\sim
    \mathbb P_{[t,1]|t}^{b,\sigma}(\cdot|x_t)}
    \left[
    J_t^{\varphi,l}(\BY_{[t,1]})
    \right]
    =
    \E_{Y_1\sim q_{1|t}^{b,\sigma}(\cdot|x_t)}
    \left[
    K_{1|t}^{b,\sigma}(Y_1|x_t)^\top
    \nabla(\varphi+h_1)(Y_1)
    \right]
    -
    \nabla h_t(x_t).
    \label{eq:229}
\end{align}
In particular, if $l_s=0$, then
\begin{align}
    &\nabla_{x_t}
    \E_{\BY_{[t,1]}\sim
    \mathbb P_{[t,1]|t}^{b,\sigma}(\cdot|x_t)}
    \left[
    \varphi\left(Y_{1}\right)
    \right]
    =
    \E_{Y_1\sim q_{1|t}^{b,\sigma}(\cdot|x_t)}
    \left[
    K_{1|t}^{b,\sigma}(Y_1|x_t)^\top
    \nabla\varphi(Y_1)
    \right].
    \label{eq:230}
\end{align}
\end{proposition}

\begin{proof}
Differentiate under the expectation and apply Lemma~\ref{lem:13} to obtain~\eqref{eq:228}. If $l_s=(\partial_s+\mathcal A_s^{b,\sigma})h_s$, Dynkin's formula in Lemma~\ref{lem:10} gives
\begin{align*}
    \E_{\BY_{[t,1]}\sim
    \mathbb P_{[t,1]|t}^{b,\sigma}(\cdot|x_t)}
    \left[
    \int_t^1
    (\partial_s+\mathcal A_s^{b,\sigma})h_s(Y_s)\dd s
    \right]
    =
    \E_{Y_1\sim q_{1|t}^{b,\sigma}(\cdot|x_t)}[h_1(Y_1)]
    -
    h_t(x_t).
\end{align*}
Consequently,~\eqref{eq:229} is given by
\begin{align*}
    &\nabla_{x_t}
    \E_{\BY_{[t,1]}\sim
    \mathbb P_{[t,1]|t}^{b,\sigma}(\cdot|x_t)}
    \left[
    J_t^{\varphi,l}(\BY_{[t,1]})
    \right]
    +
    \nabla h_t(x_t)
    \\
    =&
    \E_{\BY_{[t,1]}\sim
    \mathbb P_{[t,1]|t}^{b,\sigma}(\cdot|x_t)}
    \left[
    G_{t\to1}^b(\BY_{[t,1]})^\top
    \nabla(\varphi+h_1)(Y_1)
    \right]
    \\
    =&
    \E_{Y_1\sim q_{1|t}^{b,\sigma}(\cdot|x_t)}
    \left[
    \E_{\BY_{[t,1]}\sim
    \mathbb P_{[t,1]|t,1}^{b,\sigma}(\cdot|x_t,Y_1)}
    \left[
    G_{t\to1}^b(\BY_{[t,1]})
    \right]^\top
    \nabla(\varphi+h_1)(Y_1)
    \right]
    \\
    =&
    \E_{Y_1\sim q_{1|t}^{b,\sigma}(\cdot|x_t)}
    \left[
    K_{1|t}^{b,\sigma}(Y_1|x_t)^\top
    \nabla(\varphi+h_1)(Y_1)
    \right].
\end{align*}
Taking $l_s=0$ and $h_s=0$ gives~\eqref{eq:230}.
\end{proof}

Therefore, we have obtained the endpoint-conditioned sensitivity for SDE~\eqref{eq:186}. Appendix~\ref{subsec:49} specializes this operator to the posterior-preserving SDE and verifies that a schedule-dependent rescaling produces an exact posterior Stein kernel, which is called the posterior sensitivity kernel.

\section{Posterior Stein Kernels}\label{app:5}

This appendix develops the posterior Stein kernel, which provides the mechanism that converts the covariance-form target~\eqref{eq:56} to the gradient-form target~\eqref{eq:59}. We prove that, for every sufficiently regular terminal observable $f\in C^1(\mathbb{R}^d;\mathbb{R})$, a posterior Stein kernel $\Lambda_t^\rho(\cdot | x_t):\R^d\to \R^{d\times d}$ yields the covariance--gradient identity
\[
\operatorname{Cov}_{X_1 \sim p^\rho_{1|t}(\cdot | x_t)}\bigl(v_{t|1}(x_t | X_1), f(X_1)\bigr) = \frac{\alpha_t\kappa_t}{\beta_t^2}\mathbb{E}_{X_1 \sim p^\rho_{1|t}(\cdot | x_t)}\left[\Lambda_t^\rho(X_1 | x_t)^\top \nabla f(X_1)\right].
\]
For Newton Matching, setting \(f=\tilde r^\rho\) in the covariance--gradient identity yields covariance-form and gradient-form sample-wise targets with the same conditional mean and hence the same population minimizer under squared-loss regression. Their difference is a zero-mean control variate, so any matrix-weighted affine combination of the two remains population-exact. When the required conditional second moments exist, the matrix weight can be chosen to minimize the variance.

Appendix~\ref{subsec:46} first introduces the Stein identity for the Langevin Stein operator. Appendix~\ref{subsec:47} then uses this identity to construct a family of zero-mean Stein control variates. By specializing the construction to the posterior $p_{1|t}^\rho(\cdot|x_t)$, Appendix~\ref{subsec:48} defines posterior Stein kernels and proves the covariance--gradient identity displayed above. Building on the adjoint calculus in Appendix~\ref{subsec:45}, Appendix~\ref{subsec:49} constructs the posterior sensitivity kernel and its adjoint-based evaluation used by exact Newton Matching in Section~\ref{sec:5}. Finally, Appendix~\ref{subsec:50} records the Gaussian Stein kernels and posterior-moment formulas used by the Gaussian-kernel approximation in Section~\ref{sec:6}.

As in Appendix~\ref{app:4}, all conclusions in this appendix apply to both the standard flow matching construction and the one-sided interpolant setting. The extension of Newton Matching to one-sided interpolants is discussed in Appendix~\ref{subsec:57}.

\subsection{Langevin Stein Identities}\label{subsec:46}

In this subsection, we introduce the Langevin Stein operator \citep{gorham2015measuring,li2026reverse} and its columnwise extension to matrix-valued test fields. The Stein identity provides the zero-mean property used to construct control variates, while the product rule, combined with the defining relation of a posterior Stein kernel, yields the covariance--gradient identity.

\begin{definition}
\label{def:11}
Let $p$ be a positive $C^1$ density on $\R^d$, and let $\phi:\R^d\to\R^d$ be a continuously differentiable vector field. The \textit{Langevin Stein operator} associated with $p$ is
\begin{equation*}
    \Stein_p\phi:\R^d\to\R,
    \qquad
    (\Stein_p\phi)(x)
    :=
    \nabla\cdot\phi(x)
    +
    \phi(x)^\top\nabla\log p(x).
\end{equation*}

Let $\Phi:\R^d\to\R^{d\times d}$ be $C^1$, and write $\Phi=[\phi_1,\ldots,\phi_d]$, where each column $\phi_j:\R^d\to\R^d$ is $C^1$. Define the Langevin Stein operator for the matrix-valued test field $\Phi$ by applying $\Stein_p$ columnwise:
\begin{equation*}
    \Stein_p\Phi:\R^d\to\R^d,
    \qquad
    (\Stein_p\Phi)(x)
    :=
    \left[\begin{matrix}
        (\Stein_p\phi_1)(x)\\
        \vdots\\
        (\Stein_p\phi_d)(x)
    \end{matrix}\right]
    =
    \nabla\cdot\Phi(x)
    +
    \Phi(x)^\top\nabla\log p(x),
\end{equation*}
where $\nabla\cdot\Phi(x)\in\R^d$ denotes the columnwise divergence:
\begin{equation*}
    \bigl(\nabla\cdot\Phi(x)\bigr)_j
    =
    \sum_{i=1}^d
    \frac{\partial\Phi_{ij}(x)}{\partial x_i},
    \qquad j=1,\ldots,d.
\end{equation*}
\end{definition}

Under the appropriate integrability and boundary conditions, the Langevin Stein operator yields the following zero-mean property, which is called the Stein identity.

\begin{lemma}%
\label{lem:14}
Use the notation in Definition~\ref{def:11}. Assume that
\begin{equation}
    \E_{X\sim p}\left[\abs{(\Stein_p\phi)(X)}\right]<\infty,
    \label{eq:231}
\end{equation}
and that the boundary term vanishes:
\begin{equation}
    \lim_{R\to\infty}
    \int_{\partial B_R}
    p(x)\phi(x)^\top n(x)\dd S(x)
    =0,
    \label{eq:232}
\end{equation}
where $B_R:=\{x:\norm[2]{x}\le R\}$ and $n(x)$ is the outward unit normal vector on $\partial B_R$. Then
\begin{equation}
    \E_{X\sim p}\left[(\Stein_p\phi)(X)\right]=0.
    \label{eq:233}
\end{equation}
Consequently, if every column of $\Phi:\R^d\to\R^{d\times d}$ satisfies~\eqref{eq:231} and~\eqref{eq:232}, then
\begin{equation}
    \E_{X\sim p}\left[(\Stein_p\Phi)(X)\right]
    =0\in\R^d.
    \label{eq:234}
\end{equation}
\end{lemma}

\begin{proof}
Since $p$ is positive and continuously differentiable,
\begin{align*}
    p(x)(\Stein_p\phi)(x)
    &=
    p(x)\nabla\cdot\phi(x)
    +
    \phi(x)^\top\nabla p(x)
    =
    \nabla\cdot\left(p(x)\phi(x)\right).
\end{align*}
Integrating over $B_R$ and applying the divergence theorem gives
\begin{equation*}
    \int_{B_R}(\Stein_p\phi)(x)p(x)\dd x
    =
    \int_{\partial B_R}p(x)\phi(x)^\top n(x)\dd S(x).
\end{equation*}
By~\eqref{eq:231} and~\eqref{eq:232}, taking $R\to\infty$ proves~\eqref{eq:233}. Applying~\eqref{eq:233} to every column of $\Phi$ gives~\eqref{eq:234}.
\end{proof}

In all subsequent applications, the densities and test fields are assumed to satisfy the integrability and boundary conditions in~\eqref{eq:231} and~\eqref{eq:232}. Under these conditions, the Langevin Stein operator applied to an admissible test field has zero expectation and can therefore serve as a control variate. We next record the product rule for the Langevin Stein operator.

\begin{lemma}%
\label{lem:15}
Let $f:\R^d\to\R$ be $C^1$. For a vector field $\phi:\R^d\to\R^d$,
\begin{equation}
    \Stein_p(f\phi)(x)
    =
    \phi(x)^\top\nabla f(x)
    +
    f(x)(\Stein_p\phi)(x).
    \label{eq:235}
\end{equation}
For a matrix-valued test function $\Phi:\R^d\to\R^{d\times d}$,
\begin{equation*}
    \Stein_p(f\Phi)(x)
    =
    \Phi(x)^\top\nabla f(x)
    +
    f(x)(\Stein_p\Phi)(x).
\end{equation*}
\end{lemma}

\begin{proof}
For the vector-field identity,
\begin{align*}
    \Stein_p(f\phi)(x)
    &=
    \nabla\cdot(f(x)\phi(x))
    +
    f(x)\phi(x)^\top\nabla\log p(x)
    \\
    &=
    \phi(x)^\top\nabla f(x)
    +
    f(x)\nabla\cdot\phi(x)
    +
    f(x)\phi(x)^\top\nabla\log p(x)
    \\
    &=
    \phi(x)^\top\nabla f(x)
    +
    f(x)(\Stein_p\phi)(x).
\end{align*}
The matrix-valued identity follows by applying~\eqref{eq:235} to every column of $\Phi$.
\end{proof}

\subsection{Stein Control Variates}\label{subsec:47}

Appendix~\ref{subsec:46} provides a class of zero-mean control variates~\eqref{eq:234} through the Langevin Stein operator. In this subsection, we provide a variance-reduction result: the optimal linear coefficient in the Loewner order.

\begin{proposition}
\label{prop:41}
Consider a density $p\in\pdfspace$ and an integrable function $\psi:\R^d\to\R^d$ which satisfies
\begin{equation*}
    \E_{X\sim p}\left[\norm[2]{\psi(X)}\right]<\infty.
\end{equation*}
Let $\Phi:\R^d\to\R^{d\times d}$ be a regular matrix field whose columns satisfy~\eqref{eq:231} and~\eqref{eq:232}. Then, for every coefficient $\Omega\in\R^{d\times d}$, we have
\begin{equation}
    \E_{X\sim p}[\psi(X)]
    =
    \E_{X\sim p}
    \left[
    \psi(X)+\Omega(\Stein_p\Phi)(X)
    \right].
    \label{eq:236}
\end{equation}

Assume additionally that
\begin{equation*}
    \E_{X\sim p}\left[\norm[2]{\psi(X)}^2\right]<\infty,
    \qquad
    \E_{X\sim p}\left[\norm[2]{(\Stein_p\Phi)(X)}^2\right]<\infty.
\end{equation*}
Then
\begin{equation*}
    \Omega^\star
    =
    -\Cov_{X\sim p}
    \left(
    \psi(X),(\Stein_p\Phi)(X)
    \right)
    \Var_{X\sim p}
    \left((\Stein_p\Phi)(X)\right)^\dagger
\end{equation*}
minimizes the variance in the Loewner order:
\begin{align}
    &\Var_{X\sim p}
    \left(
    \psi(X)+\Omega(\Stein_p\Phi)(X)
    \right)
    -
    \Var_{X\sim p}
    \left(
    \psi(X)+\Omega^\star(\Stein_p\Phi)(X)
    \right)
    \in\S_+^d
    \label{eq:237}
\end{align}
for every $\Omega\in\R^{d\times d}$. Here ${}^\dagger$ denotes the Moore--Penrose pseudoinverse.
\end{proposition}

\begin{proof}
Equation~\eqref{eq:236} follows directly from Lemma~\ref{lem:14}. Denote
\begin{equation*}
    \Sigma
    :=
    \Var_{X\sim p}(\psi(X)),
    \qquad
    A
    :=
    \Var_{X\sim p}((\Stein_p\Phi)(X)),
    \qquad
    B
    :=
    \Cov_{X\sim p}(\psi(X),(\Stein_p\Phi)(X)).
\end{equation*}
Then
\begin{equation}
    \Var_{X\sim p}
    \left(
    \psi(X)+\Omega(\Stein_p\Phi)(X)
    \right)
    =
    \Sigma+\Omega B^\top+B\Omega^\top+\Omega A\Omega^\top.
    \label{eq:238}
\end{equation}

We next prove $B=BA^\dagger A$. It suffices to show $\ker(A)\subseteq\ker(B)$. For any $u\in\ker(A)$,
\begin{equation*}
    \Var_{X\sim p}
    \left(
    u^\top(\Stein_p\Phi)(X)
    \right)
    =
    u^\top Au
    =0.
\end{equation*}
Since the Stein control variate has zero mean, we have $u^\top(\Stein_p\Phi)(X)=0$ a.s. under $X\sim p$. Therefore,
\begin{equation*}
    Bu
    =
    \Cov_{X\sim p}
    \left(
    \psi(X),u^\top(\Stein_p\Phi)(X)
    \right)
    =0,
\end{equation*}
which proves $\ker(A)\subseteq\ker(B)$ and hence
\begin{equation}
    B=BA^\dagger A.
    \label{eq:239}
\end{equation}
Combining~\eqref{eq:238} and~\eqref{eq:239} gives
\begin{align*}
    &\Var_{X\sim p}
    \left(
    \psi(X)+\Omega(\Stein_p\Phi)(X)
    \right)
    =
    \Sigma-BA^\dagger B^\top
    +
    (\Omega+BA^\dagger)A(\Omega+BA^\dagger)^\top,
\end{align*}
which proves~\eqref{eq:237}.
\end{proof}

Using Proposition~\ref{prop:41}, Appendix~\ref{subsec:48} applies this construction to the posterior density and combines it with posterior Stein kernels.

\subsection{Covariance--Gradient Identity of Posterior Stein Kernels}\label{subsec:48}

Recall from Proposition~\ref{prop:2} that, in Newton Matching, the posterior covariance
\begin{equation}
    \Cov_{X_1\sim p_{1|t}^\rho(\cdot|x_t)}
    \left(X_1,f(X_1)\right)
    =
    \E_{X_1\sim p_{1|t}^\rho(\cdot|x_t)}
    \left[
    f(X_1)\left(X_1-M_t^\rho(x_t)\right)
    \right]
    \label{eq:240}
\end{equation}
is used to represent canonical tangent directions, where
\begin{equation*}
    M_t^\rho(x_t)
    :=
    \E_{X_1\sim p_{1|t}^\rho(\cdot|x_t)}[X_1]
\end{equation*}
is the posterior mean. Note that $v_{t|1}(x_t|X_1)$ is affine in $X_1$. Using the results in Appendices~\ref{subsec:46} and~\ref{subsec:47}, we equip the posterior $p_{1|t}^\rho(\cdot|x_t)$ with a Stein kernel \citep{courtade2019existence}. Then, the resulting posterior Stein kernel transforms~\eqref{eq:240} into the gradient form
\begin{equation*}
    \E_{X_1\sim p_{1|t}^\rho(\cdot|x_t)}
    \left[
    \Lambda_t^\rho(X_1|x_t)^\top\nabla f(X_1)
    \right].
\end{equation*}

\begin{definition}
\label{def:12}
Consider a terminal density $\rho\in\pdfspace$. For any $t\in(0,1)$ and $x_t\in\R^d$, a matrix-valued $C^1$ function $\Lambda_t^\rho(\cdot|x_t):\R^d\to\R^{d\times d}$ is called a posterior Stein kernel for $p_{1|t}^\rho(\cdot|x_t)$ if its columns satisfy~\eqref{eq:231}--\eqref{eq:232} and for any $x_1\in\R^d$,
\begin{equation}
    \left(
    \Stein_{p_{1|t}^\rho(\cdot|x_t)}
    \Lambda_t^\rho(\cdot|x_t)
    \right)(x_1)
    =
    -\left(x_1-M_t^\rho(x_t)\right).
    \label{eq:241}
\end{equation}
Equivalently,
\begin{equation}
    \nabla_{x_1}\cdot
    \left(
    p_{1|t}^\rho(x_1|x_t)
    \Lambda_t^\rho(x_1|x_t)
    \right)
    =
    -p_{1|t}^\rho(x_1|x_t)
    \left(x_1-M_t^\rho(x_t)\right).
    \label{eq:242}
\end{equation}
\end{definition}

Note that~\eqref{eq:242} need not determine a unique matrix field. Nevertheless, every admissible solution yields the same covariance--gradient identity below. Appendix~\ref{subsec:49} later constructs a distinguished solution by specializing the endpoint-conditioned state-transition matrix developed in Appendix~\ref{subsec:45} to the posterior-preserving SDE.

\begin{theorem}
\label{thm:15}
Let $\Lambda_t^\rho(\cdot|x_t)$ be a posterior Stein kernel for $p_{1|t}^\rho(\cdot|x_t)$, and let $f:\R^d\to\R$ be $C^1$. Assume that every column of $f\Lambda_t^\rho(\cdot|x_t)$ satisfies~\eqref{eq:231}--\eqref{eq:232}. Then, we have the covariance--gradient identity
\begin{equation}
    \Cov_{X_1\sim p_{1|t}^\rho(\cdot|x_t)}
    \left(X_1,f(X_1)\right)
    =
    \E_{X_1\sim p_{1|t}^\rho(\cdot|x_t)}
    \left[
    \Lambda_t^\rho(X_1|x_t)^\top\nabla f(X_1)
    \right].
    \label{eq:243}
\end{equation}
\end{theorem}

\begin{proof}
The Stein identity in Lemma~\ref{lem:14} gives
\begin{equation*}
    \E_{X_1\sim p_{1|t}^\rho(\cdot|x_t)}
    \left[
    \left(
    \Stein_{p_{1|t}^\rho(\cdot|x_t)}
    \left(f\Lambda_t^\rho(\cdot|x_t)\right)
    \right)(X_1)
    \right]
    =0.
\end{equation*}
The product rule in Lemma~\ref{lem:15} implies
\begin{align*}
    &\E_{X_1\sim p_{1|t}^\rho(\cdot|x_t)}
    \left[
    \Lambda_t^\rho(X_1|x_t)^\top\nabla f(X_1)
    \right]
    =
    -\E_{X_1\sim p_{1|t}^\rho(\cdot|x_t)}
    \left[
    f(X_1)
    \left(
    \Stein_{p_{1|t}^\rho(\cdot|x_t)}
    \Lambda_t^\rho(\cdot|x_t)
    \right)(X_1)
    \right].
\end{align*}
Since $\Lambda_t^\rho(\cdot|x_t)$ is a posterior Stein kernel,~\eqref{eq:241} implies~\eqref{eq:243}.
\end{proof}

For a posterior Stein kernel, the product rule gives an exact affine family interpolating between the covariance and gradient forms. Specifically, for every matrix field $\Omega_t(x_t)\in\R^{d\times d}$, we have
\begin{align*}
    &\Cov_{X_1\sim p_{1|t}^\rho(\cdot|x_t)}
    \left(X_1,f(X_1)\right)
    \notag\\
    =&
    \E_{X_1\sim p_{1|t}^\rho(\cdot|x_t)}
    \left[
    f(X_1)\left(X_1-M_t^\rho(x_t)\right)
    +
    \Omega_t(x_t)
    \left(
    \Stein_{p_{1|t}^\rho(\cdot|x_t)}
    \left(f\Lambda_t^\rho(\cdot|x_t)\right)
    \right)(X_1)
    \right]
    \notag\\
    =&
    \E_{X_1\sim p_{1|t}^\rho(\cdot|x_t)}
    \left[
    \left(I-\Omega_t(x_t)\right)
    f(X_1)\left(X_1-M_t^\rho(x_t)\right)
    +
    \Omega_t(x_t)
    \Lambda_t^\rho(X_1|x_t)^\top\nabla f(X_1)
    \right],\notag
\end{align*}
which is a specialization of~\eqref{eq:236}. Thus, the posterior covariance can be estimated by any affine combination of the covariance and gradient representations. Proposition~\ref{prop:41} gives the corresponding optimal linear coefficient whenever the required conditional second moments exist.

\subsection{Posterior Sensitivity Kernel}\label{subsec:49}

This subsection specializes the endpoint-conditioned sensitivity construction in Appendix~\ref{sub2sec:33} to the posterior-preserving SDE. Rescaling the endpoint-conditioned state-transition matrix in Definition~\ref{def:10} by \(\beta_t^2/\alpha_t\) yields the posterior sensitivity kernel, which satisfies the defining equation of a posterior Stein kernel. Its action on an endpoint gradient can be evaluated via the backward adjoint ODE.

For a terminal density $\rho\in\pdfspace$, recall the posterior-preserving SDE from Definition~\ref{def:8}:
\begin{subequations}
\label{eq:244}
    \begin{align}
        &\dd Y_{s}^{\rho}(t,x_t)
        =
        b_s^\rho\left(Y_{s}^{\rho}(t,x_t)\right)\dd s
        +
        \sqrt{2\kappa_s}\dd W_s,
        \qquad s\in[t,1],\\
        &Y_{t}^{\rho}(t,x_t)=x_t,
    \end{align}
\end{subequations}
where
\begin{equation*}
    b_s^\rho(x)
    =
    2v_s^\rho(x)
    -
    \frac{\dot\alpha_s}{\alpha_s}x,
    \qquad
    v^\rho=\canonicalmap(\rho)\in\velmancan.
\end{equation*}
For brevity, write $\mathbb P^\rho:=\mathbb P^{b^\rho,\sqrt{2\kappa}}$ and $K_{1|t}^\rho:=K_{1|t}^{b^\rho,\sqrt{2\kappa}}$ for the corresponding path distribution and endpoint-conditioned state-transition matrix.

\begin{theorem}
\label{thm:16}
Define
\begin{equation}
    \Lambda_t^\rho(x_1|x_t)
    :=
    \frac{\beta_t^2}{\alpha_t}
    K_{1|t}^{\rho}(x_1|x_t).
    \label{eq:245}
\end{equation}
Then $\Lambda_t^\rho(\cdot|x_t)$ is a posterior Stein kernel for $p_{1|t}^\rho(\cdot|x_t)$, which we call the posterior sensitivity kernel. In particular, for every regular test function $f:\R^d\to\R$ such that the columns of $f\Lambda_t^\rho(\cdot|x_t)$ satisfy the integrability and boundary conditions in~\eqref{eq:231}--\eqref{eq:232}, we have
\begin{equation}
    \Cov_{X_1\sim p_{1|t}^\rho(\cdot|x_t)}
    \left(X_1,f(X_1)\right)
    =
    \E_{X_1\sim p_{1|t}^\rho(\cdot|x_t)}
    \left[
    \Lambda_t^\rho(X_1|x_t)^\top
    \nabla f(X_1)
    \right].
    \label{eq:246}
\end{equation}
\end{theorem}

\begin{proof}
The required integrability and boundary conditions are included in the standing regularity assumptions. We therefore focus on verifying the defining equation~\eqref{eq:242}.

By Theorem~\ref{thm:13}, the terminal transition density of~\eqref{eq:244} is the posterior $p_{1|t}^\rho(\cdot|x_t)$. Therefore, for every compactly supported $C^1$ terminal test function $f$,
\begin{align*}
    \E_{X_1\sim p_{1|t}^\rho(\cdot|x_t)}[f(X_1)]
    &=
    \E_{\BY_{[t,1]}\sim
    \mathbb P_{[t,1]|t}^{\rho}(\cdot|x_t)}
    \left[
    J_t^{f,0}
    \left(
    \BY_{[t,1]}
    \right)
    \right].
\end{align*}
Taking the gradient with respect to $x_t$ and applying Proposition~\ref{prop:40} gives
\begin{align*}
    \int_{\R^d}
    f(x_1)
    \nabla_{x_t}p_{1|t}^\rho(x_1|x_t)
    \dd x_1
    =&
    \E_{X_1\sim p_{1|t}^\rho(\cdot|x_t)}
    \left[
    K_{1|t}^{\rho}(X_1|x_t)^\top
    \nabla f(X_1)
    \right]
    \notag\\
    =&
    \int_{\R^d}
    K_{1|t}^{\rho}(x_1|x_t)^\top
    \nabla f(x_1)
    p_{1|t}^\rho(x_1|x_t)
    \dd x_1.
\end{align*}

If $f$ is supported in a ball $B_R$, then the relevant terms are integrable under the standing regularity assumptions. In particular,
    \begin{align*}
        &\int_{\R^d}
        \norm[2]{
        K_{1|t}^{\rho}(x_1|x_t)^\top
        \nabla f(x_1)
        p_{1|t}^\rho(x_1|x_t)
        }
        \dd x_1
        <\infty,\\
        &\int_{\R^d}
        \norm[2]{
        f(x_1)
        \nabla_{x_1}\cdot
        \left(
        p_{1|t}^\rho(x_1|x_t)
        K_{1|t}^{\rho}(x_1|x_t)
        \right)
        }
        \dd x_1
        <\infty.
    \end{align*}
Integration by parts therefore gives
\begin{align*}
    \int_{\R^d}
    f(x_1)
    \nabla_{x_t}p_{1|t}^\rho(x_1|x_t)
    \dd x_1
    =&
    \int_{\R^d}
    K_{1|t}^{\rho}(x_1|x_t)^\top
    \nabla f(x_1)
    p_{1|t}^\rho(x_1|x_t)
    \dd x_1\\
    =&
    -\int_{\R^d}
    f(x_1)
    \nabla_{x_1}\cdot
    \left(
    p_{1|t}^\rho(x_1|x_t)
    K_{1|t}^{\rho}(x_1|x_t)
    \right)
    \dd x_1.
\end{align*}
By the arbitrariness of compactly supported $f$, we have
\begin{equation}
    \nabla_{x_t}p_{1|t}^\rho(x_1|x_t)
    =
    -\nabla_{x_1}\cdot
    \left(
    p_{1|t}^\rho(x_1|x_t)
    K_{1|t}^{\rho}(x_1|x_t)
    \right).
    \label{eq:247}
\end{equation}

The posterior score satisfies
\begin{equation*}
    \nabla_{x_t}
    \log p_{1|t}^\rho(x_1|x_t)
    =
    \frac{\alpha_t}{\beta_t^2}
    \left(
    x_1-M_t^\rho(x_t)
    \right).
\end{equation*}
Multiplying~\eqref{eq:247} by $\frac{\beta_t^2}{\alpha_t}$ and using~\eqref{eq:245} gives
\begin{equation*}
    \nabla_{x_1}\cdot
    \left(
    p_{1|t}^\rho(x_1|x_t)
    \Lambda_t^\rho(x_1|x_t)
    \right)
    =
    -p_{1|t}^\rho(x_1|x_t)
    \left(
    x_1-M_t^\rho(x_t)
    \right).
\end{equation*}
This is exactly~\eqref{eq:242}. Hence, $\Lambda_t^\rho$ is a posterior Stein kernel by Definition~\ref{def:12}, and~\eqref{eq:246} follows from Theorem~\ref{thm:15}.
\end{proof}

In practice, we do not explicitly construct the posterior sensitivity kernel $\Lambda_t^\rho$. Instead, Lemma~\ref{lem:13} and Proposition~\ref{prop:40} allow its action on an endpoint gradient to be evaluated through a pathwise adjoint.

\begin{proposition}\label{prop:42}
    For every $t\in(0,1)$ and $x_t\in\R^d$, we have
    \[
    \Lambda_t^\rho(x_1|x_t)^\top
    \nabla f(x_1)
    =
    \frac{\beta_t^2}{\alpha_t}\E_{\BY_{[t,1]}\sim
    \mathbb P_{[t,1]}^{\mathrm{uni}}(\cdot|x_t,x_1)}
    \left[
    \lambda_t^{b^\rho,f,0}(\BY_{[t,1]})
    \right],
    \]
    where the pathwise adjoint $\lambda_t^{b^\rho,f,0}(\BY_{[t,1]})$ satisfies the ODE~\eqref{eq:223}, and the bridge distribution $\mathbb P_{[t,1]}^{\mathrm{uni}}(\cdot|x_t,x_1)$ is defined in Theorem~\ref{thm:14}.
\end{proposition}

\begin{proof}
    By~\eqref{eq:245} and Definition~\ref{def:10}, we have
    \[
    \Lambda_t^\rho(x_1|x_t)^\top\nabla f(x_1)
    =\frac{\beta_t^2}{\alpha_t}
    \E_{\BY_{[t,1]}\sim
    \mathbb P_{[t,1]|t,1}^\rho(\cdot|x_t,x_1)}
    \left[
    G_{t\to1}^{b^\rho}(\BY_{[t,1]})^\top\nabla f(x_1)
    \right].
    \]
    Then,~\eqref{eq:225} implies 
    \[
    \Lambda_t^\rho(x_1|x_t)^\top\nabla f(x_1)
    =\frac{\beta_t^2}{\alpha_t}
    \E_{\BY_{[t,1]}\sim
    \mathbb P_{[t,1]|t,1}^\rho(\cdot|x_t,x_1)}
    \left[
    \lambda_t^{b^\rho,f,0}(\BY_{[t,1]})
    \right].
    \]
    Finally, Theorem~\ref{thm:14} finishes the proof.
\end{proof}

Proposition~\ref{prop:42} gives
\begin{align}
    &\frac{\alpha_t\kappa_t}{\beta_t^2}
    \E_{X_1\sim p_{1|t}^\rho(\cdot|x_t)}
    \left[
    \Lambda_t^\rho(X_1|x_t)^\top
    \nabla f(X_1)
    \right]
    \notag\\
    =&
    \kappa_t
    \E_{\BY_{[t,1]}\sim
    \mathbb P_{[t,1]|t}^{\rho}(\cdot|x_t)}
    \left[
    \lambda_t^{b^\rho,f,0}
    \left(
    \BY_{[t,1]}
    \right)
    \right]
    \notag\\
    =&
    \kappa_t
    \E_{X_1\sim p_{1|t}^\rho(\cdot|x_t)}
    \left[
    \E_{\BY_{[t,1]}\sim
    \mathbb P_{[t,1]|t,1}^{\rho}(\cdot|x_t,X_1)}
    \left[
    \lambda_t^{b^\rho,f,0}(\BY_{[t,1]})
    \right]
    \right].
    \label{eq:248}
\end{align}
Equation~\eqref{eq:248} admits two exact path-sampling realizations. Under the reverse construction, the full path is sampled by simulating the posterior-preserving SDE from \(Y_t=x_t\). Under the forward construction, the endpoint pair \((X_t,X_1)\) is generated first and then completed by sampling the universal bridge in Theorem~\ref{thm:14}. In either case, the pathwise adjoint is obtained by solving the ODE~\eqref{eq:223} along the sampled path.

\subsection{Gaussian Stein Kernels and Posterior Moments}\label{subsec:50}

The Gaussian-kernel approximation in Section~\ref{sub2sec:15} uses the covariance matrix of a Gaussian surrogate in place of the exact posterior sensitivity kernel constructed above. This subsection first records the Gaussian Stein identity, which identifies this covariance matrix as an exact Stein kernel for the surrogate distribution, and then derives exact formulas for the mean and covariance of the interpolant posterior that can be used to specify the surrogate.

Let $X\sim\Normal(m,\Sigma)$, where $m\in\R^d$ and $\Sigma\in\S_{++}^d$. For every sufficiently regular function $f:\R^d\to\R$, the Gaussian Stein identity states that
\begin{equation*}
    \E_{X\sim\Normal(m,\Sigma)}
    \left[(X-m)f(X)\right]
    =
    \Sigma\,
    \E_{X\sim\Normal(m,\Sigma)}
    \left[\nabla f(X)\right].
\end{equation*}
Equivalently, the constant matrix field $x\mapsto\Sigma$ is a Stein kernel for $\Normal(m,\Sigma)$. This property is specific to Gaussian distributions: for a non-Gaussian posterior, its covariance matrix is generally not a posterior Stein kernel.

This observation motivates approximating the posterior by a Gaussian for computational convenience and using the corresponding Stein kernel. Consider a Gaussian surrogate
\begin{equation*}
    \hat p_{1|t}^\rho(\cdot|x_t)
    :=
    \Normal
    \left(
    \widehat M_t^\rho(x_t),
    \widehat\Sigma_t^\rho(x_t)
    \right),
\end{equation*}
where $\widehat M_t^\rho(x_t)\in\R^d$ and $\widehat\Sigma_t^\rho(x_t)\in\S_{++}^d$ may be supplied by a trained model, estimated via statistical tools, or specified analytically. Its score is given by
\begin{equation*}
    \nabla_{x_1}\log\hat p_{1|t}^\rho(x_1|x_t)
    =
    -\widehat\Sigma_t^\rho(x_t)^{-1}
    \left(x_1-\widehat M_t^\rho(x_t)\right).
\end{equation*}
Hence, the matrix field
\begin{equation*}
    \widehat\Lambda_t^\rho(x_1|x_t)
    :=
    \widehat\Sigma_t^\rho(x_t)
\end{equation*}
is a Stein kernel for the surrogate posterior $\hat p_{1|t}^\rho(\cdot|x_t)$, since
\begin{align*}
    &\left(
    \Stein_{\hat p_{1|t}^\rho(\cdot|x_t)}
    \widehat\Lambda_t^\rho(\cdot|x_t)
    \right)(x_1)
    =
    \nabla_{x_1}\cdot\widehat\Sigma_t^\rho(x_t)
    +
    \widehat\Sigma_t^\rho(x_t)^\top
    \nabla_{x_1}\log\hat p_{1|t}^\rho(x_1|x_t)
    =
    -\left(x_1-\widehat M_t^\rho(x_t)\right).
\end{align*}
Therefore, we have
\begin{equation*}
    \Cov_{X_1\sim\hat p_{1|t}^\rho(\cdot|x_t)}
    \left(X_1,f(X_1)\right)
    =
    \widehat\Sigma_t^\rho(x_t)
    \E_{X_1\sim\hat p_{1|t}^\rho(\cdot|x_t)}
    \left[\nabla f(X_1)\right].
\end{equation*}
A full plug-in approximation consequently takes the form
\begin{equation*}
    \Cov_{X_1\sim p_{1|t}^\rho(\cdot|x_t)}
    \left(X_1,f(X_1)\right)
    \approx
    \widehat\Sigma_t^\rho(x_t)
    \E_{X_1\sim\hat p_{1|t}^\rho(\cdot|x_t)}
    \left[\nabla f(X_1)\right].
\end{equation*}
By contrast, the Gaussian-kernel approximation in Section~\ref{sub2sec:15} retains the exact posterior density for endpoint sampling and replaces only the exact posterior Stein kernel by
\(\widehat\Sigma_t^\rho(x_t)\). It therefore uses
\begin{equation*}
    \Cov_{X_1\sim p_{1|t}^\rho(\cdot|x_t)}
    \left(X_1,f(X_1)\right)
    \approx
    \widehat\Sigma_t^\rho(x_t)
    \E_{X_1\sim p_{1|t}^\rho(\cdot|x_t)}
    \left[\nabla f(X_1)\right].
\end{equation*}
Retaining the exact posterior density is required for the critical-point consistency. A full plug-in approximation introduces an additional source of inexactness and is generally not critical-point consistent, although it may reduce computational cost and remain useful in practice.

The exact posterior moments provide natural choices for the parameters of the Gaussian surrogate. We first consider the mean. The posterior mean admits the equivalent representations
\begin{subequations}\label{eq:249}
\begin{align}
    M_t^\rho(x_t)
    &=
    \frac{x_t}{\alpha_t}
    +
    \frac{\beta_t^2}{\alpha_t}
    \nabla\log p_t^\rho(x_t)
    \label{eq:250}\\
    &=
    \frac{\beta_t^2}{\alpha_t\kappa_t}
    \left(
        v_t^\rho(x_t)
        -
        \frac{\dot\beta_t}{\beta_t}x_t
    \right).
    \label{eq:251}
\end{align}
\end{subequations}
Accordingly, the surrogate mean can be chosen to match the exact posterior mean:
\begin{equation*}
    \widehat M_t^\rho(x_t)
    :=
    M_t^\rho(x_t)
    =
    \frac{\beta_t^2}{\alpha_t\kappa_t}
    \left(
        v_t^\rho(x_t)
        -
        \frac{\dot\beta_t}{\beta_t}x_t
    \right).
\end{equation*}
For fixed \((t,x_t)\), this choice requires only a single evaluation of the canonical anchor \(v_t^\rho\).

The covariance matrix of the exact posterior admits equivalent representations in terms of the Jacobian of the canonical velocity and the Hessian of \(\log p_t^\rho(x_t)\). Define
\begin{equation*}
    \Sigma_t^\rho(x_t)
    :=
    \Cov_{X_1\sim p_{1|t}^\rho(\cdot|x_t)}
    \left(X_1,X_1\right).
\end{equation*}
Then
\begin{subequations}\label{eq:252}
    \begin{align}
    \Sigma_t^\rho(x_t)
    &=
    \frac{\beta_t^4}{\alpha_t^2\kappa_t}
    \left[
        \nabla v_t^\rho(x_t)
        -
        \frac{\dot\beta_t}{\beta_t}I
    \right]
    \label{eq:253}\\
    &=
    \frac{\beta_t^2}{\alpha_t^2}I
    +
    \frac{\beta_t^4}{\alpha_t^2}
    \nabla^2\log p_t^\rho(x_t).
    \label{eq:254}
    \end{align}
\end{subequations}

To obtain these formulas, we start from the definition of the posterior mean:
\begin{equation*}
    M_t^\rho(x_t)
    =
    \int_{\R^d}
    x_1 p_{1|t}^\rho(x_1|x_t)\,\dd x_1.
\end{equation*}
Under the standing regularity assumptions, differentiation with respect to \(x_t\) may be interchanged with integration. Hence,
\begin{align*}
    \nabla M_t^\rho(x_t)
    &=
    \int_{\R^d}
    x_1
    \left(
        \nabla_{x_t}p_{1|t}^\rho(x_1|x_t)
    \right)^\top
    \dd x_1\\
    &=
    \E_{X_1\sim p_{1|t}^\rho(\cdot|x_t)}
    \left[
        X_1
        \nabla_{x_t}\log
        p_{1|t}^\rho(X_1|x_t)^\top
    \right]\\
    &=
    \frac{\alpha_t}{\beta_t^2}
    \E_{X_1\sim p_{1|t}^\rho(\cdot|x_t)}
    \left[
        X_1
        \left(
            X_1-M_t^\rho(x_t)
        \right)^\top
    \right]\\
    &=
    \frac{\alpha_t}{\beta_t^2}
    \Sigma_t^\rho(x_t).
\end{align*}
Therefore,
\begin{equation*}
    \Sigma_t^\rho(x_t)
    =
    \frac{\beta_t^2}{\alpha_t}
    \nabla M_t^\rho(x_t).
\end{equation*}
Differentiating the two representations of \(M_t^\rho\) in~\eqref{eq:249} with respect to \(x_t\) and substituting them into this identity gives
\eqref{eq:253}
and
\eqref{eq:254}.

A simple isotropic surrogate is obtained by omitting the Hessian correction in~\eqref{eq:254}:
\begin{equation}
    \widehat\Sigma_t^\rho(x_t)
    =
    \frac{\beta_t^2}{\alpha_t^2}I.
    \label{eq:255}
\end{equation}
More refined surrogates may incorporate estimates based on either representation in~\eqref{eq:252}. For the Gaussian-kernel approximation in Section~\ref{sub2sec:15}, however, any choice that is positive definite preserves critical-point consistency.

\section{Proofs for Approximate Newton Matching}\label{app:6}

This appendix provides the proofs of the results in Section~\ref{sec:6}. We use the notation, targets, and assumptions introduced there and follow the same organization. Appendix~\ref{subsec:51} characterizes the population-stationary points of the two covariance approximations in Section~\ref{subsec:22} and establishes their critical-point consistency. Appendix~\ref{subsec:52} proves the critical-point consistency for the reference-adjoint and Gaussian-kernel approximations in Section~\ref{subsec:23}. Finally, Appendix~\ref{subsec:53} establishes the decomposition underlying approximate regularization and proves the reward-ascent and limiting results for the unregularized iteration in Section~\ref{subsec:24}.

\subsection{Approximate Covariance Forms with Critical-Point Consistency}\label{subsec:51}

\restateproposition{\targetPosteriorConsistency}
\begin{proof}
Let $v^\star$ be a population-stationary point of the loss $\calL_{\rho,\eta,B}^{\mathrm{dir}\text{-}\mathrm{lin}}$ in~\eqref{eq:129}. Proposition~\ref{prop:20} implies that for a.e. $t$ and $x_t$,
\begin{align*}
    v_t^\star(x_t)=&
    \E_{\substack{
    X_1\sim p_{1|t}^\base(\cdot|x_t)}}
    \Big[
    \left(1-\frac{\eta}{\tau}\right)v_t^\rho(x_t)
    +
    \frac{\eta}{\tau}v_t^\base(x_t)
    \notag\\
    &\qquad\qquad
    +
    \left(
    e^{\tau(r(X_1)-B_t(x_t))}-1
    \right)
    \left(
    \left(1-\frac{\eta}{\tau}\right)v_t^\rho(x_t)
    +
    \frac{\eta}{\tau}v_{t|1}(x_t|X_1)
    -
    v_t^\star(x_t)
    \right)
    \Big].
\end{align*}
Since $v_t^\base(x_t)=\E_{X_1\sim p_{1|t}^\base(\cdot|x_t)}[v_{t|1}(x_t|X_1)]$, we have
\begin{align*}
    0=&
    \E_{\substack{
    X_1\sim p_{1|t}^\base(\cdot|x_t)}}
    \Big[
    v_t^\star(x_t)
    -
    \left(
    \left(1-\frac{\eta}{\tau}\right)v_t^\rho(x_t)
    +
    \frac{\eta}{\tau}v_{t|1}(x_t|X_1)
    \right)
    \notag\\
    &\qquad\qquad
    -
    \left(
    e^{\tau(r(X_1)-B_t(x_t))}-1
    \right)
    \left(
    \left(1-\frac{\eta}{\tau}\right)v_t^\rho(x_t)
    +
    \frac{\eta}{\tau}v_{t|1}(x_t|X_1)
    -
    v_t^\star(x_t)
    \right)
    \Big]\\
    =&
    e^{ -\tau B_t(x_t)}\E_{\substack{
    X_1\sim p_{1|t}^\base(\cdot|x_t)}}
    \left[
        e^{\tau r(X_1)}\left(v_t^\star(x_t)-\left(1-\frac{\eta}{\tau}\right)v_t^\rho(x_t)-\frac{\eta}{\tau}v_{t|1}(x_t|X_1)\right)
    \right].
\end{align*}
By~\eqref{eq:130}, we can rewrite the first-order stationarity condition as
\begin{align*}
    0=&\E_{\substack{
    X_1\sim p_{1|t}^\pi(\cdot|x_t)}}
    \left[
        v_t^\star(x_t)-\left(1-\frac{\eta}{\tau}\right)v_t^\rho(x_t)-\frac{\eta}{\tau}v_{t|1}(x_t|X_1)
    \right]\\
    =&v_t^\star(x_t)-\left(1-\frac{\eta}{\tau}\right)v_t^\rho(x_t)-\frac{\eta}{\tau}v_{t}^\pi(x_t).
\end{align*}
Therefore, the unique population-stationary point is 
\[
v^\star=\left(1-\frac{\eta}{\tau}\right)v^\rho+\frac{\eta}{\tau}v^\pi.
\]
For any $\eta\in(0,\tau]$, we have
\[
v^\star=v^\rho\quad\Leftrightarrow\quad v^\pi=v^\rho\quad\Leftrightarrow\quad\pi=\rho,
\]
which establishes critical-point consistency.
\end{proof}

\restateproposition{\negativeTiltConsistency}
\begin{proof}
    Let $v^\star$ be a population-stationary point of the loss $\calL_{\rho,\eta,B}^{\mathrm{spl}\text{-}\mathrm{lin}}$ in~\eqref{eq:133}. Proposition~\ref{prop:20} implies that for a.e. $t$ and $x_t$,
    \begin{align*}
        v_t^\star(x_t)
        =&
    \E_{\substack{
    X_1\sim p_{1|t}^\rho(\cdot|x_t)}}
    \Big[
    \left(1-\frac{\eta}{\tau}\right)v_t^\rho(x_t)
    +
    \frac{\eta}{\tau}v_t^\base(x_t)
    \notag\\
    &
    -
    \left(
    e^{-\tau(r(X_1)-B_t(x_t))}-1
    \right)
    \left(
    v_t^\rho(x_t)
    -
    \frac{\eta}{\tau}
    \left(
    v_{t|1}(x_t|X_1)-v_t^\base(x_t)
    \right)
    -
    v_t^\star(x_t)
    \right)
    \Big].
    \end{align*}
    Thus,
    \begin{align*}
        0
        =&
    \E_{\substack{
    X_1\sim p_{1|t}^\rho(\cdot|x_t)}}
    \Big[
    v_t^\star(x_t)
    -
    \left(
    \left(1-\frac{\eta}{\tau}\right)v_t^\rho(x_t)
    +
    \frac{\eta}{\tau}v_t^\base(x_t)
    \right)
    \notag\\
    &\qquad
    -
    \left(
    e^{-\tau(r(X_1)-B_t(x_t))}-1
    \right)
    \left(
    v_t^\rho(x_t)
    -
    \frac{\eta}{\tau}
    \left(
    v_{t|1}(x_t|X_1)-v_t^\base(x_t)
    \right)
    -
    v_t^\star(x_t)
    \right)
    \Big]\\
    =&
    \E_{\substack{
    X_1\sim p_{1|t}^\rho(\cdot|x_t)}}
    \Big[
    v_t^\star(x_t)
    -
    \left(
        v_t^\rho(x_t)-\frac{\eta}{\tau}v_{t|1}(x_t|X_1)
    +
    \frac{\eta}{\tau}v_t^\base(x_t)
    \right)
    \notag\\
    &\qquad
    -
    \left(
    e^{-\tau(r(X_1)-B_t(x_t))}-1
    \right)
    \left(
    v_t^\rho(x_t)
    -
    \frac{\eta}{\tau}
    \left(
    v_{t|1}(x_t|X_1)-v_t^\base(x_t)
    \right)
    -
    v_t^\star(x_t)
    \right)
    \Big]\\
    =&
    e^{\tau B_t(x_t)}\E_{\substack{
    X_1\sim p_{1|t}^\rho(\cdot|x_t)}}
    \Big[
        e^{-\tau r(X_1)}\left(
    v_t^\star(x_t)
    -
        v_t^\rho(x_t)+\frac{\eta}{\tau}v_{t|1}(x_t|X_1)
    -
    \frac{\eta}{\tau}v_t^\base(x_t)
    \right)\Big].
    \end{align*}
    By~\eqref{eq:131}, we can rewrite the first-order stationarity condition as
    \begin{align*}
        0
        =&
    \E_{\substack{
    X_1\sim p_{1|t}^{\rho^-}(\cdot|x_t)}}
    \left[
    v_t^\star(x_t)-v_t^\rho(x_t)+\frac{\eta}{\tau}v_{t|1}(x_t|X_1)-\frac{\eta}{\tau}v_t^\base(x_t)
    \right]\\
    =&v_t^\star(x_t)-v_t^\rho(x_t)+\frac{\eta}{\tau}v_{t}^{\rho^-}(x_t)-\frac{\eta}{\tau}v_t^\base(x_t).
    \end{align*}
    Therefore, the population-stationary point is
    \[
    v^\star=v^\rho+\frac{\eta}{\tau}\left(v^\base-v^{\rho^-}\right).
    \]
    For any $\eta\in(0,\tau]$, we have
    \[
    v^\star=v^\rho\quad\Leftrightarrow\quad v^\base=v^{\rho^-}\quad\Leftrightarrow\quad\rho^\base=\rho^-\quad\Leftrightarrow\quad\pi=\rho,
    \]
    which establishes critical-point consistency.
\end{proof}

\subsection{Approximate Gradient Forms with Critical-Point Consistency}\label{subsec:52}

\restateproposition{\referenceSensitivityConsistency}

\begin{proof}
    It is straightforward to verify that the unique population minimizer of the training loss~\eqref{eq:138} is $\hat{v}$ in~\eqref{eq:136}. We next prove that $\hat{v}^{\mathrm{ref}\text{-}\mathrm{adj}}=v^\rho$ if and only if $\rho=\pi$.

    Define
    \[
    \bar\lambda_t^\rho(x)
    :=
    \E_{\BY_{[t,1]}\sim
    \mathbb P_{[t,1]|t}^{\rho}(\cdot|x)}
    \left[
    \lambda_t^{b^\rho,r,l^{\rho,\mu}}(\BY_{[t,1]})
    \right],\quad
    \bar\lambda_t^\mu(x)
    :=
    \E_{\BY_{[t,1]}\sim
    \mathbb P_{[t,1]|t}^{\rho}(\cdot|x)}
    \left[
    \lambda_t^{b^\mu,r,0}(\BY_{[t,1]})
    \right].
    \]
    Applying Proposition~\ref{prop:39} to $\bar\lambda_t^\rho$ and $\bar\lambda_t^\mu$ gives
    \begin{subequations}\label{eq:256}
        \begin{align}
            &(\partial_t+\mathcal{L}_t^\rho)\bar\lambda_t^\rho(x)+\nabla b_t^\rho(x)^\top\bar\lambda_t^\rho(x)+\nabla l_t^{\rho,\mu}(x)=0,\qquad t\in(0,1),\\
            &\bar\lambda_1^\rho(x)=\nabla r(x),
        \end{align}
    \end{subequations}
    and
    \begin{subequations}\label{eq:257}
        \begin{align}
            &(\partial_t+\mathcal{L}_t^\rho)\bar\lambda_t^\mu(x)+\nabla b_t^\mu(x)^\top\bar\lambda_t^\mu(x)=0,\qquad t\in(0,1),\label{eq:258}\\
            &\bar\lambda_1^\mu(x)=\nabla r(x),
        \end{align}
    \end{subequations}
    respectively.

    According to Proposition~\ref{prop:7}, we have
    \[
    \rho=\pi\quad\Leftrightarrow\quad\bar\lambda_t^\rho\equiv\frac{v_t^\rho-v_t^\mu}{\tau\kappa_t}\text{ for all }t\in(0,1).
    \]
    By~\eqref{eq:136}, we have
    \[
    \hat{v}^{\mathrm{ref}\text{-}\mathrm{adj}}=v^\rho\quad\Leftrightarrow\quad\bar\lambda_t^\mu\equiv \frac{v_t^\rho-v_t^\mu}{\tau\kappa_t}\text{ for all }t\in(0,1).
    \]
    Therefore, we need to prove that
    \[
    \bar\lambda_t^\rho\equiv\frac{v_t^\rho-v_t^\mu}{\tau\kappa_t}\text{ for all }t\in(0,1)\quad\Leftrightarrow\quad\bar\lambda_t^\mu\equiv \frac{v_t^\rho-v_t^\mu}{\tau\kappa_t}\text{ for all }t\in(0,1).
    \]
    
    Assume that $\bar\lambda_t^\rho\equiv\frac{v_t^\rho-v_t^\mu}{\tau\kappa_t}$ for all $t\in(0,1)$. Substituting this into~\eqref{eq:256} implies that
    \[
    (\partial_t+\mathcal{L}_t^\rho)\bar\lambda_t^\rho(x)+\nabla b_t^\mu(x)^\top\bar\lambda_t^\rho(x)=0.
    \]
    In other words, $\bar\lambda_t^\rho$ satisfies the same equation~\eqref{eq:257} as $\bar\lambda_t^\mu$. Therefore, $\bar\lambda_t^\rho\equiv \bar\lambda_t^\mu$ for all $t\in(0,1)$; hence, $\bar\lambda_t^\mu\equiv\frac{v_t^\rho-v_t^\mu}{\tau\kappa_t}$ for all $t\in(0,1)$.

    Assume that $\bar\lambda_t^\mu\equiv\frac{v_t^\rho-v_t^\mu}{\tau\kappa_t}$ for all $t\in(0,1)$. Substituting this into~\eqref{eq:257} implies that
    \[
    (\partial_t+\mathcal{L}_t^\rho)\bar\lambda_t^\mu(x)+\nabla b_t^\rho(x)^\top\bar\lambda_t^\mu(x)+\nabla l_t^{\rho,\mu}(x)=0.
    \]
    For the same reason, $\bar\lambda_t^\rho\equiv\frac{v_t^\rho-v_t^\mu}{\tau\kappa_t}$ for all $t\in(0,1)$ in this case.
\end{proof}

\restateproposition{\gaussianKernelConsistency}

\begin{proof}
    It is straightforward to verify that the unique population minimizer of the training loss~\eqref{eq:142} is $\hat{v}$ in~\eqref{eq:141}. Since $\widehat\Sigma_t^\rho(x_t)\in\S_{++}^d$, we have
    \[
    \hat{v}=v^\rho\quad\Leftrightarrow\quad
    \E_{X_1\sim p_{1|t}^\rho(\cdot|x_t)}
    [\nabla\tilde r^\rho(X_1)]
    =0\text{ for all }t\in(0,1),\,x_t\in\R^d.
    \]

    Assume that $\rho=\pi$. Then, $\tilde{r}^\rho\equiv\const$, which gives $\E_{X_1\sim p_{1|t}^\rho(\cdot|x_t)}
    [\nabla\tilde r^\rho(X_1)]
    =0$.

    Assume that $\E_{X_1\sim p_{1|t}^\rho(\cdot|x_t)}
    [\nabla\tilde r^\rho(X_1)]
    =0$. Then, we have
    \begin{equation*}
        \int_{\R^d}
        p_{t|1}(x_t|x_1)
        \rho(x_1)
        \nabla\tilde r^\rho(x_1)
        \dd x_1
        =0.
    \end{equation*}
    For any interior time $t\in(0,1)$, the left-hand side is a nondegenerate scaled Gaussian convolution. Injectivity gives
    \begin{equation*}
        \rho(x)\nabla\tilde r^\rho(x)\equiv0.
    \end{equation*}
    Since $\rho$ is positive, we have $\tilde r^\rho\equiv\const$, which implies $\rho=\pi$.
\end{proof}

\subsection{Regularization Trade-Offs}\label{subsec:53}

\restateproposition{\partialRegularizationDecomposition}
\begin{proof}
By Bayes' rule, for every interior $(t,x_t)$, we have
\begin{align*}
    V_t^\rho\left[\log\frac{\rho}{\rho^\base}\right](x_t)
    &=
    \log\frac{p_t^\rho(x_t)}{p_t^\base(x_t)}
    +
    \KL{p_{1|t}^{\rho}(\cdot|x_t)}{p_{1|t}^{\base}(\cdot|x_t)}.
\end{align*}
Then, 
\begin{equation*}
    \Gamma^{\rho,\tilde r^\rho}
    =
    \Gamma^{\rho,r}
    -
    \frac{1}{\tau}
    \Gamma^{\rho,\log(\rho/\rho^\base)},
    \qquad
    \kappa_t\nabla_{x_t}
    \log\frac{p_t^\rho(x_t)}{p_t^\base(x_t)}
    =
    v_t^\rho(x_t)-v_t^\base(x_t),
\end{equation*}
which proves the conclusion.
\end{proof}

\restateproposition{\noRegularizationRewardImprovementNotCPC}

\begin{proof}
    The value-ascent certificate is a specialization of Theorem~\ref{thm:2} with $f:=r$. If $\rho=\pi$ but $r$ is nonconstant, then the strict value ascent implies that $q\neq\rho$; hence, $\Gamma^{\rho,r}\neq 0$ even at $\rho=\pi$. Therefore, the approximation is not critical-point consistent.
\end{proof}

\restateproposition{\noRegularizationConvergeToArgmax}

\begin{proof}
    We have
    \[
        \Diss_{\rho_k,\eta_k,r}(x)=\eta_k^2\int_{0}^{1}\kappa_t\norm[2]{\nabla V_t^{\rho_k}[r]\left(\flow_{1\to t}^{\rho_k,\eta_k,r}(x)\right)}^2\dd t\geq0.
    \]
    Denote $A_\varepsilon:=\{x\in\R^d:r(x)<r_{\max}-\varepsilon\}$. Denote $S_k:=\sum_{i=0}^{k-1}\Diss_{\rho_i,\eta_i,r}$. Theorem~\ref{thm:2} implies
    \[
    \rho_k(x)\propto\rho_0(x)\exp\left(\sum_{i=0}^{k-1}\eta_ir(x)-S_k(x)\right).
    \]
    For any $\delta\in(0,\varepsilon)$, we have
    \[
    \begin{aligned}
        \rho_k(A_{\varepsilon})
        =&\frac{\int_{A_{\varepsilon}}\rho_0(x)\exp\left(\sum_{i=0}^{k-1}\eta_ir(x)-S_k(x)\right)\dd x}{\int_{\R^d}\rho_0(x)\exp\left(\sum_{i=0}^{k-1}\eta_ir(x)-S_k(x)\right)\dd x}\\
        \leq&\frac{\int_{A_{\varepsilon}}\rho_0(x)\exp\left(\sum_{i=0}^{k-1}\eta_i(r_{\max}-\varepsilon)-S_k(x)\right)\dd x}{\int_{A_{\delta}^\mathrm{c}}\rho_0(x)\exp\left(\sum_{i=0}^{k-1}\eta_i(r_{\max}-\delta)-S_k(x)\right)\dd x}\\
        \leq&\exp\left((\delta-\varepsilon)\sum_{i=0}^{k-1}\eta_i\right)\left(\int_{A_{\delta}^\mathrm{c}}\rho_0(x)\exp\left(-S_k(x)\right)\dd x\right)^{-1}.
    \end{aligned}
    \]
    By the definition of $r_{\max}$ and the continuity of $r$, we have $\rho_0(A_\delta^\mathrm{c})>0$. Let $\rho_{0|A_{\delta}^\mathrm{c}}$ denote the normalized restriction of $\rho_0$ to $A_{\delta}^\mathrm{c}$. By Jensen's inequality, we have
    \[
    \begin{aligned}
        \int_{A_{\delta}^\mathrm{c}}\rho_0(x)\exp\left(-S_k(x)\right)\dd x
        =&\rho_0(A_{\delta}^\mathrm{c})\E_{X\sim\rho_{0|A_{\delta}^\mathrm{c}}}\left[\exp\left(-S_k(X)\right)\right]\\
        \geq&\rho_0(A_{\delta}^\mathrm{c})\exp\left(-\E_{X\sim\rho_{0|A_{\delta}^\mathrm{c}}}\left[S_k(X)\right]\right)\\
        \geq&\rho_0(A_{\delta}^\mathrm{c})\exp\left(-\frac{\E_{X\sim\rho_{0}}\left[S_k(X)\right]}{\rho_0(A_{\delta}^\mathrm{c})}\right).
    \end{aligned}
    \]
    Therefore, we have
    \[
    \begin{aligned}
        \rho_k(A_{\varepsilon})\leq&\frac{1}{\rho_0(A_{\delta}^\mathrm{c})}\exp\left((\delta-\varepsilon)\sum_{i=0}^{k-1}\eta_i+\frac{1}{\rho_0(A_{\delta}^\mathrm{c})}\sum_{i=0}^{k-1}\E_{X\sim\rho_{0}}\left[\Diss_{\rho_i,\eta_i,r}(X)\right]\right)\\
        \leq&\frac{1}{\rho_0(A_{\delta}^\mathrm{c})}\exp\left(\left(\sum_{i=0}^{k-1}\eta_i\right)\left(\frac{\sum_{i=0}^{k-1}\E_{X\sim\rho_{0}}\left[\Diss_{\rho_i,\eta_i,r}(X)\right]}{\rho_0(A_{\delta}^\mathrm{c})\left(\sum_{i=0}^{k-1}\eta_i\right)}+\delta-\varepsilon\right)\right).
    \end{aligned}
    \]
    By condition~\eqref{eq:150}, the Stolz--Ces\`aro theorem implies
    \[
    \lim_{k\to\infty}\frac{\sum_{i=0}^{k-1}\E_{X\sim\rho_{0}}\left[\Diss_{\rho_i,\eta_i,r}(X)\right]}{\sum_{i=0}^{k-1}\eta_i}=\lim_{k\to\infty}\frac{1}{\eta_k}\E_{X\sim\rho_{0}}\left[\Diss_{\rho_k,\eta_k,r}(X)\right]=0.
    \]
    Since $\delta-\varepsilon<0$ is a fixed negative number, we have $\lim_{k\to\infty}\rho_k(A_{\varepsilon})=0$.
\end{proof}

To prove Proposition~\ref{prop:19}, we separate the terminal-density error into two components: a residual exponential tilt caused by the mismatch $\hat{\tau}-\tau$, and an accumulated path-dissipation correction. The former can be controlled uniformly for bounded rewards through Hoeffding's lemma~\cite[Equation (4.16)]{hoeffding1963probability}.

\begin{lemma}[Hoeffding's lemma]\label{lem:16}
    Let $X\sim p$ be a bounded real-valued random variable satisfying
    \[
    a\leq X\leq b\qquad\text{a.s.}
    \]
    Denote $m:=\E_{X\sim p}[X]$. Then, for any $\tau\in\R$,
    \[
    \E_{X\sim p}\left[e^{\tau(X-m)}\right]\leq\exp\left(\frac{\tau^2(b-a)^2}{8}\right).
    \]
\end{lemma}

\begin{proof}
    The case where $a=b$ is immediate. Assume that $a<b$. For all $x\in[a,b]$,
    \[
    e^{\tau x}\leq\frac{b-x}{b-a}e^{\tau a}+\frac{x-a}{b-a}e^{\tau b}.
    \]
    Taking the expectation over $p$ gives
    \begin{equation}\label{eq:259}
        \E_{X\sim p}\left[e^{\tau(X-m)}\right]
        \leq\frac{b-m}{b-a}e^{\tau(a-m)}+\frac{m-a}{b-a}e^{\tau(b-m)}
        =e^{L(\tau(b-a))},
    \end{equation}
    where
    \[
    L(y):=\frac{y(a-m)}{b-a}+\log\left(1-\frac{(e^y-1)(a-m)}{b-a}\right).
    \]
    Then,
    \[
    L(0)=0,\qquad
    L'(0)=0,\qquad
    L''(y)=(m-a)(b-m)e^{y}\left((m-a)e^y+b-m\right)^{-2}\in\left[0,\frac14\right].
    \]
    By the Taylor expansion, for every $y\in\R$, there exists $z$ between $0$ and $y$ such that
    \[
    L(y)=\frac{1}{2}L''(z)y^2\leq\frac{y^2}{8}.
    \]
    Substituting this back into \eqref{eq:259} finishes the proof.
\end{proof}

Based on Hoeffding's lemma, the following lemma bounds the KL divergences in both directions when a probability density is exponentially tilted by a bounded reward.

\begin{lemma}\label{lem:17}
    Let $r:\R^d\to\R$ be a bounded reward function satisfying
    \[
    \norm[\infty]{r}:=\sup_{x\in\R^d}\abs{r(x)}<\infty.
    \]
    Let $\mu\in\pdfspace$ and $\pi\in\pdfspace$ be two probability densities satisfying
    \[
    \pi(x)\propto \mu(x)e^{\tau r(x)}
    \]
    for some constant $\tau\in\R$. Then,
    \[
    \max\left\{\KL{\pi}{\mu},\KL{\mu}{\pi}\right\}\leq\frac{1}{2}\tau^2\norm[\infty]{r}^2.
    \]
\end{lemma}

\begin{proof}
    Since
    \[
    \pi(x)=\frac{\mu(x)e^{\tau r(x)}}{\E_{X\sim\mu}\left[e^{\tau r(X)}\right]},\qquad
    -\norm[\infty]{r}\leq r(x)\leq\norm[\infty]{r},
    \]
    applying Hoeffding's lemma in Lemma~\ref{lem:16} gives
    \begin{align*}
        \KL{\mu}{\pi}
        =&\E_{X\sim\mu}\left[\log\frac{\mu(X)}{\pi(X)}\right]\\
        =&\log\E_{X\sim\mu}\left[e^{\tau r(X)}\right]-\tau \E_{X\sim\mu}\left[r(X)\right]\\
        =&\log\E_{X\sim\mu}\left[e^{\tau\left(r(X)-\E_{Y\sim\mu}[r(Y)]\right)}\right]\\
        \leq&\log\left(\exp\left(\frac{\tau^2\left(2\norm[\infty]{r}\right)^2}{8}\right)\right)\\
        =&\frac{1}{2}\tau^2\norm[\infty]{r}^2.
    \end{align*}
    Then, the same argument gives
    \[
    \KL{\pi}{\mu}\leq\frac{1}{2}\tau^2\norm[\infty]{r}^2
    \]
    since $\mu(x)\propto \pi(x)e^{-\tau r(x)}$.
\end{proof}

\restateproposition{\noRegularizationConvergeToPi}

\begin{proof}
    Denote $S(x):=\sum_{k=0}^{K-1}\Diss_{\rho_k,\eta_k,r}(x)\geq0$. Then,
    \[
    \E_{X\sim\pi}\left[S(X)\right]=\sum_{k=0}^{K-1}\E_{X\sim\pi}\left[\Diss_{\rho_k,\eta_k,r}(X)\right]\leq C\sum_{k=0}^{K-1}\eta_k^2\leq C\eta_\mathrm{max}\hat{\tau}.
    \]
    Theorem~\ref{thm:2} implies
    \[
    \rho_K(x)\propto\pi(x)\exp\left(\left(\hat{\tau}-\tau\right)r(x)-S(x)\right).
    \]

    We first investigate the forward KL. Let
    \[
    \hat{\pi}(x):=\frac{\pi(x)e^{\left(\hat{\tau}-\tau\right)r(x)}}{\E_{X\sim\pi}\left[e^{\left(\hat{\tau}-\tau\right)r(X)}\right]}.
    \]
    Then, Lemma~\ref{lem:17} implies
    \begin{align*}
        \KL{\pi}{\hat{\pi}}\leq\frac{1}{2}\left(\hat{\tau}-\tau\right)^2\norm[\infty]{r}^2.
    \end{align*}
    Since
    \[
    \rho_K(x)=\frac{\hat{\pi}(x)e^{-S(x)}}{\E_{X\sim\hat\pi}\left[e^{-S(X)}\right]},\qquad
    S(x)\geq0,
    \]
    the forward KL can be bounded by
    \begin{align*}
        \KL{\pi}{\rho_K}
        =&\E_{X\sim\pi}\left[\log\frac{\pi(X)}{\rho_K(X)}\right]\\
        =&\E_{X\sim\pi}\left[\log\frac{\hat{\pi}(X)}{\rho_K(X)}\right]+\E_{X\sim\pi}\left[\log\frac{\pi(X)}{\hat{\pi}(X)}\right]\\
        \leq&\log\E_{X\sim\hat\pi}\left[e^{-S(X)}\right]+\E_{X\sim\pi}\left[S(X)\right]+\frac{1}{2}\left(\hat{\tau}-\tau\right)^2\norm[\infty]{r}^2\\
        \leq&C\eta_{\mathrm{max}}\hat{\tau}+\frac{1}{2}\left(\hat{\tau}-\tau\right)^2\norm[\infty]{r}^2.
    \end{align*}
    
    We then bound the reverse KL. Let
    \[
    \tilde{\pi}(x):=\frac{\pi(x)e^{-S(x)}}{\E_{X\sim\pi}\left[e^{-S(X)}\right]}.
    \]
    Since
    \[
    \rho_K(x)=\frac{\tilde{\pi}(x)e^{(\hat{\tau}-\tau)r(x)}}{\E_{X\sim\tilde{\pi}}\left[e^{(\hat{\tau}-\tau)r(X)}\right]},
    \]
    Lemma~\ref{lem:17} implies
    \[
    \KL{\rho_K}{\tilde{\pi}}\leq\frac{1}{2}\left(\hat{\tau}-\tau\right)^2\norm[\infty]{r}^2.
    \]
    The reverse KL can be decomposed as
    \begin{align*}
        \KL{\rho_K}{\pi}
        =&\E_{X\sim\rho_K}\left[\log\frac{\rho_K(X)}{\pi(X)}\right]\\
        =&\E_{X\sim\rho_K}\left[\log\frac{\rho_K(X)}{\tilde{\pi}(X)}\right]+\E_{X\sim\rho_K}\left[\log\frac{\tilde{\pi}(X)}{\pi(X)}\right]\\
        \leq&\frac{1}{2}\left(\hat{\tau}-\tau\right)^2\norm[\infty]{r}^2-\E_{X\sim\rho_K}\left[S(X)\right]-\log\E_{X\sim\pi}\left[e^{-S(X)}\right]\\
        \leq&\frac{1}{2}\left(\hat{\tau}-\tau\right)^2\norm[\infty]{r}^2-\log\E_{X\sim\pi}\left[e^{-S(X)}\right].
    \end{align*}
    By Jensen's inequality, we have
    \[
    \log\E_{X\sim\pi}\left[e^{-S(X)}\right]\geq\E_{X\sim\pi}\left[-S(X)\right]\geq-C\eta_{\mathrm{max}}\hat{\tau}.
    \]
    Therefore, we have
    \begin{align*}
        \KL{\rho_K}{\pi}
        \leq&\frac{1}{2}\left(\hat{\tau}-\tau\right)^2\norm[\infty]{r}^2+C\eta_{\mathrm{max}}\hat{\tau}.
    \end{align*}
    This finishes the proof.
\end{proof}

\section{Canonical Updates with Time-Dependent Stepsizes}
\label{app:7}

Section~\ref{sec:2} studies the finite-stepsize canonical retraction. The corresponding tangential update is
\[
\bar v_t
=
v_t^\rho
+
\eta\Gamma_t^{\rho,f},
\qquad
\eta\in(0,\infty)
\text{ is constant}.
\]
Since $\eta\Gamma^{\rho,f}\in T_{v^\rho}\velmancan$, composing this velocity-space addition with the canonical projection implements the canonical retraction introduced in Section~\ref{subsec:7}.

This appendix replaces the constant stepsize \(\eta\) by a time-dependent modulation:
\[
\bar v_t
=
v_t^\rho
+
\eta_t\Gamma_t^{\rho,f},
\qquad
\eta\in C^1([0,1];(0,\infty))
\text{ is time-dependent}.
\]
Although $\Gamma^{\rho,f}$ is a tangent vector in $T_{v^\rho}\velmancan$, its time-dependent modulation $\left(\eta_t\Gamma_t^{\rho,f}\right)_{t\in[0,1]}$ need not belong to $T_{v^\rho}\velmancan$. The resulting velocity-space addition followed by canonicalization is therefore generally not a retraction in the geometric sense. Nevertheless, the resulting velocity field generates a well-defined probability path and terminal density. We derive an exact formula for this terminal density in Appendix~\ref{subsec:54}. Then, Appendix~\ref{subsec:55} shows that the monotonicity condition
\[
\dot\eta_t\geq0,
\qquad
t\in(0,1),
\]
is sufficient to extend the finite-stepsize value-ascent certificate in Section~\ref{subsec:8} to this new setting. Furthermore, if we set $f=\tilde{r}^\rho$ and use a nondecreasing $\eta\in C^1([0,1];(0,\tau])$ for Newton Matching, the reverse-KL descent guarantee remains valid.

\subsection{Terminal-Density Characterization}
\label{subsec:54}

Since the canonical projection preserves terminal densities, the terminal density of the canonical updates with time-dependent stepsizes is determined entirely by the ambient velocity field before canonicalization. We therefore analyze the density path generated by
\[
\bar v_t^{\rho,\eta,f}
=
v_t^\rho
+
\eta_t\Gamma_t^{\rho,f},
\]
and recover the same terminal density as the field obtained after canonicalization.

This subsection derives the resulting terminal-density formula in two steps. First, Appendix~\ref{sub2sec:34} establishes a general terminal-density formula for velocity-space additive updates with generic gradient fields:
\[
\bar v_t^{\rho,\Psi}
=
v_t^\rho
+
\kappa_t\nabla\Psi_t.
\]
This calculation uses neither the posterior-value structure nor tangency to the canonical manifold. Appendix~\ref{sub2sec:35} then chooses $\Psi_t=\eta_t V_t^\rho[f]$. This specialization provides the terminal-density formula for the update $\bar v_t^{\rho,\eta,f}=v_t^\rho+\eta_t\Gamma_t^{\rho,f}$.

\subsubsection{Additive Updates with Generic Gradients}\label{sub2sec:34}

We derive the general formula for the terminal density of a velocity-space additive update with a generic gradient field.
Let $\Psi_t:\R^d\to\R$ be a time-dependent potential. Define
\begin{equation}
    \bar{v}_t^{\rho,\Psi}(x)
    :=
    v_t^\rho(x)+\kappa_t\nabla\Psi_t(x),
    \label{eq:260}
\end{equation}
and
\begin{equation}
    \mathfrak R_t^{\rho,\Psi}(x)
    :=
    (\partial_t+\calL_t^\rho)\Psi_t(x)
    +
    \kappa_t\norm[2]{\nabla\Psi_t(x)}^2.
    \label{eq:261}
\end{equation}
Let $\nu_t^{\rho,\Psi}$ and $\flow^{\rho,\Psi}$ denote the density path and the ODE flow generated by $\bar{v}_t^{\rho,\Psi}$, respectively.

\begin{proposition}
\label{prop:43}
Define
\begin{equation*}
    \widetilde G_t^{\rho,\Psi}(x)
    :=
    p_t^\rho(x)e^{\Psi_t(x)}.
\end{equation*}
Then, we have the following transport identity:
\begin{equation*}
    \partial_t\widetilde G_t^{\rho,\Psi}
    +
    \nabla\cdot
    \left(
    \widetilde G_t^{\rho,\Psi}\bar{v}_t^{\rho,\Psi}
    \right)
    =
    \widetilde G_t^{\rho,\Psi}\mathfrak R_t^{\rho,\Psi}.
\end{equation*}
\end{proposition}

\begin{proof}
Expand the left-hand side using $\widetilde G_t^{\rho,\Psi}=p_t^\rho e^{\Psi_t}$ and~\eqref{eq:260}. The continuity equation cancels the terms without $\Psi_t$, and the remaining factor is~\eqref{eq:261}.
\end{proof}

\begin{proposition}
\label{prop:44}
Let $Y_t$ be the characteristic
\[
\frac{\dd Y_t}{\dd t}=\bar v_t^{\rho,\Psi}(Y_t).
\]
Then, we have
\begin{equation}\label{eq:262}
\frac{\dd}{\dd t}\left[\log\frac{p_t^\rho(Y_t)}{\nu_t^{\rho,\Psi}(Y_t)}+\Psi_t(Y_t)\right]=\mathfrak R_t^{\rho,\Psi}(Y_t).
\end{equation}
Consequently, the updated terminal density is
\begin{align*}
    \nu_1^{\rho,\Psi}(x)
    =&
    \rho(x)
    \exp
    \left(
    \Psi_1(x)
    -
    \Psi_0
    \left(
    \flow_{1\to0}^{\rho,\Psi}(x)
    \right)
    -
    \int_0^1
    \mathfrak R_t^{\rho,\Psi}
    \left(
    \flow_{1\to t}^{\rho,\Psi}(x)
    \right)\dd t
    \right).
\end{align*}
\end{proposition}

\begin{proof}
By Proposition~\ref{prop:43}, we have
\[
\frac{\dd}{\dd t}\log\widetilde G_t^{\rho,\Psi}(Y_t)=-\nabla\cdot \bar{v}_t^{\rho,\Psi}(Y_t)+\mathfrak R_t^{\rho,\Psi}(Y_t).
\]
By the continuity equation, we have
\[
\frac{\dd}{\dd t}\log\nu_t^{\rho,\Psi}(Y_t)=-\nabla\cdot \bar{v}_t^{\rho,\Psi}(Y_t),
\]
which gives~\eqref{eq:262}. Integrating from $t=0$ to $t=1$ finishes the proof.
\end{proof}

\subsubsection{Additive Updates with Posterior-Value Gradients}\label{sub2sec:35}

Here, we specialize the generic gradient field to the posterior-value gradient, i.e., choosing $\Psi_t=\eta_tV_t^\rho[f]$ in~\eqref{eq:260}, where the time-dependent stepsize $\eta\in C^1([0,1];\R)$. Consider the following update:
\begin{subequations}\label{eq:263}
    \begin{align}
    &\bar v_t^{\rho,\eta,f}(x)
    :=
    v_t^\rho(x)
    +
    \eta_t\Gamma_t^{\rho,f}(x),\label{eq:264}\\
    &q:=\terminalmap(\bar v^{\rho,\eta,f}).
    \end{align}
\end{subequations}
Here $\Gamma_t^{\rho,f}(x)=\kappa_t\nabla V_t^{\rho}[f](x)$. Let $\nu_t^{\rho,\eta,f}$ and $\flow^{\rho,\eta,f}$ denote the density path and the ODE flow generated by $\bar{v}_t^{\rho,\eta,f}$, respectively.

\begin{proposition}
\label{prop:45}
Along the characteristic
\[
\frac{\dd Y_t}{\dd t}=\bar v_t^{\rho,\eta,f}(Y_t),
\]
we have
\begin{equation}\label{eq:265}
\frac{\dd}{\dd t}\left[\log\frac{p_t^\rho(Y_t)}{\nu_t^{\rho,\eta,f}(Y_t)}+\eta_t\left(V_t^\rho[f](Y_t)-\E_\rho[f]\right)\right]=\dot\eta_t\left(V_t^\rho[f](Y_t)-\E_\rho[f]\right)+\eta_t^2\kappa_t\norm[2]{\nabla V_t^\rho[f](Y_t)}^2.
\end{equation}
Consequently,
\begin{equation*}
    q(x)
    =
    \rho(x)
    \exp
    \left(
    \eta_1
    \left(
    f(x)-\E_{X\sim\rho}[f(X)]
    \right)
    -
    \Diss_{\rho,\eta,f}(x)
    \right),
\end{equation*}
where
\begin{align*}
    \Diss_{\rho,\eta,f}(x)
    :=&
    \int_0^1
    \left[
    \dot\eta_t
    \left(
    V_t^\rho[f]
    \left(
    \flow_{1\to t}^{\rho,\eta,f}(x)
    \right)
    -
    \E_{X\sim\rho}[f(X)]
    \right)
    +
    \eta_t^2\kappa_t
    \norm[2]{
    \nabla V_t^\rho[f]
    \left(
    \flow_{1\to t}^{\rho,\eta,f}(x)
    \right)
    }^2
    \right]\dd t.
\end{align*}
\end{proposition}

\begin{proof}
    According to Theorem~\ref{thm:13}, we have $(\partial_t+\calL_t^\rho)V_t^\rho[f]=0$. Substituting this into~\eqref{eq:261} gives
    \[
    \mathfrak R_t^{\rho,\eta V^\rho[f]}(x)
    =
    \dot{\eta}_t V_t^\rho[f](x)
    +
    \kappa_t\eta_t^2\norm[2]{\nabla V_t^\rho[f](x)}^2.
    \]
    By Proposition~\ref{prop:44}, 
    \[
    \frac{\dd}{\dd t}\left[\log\frac{p_t^\rho(Y_t)}{\nu_t^{\rho,\eta,f}(Y_t)}+\eta_tV_t^\rho[f](Y_t)\right]=\dot\eta_tV_t^\rho[f](Y_t)+\eta_t^2\kappa_t\norm[2]{\nabla V_t^\rho[f](Y_t)}^2,
    \]
    which gives~\eqref{eq:265}. Integrating from $t=0$ to $t=1$ finishes the proof.
\end{proof}

When $\eta_t\equiv\eta$ is constant, the term involving $\dot\eta_t$ vanishes, and $\Diss_{\rho,\eta,f}$ reduces exactly to the nonnegative path dissipation introduced in Section~\ref{subsec:8}. However, when $\eta$ is time-dependent, the term
\[
\dot\eta_t
\left(
    V_t^\rho[f]
    \left(
    \flow_{1\to t}^{\rho,\eta,f}(x)
    \right)
    -
    \E_{X\sim\rho}[f(X)]
    \right)
\]
need not be pointwise nonnegative. The terminal-density formula in Proposition~\ref{prop:45} alone therefore does not yet imply value ascent. The next subsection shows that a nondecreasing time-dependent stepsize is sufficient for value ascent and, in the Newton Matching specialization, for reverse-KL descent under the additional condition \(\eta_1\leq\tau\).

\subsection{Monotonicity Guarantees}
\label{subsec:55}

Appendix~\ref{subsec:54} presents the terminal-density characterization for general additive updates with time-dependent stepsizes. We now identify conditions under which the monotonicity guarantees for constant stepsizes extend to this setting. A nondecreasing time-dependent stepsize yields value ascent for general \(f\); for \(f=\tilde r^\rho\), the additional condition \(\eta_1\leq\tau\) yields reverse-KL descent.

Appendix~\ref{sub2sec:36} first derives an identity for the expectation of the pathwise correction under the updated density and uses it to prove value ascent. Appendix~\ref{sub2sec:37} then specializes to \(f=\tilde r^\rho\), yielding the time-dependent-stepsize counterparts of the Newton Matching terminal-density formula and three-point identity. Under the additional bound \(\eta_1\leq\tau\), the latter implies one-stage reverse-KL descent.

\subsubsection{Value-Ascent Certificate}
\label{sub2sec:36}

The terminal-density formula in Appendix~\ref{sub2sec:35} contains a pathwise correction that need not be nonnegative pointwise. To establish value ascent, it suffices to show that its expectation under the updated density \(q\) is nonnegative. We first derive an exact identity for this expectation and then combine it with the terminal-density formula to obtain the time-dependent-stepsize counterpart of the decomposition in Section~\ref{subsec:8}.

\begin{proposition}
\label{prop:46}
Assume
\[
\forall t\in(0,1),\,\eta_t>0
\]
and the boundary conditions required for the integration by parts below. Then, we have
\begin{align*}
    &\E_{X\sim q}
    \left[
    \Diss_{\rho,\eta,f}(X)
    \right]
    =
    \eta_1
    \int_0^1
    \left[
    \frac{\dot\eta_t}{\eta_t^2}
    \KL{\nu_t^{\rho,\eta,f}}{p_t^\rho}
    +
    \eta_t\kappa_t
    \E_{X\sim\nu_t^{\rho,\eta,f}}
    \left[
    \norm[2]{\nabla V_t^\rho[f](X)}^2
    \right]
    \right]\dd t.
\end{align*}
In particular, if we additionally assume 
\[
\forall t\in(0,1),\,\dot\eta_t\ge0,
\]
then
\[
\E_{X\sim q}
    \left[
    \Diss_{\rho,\eta,f}(X)
    \right]\geq0.
\]
\end{proposition}

\begin{proof}
For simplicity, denote $\nu_t=\nu_t^{\rho,\eta,f}$ and
\[
    \Diss_t(x)
    :=
    \int_0^t
    \left[
    \dot\eta_s
    \left(
    V_s^\rho[f]
    \left(
    \flow_{t\to s}^{\rho,\eta,f}(x)
    \right)
    -
    \E_{X\sim\rho}[f(X)]
    \right)
    +
    \eta_s^2\kappa_s
    \norm[2]{
    \nabla V_s^\rho[f]
    \left(
    \flow_{t\to s}^{\rho,\eta,f}(x)
    \right)
    }^2
    \right]\dd s.
\]
Integrating~\eqref{eq:265} from $s=0$ to $s=t$ gives
\[
    \nu_t(x)
    =
    p_t^\rho(x)
    \exp
    \left(
    \eta_t
    \left(
    V_t^\rho[f](x)-\E_{X\sim\rho}[f(X)]
    \right)
    -
    \Diss_t(x)
    \right).
\]
Therefore, we have
\[
\E_{X\sim\nu_t}\left[\Diss_t(X)\right]=-\KL{\nu_t}{p_t^\rho}+\eta_t\left(\E_{X\sim\nu_t}\left[V_t^\rho[f](X)\right]-\E_{X\sim\rho}[f(X)]\right).
\]
Then, taking the expectation of~\eqref{eq:265} gives
\[
\frac{\dd}{\dd t}\E_{X\sim\nu_t}\left[\Diss_t(X)\right]=\dot\eta_t\left(\E_{X\sim\nu_t}\left[V_t^\rho[f](X)\right]-\E_{X\sim\rho}[f(X)]\right)+\eta_t^2\kappa_t\E_{X\sim\nu_t}\left[\norm[2]{\nabla V_t^\rho[f](X)}^2\right].
\]
Therefore,
\begin{align*}
    \frac{\dd}{\dd t}\left(\frac{1}{\eta_t}\E_{X\sim\nu_t}\left[\Diss_t(X)\right]\right)
    =&\frac{1}{\eta_t}\frac{\dd}{\dd t}\E_{X\sim\nu_t}\left[\Diss_t(X)\right]-\frac{\dot\eta_t}{\eta_t^2}\E_{X\sim\nu_t}\left[\Diss_t(X)\right]\\
    =&\frac{\dot\eta_t}{\eta_t^2}\KL{\nu_t}{p_t^\rho}+\eta_t\kappa_t\E_{X\sim\nu_t}\left[\norm[2]{\nabla V_t^\rho[f](X)}^2\right].
\end{align*}
Integrating from $t=0$ to $t=1$ finishes the proof.
\end{proof}

The following theorem provides the value-ascent certificate.

\begin{theorem}\label{thm:17}
    Consider the ideal stage~\eqref{eq:263}. For any test density $p\in\pdfspace$, we have%
    \begin{equation}\label{eq:266}
        \KL{p}{q}=\KL{p}{\rho}-\eta_1(\E_{X\sim p}[f(X)]-\E_{X\sim\rho}[f(X)])+\E_{X\sim p}\left[\Diss_{\rho,\eta,f}(X)\right].
    \end{equation}
    In particular, when $\eta_1>0$, letting $p=q$ gives
    \begin{equation}\label{eq:267}
    \E_{X\sim q}[f(X)]-\E_{X\sim\rho}[f(X)]=\frac{1}{\eta_1}\KL{q}{\rho}+\frac{1}{\eta_1}\E_{X\sim q}\left[\Diss_{\rho,\eta,f}(X)\right].
    \end{equation}

    Consequently, for any nondecreasing time-dependent stepsize $\eta$ which satisfies
    \[
    \forall t\in(0,1),\,\eta_t>0,\,\dot{\eta}_t\geq0,
    \]
    we have the value-ascent certificate:
    \begin{equation*}
        \E_{X\sim q}[f(X)]\ge\E_{X\sim\rho}[f(X)].
    \end{equation*}
    Furthermore,
    \[
    \E_{X\sim q}[f(X)]=\E_{X\sim\rho}[f(X)]\quad\Leftrightarrow\quad\rho=q\quad\Leftrightarrow\quad
    f\equiv\const.
    \]
\end{theorem}

\begin{proof}
    According to Proposition~\ref{prop:45}, we have
    \[
    \begin{aligned}
        \KL{p}{q}
        =&\E_{X\sim p}[\log p(X)]-\E_{X\sim p}[\log q(X)]\\
        =&\E_{X\sim p}[\log p(X)]-\E_{X\sim p}[\log\rho(X)+\eta_1\left(f(X)-\E_{X\sim\rho}[f(X)]\right)-\Diss_{\rho,\eta,f}(X)],
    \end{aligned}
    \]
    which gives~\eqref{eq:266}. Setting $p=q$ gives~\eqref{eq:267}.

    If we further assume that $\eta$ is nondecreasing, Proposition~\ref{prop:46} implies
    \[
        \E_{X\sim q}
            \left[
            \Diss_{\rho,\eta,f}(X)
            \right]\geq0.
    \]
    Therefore, we have $\E_{X\sim q}[f(X)]\ge\E_{X\sim\rho}[f(X)]$.

    The remaining task is to prove the sufficiency and necessity of the equality. First, if $\E_{X\sim q}[f(X)]=\E_{X\sim\rho}[f(X)]$ holds, then~\eqref{eq:267} implies $\KL{q}{\rho}=0$; hence, we have $\rho=q$. Second, $\rho=q$ implies $\E_{X\sim q}\left[\Diss_{\rho,\eta,f}(X)\right]=0$. Therefore, we have $\nabla V_t^\rho[f]\aeq 0$. Since $\E_{\rho}[\abs{f}]<\infty$ and the convolution kernel is a nondegenerate Gaussian, $V_t^\rho[f]$ is smooth; hence, $V_t^\rho[f]\aeq c_t$ is constant. In other words, we have
    \[
        \int_{\R^d}(f(x_1)-c_t)\rho(x_1)p_{t|1}(x_t|x_1)\,\dd x_1=0
    \]
    for a.e. $x_t\in\R^d$. Since the Gaussian convolution is injective and $f$, $\rho$ are continuous, we have $(f(x_1)-c_t)\rho(x_1)\equiv0$. Therefore, $f\equiv\const$. Third, if $f\equiv\const$, then $\E_{X\sim q}[f(X)]=f(0)=\E_{X\sim\rho}[f(X)]$.
\end{proof}

\subsubsection{Finite-Stepsize KL Descent}
\label{sub2sec:37}

Section~\ref{sec:3} identifies \(\tau\Gamma^{\rho,\tilde r^\rho}\) as the canonical Newton direction, while Section~\ref{sub2sec:7} proves finite-stepsize reverse-KL descent for the associated canonical retraction with a constant stepsize \(\eta\in(0,\tau]\). Here, we instead scale the tangent \(\Gamma_t^{\rho,\tilde r^\rho}\) by a time-dependent stepsize \(\eta_t\). The results below characterize the resulting single-stage canonical update and establish reverse-KL descent for a nondecreasing time-dependent stepsize satisfying \(\eta_1\leq\tau\).

\begin{proposition}%
    \label{prop:47}
    For any terminal density $\rho\in\pdfspace$ and time-dependent stepsize $\eta\in C^1([0,1];\R)$, we define the path dissipation $\Diss_{\rho,\eta,\tilde r^\rho}:\R^d\to\R$ by
    \[
        \Diss_{\rho,\eta,\tilde r^\rho}(x):=\int_{0}^{1}\left(\dot{\eta}_t\left(V_t^\rho[\tilde r^\rho]\left(\flow_{1\to t}^{\rho,\eta,\tilde{r}^\rho}(x)\right)-\E_{\rho}\left[\tilde r^\rho\right]\right)+\eta_t^2\kappa_t\norm[2]{\nabla V_t^\rho[\tilde r^\rho]\left(\flow_{1\to t}^{\rho,\eta,\tilde r^\rho}(x)\right)}^2\right)\dd t,
    \]
    where $\flow^{\rho,\eta,\tilde r^\rho}$ is the ODE flow generated by the velocity field $v^\rho+\eta\Gamma^{\rho,\tilde r^\rho}$. Then, for an ideal Newton step~\eqref{eq:42} where $q=\terminalmap\left(v^\rho+\eta\Gamma^{\rho,\tilde{r}^\rho}\right)$, we have
    \begin{align*}
        q(x)
        &=
        \rho(x)^{1-\eta_1/\tau}
        \pi_{\mu,\tau,r}(x)^{\eta_1/\tau}
        \exp
        \left(
        \frac{\eta_1}{\tau}
        \KL{\rho}{\pi_{\mu,\tau,r}}
        -
        \Diss_{\rho,\eta,\tilde r^\rho}(x)
        \right).
    \end{align*}
\end{proposition}
\begin{proof}
    By Proposition~\ref{prop:45}, letting $f=\tilde{r}^\rho$ gives
    \[
    \begin{aligned}
        q(x)
        =&\rho(x)\exp\left(\eta_1\left(\tilde{r}^\rho(x)-\E_{X\sim\rho}[\tilde{r}^\rho(X)]\right)-\Diss_{\rho,\eta,\tilde{r}^\rho}(x)\right)\\
        =&\rho(x)\exp\left(\frac{\eta_1}{\tau}\left(\log\frac{\pi_{\mu,\tau,r}(x)}{\rho(x)}-\E_{X\sim\rho}\left[\log\frac{\pi_{\mu,\tau,r}(X)}{\rho(X)}\right]\right)-\Diss_{\rho,\eta,\tilde{r}^\rho}(x)\right),
    \end{aligned}
    \]
    which finishes the proof.
\end{proof}

The following three-point identity is obtained by specializing Theorem~\ref{thm:17} to \(f=\tilde r^\rho\). When \(\eta_t\equiv\eta\), it reduces to the constant-stepsize identity in Section~\ref{sub2sec:7}.

\begin{theorem}
\label{thm:18}
Let $q=\terminalmap(v^\rho+\eta\Gamma^{\rho,\tilde r^\rho})$, where $\eta\in C^1([0,1];\R)$ is a time-dependent stepsize. Then, for any density $p\in\pdfspace$, we have
\begin{equation*}
    \KL{p}{q}=\left(1-\frac{\eta_1}{\tau}\right)\KL{p}{\rho}+\frac{\eta_1}{\tau}\left(\KL{p}{\pi}-\KL{\rho}{\pi}\right)+\E_{X\sim p}\left[\Diss_{\rho,\eta,\tilde{r}^\rho}(X)\right].
\end{equation*}
Consequently, when $\eta_1>0$, applying $p=q$ gives
\begin{equation}\label{eq:268}
\KL{\rho}{\pi}-\KL{q}{\pi}=\left(\frac{\tau}{\eta_1}-1\right)\KL{q}{\rho}+\frac{\tau}{\eta_1}\E_{X\sim q}\left[\Diss_{\rho,\eta,\tilde{r}^\rho}(X)\right].
\end{equation}
Applying $p=\pi$ gives
\begin{equation*}
    \KL{\pi}{\rho}-\KL{\pi}{q}=\frac{\eta_1}{\tau}(\KL{\pi}{\rho}+\KL{\rho}{\pi})-\E_{X\sim \pi}\left[\Diss_{\rho,\eta,\tilde{r}^\rho}(X)\right].
\end{equation*}
Applying $p=\rho$ gives
\begin{equation*}
    \KL{\rho}{q}=\E_{X\sim \rho}\left[\Diss_{\rho,\eta,\tilde{r}^\rho}(X)\right].
\end{equation*}
\end{theorem}

\begin{proof}
    By Theorem~\ref{thm:17}, letting $f=\tilde{r}^\rho$ gives
    \[
        \begin{aligned}
            \KL{p}{q}
            =&\KL{p}{\rho}-\eta_1(\E_{X\sim p}[\tilde{r}^\rho(X)]-\E_{X\sim\rho}[\tilde{r}^\rho(X)])+\E_{X\sim p}\left[\Diss_{\rho,\eta,\tilde{r}^\rho}(X)\right]\\
            =&\KL{p}{\rho}-\frac{\eta_1}{\tau}\left(\E_{X\sim p}\left[\log\frac{\pi(X)}{\rho(X)}\right]-\E_{X\sim\rho}\left[\log\frac{\pi(X)}{\rho(X)}\right]\right)+\E_{X\sim p}\left[\Diss_{\rho,\eta,\tilde{r}^\rho}(X)\right]\\
            =&\left(1-\frac{\eta_1}{\tau}\right)\KL{p}{\rho}+\frac{\eta_1}{\tau}\left(\KL{p}{\pi}-\KL{\rho}{\pi}\right)+\E_{X\sim p}\left[\Diss_{\rho,\eta,\tilde{r}^\rho}(X)\right].
        \end{aligned}
    \]
    Applying $p=\rho,\pi,q$ finishes the proof.
\end{proof}

\begin{theorem}%
\label{thm:19}
Let $q=\terminalmap(v^\rho+\eta\Gamma^{\rho,\tilde r^\rho})$. Then, for every nondecreasing time-dependent stepsize $\eta$ which satisfies
    \[
    \forall t\in(0,1),\qquad\eta_t\in(0,\tau],\qquad\dot{\eta}_t\geq0,
    \]
    the reverse KL decreases:
\begin{equation*}%
    \KL{q}{\pi} \le \KL{\rho}{\pi}.
\end{equation*}
Furthermore, since $\pi,\rho,q$ are positive densities on $\R^d$, we have
\[
\KL{q}{\pi}=\KL{\rho}{\pi}\quad\Leftrightarrow\quad\rho=q=\pi.%
\]
\end{theorem}

\begin{proof}
Since $\eta_t>0$ and $\dot{\eta}_t\geq0$ for all $t\in(0,1)$, Proposition~\ref{prop:46} implies $\E_{X\sim q}\left[\Diss_{\rho,\eta,\tilde{r}^\rho}(X)\right]\geq0$. Since $\eta_1\in(0,\tau]$,~\eqref{eq:268} implies $\KL{q}{\pi}\le\KL{\rho}{\pi}$.

The implication $\rho=q=\pi\Rightarrow\KL{q}{\pi}=\KL{\rho}{\pi}$ is immediate. For the opposite direction, assume that $\KL{q}{\pi}=\KL{\rho}{\pi}$. Proposition~\ref{prop:46} and~\eqref{eq:268} imply that $\left(\frac{\tau}{\eta_1}-1\right)\KL{q}{\rho}=0$ and $\E_{X\sim q}\left[\Diss_{\rho,\eta,\tilde{r}^\rho}(X)\right]=0$. Therefore, $\nabla V_t^\rho[\tilde{r}^\rho]\aeq0$. Since $\E_{\rho}[\abs{\tilde{r}^\rho}]<\infty$ and the convolution kernel is a nondegenerate Gaussian, $V_t^\rho[\tilde{r}^\rho]$ is smooth; hence, $V_t^\rho[\tilde{r}^\rho]\equiv c_t$ is constant. Then, Proposition~\ref{prop:7} implies $\rho=\pi$; hence, $\Gamma^{\rho,\tilde{r}^\rho}=0$, which gives $q=\rho$.
\end{proof}

Theorem~\ref{thm:19} establishes one-stage reverse-KL descent for any nondecreasing time-dependent stepsize satisfying \(\eta_t\in(0,\tau]\).

\section{Extensions to Alternative Coordinates and Interpolants}
\label{app:8}

This appendix develops two complementary representation-level extensions of Newton Matching. Appendix~\ref{subsec:56} changes the prediction coordinate: it rewrites regression objectives for exact and approximate tangential updates and for canonicalization in score and drift coordinates. Appendix~\ref{subsec:57} instead shows that Newton Matching extends beyond the standard CFM construction adopted in the main text, which uses a Gaussian source and a linear interpolant. As a representative alternative, it develops the corresponding canonical representations and regression objectives for a one-sided interpolant with deterministic initial state \(X_0=0\). In both extensions, the density-space objective, Fisher--Rao geometry, and Newton direction remain unchanged; only the prediction coordinate or interpolant construction used to realize a Newton Matching stage is modified.

\subsection{Newton Matching under Score and Drift Coordinates}\label{subsec:56}

Both exact and approximate Newton Matching are formulated through velocity-field regression. At one stage, the tangential-update and canonicalization targets share the generic conditional-expectation representation
\begin{align}
    \hat v_t(x_t)
    =&\E_{\BY_{[t,1]}\sim\mathbb P_{[t,1]|t}^{\hat\rho}(\cdot|x_t)}
    \left[\target^{\mathrm{vel}}(t,\BY_{[t,1]})\right],
    \label{eq:269}
\end{align}
where $\target^{\mathrm{vel}}$ is a sample-wise velocity target and $\hat\rho\in\pdfspace$ denotes the terminal density associated with the conditional path distribution. For endpoint-based sampling,~\eqref{eq:269} reduces to
\[
    \hat v_t(x_t)
    =
    \E_{X_1\sim p_{1|t}^{\hat\rho}(\cdot|x_t)}
    \left[\widehat{\target}^{\mathrm{vel}}(t,x_t,X_1)\right].
\]
The corresponding velocity-coordinate regression loss is
\begin{equation}
    \mathcal L^{\mathrm{vel}}(\theta)
    :=
    \E_{\substack{
        t\sim\U(0,1),\,
        \BY_{[t,1]}\sim\Xi_{[t,1]}^{\hat\rho}}}
    \left[
        \norm[2]{v_t^\theta(Y_t)-\target^{\mathrm{vel}}(t,\BY_{[t,1]})}^2
    \right].
    \label{eq:270}
\end{equation}
Here $\Xi_{[t,1]}^{\hat\rho}$ follows the requirement in~\eqref{eq:95}. By definition, the population minimizer of~\eqref{eq:270} is exactly the field in~\eqref{eq:269}. When the target involves stop-gradient operations, the corresponding regression loss can be defined analogously, following the construction in Section~\ref{subsec:25}.

We now express this regression template in score and drift coordinates.

\subsubsection{Score Coordinate}

\paragraph{Relation between score and velocity.}

Under Gaussian source and linear interpolant, the conditional score is
\[
    s_{t|1}(x_t|x_1)
    :=
    \nabla_{x_t}\log p_{t|1}(x_t|x_1)
    =
    -\frac{x_t-\alpha_t x_1}{\beta_t^2},
    \qquad
    t\in[0,1).
\]
It is related to the conditional velocity by
\[
    v_{t|1}(x_t|x_1)
    =
    \frac{\dot\alpha_t}{\alpha_t}x_t
    +
    \kappa_t s_{t|1}(x_t|x_1),
    \qquad
    t\in(0,1).
\]
Given a terminal density $\rho\in\pdfspace$, define its \textit{canonical score} $s^\rho$ by
\[
    s_t^\rho(x_t):=\nabla_{x_t}\log p_t^\rho(x_t).
\]
It admits the conditional-expectation representation
\[
    s_t^\rho(x_t)
    =
    \E_{X_1\sim p_{1|t}^\rho(\cdot|x_t)}
    \left[s_{t|1}(x_t|X_1)\right],
    \qquad
    t\in[0,1).
\]
We then have the following velocity--score relation
\begin{equation}
    v_t^\rho(x_t)
    =
    \frac{\dot\alpha_t}{\alpha_t}x_t
    +
    \kappa_t s_t^\rho(x_t),
    \qquad
    t\in(0,1).
    \label{eq:271}
\end{equation}

\paragraph{Tangential updates in score coordinates.}

The relation above defines an affine change of prediction coordinates at every interior time. For the velocity field $\hat v$ in~\eqref{eq:269}, define its score-coordinate representation by
\[
    \hat s_t(x_t)
    :=
    \frac{1}{\kappa_t}
    \left(
        \hat v_t(x_t)
        -
        \frac{\dot\alpha_t}{\alpha_t}x_t
    \right),
    \qquad
    t\in(0,1).
\]
When $\hat v$ is canonical, $\hat s$ coincides with the marginal score of the associated canonical probability path. For a general noncanonical field $\hat v$, however, $\hat s$ is only its score-coordinate representation.

Applying the same coordinate transformation to the sample-wise velocity target, define
\[
    \target^{\mathrm{score}}(t,\BY_{[t,1]})
    :=
    \frac{1}{\kappa_t}
    \left(
        \target^{\mathrm{vel}}(t,\BY_{[t,1]})
        -
        \frac{\dot\alpha_t}{\alpha_t}Y_t
    \right).
\]
Since $Y_t=x_t$ under the conditional path distribution,~\eqref{eq:269} becomes
\begin{align*}
    \hat s_t(x_t)
    =&\E_{\BY_{[t,1]}\sim\mathbb P_{[t,1]|t}^{\hat\rho}(\cdot|x_t)}
    \left[\target^{\mathrm{score}}(t,\BY_{[t,1]})\right].
\end{align*}

Suppose that the velocity model is parameterized through a score model as
\[
    v_t^\theta(x)
    =
    \frac{\dot\alpha_t}{\alpha_t}x
    +
    \kappa_t s_t^\theta(x).
\]
Then, the velocity loss in~\eqref{eq:270} can be written as
\begin{align*}
    \mathcal L^{\mathrm{vel}}(\theta)
    =&\E_{\substack{
        t\sim\U(0,1),\,
        \BY_{[t,1]}\sim\Xi_{[t,1]}^{\hat\rho}}}
    \left[
        \norm[2]{
            \kappa_t s_t^\theta(Y_t)
            +
            \frac{\dot\alpha_t}{\alpha_t}Y_t
            -
            \target^{\mathrm{vel}}(t,\BY_{[t,1]})
        }^2
    \right]\\
    =&\E_{\substack{
        t\sim\U(0,1),\,
        \BY_{[t,1]}\sim\Xi_{[t,1]}^{\hat\rho}}}
    \left[
        \kappa_t^2
        \norm[2]{
            s_t^\theta(Y_t)
            -
            \target^{\mathrm{score}}(t,\BY_{[t,1]})
        }^2
    \right].
\end{align*}
More generally, consider the weighted score-coordinate regression loss
\[
    \mathcal L_\omega^{\mathrm{score}}(\theta)
    :=
    \E_{\substack{
        t\sim\U(0,1),\,
        \BY_{[t,1]}\sim\Xi_{[t,1]}^{\hat\rho}}}
    \left[
        \omega_t(Y_t)
        \norm[2]{
            s_t^\theta(Y_t)
            -
            \target^{\mathrm{score}}(t,\BY_{[t,1]})
        }^2
    \right],
\]
where the positive weight $\omega_t(x_t)>0$ preserves the same population minimizer and may be chosen to mitigate endpoint singularities. The choice $\omega_t\equiv\kappa_t^2$ makes $\mathcal L_\omega^{\mathrm{score}}(\theta)$ and $\mathcal L^{\mathrm{vel}}(\theta)$ equal sample-wise.

\paragraph{Examples of tangential updates.}

The coordinate transformation above applies directly to the exact and approximate tangential-update losses. In exact Newton Matching, the tangential update in score coordinates is
\[
    \hat s_t
    =
    s_t^\rho
    +
    \frac{\eta}{\kappa_t}\Gamma_t^{\rho,\tilde r^\rho}
    =
    s_t^\rho
    +
    \eta\nabla V_t^\rho[\tilde r^\rho].
\]
For example, the covariance-form of exact Newton Matching in~\eqref{eq:67}--\eqref{eq:68} becomes
\begin{align*}
    \calL_{\rho,\eta,B,\omega}^{\mathrm{cov},\mathrm{score}}(\theta)
    :=&
    \E_{\substack{
        t\sim\U(0,1),\,(X_t,X_1)\sim\Xi_{t,1}^\rho}}
    \Big[
        \omega_t(X_t)
        \big\|
            s_t^\theta(X_t)
            -
            s_{t|1}(X_t|X_1)\\
    &\qquad\qquad
            -
            \eta
            \left(
                \tilde r^\rho(X_1)-B_t(X_t)
            \right)
            \left(
                s_{t|1}(X_t|X_1)-s_t^\rho(X_t)
            \right)
        \big\|_2^2
    \Big].
\end{align*}
Here $\tilde r^\rho(X_1)$ can be computed through the ODE-based or SDE-based methods introduced in Section~\ref{subsec:16}, with the canonical velocity $v^\rho$ recovered from the canonical score through~\eqref{eq:271}.

The Bolza realization of exact Newton Matching in~\eqref{eq:109}--\eqref{eq:110} becomes
\begin{align*}
    \calL_{\rho,\eta,\omega}^{\mathrm{grad,Bolza},\mathrm{score}}(\theta)
    :=
    \E_{\substack{
            t\sim\U(0,1),\,\BY_{[t,1]}\sim\Xi_{[t,1]}^\rho}}
        \bigg[
            \omega_t(Y_t)\Big\|&
               s_t^\theta(Y_t)
                -
    \left(
        1-\frac{\eta}{\tau}
    \right)
    s_{t|1}(Y_t|Y_1)\\
    &-
    \frac{\eta}{\tau}
    s_t^{\mu}(Y_t)
    -
    \eta
    \lambda_t^{b^\rho,r,l^{\rho,\mu}}(\BY_{[t,1]})
            \Big\|_2^2
        \bigg].
\end{align*}
Under the reverse construction, the conditional path $\BY_{[t,1]}\sim\mathbb P_{[t,1]|t}^\rho(\cdot|x_t)$ can be sampled from the posterior-preserving SDE
\begin{align*}
    &\dd Y_u^\rho=\left(2\kappa_u s_u^\rho(Y_u^\rho)+\frac{\dot{\alpha}_u}{\alpha_u}Y_u^\rho\right)\dd u+\sqrt{2\kappa_u}\dd W_u,\qquad u\in[t,1],\\
    &Y_t^\rho=x_t.    
\end{align*}
The pathwise adjoint satisfies
\begin{align*}
    &\frac{\dd}{\dd u}
    \lambda_u^{b^\rho,r,l^{\rho,\mu}}(\BY_{[t,1]})
    =
    -\nabla \left(2\kappa_u s_u^\rho(Y_u^\rho)
    +\frac{\dot{\alpha}_u}{\alpha_u}Y_u^\rho\right)^\top
    \lambda_u^{b^\rho,r,l^{\rho,\mu}}(\BY_{[t,1]})\\
    &\qquad\qquad\qquad\qquad\qquad\qquad\qquad
    +
    \frac{\kappa_u}{\tau}
    \nabla \left(
    \norm[2]{
        s_u^\rho(Y_u)
        -
        s_u^{\mu}(Y_u)
    }^2\right),
    \qquad u\in[t,1],
    \\
    &\lambda_1^{b^\rho,r,l^{\rho,\mu}}(\BY_{[t,1]})
    =
    \nabla r(Y_1).
\end{align*}
In particular, when $\mu=1$, we use \[s_t^1(x_t)=0.\]

As an example of applying the stop-gradient trick in score coordinates, the Gaussian-kernel approximation objective in~\eqref{eq:162} becomes
\begin{align*}
    &\calL_{\eta,\widehat{\Sigma},\omega}^{\mathrm{Gau}\text{-}\mathrm{ker},\mathrm{score},\mathrm{sg}}(\theta)
    :=
    \E_{\substack{
    t\sim\U(0,1),\,(X_t,X_1)\sim \Xi_{t,1}^{\sg(\theta)}}}
    \Bigg[
        \omega_t(X_t)\bigg\|
        s_t^\theta(X_t)
        -
        s_{t|1}(X_t|X_1)
        \notag\\
    &\qquad\qquad
        -
        \eta\frac{\alpha_t}{\beta_t^2}
        \widehat\Sigma_t(X_t)
        \left(
            \nabla r(X_1)
            +
            \frac{1}{\tau}\nabla\log\mu(X_1)
            +
            \frac{\alpha_t}{\tau\beta_t^2}
            \left(
                X_t-\alpha_tX_1
            \right)
        \right)
        \bigg\|_2^2
    \Bigg].
\end{align*}

\paragraph{Score canonicalization.}

We demonstrate how to perform the explicit canonicalization in~\eqref{eq:121} for score coordinates. We provide two options.

Given a noncanonical velocity field $\hat v$ with score-coordinate representation $\hat s$, let
\[
    q=\terminalmap(\hat v)
\]
denote its terminal density. A direct score-coordinate realization for recovering the canonical score $s^q$ is
\[
    \calL^{\mathrm{can},\mathrm{score}}_{\omega}(\theta)
    :=
    \E_{\substack{
        t\sim\U(0,1),\,X_1\sim q,\,X_0\sim p_0,\,
        X_t=\alpha_tX_1+\beta_tX_0}}
    \left[
        \omega_t(X_t)
        \norm[2]{
            s_t^\theta(X_t)
            -
            s_{t| 1}(X_t| X_1)
        }^2
    \right].
\]
The endpoint $X_1\sim q$ can be generated from the ODE
\[
    \frac{\dd Y_t}{\dd t}
    =
    \kappa_t\hat s_t(Y_t)
    +
    \frac{\dot\alpha_t}{\alpha_t}Y_t,
    \qquad
    Y_0\sim p_0,
\]
with $X_1:=Y_1$.

Alternatively, we can generalize the notion of the terminal-density map to score coordinates. Given a score field $\hat s$, consider the SDE
\begin{align*}
    &\dd Y_t=\left(2\kappa_t \hat{s}_t(Y_t)+\frac{\dot{\alpha}_t}{\alpha_t}Y_t\right)\dd t+\sqrt{2\kappa_t}\dd W_t,\qquad t\in(0,1),\\
    &Y_0\sim p_0,
\end{align*}
and define
\[
    \hat q
    :=
    \terminalmap^{\mathrm{score}}(\hat s)
\]
to be its terminal density. We again set $X_1:=Y_1$. For a canonical score $s^\rho$, this score-coordinate terminal-density map agrees with the original terminal-density map:
\[
    \terminalmap^{\mathrm{score}}(s^\rho)
    =
    \terminalmap(v^\rho)
    =
    \rho.
\]
By contrast, for a general noncanonical velocity field $\hat v$ and its score-coordinate representation $\hat s$, the identity
\[
    \terminalmap^{\mathrm{score}}(\hat s)
    =
    \terminalmap(\hat v)
\]
need not hold. Nevertheless, $\hat q$ is a well-defined terminal density induced by the score field $\hat s$, and endpoint samples $X_1\sim\hat q$ can be used in the standard conditional score regression. The corresponding SDE-based canonicalization loss is
\[
    \widehat{\calL}^{\mathrm{can},\mathrm{score}}_{\omega}(\theta)
    :=
    \E_{\substack{
        t\sim\U(0,1),\,X_1\sim \hat q,\,X_0\sim p_0,\,
        X_t=\alpha_tX_1+\beta_tX_0}}
    \left[
        \omega_t(X_t)
        \norm[2]{
            s_t^\theta(X_t)
            -
            s_{t| 1}(X_t| X_1)
        }^2
    \right].
\]
Its population minimizer is exactly the canonical score $s_t^{\hat q}(x_t)$ of the SDE-induced terminal density $\hat q$.

More broadly, the canonical-score assignment \(\rho\mapsto s^\rho\) and the score-coordinate terminal-density map \(\terminalmap^{\mathrm{score}}\) suggest a score-space counterpart to the pair \((\canonicalmap,\terminalmap)\). Developing this viewpoint into a complete score-space formulation, including an appropriate canonical manifold, transported density-space geometry, canonical retraction, Newton direction, and convergence theory, would require further careful analysis. We leave this extension, together with its drift-coordinate analogue, to future work.

\subsubsection{Drift Coordinate}\label{sub2sec:38}

Having rewritten the common regression structure in score coordinates, we now carry out the analogous construction in drift coordinates. For simplicity, we consider the diffusion coefficient \(\sqrt{2\kappa_t}\) from the posterior-preserving SDE.

\paragraph{Relation between drift and velocity.} For a marginal probability path $(p_t)_{t\in[0,1]}$ associated with velocity $v$ and score $s$, define the drift by
\[
    b_t(x_t)
    :=
    v_t(x_t)
    +
    \kappa_t s_t(x_t).
\]
The corresponding SDE
\[
    \dd Y_t
    =
    b_t(Y_t)\dd t
    +
    \sqrt{2\kappa_t}\dd W_t
\]
shares the same marginal probability path as the ODE of $v$.

Under Gaussian source and linear interpolant, the conditional drift is
\[
    b_{t|1}(x_t|x_1)
    :=
    v_{t|1}(x_t|x_1)
    +
    \kappa_t s_{t|1}(x_t|x_1)
    =
    2v_{t|1}(x_t|x_1)
    -
    \frac{\dot\alpha_t}{\alpha_t}x_t,
    \qquad
    t\in(0,1).
\]
Given a terminal density $\rho\in\pdfspace$, define its \textit{canonical drift} $b^\rho$ by
\[
    b_t^\rho(x_t)
    :=
    v_t^\rho(x_t)
    +
    \kappa_t s_t^\rho(x_t)
    =
    2v_t^\rho(x_t)
    -
    \frac{\dot\alpha_t}{\alpha_t}x_t,
    \qquad
    t\in(0,1).
\]
It admits the conditional-expectation representation
\[
    b_t^\rho(x_t)
    =
    \E_{X_1\sim p_{1|t}^\rho(\cdot|x_t)}
    \left[b_{t|1}(x_t|X_1)\right].
\]
Equivalently, the canonical velocity can be recovered from the canonical drift through
\begin{equation}
    v_t^\rho(x_t)
    =
    \frac{1}{2}
    \left(
        b_t^\rho(x_t)
        +
        \frac{\dot\alpha_t}{\alpha_t}x_t
    \right),
    \qquad
    t\in(0,1).
    \label{eq:272}
\end{equation}

\paragraph{Tangential updates in drift coordinates.}

The relation above defines another affine change of prediction coordinates at every interior time. For the velocity field $\hat v$ in~\eqref{eq:269}, define its drift-coordinate representation by
\[
    \hat b_t(x_t)
    :=
    2\hat v_t(x_t)
    -
    \frac{\dot\alpha_t}{\alpha_t}x_t,
    \qquad
    t\in(0,1).
\]
When $\hat v$ is canonical, $\hat b$ coincides with the marginal drift of the associated canonical probability path. For a general noncanonical field $\hat v$, however, $\hat b$ is only its drift-coordinate representation.

Applying the same coordinate transformation to the sample-wise velocity target, define
\[
    \target^{\mathrm{drift}}(t,\BY_{[t,1]})
    :=
    2\target^{\mathrm{vel}}(t,\BY_{[t,1]})
    -
    \frac{\dot\alpha_t}{\alpha_t}Y_t.
\]
Since $Y_t=x_t$ under the conditional path distribution,~\eqref{eq:269} becomes
\begin{align*}
    \hat b_t(x_t)
    =&\E_{\BY_{[t,1]}\sim\mathbb P_{[t,1]|t}^{\hat\rho}(\cdot|x_t)}
    \left[\target^{\mathrm{drift}}(t,\BY_{[t,1]})\right].
\end{align*}

Suppose that the velocity model is parameterized through a drift model as
\[
    v_t^\theta(x)
    =
    \frac{1}{2}
    \left(
        b_t^\theta(x)
        +
        \frac{\dot\alpha_t}{\alpha_t}x
    \right).
\]
Then, the velocity loss in~\eqref{eq:270} can be written as
\begin{align*}
    \mathcal L^{\mathrm{vel}}(\theta)
    =&\E_{\substack{
        t\sim\U(0,1),\,\BY_{[t,1]}\sim\Xi_{[t,1]}^{\hat\rho}}}
    \left[
        \norm[2]{
            \frac{1}{2}
            \left(
                b_t^\theta(Y_t)
                +
                \frac{\dot\alpha_t}{\alpha_t}Y_t
            \right)
            -
            \target^{\mathrm{vel}}(t,\BY_{[t,1]})
        }^2
    \right]\\
    =&\E_{\substack{
        t\sim\U(0,1),\,\BY_{[t,1]}\sim\Xi_{[t,1]}^{\hat\rho}}}
    \left[
        \frac{1}{4}
        \norm[2]{
            b_t^\theta(Y_t)
            -
            \target^{\mathrm{drift}}(t,\BY_{[t,1]})
        }^2
    \right].
\end{align*}
More generally, consider the weighted drift-coordinate regression loss
\[
    \mathcal L_\omega^{\mathrm{drift}}(\theta)
    :=
    \E_{\substack{
        t\sim\U(0,1),\,\BY_{[t,1]}\sim\Xi_{[t,1]}^{\hat\rho}}}
    \left[
        \omega_t(Y_t)
        \norm[2]{
            b_t^\theta(Y_t)
            -
            \target^{\mathrm{drift}}(t,\BY_{[t,1]})
        }^2
    \right],
\]
where the positive weight $\omega_t(x_t)>0$ preserves the same population minimizer. The choice $\omega_t\equiv\frac14$ makes $\mathcal L_\omega^{\mathrm{drift}}(\theta)$ and $\mathcal L^{\mathrm{vel}}(\theta)$ equal sample-wise.

\paragraph{Examples of tangential updates.}

As in score coordinates, the transformation above applies directly to the exact and approximate tangential-update losses. In exact Newton Matching, the tangential update under the drift coordinate is
\[
    \hat b_t
    =
    b_t^\rho
    +
    2\eta\Gamma_t^{\rho,\tilde r^\rho}.
\]
For example, the covariance-form of exact Newton Matching in~\eqref{eq:67}--\eqref{eq:68} becomes
\begin{align*}
    \calL_{\rho,\eta,B,\omega}^{\mathrm{cov},\mathrm{drift}}(\theta)
    :=&
    \E_{\substack{
        t\sim\U(0,1),\,(X_t,X_1)\sim\Xi_{t,1}^\rho}}
    \Big[
        \omega_t(X_t)
        \big\|
            b_t^\theta(X_t)
            -
            b_{t|1}(X_t|X_1)\\
    &\qquad\qquad
            -
            \eta
            \left(
                \tilde r^\rho(X_1)-B_t(X_t)
            \right)
            \left(
                b_{t|1}(X_t|X_1)-b_t^\rho(X_t)
            \right)
        \big\|_2^2
    \Big].
\end{align*}
Here $\tilde r^\rho(X_1)$ can be computed through the ODE-based or SDE-based methods introduced in Section~\ref{subsec:16}, with the canonical velocity $v^\rho$ recovered from the drift through~\eqref{eq:272}.

The Bolza realization of exact Newton Matching in~\eqref{eq:109}--\eqref{eq:110} becomes
\begin{align*}
    \calL_{\rho,\eta,\omega}^{\mathrm{grad,Bolza},\mathrm{drift}}(\theta)
    :=
    \E_{\substack{
            t\sim\U(0,1),\,\BY_{[t,1]}\sim\Xi_{[t,1]}^\rho}}
        \bigg[
            \omega_t(Y_t)\Big\|&
               b_t^\theta(Y_t)
                -
    \left(
        1-\frac{\eta}{\tau}
    \right)
    b_{t|1}(Y_t|Y_1)\\
    &-
    \frac{\eta}{\tau}
    b_t^{\mu}(Y_t)
    -
    2\eta\kappa_t
    \lambda_t^{b^\rho,r,l^{\rho,\mu}}(\BY_{[t,1]})
            \Big\|_2^2
        \bigg].
\end{align*}
Under the reverse construction, the conditional path $\BY_{[t,1]}\sim\mathbb P_{[t,1]|t}^\rho(\cdot|X_t)$ can be sampled directly from the posterior-preserving SDE
\begin{align*}
    &\dd Y_u^\rho=b_u^\rho(Y_u^\rho)\dd u+\sqrt{2\kappa_u}\dd W_u,\qquad u\in[t,1],\\
    &Y_t^\rho=x_t.
\end{align*}
The pathwise adjoint satisfies
\begin{align*}
    &\frac{\dd}{\dd u}
    \lambda_u^{b^\rho,r,l^{\rho,\mu}}(\BY_{[t,1]})
    =
    -\nabla b_u^\rho(Y_u^\rho)^\top
    \lambda_u^{b^\rho,r,l^{\rho,\mu}}(\BY_{[t,1]})\\
    &\qquad\qquad\qquad\qquad\qquad\qquad\qquad
    +
    \frac{1}{4\tau\kappa_u}
    \nabla \left(
    \norm[2]{
        b_u^\rho(Y_u)
        -
        b_u^{\mu}(Y_u)
    }^2\right),
    \qquad u\in[t,1],
    \\
    &\lambda_1^{b^\rho,r,l^{\rho,\mu}}(\BY_{[t,1]})
    =
    \nabla r(Y_1).
\end{align*}
In particular, when $\mu=1$, we use
\[
    b_t^1(x_t)=\frac{\dot\alpha_t}{\alpha_t}x_t.
\]

As an example of applying the stop-gradient trick in drift coordinates, the Gaussian-kernel approximation objective in~\eqref{eq:162} becomes
\begin{align*}
    &\calL_{\eta,\widehat{\Sigma}}^{\mathrm{Gau}\text{-}\mathrm{ker},\mathrm{drift},\mathrm{sg}}(\theta)
    :=
    \E_{\substack{
    t\sim\U(0,1),\,(X_t,X_1)\sim \Xi_{t,1}^{\sg(\theta)}}}
    \Bigg[
        \bigg\|
        b_t^\theta(X_t)
        -
        b_{t|1}(X_t|X_1)
        \notag\\
    &\qquad\qquad
        -
        2\eta\frac{\alpha_t\kappa_t}{\beta_t^2}
        \widehat\Sigma_t(X_t)
        \left(
            \nabla r(X_1)
            +
            \frac{1}{\tau}\nabla\log\mu(X_1)
            +
            \frac{\alpha_t}{\tau\beta_t^2}
            \left(
                X_t-\alpha_tX_1
            \right)
        \right)
        \bigg\|_2^2
    \Bigg].
\end{align*}

\paragraph{Drift canonicalization.}

We demonstrate how to perform the explicit canonicalization in~\eqref{eq:121} for drift coordinates. We provide two options.

Given a noncanonical velocity field $\hat v$ with drift-coordinate representation $\hat b$, let
\[
    q=\terminalmap(\hat v)
\]
denote its terminal density. A direct drift-coordinate realization for recovering the canonical drift $b^q$ is
\[
    \calL^{\mathrm{can},\mathrm{drift}}_{\omega}(\theta)
    :=
    \E_{\substack{
        t\sim\U(0,1),\,X_1\sim q,\,X_0\sim p_0,\,
        X_t=\alpha_tX_1+\beta_tX_0}}
    \left[
        \omega_t(X_t)
        \norm[2]{
            b_t^\theta(X_t)
            -
            b_{t| 1}(X_t| X_1)
        }^2
    \right].
\]
The endpoint $X_1\sim q$ can be generated from the ODE
\[
    \frac{\dd Y_t}{\dd t}
    =
    \frac{1}{2}
    \left(
        \hat b_t(Y_t)
        +
        \frac{\dot\alpha_t}{\alpha_t}Y_t
    \right),
    \qquad
    Y_0\sim p_0,
\]
with $X_1:=Y_1$.

Alternatively, we can generalize the notion of the terminal-density map to drift coordinates. Given a drift field $\hat b$, consider the SDE
\begin{align*}
    &\dd Y_t=\hat b_t(Y_t)\dd t+\sqrt{2\kappa_t}\dd W_t,\qquad t\in(0,1),\\
    &Y_0\sim p_0,
\end{align*}
and define
\[
    \hat q
    :=
    \terminalmap^{\mathrm{drift}}(\hat b)
\]
to be its terminal density. We again set $X_1:=Y_1$. For a canonical drift $b^\rho$, this drift-coordinate terminal-density map agrees with the original terminal-density map:
\[
    \terminalmap^{\mathrm{drift}}(b^\rho)
    =
    \terminalmap(v^\rho)
    =
    \rho.
\]
By contrast, for a general noncanonical velocity field $\hat v$ and its drift-coordinate representation $\hat b$, the identity
\[
    \terminalmap^{\mathrm{drift}}(\hat b)
    =
    \terminalmap(\hat v)
\]
need not hold. Nevertheless, $\hat q$ is a well-defined terminal density induced by the drift field $\hat b$, and endpoint samples $X_1\sim\hat q$ can be used in the standard conditional drift regression. The corresponding SDE-based canonicalization loss is
\[
    \widehat{\calL}^{\mathrm{can},\mathrm{drift}}_{\omega}(\theta)
    :=
    \E_{\substack{
        t\sim\U(0,1),\,X_1\sim \hat q,\,X_0\sim p_0,\,
        X_t=\alpha_tX_1+\beta_tX_0}}
    \left[
        \omega_t(X_t)
        \norm[2]{
            b_t^\theta(X_t)
            -
            b_{t| 1}(X_t| X_1)
        }^2
    \right].
\]
Its population minimizer is exactly the canonical drift $b_t^{\hat q}(x_t)$ of the SDE-induced terminal density $\hat q$.

\subsection{Newton Matching under One-Sided Interpolants}\label{subsec:57}

In this subsection, we extend Newton Matching beyond the standard CFM construction adopted in the main text, which uses a Gaussian source and a linear interpolant. Specifically, we consider a one-sided interpolant \citep{albergo2025stochastic} with deterministic initial state \(X_0=0\). For suitable interpolant schedules, this one-sided construction also admits a Schr\"odinger half-bridge interpretation with a linear-Gaussian reference \citep{peluchetti2023diffusion}. The constructions in Appendices~\ref{app:4} and~\ref{app:5} remain applicable in this setting.

\subsubsection{Canonical Velocity, Score, and Drift}

The linear interpolant \(X_t = \alpha_t X_1 + \beta_t X_0\) in this paper is associated with a standard Gaussian source and the interpolant schedule satisfies the boundary conditions $\alpha_0=0$, $\beta_0=1$, $\alpha_1=1$, and $\beta_1=0$. For the one-sided interpolant considered here, the source is instead the deterministic endpoint $X_0=0$. Thus, for a terminal density $\rho\in\pdfspace$, we consider
\begin{equation}
X_t=\widetilde\alpha_tX_1+\widetilde\beta_tZ,\qquad X_1\sim\rho,\qquad Z\sim\Normal(0,I),
\label{eq:273}
\end{equation}
with the boundary conditions
\begin{equation*}
\widetilde\alpha_0=0,\qquad \widetilde\beta_0=0,\qquad \widetilde\alpha_1=1,\qquad \widetilde\beta_1=0.
\end{equation*}
We assume that $\widetilde\alpha$ and $\widetilde\beta$ are continuous on $[0,1]$, differentiable on $(0,1)$, and strictly positive at every interior time. Define
\[
\widetilde{\kappa}_t:=\frac{\widetilde\beta_t}{\widetilde\alpha_t}\left(\dot{\widetilde\alpha}_t\widetilde\beta_t-\widetilde\alpha_t\dot{\widetilde\beta}_t\right),
\]
and assume $\widetilde\kappa_t>0$ for every $t\in(0,1)$.

For every interior time $t\in(0,1)$, define
\begin{equation*}
\widetilde p_{t|1}(x_t|x_1):=\Normal(x_t;\widetilde\alpha_tx_1,\widetilde\beta_t^2I),\qquad \widetilde p_t^\rho(x_t):=\int_{\R^d}\rho(x_1)\widetilde p_{t|1}(x_t|x_1)\dd x_1.
\end{equation*}
The corresponding endpoint posterior is
\begin{equation*}
\widetilde p_{1|t}^\rho(x_1|x_t):=\frac{\rho(x_1)\widetilde p_{t|1}(x_t|x_1)}{\widetilde p_t^\rho(x_t)}.
\end{equation*}

The conditional velocity associated with~\eqref{eq:273} is
\begin{equation*}
\widetilde v_{t|1}(x_t|x_1)=\frac{\dot{\widetilde\beta}_t}{\widetilde\beta_t}x_t+\frac{\widetilde\alpha_t\widetilde{\kappa}_t}{\widetilde\beta_t^2}x_1.
\end{equation*}
Under~\eqref{eq:273}, we have
\[
\widetilde v_{t|1}(X_t|X_1)=\dot{\widetilde\alpha}_tX_1+\dot{\widetilde\beta}_tZ.
\]
Under this one-sided setting, we define the \textit{canonical velocity field} of $\rho$ by
\begin{equation*}
\widetilde v_t^\rho(x_t):=\E_{X_1\sim\widetilde p_{1|t}^\rho(\cdot|x_t)}\left[\widetilde v_{t|1}(x_t|X_1)\right]=\frac{\dot{\widetilde\alpha}_t}{\widetilde\alpha_t}x_t+\widetilde\kappa_t\nabla\log\widetilde p_t^\rho(x_t).
\end{equation*}

The conditional score is
\[
\widetilde{s}_{t|1}(x_t|x_1):=\nabla_{x_t}\log\widetilde p_{t|1}(x_t|x_1)=-\frac{x_t-\widetilde\alpha_tx_1}{\widetilde\beta_t^2}.
\]
The \textit{canonical score} is
\[
\widetilde s_t^\rho(x_t):=\E_{X_1\sim\widetilde p_{1|t}^\rho(\cdot|x_t)}\left[\widetilde s_{t|1}(x_t|X_1)\right]=\nabla_{x_t}\log\widetilde p_{t}^\rho(x_t).
\]

Define the conditional drift as
\begin{equation*}
\widetilde b_{t|1}(x_t|x_1):=\widetilde v_{t|1}(x_t|x_1)+\widetilde\kappa_t\nabla_{x_t}\log\widetilde p_{t|1}(x_t|x_1)=\widetilde v_{t|1}(x_t|x_1)-\frac{\widetilde\kappa_t}{\widetilde\beta_t^2}(x_t-\widetilde\alpha_tx_1),
\end{equation*}
and its marginalization gives the \textit{canonical drift}
\begin{equation*}
\widetilde b_t^\rho(x_t):=\E_{X_1\sim\widetilde p_{1|t}^\rho(\cdot|x_t)}\left[\widetilde b_{t|1}(x_t|X_1)\right]=\widetilde v_t^\rho(x_t)+\widetilde\kappa_t\nabla\log\widetilde p_t^\rho(x_t)=2\widetilde v_t^\rho(x_t)-\frac{\dot{\widetilde\alpha}_t}{\widetilde\alpha_t}x_t.
\end{equation*}

In particular, for $\mu=1$, we use
\[
    \widetilde s_t^1(x)=0,
    \qquad
    \widetilde v_t^{1}(x)
    =
    \widetilde b_t^1(x)
    =
    \frac{\dot{\widetilde\alpha}_t}{\widetilde\alpha_t}x.
\]

Consider the posterior-preserving SDE
\begin{subequations}\label{eq:274}
    \begin{align}
        &\dd Y_u=\widetilde b_u^\rho(Y_u)\dd u+\sqrt{2\widetilde\kappa_u}\dd W_u,\qquad u\in[t,1],\\
        &Y_t=x_t.
    \end{align}
\end{subequations}
Then, we have
\[
Y_1\sim \widetilde p_{1|t}^\rho(\cdot|x_t).
\]
The path distribution is accordingly denoted by
\[
\BY_{[t,1]}:=(Y_u)_{u\in[t,1]}\sim\widetilde{\mathbb{P}}_{[t,1]|t}^\rho(\cdot|x_t).
\]
Conditioning further on an endpoint $Y_u=x_u$, where $t<u\leq1$, gives the universal bridge in Theorem~\ref{thm:14}:
\[
\BY_{[t,u]}|(Y_t=x_t,Y_u=x_u)\sim \widetilde{\mathbb{P}}_{[t,u]|t,u}^\rho(\cdot|x_t,x_u)=\widetilde{\mathbb{P}}_{[t,u]}^\mathrm{uni}(\cdot|x_t,x_u).
\]
Here, we use the notation $\widetilde{\mathbb{P}}_{[t,u]}^\mathrm{uni}(\cdot|x_t,x_u)$ to emphasize that the Gaussian coefficients in Theorem~\ref{thm:14} are evaluated with $\widetilde\alpha$ and $\widetilde\beta$.

Whenever the specific construction is immaterial, we use the generic notation
\[
(X_t,X_1)\sim\widetilde\Xi_{t,1}^{\hat\rho}\quad\Rightarrow\quad 
X_1|X_t\sim\widetilde{p}_{1|t}^{\hat\rho}(\cdot|X_t)
\qquad\text{for endpoint sampling},
\]
and
\[
\BY_{[t,1]}\sim\widetilde\Xi_{[t,1]}^{\hat\rho}\quad\Rightarrow\quad 
Y_{[t,1]}|Y_t\sim\widetilde{\mathbb{P}}_{[t,1]|t}^{\hat\rho}(\cdot|Y_t)
\qquad\text{for path sampling}.
\]
Here, the terminal density \(\hat\rho\) is typically \(\rho\) or, in certain cases, \(\rho^\base\). The marginal distribution of \(X_t\) or \(Y_t\) is left implicit.

\subsubsection{Training Losses}

The regression constructions in Sections~\ref{sec:5} and~\ref{sec:6}, together with their score-coordinate and drift-coordinate reformulations in Appendix~\ref{subsec:56}, carry over to the one-sided setting by replacing \((\alpha_t,\beta_t,\kappa_t,p_t^\rho,p_{1|t}^\rho,
    \mathbb P_{[t,1]|t}^\rho)\) with \((\widetilde\alpha_t,\widetilde\beta_t,\widetilde\kappa_t,
    \widetilde p_t^\rho,\widetilde p_{1|t}^\rho,
    \widetilde{\mathbb P}_{[t,1]|t}^\rho)\).

The forward construction in the one-sided setting is given by
\[
t\sim\U(0,1),\qquad X_1\sim\rho,\qquad Z\sim\Normal(0,I),\qquad X_t=\widetilde\alpha_tX_1+\widetilde\beta_tZ.
\]
In particular, $X_1\sim\rho$ can be sampled from the SDE
    \begin{align*}
        &\dd Y_t=\widetilde b_t^\rho(Y_t)\dd t+\sqrt{2\widetilde\kappa_t}\dd W_t,\qquad t\in[0,1],\\
        &Y_0=0,
    \end{align*}
with $X_1:=Y_1$.

The reverse construction is given by
\[
t\sim\U(0,1),\qquad X_t\sim\hat{p}_t,\qquad X_1\sim \widetilde{p}_{1|t}^\rho(\cdot|X_t)\text{ or }\BY_{[t,1]}\sim\widetilde{\mathbb{P}}_{[t,1]|t}^\rho(\cdot|X_t).
\]
In particular, $X_1\sim \widetilde{p}_{1|t}^\rho(\cdot|X_t)$ or $\BY_{[t,1]}\sim\widetilde{\mathbb{P}}_{[t,1]|t}^\rho(\cdot|X_t)$ can both be sampled from SDE~\eqref{eq:274}. 

Below we provide some examples of performing exact or approximate tangential updates in the one-sided setting.

The covariance form of exact Newton Matching in~\eqref{eq:67}--\eqref{eq:68} can be generalized to the one-sided setting as:
\begin{align*}
    \widetilde{\calL}_{\rho,\eta,B,\omega}^{\mathrm{cov},\mathrm{vel}}(\theta)
    :=&
    \E_{\substack{
        t\sim\U(0,1),\,(X_t,X_1)\sim\widetilde\Xi_{t,1}^\rho}}
    \Big[
        \omega_t(X_t)\big\|
            v_t^\theta(X_t)
            -\widetilde v_{t|1}(X_t|X_1)\\
            &\qquad
        -\eta
        \left(
            \tilde r^\rho(X_1)-B_t(X_t)
        \right)
        \left(
            \widetilde v_{t|1}(X_t|X_1)-\widetilde v_t^\rho(X_t)
        \right)
            \big\|_2^2
    \Big],
\end{align*}
whose unique population minimizer is
\[
\widetilde v^\rho+\eta\widetilde{\Gamma}^{\rho,\tilde{r}^\rho}.
\]
Here we define
\[
\widetilde{\Gamma}^{\rho,f}_t(x_t):=\Cov_{X_1\sim\widetilde{p}_{1|t}^\rho(\cdot|x_t)}\left(\tilde{v}_{t|1}(x_t|X_1),f(X_1)\right)=\widetilde\kappa_t\nabla_{x_t}\E_{X_1\sim\widetilde{p}_{1|t}^\rho(\cdot|x_t)}\left[f(X_1)\right].
\]

The Bolza realization of exact Newton Matching in~\eqref{eq:109}--\eqref{eq:110} can be generalized to the one-sided setting with drift coordinates as:
\begin{align*}
    \widetilde{\calL}_{\rho,\eta,\omega}^{\mathrm{grad,Bolza},\mathrm{drift}}(\theta)
    :=
    \E_{\substack{
            t\sim\U(0,1),\,\BY_{[t,1]}\sim\widetilde\Xi_{[t,1]}^\rho}}
        \bigg[
            \omega_t(Y_t)\Big\|&
               b_t^\theta(Y_t)
                -
    \left(
        1-\frac{\eta}{\tau}
    \right)
    \widetilde b_{t|1}(Y_t|Y_1)\\
    &-
    \frac{\eta}{\tau}
    \widetilde b_t^{\mu}(Y_t)
    -
    2\eta\widetilde\kappa_t
    \lambda_t^{\widetilde b^\rho,r,\widetilde l^{\rho,\mu}}(\BY_{[t,1]})
            \Big\|_2^2
        \bigg].
\end{align*}
The pathwise adjoint satisfies
\begin{align*}
    &\frac{\dd}{\dd u}
    \lambda_u^{\widetilde b^\rho,r,\widetilde l^{\rho,\mu}}(\BY_{[t,1]})
    =
    -\nabla \widetilde b_u^\rho(Y_u^\rho)^\top
    \lambda_u^{\widetilde b^\rho,r,\widetilde l^{\rho,\mu}}(\BY_{[t,1]})\\
    &\qquad\qquad\qquad\qquad\qquad\qquad\qquad
    +
    \frac{1}{4\tau\widetilde \kappa_u}
    \nabla \left(
    \norm[2]{
        \widetilde b_u^\rho(Y_u)
        -
        \widetilde b_u^{\mu}(Y_u)
    }^2\right),
    \qquad u\in[t,1],
    \\
    &\lambda_1^{\widetilde{b}^\rho,r,\widetilde l^{\rho,\mu}}(\BY_{[t,1]})
    =
    \nabla r(Y_1).
\end{align*}

The Gaussian-kernel approximation with the stop-gradient trick in~\eqref{eq:162} can be generalized to the one-sided setting with score coordinates as:
\begin{align*}
    &\widetilde{\calL}_{\eta,\widehat{\Sigma},\omega}^{\mathrm{Gau}\text{-}\mathrm{ker},\mathrm{score},\mathrm{sg}}(\theta)
    :=
    \E_{\substack{
    t\sim\U(0,1),\,(X_t,X_1)\sim \widetilde\Xi_{t,1}^{\sg(\theta)}}}
    \Bigg[
        \omega_t(X_t)\bigg\|
        s_t^\theta(X_t)
        -
        \widetilde{s}_{t|1}(X_t|X_1)
        \notag\\
    &\qquad\qquad
        -
        \eta\frac{\widetilde\alpha_t}{\widetilde\beta_t^2}
        \widehat\Sigma_t(X_t)
        \left(
            \nabla r(X_1)
            +
            \frac{1}{\tau}\nabla\log\mu(X_1)
            +
            \frac{\widetilde\alpha_t}{\tau\widetilde\beta_t^2}
            \left(
                X_t-\widetilde\alpha_tX_1
            \right)
        \right)
        \bigg\|_2^2
    \Bigg].
\end{align*}

The canonicalization losses can be generalized accordingly.

\section{Detailed Derivations for Existing Works}\label{app:9}

This appendix provides the detailed derivations underlying the correspondences stated in Section~\ref{sec:7}.

\subsection{Tilt Matching}\label{subsec:58}

Explicit Tilt Matching (ETM) and Implicit Tilt Matching (ITM) are organized along the same exponential-tilt continuation path, but differ in the population-level update realized at each stage. Following the notation of Section~\ref{subsec:26}, we relabel the current continuation density, its canonical velocity, and the inverse-temperature increment by \(\rho^\base\), \(v^\base=\canonicalmap(\rho^\base)\), and \(\tau\), respectively. We analyze the two methods separately.

\subsubsection{Explicit Tilt Matching}\label{sub2sec:39}

Using our notation, the ETM loss is
\begin{align*}
    \mathcal{L}^{\mathrm{ETM}}(\theta)
    :=&\E_{\substack{t\sim\U(0,1),\,X_1\sim\rho^\base,\,X_0\sim p_0,\,X_t=\alpha_tX_1+\beta_tX_0}}
    \bigg[\Big\|v_t^\theta(X_t)-v_t^\base(X_t)\notag\\
    &\qquad\qquad\qquad\qquad
    -\tau\left(v_{t|1}(X_t|X_1) r(X_1)-v_t^\base(X_t) r(X_1)\right)\Big\|_2^2\bigg].
\end{align*}
Rearranging the ETM target gives
\begin{align*}
    &v_t^\theta(X_t)-v_t^\base(X_t)-\tau\left(v_{t|1}(X_t|X_1) r(X_1)-v_t^\base(X_t) r(X_1)\right)\\
    =&v_t^\theta(X_t)-v_{t|1}(X_t|X_1)+v_{t|1}(X_t|X_1)-v_t^\base(X_t)-\tau r(X_1)\left(v_{t|1}(X_t|X_1)-v_t^\base(X_t)\right)\\
    =&v_t^\theta(X_t)-v_{t|1}(X_t|X_1)-\tau\left(r(X_1)-\frac{1}{\tau}\right)\left(v_{t|1}(X_t|X_1)-v_t^\base(X_t)\right).
\end{align*}
Comparing the ETM loss with our unregularized covariance-form loss~\eqref{eq:148} yields the exact sample-wise correspondence
\[
\mathcal{L}^{\mathrm{ETM}}(\theta)=\where{\calL_{\rho,\eta,B}^{\mathrm{no}\text{-}\mathrm{reg}\text{-}\mathrm{cov}}(\theta)}{
    \rho=\rho^\base,\,\eta=\tau,\,B_t\equiv\frac{1}{\tau}
}
\qquad\text{under the forward construction.}
\]
Therefore, ETM is the covariance form of the full-step unregularized update at the base anchor with the forward construction and a constant baseline. It is not critical-point consistent whenever $r$ is nonconstant. Its unique population minimizer is
\[
\bar{v}^{\mathrm{ETM}}=v^\base+\tau\Gamma^{\rho^\base,r}.
\]
Theorem~\ref{thm:2} gives the terminal density $q^{\mathrm{ETM}}:=\terminalmap(\bar{v}^{\mathrm{ETM}})$ as
\[
q^{\mathrm{ETM}}(x)=\pi_{\rho^\base,\tau,r}(x)\exp\left(\KL{\rho^\base}{\pi_{\rho^\base,\tau,r}}-\Diss_{\rho^\base,\tau,r}(x)\right).
\]

The original analysis in Tilt Matching attributes the mismatch between $\bar{v}^{\mathrm{ETM}}$ and $v^{\pi_{\rho^\base,\tau,r}}$ to the discretization error defined in \cite[Section 3.1]{potaptchik2026tilt}. Newton Matching further separates this mismatch into two mechanisms through the decomposition
\[
    \bar{v}^{\mathrm{ETM}}
    -
    v^{\pi_{\rho^\base,\tau,r}}
    =
    \left(
        \bar{v}^{\mathrm{ETM}}-v^{q^{\mathrm{ETM}}}
    \right)
    +
    \left(
        v^{q^{\mathrm{ETM}}}
        -
        v^{\pi_{\rho^\base,\tau,r}}
    \right).
\]
The first term is the velocity-level \textit{canonicality error} since $\bar{v}^{\mathrm{ETM}}\notin\velmancan$ in general. The second term arises from the terminal-density mismatch induced by the \textit{path dissipation} $\Diss_{\rho^\base,\tau,r}$.

\subsubsection{Implicit Tilt Matching}

Using our notation, the ITM loss is
\begin{align*}
    \mathcal{L}_c^{\mathrm{ITM}}(\theta)
    :=&\E_{\substack{t\sim\U(0,1),\,X_1\sim\rho^\base,\,X_0\sim p_0,\,X_t=\alpha_tX_1+\beta_tX_0}}
    \bigg[\Big\|c(X_t)\left(v_t^\theta(X_t)-v_t^\base(X_t)\right)\notag\\
    &\qquad\quad
    +\left(e^{\tau r(X_1)}-c(X_t)\right)\left(\sg\left(v_t^\theta(X_t)\right)-v_{t|1}(X_t|X_1)\right)\Big\|_2^2\bigg],
\end{align*}
where $c(x_t)>0$. Define the corresponding baseline and weight by
\[
B_t^c(x_t):=\frac{1}{\tau}\log c(x_t),\qquad
\omega_t^c(x_t):=c(x_t)^2.
\]
The ITM loss can be rewritten as
\begin{align*}
    \mathcal{L}_c^{\mathrm{ITM}}(\theta)
    =&\E_{\substack{t\sim\U(0,1),\,X_t\sim p_t^\base,\,X_1\sim p_{1|t}^\base(\cdot|X_t)}}
    \bigg[\omega_t^c(X_t)\Big\|v_t^\theta(X_t)-v_t^\base(X_t)\notag\\
    &\qquad\quad
    +\left(e^{\tau (r(X_1)-B_t^c(X_t))}-1\right)\left(\sg\left(v_t^\theta(X_t)\right)-v_{t|1}(X_t|X_1)\right)\Big\|_2^2\bigg].
\end{align*}

To match the ITM weighting exactly, we generalize the direct-linearization loss~\eqref{eq:129} by introducing a positive weight $\omega_t(x_t)>0$:
\begin{align}
    &\calL_{\rho,\eta,B,\omega}^{\mathrm{dir}\text{-}\mathrm{lin}}(\theta)
    :=
    \E_{\substack{
    t\sim\U(0,1),\,
    (X_t,X_1)\sim \Xi_{t,1}^\base}}
    \bigg[\omega_t(X_t)
    \Big\|
    v_t^\theta(X_t)
    -
    \left[
        \left(1-\frac{\eta}{\tau}\right)v_t^\rho(X_t)
        +
        \frac{\eta}{\tau}v_t^\base(X_t)
    \right]
    \notag\\
    &\quad
    -
    \left(
        e^{\tau(r(X_1)-B_t(X_t))}-1
    \right)
    \left[
        \left(1-\frac{\eta}{\tau}\right)v_t^\rho(X_t)
        +
        \frac{\eta}{\tau}v_{t|1}(X_t|X_1)
        -
        \sg\left(v_t^\theta(X_t)\right)
    \right]
    \Big\|_2^2
    \bigg].
    \label{eq:275}
\end{align}
Here, $\Xi_{t,1}^\base:=\Xi_{t,1}^{\rho^\base}$ satisfies the posterior requirement in~\eqref{eq:60}. At the function-space level, the positive weight $\omega_t(x_t)$ does not change the pointwise population-stationarity condition. Comparing the ITM loss $\mathcal{L}_c^{\mathrm{ITM}}$ with~\eqref{eq:275} yields the exact sample-wise correspondence
\[
\mathcal{L}_c^{\mathrm{ITM}}(\theta)=\where{\calL_{\rho,\eta,B,\omega}^{\mathrm{dir}\text{-}\mathrm{lin}}(\theta)}{\rho=\rho^\base,\,\eta=\tau,\,B_t=B_t^c,\,\omega_t=\omega_t^c}
\qquad\text{under the forward construction.}
\]
Therefore, ITM is equivalent to the full-step update of the direct-linearization approximation under the forward construction, with specific choices of the baseline and weighting. Its unique population-stationary point is
\[
v^{\mathrm{ITM}}=v^{\pi_{\rho^\base,\tau,r}}.
\]
Consequently, ITM is critical-point consistent.

\subsection{DiffusionNFT}\label{subsec:59}

Using our notation, the DiffusionNFT loss is
\begin{align*}
    \mathcal{L}_{\rho,c}^{\mathrm{NFT}}(\theta)
    :=&\E_{\substack{t\sim\U(0,1),\,X_1\sim\rho,\,X_0\sim p_0,\\X_t=\alpha_tX_1+\beta_tX_0}}
    \bigg[
    r(X_1)\norm[2]{(1-c)v_t^\rho(X_t)+c v_t^\theta(X_t)-v_{t|1}(X_t|X_1)}^2\\
    &\qquad\qquad\quad
    +(1-r(X_1))\norm[2]{(1+c)v_t^\rho(X_t)-c v_t^\theta(X_t)-v_{t|1}(X_t|X_1)}^2
    \bigg],
\end{align*}
where $c>0$ is a user-specified scalar value. To expose the regression target without changing the sample-wise gradient with respect to $\theta$, define
\[
u_t^\theta(x_t):=v_t^\theta(x_t)-v_t^\rho(x_t),\qquad
u_t^\rho(x_t,x_1):=v_t^\rho(x_t)-v_{t|1}(x_t|x_1).
\]
Then, the DiffusionNFT loss $\mathcal{L}_{\rho,c}^{\mathrm{NFT}}$ can be rewritten as
\begin{align*}
    \mathcal{L}_{\rho,c}^{\mathrm{NFT}}(\theta)
    =&\E_{\substack{t\sim\U(0,1),\,X_1\sim\rho,\,X_0\sim p_0,\,X_t=\alpha_tX_1+\beta_tX_0}}
    \bigg[
    r(X_1)\norm[2]{c u_t^\theta(X_t)+u_t^\rho(X_t,X_1)}^2\\
    &\qquad\qquad\qquad\qquad\qquad\qquad
    +(1-r(X_1))\norm[2]{c u_t^\theta(X_t)-u_t^\rho(X_t,X_1)}^2
    \bigg].
\end{align*}

Since
\begin{align*}
    &\norm[2]{c u_t^\theta(X_t)+u_t^\rho(X_t,X_1)}^2-\norm[2]{c u_t^\theta(X_t)-u_t^\rho(X_t,X_1)}^2
    =4c u_t^\theta(X_t)^\top u_t^\rho(X_t,X_1),
\end{align*}
and
\begin{align*}
    &\norm[2]{c u_t^\theta(X_t)-u_t^\rho(X_t,X_1)}^2
    =c^2\norm[2]{u_t^\theta(X_t)}^2+\norm[2]{u_t^\rho(X_t,X_1)}^2
    -2c u_t^\theta(X_t)^\top u_t^\rho(X_t,X_1),
\end{align*}
we have
\begin{align*}
    &r(X_1)\norm[2]{c u_t^\theta(X_t)+u_t^\rho(X_t,X_1)}^2+(1-r(X_1))\norm[2]{c u_t^\theta(X_t)-u_t^\rho(X_t,X_1)}^2\\
    =&c^2\norm[2]{u_t^\theta(X_t)}^2+\norm[2]{u_t^\rho(X_t,X_1)}^2
    +2c(2r(X_1)-1) u_t^\theta(X_t)^\top u_t^\rho(X_t,X_1)\\
    =&\norm[2]{c u_t^\theta(X_t)+(2r(X_1)-1)u_t^\rho(X_t,X_1)}^2+\left(1-(2r(X_1)-1)^2\right)\norm[2]{u_t^\rho(X_t,X_1)}^2.
\end{align*}
Here, the second term $\left(1-(2r(X_1)-1)^2\right)\norm[2]{u_t^\rho(X_t,X_1)}^2$ is independent of $\theta$ and therefore does not affect the sample-wise gradient. The first term satisfies
\begin{align*}
    &\norm[2]{c u_t^\theta(X_t)+(2r(X_1)-1)u_t^\rho(X_t,X_1)}^2\\
    =&c^2\norm[2]{u_t^\theta(X_t)+\frac{2}{c}\left(r(X_1)-\frac12\right)u_t^\rho(X_t,X_1)}^2\\
    =&c^2\norm[2]{v_t^\theta(X_t)-v_t^\rho(X_t)+\frac{2}{c}\left(r(X_1)-\frac12\right)\left(v_t^\rho(X_t)-v_{t|1}(X_t|X_1)\right)}^2\\
    =&c^2\norm[2]{v_t^\theta(X_t)-v_{t|1}(X_t|X_1)+\frac{2}{c}\left(r(X_1)-\frac{c+1}{2}\right)\left(v_t^\rho(X_t)-v_{t|1}(X_t|X_1)\right)}^2.
\end{align*}

Consequently, the DiffusionNFT loss $\mathcal{L}_{\rho,c}^{\mathrm{NFT}}$ can be written as
\begin{align*}
    \mathcal{L}_{\rho,c}^{\mathrm{NFT}}(\theta)
    =&c^2\E_{\substack{t\sim\U(0,1),\,X_1\sim\rho,\,X_0\sim p_0,\,X_t=\alpha_tX_1+\beta_tX_0}}
    \Bigg[
    \bigg\|v_t^\theta(X_t)-v_{t|1}(X_t|X_1)\\
    &\qquad
    +\frac{2}{c}\left(r(X_1)-\frac{c+1}{2}\right)\left(v_t^\rho(X_t)-v_{t|1}(X_t|X_1)\right)\bigg\|_2^2
    \Bigg]+\const,
\end{align*}
where the additive constant is independent of $\theta$. Comparing the DiffusionNFT loss $\mathcal{L}_{\rho,c}^{\mathrm{NFT}}$ with~\eqref{eq:148} yields the exact sample-wise correspondence up to this $\theta$-independent additive term:
\[
\mathcal{L}_{\rho,c}^{\mathrm{NFT}}(\theta)=c^2\cdot\where{\calL_{\rho,\eta,B}^{\mathrm{no}\text{-}\mathrm{reg}\text{-}\mathrm{cov}}(\theta)}{\eta=\frac{2}{c},\,B_t\equiv \frac{c+1}{2}}+\const
\qquad\text{under the forward construction.}
\]
Therefore, DiffusionNFT is the covariance form of the unregularized update under the forward construction, stepsize $\eta=\frac2c$, and constant baseline $B_t\equiv\frac{c+1}2$. Its unique population minimizer is
\[
\bar{v}^{\mathrm{NFT}}=v^\rho+\frac{2}{c}\Gamma^{\rho,r}.
\]
It is not critical-point consistent whenever $r$ is nonconstant.

\subsection{Reinforce Adjoint Matching}\label{subsec:60}

Using our notation, the Reinforce Adjoint Matching (RAM) loss is
\begin{align*}
\mathcal{L}^{\mathrm{RAM}}(\theta)
=&\E_{\substack{t\sim\U(0,1),\,X_1\sim \rho^{\sg(\theta)},\,X_0\sim p_0,\,X_t=\alpha_tX_1+\beta_tX_0}}\bigg[\Big\|v_t^\theta(X_t)-v_t^\base(X_t)\\
&\qquad\qquad\qquad\qquad\qquad
-r(X_1)\left(v_{t|1}(X_t|X_1)-\sg\left(v_t^\theta(X_t)\right)\right)\Big\|_2^2\bigg].
\end{align*}
Here, \(X_1\sim\rho^{\sg(\theta)}\) denotes an endpoint generated by the ODE of the current model \(v^\theta\), with the dependence of the sampling procedure on \(\theta\) stopped during differentiation.

From the fixed-point perspective of Section~\ref{subsec:25}, the approximate-regularization update \eqref{eq:143} induces the fixed-point condition
\[
v^\rho
=
v^\rho
+
\eta\widehat{\Gamma}^{\rho,\mathrm{appr}\text{-}\mathrm{reg}}.
\]
Accordingly, the covariance-form approximate-regularization loss \eqref{eq:144} admits the following stop-gradient form:
\begin{align}
    &\calL_{\rho,\eta,B}^{\mathrm{appr}\text{-}\mathrm{reg}\text{-}\mathrm{cov},\mathrm{sg}}(\theta)
    :=
    \E_{\substack{
    t\sim\U(0,1),\,(X_t,X_1)\sim \Xi_{t,1}^{\sg(\theta)}}}
    \bigg[
        \Big\|
        v_t^\theta(X_t)
        -
        \left(1-\frac{\eta}{\tau}\right)v_{t|1}(X_t|X_1)
        \notag\\
    &\quad
        -
        \frac{\eta}{\tau}v_t^\base(X_t)
        -
        \eta\left(r(X_1)-B_t(X_t)\right)
        \left(
            v_{t|1}(X_t|X_1)-\sg(v_t^\theta)(X_t)
        \right)
        +
        \frac{\eta}{\tau}\kappa_t\widehat K_t(X_t)
        \Big\|_2^2
    \bigg].
    \label{eq:276}
\end{align}
Here, $\Xi_{t,1}^{\sg(\theta)}$ denotes the joint distribution constructed from the current model $v^\theta$, with its dependence on $\theta$ placed under stop-gradient. Under the specialization
\begin{align*}
&\where{\calL_{\rho,\eta,B}^{\mathrm{appr}\text{-}\mathrm{reg}\text{-}\mathrm{cov},\mathrm{sg}}(\theta)}{\eta=\tau=1,\,\widehat{K}_t\equiv 0,\,B_t\equiv 0}\\
=&
    \E_{\substack{
    t\sim\U(0,1),\,(X_t,X_1)\sim \Xi_{t,1}^{\sg(\theta)}}}
    \bigg[
        \Big\|
        v_t^\theta(X_t)
        -v_t^\base(X_t)
        -
        r(X_1)
        \left(
            v_{t|1}(X_t|X_1)-\sg\left(v_t^\theta(X_t)\right)
        \right)\Big\|_2^2
    \bigg],
\end{align*}
the fixed-point condition reduces to
\[
v^\rho
=
\where{\hat v^{\mathrm{appr}\text{-}\mathrm{reg}}}{\eta=\tau=1,\,\widehat{K}_t\equiv 0}
=
v^\base+\Gamma^{\rho,r}.
\]

Comparing $\mathcal{L}^{\mathrm{RAM}}$ with the loss above yields the exact sample-wise correspondence
\[
\mathcal{L}^{\mathrm{RAM}}(\theta)=\where{\calL_{\rho,\eta,B}^{\mathrm{appr}\text{-}\mathrm{reg}\text{-}\mathrm{cov},\mathrm{sg}}(\theta)}{\eta=\tau=1,\,\widehat{K}_t\equiv 0,\,B_t\equiv 0}
\qquad\text{under the forward construction.}
\]
Therefore, RAM is the stop-gradient covariance-form realization of the fixed-point condition~\eqref{eq:166} associated with the full-step approximate-regularization update, using the forward construction and a zero baseline. Equivalently, RAM uses the stop-gradient trick to enforce the fixed-point condition
\[
v^\rho=v^\base+\Gamma^{\rho,r}.
\]
RAM is not critical-point consistent in general.

\subsection{Adjoint Matching}\label{subsec:61}

We first translate the SOC formulation used by Basic Adjoint Matching (BAM) and Lean Adjoint Matching (LAM) into our notation. In Adjoint Matching \citep{domingo2025adjoint}, the SOC problem for reward-based fine-tuning is
\begin{align*}
    \min_{u}\,\,&\E\left[\int_{0}^{1}\frac{1}{2}\norm[2]{u_t\left(Y_t\right)}^2\dd t-r\left(Y_1\right)\right],\\
    \mathrm{s.t.}\,\,&\dd Y_t=\left(b_t^\base\left(Y_t\right)+\sqrt{2\kappa_t}u_t\left(Y_t\right)\right)\dd t+\sqrt{2\kappa_t}\dd W_t,\\
    &Y_0\sim p_0=\Normal(0,I),
\end{align*}
where the inverse temperature is set to $\tau=1$. Here, the base drift is
\[
b_t^\base(x)=2v_t^\base(x)-\frac{\dot{\alpha}_t}{\alpha_t}x.
\]

Under a parameterized velocity model $v^\theta$, Adjoint Matching rewrites the control as
\begin{equation*}
u_t^\theta(x)=\sqrt{\frac{2}{\kappa_t}}\left(v_t^\theta(x)-v_t^\base(x)\right).
\end{equation*}
The SOC problem then becomes
\begin{subequations}
\begin{align}
    \min_{\theta}\,\,&\E\left[\int_{0}^{1}\frac{1}{\kappa_t}\norm[2]{v_t^\theta(Y_t)-v_t^\base(Y_t)}^2\dd t-r\left(Y_1\right)\right],\\
    \mathrm{s.t.}\,\,&\dd Y_t=\left(2v_t^\theta(Y_t)-\frac{\dot{\alpha}_t}{\alpha_t}Y_t\right)\dd t+\sqrt{2\kappa_t}\dd W_t,\qquad t\in(0,1),\label{eq:277}\\
    &Y_0\sim p_0=\Normal(0,I).
\end{align}
\end{subequations}

When $v^\theta$ is canonical, the SDE~\eqref{eq:277} under the ``memoryless noise schedule'' $\sigma_t=\sqrt{2\kappa_t}$ \citep{domingo2025adjoint} coincides exactly with the posterior-preserving SDE~\eqref{eq:200}. In the stop-gradient implementations of BAM and LAM, the current model $v^\theta$ is used as a self-anchor and treated as canonical, as discussed in Section~\ref{subsec:25}. The corresponding sampling scheme is
\begin{equation}\label{eq:278}
    X_0\sim p_0,\,\BY_{[0,1]}\sim\mathbb{P}^\theta_{[0,1]|0}(\cdot|X_0)\quad\Rightarrow \quad\BY_{[0,1]}\sim\mathbb{P}^\theta_{[0,1]}.
\end{equation}
For every $t\in[0,1)$, we denote $\mathbb{P}^\theta_{[t,1]}$ by the path distribution of $\BY_{[t,1]}$ under the current model $v^\theta$. Using this notation, the losses of both BAM and LAM take the form
\[
\mathcal{L}(\theta)=\frac{1}{2}\E_{\BY_{[0,1]}\sim\mathbb{P}^{\sg(\theta)}_{[0,1]}}\left[\int_{0}^{1}\norm[2]{\sqrt{\frac{2}{\kappa_t}}\left(v_t^\theta(Y_t)-v_t^\base(Y_t)\right)+\sqrt{2\kappa_t}\sg\left(\lambda_t^{\theta}(\BY_{[0,1]})\right)}^2\dd t\right],
\]
where the pathwise adjoint $\lambda_t^{\theta}(\BY_{[0,1]})$ satisfies the basic or lean adjoint ODE in BAM and LAM, respectively. The full-trajectory implementation corresponds to the multi-time supervision scheme for the gradient-form (under the reverse construction) introduced in Section~\ref{subsec:20}.
For ease of presentation, we rewrite the full-trajectory loss in the equivalent single-time supervision form:
\begin{align*}
    \mathcal{L}(\theta)
    =&\frac{1}{2}\E_{\BY_{[0,1]}\sim\mathbb{P}^{\sg(\theta)}_{[0,1]}}\left[\int_{0}^{1}\norm[2]{\sqrt{\frac{2}{\kappa_t}}\left(v_t^\theta(Y_t)-v_t^\base(Y_t)\right)+\sqrt{2\kappa_t}\sg\left(\lambda_t^{\theta}(\BY_{[0,1]})\right)}^2\dd t\right]\\
    =&\E_{\BY_{[0,1]}\sim\mathbb{P}^{\sg(\theta)}_{[0,1]}}\left[\int_{0}^{1}\frac{1}{\kappa_t}\norm[2]{\left(v_t^\theta(Y_t)-v_t^\base(Y_t)\right)+\kappa_t\sg\left(\lambda_t^{\theta}(\BY_{[t,1]})\right)}^2\dd t\right]\\
    =&\E_{t\sim\U(0,1),\,\BY_{[t,1]}\sim\mathbb{P}^{\sg(\theta)}_{[t,1]}}\left[\frac{1}{\kappa_t}\norm[2]{v_t^\theta(Y_t)-v_t^\base(Y_t)+\kappa_t\sg\left(\lambda_t^{\theta}(\BY_{[t,1]})\right)}^2\right].
\end{align*}
Here, \(\mathbb{P}^{\sg(\theta)}_{[t,1]}\) is a particular choice of \(\Xi_{[t,1]}^{\sg(\theta)}\) for which the \(Y_t\)-marginal is given by the marginal probability path of the current model \(v^\theta\). The notation \(\Xi_{[t,1]}^{\sg(\theta)}\) imposes the same posterior-compatibility requirement as
\(\Xi_{[t,1]}^\rho\) in~\eqref{eq:95}, with \(v^\theta\) treated as canonical in place of \(v^\rho\) and all resulting dependence on \(\theta\) held fixed during differentiation.
For simplicity, we use the following more general form of the loss throughout the remainder of this subsection:
\begin{align*}
    \mathcal{L}(\theta)
    =&\E_{t\sim\U(0,1),\,\BY_{[t,1]}\sim\Xi^{\sg(\theta)}_{[t,1]}}\left[\frac{1}{\kappa_t}\norm[2]{v_t^\theta(Y_t)-v_t^\base(Y_t)+\kappa_t\sg\left(\lambda_t^{\theta}(\BY_{[t,1]})\right)}^2\right].
\end{align*}

\subsubsection{Basic Adjoint Matching}

The BAM loss is
\[
\mathcal{L}^{\mathrm{BAM}}(\theta)=\E_{t\sim\U(0,1),\,\BY_{[t,1]}\sim\Xi^{\sg(\theta)}_{[t,1]}}\left[\frac{1}{\kappa_t}\norm[2]{v_t^\theta(Y_t)-v_t^\base(Y_t)+\kappa_t\sg\left(\lambda_t^{\mathrm{BAM},\theta}(\BY_{[t,1]})\right)}^2\right],
\]
where the pathwise basic adjoint $\lambda_s^{\mathrm{BAM},\theta}(\BY_{[t,1]})$ satisfies the ODE
\begin{align*}
    &\frac{\dd}{\dd s}\lambda_s^{\mathrm{BAM},\theta}(\BY_{[t,1]})=-\nabla_{Y_s}\left(2v_s^\theta(Y_s)-\frac{\dot{\alpha}_s}{\alpha_s}Y_s\right)^\top\lambda_s^{\mathrm{BAM},\theta}(\BY_{[t,1]})\\
    &\qquad\qquad\qquad\qquad\qquad\qquad
    -\nabla_{Y_s}\left(\frac12\norm[2]{\sqrt{\frac{2}{\kappa_s}}\left(v_s^\theta(Y_s)-v_s^\base(Y_s)\right)}^2\right),
    \qquad s\in(t,1),\\
    &\lambda_1^{\mathrm{BAM},\theta}(\BY_{[t,1]})=-\nabla r(Y_1).
\end{align*}
Since \citep{domingo2025adjoint} sets $\tau=1$, comparing this ODE with~\eqref{eq:160} gives
\[
\lambda_t^{\mathrm{BAM},\theta}(\BY_{[t,1]})=-\where{\lambda_t^{\mathrm{Bolza},\theta}(\BY_{[t,1]})}{\tau=1,\,\mu=\rho^\base}.
\]
Therefore, the BAM loss can be rewritten as
\[
\mathcal{L}^{\mathrm{BAM}}(\theta)=\E_{\substack{t\sim\U(0,1),\\\BY_{[t,1]}\sim\Xi^{\sg(\theta)}_{[t,1]}}}\left[\frac{1}{\kappa_t}\norm[2]{v_t^\theta(Y_t)-v_t^\base(Y_t)-\kappa_t\sg\left(\where{\lambda_t^{\mathrm{Bolza},\theta}(\BY_{[t,1]})}{\tau=1,\,\mu=\rho^\base}\right)}^2\right].
\]

To make the correspondence exact, we generalize the stop-gradient loss~\eqref{eq:159} of the Bolza realization by introducing a positive weight $\omega_t(x_t)>0$:
\begin{align}
    \calL_{\eta,\omega}^{\mathrm{grad,Bolza},\mathrm{sg}}(\theta)
    :=&
    \E_{\substack{
        t\sim\U(0,1),\,\BY_{[t,1]}\sim\Xi_{[t,1]}^{\sg(\theta)}}}
    \bigg[\omega_t(Y_t)
        \Big\|
            v_t^\theta(Y_t)
            -
            \left(1-\frac{\eta}{\tau}\right)v_{t|1}(Y_t|Y_1)\notag\\
            &\qquad\qquad\qquad\qquad
            -\frac{\eta}{\tau}v_t^{\mu}(Y_t)-\eta\kappa_t\sg\left(\lambda_t^{\mathrm{Bolza},\theta}(\BY_{[t,1]})\right)
        \Big\|_2^2
    \bigg].
    \label{eq:279}
\end{align}
Comparing the BAM loss $\mathcal{L}^{\mathrm{BAM}}$ with $\calL_{\eta,\omega}^{\mathrm{grad,Bolza},\mathrm{sg}}$ yields the exact sample-wise correspondence
\[
\mathcal{L}^{\mathrm{BAM}}(\theta)=\where{\calL_{\eta,\omega}^{\mathrm{grad,Bolza},\mathrm{sg}}(\theta)}{\tau=\eta=1,\,\omega_t\equiv\frac{1}{\kappa_t},\,\mu=\rho^\base}
\qquad\text{under the reverse construction}.
\]
Therefore, BAM is the stop-gradient Bolza realization of the fixed-point condition associated with the full-step exact tangential update, using the reverse construction and a specific weighting. According to Propositions~\ref{prop:7} and~\ref{prop:21}, the stop-gradient loss has $\canonicalmap(\pi_{\rho^\base,1,r})$ as its unique canonical population-stationary point.

\subsubsection{Lean Adjoint Matching}

The LAM loss is
\[
\mathcal{L}^{\mathrm{LAM}}(\theta)=\E_{t\sim\U(0,1),\,\BY_{[t,1]}\sim\Xi^{\sg(\theta)}_{[t,1]}}\left[\frac{1}{\kappa_t}\norm[2]{v_t^\theta(Y_t)-v_t^\base(Y_t)+\kappa_t\lambda_t^{\mathrm{LAM}}(\BY_{[t,1]})}^2\right],
\]
where the pathwise lean adjoint satisfies the ODE
\begin{align*}
    &\frac{\dd}{\dd s}\lambda_s^{\mathrm{LAM}}(\BY_{[t,1]})
    =-\nabla_{Y_s}\left(2v_s^\base(Y_s)-\frac{\dot{\alpha}_s}{\alpha_s}Y_s\right)^\top\lambda_s^{\mathrm{LAM}}(\BY_{[t,1]}),
    \qquad s\in(t,1),\\
    &\lambda_1^{\mathrm{LAM}}(\BY_{[t,1]})=-\nabla r(Y_1).
\end{align*}
Comparing this ODE with~\eqref{eq:134} yields the pathwise identity
\[
\lambda_t^{\mathrm{LAM}}(\BY_{[t,1]})=-\where{\lambda_t^{b^\mu,r,0}(\BY_{[t,1]})}{\mu=\rho^\base}.
\]
Therefore, the LAM loss can be rewritten as
\[
\mathcal{L}^{\mathrm{LAM}}(\theta)=\E_{t\sim\U(0,1),\,\BY_{[t,1]}\sim\Xi^{\sg(\theta)}_{[t,1]}}\left[\frac{1}{\kappa_t}\norm[2]{v_t^\theta(Y_t)-v_t^\base(Y_t)-\where{\kappa_t\lambda_t^{b^\mu,r,0}(\BY_{[t,1]})}{\mu=\rho^\base}}^2\right].
\]

To match the LAM weighting exactly, we generalize the stop-gradient loss~\eqref{eq:161} of the reference-adjoint approximation by introducing a positive weight $\omega_t(x_t)>0$:
\begin{align}
    \calL_{\eta,\omega}^{\mathrm{ref}\text{-}\mathrm{adj},\mathrm{sg}}(\theta)
    :=
    \E_{\substack{
    t\sim\U(0,1),\,
    \BY_{[t,1]}\sim\Xi_{[t,1]}^{\sg(\theta)}}}
    \bigg[\omega_t(Y_t)
        \Big\|&
        v_t^\theta(Y_t)
        -
        \left(1-\frac{\eta}{\tau}\right)v_{t|1}(Y_t|Y_1)
        \notag\\
        &
        -
        \frac{\eta}{\tau}v_t^\mu(Y_t)
        -
        \eta\kappa_t
        \lambda_t^{b^\mu,r,0}(\BY_{[t,1]})
        \Big\|_2^2
    \bigg].
    \label{eq:280}
\end{align}
Comparing the LAM loss $\mathcal{L}^{\mathrm{LAM}}$ with $\calL_{\eta,\omega}^{\mathrm{ref}\text{-}\mathrm{adj},\mathrm{sg}}$ yields the exact sample-wise correspondence
\[
\mathcal{L}^{\mathrm{LAM}}(\theta)=\where{\calL_{\eta,\omega}^{\mathrm{ref}\text{-}\mathrm{adj},\mathrm{sg}}(\theta)}{\tau=\eta=1,\,\omega_t\equiv\frac{1}{\kappa_t},\,\mu=\rho^\base}
\qquad\text{under the reverse construction}.
\]
Therefore, LAM is the stop-gradient realization of the full-step reference-adjoint approximation update, using the reverse construction and a specific weighting.
According to Propositions~\ref{prop:14} and~\ref{prop:21}, the stop-gradient loss has $\canonicalmap(\pi_{\rho^\base,1,r})$ as its unique canonical population-stationary point.

\subsection{Flow Sampling}\label{subsec:62}

Using our notation, the Flow Sampling (FS) loss can be written as
\[
    \mathcal{L}^{\mathrm{FS}}(\theta)
    :=\E_{\substack{t\sim\U(0,1),\,X_1\sim\rho^{\sg(\theta)},\,X_0\sim p_0,\,X_t=\alpha_tX_1+\beta_tX_0}}\left[
        \norm[2]{
    b_t^\theta(X_t)-v_{t|1}(X_t|X_1)-\frac{\kappa_t}{\alpha_t}\nabla r(X_1)
    }^2
    \right].
\]
Here, $X_1\sim\rho^{\sg(\theta)}$ denotes the endpoint $X_1:=Y_1$ generated by
\begin{equation}\label{eq:281}
\dd Y_t=b_t^\theta(Y_t)\,\dd t+\sigma_t\,\dd W_t,\qquad Y_0\sim p_0,
\end{equation}
with the dependence of the sampling construction on $\theta$ placed under stop-gradient.

Since
\[
b_{t|1}(x_t|x_1)=2v_{t|1}(x_t|x_1)-\frac{\dot\alpha_t}{\alpha_t}x_t,
\]
we have
\begin{align*}
    v_{t|1}(X_t|X_1)+\frac{\kappa_t}{\alpha_t}\nabla r(X_1)
    =&b_{t|1}(X_t|X_1)-v_{t|1}(X_t|X_1)+\frac{\dot\alpha_t}{\alpha_t}X_t+\frac{\kappa_t}{\alpha_t}\nabla r(X_1)\\
    =&b_{t|1}(X_t|X_1)+\frac{\kappa_t}{\alpha_t}\nabla r(X_1)+\frac{\dot\alpha_t}{\alpha_t}X_t-\left(\dot\alpha_tX_1+\frac{\dot\beta_t}{\beta_t}\left(X_t-\alpha_tX_1\right)\right)\\
    =&b_{t|1}(X_t|X_1)+\frac{\kappa_t}{\alpha_t}\left(\nabla r(X_1)+\frac{\alpha_t}{\beta_t^2}\left(X_t-\alpha_tX_1\right)\right).
\end{align*}
Therefore, the FS loss can be equivalently rewritten as
\begin{align*}
    \mathcal{L}^{\mathrm{FS}}(\theta)
    :=\E_{\substack{t\sim\U(0,1),\,X_1\sim\rho^{\sg(\theta)},\,X_0\sim p_0,\,X_t=\alpha_tX_1+\beta_tX_0}}&\Bigg[
        \bigg\|
    b_t^\theta(X_t)-b_{t|1}(X_t|X_1)\\
    &-\frac{\kappa_t}{\alpha_t}\left(\nabla r(X_1)+\frac{\alpha_t}{\beta_t^2}\left(X_t-\alpha_tX_1\right)\right)
    \bigg\|_2^2
    \Bigg].
\end{align*}

By the drift-coordinate extension in Appendix~\ref{sub2sec:38}, the Gaussian-kernel approximation~\eqref{eq:142} takes the form
\begin{align}
    &\calL_{\rho,\eta,\widehat{\Sigma}}^{\mathrm{Gau}\text{-}\mathrm{ker},\mathrm{drift}}(\theta)
    :=
    \E_{\substack{
    t\sim\U(0,1),\,(X_t,X_1)\sim\Xi_{t,1}^\rho}}
    \Bigg[
        \bigg\|
        b_t^\theta(X_t)
        -
        b_{t|1}(X_t|X_1)
        \notag\\
    &\qquad
        -
        2\eta\frac{\alpha_t\kappa_t}{\beta_t^2}
        \widehat\Sigma_t^\rho(X_t)
        \left(
            \nabla r(X_1)
            +
            \frac{1}{\tau}\nabla\log\mu(X_1)
            +
            \frac{\alpha_t}{\tau\beta_t^2}
            \left(
                X_t-\alpha_tX_1
            \right)
        \right)
        \bigg\|_2^2
    \Bigg].
    \label{eq:282}
\end{align}
At a canonical anchor, the associated fixed-point condition is
\[
b_t^\rho(x_t)
=
b_t^\rho(x_t)
+
2\eta\frac{\alpha_t\kappa_t}{\beta_t^2}
\widehat\Sigma_t^\rho(x_t)
\E_{X_1\sim p_{1|t}^\rho(\cdot|x_t)}
\left[
\nabla r(X_1)
+
\frac{1}{\tau}\nabla\log\mu(X_1)
+
\frac{\alpha_t}{\tau\beta_t^2}
\left(
x_t-\alpha_tX_1
\right)
\right].
\]
By the stop-gradient construction of Section~\ref{subsec:25}, this fixed-point condition admits the following stop-gradient realization:
\begin{align}
    &\calL_{\eta,\widehat{\Sigma}}^{\mathrm{Gau}\text{-}\mathrm{ker},\mathrm{drift},\mathrm{sg}}(\theta)
    :=
    \E_{\substack{
    t\sim\U(0,1),\,(X_t,X_1)\sim\Xi_{t,1}^{\sg(\theta)}}}
    \Bigg[
        \bigg\|
        b_t^\theta(X_t)
        -
        b_{t|1}(X_t|X_1)
        \notag\\
    &\qquad
        -
        2\eta\frac{\alpha_t\kappa_t}{\beta_t^2}
        \widehat\Sigma_t(X_t)
        \left(
            \nabla r(X_1)
            +
            \frac{1}{\tau}\nabla\log\mu(X_1)
            +
            \frac{\alpha_t}{\tau\beta_t^2}
            \left(
                X_t-\alpha_tX_1
            \right)
        \right)
        \bigg\|_2^2
    \Bigg].
    \label{eq:283}
\end{align}

With the isotropic choice $\widehat\Sigma_t(X_t)=\frac{\beta_t^2}{\alpha_t^2}I$, comparing the two losses yields the exact sample-wise correspondence
\begin{equation*}
\mathcal{L}^{\mathrm{FS}}(\theta)
=
\where{\calL_{\eta,\widehat{\Sigma}}^{\mathrm{Gau}\text{-}\mathrm{ker},\mathrm{drift},\mathrm{sg}}(\theta)}{\mu=1,\,\tau=1,\,\eta=\frac12,\,\widehat\Sigma_t\equiv \frac{\beta_t^2}{\alpha_t^2}I}
\qquad
\text{under the forward construction}.
\end{equation*}
Therefore, FS is the stop-gradient realization of the damped Gaussian-kernel approximation update with stepsize \(\eta=\frac12\), using the drift coordinate and the forward construction with SDE-based sampling. According to Propositions~\ref{prop:15} and~\ref{prop:21}, the stop-gradient loss has $\canonicalmap(\pi_{1,1,r})$ as its unique canonical population-stationary point.

\subsection{Adjoint Sampling}\label{subsec:63}

Using our notation, Adjoint Sampling (AS) considers the SDE
\begin{subequations}\label{eq:284}
\begin{align}
    &\dd Y_t=b_t^\theta(Y_t)\dd t+\sigma_t\dd W_t,\qquad t\in(0,1),\\
    &Y_0=0.
\end{align}
\end{subequations}
Let $\rho^\theta$ denote the terminal density induced by this SDE. AS then uses the following stop-gradient loss:
\begin{align*}
    \mathcal{L}^{\mathrm{AS}}(\theta)
    :=\E_{\substack{
        t\sim\U(0,1),\,X_1\sim\rho^{\sg(\theta)},\,\\
        X_t\sim\Normal\left(\frac{\widetilde\gamma_t}{\widetilde\gamma_1}X_1,\frac{\widetilde\gamma_t(\widetilde\gamma_1-\widetilde\gamma_t)}{\widetilde\gamma_1}I\right)}}
        \left[\frac{1}{2\sigma_t^2}\norm[2]{\frac{1}{\sigma_t}b_t^\theta(X_t)+\sigma_t\nabla\left(\log\Normal\left(X_1;0,\widetilde{\gamma}_1I\right)-\tau r(X_1)\right)}^2\right],
\end{align*}
where
\[
\widetilde\gamma_t:=\int_{0}^{t}\sigma_u^2\,\dd u.
\]
The reciprocal sampling kernel used by AS,
\[
X_t|X_1\sim\Normal\left(\frac{\widetilde\gamma_t}{\widetilde\gamma_1}X_1,\frac{\widetilde\gamma_t(\widetilde\gamma_1-\widetilde\gamma_t)}{\widetilde\gamma_1}I\right),
\]
can be recovered from the one-sided interpolant setting. Using the notation of Appendix~\ref{subsec:57}, we define the AS-specific schedule
\begin{equation}\label{eq:285}
\widetilde\alpha_t:=\frac{\widetilde\gamma_t}{\widetilde\gamma_1},\qquad
\widetilde\beta_t:=\sqrt{\frac{\widetilde\gamma_t(\widetilde\gamma_1-\widetilde\gamma_t)}{\widetilde\gamma_1}},\qquad
\widetilde{\kappa}_t:=\frac{\widetilde\beta_t}{\widetilde\alpha_t}\left(\dot{\widetilde\alpha}_t\widetilde\beta_t-\widetilde\alpha_t\dot{\widetilde\beta}_t\right).
\end{equation}
This schedule satisfies
\[
\sigma_t=\sqrt{2\widetilde{\kappa}_t}.
\]
The sampling rule used by AS is equivalent to the forward construction under the one-sided interpolant setting:
\[
X_1\sim\rho^{\sg(\theta)},\, X_t\sim\Normal\left(\widetilde\alpha_tX_1,\widetilde\beta_t^2I\right)
\quad\Leftrightarrow\quad
X_1\sim\rho^{\sg(\theta)},\,Z\sim\Normal(0,I),\,X_t=\widetilde\alpha_tX_1+\widetilde\beta_tZ.
\]
Accordingly, the AS loss can be rewritten as
\begin{align*}
    \mathcal{L}^{\mathrm{AS}}(\theta)
    :=\E_{\substack{
        t\sim\U(0,1),\,X_1\sim\rho^{\sg(\theta)},\,Z\sim\Normal(0,I),\\
        X_t=\widetilde\alpha_tX_1+\widetilde\beta_tZ}}
        \left[\frac{1}{2\sigma_t^2}\norm[2]{\frac{1}{\sigma_t}b_t^\theta(X_t)+\sigma_t\nabla\left(\log\Normal\left(X_1;0,\widetilde{\gamma}_1I\right)-\tau r(X_1)\right)}^2\right].
\end{align*}
We have
\begin{align*}
    &\frac{1}{2\sigma_t^2}\norm[2]{\frac{1}{\sigma_t}b_t^\theta(X_t)+\sigma_t\nabla\left(\log\Normal\left(X_1;0,\widetilde{\gamma}_1I\right)-\tau r(X_1)\right)}^2\\
    =&\frac{1}{2\sigma_t^4}\norm[2]{b_t^\theta(X_t)+\sigma_t^2\left(-\frac{1}{\widetilde{\gamma}_1}X_1-\tau\nabla r(X_1)\right)}^2\\
    =&\frac{1}{8\widetilde\kappa_t^2}\norm[2]{b_t^\theta(X_t)-2\widetilde\kappa_t\left(\frac{1}{\widetilde{\gamma}_1}X_1+\tau\nabla r(X_1)\right)}^2.
\end{align*}

Moreover, the schedule~\eqref{eq:285} satisfies
\[
\dot{\widetilde{\alpha}}_t=\frac{2\widetilde{\kappa}_t}{\widetilde{\gamma}_1},\qquad
\dot{\widetilde{\beta}}_t=\frac{\widetilde{\gamma}_1-2\widetilde{\gamma}_t}{2\widetilde{\beta}_t\widetilde{\gamma}_1}\cdot 2\widetilde{\kappa}_t
=\frac{\widetilde{\kappa}_t}{\widetilde{\beta}_t}\left(1-2\widetilde{\alpha}_t\right).
\]
It follows that
\begin{align*}
    \widetilde{b}_{t|1}(x_t|x_1)=\dot{\widetilde{\alpha}}_t x_1+\left(\dot{\widetilde{\beta}}_t-\frac{\widetilde{\kappa}_t}{\widetilde{\beta}_t}\right)\frac{x_t-\widetilde{\alpha}_t x_1}{\widetilde{\beta}_t}=\dot{\widetilde{\alpha}}_t x_1-\frac{2\widetilde{\alpha}_t\widetilde{\kappa}_t}{\widetilde{\beta}_t^2}\left(x_t-\widetilde{\alpha}_t x_1\right).
\end{align*}
Thus,
\begin{align*}
    &\frac{1}{8\widetilde\kappa_t^2}\norm[2]{b_t^\theta(X_t)-2\widetilde\kappa_t\left(\frac{1}{\widetilde{\gamma}_1}X_1+\tau\nabla r(X_1)\right)}^2\\
    =&\frac{1}{8\widetilde\kappa_t^2}\norm[2]{b_t^\theta(X_t)-\left(\dot{\widetilde{\alpha}}_tX_1+2\widetilde\kappa_t\tau\nabla r(X_1)\right)}^2\\
    =&\frac{1}{8\widetilde\kappa_t^2}\norm[2]{b_t^\theta(X_t)-\left(\widetilde{b}_{t|1}(X_t|X_1)+\frac{2\widetilde{\alpha}_t\widetilde{\kappa}_t}{\widetilde{\beta}_t^2}\left(X_t-\widetilde{\alpha}_t X_1\right)+2\widetilde\kappa_t\tau\nabla r(X_1)\right)}^2.
\end{align*}
Therefore, the AS loss can be rewritten as
\begin{align}
    \mathcal{L}^{\mathrm{AS}}(\theta)
    &=\E_{\substack{
        t\sim\U(0,1),\,X_1\sim\rho^{\sg(\theta)},\,Z\sim\Normal(0,I),\,
        X_t=\widetilde\alpha_tX_1+\widetilde\beta_tZ}}
        \Bigg[\frac{1}{8\widetilde\kappa_t^2}\bigg\|b_t^\theta(X_t)-\widetilde{b}_{t|1}(X_t|X_1)\notag\\
        &\qquad\qquad\qquad\qquad\qquad\qquad\qquad
        -2\widetilde\kappa_t\tau\bigg(\nabla r(X_1)+\frac{\widetilde{\alpha}_t}{\tau\widetilde{\beta}_t^2}\left(X_t-\widetilde{\alpha}_t X_1\right)\bigg)\bigg\|_2^2\Bigg].\label{eq:286}
\end{align}

Using the one-sided-interpolant construction in Appendix~\ref{subsec:57}, we extend the drift-coordinate Gaussian-kernel approximation of Newton Matching in~\eqref{eq:282} to the one-sided setting:
\begin{align}
    &\widetilde{\calL}_{\rho,\eta,\widehat{\Sigma},\omega}^{\mathrm{Gau}\text{-}\mathrm{ker},\mathrm{drift}}(\theta)
    :=
    \E_{\substack{
    t\sim\U(0,1),\,(X_t,X_1)\sim\widetilde{\Xi}_{t,1}^\rho}}
    \Bigg[
        \omega_t(X_t)\bigg\|
        b_t^\theta(X_t)
        -
        \widetilde b_{t|1}(X_t|X_1)
        \notag\\
    &\qquad
        -
        2\eta\frac{\widetilde\alpha_t\widetilde\kappa_t}{\widetilde\beta_t^2}
        \widehat\Sigma_t^\rho(X_t)
        \bigg(
            \nabla r(X_1)
            +
            \frac{1}{\tau}\nabla\log\mu(X_1)
            +
            \frac{\widetilde\alpha_t}{\tau\widetilde\beta_t^2}
            \left(
                X_t-\widetilde\alpha_tX_1
            \right)
        \bigg)
        \bigg\|_2^2
    \Bigg].
    \label{eq:287}
\end{align}
Here, the positive weight $\omega_t(x_t)>0$ does not affect the population-stationary point. At a canonical drift $\widetilde b^\rho$, the associated fixed-point condition is
\begin{align*}
    \widetilde b_t^\rho(x_t)
    =\widetilde b_t^\rho(x_t)+
        2\eta\frac{\widetilde\alpha_t\widetilde\kappa_t}{\widetilde\beta_t^2}
        \widehat\Sigma_t^\rho(x_t)
    \E_{X_1\sim \widetilde{p}_{1|t}^\rho(\cdot|x_t)}
    \left[
            \nabla r(X_1)
            +
            \frac{1}{\tau}\nabla\log\mu(X_1)
            +
            \frac{\widetilde\alpha_t}{\tau\widetilde\beta_t^2}
            \left(
                x_t-\widetilde\alpha_tX_1
            \right)
    \right].
\end{align*}
By the stop-gradient construction of Section~\ref{subsec:25}, this fixed-point condition admits the following stop-gradient realization:
\begin{align}
    &\widetilde{\calL}_{\rho,\eta,\widehat{\Sigma},\omega}^{\mathrm{Gau}\text{-}\mathrm{ker},\mathrm{drift},\mathrm{sg}}(\theta)
    :=
    \E_{\substack{
    t\sim\U(0,1),\,(X_t,X_1)\sim\widetilde{\Xi}_{t,1}^{\sg(\theta)}}}
    \Bigg[
        \omega_t(X_t)\bigg\|
        b_t^\theta(X_t)
        -
        \widetilde b_{t|1}(X_t|X_1)
        \notag\\
    &\qquad\qquad
        -
        2\eta\frac{\widetilde\alpha_t\widetilde\kappa_t}{\widetilde\beta_t^2}
        \widehat\Sigma_t(X_t)
        \bigg(
            \nabla r(X_1)
            +
            \frac{1}{\tau}\nabla\log\mu(X_1)
            +
            \frac{\widetilde\alpha_t}{\tau\widetilde\beta_t^2}
            \left(
                X_t-\widetilde\alpha_tX_1
            \right)
        \bigg)
        \bigg\|_2^2
    \Bigg].
    \label{eq:288}
\end{align}

Comparing this loss with~\eqref{eq:286} yields the exact sample-wise correspondence
\begin{equation*}
\mathcal{L}^{\mathrm{AS}}(\theta)
=
\where{\widetilde{\calL}_{\rho,\eta,\widehat{\Sigma},\omega}^{\mathrm{Gau}\text{-}\mathrm{ker},\mathrm{drift},\mathrm{sg}}(\theta)}{\mu=1,\,\widetilde\alpha_t=\frac{\widetilde\gamma_t}{\widetilde\gamma_1},\,
\widetilde\beta_t=\sqrt{\frac{\widetilde\gamma_t(\widetilde\gamma_1-\widetilde\gamma_t)}{\widetilde\gamma_1}},\,\eta=\tau,\,\widehat\Sigma_t\equiv \frac{\widetilde\beta_t^2}{\widetilde\alpha_t}I,\,\omega_t\equiv\frac{1}{8\widetilde\kappa_t^2}}
\end{equation*}
under the forward construction.

Therefore, AS is the stop-gradient realization of the full-step Gaussian-kernel approximation update under the forward construction with SDE-based sampling, using the drift coordinate and the one-sided interpolant. The stop-gradient loss has $\canonicalmap(\pi_{1,\tau,r})$ as its unique canonical population-stationary point.

\end{appendices}

\newpage

\phantomsection
\addcontentsline{toc}{section}{References}
\bibliography{ref}

\end{document}